\documentclass[acmtog]{acmart} 
\acmSubmissionID{279}

\usepackage{booktabs}
\usepackage{wrapfig}
\usepackage{xcolor}
\usepackage{lipsum}
\usepackage{floatflt}
\usepackage{algorithm}
\usepackage{algpseudocode}
\usepackage{enumitem}
\usepackage{amsmath}
\usepackage{nicefrac}
\usepackage{makecell}
\usepackage{multirow}
\usepackage{color}
\usepackage{bm}
\usepackage{microtype}
\usepackage{tabularx}
\usepackage{xspace}

\usepackage{graphicx}

\newif\ifshownotes
\shownotestrue 

\ifdefined\MakeWithNotes
\shownotestrue
\fi
\ifdefined\MakeWithoutNotes
\shownotesfalse
\fi

\ifshownotes
\newcommand{\colornote}[3]{{\color{#1}\bf{#2: #3}\normalfont}}
\newcommand{\colornoteTwo}[3]{{\color{#1}\bf{#3}\normalfont}}
\newcommand{\colornoteThree}[2]{{\color{#1}\bf{#2}\normalfont}}      
\else
\newcommand{\colornote}[3]{}
\newcommand{\colornoteTwo}[3]{}
\newcommand{\colornoteThree}[2]{}      
\fi

\definecolor{darkgreen}{rgb}{0.0,0.75,0.0}

\definecolor{darkblue}{rgb}{0.0,0.0,0.45}

\acmJournal{TOG}

\makeatletter
\DeclareRobustCommand\onedot{\futurelet\@let@token\@onedot}
\def\@onedot{\ifx\@let@token.\else.\null\fi\xspace}
\def\eg{\emph{e.g}\onedot}
\def\ie{\emph{i.e}\onedot}
\def\etal{\emph{et al}\onedot}
\def\etc{\emph{etc}\onedot}
\makeatother

\def\revision#1{#1}

\author{Lvmin Zhang}
\email{lvmin@stanford.edu}
\affiliation{\institution{Stanford University}\country{USA}}

\author{Maneesh Agrawala}
\email{maneesh@cs.stanford.edu}
\affiliation{\institution{Stanford University}\country{USA}}

\setcopyright{acmlicensed}
\acmJournal{TOG}
\acmYear{2024} \acmVolume{43} \acmNumber{4} \acmArticle{100} \acmMonth{7}\acmDOI{10.1145/3658150}

\begin{document}

\title{Transparent Image Layer Diffusion using Latent Transparency}

\begin{abstract}
We present an approach enabling large-scale pretrained latent diffusion models to generate transparent images. 
The method allows generation of single transparent images or of multiple transparent layers. 
The method learns a ``latent transparency'' that encodes alpha channel transparency into the latent manifold of a pretrained latent diffusion model. 
It preserves the production-ready quality of the large diffusion model by regulating the added transparency as a latent offset with minimal changes to the original latent distribution of the pretrained model.
In this way, any latent diffusion model can be converted into a transparent image generator by finetuning it with the adjusted latent space.
We train the model with 1M transparent image layer pairs collected using a human-in-the-loop collection scheme.
We show that latent transparency can be applied to different open source image generators, or be adapted to various conditional control systems to achieve applications like foreground/background-conditioned layer generation, joint layer generation, structural control of layer contents, \etc.
A user study finds that in most cases (97\%) users prefer our natively generated transparent content over previous ad-hoc solutions such as generating and then matting. Users also report the quality of our generated transparent images is comparable to real commercial transparent assets like Adobe Stock. 
\end{abstract}

\ccsdesc[500]{Applied computing~Fine arts}
\ccsdesc[500]{Applied computing~Media arts}

\keywords{Transparent images, image editing, image layer, text-to-image diffusion}

\begin{teaserfigure}
\begin{tabularx}{\linewidth}{*{2}{>{\centering\arraybackslash}X}}
\small "Woman with messy hair, in the bedroom" & 
\small "Burning firewood, on a table, in the countryside"
\end{tabularx}
\includegraphics[width=\textwidth]{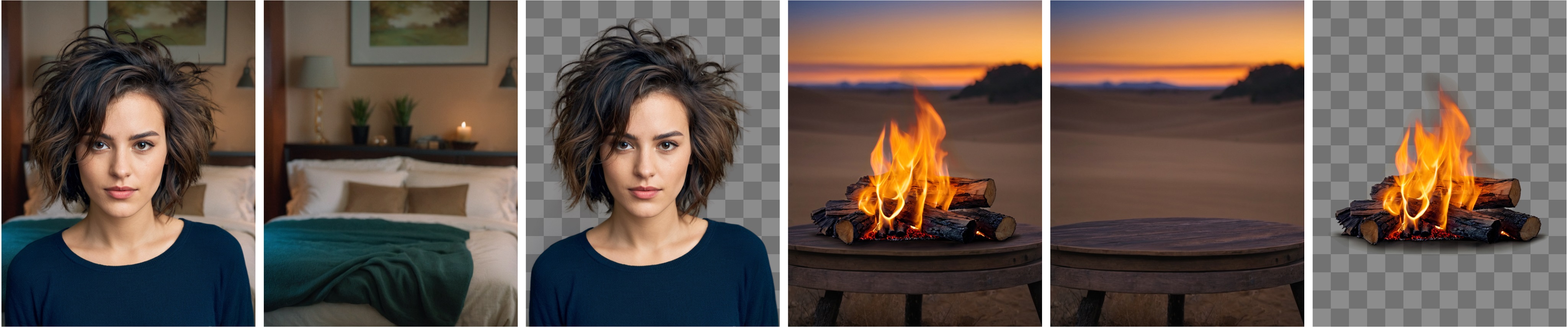}
\begin{tabularx}{\linewidth}{*{6}{>{\centering\arraybackslash}X}}
\small Blended output & 
\small Output layer 1 & 
\small Output layer 2 & 
\small Blended output & 
\small Output layer 1 & 
\small Output layer 2 
\end{tabularx}
\caption{Generating transparent images and layers. For the given text prompts (top), our framework is capable of generating multiple layers with transparency. These layers can be blended to produce images corresponding to the prompts. Zoom in to see details including messy hair and semi-transparent fire.}
\label{fig:teaser}
\end{teaserfigure}

\maketitle

\section{Introduction}

While large-scale models for generating images have become foundational in computer vision and graphics, surprisingly little research attention has been given to layered content generation or transparent image generation. This situation is in stark contrast to  substantial market demand. The vast majority of visual content editing software and workflows are layer-based, relying heavily on transparent or layered elements to compose and create content.

The primary factors contributing to this research gap are the lack of training data and the difficulty in manipulating the data representation of existing large-scale image generators. High-quality transparent image elements on the Internet are typically hosted by commercial image stocks with limited (and costly) access, in contrast to text-image datasets that already include billions of images (\eg, LAION\,\cite{Schuhmann2022}). The largest open-source transparent image datasets are often less than 50K in size (\eg, DIM\,\cite{Xu2017} includes 45,500 transparent images). Meanwhile, most open-source image generation models, \eg, Stable Diffusion, are latent diffusion models that are sensitive to their latent space  data representations. Even minor changes to the latent distribution could severely degrade inference or finetuning. For instance, Stable Diffusion 1.5 and XL use different latent spaces, and finetuning with mismatched latents can cause significant degradation in output image quality\,\cite{sdd}. This adds to the challenge of manipulating the data representation of existing models to support additional formats like transparent images.

We present a "latent transparency" approach that enables large-scale pretrained latent diffusion models to generate transparent images as well as multiple transparent layers. This method encodes image transparency into a latent offset that is explicitly regulated to avoid disrupting the latent distribution. The latent transparency is encoded and decoded by external independent models, ensuring that the original pretrained latent encoder/decoder is preserved, so as to maintain high-quality results of state-of-the-art diffusion models. To generate multiple layers together, we use a shared attention mechanism that ensures consistency and harmonious blending between image layers, and we train LoRAs to adapt the models to different layer conditions.

We employ a human-in-the-loop scheme to train our framework and collect data simultaneously. We finalize the scale of our dataset at 1M transparent images, covering a diversity of content topics and styles. We then use state-of-the-art methods to extend the dataset to multi-layer samples. This dataset not only enables the training of transparent image generators but can also be used in different applications like background/foreground-conditioned generation, structure-guided generation, style transfer, \etc.

Experiments show that in a majority of cases (97\%), users prefer the transparent content generated natively by our method over previous ad-hoc solutions like generating-then-matting. When we compare the quality of our generated results with the search results from commercial transparent assets sites like Adobe Stock, user preference rates suggest that quality is comparable.

\revision{In summary, we
(1) propose ``latent transparency'', an approach to enable large-scale pretrained latent diffusion models to generate single transparent images or multiple transparent layers,
(2) we present a shared attention mechanism to generate layers with consistent and harmonious blending, and
(3) we present a pretrained model for transparent image generation, two pretrained LoRAs for multiple layer generation, as well as several additional ablative architectures for multi-layer generation.}

\section{Related Work}

\subsection{Hiding Images inside Perturbations}

Research in multiple fields point out a phenomenon: neural networks have the ability to ``hide'' features in perturbations inside existing features without changing the overall feature distributions, \eg, hiding an image inside another image through small, invisible pixel perturbations. 
A typical CycleGAN\,\cite{CycleGAN2017} experiment showcases \emph{face-to-ramen}, where the human face identity could be hidden in a picture of ramen. 
Similarly, invertible downscaling\,\cite{Xiao2020} and invertible grayscale\,\cite{xia-2018-invertible} indicate that neural networks can hide a large image inside a smaller one, or hide a colorful image inside a grayscale one, and then reconstruct the original image. 
In another widely verified experiment \citet{Goodfellow2015} show that adversarial example signals can be hidden inside feature perturbations to influence the behaviors of other neural networks. 
In this paper, our proposed ``latent transparency'' utilizes similar principles: hiding image transparency features inside a small perturbation added to the latent space of Stable Diffusion\,\cite{sd15}, while at the same time avoiding changes to the overall distribution of the latent space. 

\subsection{Diffusion Probabilistic Models and Latent Diffusion}
Diffusion Probabilistic Model\,\cite{DBLP:journals/corr/Sohl-DicksteinW15} and related training and sampling methods like Denoising Diffusion Probabilistic Model (DDPM)\,\cite{DBLP:conf/nips/HoJA20}, Denoising Diffusion Implicit Model (DDIM)\, \cite{DBLP:conf/iclr/SongME21}, and score-based diffusion\, \cite{DBLP:journals/corr/abs-2011-13456} contribute to the foundations of recent large-scale image generators. Early image diffusion methods usually directly use pixel colors as training data\,\cite{DBLP:conf/iclr/SongME21, DBLP:journals/corr/abs-2104-02600, DBLP:journals/corr/abs-2106-00132}. In contrast, the Latent Diffusion Model (LDM)\,\cite{rombach2021highresolution} operates in latent space and has been shown to enable easier training while lowering computation requirements. This method has been further extended to create Stable Diffusion\,\cite{sd15}. Recently, eDiff-I\,\cite{balaji2022ediffi} has used an ensemble of multiple conditions including a T5 text encoder\,\cite{Raffel2019}, a CLIP text and image embedding encoder\,\cite{ilharco2021openclip}. Versatile Diffusion\,\cite{xu2022versatile} adopts a multi-purpose diffusion framework to process text, an image, and variations within a single model. 

\subsection{Customized Diffusion Models and Image Editing}

Early methods to customize diffusion models have focused on text-guidance\,\cite{nichol2021glide,kim2022diffusionclip,avrahami2022blended}. 
Image diffusion algorithms also naturally support inpainting\,\cite{ramesh2022hierarchical,avrahami2022blended}. Textual Inversion\,\cite{gal2022image} and DreamBooth\,\cite{ruiz2022dreambooth} can personalize the contents of generated results based on a small set of examplar images  of the same topic or object.
Recently, control models have also been used to add additional conditions for the generation of text-to-image models, \eg, ControlNet~\cite{zhang2023adding}, lightweight T2I-adapter~\cite{mou2023t2i}, etc. IP-Adapter\cite{ye2023ip-adapter} uses a cross-attention mechanism to separate text and image features, allowing for the control signals of the reference image as a visual prompt.
\cite{li2023layerdiffusion} uses masks in neural network features to achieve semantic region control.
\revision{Inversion-based methods are also popular in editing images. 
The DDPM~\cite{DBLP:conf/nips/HoJA20} theory indicates that a diffusion algorithm constructs data with accumulated small variations and those variations, conditioned on noise, can be manipulated with inverted optimization. 
Mokady \etal~\shortcite{Mokady22} shows that DDIM inversion can optimize images without requiring inputs to be generated by a previously known diffusion process (null-text embedding).
Cao \etal~\shortcite{cao23} and Narek \etal~\shortcite{Narek22} manipulate spatial cross-attention features of Stable Diffusion layers together with DDPM inversion.
Hertz \etal~\shortcite{Hertz22} edit attention activations of the input images with user-given text prompts and feed them back to the diffusion models. 
DiffEdit~\shortcite{Guillaume22} generates region masks for image editing, given input images and user prompts.
DiffusionCLIP \cite{Kim22} finetunes diffusion models with CLIP loss against prompts. 
Imagic \cite{Bahjat22} jointly optimizes text embedding of user prompts and the model gradients to reconstruct the image for image editing applications.
}

\begin{figure*}
\includegraphics[width=\linewidth]{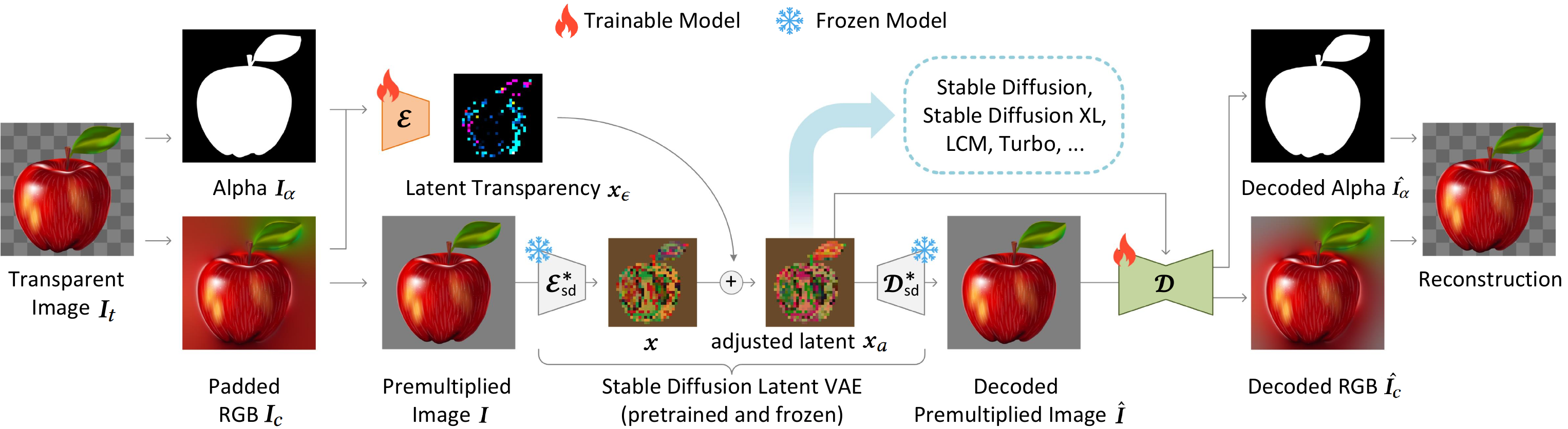}
\caption{\textbf{Latent Transparency.} Given an input transparent image, our framework encode a ``latent transparency'' to adjust the latent space of Stable Diffusion. The adjusted latent images can be decoded to reconstruct the color and alpha. This latent space with transparency can be further used in training or fine-tuning pretrained image diffusion models.}
\label{fig:vae}
\end{figure*}

\subsection{Transparent Image Layer Processing}

Transparent image processing is closely related to image decomposition, layer extraction, color palette processing, as well as image matting\,\cite{samplenet,ifm,keying}.
Typical color-based decomposition can be viewed as a RGB color space geometry problem \cite{Tan:2015:DTL,Tan:2016:DIL,Tan:2018:EPD,Tan:2018:PPB,Du:2023:IVE}.
These ideas have also been extended to more advanced blending of image layers \cite{Koyama2018}.
Unmixing-based color separation also contributes to image decomposition \cite{scs}, and semantic features can be used in image soft segmentation \cite{sss}.
We compare our approach to several state-of-the-art deep-learning based matting methods in our experiments and discussion.
\emph{PPMatting}~\cite{Chen2022} is a neural network image matting model trained from scratch using standard matting datasets. 
\emph{Matting Anything}~\cite{li2023matting} is a image matting model using the Segment Anything Model (SAM)~\cite{kirillov2023segany} as a backbone.
\emph{VitMatte}~\cite{yao2024vitmatte} is a tri-map-based matting method using a Vision Transformer (ViT).
\revision{Text2Layer~\cite{zhang2023text2layer} attempts to use foreground segmentation guidance to achieved layered effects in diffusion models, and indicates that its main bottleneck is the quality of foreground matting method since its learning objective is constructed from the image segmentation of matting models. Our approach starts from native generation of transparent images rather than post-processing of image matting, and is fundamentally different from previous approaches that use matting as post-processing of model outputs or use matting for dataset synthesizing.}

\subsection{Image Harmonization}

\revision{
Harmonious blending of transparent image layers is closely related to image harmonization research.
Achieving ``harmony'' is usually seen as a problem of correlating color, contrast, and style constituents between foreground and background to ensure natural appearance and consistent composition. 
Deep learning approaches~\cite{zhu2015learning,tsai2017deep,guo2021intrinsic, INR,PCT_Guerreiro_2023_CVPR,tan2023deep} have been proposed to harmonize images, using annotated datasets~\cite{cong2020dovenet, SycoNet2023}.
These works utilize the learning capabilities of neural networks to acquire the prior knowledge of harmonization.
}

\section{Method}

Our approach enables a Latent Diffusion Model (LDM), like Stable Diffusion, to generate transparent images, and then extends the model to jointly generate multiple transparent layers together. In section \ref{sec:31}, we introduce the method to adjust the LDM latent space to support transparent image encoding/decoding. In section \ref{sec:32}, we adapt pretrained latent diffusion models with the adjusted latent space to generate transparent images. In section \ref{sec:33}, we describe the method for joint or conditional layer generating. Finally, we detail the dataset preparation and implementation details for neural network training in section \ref{sec:34}. 

\paragraph{Definitions} To clarify the presentation we first define some terms.
For any transparent image $\bm{I}_t\in\mathbb{R}^{h\times w\times 4}$ with RGBA channels, we denote the first 3 RGB color channels as $\bm{I}_c\in\mathbb{R}^{h\times w\times 3}$ and the alpha channel as $\bm{I}_\alpha\in\mathbb{R}^{h\times w\times 1}$. Since the colors are physically undefined at pixels where the alpha value is strictly zero, in this paper, all undefined areas in $\bm{I}_c$ are always padded by an iterative Gaussian filter (see also supplementary material) to avoid aliasing and unnecessary edge patterns. We call $\bm{I}_c$ the ``padded RGB image'' (Fig.~\ref{fig:vae}). The $\bm{I}_t$ can be converted to a ``premultiplied image'' as $\bm{I} = \bm{I}_c * \bm{I}_a$ where $*$ denotes pixelwise multiplication. In this paper, all RGB values are in range $[-1, 1]$ (consistent with Stable Diffusion) while all alpha values are in range $[0, 1]$. The premultiplied image $\bm{I}$ can be seen as a common non-transparent RGB image that can be processed by any RGB-formatted neural networks. Visualizations of these images are shown in Fig.~\ref{fig:vae}.

\subsection{Latent Transparency}
\label{sec:31}

Our goal is to add transparency support to large-scale latent diffusion models, like Stable Diffusion (SD), that typically uses a latent encoder (VAE) to convert RGB images to latent images before feeding it to a diffusion model.
Herein, the VAE and the diffusion model should share the same latent distribution, as any major mismatch can significantly degrade the inference/training/fine-tuning of the latent diffusion framework.
When we adjust the latent space to support transparency, the original latent distribution must be preserved as much as possible.
These seemingly conflicting goals (adding transparency support while preserving the original latent distribution) can be handled with a straight-forward measurement: we can check how well the modified latent distribution can be decoded by the original pretrained frozen latent decoder --- if decoding a modified latent image creates severe artifacts, the latent distribution is misaligned or broken.

We can write this "harmfulness" measurement mathematically as follows.
Given an RGB image $\bm{I}$, the pretrained and frozen Stable Diffusion latent encoder $\mathcal{E}^*_{sd}(\cdot)$ and decoder $\mathcal{D}^*_{sd}(\cdot)$, where the $*$ indicates frozen models, we denote the latent image as $\bm{x}=\mathcal{E}^*_{sd}(\bm{I})$. Assuming this latent image $\bm{x}$ is modified by any offset $\bm{x}_\epsilon$, produces an adjusted latent $\bm{x}_a = \bm{x} + \bm{x}_\epsilon$. The decoded RGB reconstruction can then be written as $\hat{\bm{I}} = \mathcal{D}^*_{sd}(\bm{x}_a)$ and 
we can evaluate how ``harmful'' the offset $\bm{x}_\epsilon$ is as
\begin{equation}
\mathcal{L}_{\text{identity}} = ||\bm{I} - \hat{\bm{I}}||_2  = ||\bm{I} - \mathcal{D}^*_{sd}(\mathcal{E}^*_{sd}(\bm{I}) + \bm{x}_\epsilon)||_2 \,,
\label{eq1}
\end{equation}
where $||\cdot||_2$ is the L2 norm distance (mean squared error). Intuitively, if $\mathcal{L}_{\text{identity}}$ is relatively high, the $\bm{x}_\epsilon$ could be harmful and may have destroyed the reconstruction functionality of SD encoder-decoder, otherwise if $\mathcal{L}_{\text{identity}}$ is relatively low, the offset $\bm{x}_\epsilon$ does not break the latent reconstruction and the modified latent can still be handled by the pretrained Stable Diffusion.

Besides, since most mainstream VAE for diffusion models are KL-Divergence or Diagonal Gaussian Distribution models, these models often has a naively trained parameter for the standard deviation as an offset in the latent space. Considering such deviation denoted as $\bm{x}_{\text{std}}$, we can make use of this pretrained parameter to construct $\bm{x}_\epsilon=\lambda_{\text{offset}}\bm{x}_{\text{std}}\bm{x}_{\text{offset}}$ where $\bm{x}_{\text{offset}}$ is the raw output from newly added encoder, $\bm{x}_{\text{std}}$ is the deviation output of pretrained VAE, and $\lambda_{\text{offset}}$ is a weighting parameter with default $\lambda_{\text{offset}}=1e2$.

We make use of the latent offset $\bm{x}_\epsilon$ to establish ``latent transparency'' for encoding/decoding transparent images. 
More specifically, we train from scratch a latent transparency encoder $\mathcal{E}(\cdot, \cdot)$ that takes the RGB channels $\bm{I}_c$ and alpha channel $\bm{I}_\alpha$ as input to convert pixel-space transparency into a latent offset
\begin{equation}
\bm{x}_\epsilon = \mathcal{E}(\bm{I}_c, \bm{I}_\alpha) \,.
\end{equation}
We then train from scratch another latent transparency decoder $\mathcal{D}(\cdot, \cdot)$ that takes the adjusted latent $\bm{x}_a = \bm{x} + \bm{x}_\epsilon$ and the aforementioned RGB reconstruction $\hat{\bm{I}} = \mathcal{D}^*_{sd}(\bm{x}_a)$ to extract the transparent image from the adjusted latent space
\begin{equation}
[\hat{\bm{I}_c} \,\,\, \hat{\bm{I}_\alpha}] = \mathcal{D}(\hat{\bm{I}}, \bm{x}_a) \,,
\end{equation}
where $\hat{\bm{I}_c}, \hat{\bm{I}_\alpha}$ are the reconstructed color and alpha channels. The neural network layer architecture of $\mathcal{E}(\cdot, \cdot)$ and $\mathcal{D}(\cdot, \cdot)$ is in the supplementary material. We evaluate the reconstruction with
\begin{equation}
\mathcal{L}_{\text{recon}} = ||\bm{I_c} - \hat{\bm{I}_c}||_2 + ||\bm{I_a} - \hat{\bm{I}_a}||_2 \,,
\end{equation}
and we experimentally find that the result quality can be further improved by introducing a PatchGAN discriminator loss
\begin{equation}
\mathcal{L}_{\text{disc}} = \mathbb{L}_{\text{disc}}([\hat{\bm{I}_c}, \hat{\bm{I}_a}]) \,,
\end{equation}
where $\mathbb{L}_{\text{disc}}(\cdot, \cdot)$ is a GAN objective from a 5-layer patch discriminator (details in supplementary material). The final objective can be jointly written as 
\begin{equation}
\mathcal{L}_\text{vae} = \lambda_{\text{recon}}\mathcal{L}_{\text{recon}} + \lambda_{\text{identity}}\mathcal{L}_{\text{identity}} + \lambda_{\text{disc}}\mathcal{L}_{\text{disc}} \,,
\end{equation}
where $\lambda_{...}$ are weighting parameters: by default we use $\lambda_{\text{recon}} = 1, \lambda_{\text{identity}} = 1, \lambda_{\text{disc}} = 0.01$. By training this framework with $\mathcal{L}_\text{vae}$, the adjusted latent $\bm{x}_a$ can be encoded from transparent images or vise versa, and those latent images can be used in fine-tuning Stable Diffusion. We visualize the pipeline in Fig.~\ref{fig:vae}.

\subsection{Diffusion Model with Latent Transparency}
\label{sec:32}

\begin{figure}
\includegraphics[width=\linewidth]{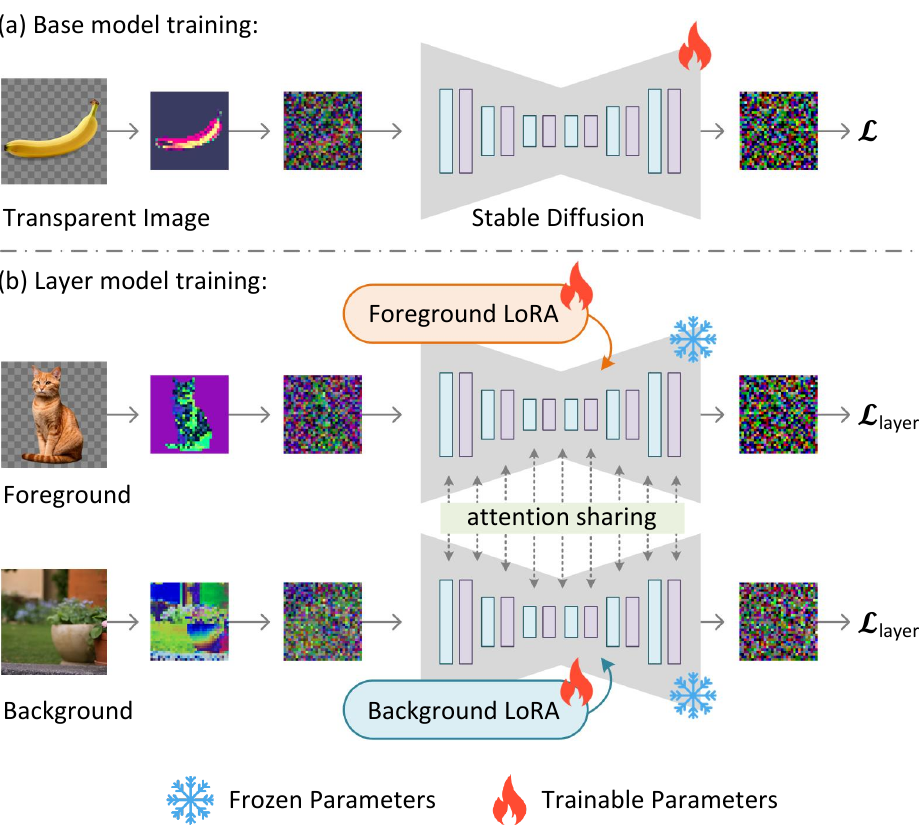}
\caption{\textbf{Model Training.} We visualize the training of the base model to generate transparent images, and the training of the multi-layer model to generate multiple layers together. When training the base diffusion model (a), all model weights are trainable, whereas for training the multi-layer model (b), only two LoRAs are trainable (the foreground LoRA and background LoRA).}
\label{fig:train}
\end{figure}

Since the altered latent space with latent transparency is explicitly regulated to align with the original pretrained latent distribution (Eq.~\ref{eq1}), Stable Diffusion can be directly fine-tuned on the altered latent space. Given the adjusted latent $\bm{x}_a$, diffusion algorithms progressively add noise to the image and produce a noisy image $\bm{x}_t$, with $t$ denoting how many times noise is added. When $t$ is large enough, the latent image approximates pure noise. Given a set of conditions including the time step $t$ and text prompt $\bm{c}_t$, image diffusion algorithms learn a network $\epsilon_\theta$ that predicts the noise added to the noisy latent image $\bm{x}_t$ with
\begin{equation}
	\mathcal{L} = \mathbb{E}_{\bm{x}_t, t, \bm{c}_t, \epsilon \sim \mathcal{N}(0, 1) }\Big[ \Vert \epsilon - \epsilon_\theta(\bm{x}_{t}, t, \bm{c}_t)) \Vert_{2}^{2}\Big]
\end{equation}
where $\mathcal{L}$ is the overall learning objective of the entire diffusion model. This training is visualized in Fig.~\ref{fig:train}-(a).

\subsection{Generating Multiple Layers}
\label{sec:33}

\begin{figure}
\includegraphics[width=\linewidth]{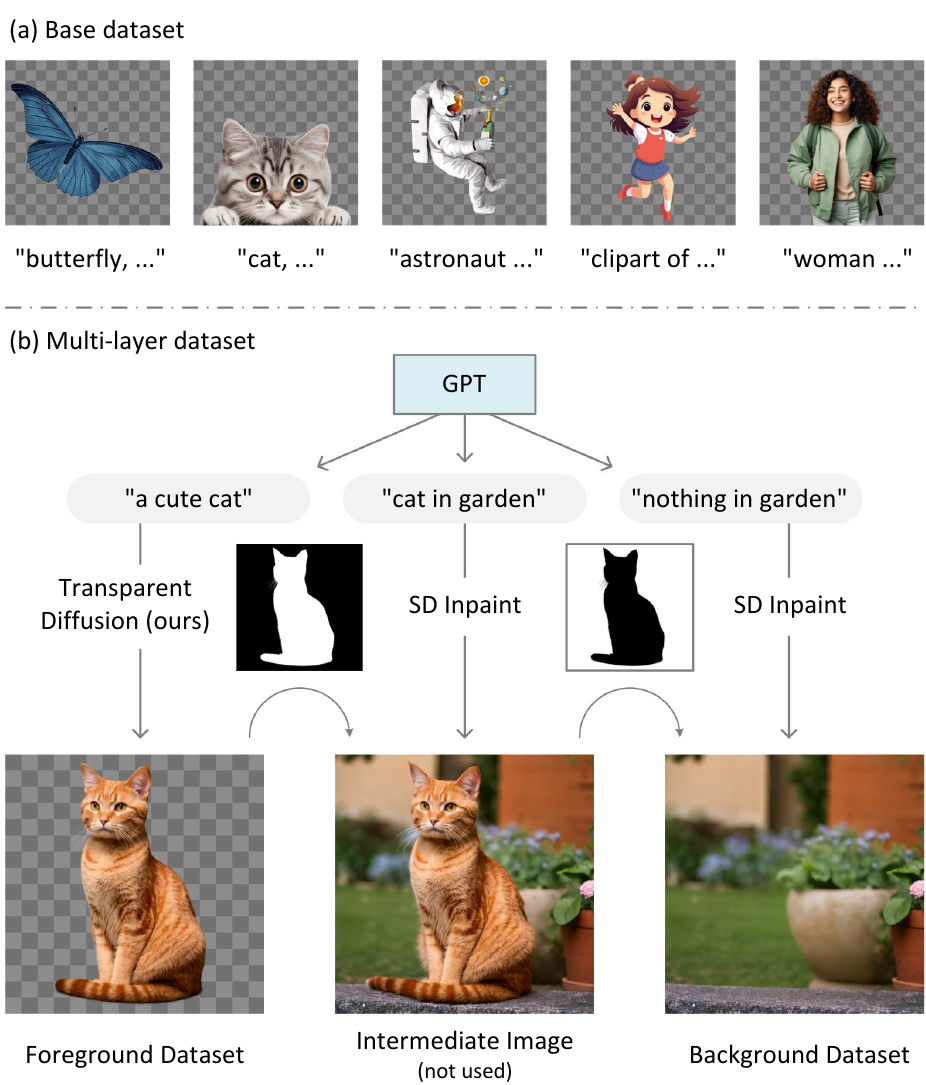}
\caption{\textbf{Dataset Preparation.} \revision{We demonstrate the preparation of the two datasets: the transparent image dataset (base dataset) and multi-layer dataset. The base dataset is collected by downloading online transparent images and a human-in-the-loop training method. The multi-layer dataset is synthesized with our transparent diffusion model and several state-of-the-art models including ChatGPT, SDXL inpaint model, \etc. The final scale of each dataset is around 1M.}}
\label{fig:data}
\end{figure}

\begin{figure}
\includegraphics[width=\linewidth]{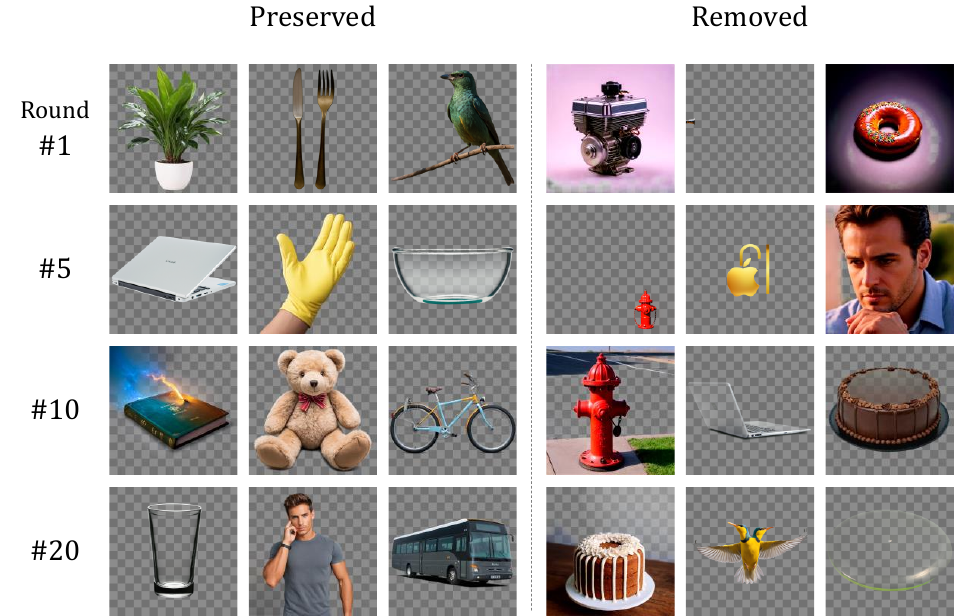}
\caption{\textbf{Human-in-the-loop data screening.} \revision{We visualize sample examples that are preserved versus removed in each round during the dataset collection process. We show examples from the round 1, 5, 10, and 20. The prompts are randomly sampled during the collecting process.}}
\label{fig:humanloop}
\end{figure}

\begin{figure*}
\begin{tabularx}{\linewidth}{*{4}{>{\centering\arraybackslash}X}}
\small "glass cup" & 
\small "man, 4k, best-quality" & 
\small "animal" & 
\small "game assets with magic effects"
\end{tabularx}
\includegraphics[width=\linewidth]{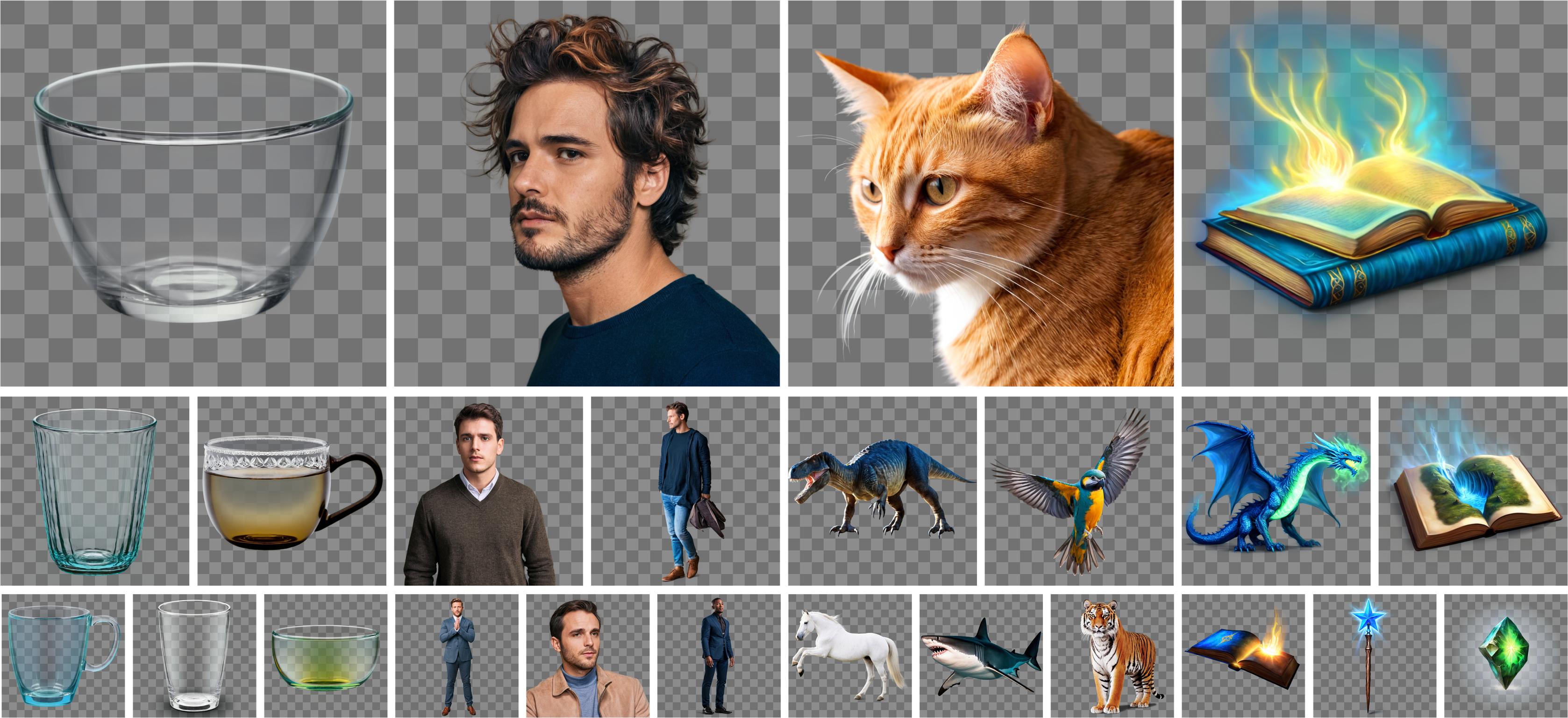}
\caption{\textbf{Qualitative Results.} We showcase various examples of transparent images generated by our model. The prompts for each group is given at the top of the examples. These examples only use our base single-layer model.}
\label{fig:qualitative}
\end{figure*}

\begin{figure*}
\includegraphics[width=\linewidth]{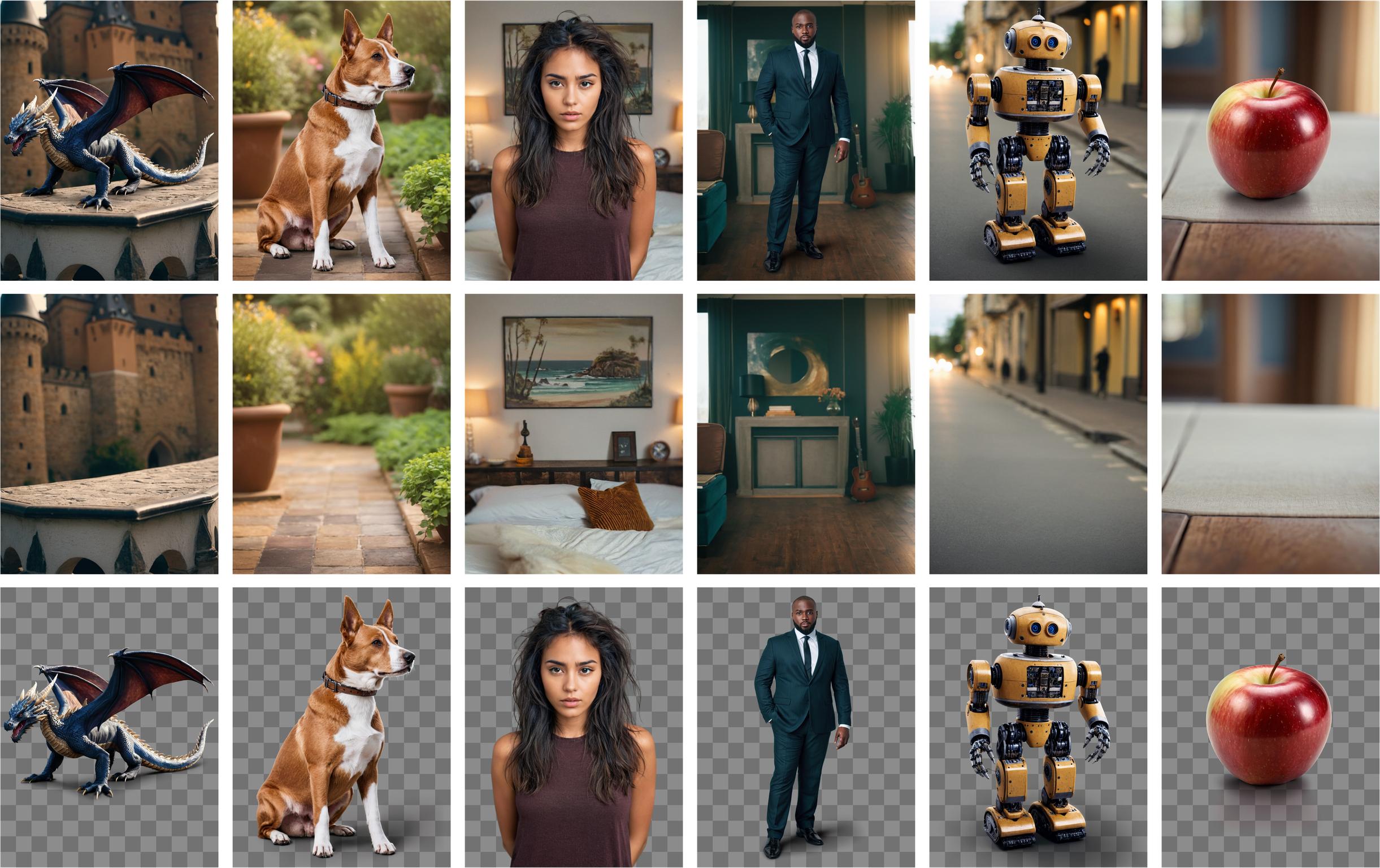}
\begin{tabularx}{\linewidth}{*{6}{>{\centering\arraybackslash}X}}
\small "dragon over castle" & 
\small "dog in garden" & 
\small "woman, messy hair" & 
\small "man in room" & 
\small "robot in street" & 
\small "apple on table" 
\end{tabularx}
\caption{\textbf{Multi-Layer Qualitative Results.} We presents qualitative results generated by our model using prompts with diverse topics. For each example, we show the blended image, and two output layers. More results are available in supplementary materials.}
\label{fig:multilayer}
\end{figure*}

We further extend the base model to a multi-layer model using attention sharing and LoRAs~\cite{hu2021lora}, as shown in Fig.~\ref{fig:train}-(b). We denote the foreground noisy latent as $\bm{x}_f$ and background as $\bm{x}_b$, and train two LoRAs, a foreground LoRA parameterized by $\theta_\text{f}$ and a background LoRA by $\theta_\text{b}$, to denoise the latent images. If the two models independently denoise the two images, we have the two objectives with
\begin{equation}
\left\{
\begin{aligned}
\mathbb{E}_{\bm{x}_f, t, \bm{c}_t, \epsilon_f \sim \mathcal{N}(0, 1) }&\Big[ \Vert \epsilon_f - \epsilon_{\theta, \theta_\text{f}}(\bm{x}_{f}, t, \bm{c}_t)) \Vert_{2}^{2}\Big]\\
\mathbb{E}_{\bm{x}_b, t, \bm{c}_t, \epsilon_b \sim \mathcal{N}(0, 1) }&\Big[ \Vert \epsilon_b - \epsilon_{\theta, \theta_\text{b}}(\bm{x}_{b}, t, \bm{c}_t)) \Vert_{2}^{2}\Big]
\end{aligned}
\right.
\end{equation}
where $\epsilon_f$, $\epsilon_b$ are latent noise for the foreground and background. We then merge the two independent diffusion process to achieve coherent generation. For each attention layer in the diffusion model, we concatenate all \{key, query, value\} vectors activated by the two images, so that the two passes can be merged into a jointly optimized big model $\epsilon_{\theta, \theta_\text{f}, \theta_\text{g}}(\cdot)$. We denote the merged noise as concatenated $\epsilon_m= [\epsilon_f,\epsilon_b]$, and we have the final objective
\begin{equation}
	\mathcal{L}_\text{layer} = \mathbb{E}_{\bm{x}_f,\bm{x}_b, t, \bm{c}_t, \epsilon_m \sim \mathcal{N}(0, 1) }\Big[ \Vert \epsilon_m - \epsilon_{\theta, \theta_\text{f}, \theta_\text{g}}(\bm{x}_{f},\bm{x}_{b}, t, \bm{c}_t)) \Vert_{2}^{2}\Big]
\end{equation}
to coherently generate multiple layers together. We can also make simple modifications to this objective to support conditional layer generation (\eg, foreground-conditioned background generation or background-conditioned foreground generation). More specifically, by using a clean latent for the foreground instead of noisy latent (\ie, by always setting $\epsilon_f=\bm{0}$), the model will not denoise foreground, and the framework becomes a foreground-conditioned generator. Similarly, by setting $\epsilon_b=\bm{0}$, the framework becomes a background-conditioned generator. We implement all these conditional variations in experiments.

\subsection{Dataset Preparation and Training Details}
\label{sec:34}

\paragraph{Base Dataset} We use a human-in-the-loop method to collect a dataset of transparent images and train our models. 
The dataset initially contains 20k high-quality transparent PNG images purchased or downloaded free from 5 online image stocks (all images include commercial use permission (examples in Fig.~\ref{fig:data}-(a)).
We then train the SDXL VAE with latent transparency using randomly sampled images with equal probability (at batch size 8), and then train the SDXL diffusion model using the same data with adjusted latents. 
Next we repeat the following steps for a total of 25 rounds.
At the beginning of each round, we generate 10k random samples using the last model in the previous round.
and the random prompts from LAIONPOP \cite{LAION_POP}. We then manually pick 1000 samples to add back to the training dataset.
The newly added samples are given a 2x higher probability of appearing in training batches in the next round.
We then train the latent transparency encoder-decoder and diffusion models again.
After 25 rounds, the size of the dataset increases to 45K. 
Afterwards, we generate 5M sample pairs without human interaction and use the LAION Aesthetic threshold\,\cite{Schuhmann2022} setting of 5.5 and clip score sorting to obtain 1M sample pairs.
We automatically remove samples that do not contain any transparent pixels as well as those that do not contain any visible pixels.
Finally, all images are captioned with LLaVA \cite{liu2023llava} (an open-source multi-modal GPT similar to GPT4v) to get detailed text prompts.
The training of both the VAE and the diffusion model is finalized with another 15k iterations using the final 1M dataset.

\begin{table}
\caption{\textbf{Statistical Record of Human-in-the-loop Collection.} \revision{We report the Defective Sample Count (DSC) per 100 sampling during the rounds of human data selection. We resume the model checkpoints recorded after each round of data collection and generate 100 samples for each checkpoint. Users find how many samples are of obvious defects (like fully empty image, or fully non-transparent image, or obvious errors like opaque glass, etc) and report the number as *DSC* (lower is better $\downarrow$).}}
\label{tab:loopfault}
\resizebox{\linewidth}{!}{
\begin{tabular}{@{}lrrrrrrrrrrrrrrrr@{}}
\toprule
Round  & \#0 & \#1 & \#2& \#3& \#4& \#5& \#6& \#7& \#8& \#9& \#10& \#12 & \#14 & \#16 & \#18 & \#20 \\ \midrule
DSC $\downarrow$  & 61 & 62& 37& 53& 41& 25& 17& 15& 23& 21& 25 & 11 & 6 & 9 & 3 & 5 \\
\bottomrule
\end{tabular}
}
\end{table}

\paragraph{Statistical Analysis} \revision{
We briefly analyze here how human data selection improves the quality of the dataset as well as the model capabilities. 
As shown in Fig.~\ref{fig:humanloop}, we visualize samples that are preserved or removed in each round of the human-in-the-loop selection. 
We can see that human efforts removes some obvious flaws (\eg, empty images, fully opaque colors for glass, \etc) and enhances the diversity of the dataset content (\eg, the glowing effects on magic books, \etc). In Table~\ref{tab:loopfault}, we resume the checkpoint from each round of data collection to sample images, and ask the users to review the images and count images with obvious defects. We can see that as the number of rounds increases, the rate of defective outputs gradually decreases.
}

\paragraph{Multi-layer Dataset} We further extend our \{\emph{text, transparent image}\} dataset into a \{\emph{text, foreground layer, background layer}\} dataset, so as to train the multi-layer models. As shown in Fig.~\ref{fig:data}-(b), we ask GPTs (we used ChatGPT for 100k requests and then moved to LLAMA2 for 900k requests) to generate structured prompts pairs for foreground like ``a cute cat'', entire image like ``cat in garden'', and background like ``nothing in garden'' (we ask GPT to add the word ``nothing'' to the background prompt). The foreground prompt is processed by our trained transparent image generator (Section \ref{sec:32}) to obtain the transparent images. Then, we use Diffusers Stable Diffusion XL Inpaint model \cite{diffusers2024} to inpaint all pixels with alpha less than one to obtain intermediate images using the prompt for the entire images. Finally, we invert the alpha mask, erode $k=8$ pixels and inpaint again with the background prompt to get the background layer. We repeat this process 1M times to generate 1M layer pairs.

\paragraph{Training Details} We use the AdamW optimizer at learning rate 1e-5 for both VAE and diffusion model. The pretrained Stable Diffusion model is SDXL \cite{Podell2023}. For the the LoRA\,\cite{hu2021lora} training, we always use rank 256 for all layers. We use the Diffusers' standard for naming and extracting LoRA keys. In the human-in-the-loop data collection, each round contains 10k iterations at batch size 16. The training devices are 4x A100 80G NV-link, and the entire training takes one week (to reduce budget, the training is paused when human are collection data for the next round of optimization) and the real GPU time is about 350 A100 hours. Our approach is training friendly for personal-scale or lab-scale research as the 350 GPU hours can often be processed within 1K USD.

\section{Experiments}

We detail qualitative and quantitative experiments with our system.
We first present qualitative results with single images (Section~\ref{sec:41}), multiple layers (Section~\ref{sec:42}), as well as iterative generation (Section~\ref{sec:43}), and then show that our framework can also be combined with control modules for wider applications (Section~\ref{sec:44}).
We then analysis the importance of each component with ablative study  (Section~\ref{sec:45}), and then discuss the difference and connection between our approach and image matting (Section~\ref{sec:46}).
Finally, we conduct perceptual user study (Section~\ref{sec:47}) and present a range of discussions to further study the behaviors of our framework (Section~\ref{sec:48},\ref{sec:49},\ref{sec:410}).

\subsection{Qualitative Results}

\label{sec:41}

\begin{figure*}
\includegraphics[width=\linewidth]{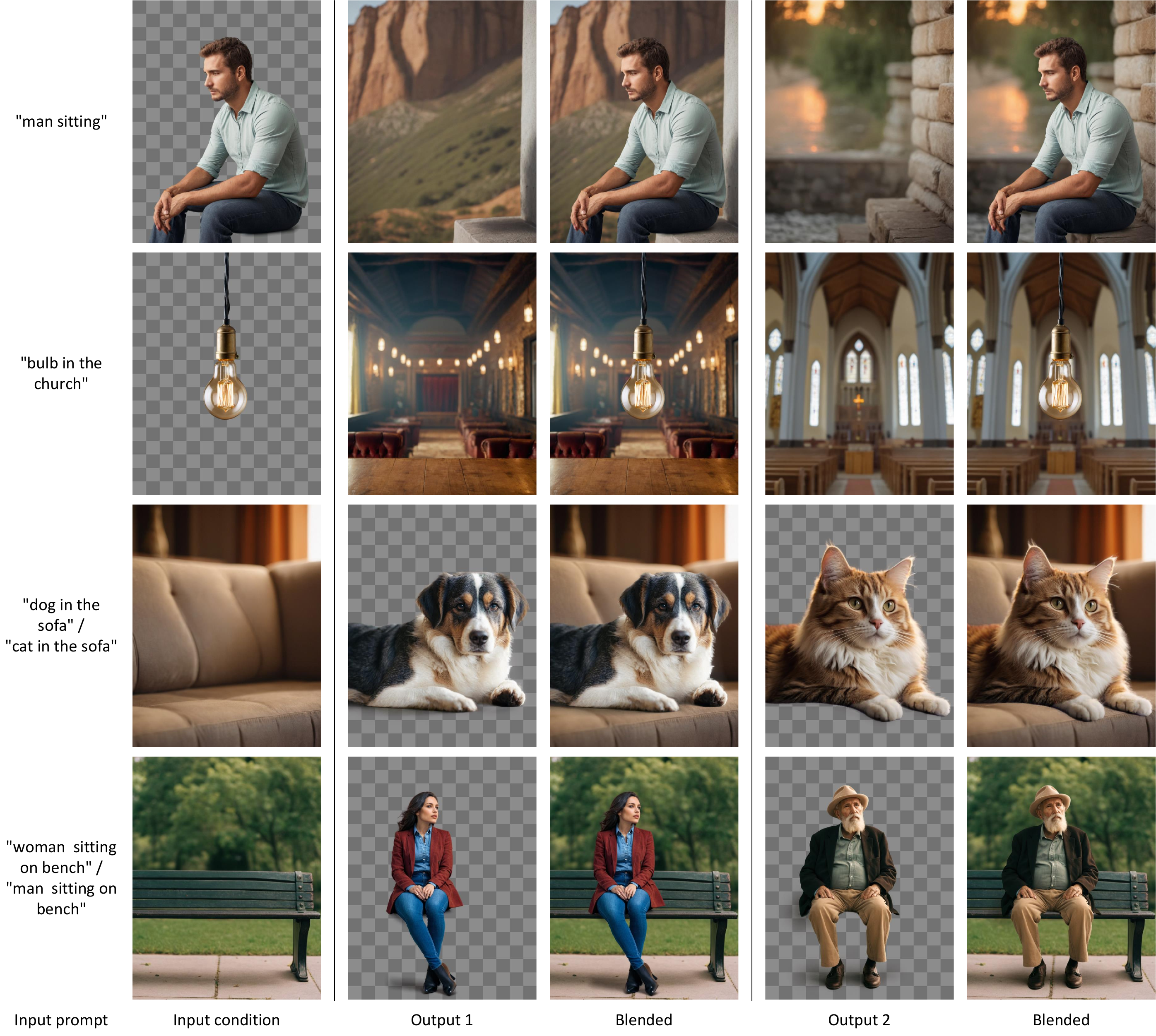}
\caption{\textbf{Conditional Layer Generating.} We presents results with foreground-conditioned background (the first two rows) and background-conditioned foreground (the last two rows). For each example, we generate two foregrounds/backgrounds.}
\label{fig:condition}
\end{figure*}

\begin{figure*}
\includegraphics[width=\linewidth]{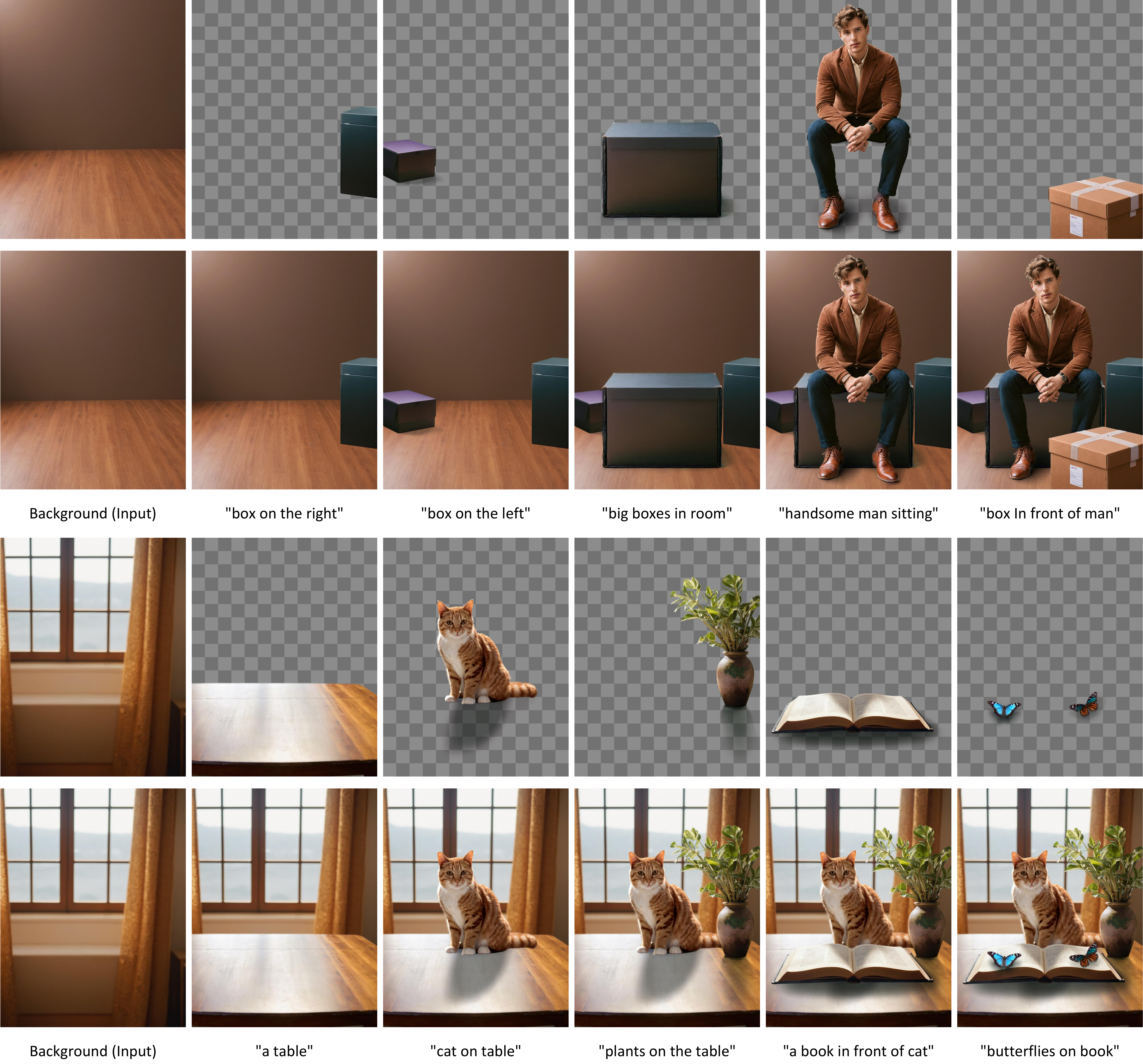}
\caption{\textbf{Generating Multiple Layers.} We show that our framework can compose multiple layers iteratively, by repeating the background-conditioned foreground model. At each step, we blend al existing layers and feed the blended result to the background-conditioned generator. The prompts at each step is at the bottom of outputs.}
\label{fig:many}
\end{figure*}

\begin{figure*}
\includegraphics[width=\linewidth]{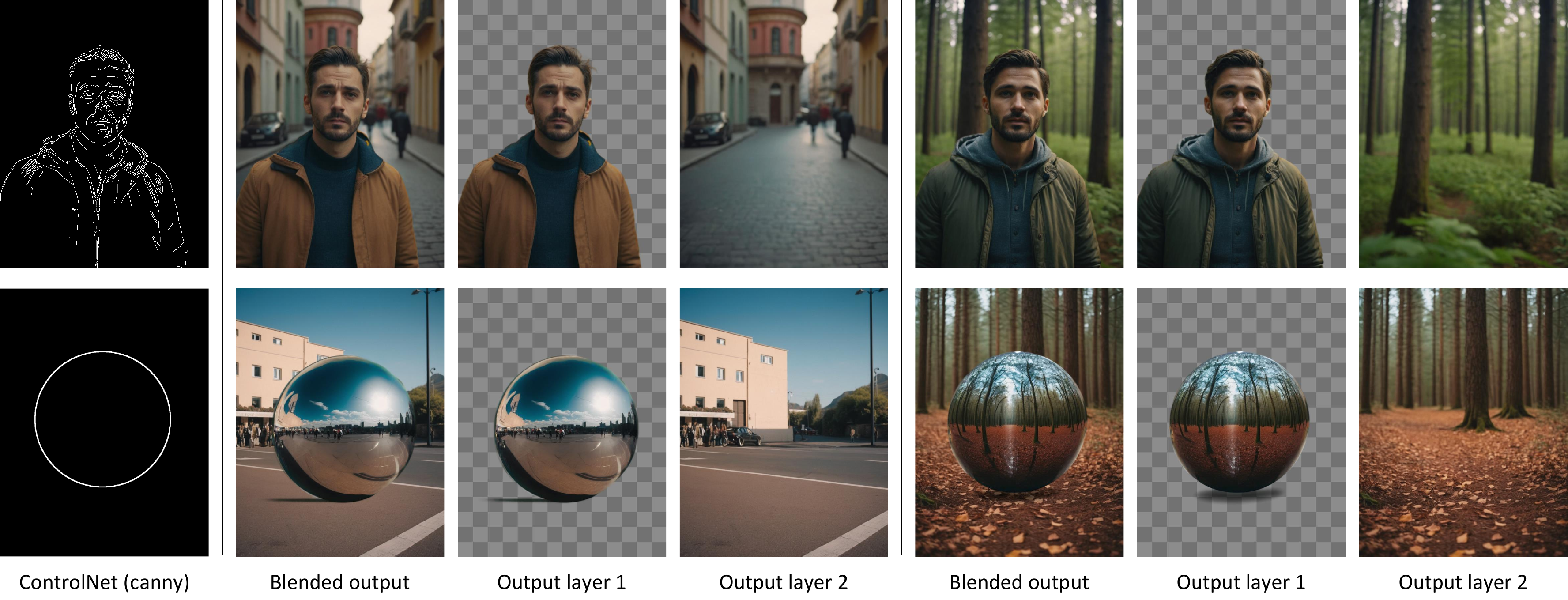}
\caption{\textbf{Combining with Control Models.} We show that our approach can directly be combined with control models like ControlNet\,\cite{zhang2023adding} to enhance the functionality. The prompts are ``human in street'', ``human in forest'', ``big reflective ball in street'', and ``big reflective ball in forest''.}
\label{fig:controlnet}
\end{figure*}

\begin{figure*}
\includegraphics[width=\linewidth]{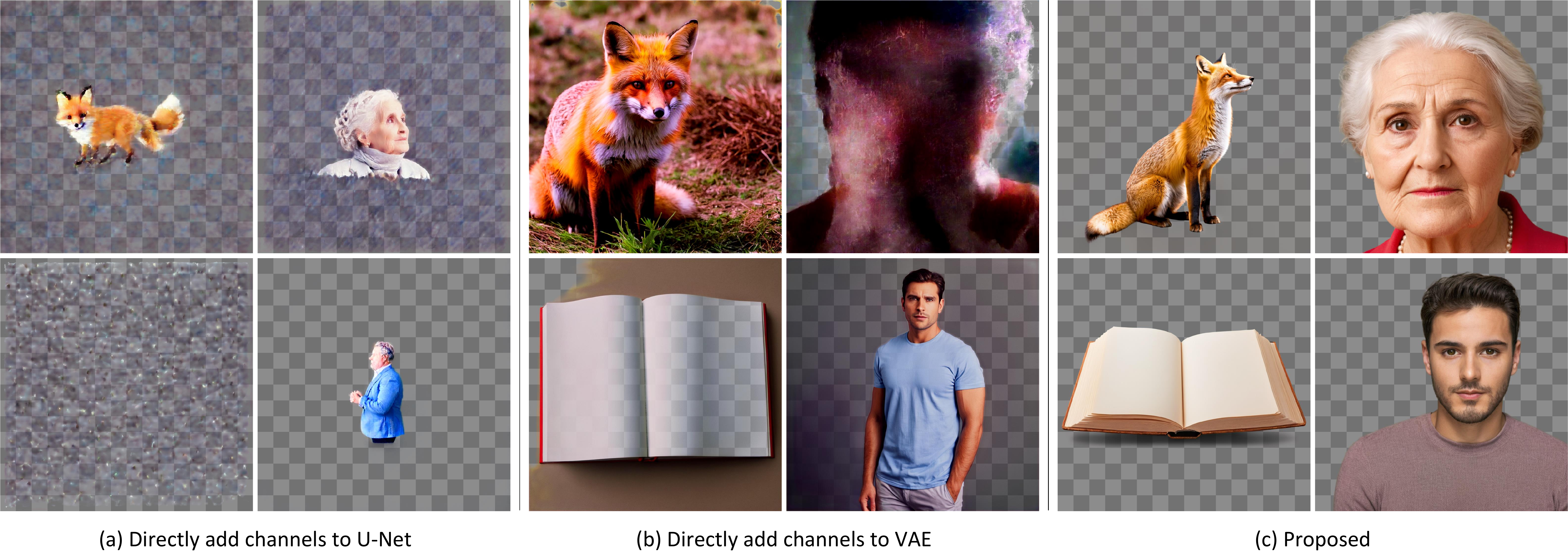}
\caption{\textbf{Ablative Study.} We compare our approach to two alternative architecture: directly adding channels to UNet and directly adding channels to VAE. When adding channel to UNet, we directly encode alpha channel as an external image and add 4 channels to UNet. When adding channels to VAE, the UNet is finetuned on the latent images encoded by the newer VAE. The test prompts are ``fox'', ``elder woman'', ``a book'', ``man''.}
\label{fig:ablation}
\end{figure*}

\begin{figure*}
\includegraphics[width=\linewidth]{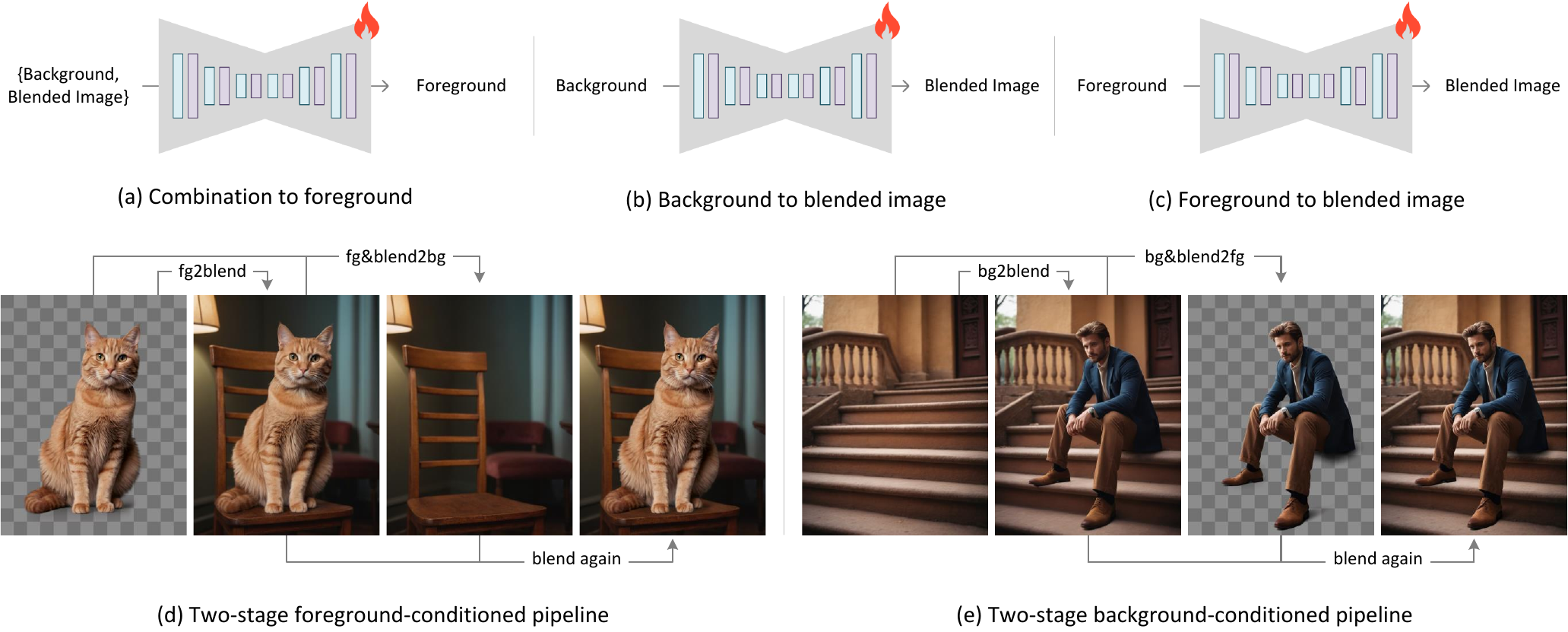}
\caption{\textbf{Additional Ablative Architectures} \revision{We also include several alternative models for more complicated workflows. These include generating a blended image from background or foreground, as well as generating background/foreground from other combined layers. We also demonstrate the use of two-step pipelines to generate/decompose independent layers.}}
\label{fig:morearch}
\end{figure*}

\begin{figure*}
\includegraphics[width=\linewidth]{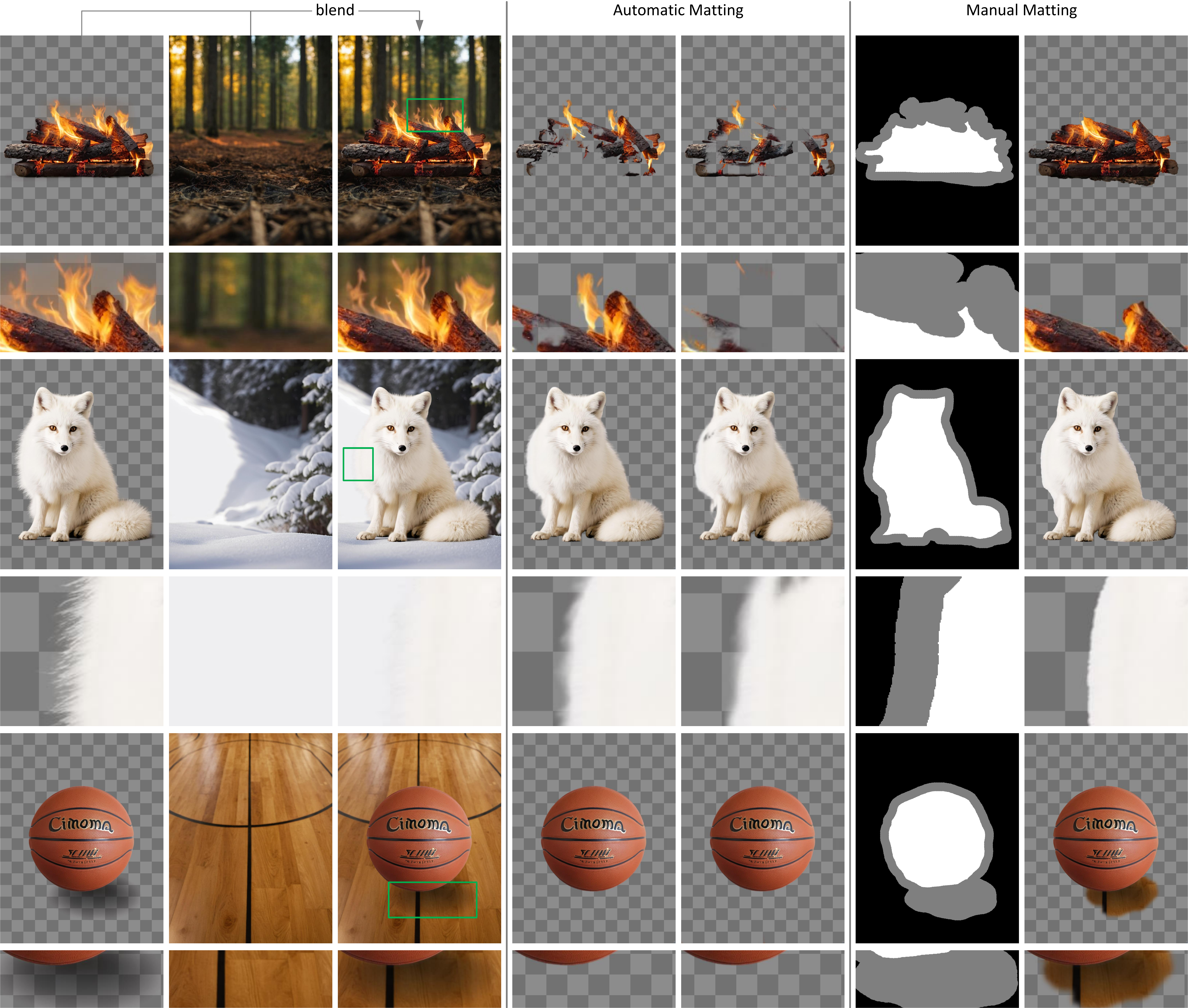}
\begin{tabularx}{\linewidth}{*{7}{>{\centering\arraybackslash}X}}
\small Foreground (ours) & 
\small Background (ours) & 
\small Blended (ours) & 
\small \cite{Chen2022} & 
\small \cite{li2023matting} & 
\small User tri-map & 
\small \cite{yao2024vitmatte}
\end{tabularx}
\caption{\textbf{Difference between Joint Layer Generating and Generating-then-matting.} \revision{This is \emph{not} a result comparison since the left images are outputs layers of our method. The blended images are alpha blending of the generated layers. (Our method does not decompose images.) We try to reproduce similar results using matting approaches. The prompts are ``fire on burning wood in forest'', ``white fox in white snow ground, all white, very white'', and ``basketball''.}}
\label{fig:comparison}
\end{figure*}

We present qualitative results in Fig.~\ref{fig:qualitative} with a diverse set of transparent images generated using our single-image base model. These results showcase the model's capability to generate \emph{natively} transparent images that yield high-quality glass transparency, hair, fur, and semi-transparent effects like glowing light, fire, magic effect, \etc. These results also demonstrate the model's capability to generalize to diverse content topics.

We further present multi-layer results in Fig.~\ref{fig:multilayer} with transparent layers generated by our multi-layer model and the blended images. These results showcase the model's capability to generate harmonious compositions of objects that can be blended together seamlessly. The layers are not only consistent with respect to illumination and geometric relationships, but also demonstrate the aesthetic quality of Stable Diffusion (\eg, the color choice of the background and foreground follows a learned distribution that looks harmonious and aesthetic).

\subsection{Conditional Layer Generation}

\label{sec:42}

We present conditional layer generation results (\ie, foreground-conditioned background and background-conditioned foreground generation) in Fig.~\ref{fig:condition}, 
We can see that the model is able to generate consistent composition with coherent geometry and illumination. In the  ``bulb in the church'' example, the model tries to generate a aesthetic symmetric design to match the foreground. The ``sitting on bench''/``sitting on sofa'' examples demonstrate that the model is able to infer the interaction between foreground and background and generate corresponding geometry.

\subsection{Iterative Generation}

\label{sec:43}

Fig.~\ref{fig:many} shows that we can iteratively use the background-conditioned foreground generation model to achieve composition or arbitrary number of layers.For each new layer, we blend all previously generated layers into one RGB image and feed it to the background-conditioned foreground model. We also observe that the model is able to interpret natural language in the context of the background image, \eg, generating a book in front of the cat.
The model displays strong geometric composition capabilitites, \eg, composing a human sitting on a box. 

\subsection{Controllable Generation}

\label{sec:44}

As shown in Fig.~\ref{fig:controlnet}, we demonstrate that existing control models like ControlNet\,\cite{zhang2023adding} can be applied to our model for enriched functionality. We can see that the model is able to preserve the global structure according to the ControlNet signal to generate harmonious compositions with consistent illumination effects. We also use a ``reflective ball'' example to show that the model is able to interact with the content of the foreground and background to generate consistent illumination like the reflections.

\subsection{Ablative Study}

\label{sec:45}

We conduct an ablative study to evaluate the contribution of each component in our framework. We are interested in a possible architecture that does not modify Stable Diffusion's latent VAE encoder/decoder, but only adds channels to the UNet. 
In the original Stable Diffusion, a $512\times512\times3$ image is encoded to a latent image of size $64\times64\times4$. This indicates that if we duplicate the $512\times512\times1$ alpha channel 3 times into a $512\times512\times3$ matrix, the alpha could be directly encoded into a $64\times64\times4$ latent image. By concatenating this with the original latent image, the final latent image would form a a $64\times64\times8$ matrix. This means we could add 4 channels to Stable Diffusion UNet to force it support an alpha channel. We present the results of this approach in Fig.~\ref{fig:ablation}-(a). We can see that this method severely degrades the generation quality of the pretrained large model, because its latent distribution is changed; although the VAE is unchanged (it is frozen), the additional 4 channels significantly change the feature distribution after the first convolution layer in the VAE UNet. Note that this is different from adding a control signal to the UNet --- the UNet must generate and recognize the added channels all at the same time because diffusion is a iterative process, and the outputs of any diffusion step become the input of the next diffusion step.

In Fig.~\ref{fig:ablation}-(b), we test another architecture that directly adds a channel to the VAE encoder and decoder. We train the VAE to include an alpha channel, and then further train the UNet. We observe that such training is very unstable, and the results suffer from different types of collapse from time to time. The essential reason leading to this phenomenon is that the latent distribution is changed too much during in the VAE fine-tuning.

\revision{We also introduce several alternative architectures in Fig.~\ref{fig:morearch} for more complicated workflows. We can add zero-initialized channels to the UNet and use the VAE (with or without latent transparency) to encode the foreground, or background, or layer combinations into conditions, and train the model to generate foreground or background or directly generate blended images (\eg, Fig.~\ref{fig:morearch}-(a, b, c)). We visualize examples of this two-stage pipeline in Fig.~\ref{fig:morearch}-(d, e).}

\subsection{Relationship to Image Matting}

\label{sec:46}

We discuss the difference and connection between native transparent image generation and image matting.
To be specific, we test the following matting methods:
(1) \emph{PPMatting}~\cite{Chen2022} is a state-of-the-art neural network image matting model. This model reports to achieve the highest precision among all ``classic'' neural network based matting methods, \ie, neural models trained from scratch on a collected dataset of transparent images. This model is fully automatic and does not need a user-specified tri-map.
(2) \emph{Matting Anything}~\cite{li2023matting} is a new type of image matting model based on the recently released
Segment Anything Model (SAM)~\cite{kirillov2023segany}. This model uses pretrained SAM as a base and finetunes it to perform matting. This model also does not need a user-specified tri-map.
We also include a tri-map-based method to study the potential for user-guided matte extraction.
(3) \emph{VitMatte}~\cite{yao2024vitmatte} is a state-of-the-art matting model that uses tri-maps. The architecture is a Vision Transformer (ViT) and represents the highest quality of current user-guided matting models.

\begin{table}
\caption{\textbf{User Study.} We present the results from user study. We conduct user study in two groups: the first group compares outputs between different methods, while the second group directly compare our generated results to the search result of a commercial transparent image assets (Adobe Stock). Higher is better and best in bold.}
\label{tab:user}
\resizebox{\linewidth}{!}{
\begin{tabular}{@{}lrr@{}}
\toprule
Candidate   & Group 1& Group 2\\ \midrule
SD + PPMatting \cite{Chen2022} & 2.1$\pm$1.2\% & /\\
SD + Matting Anything \cite{li2023matting} & 0.8$\pm$0.5\% & /\\
Ours (base model) & \textbf{97.1$\pm$1.9\%} & 45.3$\pm$9.1\%\\
\midrule
Commercial Transparent Asset Stock & /& \textbf{54.7$\pm$8.3\%}\\
\bottomrule
\end{tabular}
}
\end{table}

As shown in Fig.~\ref{fig:comparison}, we can see that several types of patterns are difficult for matting approaches, \eg, semi-transparent effects like fire, pure white fur against a pure white background, shadow separation,\etc. For semi-transparent contents like fire and shadows, once these patterns are blended with complicated background, separating them becomes a nearly impossible task. To obtain perfectly clean elements, probably the only method is to synthesize elements from scratch, using a native transparent layer generator. We further notice the potential to use outputs of our framework to train matting models. 

\subsection{Perceptual User Study}

\label{sec:47}

\begin{figure}
\includegraphics[width=\linewidth]{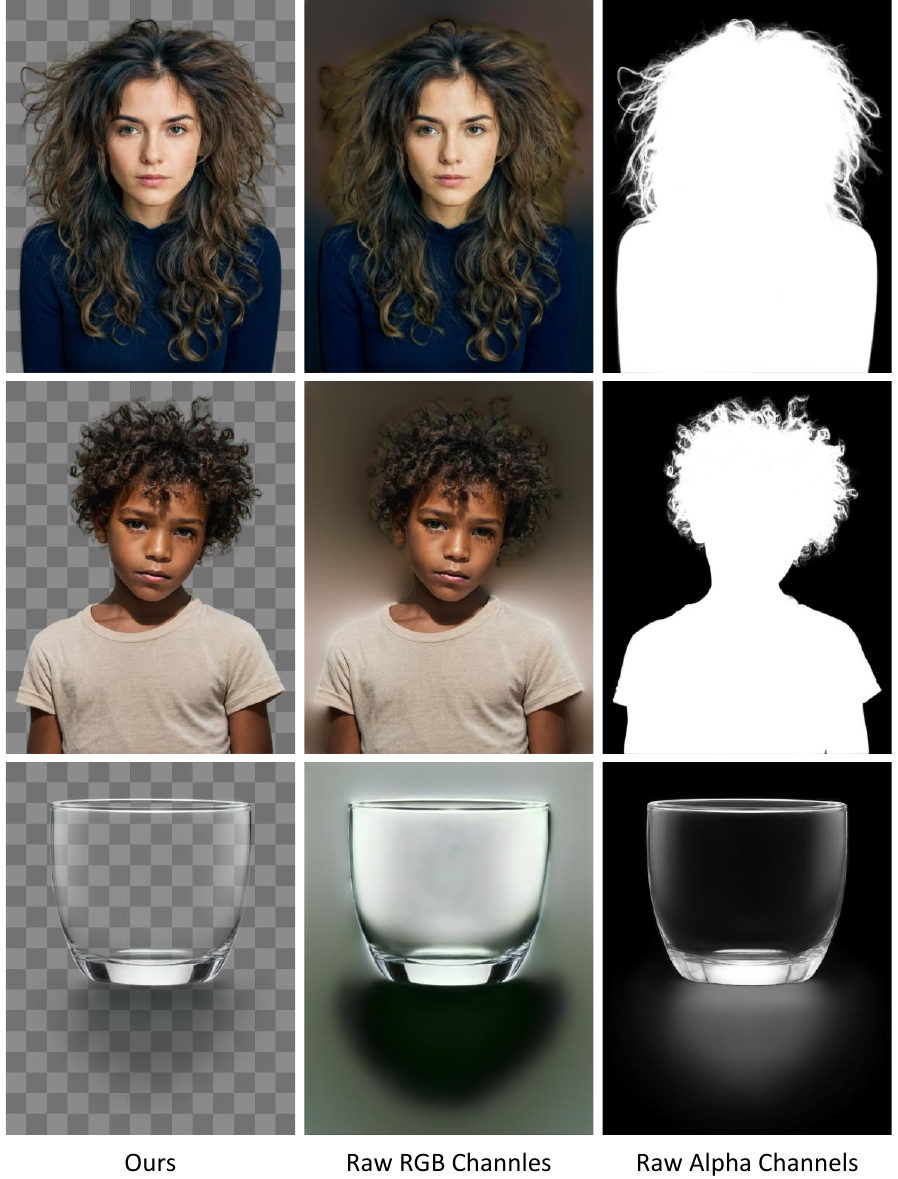}
\caption{\textbf{Raw outputs of the RGB channels and alpha channel.} We present the raw RGB and alpha channel for evaluation. The prompts are ``woman with messy hair'', 
 ``boy with messy hair'', and ``glass cup''.}
\label{fig:raw}
\end{figure}

\begin{figure}
\includegraphics[width=\linewidth]{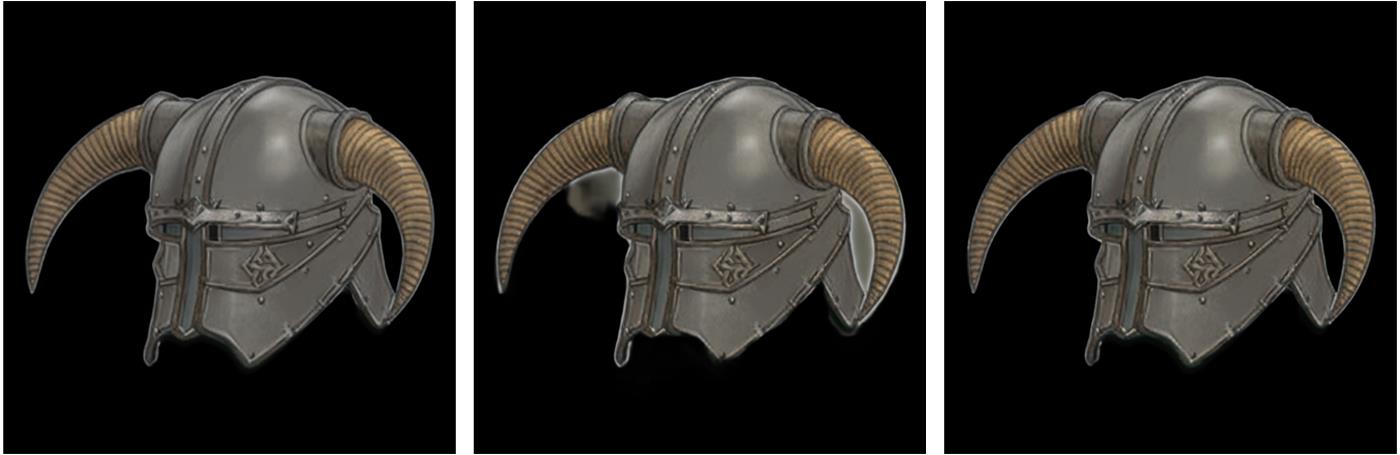}
\begin{tabularx}{\linewidth}{*{3}{>{\centering\arraybackslash}X}}
\small w/ latent offset & 
\small w/o latent offset & 
\small + data augmentation
\end{tabularx}
\caption{\textbf{Robust decoder with data augmentations.} We show that it is possible to use data augmentation methods to train a robust decoder to handle situations when the UNet cannot diffuse the desired latent offsets. Samples are transparent images on black backgrounds.}
\label{fig:fabb}
\end{figure}

\revision{In order to perceptually evaluate and compare our approach with existing methods, we perform a perceptual user study focusing on human aspects of our native transparent results and ad-hoc methods like Stable Diffusion + generation-and-matting. We target real-world use cases where users want to get transparent elements given specific demands (prompts). Our study tests multiple types of methods (native transparent generation, generating-and-matting, online commercial stock) to see how they fulfill such demands (by asking users which they prefer).}

Specifically, our user study involves 14 individuals, where 11 individuals are online crowd-source workers, 1 is a computer science student, and the other 2 are professional content creators. We sample 100 results using the 3 methods (prompts are randomly sampled from PickaPic~\cite{Kirstain2023PickaPicAO}), and this leads to 100 result groups, with each group containing 3 results from 3 methods. The participants are invited to rank the results in each group. When ranking the results in each group, we ask users the question -- ``Which of the following results do you prefer most? Please rank the following transparent elements according to your preference''. We use the preference rate as the testing metric. This process is repeated 4 times to compute the standard deviation. Afterwards, we calculate the average preference rate of each method. We call this user study ``group 1''.

We compare our approach with SD+PPMatting\,\cite{Chen2022}, SD+Matting Anything\,\cite{li2023matting}. Herein, ``SD+'' means we first use Stable Diffusion XL to generate an RGB image, and then perform matting using the corresponding method. Results are shown in Table.~\ref{tab:user}, group 1. We find that users prefer our approach over all other approaches (in more than 97\% cases). This demonstrates the advantage of native transparent image generation over ad-hoc solution like generation-then-matting.

We also perform another user preference experiment in ``group 2'', comparing our results against searching for commercial transparent assets from Adobe Stock, using the same aforementioned user preference metric. In Table.~\ref{tab:user}, group 2, we report that the preference rate of our method is close to commercial stock (45.3\% v.s. 54.7\%). Though the high-quality paid content from commercial stock is still preferred marginally. This result suggests that our generated transparent content is competitive to commercial sources that require users to pay for each image.

\subsection{Raw RGBA Channels}

\label{sec:48}

Fig.~\ref{fig:raw} shows the raw outputs with each channel in our generated transparent images. We can see that the model avoids aliasing by padding the RGB channel with smooth ``bleeding'' colors. This approach ensure high-quality foreground color 
in areas of alpha blending.

\subsection{Robust Decoder with Data Augmentations}

\label{sec:48b}

Fig.~\ref{fig:fabb} shows that we can use data augmentation methods to achieve more robust decoder to handle situations when the latent offset is missing or wrong. This can be useful when certain community models (\eg, anime, cartoon, \etc) fail to produce the desired latent offset during the diffusion process, and we still want to decode useful transparent images from those slightly mismatched latent spaces yielded by those fine-tuned models. To be specific, Fig.~\ref{fig:fabb} simply dropout 30\% offsets when training the decoder.

\subsection{Community Models}

\label{sec:49}

\begin{figure}
\includegraphics[width=\linewidth]{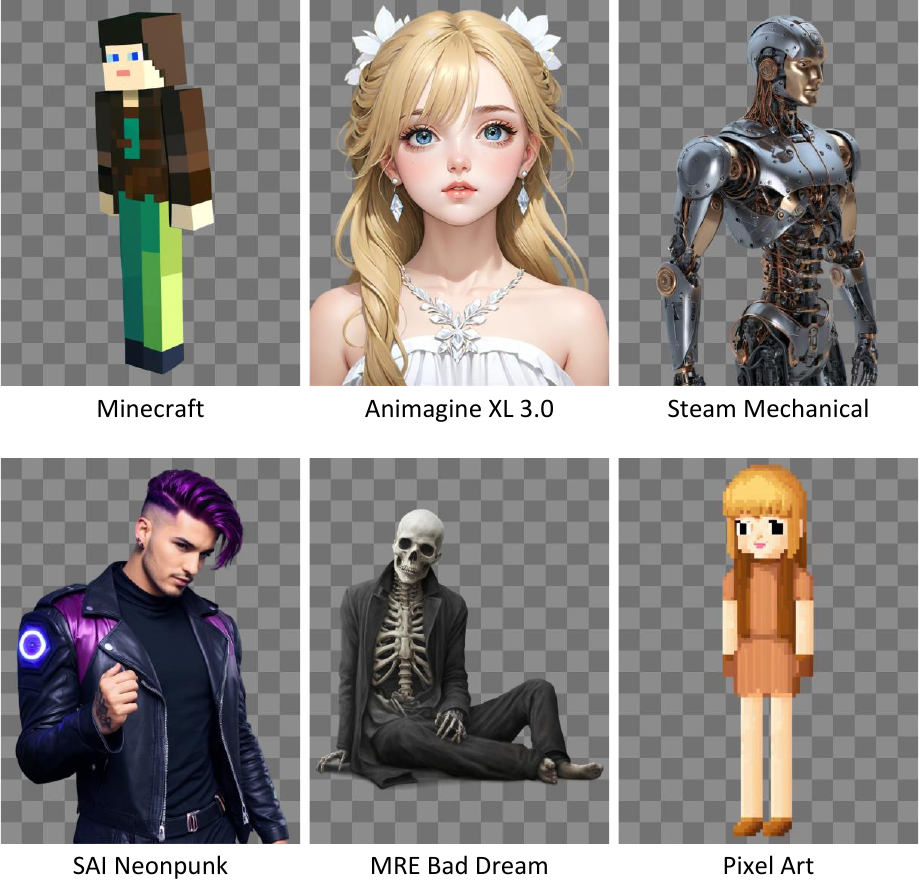}
\caption{\textbf{Applying to Community Models.} We show that our model can be applied to community LoRAs/Models/PromptStyles to achieve diverse results. All images are achieved using prompt ``person'', excepting Animagine using ``1girl, masterpiece, fantastic art''. }
\label{fig:community}
\end{figure}

As shown in Fig.~\ref{fig:community}, our method can be applied to various community models, LoRAs, and prompt styles, without additional training. More specifically, we try a Minecraft LoRA, a pixel art LoRA, an anime model\,\cite{cagliostrolab2024}, and several community prompt styles. We can see that applying to different models neither degrades the quality of target model/LoRAs nor degrades the quality of image transparency. This integration capability suggests the model potential for wider use in diverse creative and professional domains.

\subsection{Inference Speed}

\begin{table}
\caption{\textbf{Inference Speed.} \revision{We report on the inference speed of our framework under different base diffusion models and architectures. We test with a Nvidia RTX 3070 device and a RTX 4090 device. All models use 30 diffusion sampling steps. The reported data are averages of 64 runs. All two-stage model needs to be run twice for two layers, doubling the inference time. This speed may be affected by different inference software.}}
\label{tab:infertime}
\resizebox{\linewidth}{!}{
\begin{tabular}{@{}lrr@{}}
\toprule
Candidate   & SD 1.5 & SDXL \\ \midrule
Only transparent image (RTX4090) & 1.85s & 7.3s\\
Only transparent image (RTX3070) & 4.13s & 12.5s\\
Two layers (joint, RTX4090) & 5.71s & 21.5s\\
Two layers (joint, RTX3070) & 13.01s & 41.28s\\
Two layers (two-stage method, RTX4090) & 1.92s $\times$ 2 & 4.37s $\times$ 2 \\
Two layers (two-stage method, RTX3070) & 4.77s $\times$ 2 & 14.71s $\times$ 2 \\
\bottomrule
\end{tabular}
}
\end{table}

\revision{In Table~\ref{tab:infertime}, we report the inference speed with different base diffusion models and architectures. All tests are based on personal level computation devices. We tested SD1.5 and SDXL with Nvidia RTX 3070 and RTX 4090. Our tests include generating single transparent images, generating multiple layers jointly, and generating multiple layers using the two-stage pipelines.}

\subsection{Limitations}

\label{sec:410}

\begin{figure}
\includegraphics[width=\linewidth]{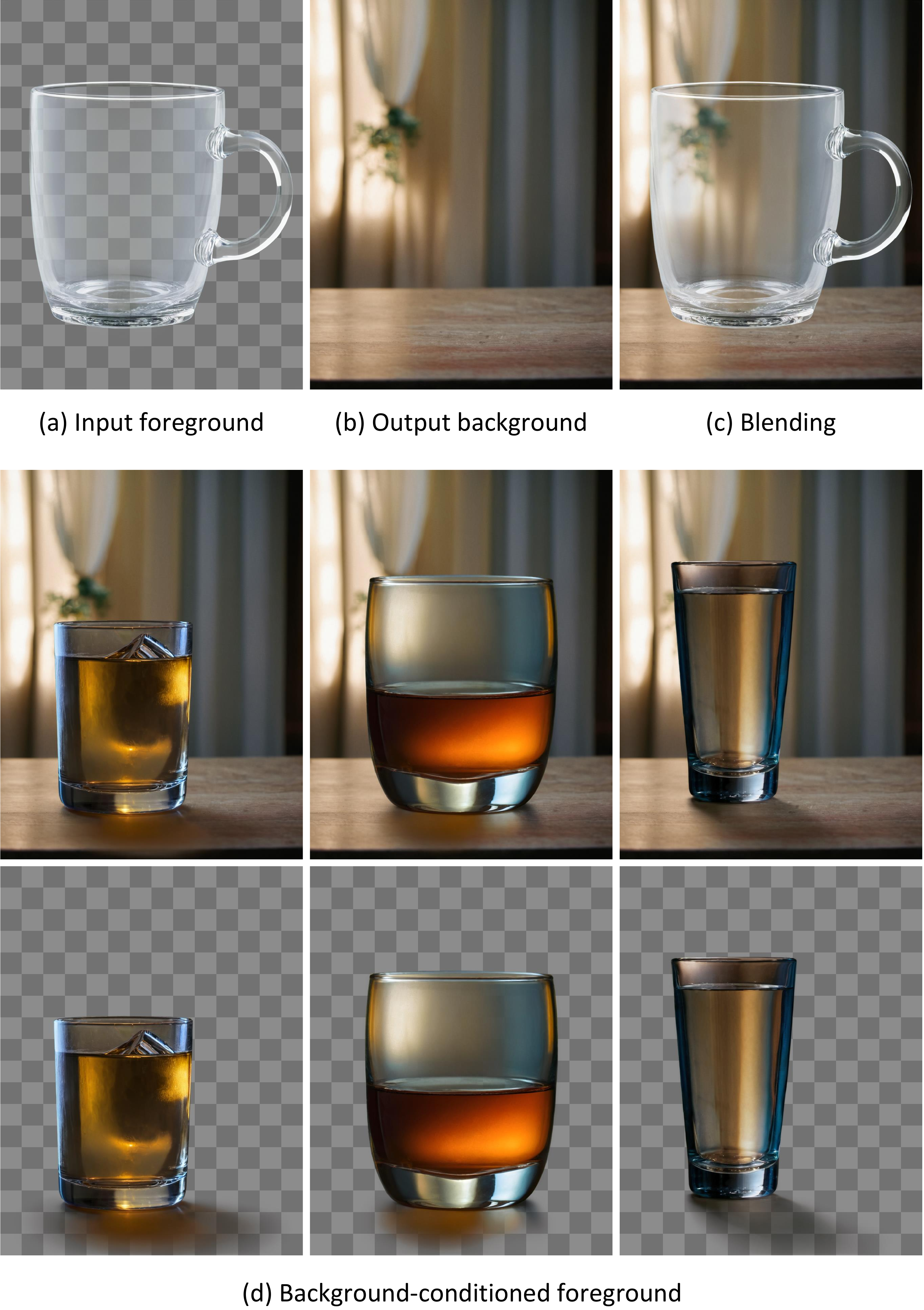}
\caption{\textbf{Limitation.} The prompt in this example is ``glass cup on table in a warm room''. If the input foreground is a clean transparent object without any illumination or shadow effects, harmonious blending is very difficult since the alpha blending does not create deformation of light or casting of shadows. This can be resolved to some extent when using the background as a condition to generate the foreground. But in this case, getting a clean and reusable transparent object without the influence of illumination  is difficult.}
\label{fig:limitation}
\end{figure}

As shown in Fig.~\ref{fig:limitation}, one trade-off with our framework is between generating ``clean transparent elements'' and ``harmonious blending''. For instance, if the transparent image is a clean and resuable element without any special illumination or shadow effects, generating a background that can be harmoniously blended with the foreground can be very challenging and the model may not succeed in every cases (Fig.~\ref{fig:limitation}-(c) is a failure case).
This phenomenon can be cured to some extent if we only use backgrounds as conditions to generate foregrounds to force a harmonious blending (Fig.~\ref{fig:limitation}-(d)). Nevertheless, this will also lead to illumination influencing the transparent object, making the transparent objects less reusable. One may argue that the image in Fig.~\ref{fig:limitation}-(a) is much more reuseable for designers and in-the-wild applications than the transparent images in Fig.~\ref{fig:limitation}-(d) which contain many specific patterns bound to the background.

\section{Conclusion}

In summary, this paper introduces ``latent transparency'', an approach to create either individual transparent images or a series of coherent transparent layers. The method encodes the transparent alpha channel into the latent distribution of Stable Diffusion. This process ensures that the high-quality output of large-scale image diffusion models, by regulating an offset added to the latent space. The training of the models involved 1M pairs of transparent image layers, gathered using a human-in-the-loop collection scheme. We present a range of applications, such as generating layers conditioned on foreground/background, combining layers, structure-controlled layer generating, \etc. User study results indicate that in a vast majority of cases, users favor the transparent content produced natively by our method over traditional methods like generation-then-matting. The quality of the transparent images generated was found to be comparable to the assets in commercial stocks.

\begin{acks}
This work was partially supported by Google through their affiliation with Stanford Institute for Human-centered Artificial Intelligence (HAI).
\end{acks}

\bibliographystyle{ACM-Reference-Format}
\bibliography{layer2024}

\appendix

\section{Padded RGB Channels}
\label{app:1}

In the RGB channel of a transparent RGBA image, we refer to pixels that are completely invisible as ``undefined'' pixels, \ie, pixels with alpha value strictly equal to zero. Since these pixels are strictly invisible, processing them with arbitrary color does not influence the appearance of images after alpha blending. Nevertheless, since neural networks tends to produce high-frequency patterns surrounding image edges, we avoid unnecessary edges in the RGB channels to avoid potential artifacts. We define a local Gaussion filter 
\begin{equation}
G(\bm{I}_c)_p = \left\{
\begin{array}{rcl}
\phi(\bm{I}_c)_p &      & {\text{, if} (\bm{I}_a)_p = 0}\\
(\bm{I}_c)_p &      & {\text{, otherwise}}\\
\end{array}
\right.
\end{equation}
where $\phi(\cdot)$ is a standard Gaussian filter with $13*13$ kernel, and $p$ is pixel position. We perform this filter 64 times to completely propagate colors to all ``undefined'' pixels.

\section{Neural Network Architecture}
\label{app:2}

The latent transparency encoder has exactly same neural network architecture with Stable Diffusion latent VAE encoder \cite{Podell2023} (but the input contains 4 channels for RGBA). This model is trained from scratch. The output convolution layer is zero-initialized to avoid initial harmful noise.

The latent transparency decoder is a UNet. The encoding part of this UNet has same architecture as Stable Diffusion's latent VAE encoder, while the decoding part has same architecture as Stable Diffusion's VAE decoder. The input latent is added to the middle block, and all the encoder's feature maps are added to the input of each decoder block with skip connection. To be specific, assuming the input image is $512\times512\times3$ and the input latent is $64\times64\times4$, the feature map goes through $512\times512\times3 \rightarrow 512\times512\times128 \rightarrow 256\times256\times256 \rightarrow 128\times128\times512 \rightarrow 64\times64\times512$ where each $\rightarrow$ is two resnet blocks. Then input latent is projected by a convolution layer to match channel and then added to the middle feature. Then the decoder goes through $64\times64\times512 \rightarrow 128\times128\times512 \rightarrow 256\times256\times256 \rightarrow 512\times512\times128 \rightarrow 512\times512\times3$ and here each $\rightarrow$ also adds the skip features from the encoder's corresponding layers.

\section{PatchGAN Discriminator}
\label{app:3}

We use exactly same PatchGAN Discriminator architecture, learning objective, and training scheduling with Latent Diffusion VAE \cite{rombach2021highresolution}. We directly use the python class \textit{LPIPSWithDiscriminator} from their official code base (the input channel is set to 4). The generator side objective (from \cite{rombach2021highresolution}) can be written as
\begin{equation}
\mathbb{L}_{\text{disc}}(\bm{z}) = \text{relu}(1 - D_{\text{disc}}(\bm{z})) \,,
\end{equation}
where $\bm{z}$ is a matrx with shape $h\times w\times 4$ and $\text{relu}(\cdot)$ is rectified linear unit. The $D_{\text{disc}}(\cdot)$ is a neural network with 5 convolution-normalization-silu layers $512\times512\times3 \rightarrow 512\times512\times64 \rightarrow 256\times256\times128 \rightarrow 128\times128\times256 \rightarrow 64\times64\times512\rightarrow 64\times64\times1$ and the last layer is a patch-wise real/fake classification layer. The last layer does not use normalization and activation.

\section{Single Transparent Images}
	
\begin{figure*}

\begin{minipage}{\linewidth}
\includegraphics[width=0.245\linewidth]{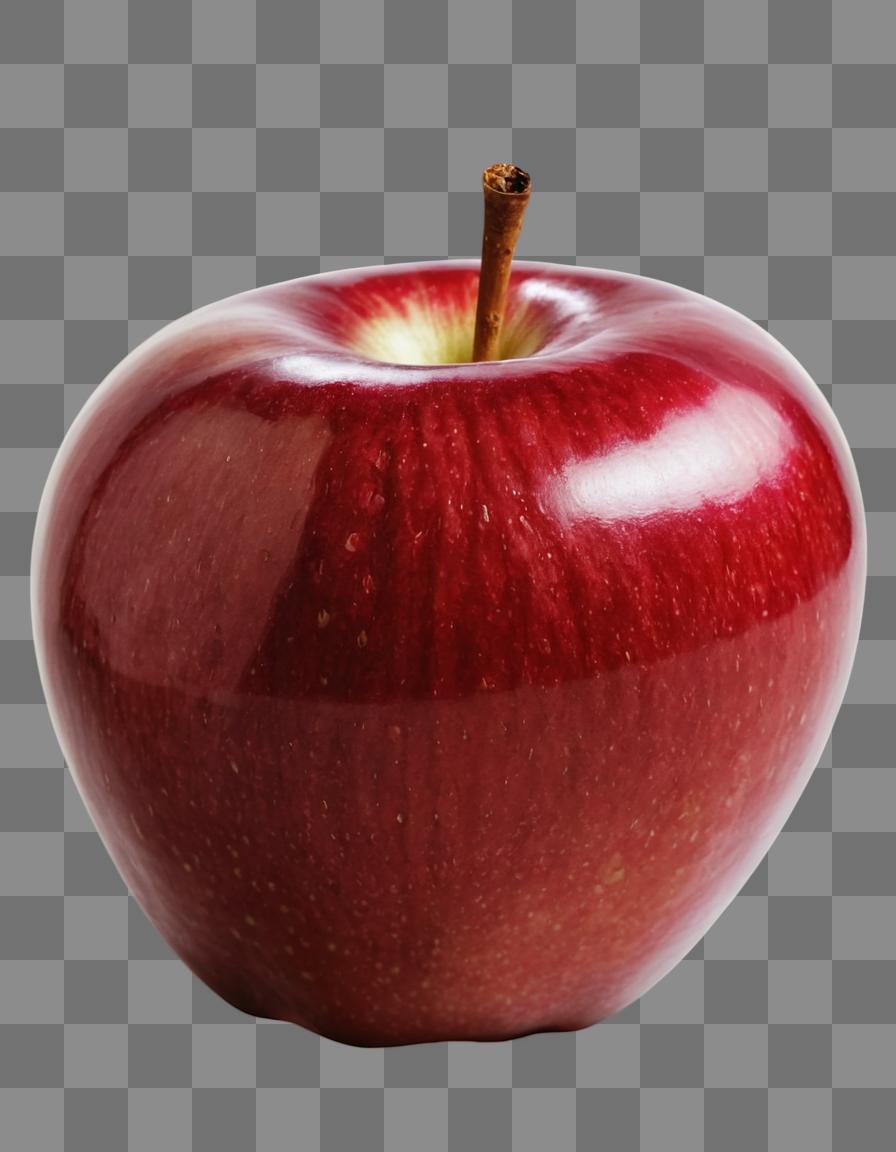}\hfill
\includegraphics[width=0.245\linewidth]{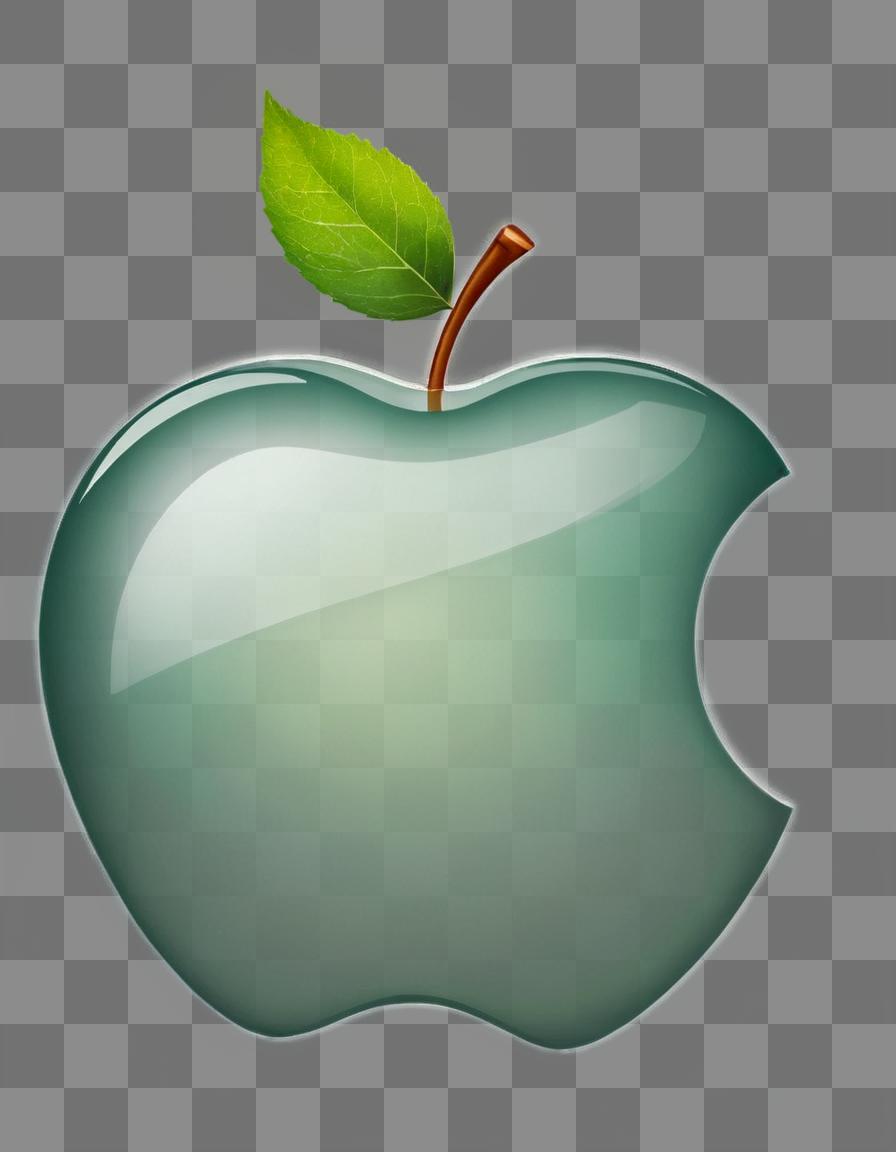}\hfill
\includegraphics[width=0.245\linewidth]{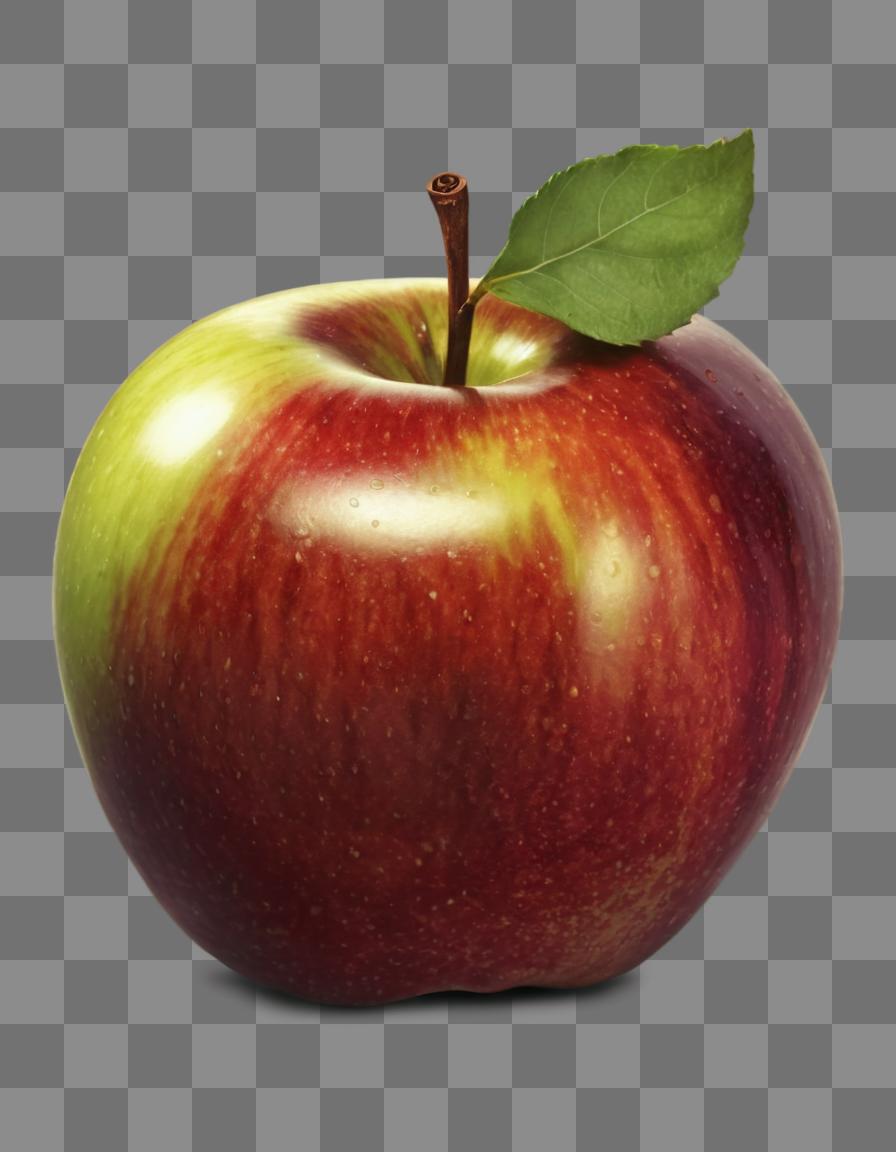}\hfill
\includegraphics[width=0.245\linewidth]{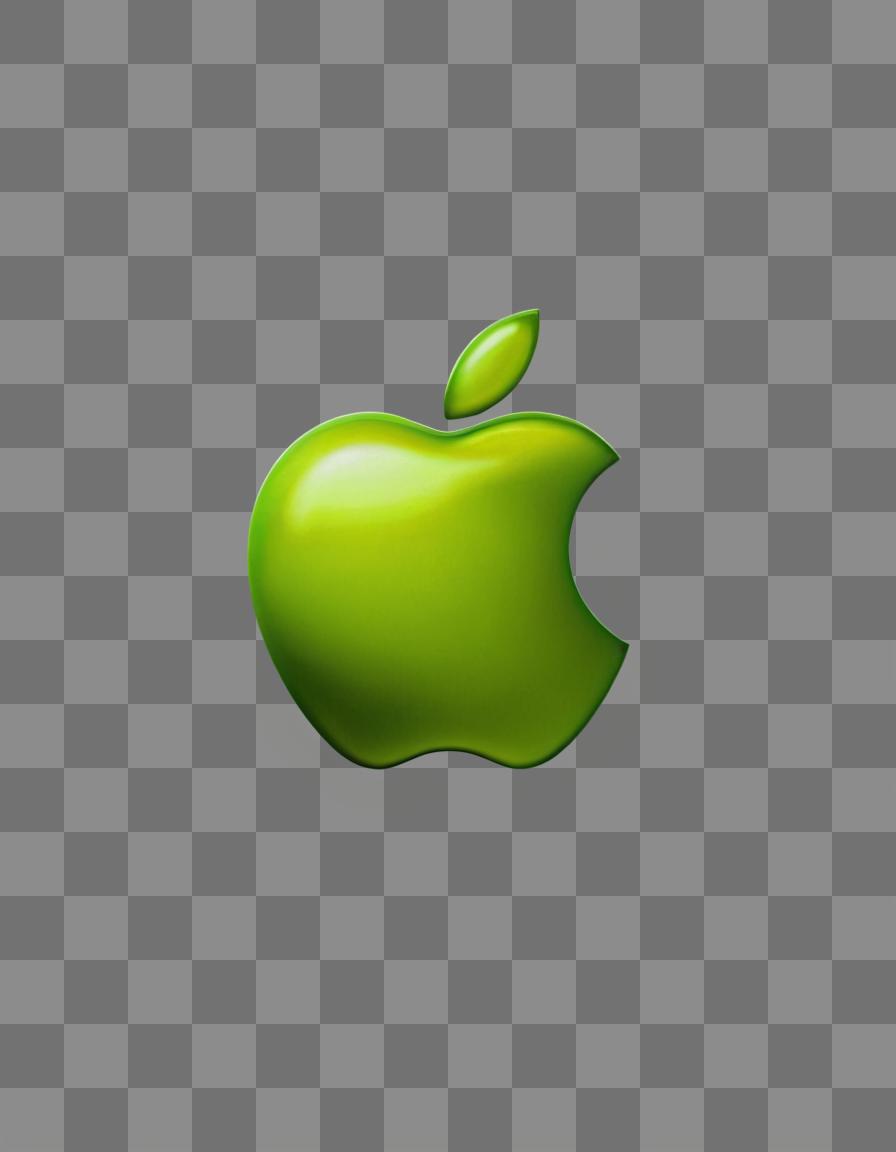}

\vspace{1pt}
\includegraphics[width=0.245\linewidth]{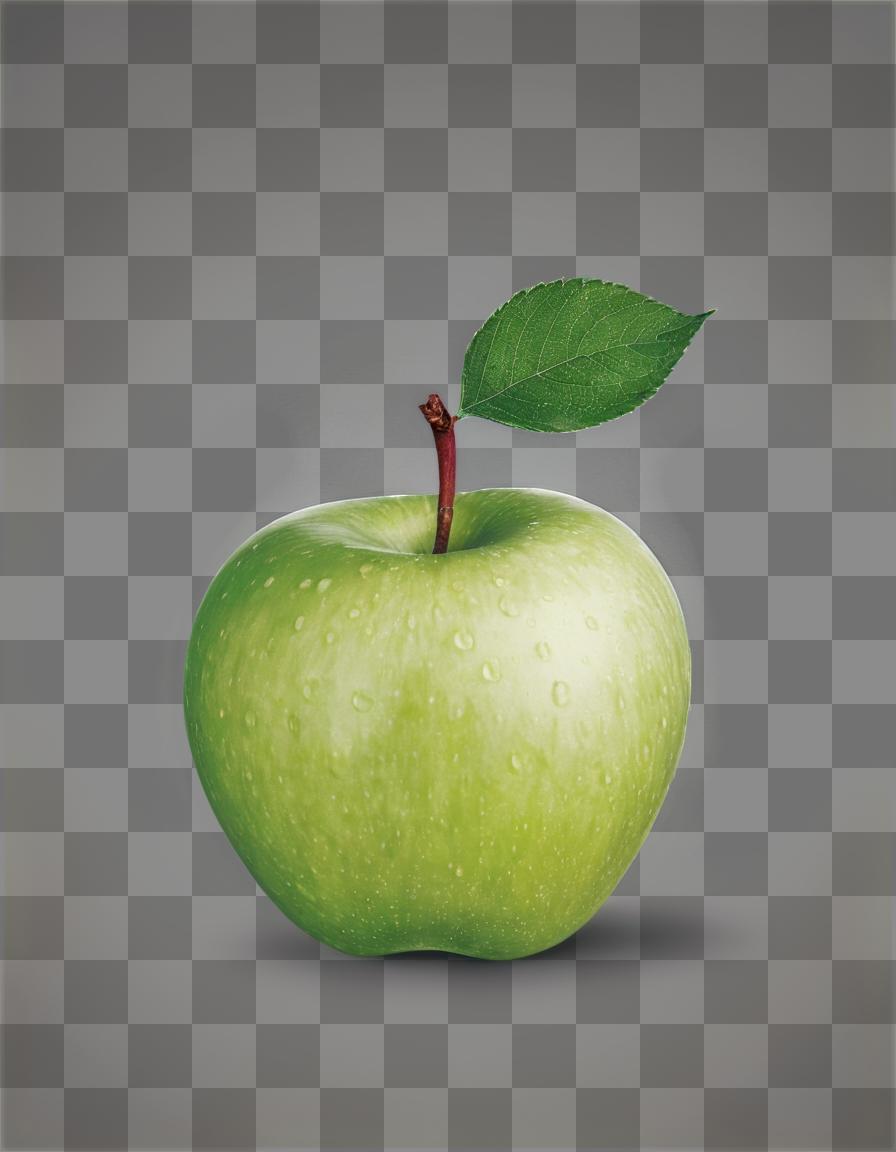}\hfill
\includegraphics[width=0.245\linewidth]{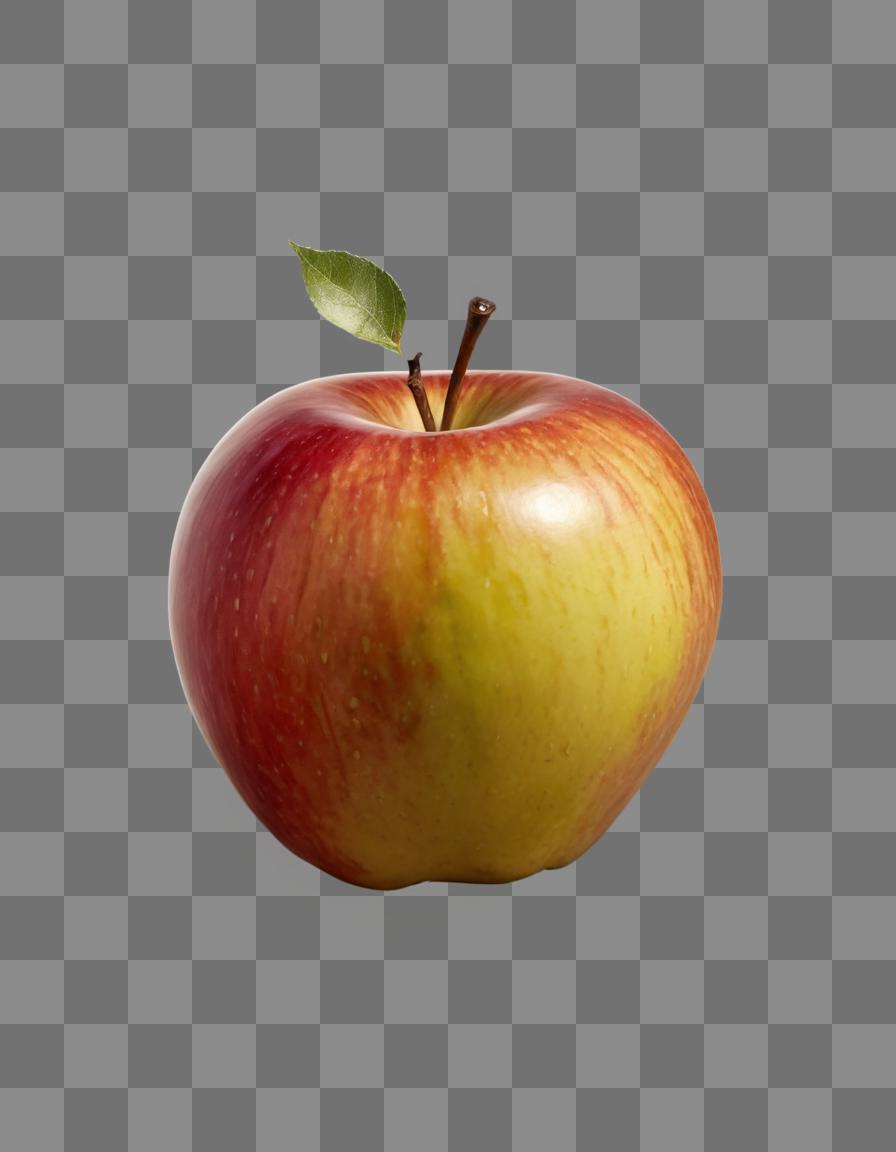}\hfill
\includegraphics[width=0.245\linewidth]{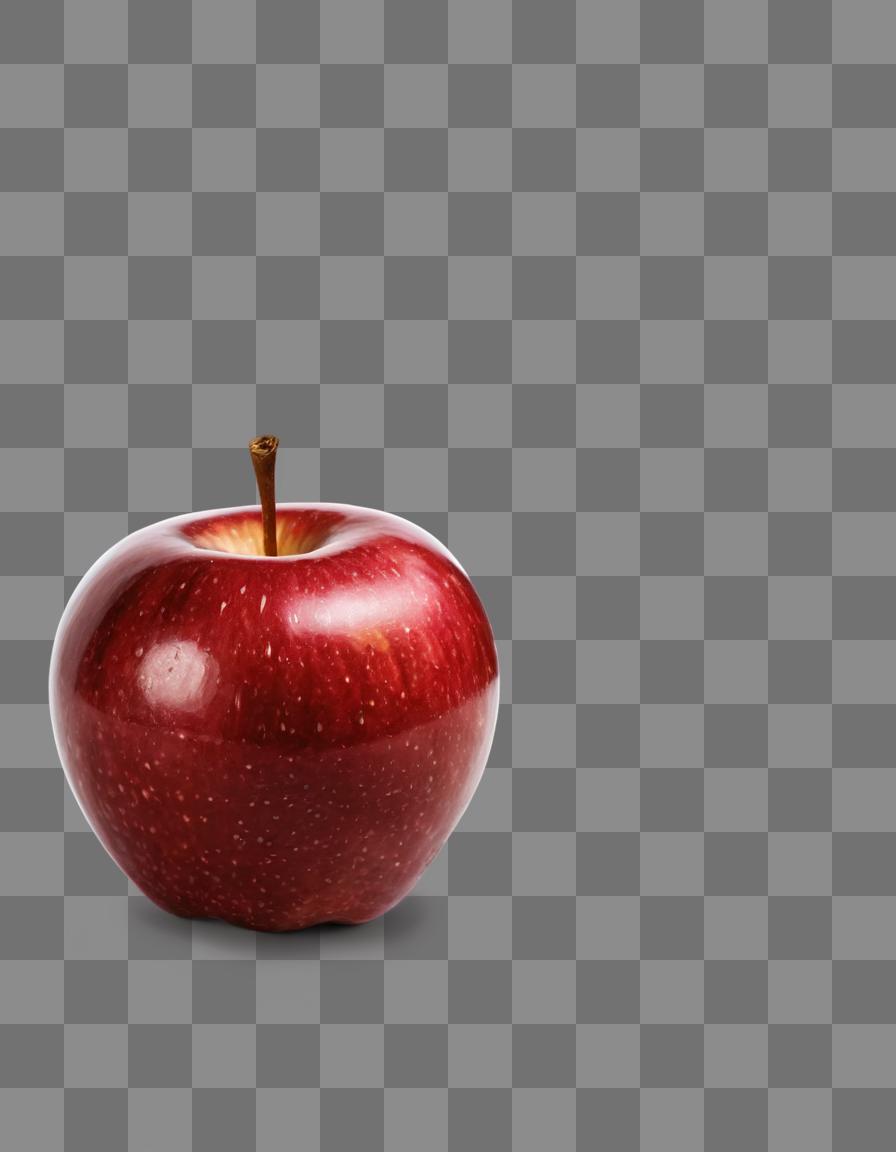}\hfill
\includegraphics[width=0.245\linewidth]{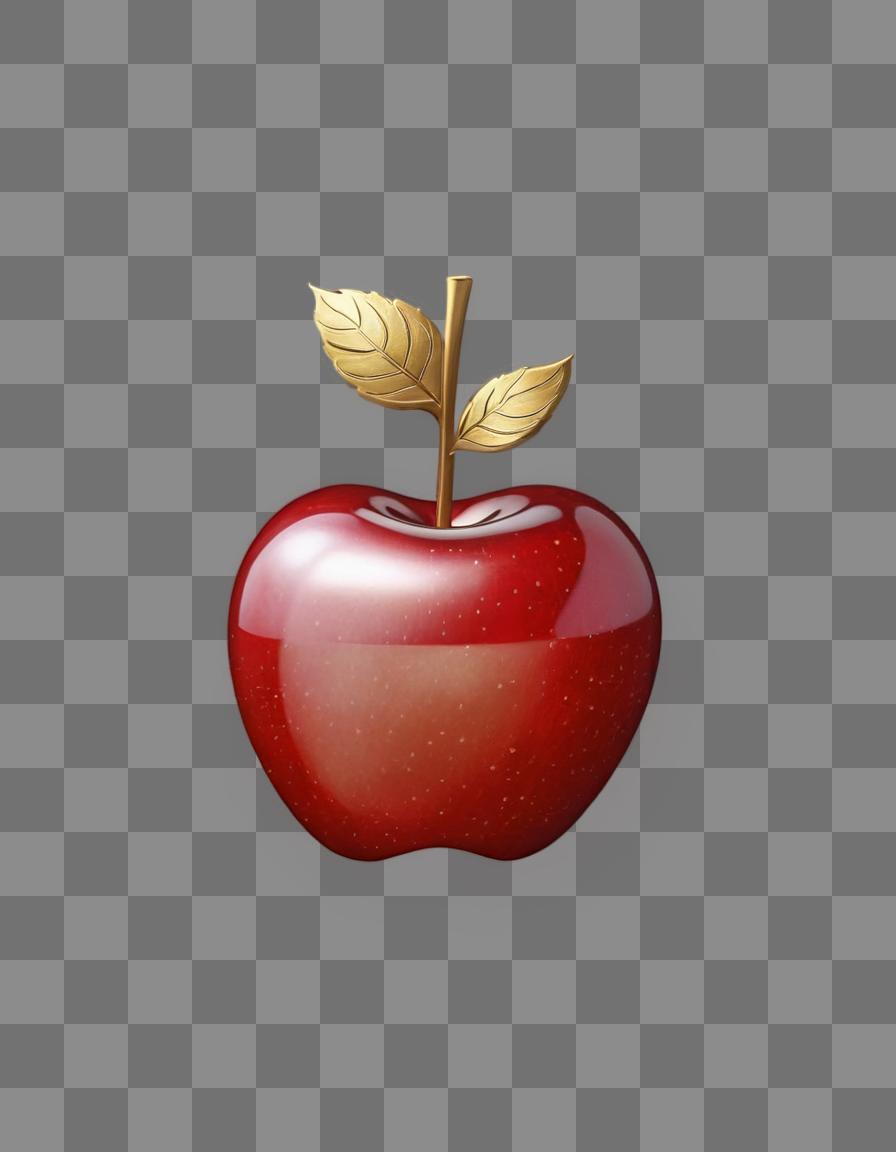}

\vspace{1pt}
\includegraphics[width=0.245\linewidth]{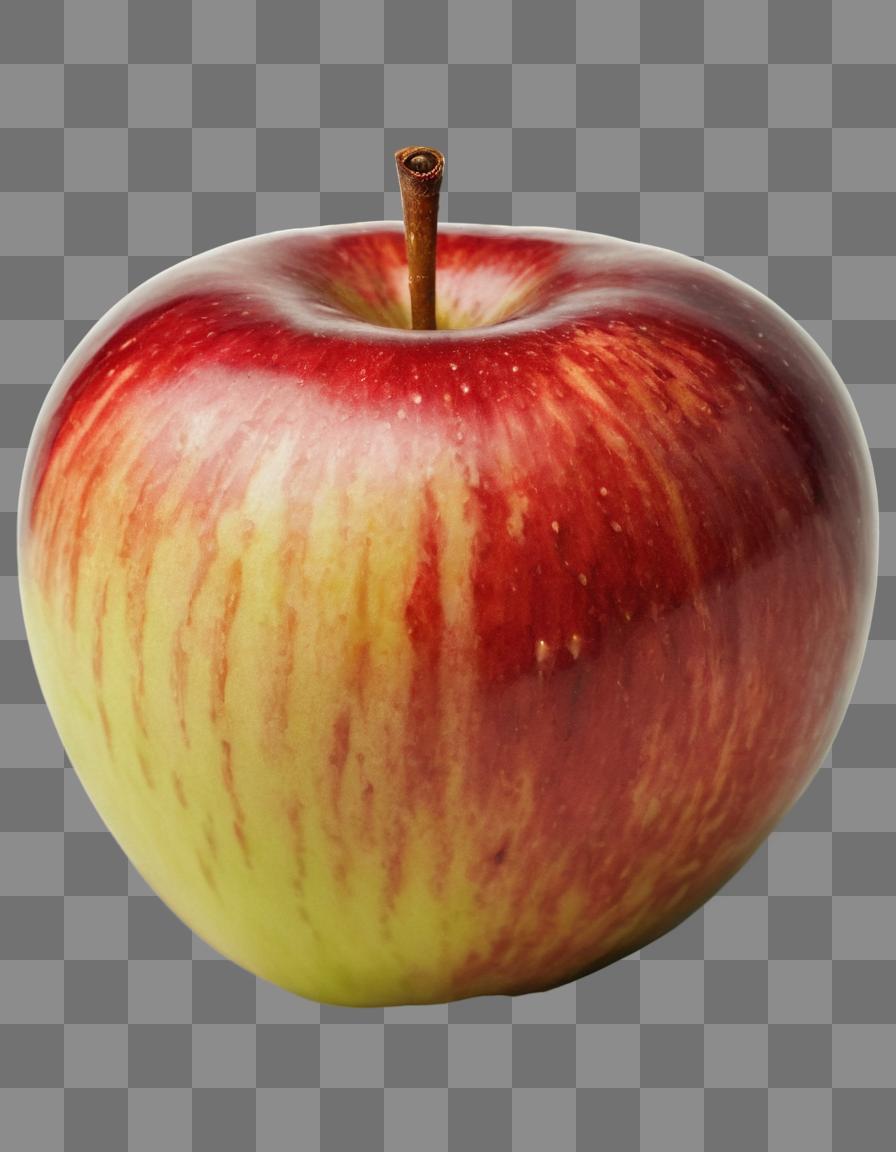}\hfill
\includegraphics[width=0.245\linewidth]{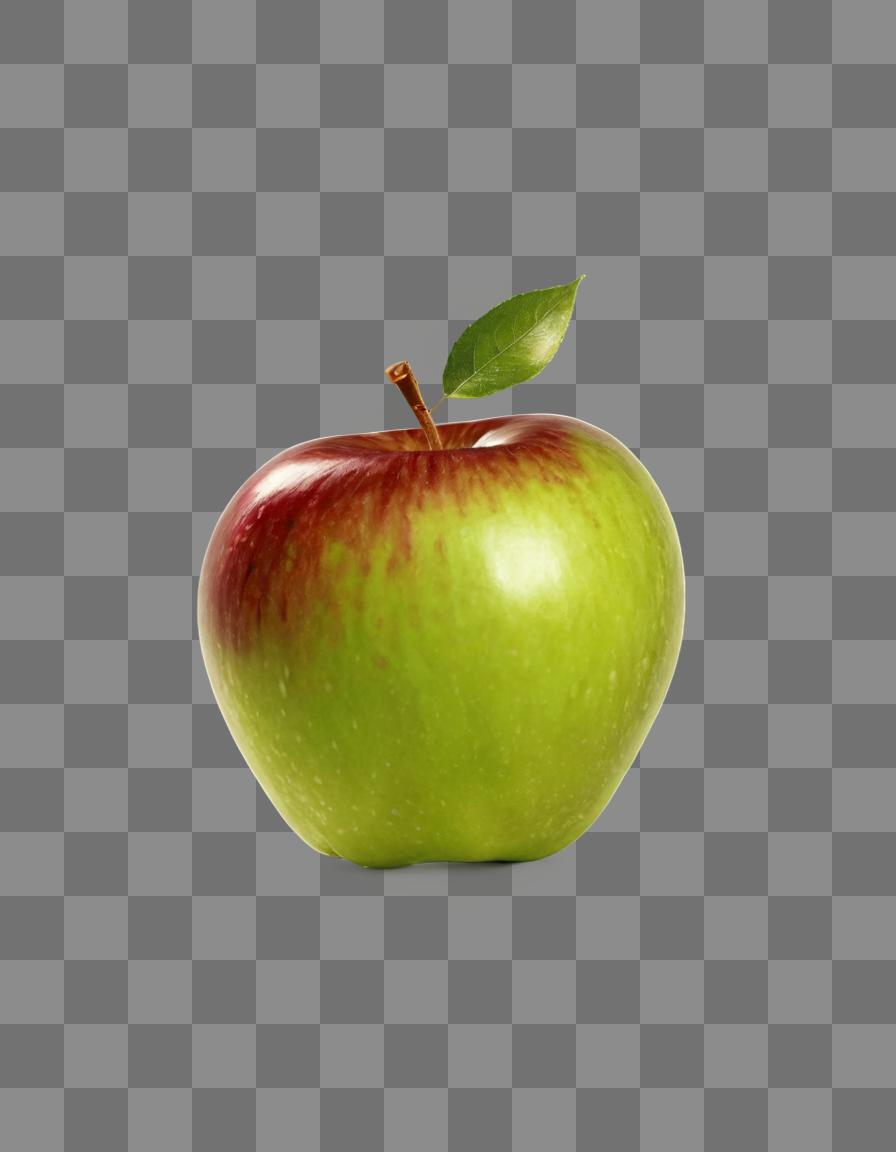}\hfill
\includegraphics[width=0.245\linewidth]{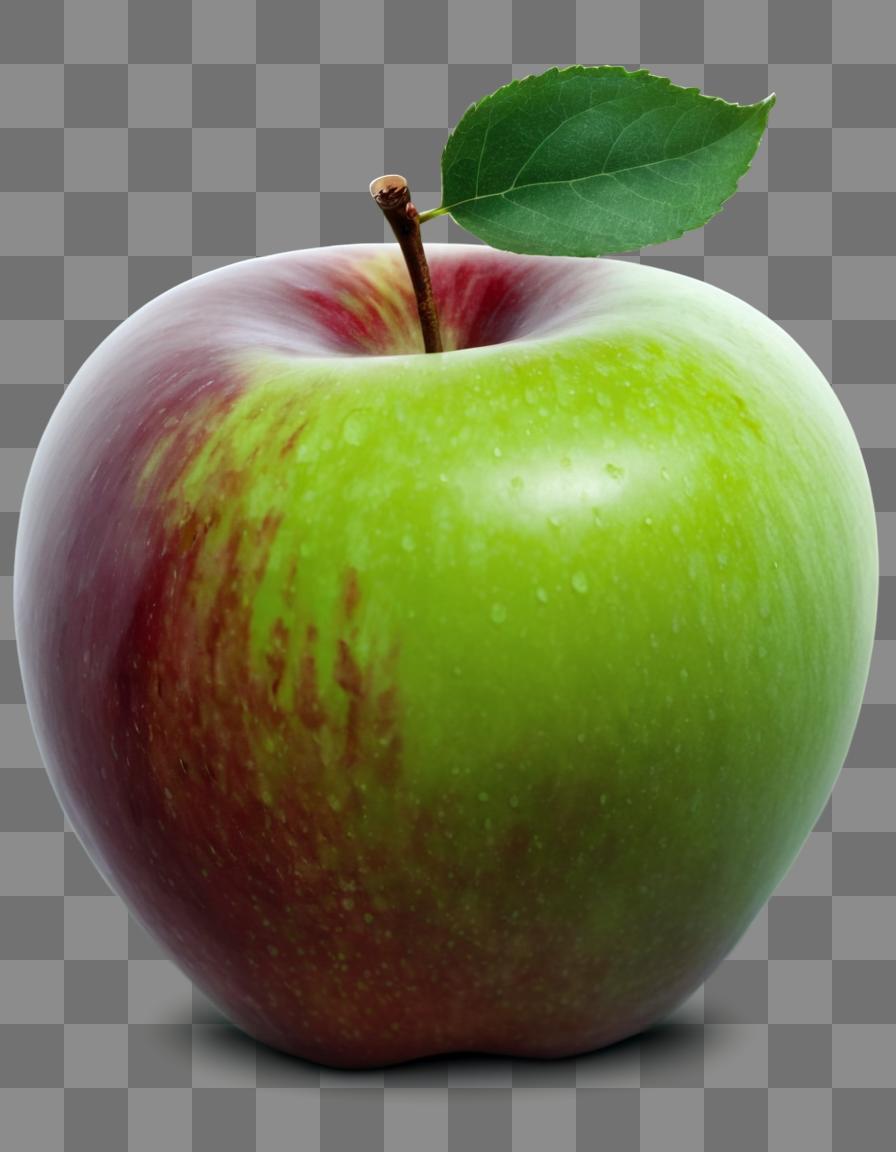}\hfill
\includegraphics[width=0.245\linewidth]{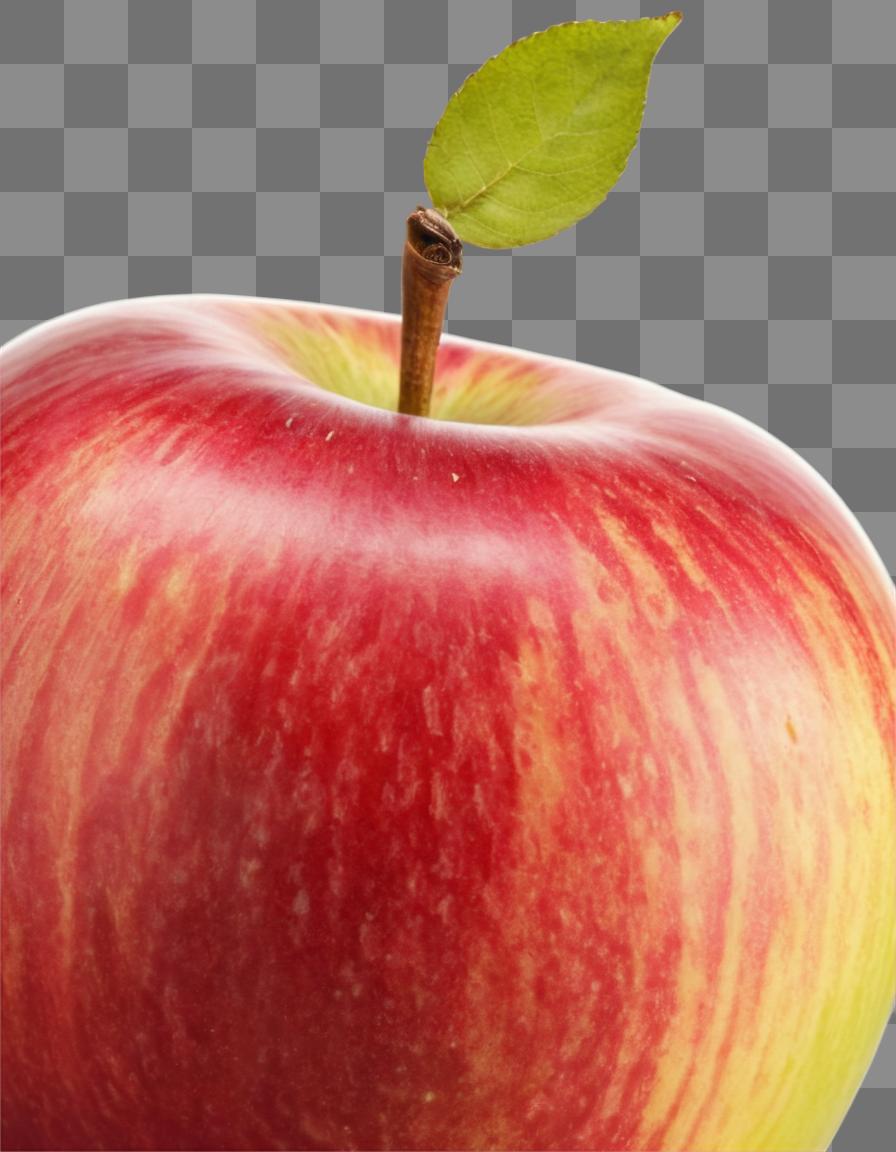}
\caption{Single Transparent Image Results \#1. The prompt is ``apple''. Resolution is $896\times1152$.}
\label{fig:a1}
\end{minipage}
\end{figure*}

\begin{figure*}

\begin{minipage}{\linewidth}
\includegraphics[width=0.245\linewidth]{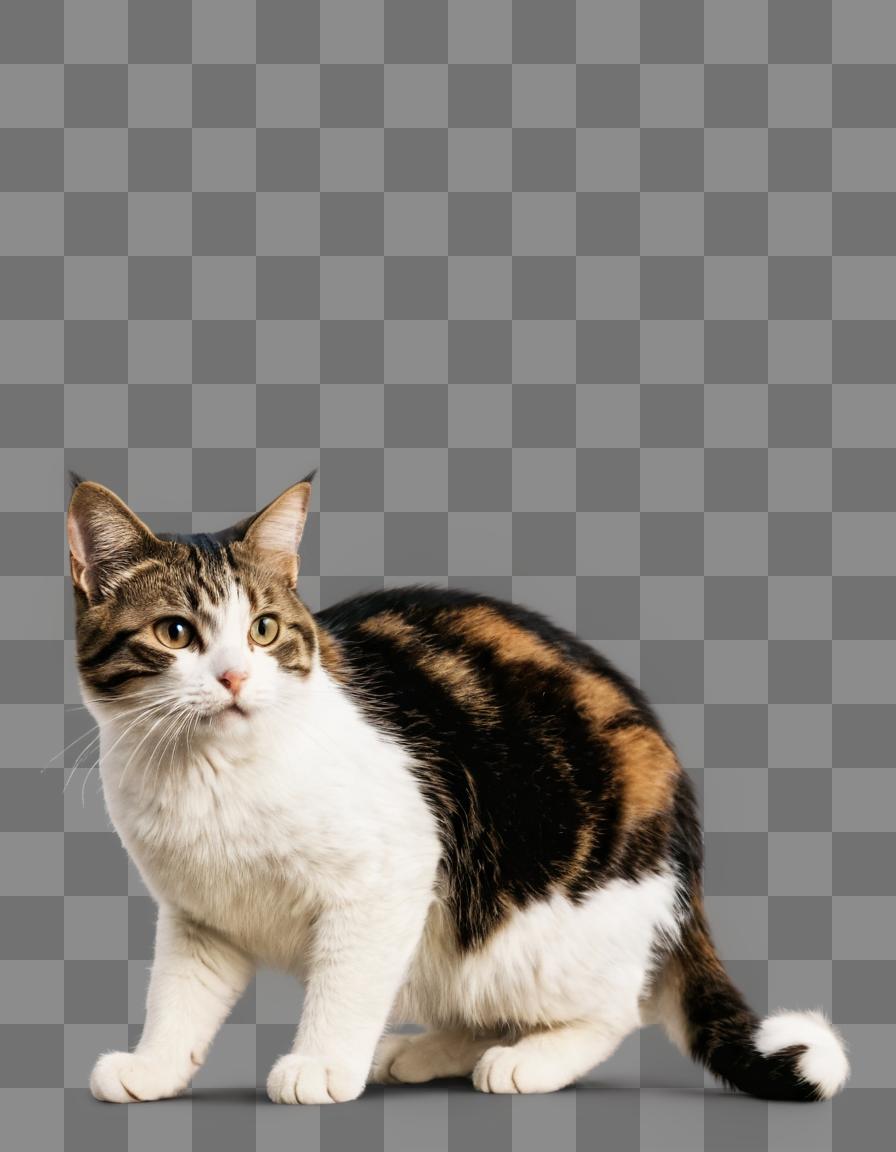}\hfill
\includegraphics[width=0.245\linewidth]{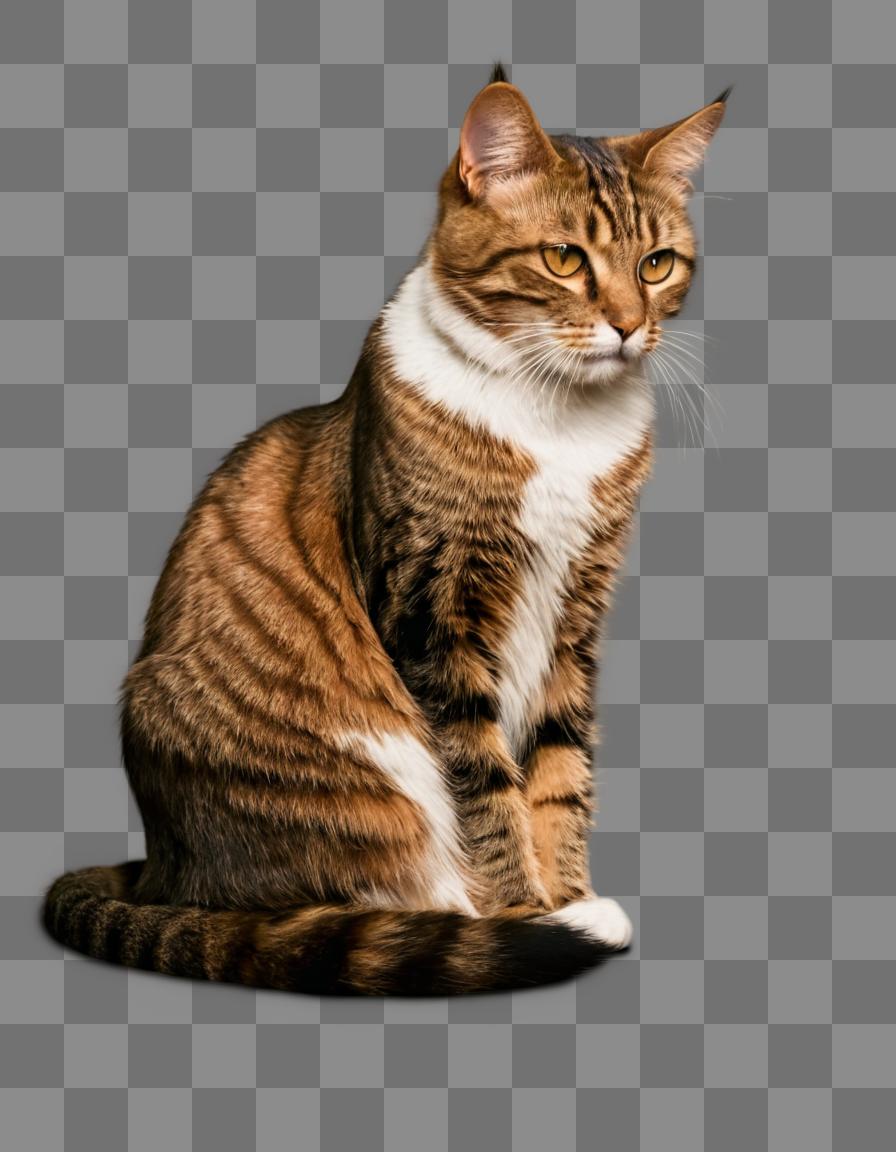}\hfill
\includegraphics[width=0.245\linewidth]{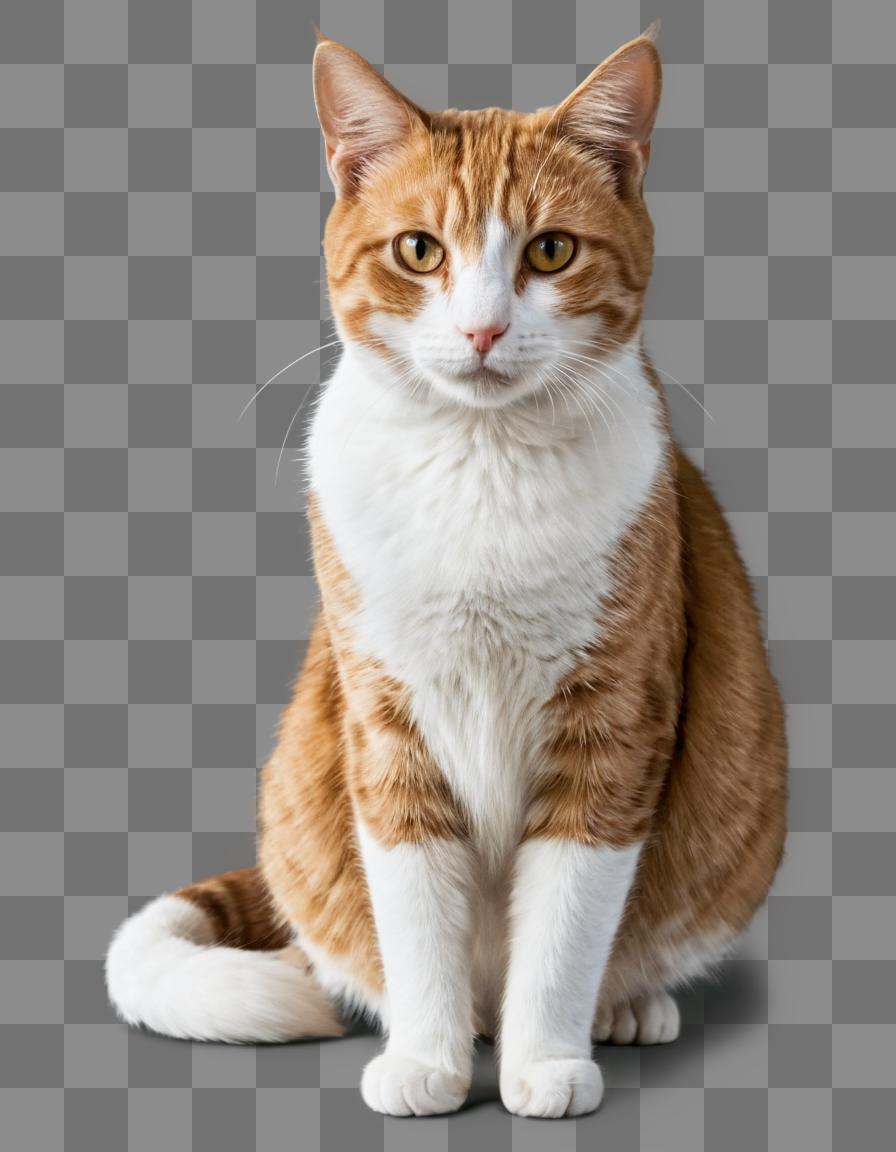}\hfill
\includegraphics[width=0.245\linewidth]{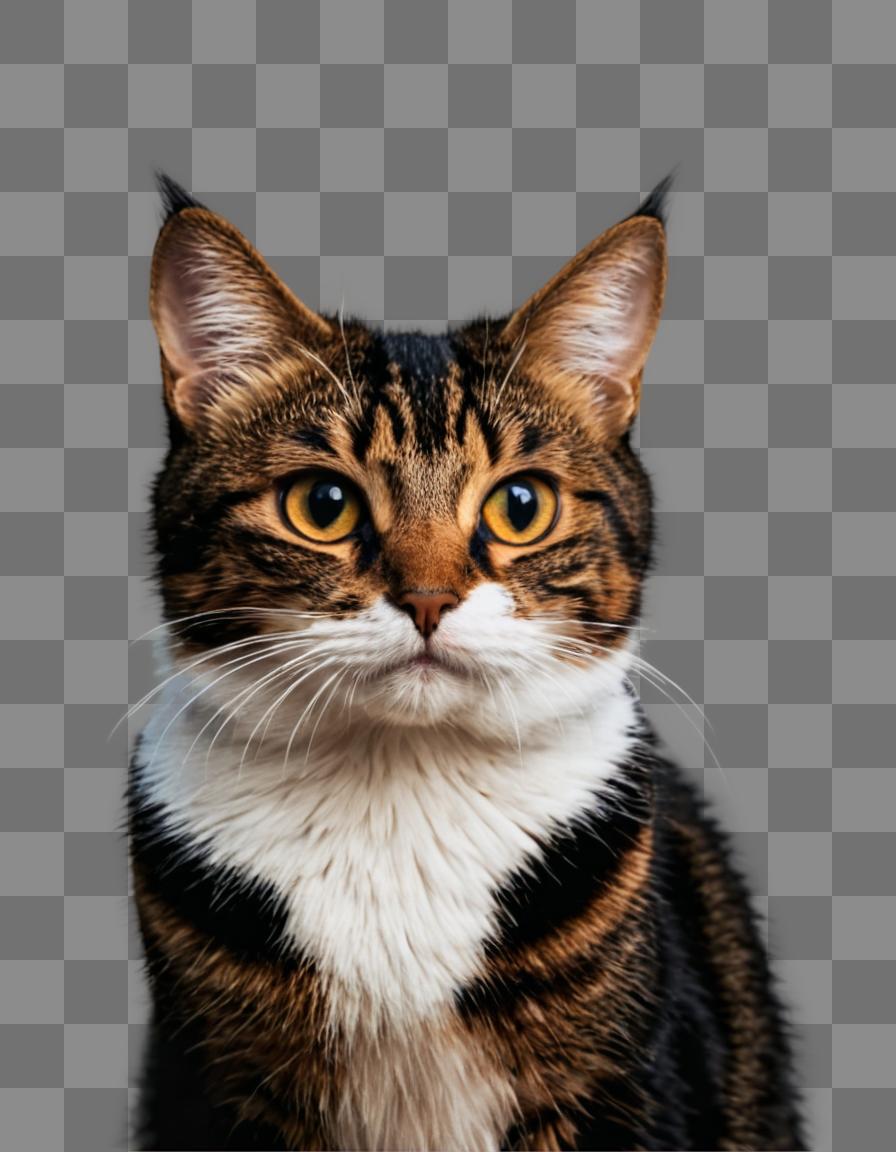}

\vspace{1pt}
\includegraphics[width=0.245\linewidth]{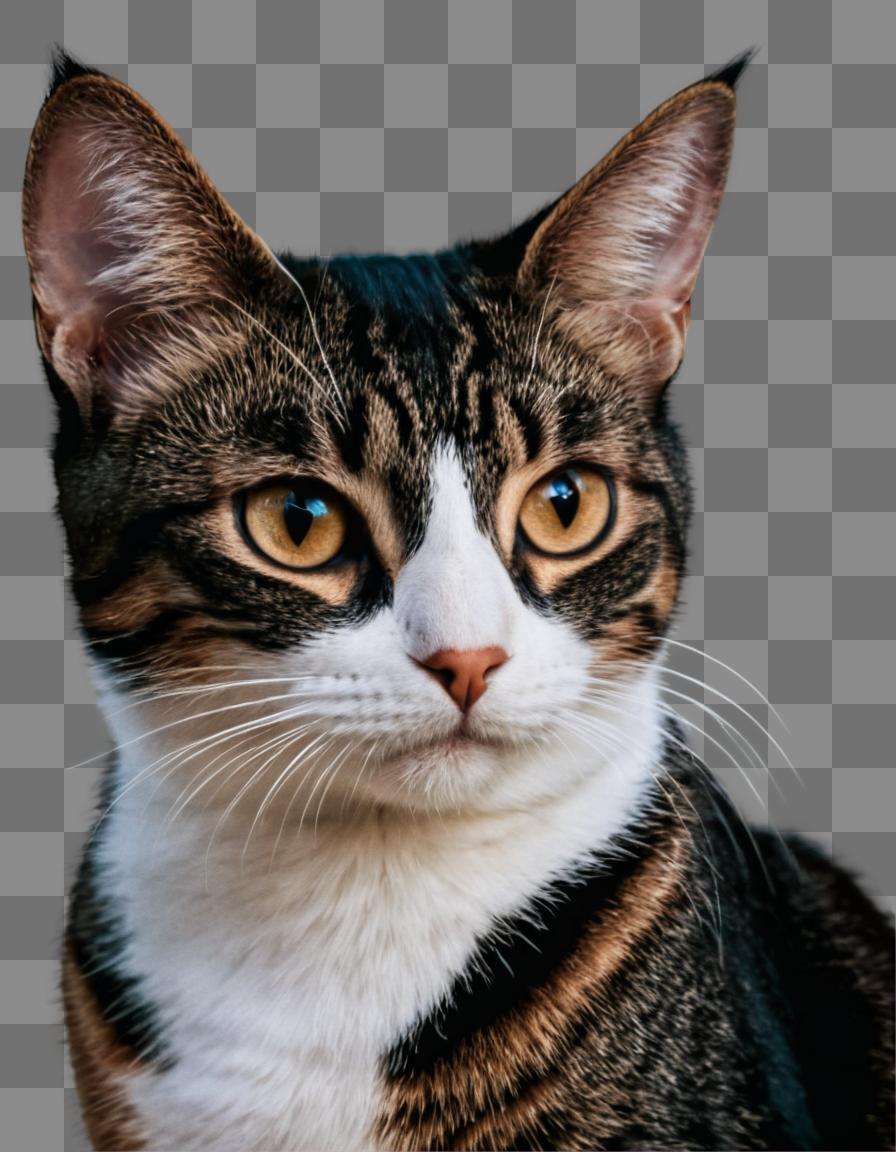}\hfill
\includegraphics[width=0.245\linewidth]{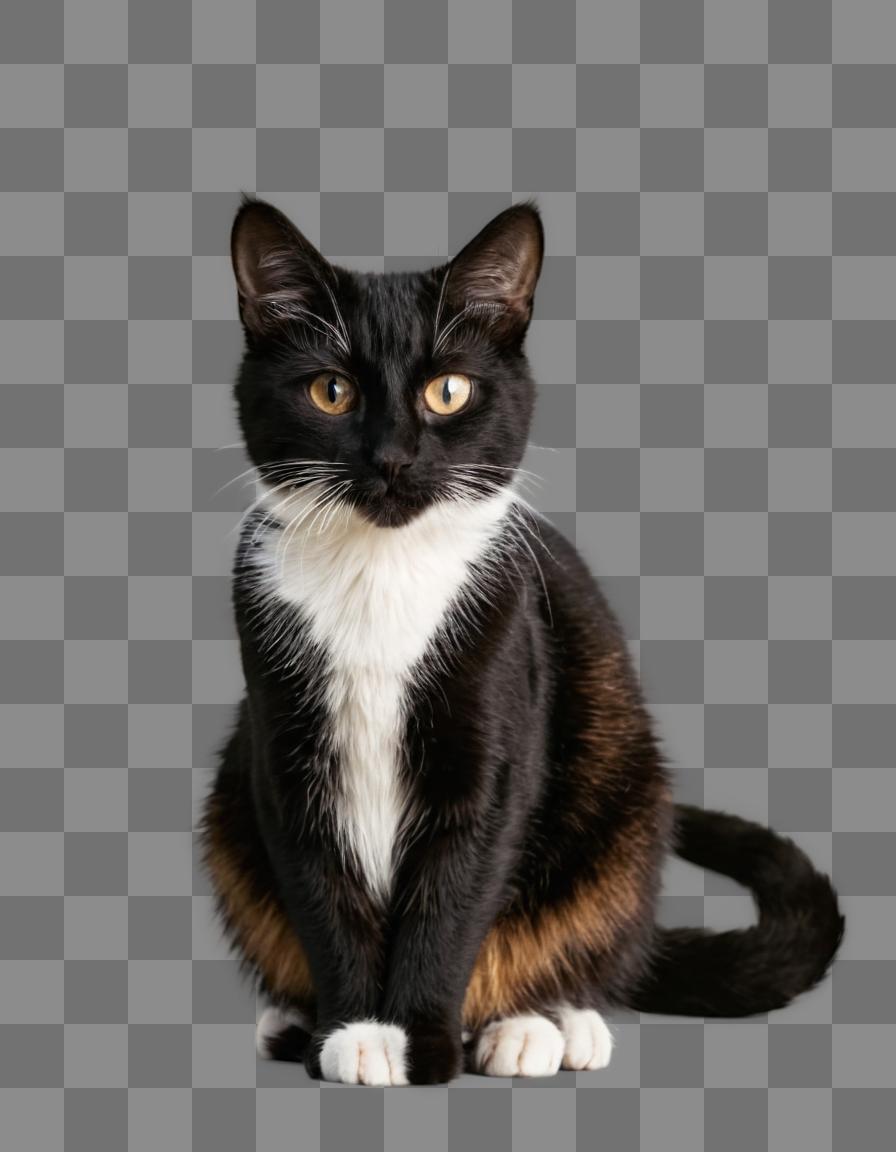}\hfill
\includegraphics[width=0.245\linewidth]{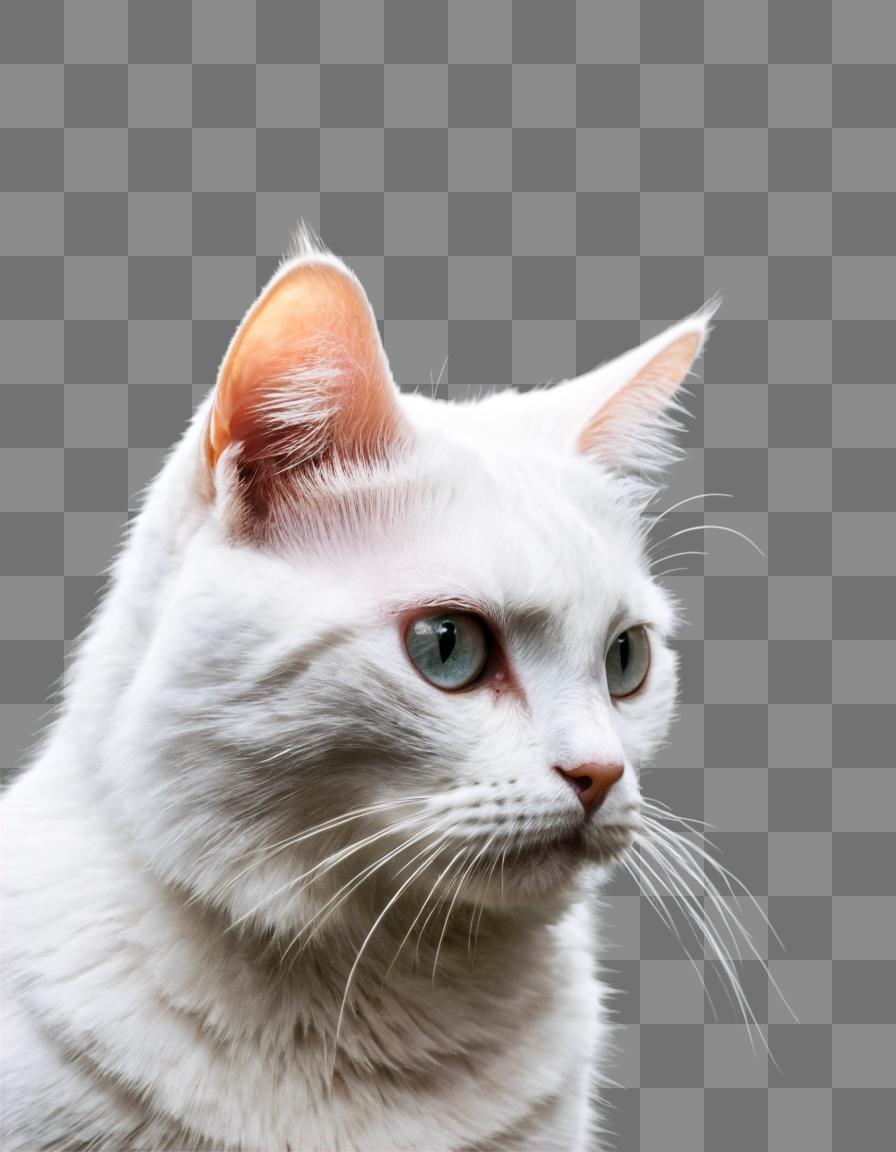}\hfill
\includegraphics[width=0.245\linewidth]{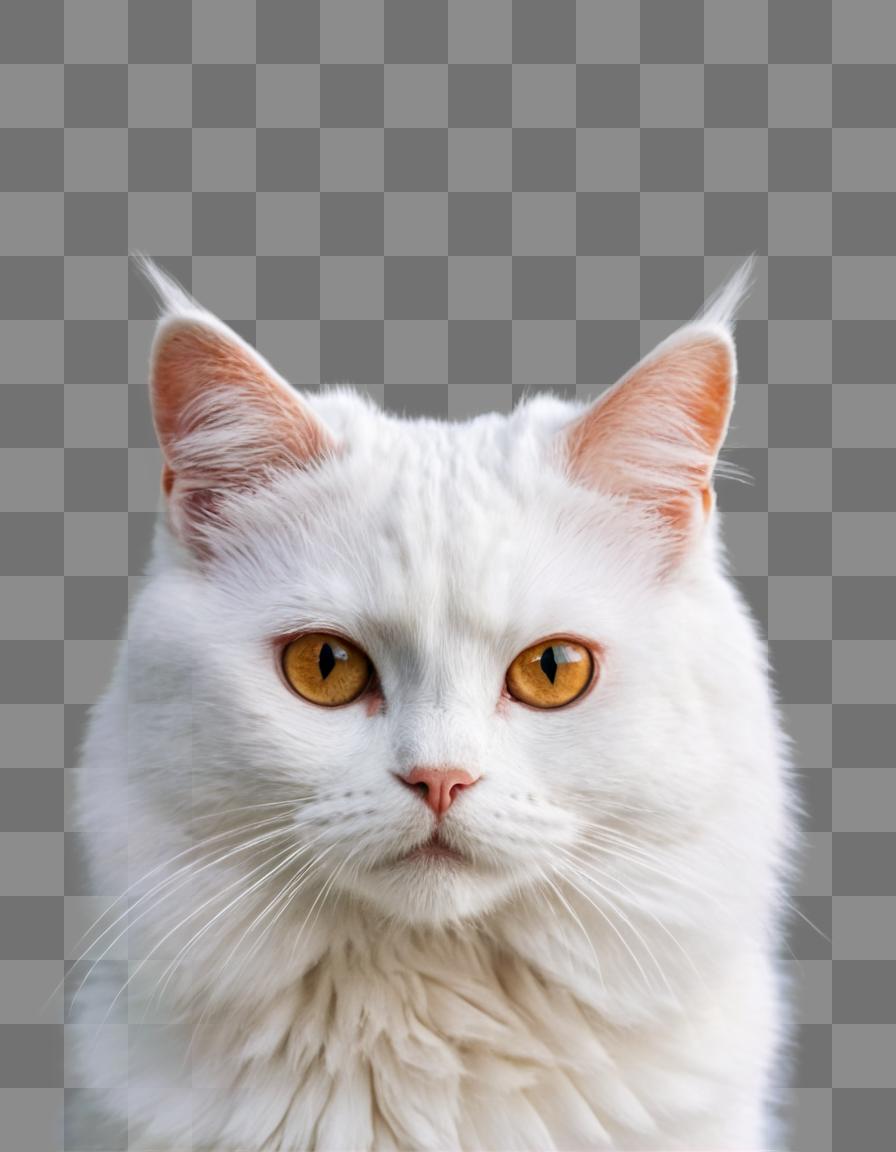}

\vspace{1pt}
\includegraphics[width=0.245\linewidth]{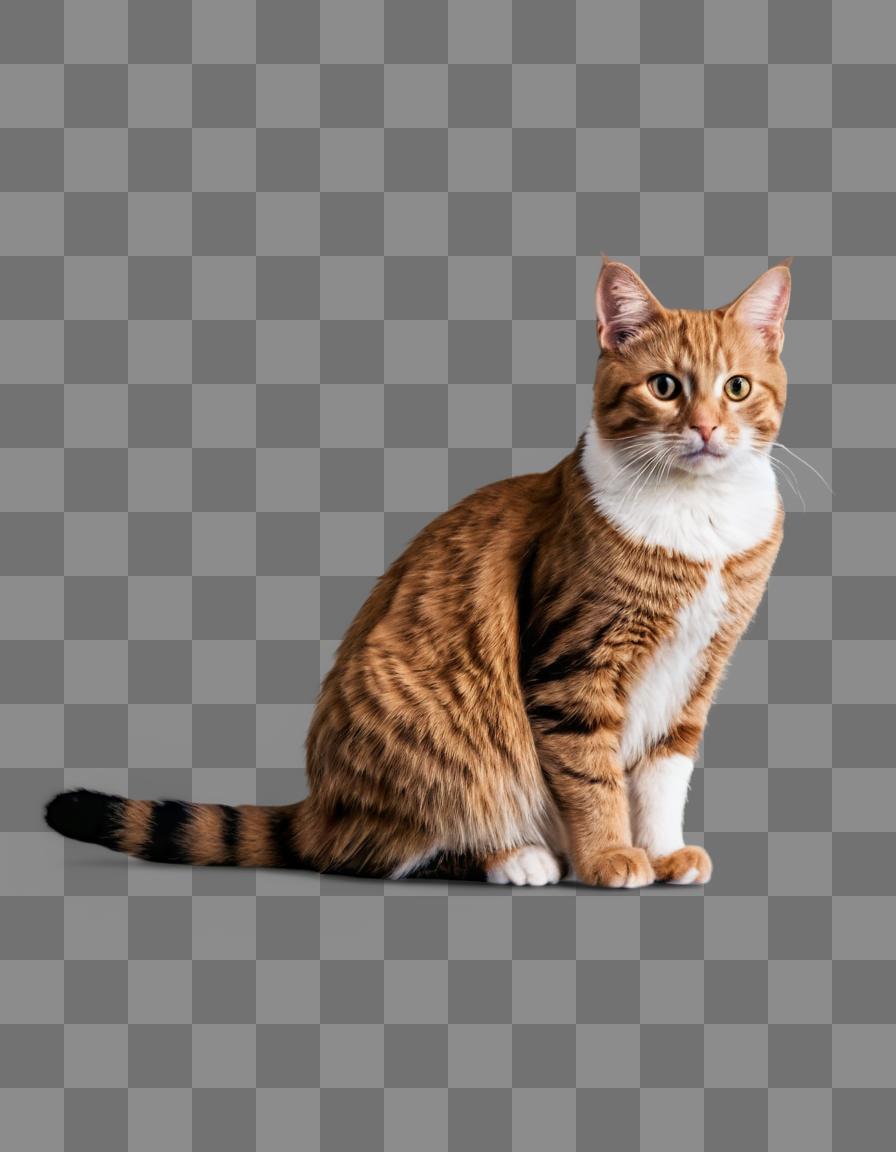}\hfill
\includegraphics[width=0.245\linewidth]{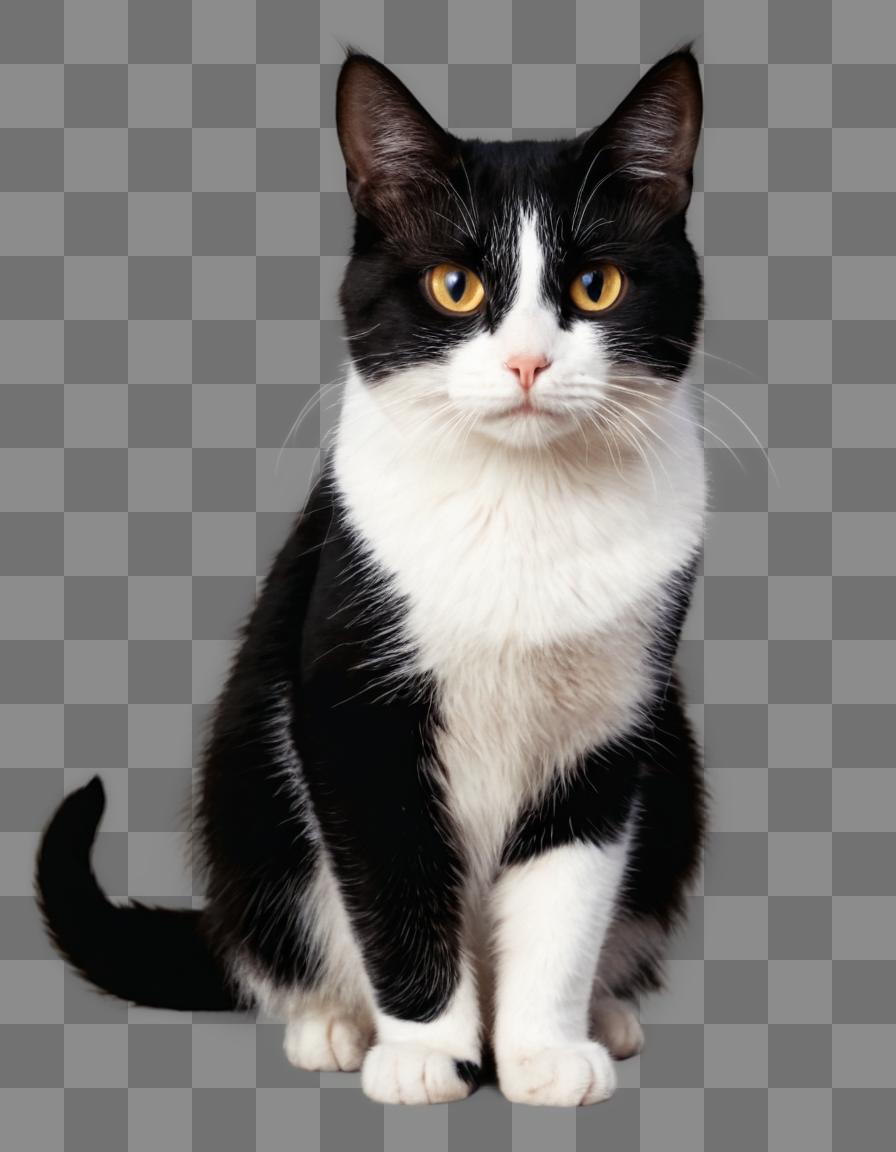}\hfill
\includegraphics[width=0.245\linewidth]{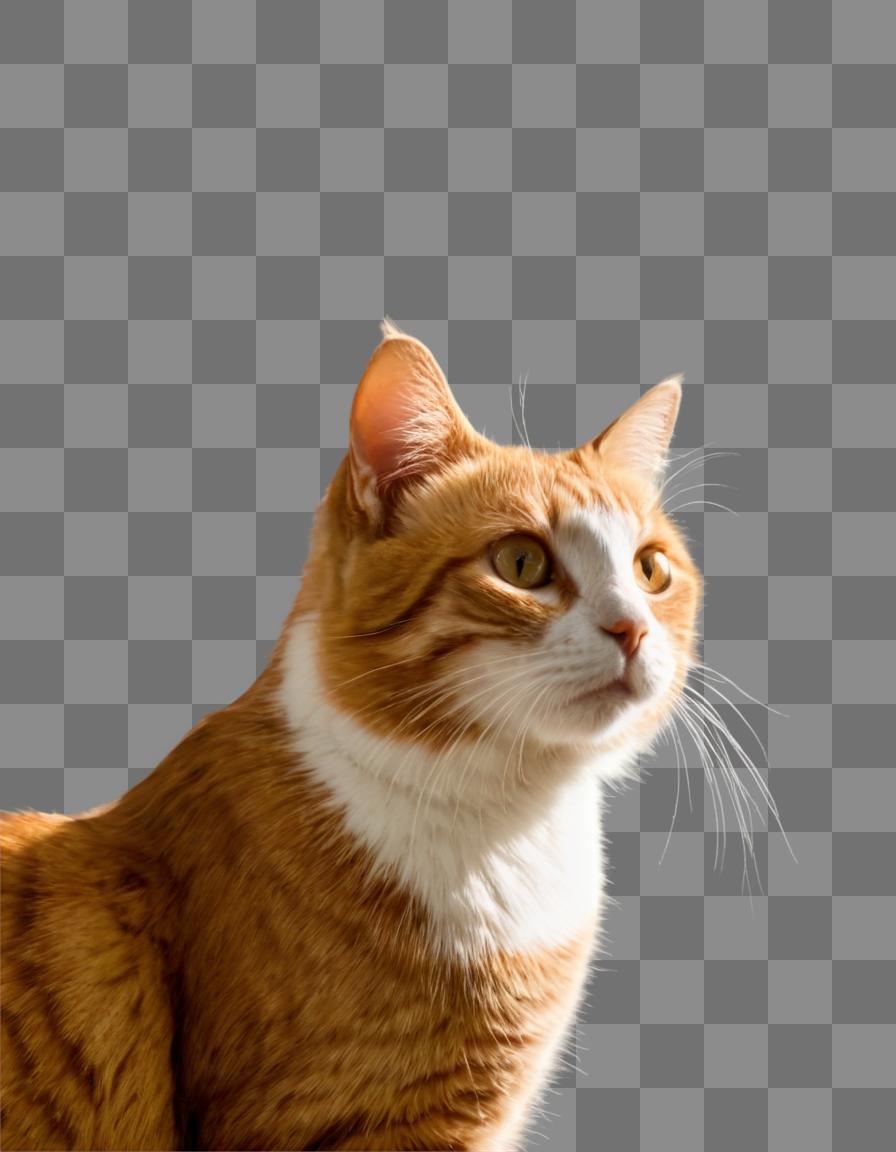}\hfill
\includegraphics[width=0.245\linewidth]{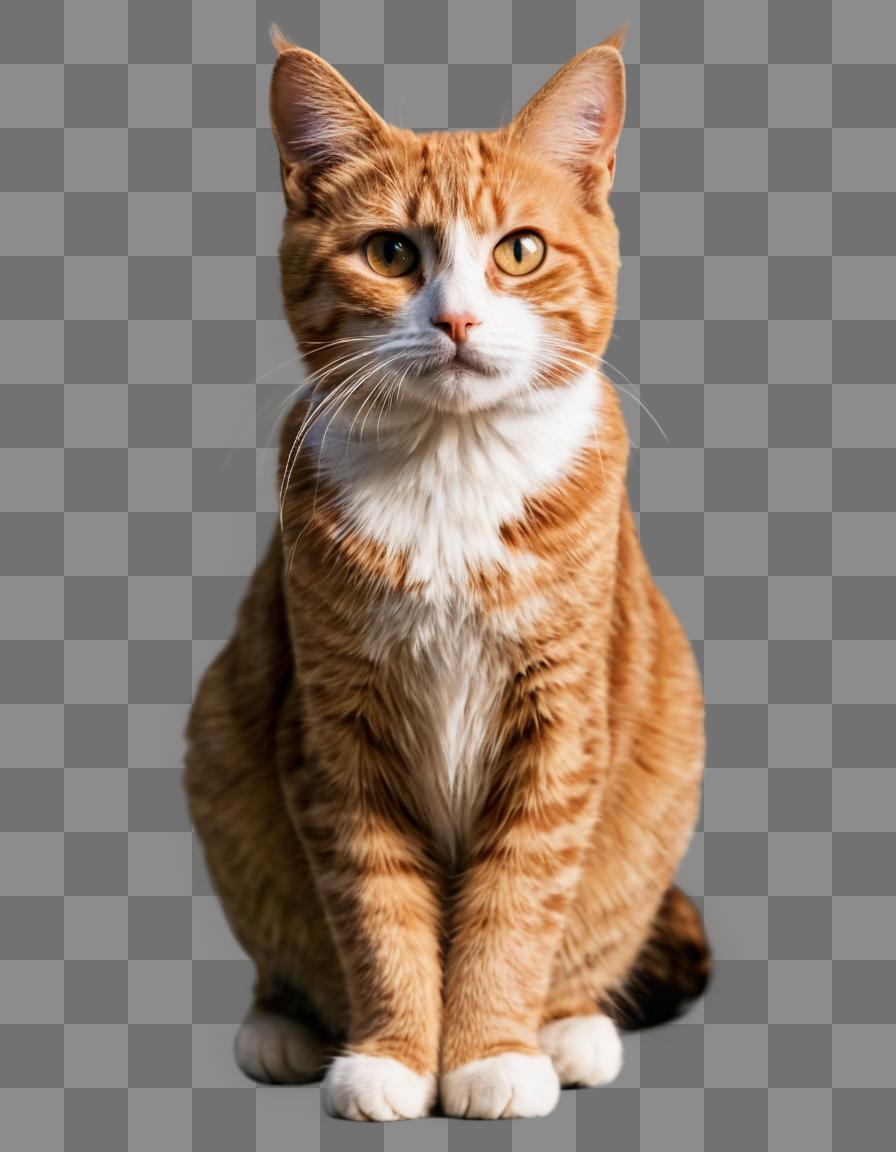}
\caption{Single Transparent Image Results \#2. The prompt is ``a cat''. Resolution is $896\times1152$.}
\label{fig:a2}
\end{minipage}
\end{figure*}

\begin{figure*}

\begin{minipage}{\linewidth}
\includegraphics[width=0.245\linewidth]{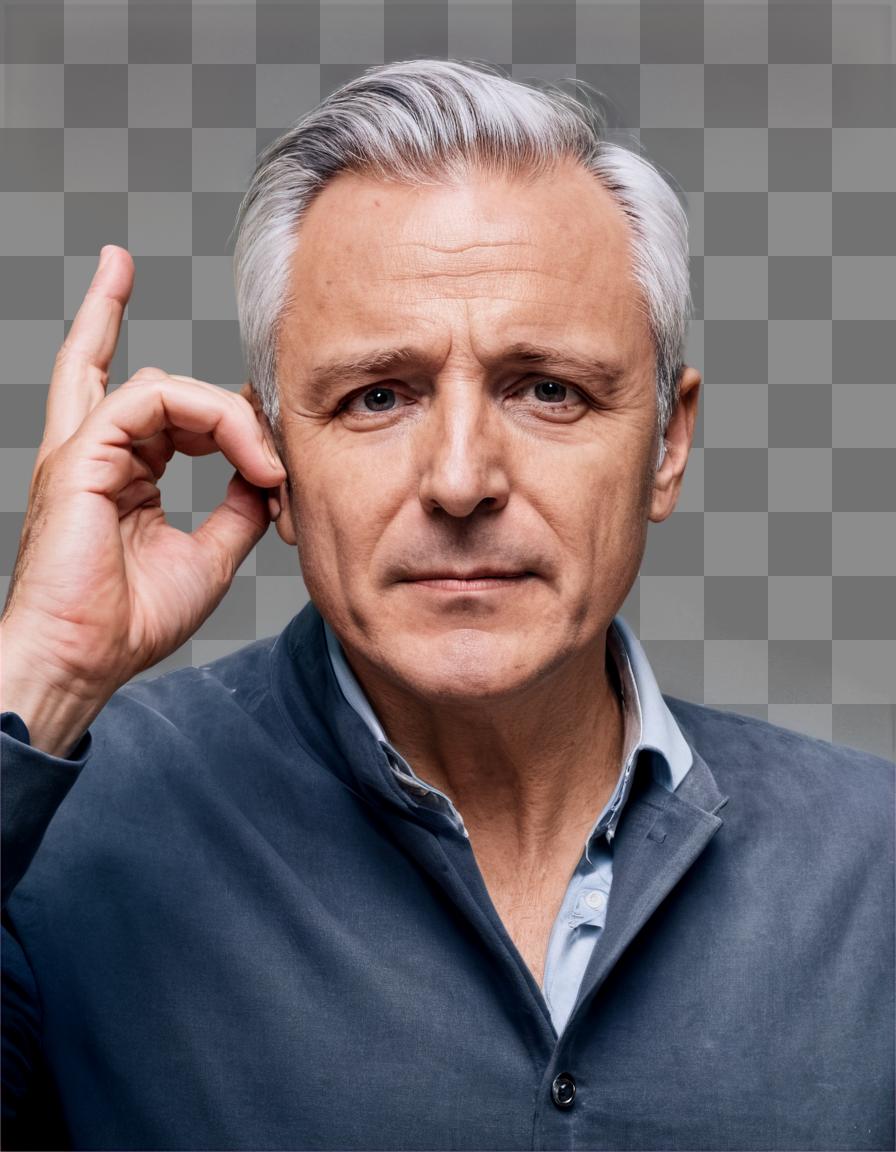}\hfill
\includegraphics[width=0.245\linewidth]{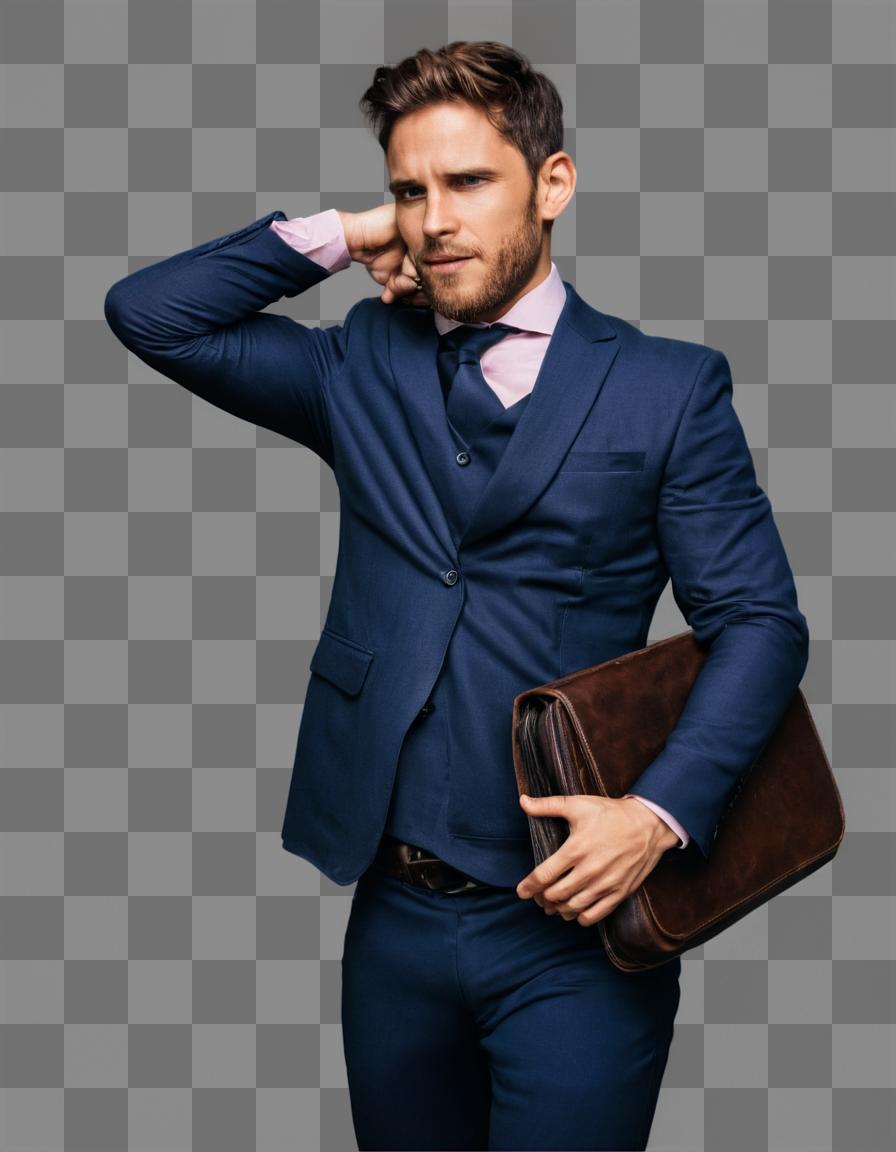}\hfill
\includegraphics[width=0.245\linewidth]{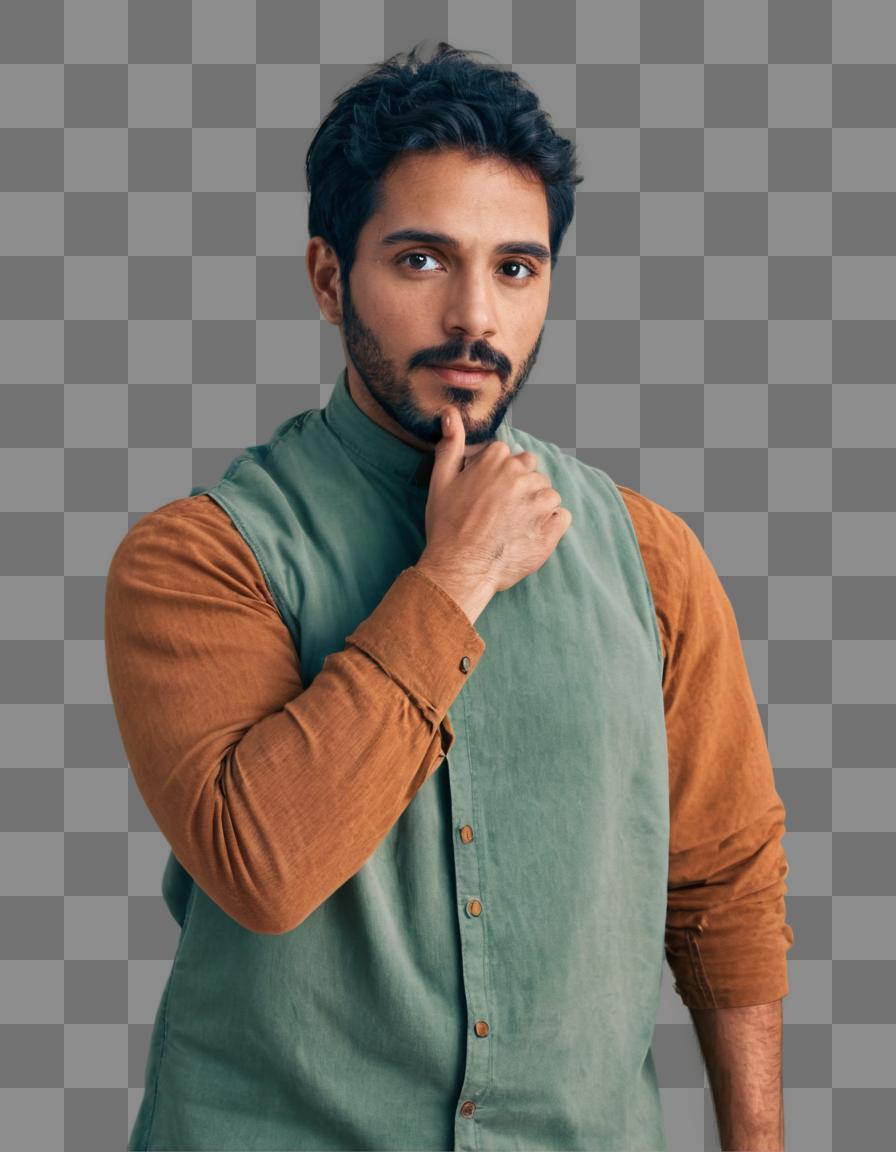}\hfill
\includegraphics[width=0.245\linewidth]{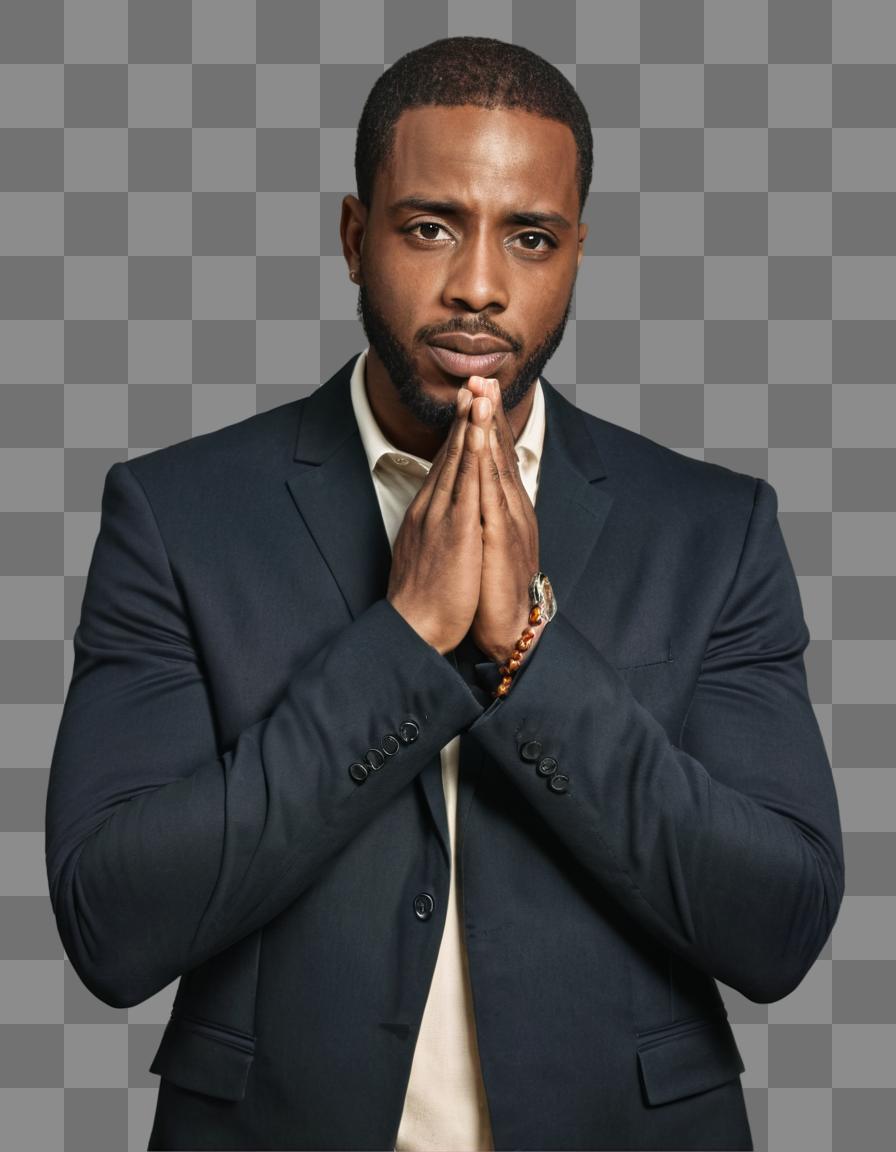}

\vspace{1pt}
\includegraphics[width=0.245\linewidth]{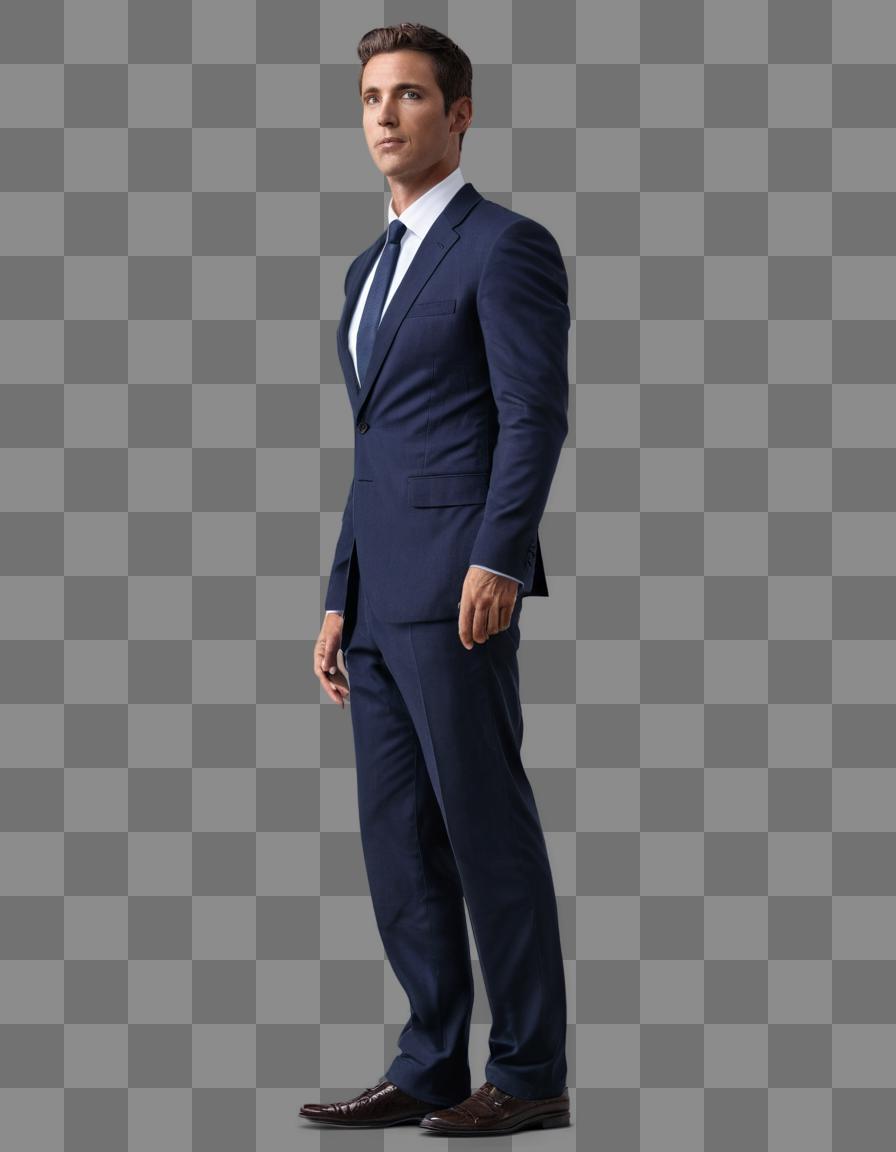}\hfill
\includegraphics[width=0.245\linewidth]{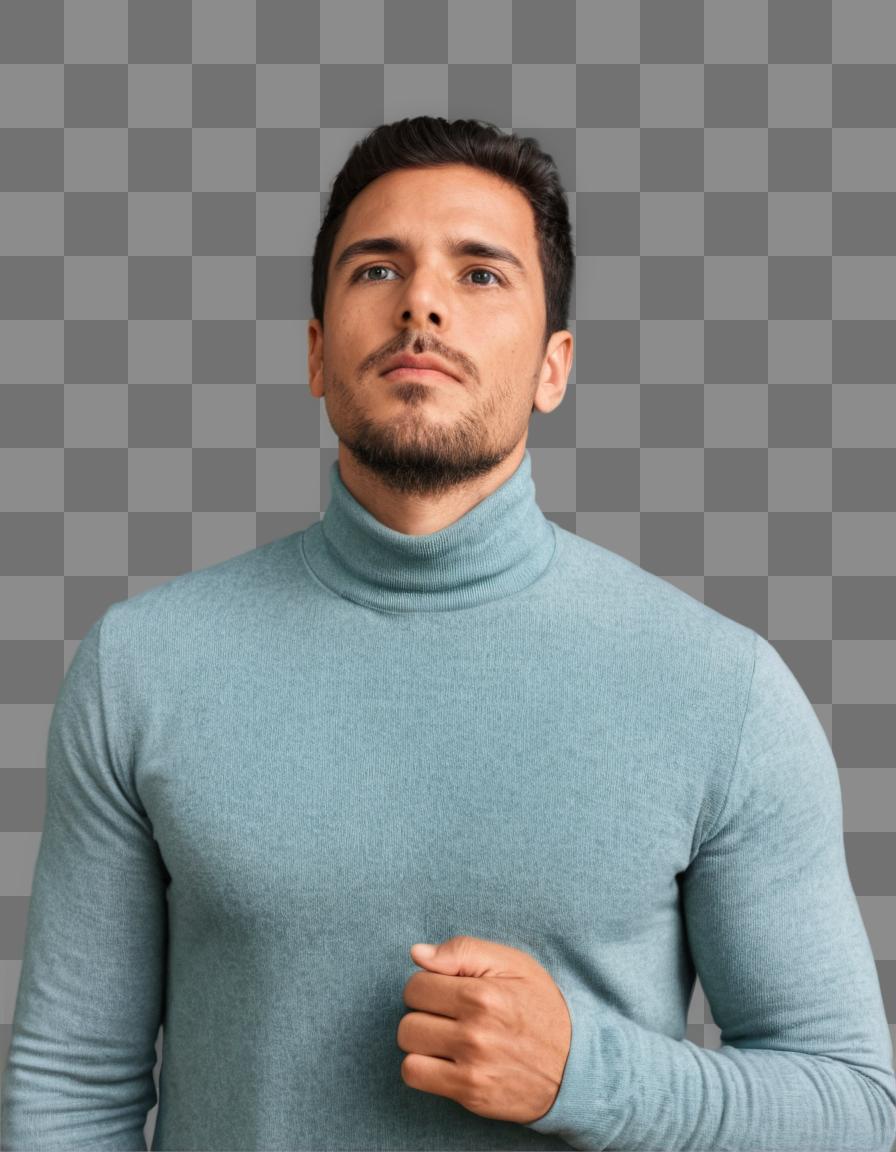}\hfill
\includegraphics[width=0.245\linewidth]{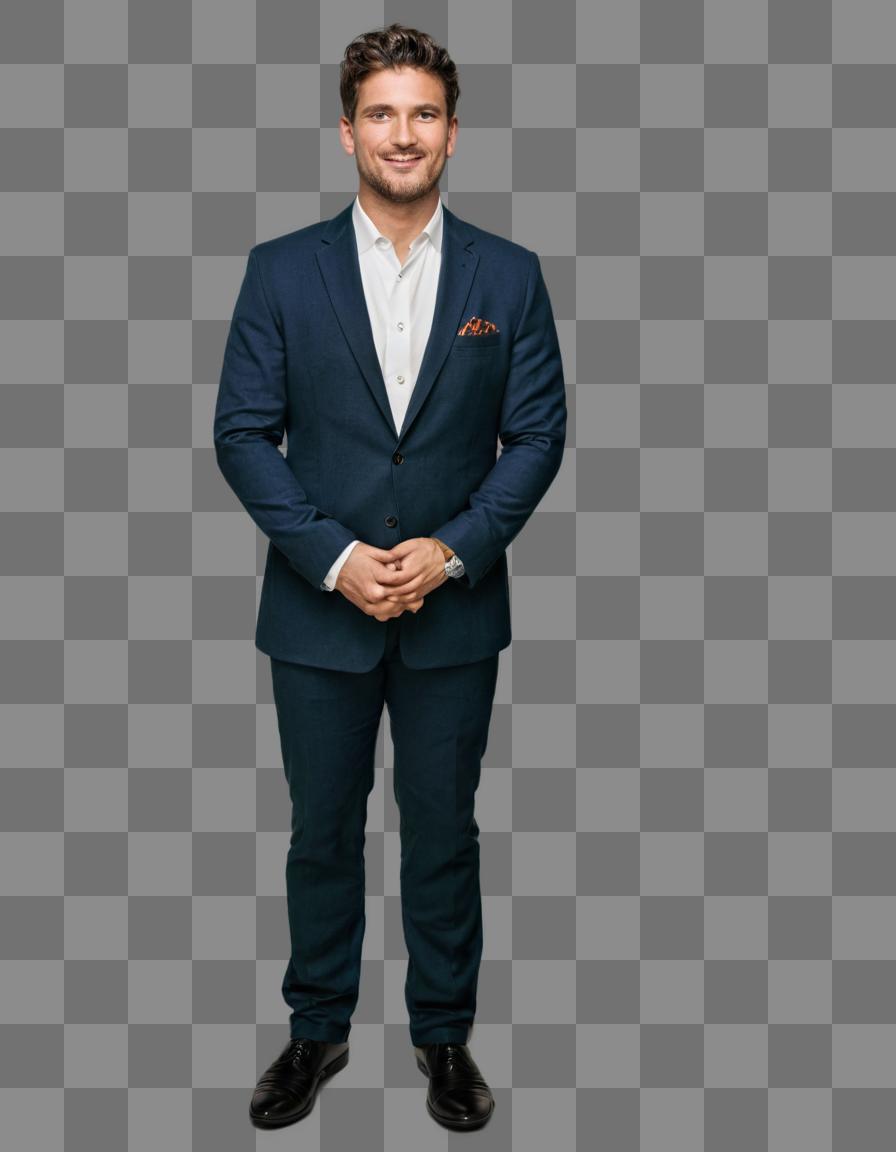}\hfill
\includegraphics[width=0.245\linewidth]{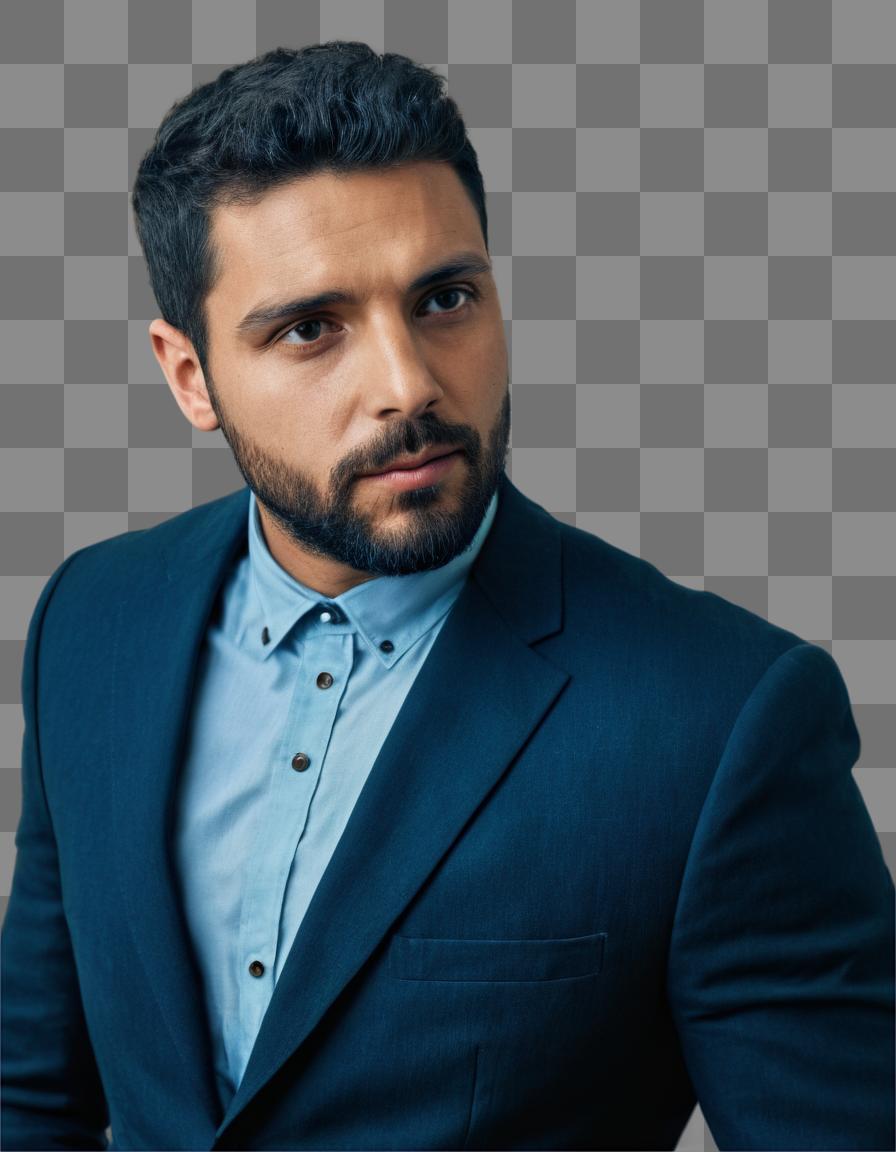}

\vspace{1pt}
\includegraphics[width=0.245\linewidth]{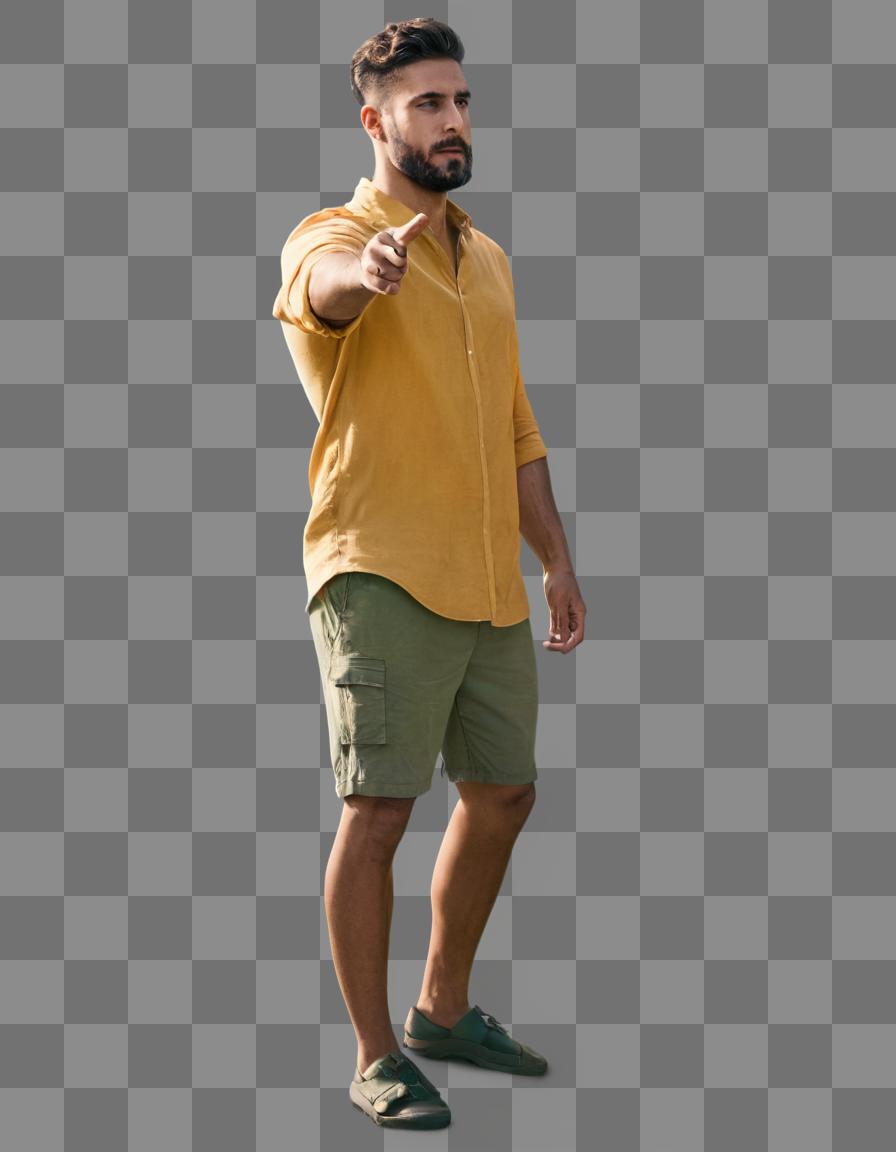}\hfill
\includegraphics[width=0.245\linewidth]{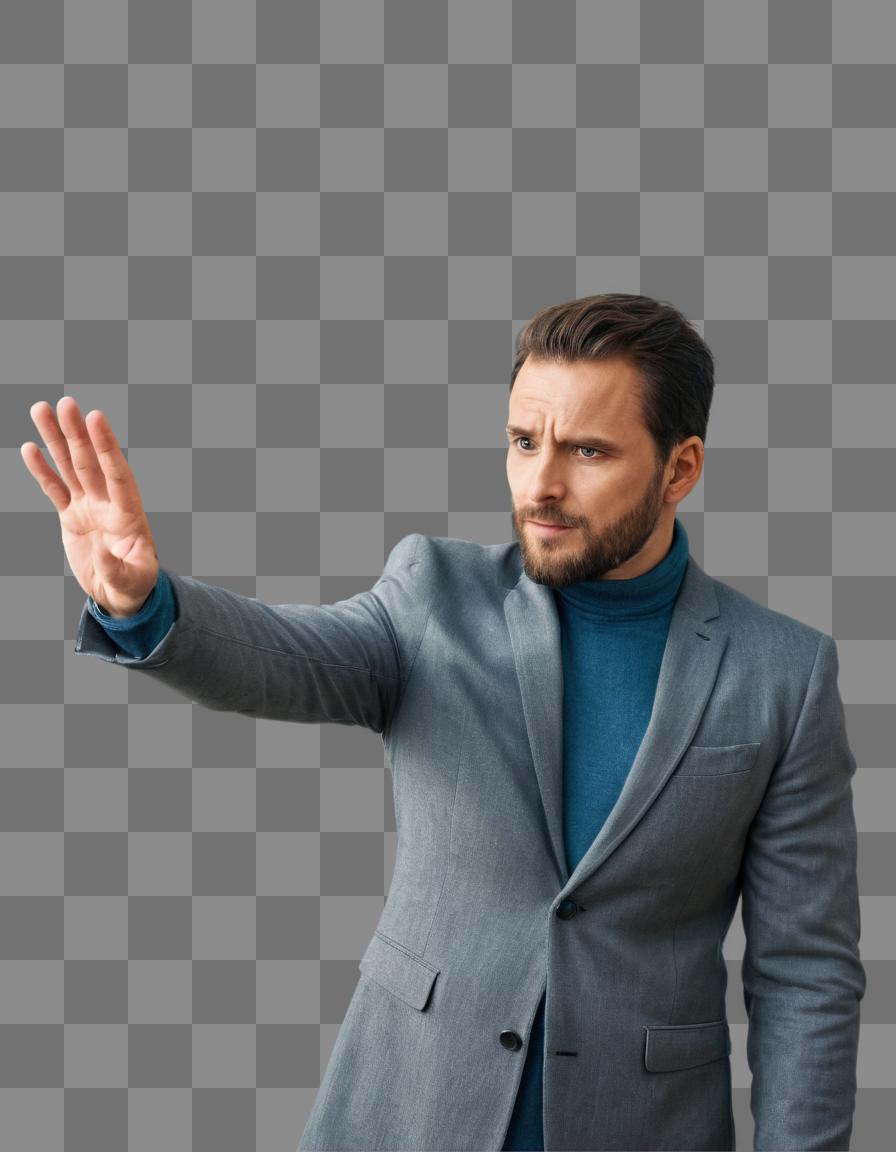}\hfill
\includegraphics[width=0.245\linewidth]{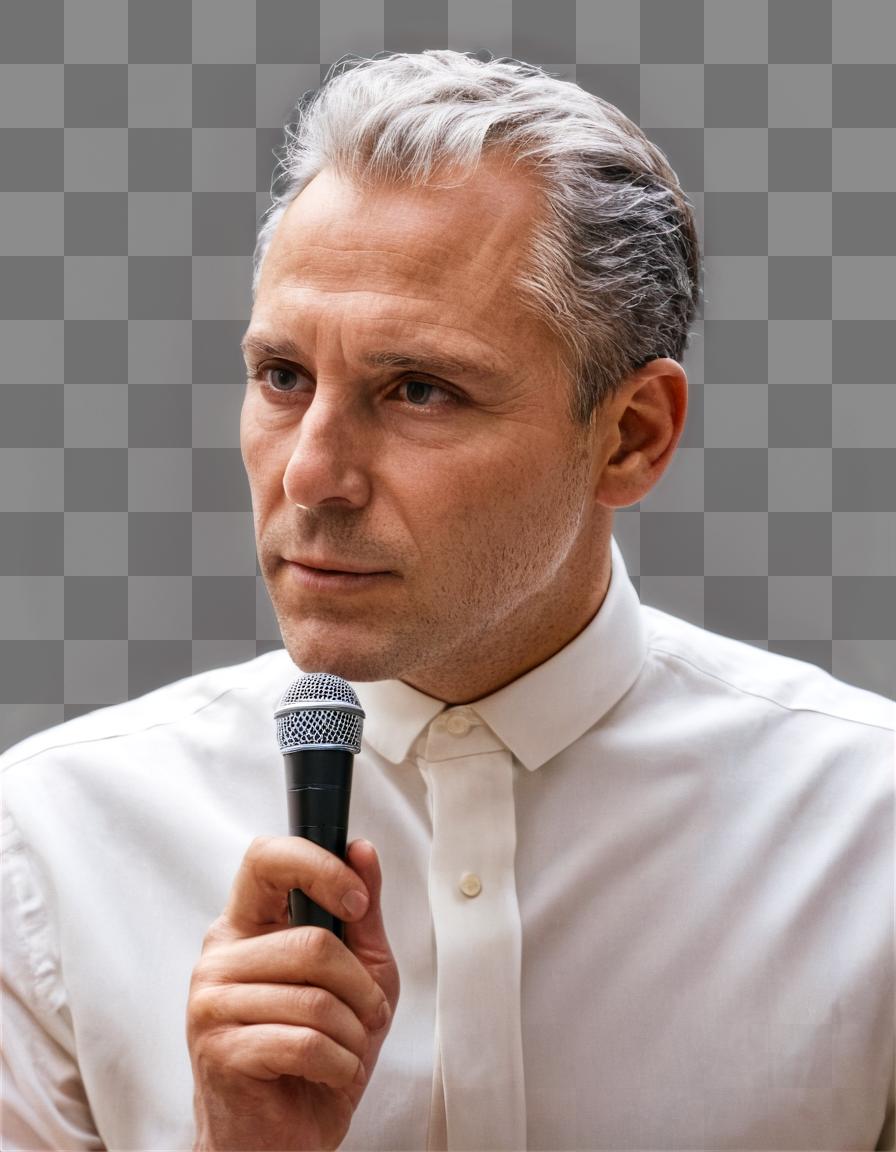}\hfill
\includegraphics[width=0.245\linewidth]{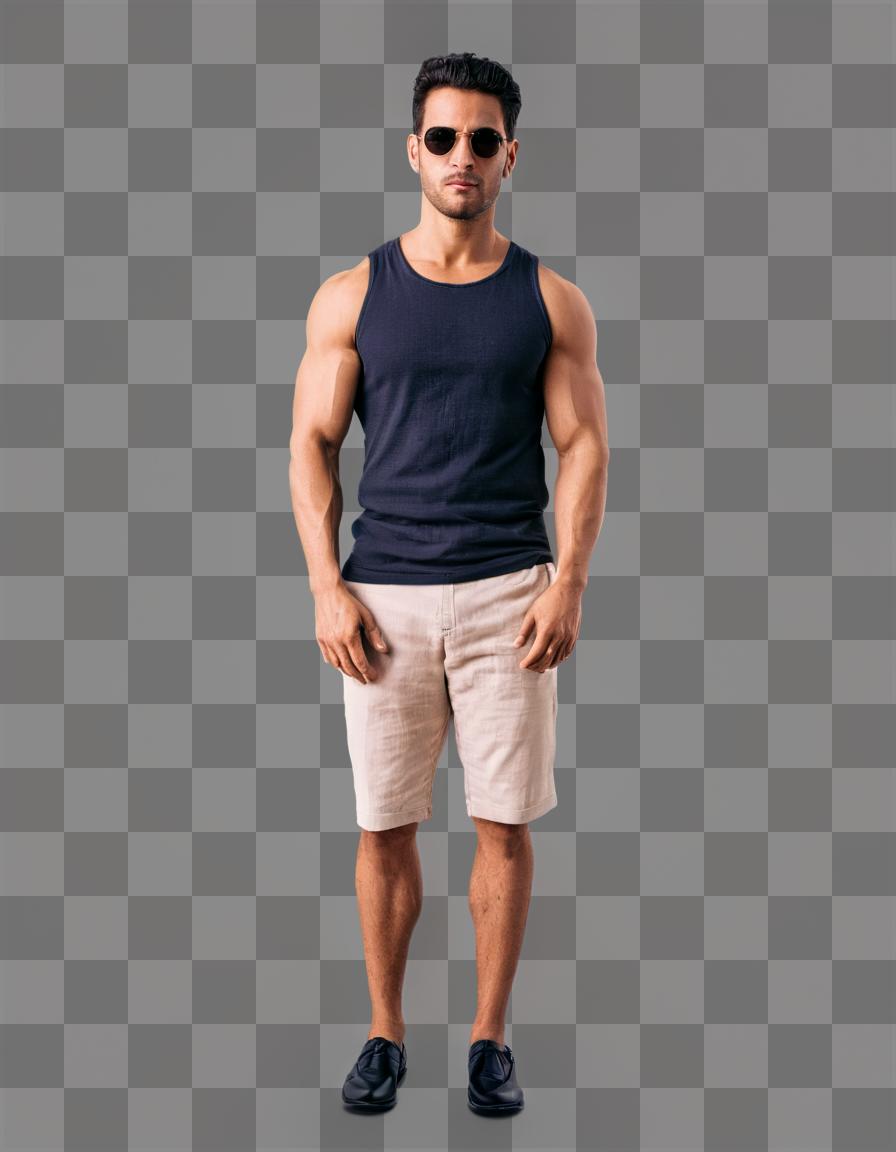}
\caption{Single Transparent Image Results \#3. The prompt is ``a man''. Resolution is $896\times1152$.}
\label{fig:a3}
\end{minipage}
\end{figure*}

\begin{figure*}

\begin{minipage}{\linewidth}
\includegraphics[width=0.245\linewidth]{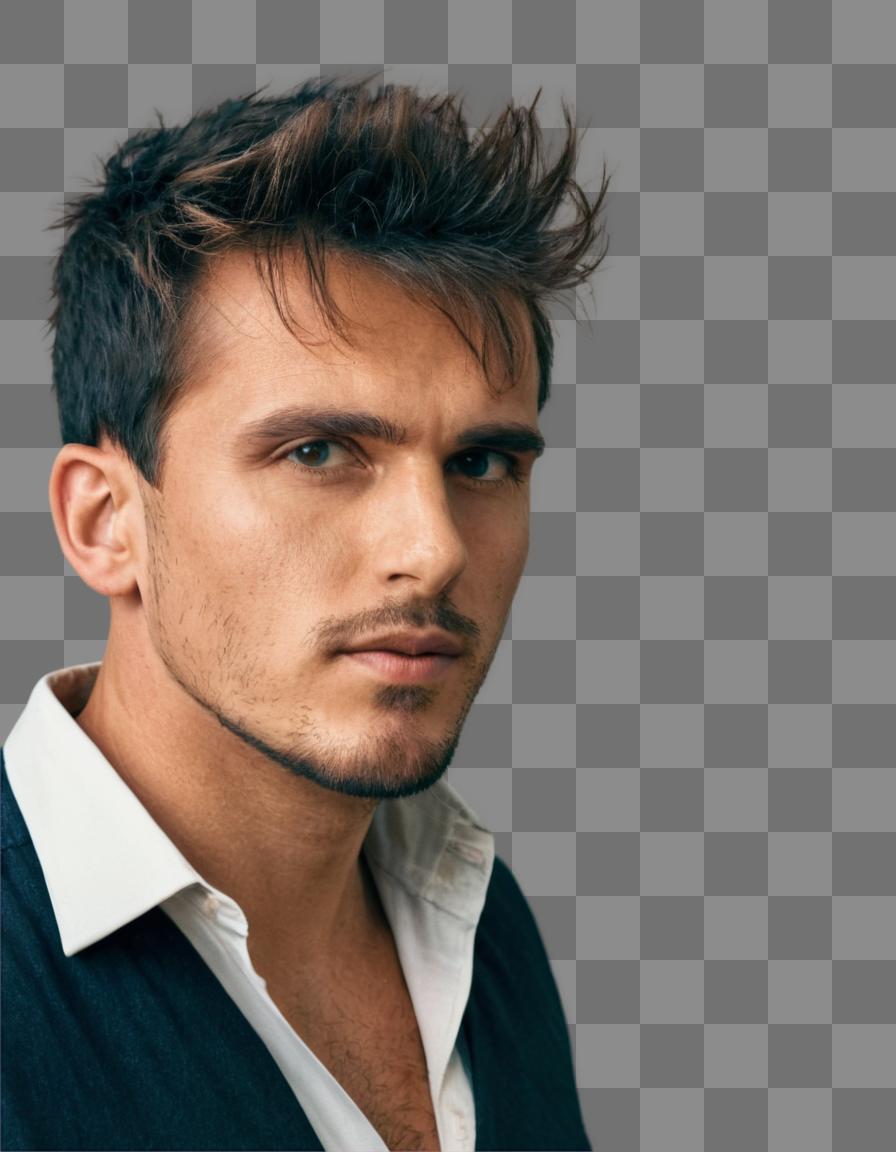}\hfill
\includegraphics[width=0.245\linewidth]{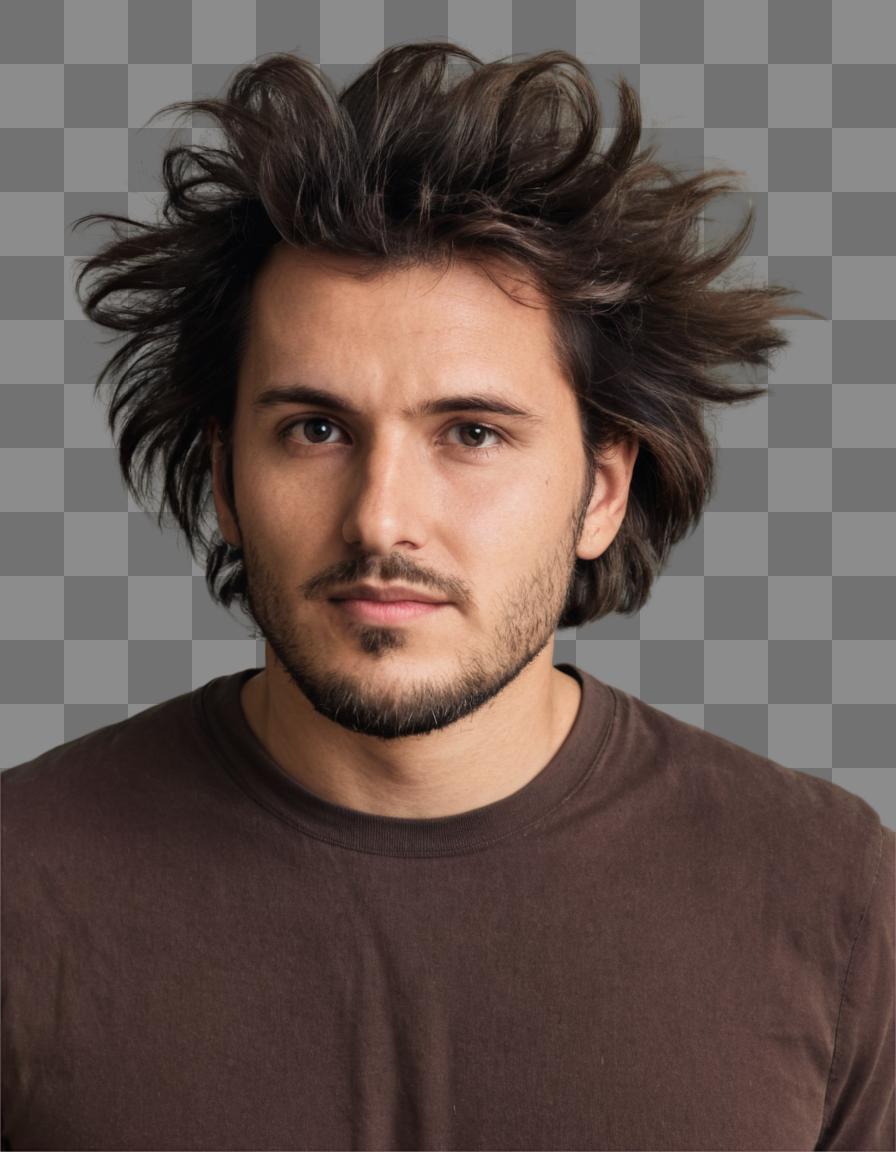}\hfill
\includegraphics[width=0.245\linewidth]{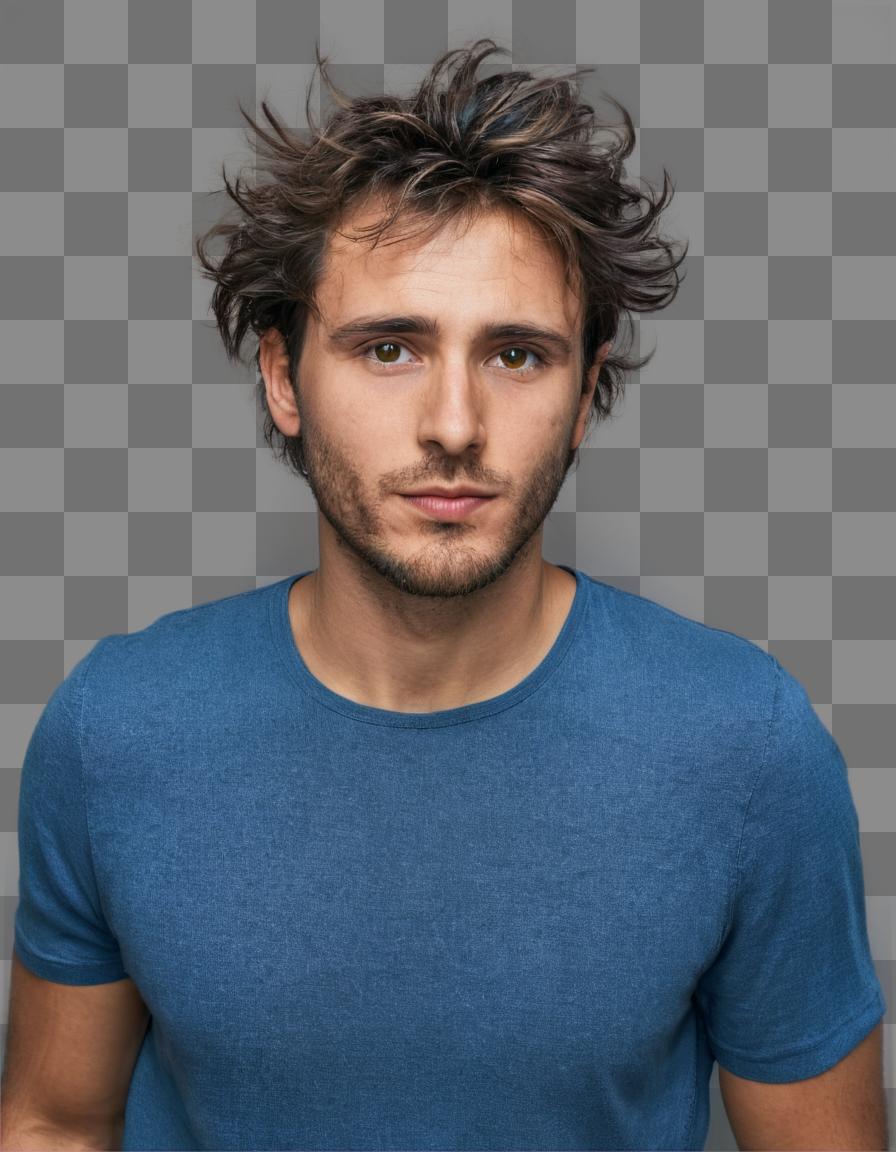}\hfill
\includegraphics[width=0.245\linewidth]{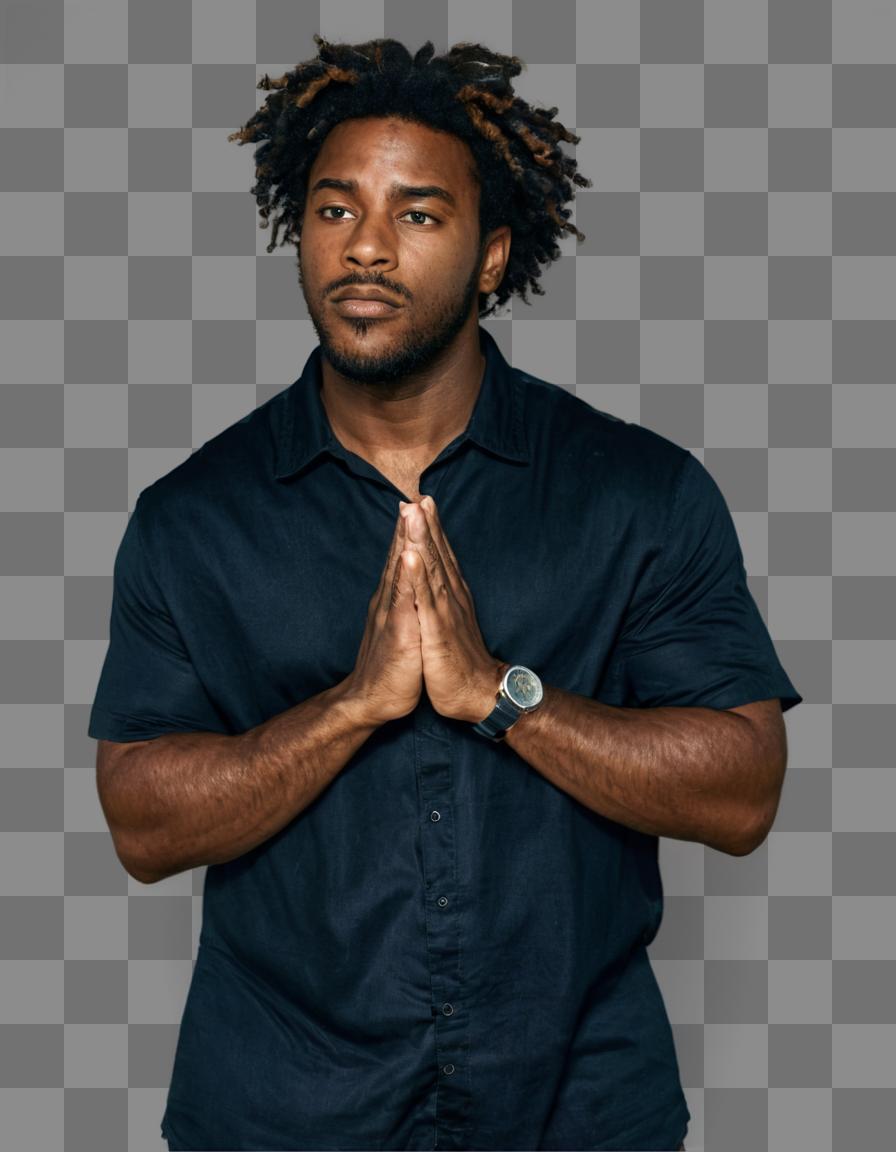}

\vspace{1pt}
\includegraphics[width=0.245\linewidth]{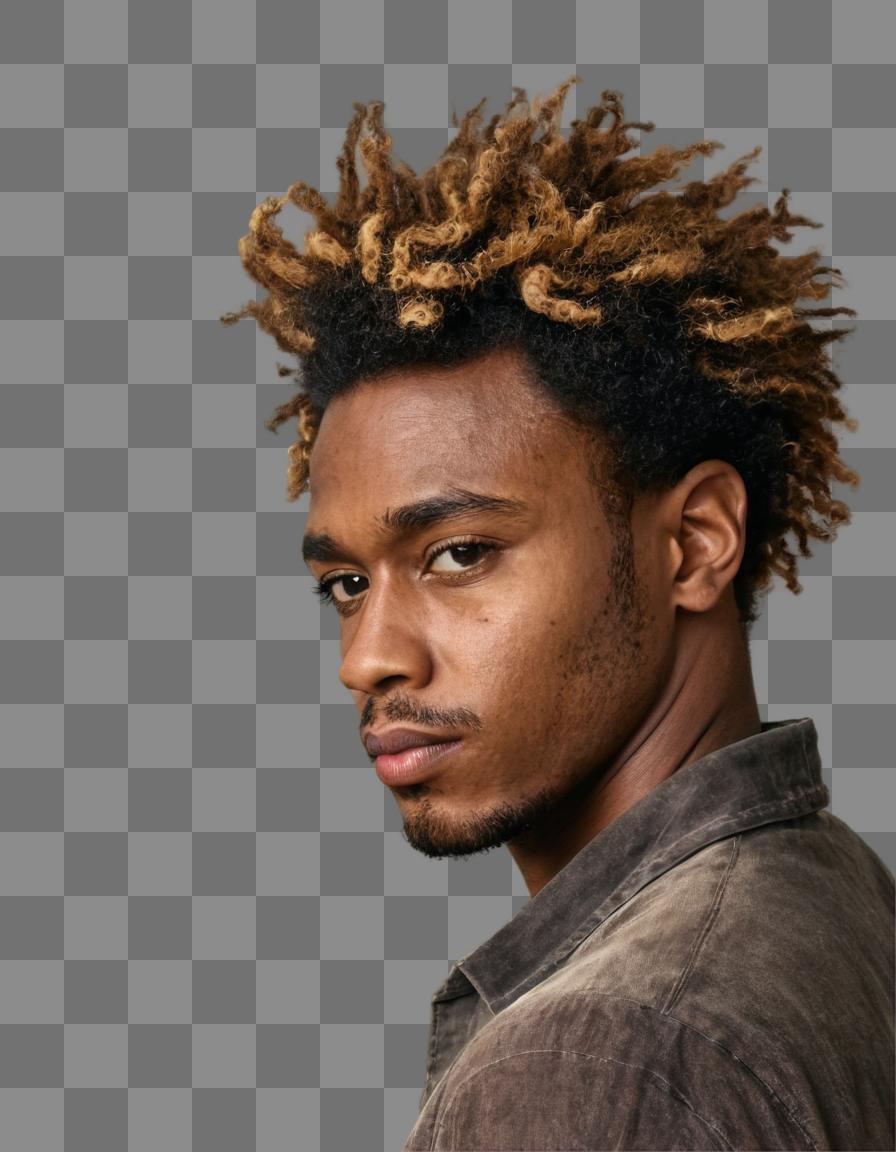}\hfill
\includegraphics[width=0.245\linewidth]{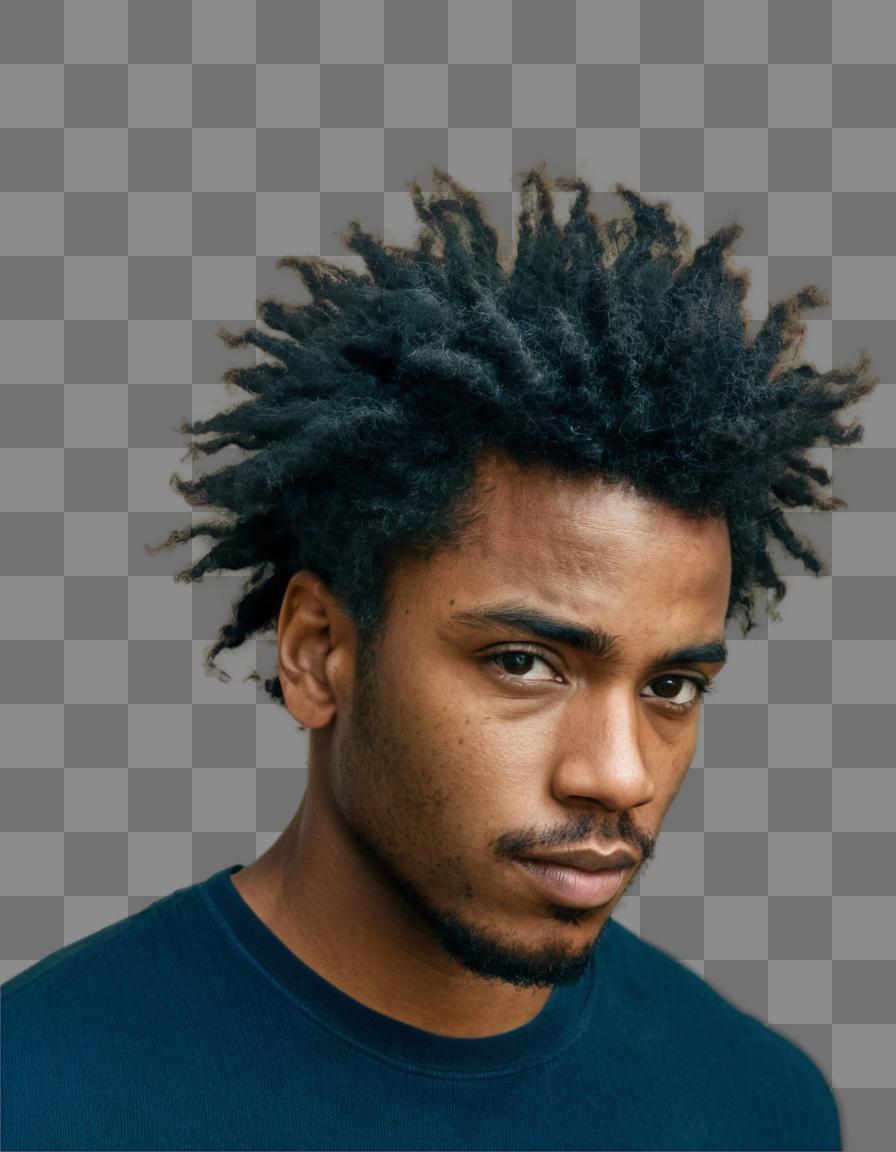}\hfill
\includegraphics[width=0.245\linewidth]{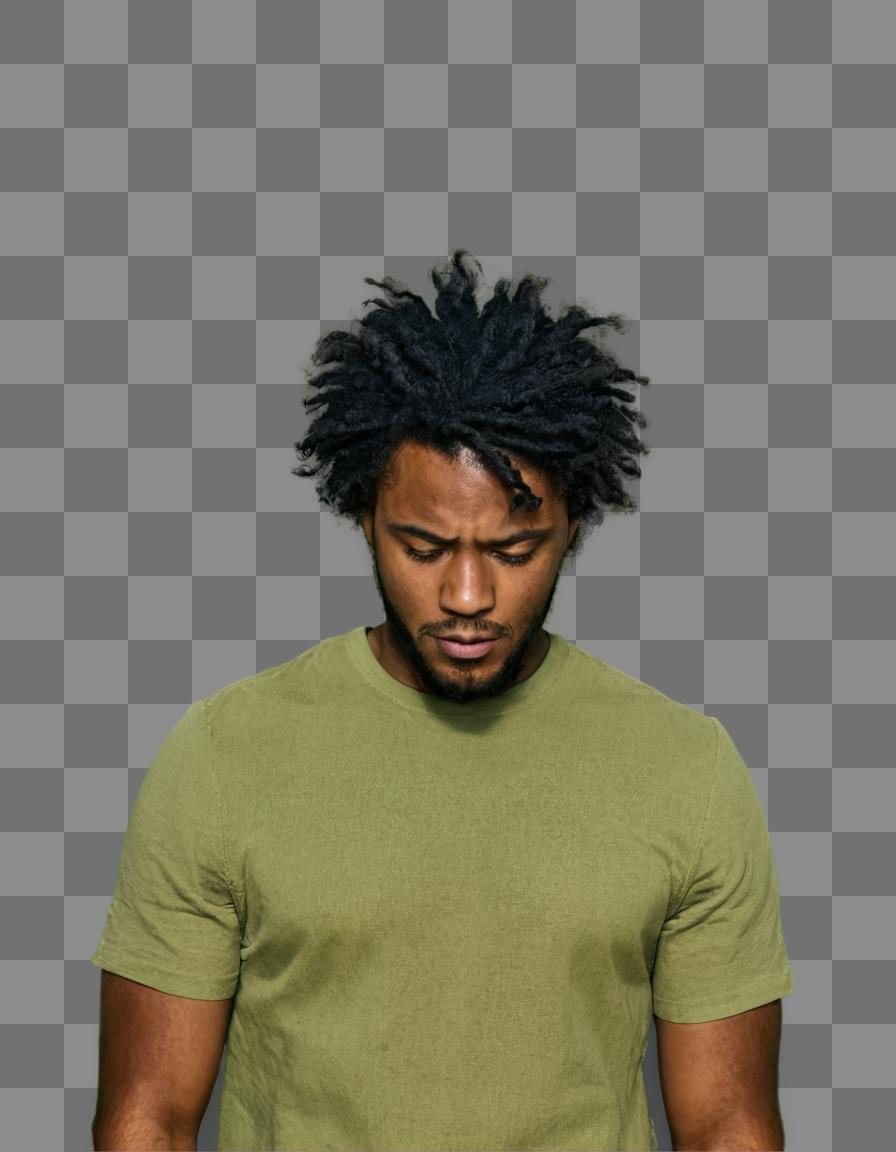}\hfill
\includegraphics[width=0.245\linewidth]{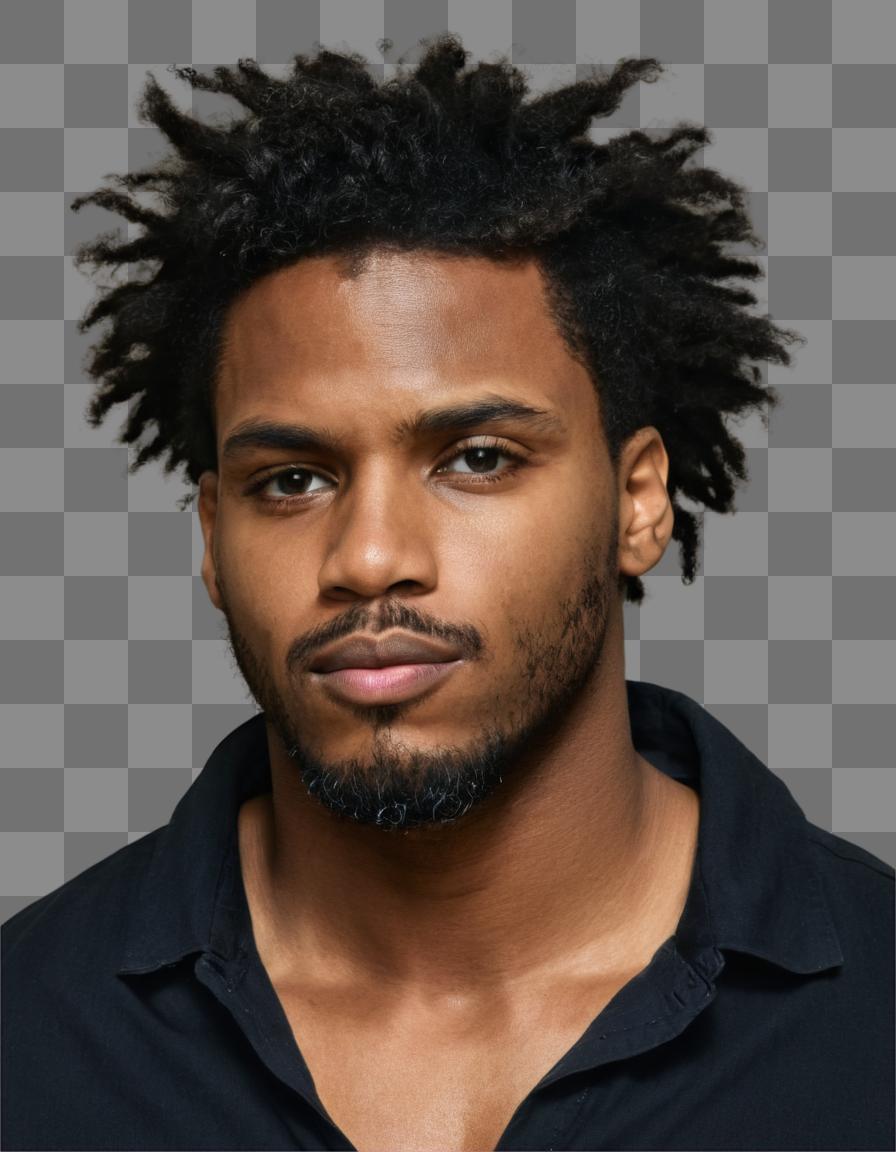}

\vspace{1pt}
\includegraphics[width=0.245\linewidth]{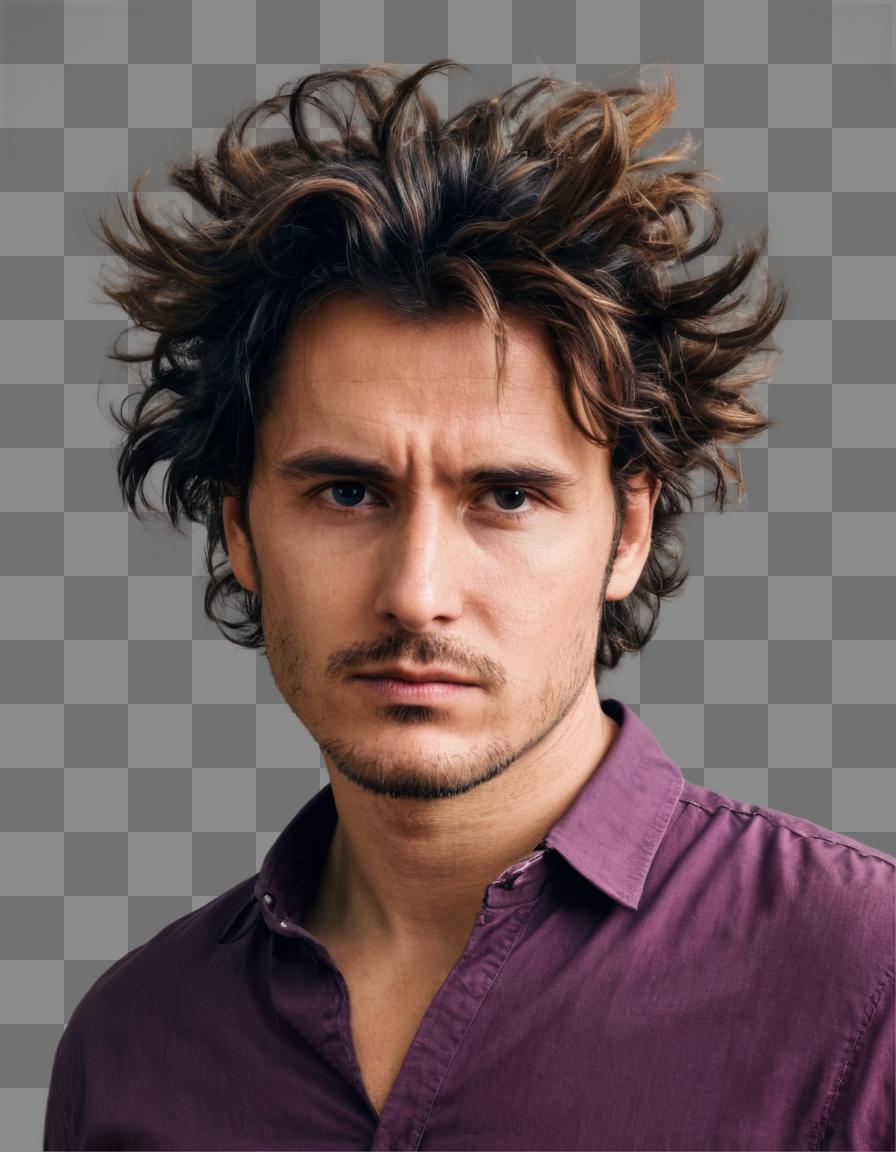}\hfill
\includegraphics[width=0.245\linewidth]{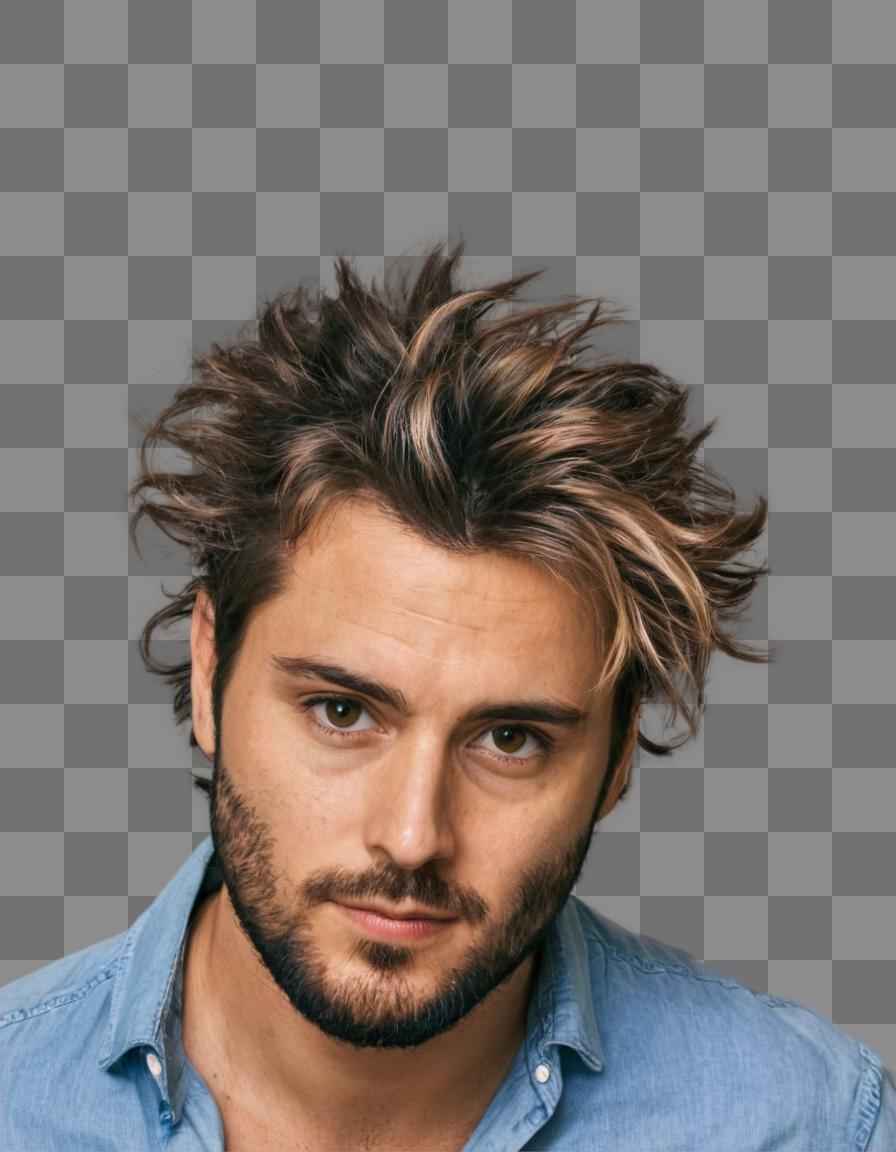}\hfill
\includegraphics[width=0.245\linewidth]{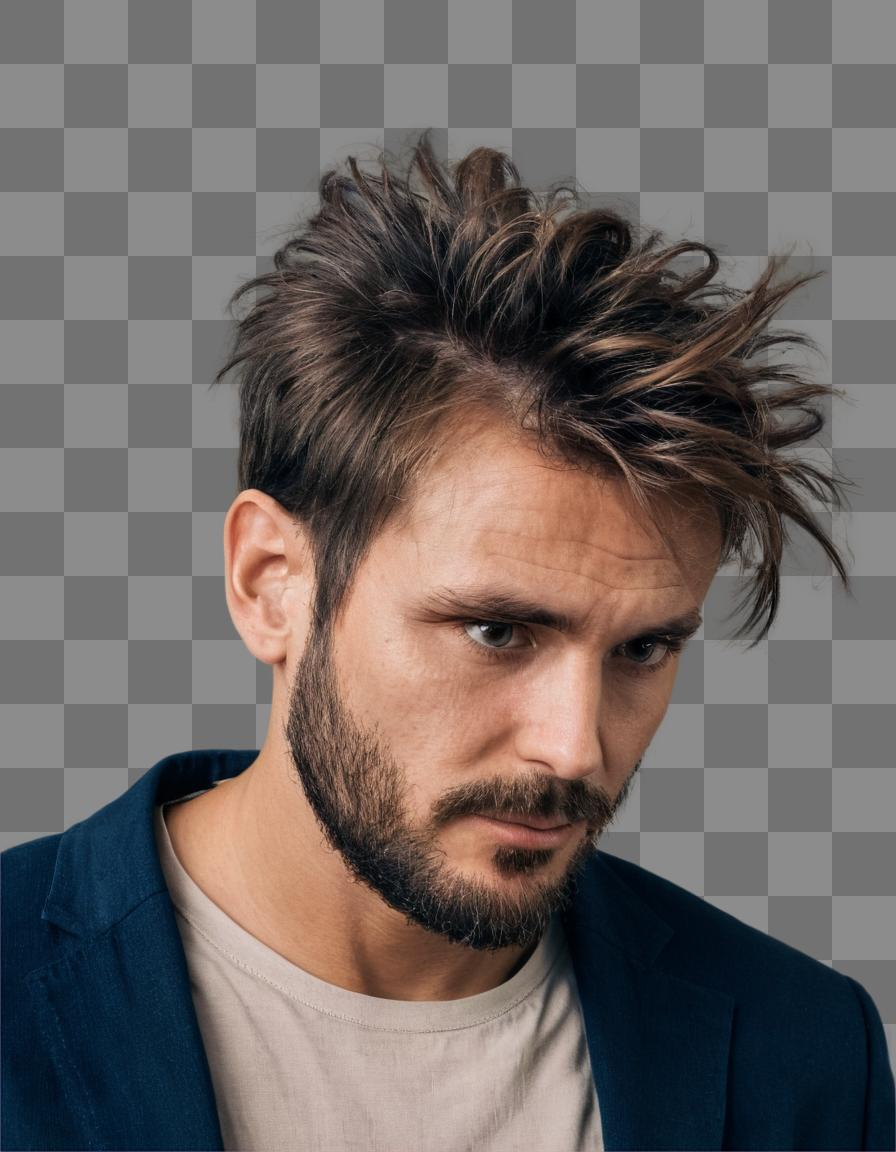}\hfill
\includegraphics[width=0.245\linewidth]{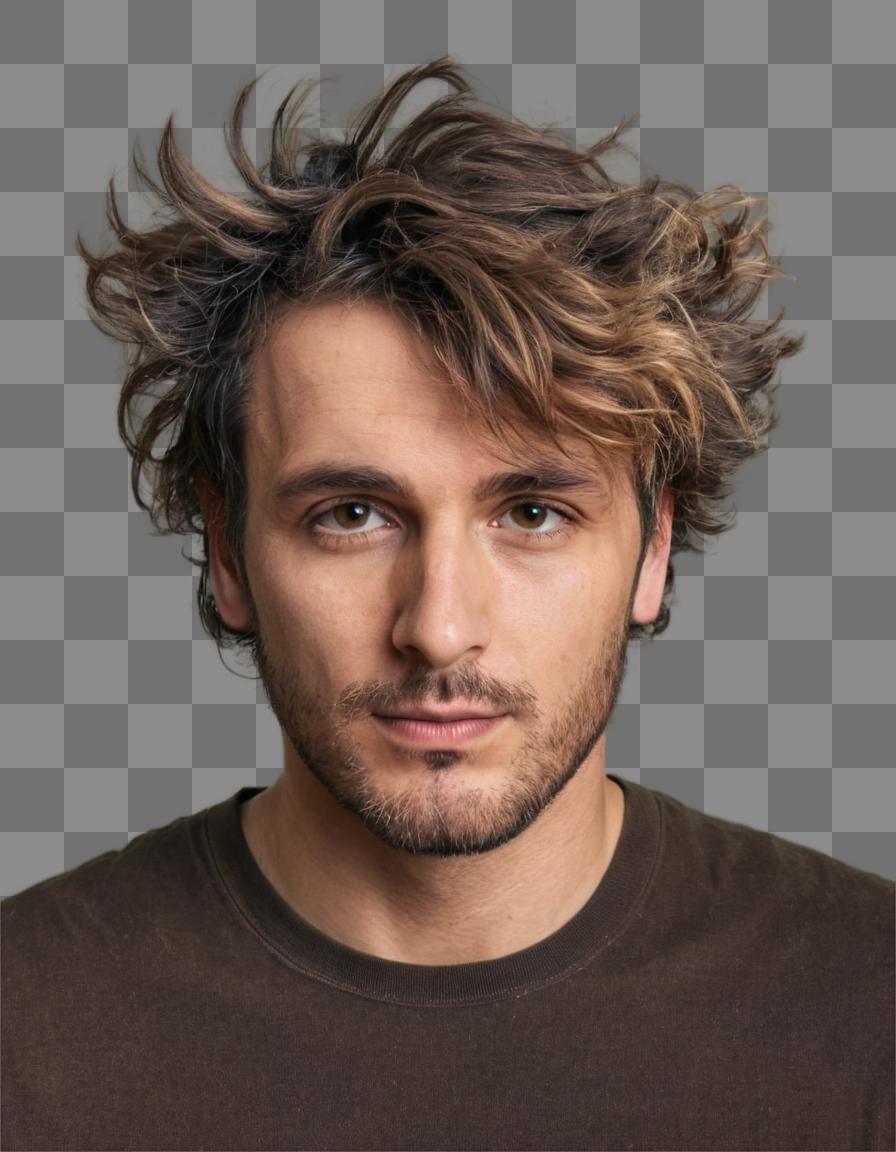}
\caption{Single Transparent Image Results \#4. The prompt is ``a man with messy hair''. Resolution is $896\times1152$.}
\label{fig:a4}
\end{minipage}
\end{figure*}

\begin{figure*}

\begin{minipage}{\linewidth}
\includegraphics[width=0.245\linewidth]{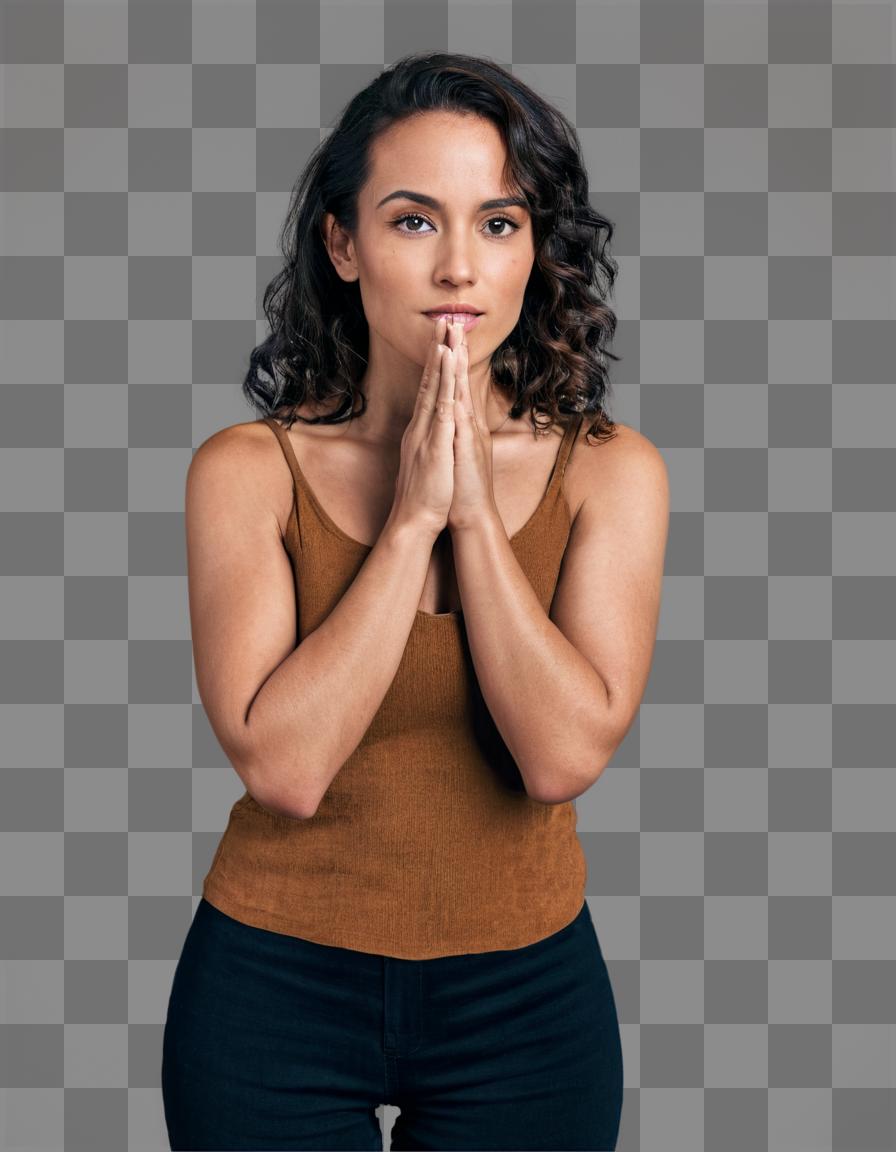}\hfill
\includegraphics[width=0.245\linewidth]{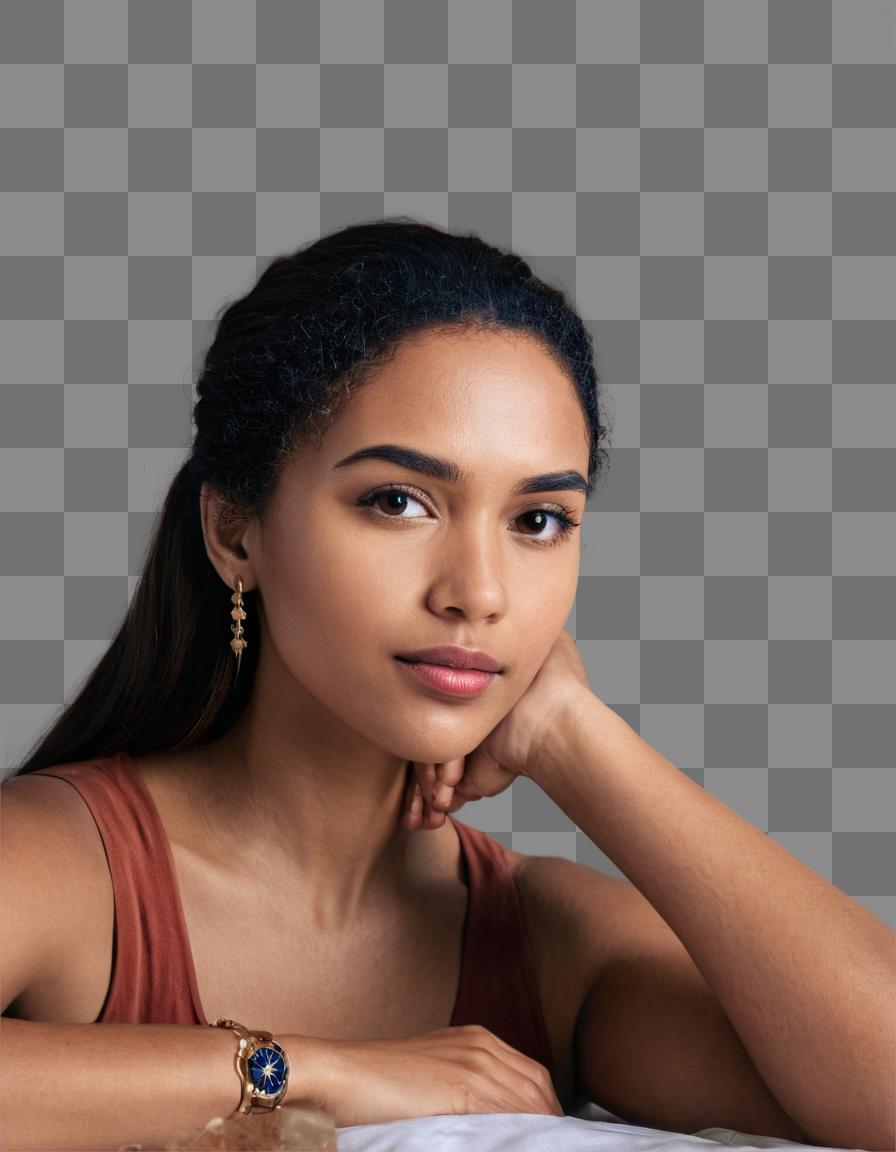}\hfill
\includegraphics[width=0.245\linewidth]{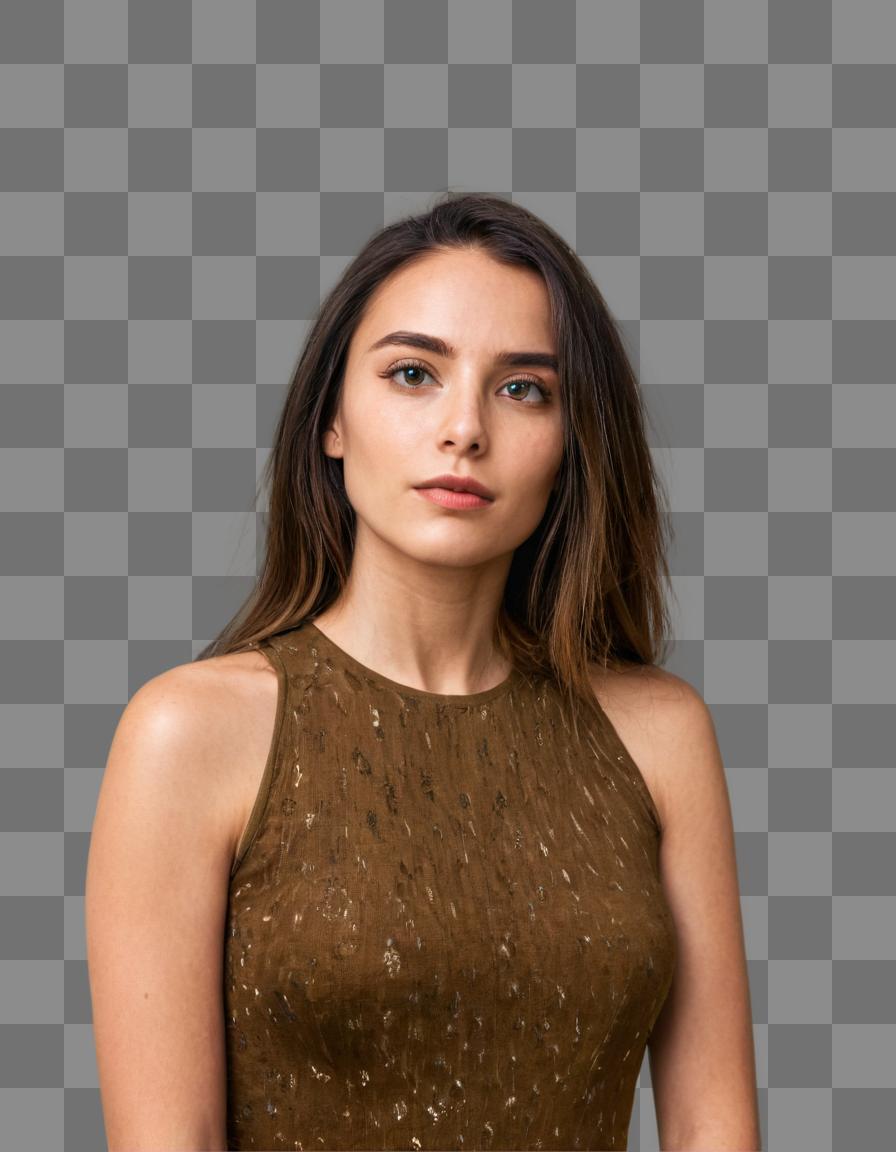}\hfill
\includegraphics[width=0.245\linewidth]{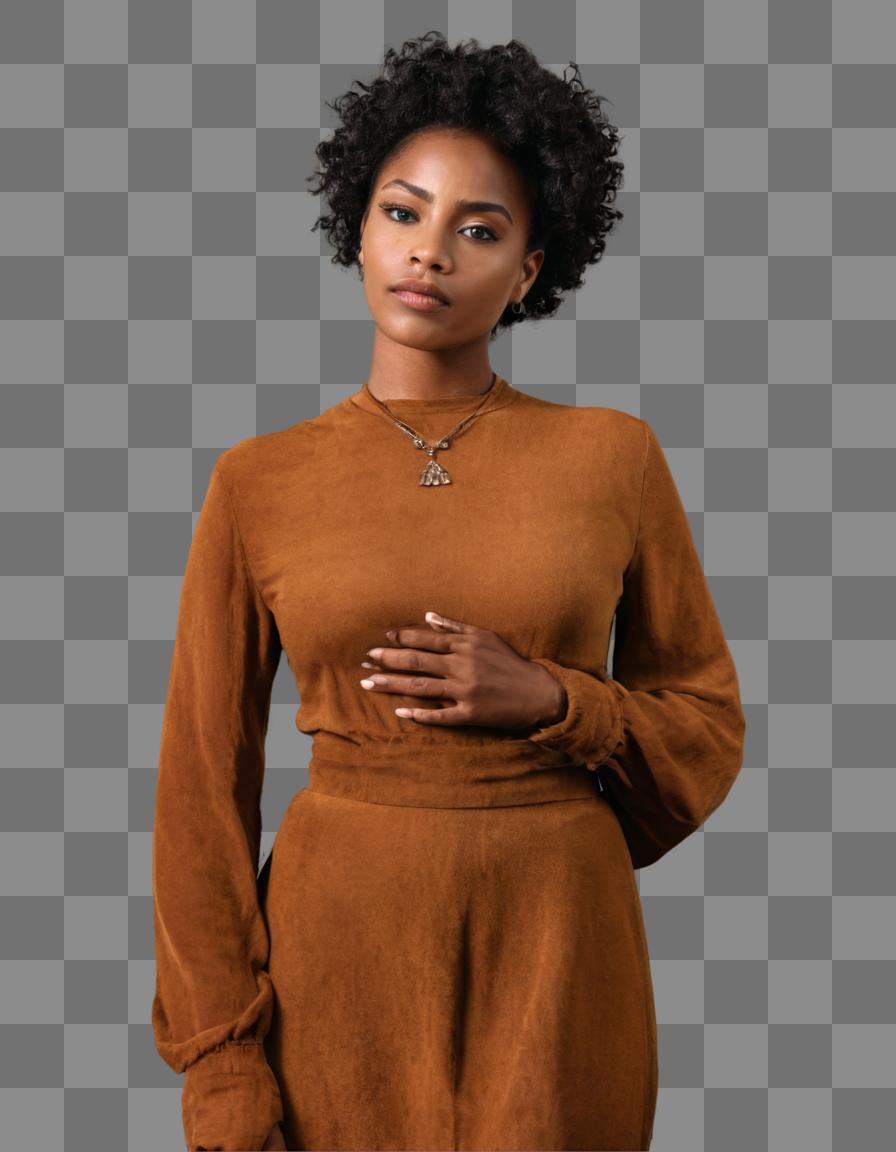}

\vspace{1pt}
\includegraphics[width=0.245\linewidth]{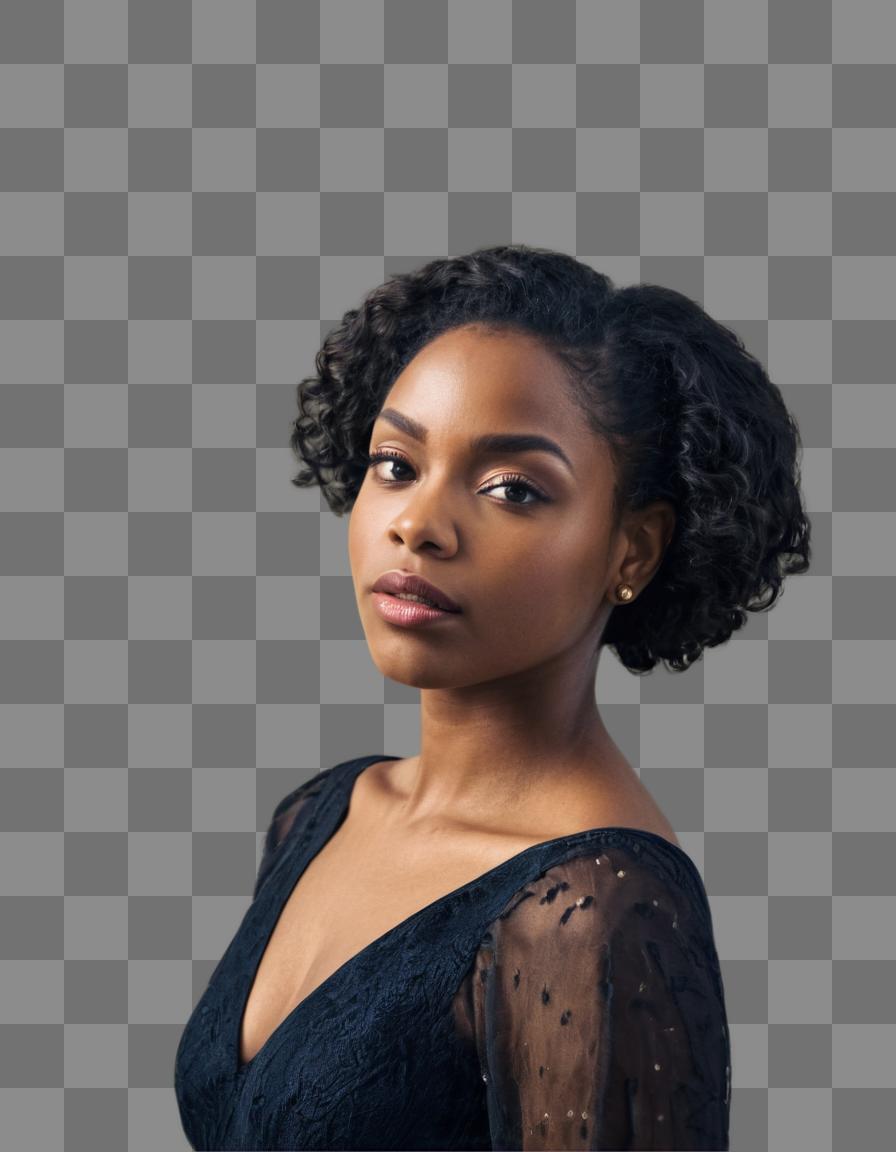}\hfill
\includegraphics[width=0.245\linewidth]{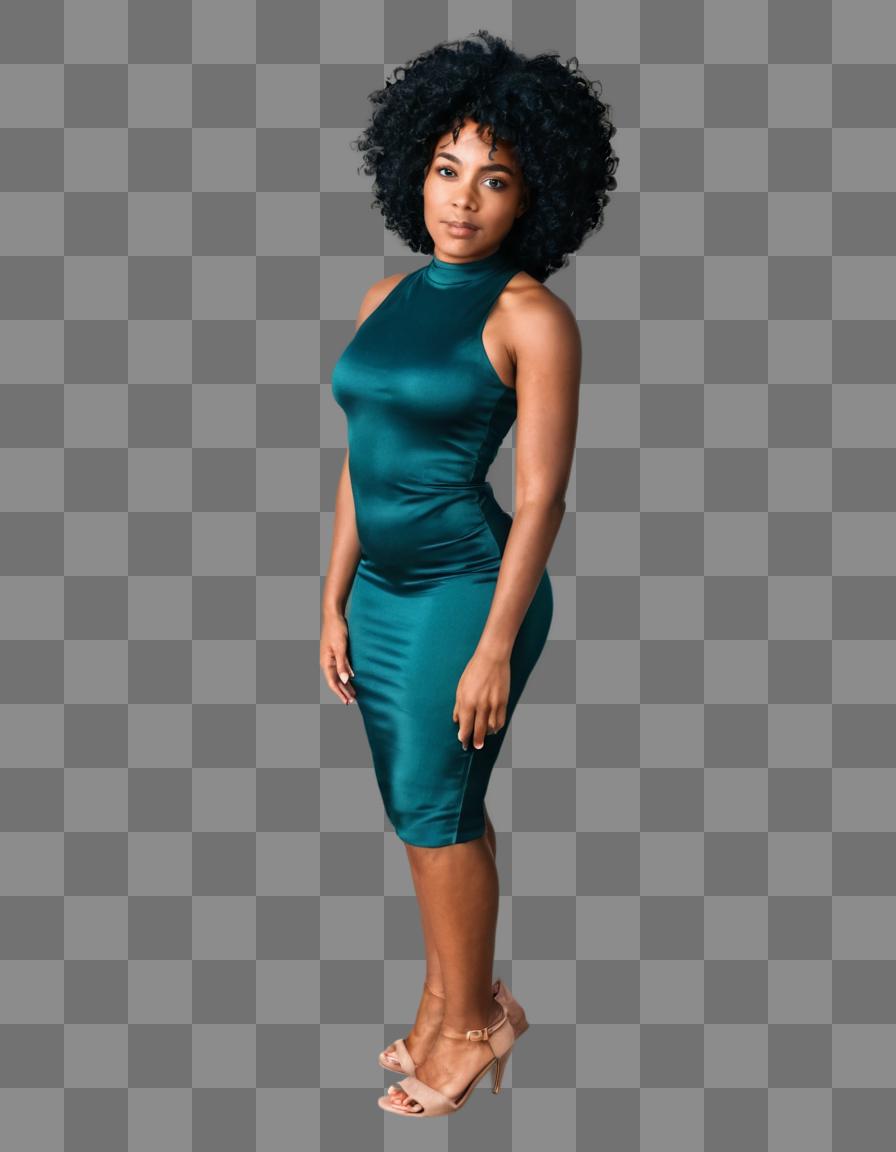}\hfill
\includegraphics[width=0.245\linewidth]{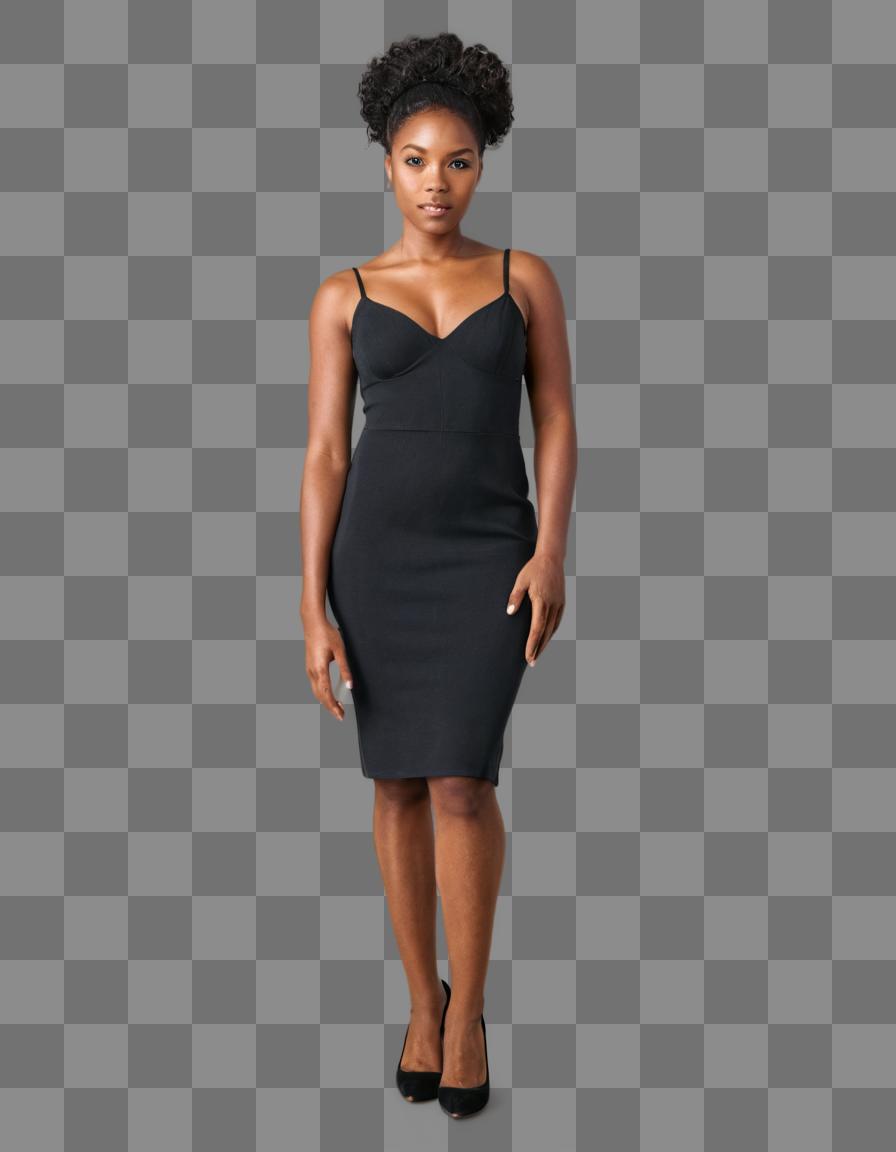}\hfill
\includegraphics[width=0.245\linewidth]{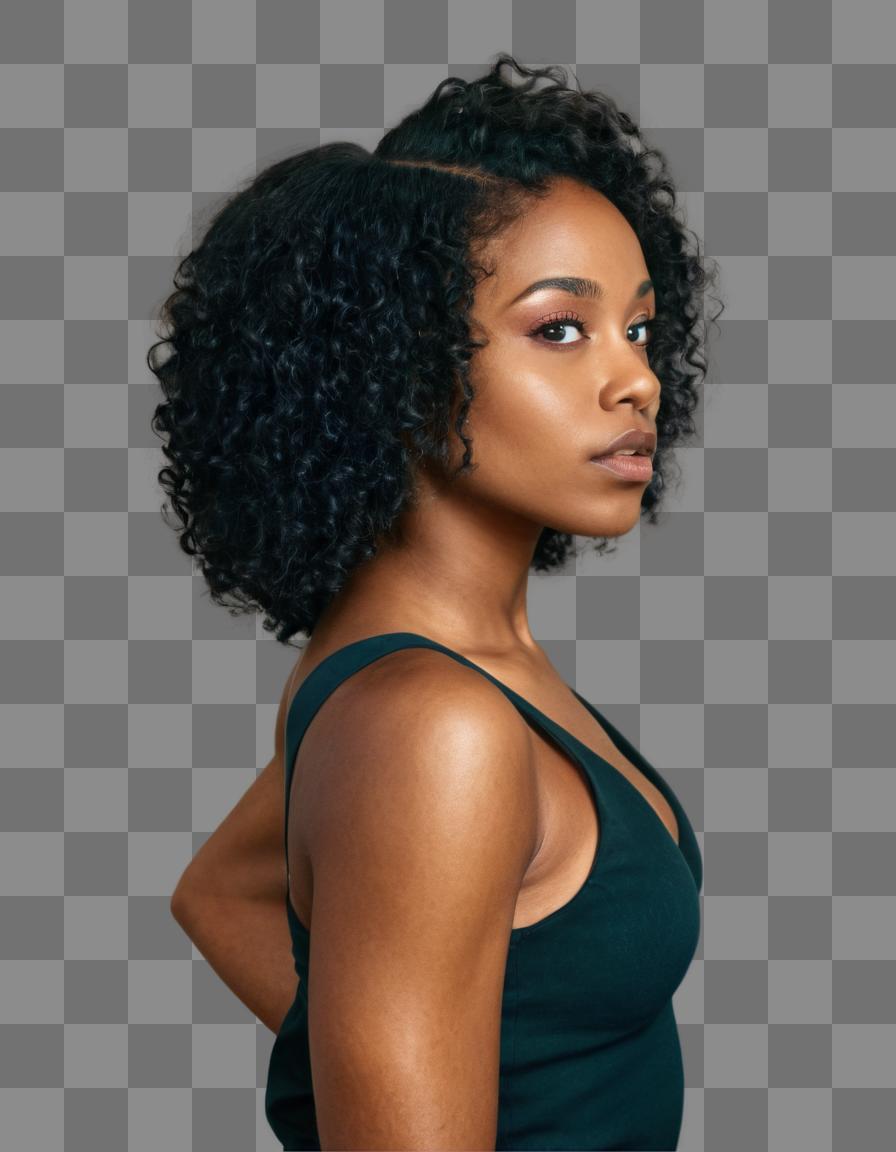}

\vspace{1pt}
\includegraphics[width=0.245\linewidth]{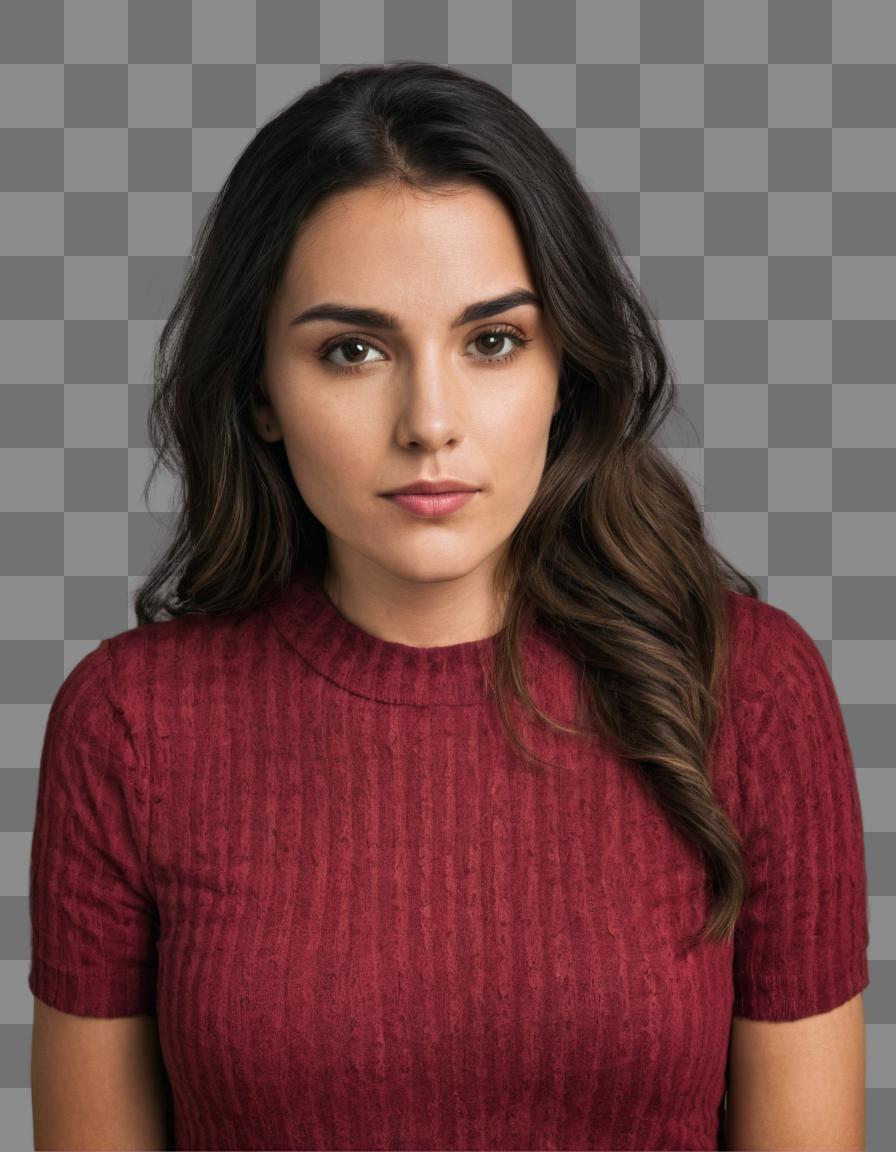}\hfill
\includegraphics[width=0.245\linewidth]{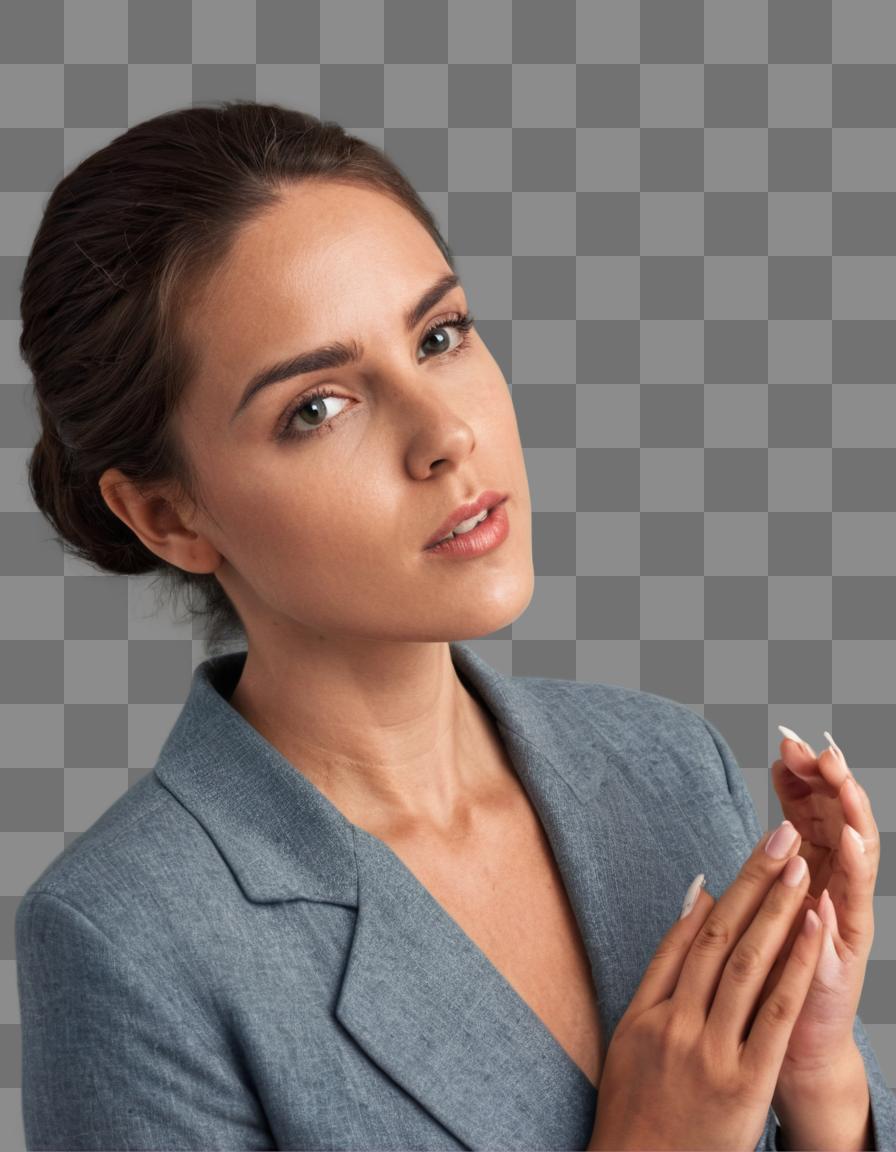}\hfill
\includegraphics[width=0.245\linewidth]{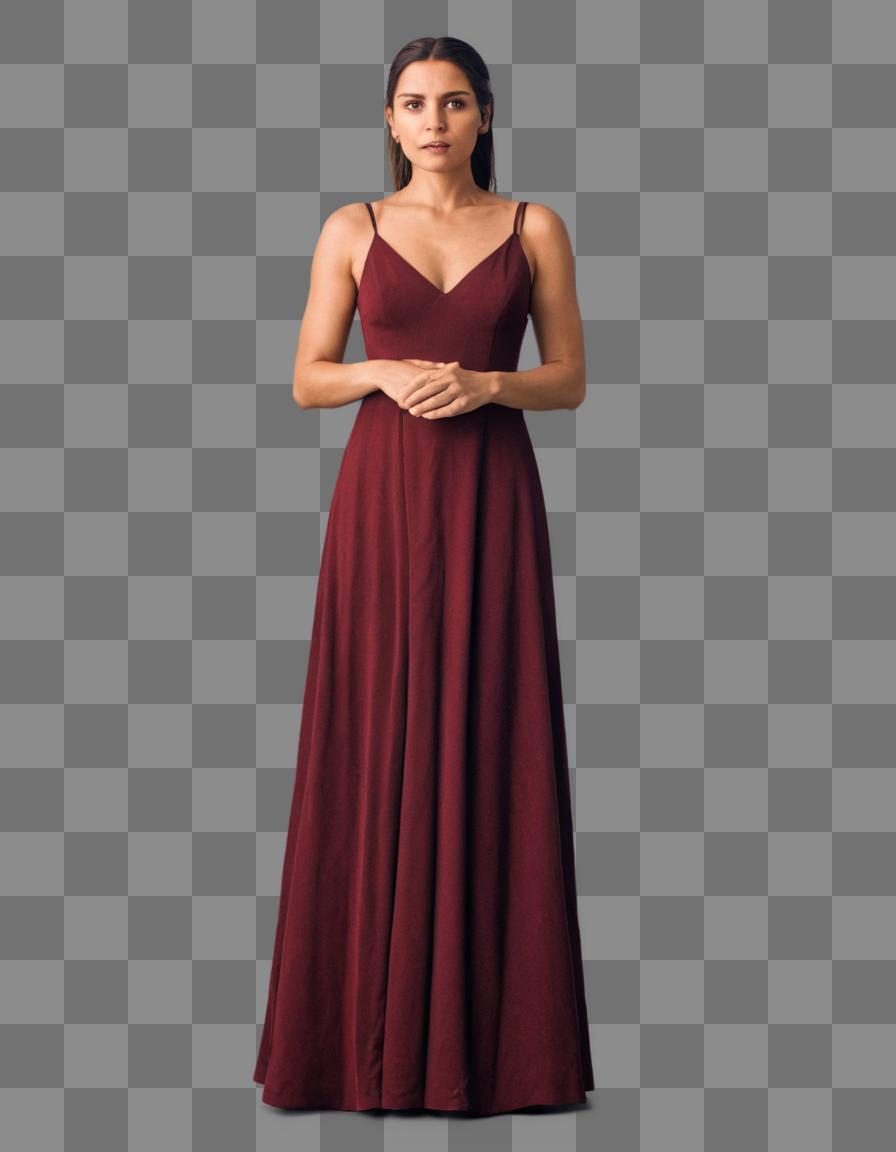}\hfill
\includegraphics[width=0.245\linewidth]{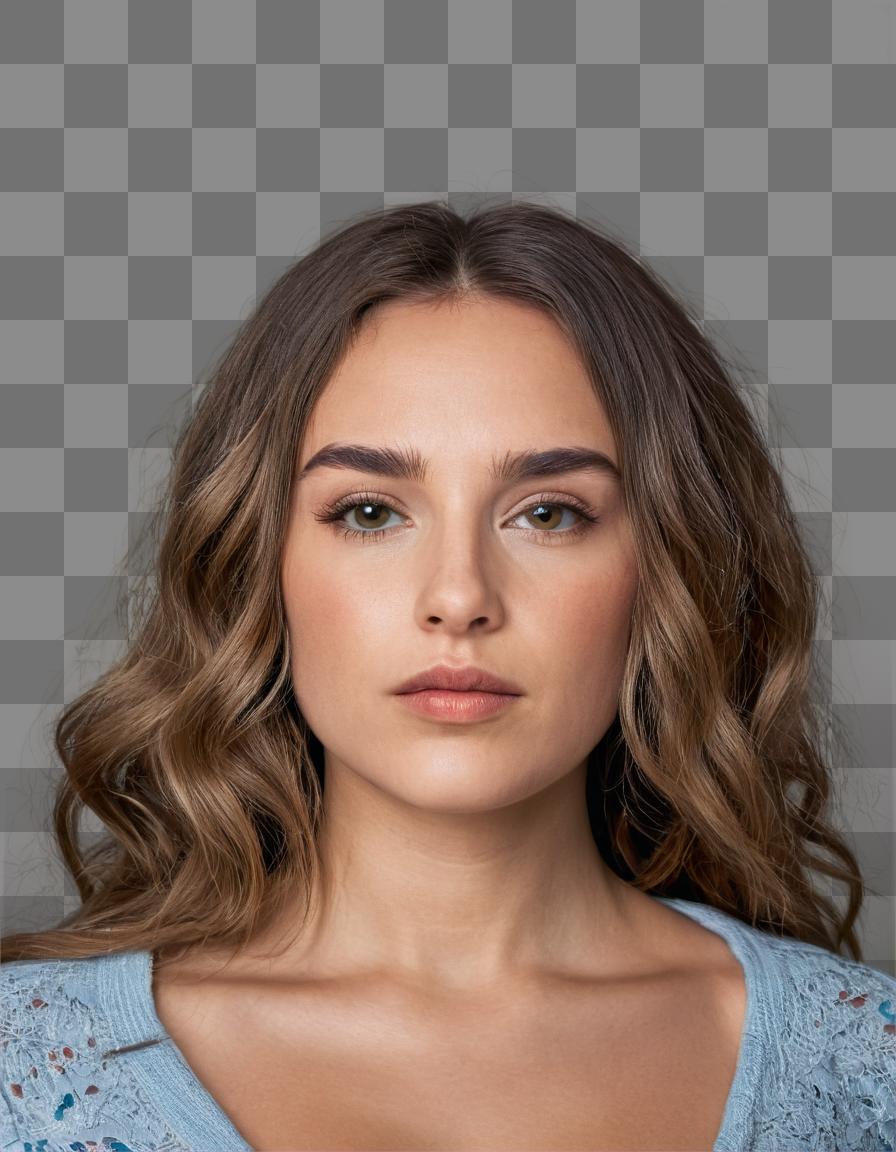}
\caption{Single Transparent Image Results \#5. The prompt is ``woman''. Resolution is $896\times1152$.}
\label{fig:a5}
\end{minipage}
\end{figure*}

\begin{figure*}

\begin{minipage}{\linewidth}
\includegraphics[width=0.245\linewidth]{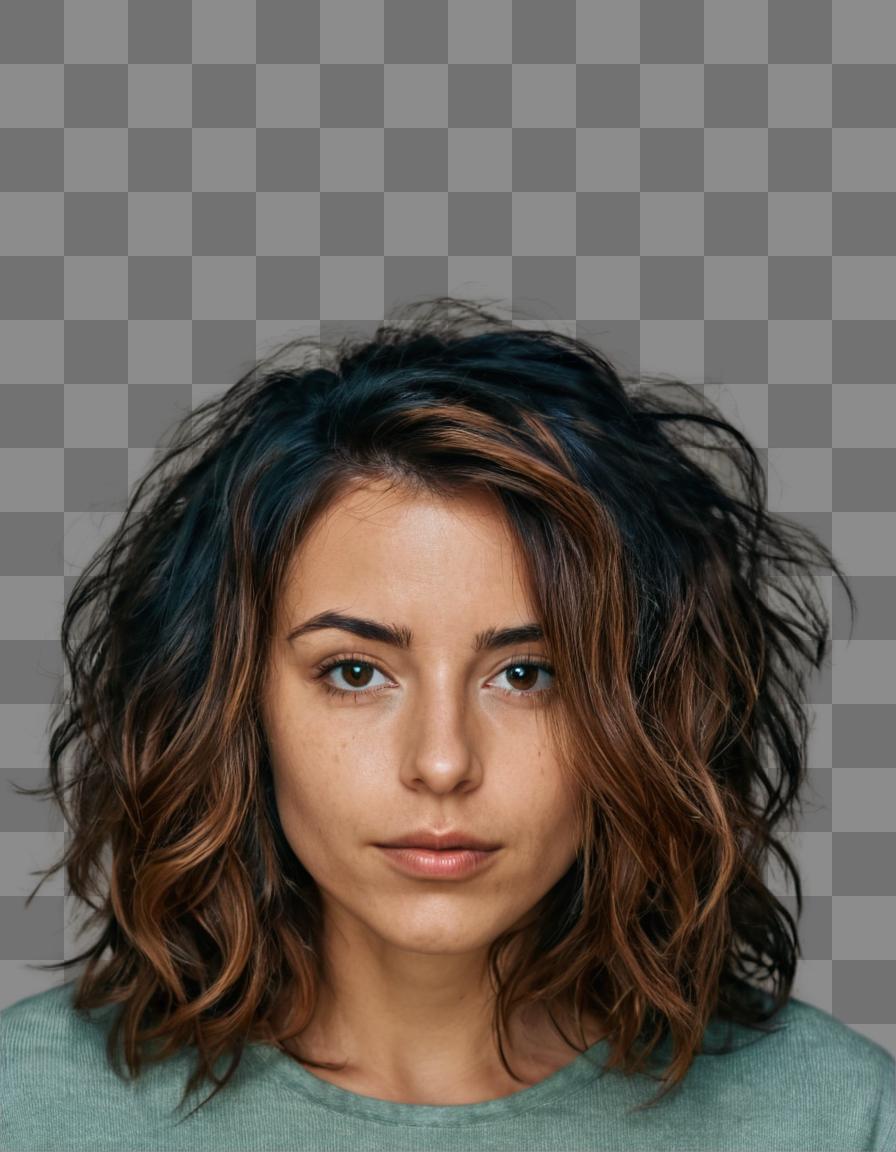}\hfill
\includegraphics[width=0.245\linewidth]{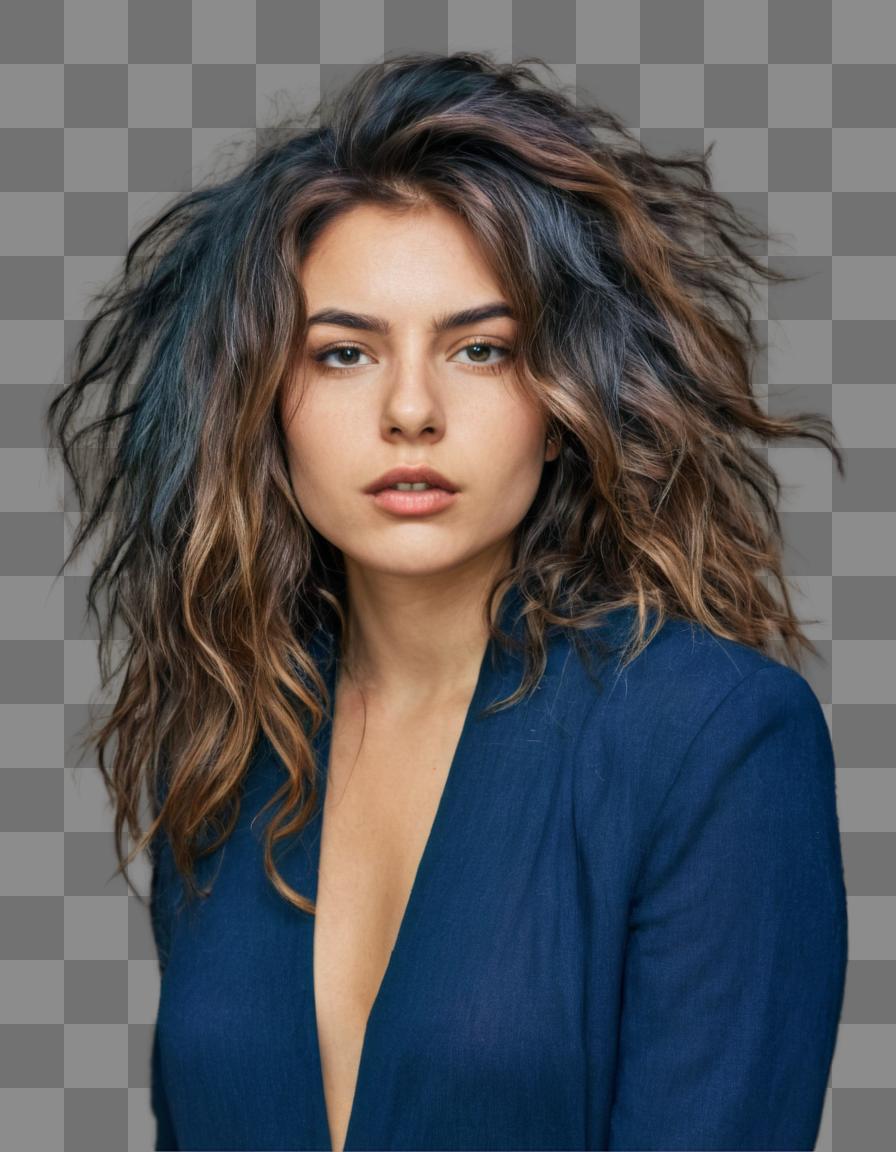}\hfill
\includegraphics[width=0.245\linewidth]{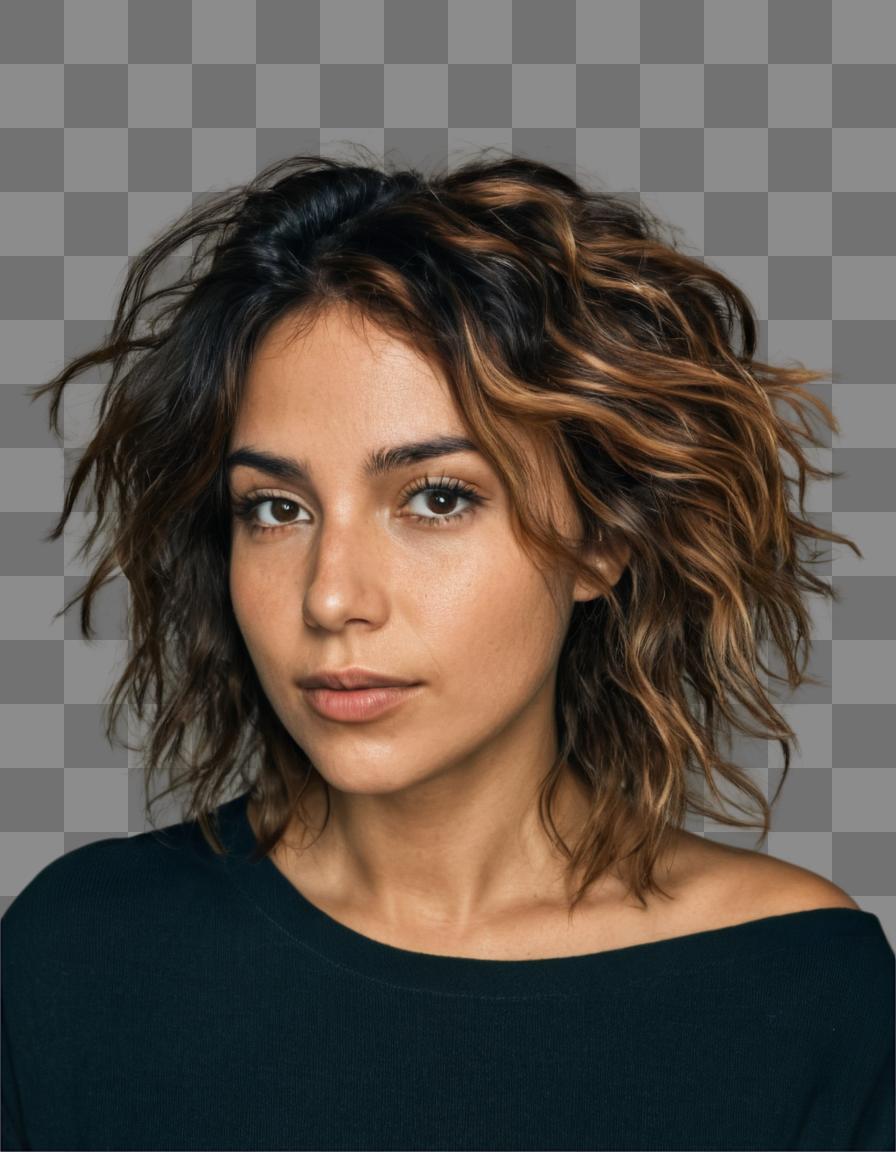}\hfill
\includegraphics[width=0.245\linewidth]{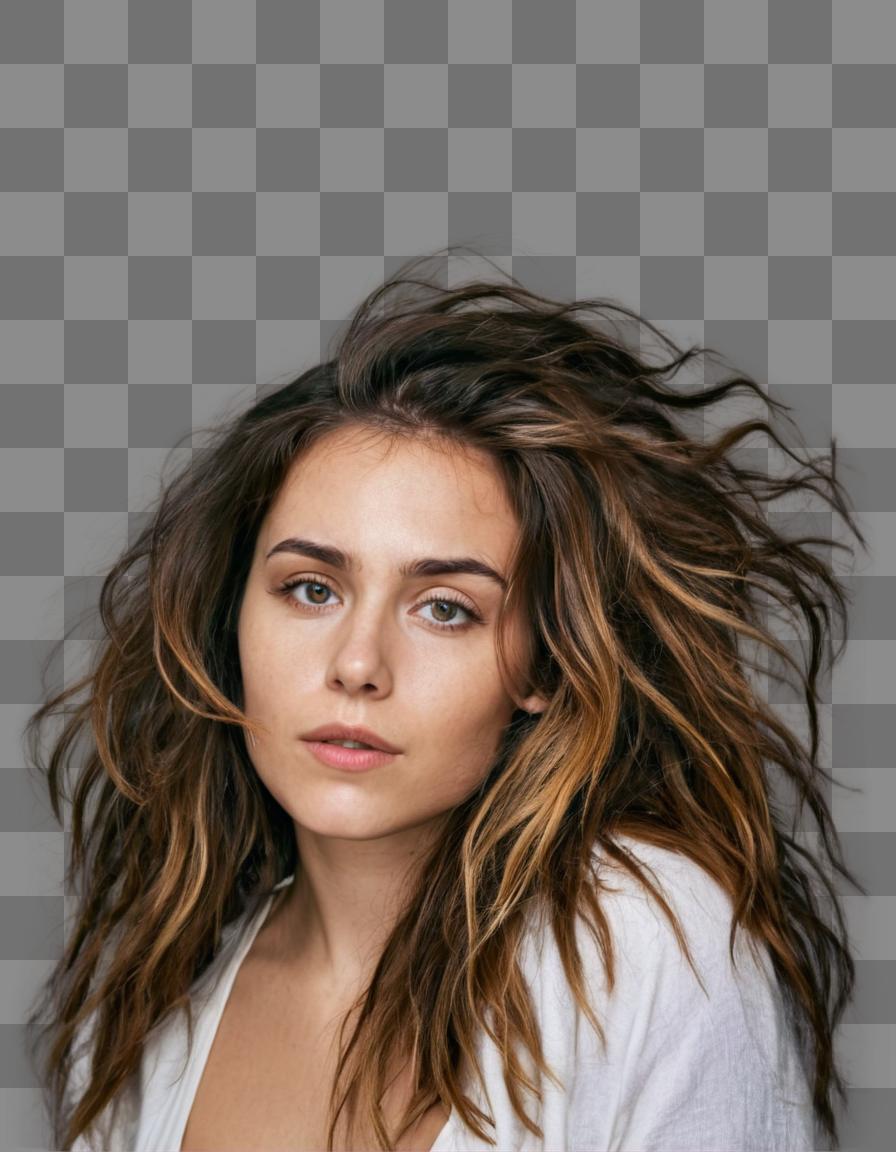}

\vspace{1pt}
\includegraphics[width=0.245\linewidth]{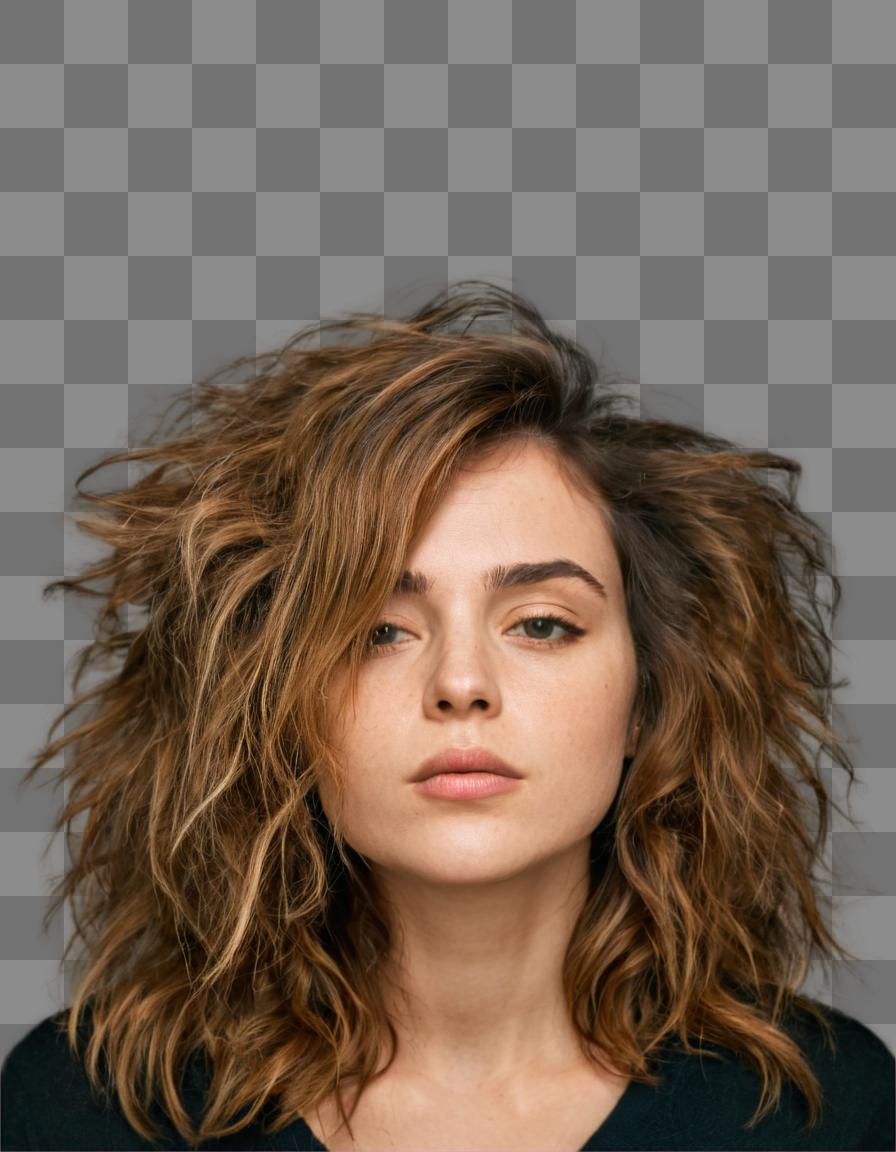}\hfill
\includegraphics[width=0.245\linewidth]{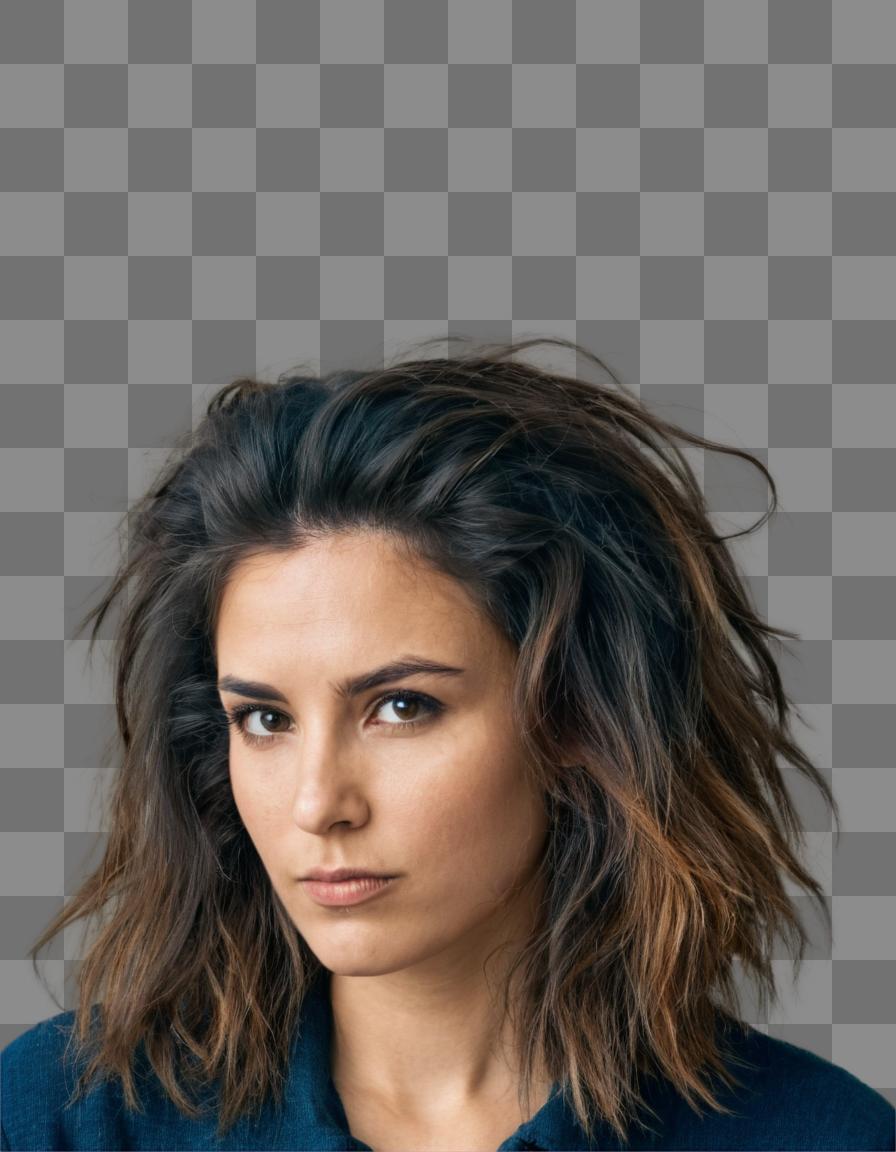}\hfill
\includegraphics[width=0.245\linewidth]{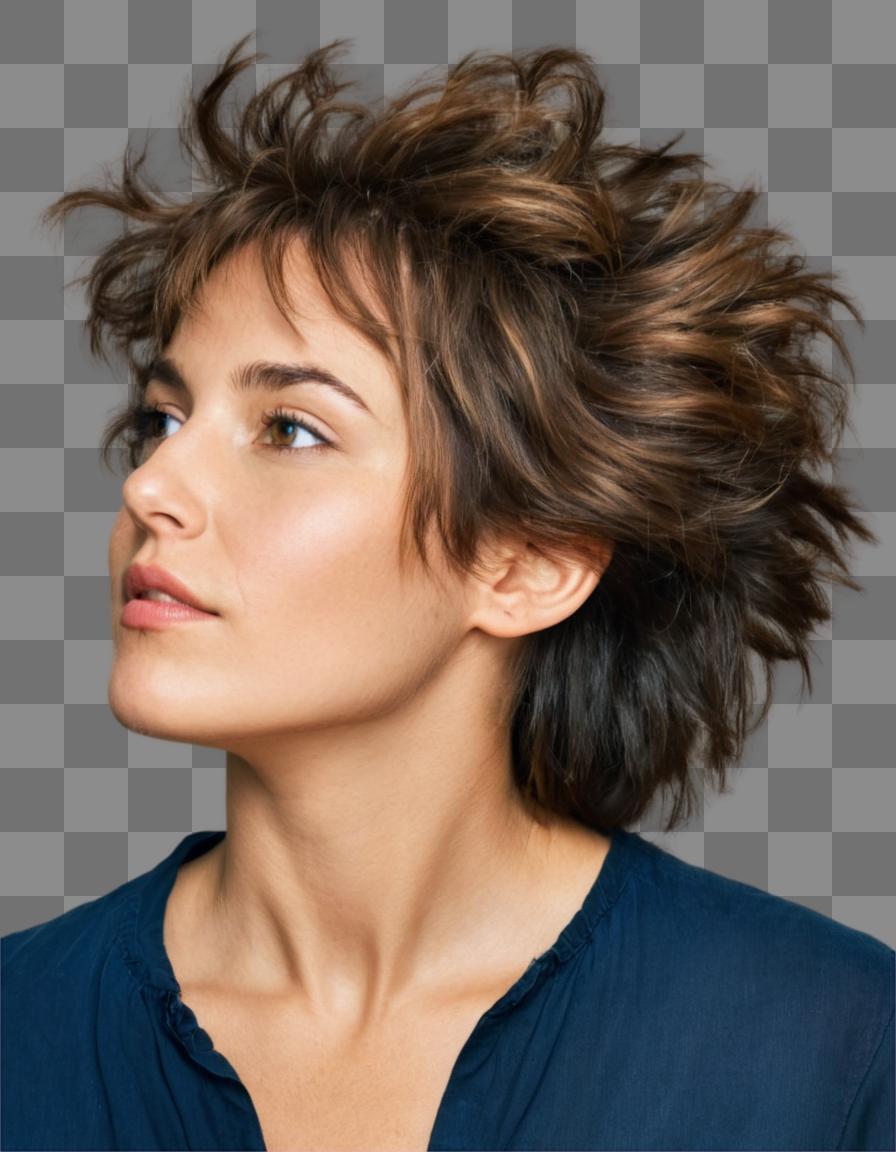}\hfill
\includegraphics[width=0.245\linewidth]{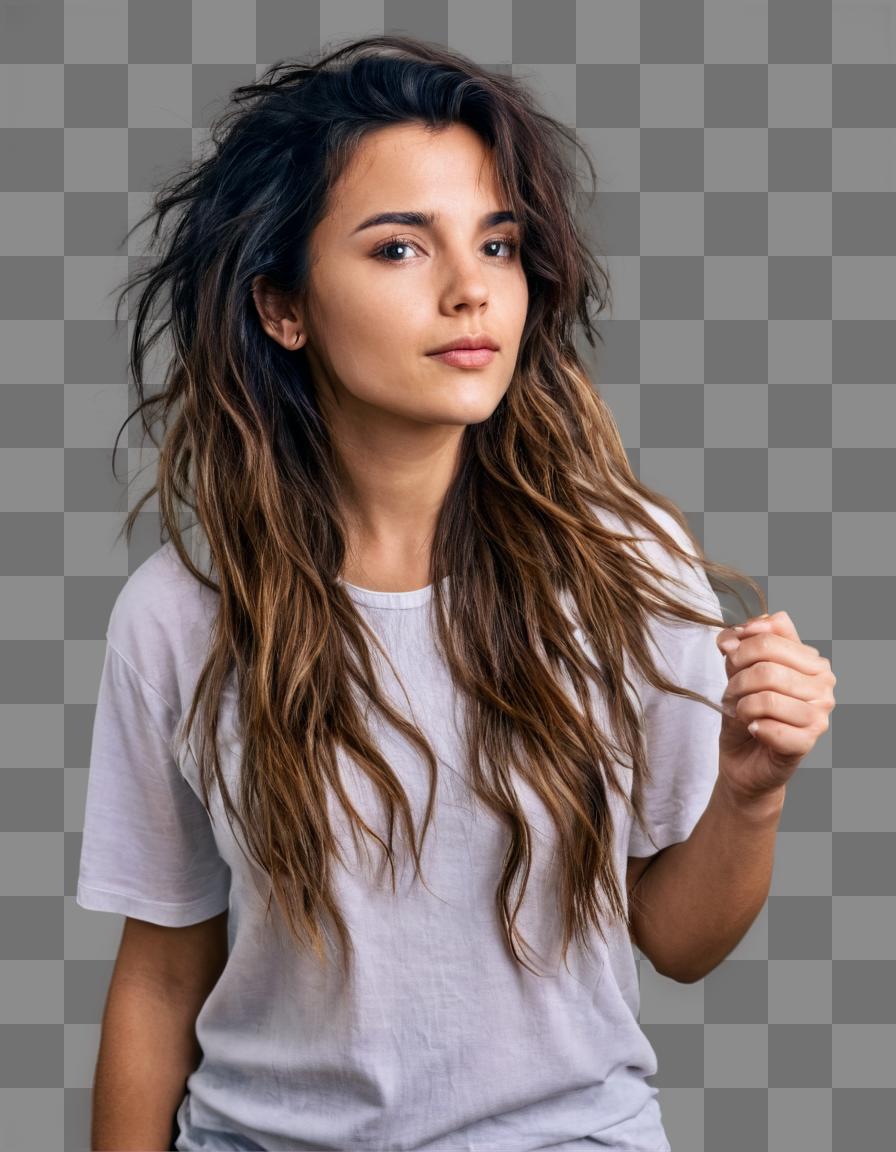}

\vspace{1pt}
\includegraphics[width=0.245\linewidth]{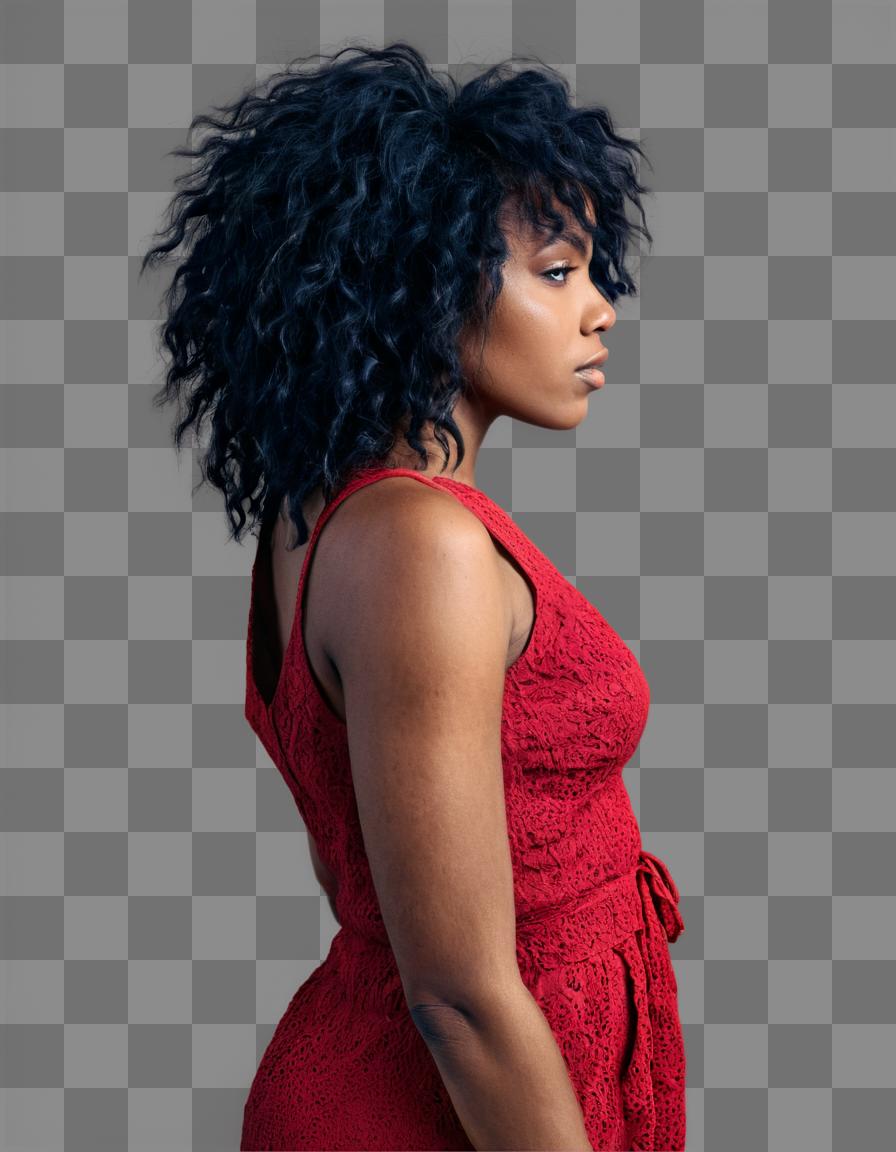}\hfill
\includegraphics[width=0.245\linewidth]{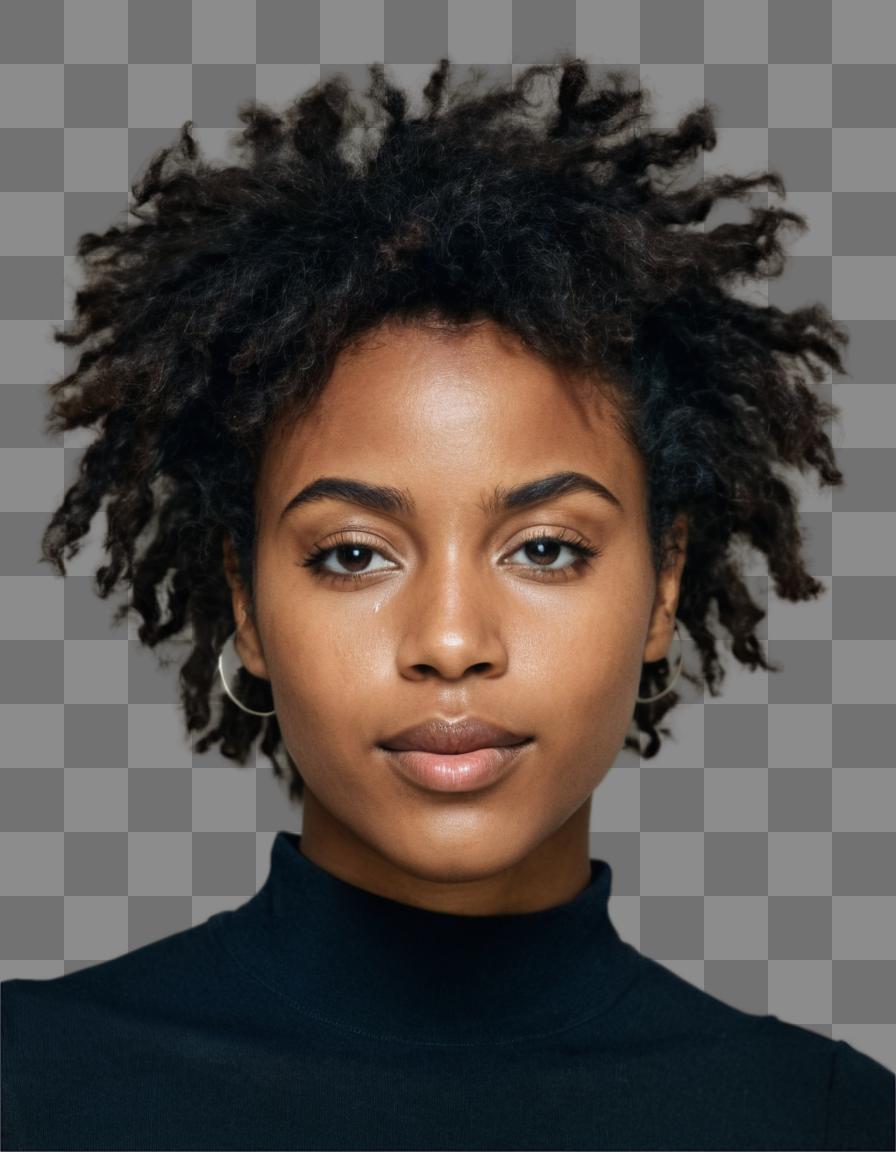}\hfill
\includegraphics[width=0.245\linewidth]{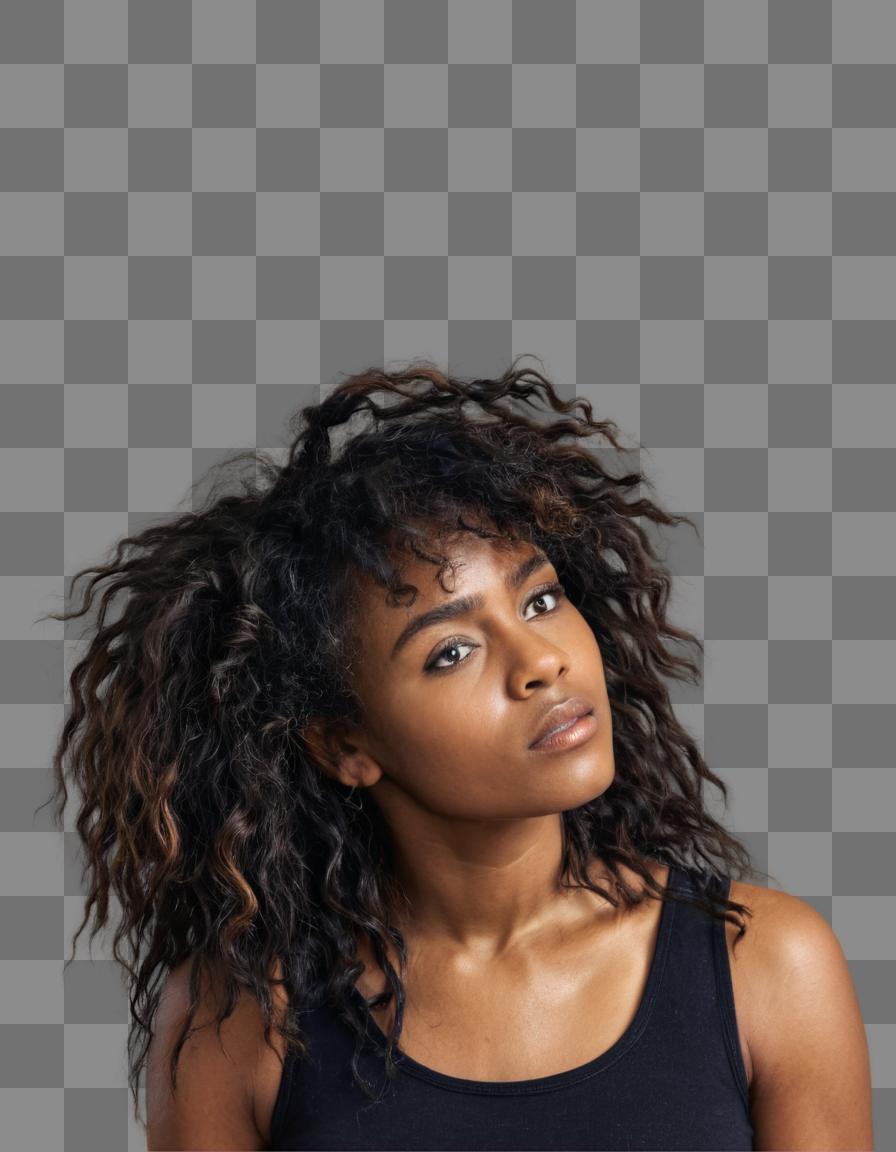}\hfill
\includegraphics[width=0.245\linewidth]{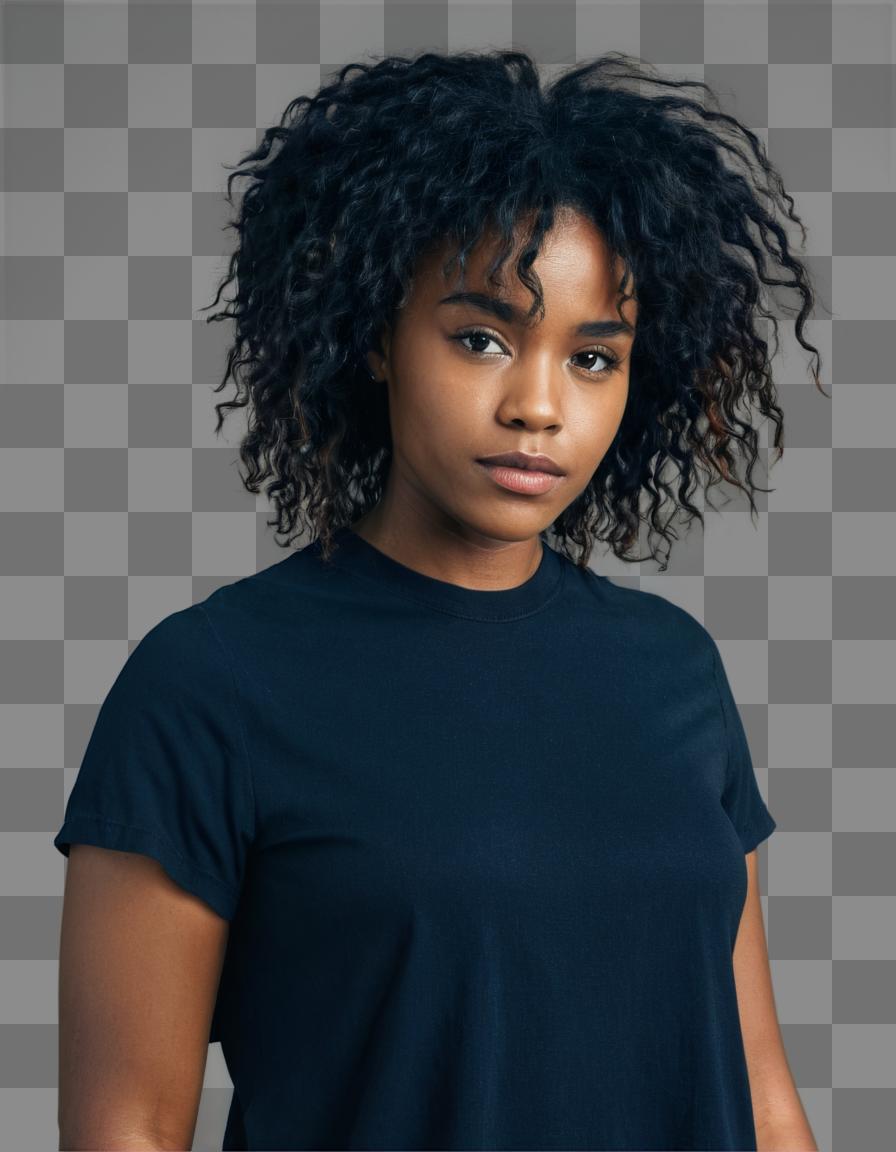}
\caption{Single Transparent Image Results \#6. The prompt is ``woman with messy hair''. Resolution is $896\times1152$.}
\label{fig:a6}
\end{minipage}
\end{figure*}

\begin{figure*}

\begin{minipage}{\linewidth}
\includegraphics[width=0.245\linewidth]{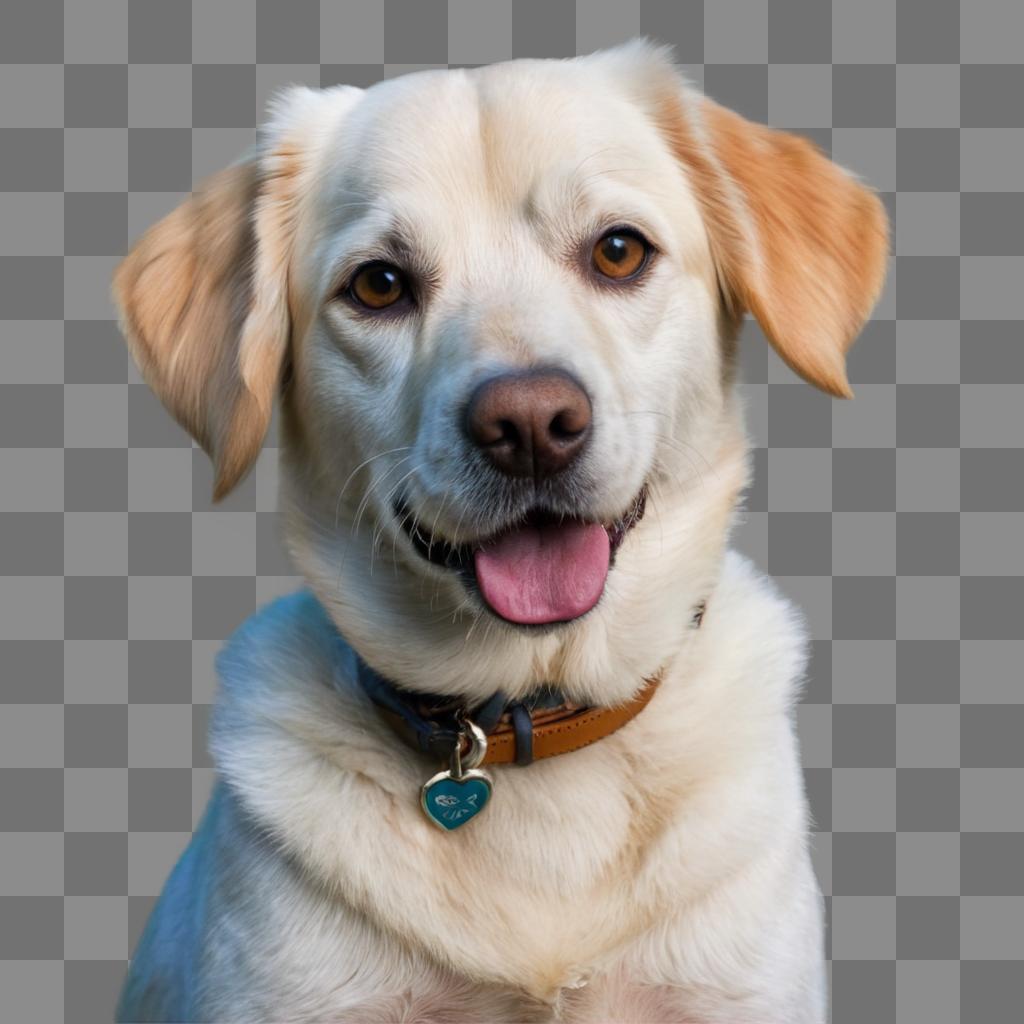}\hfill
\includegraphics[width=0.245\linewidth]{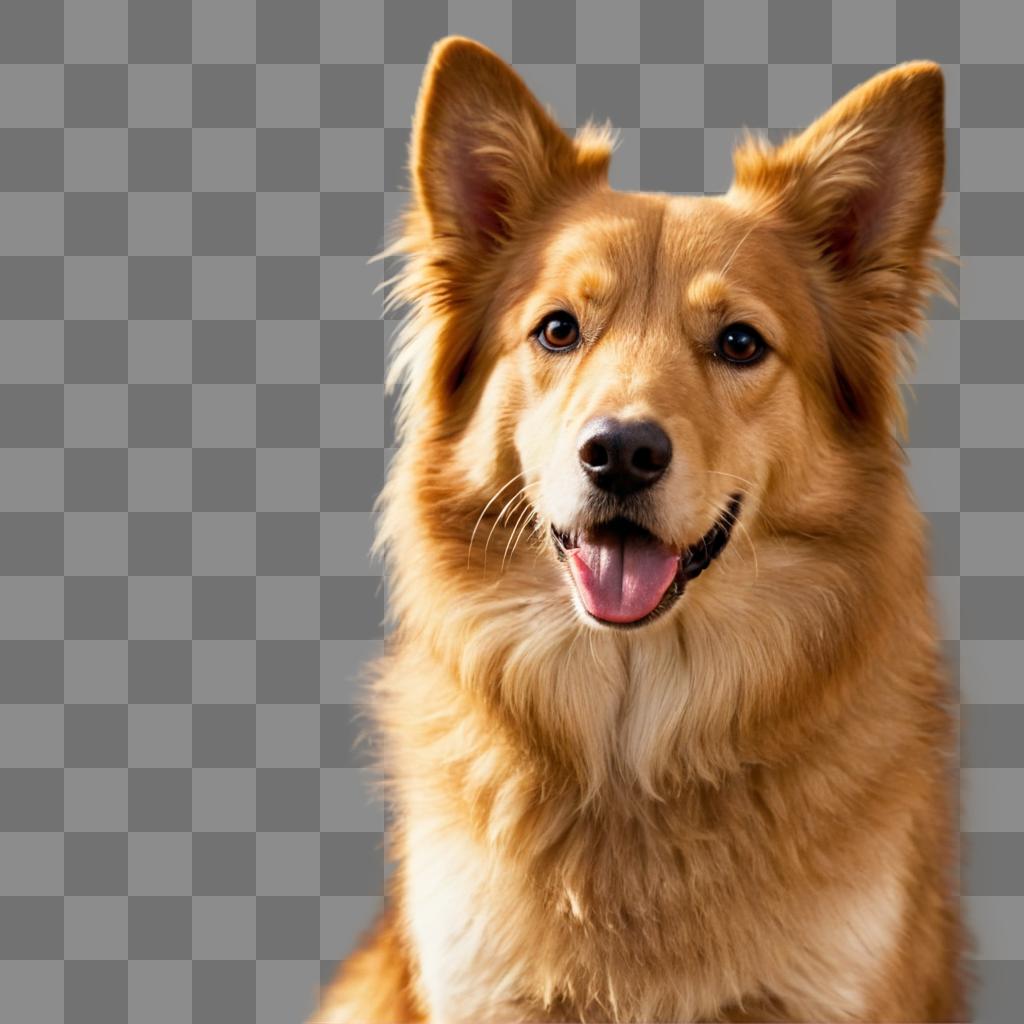}\hfill
\includegraphics[width=0.245\linewidth]{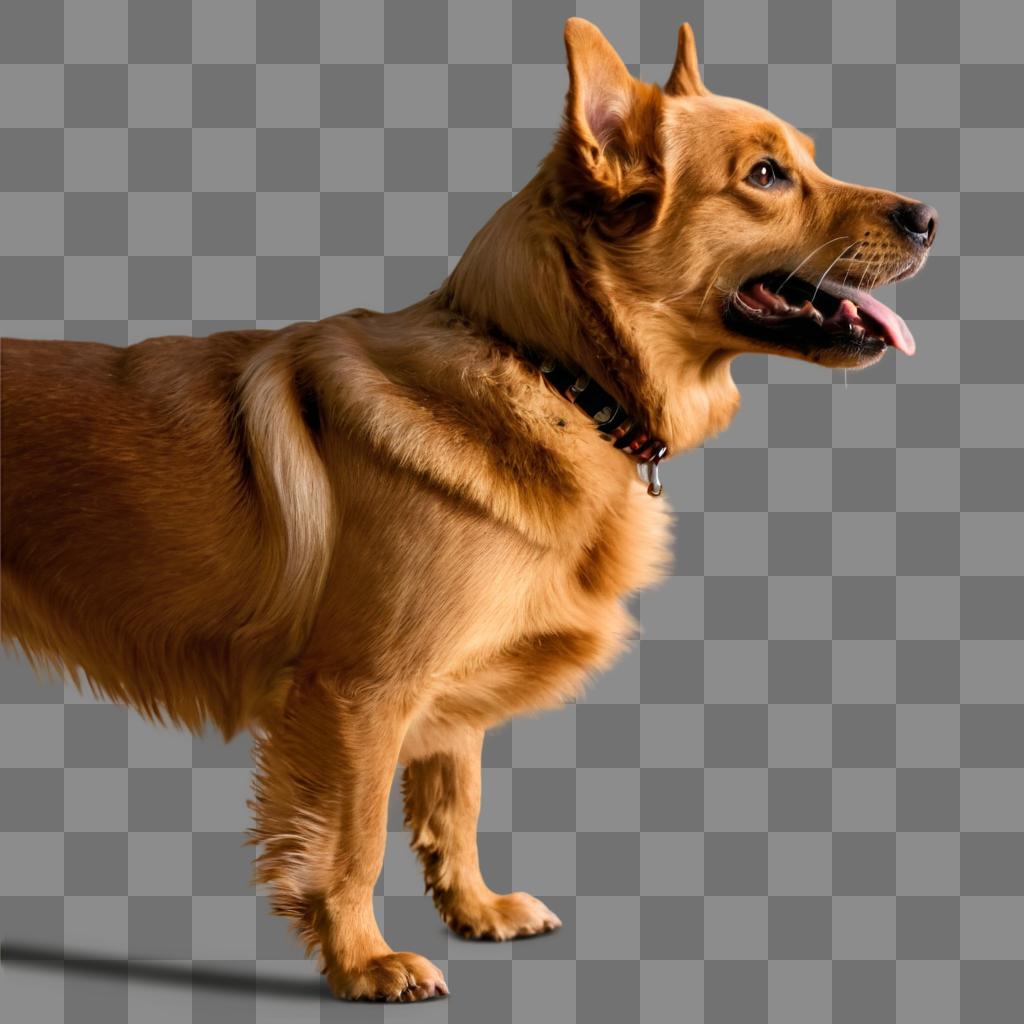}\hfill
\includegraphics[width=0.245\linewidth]{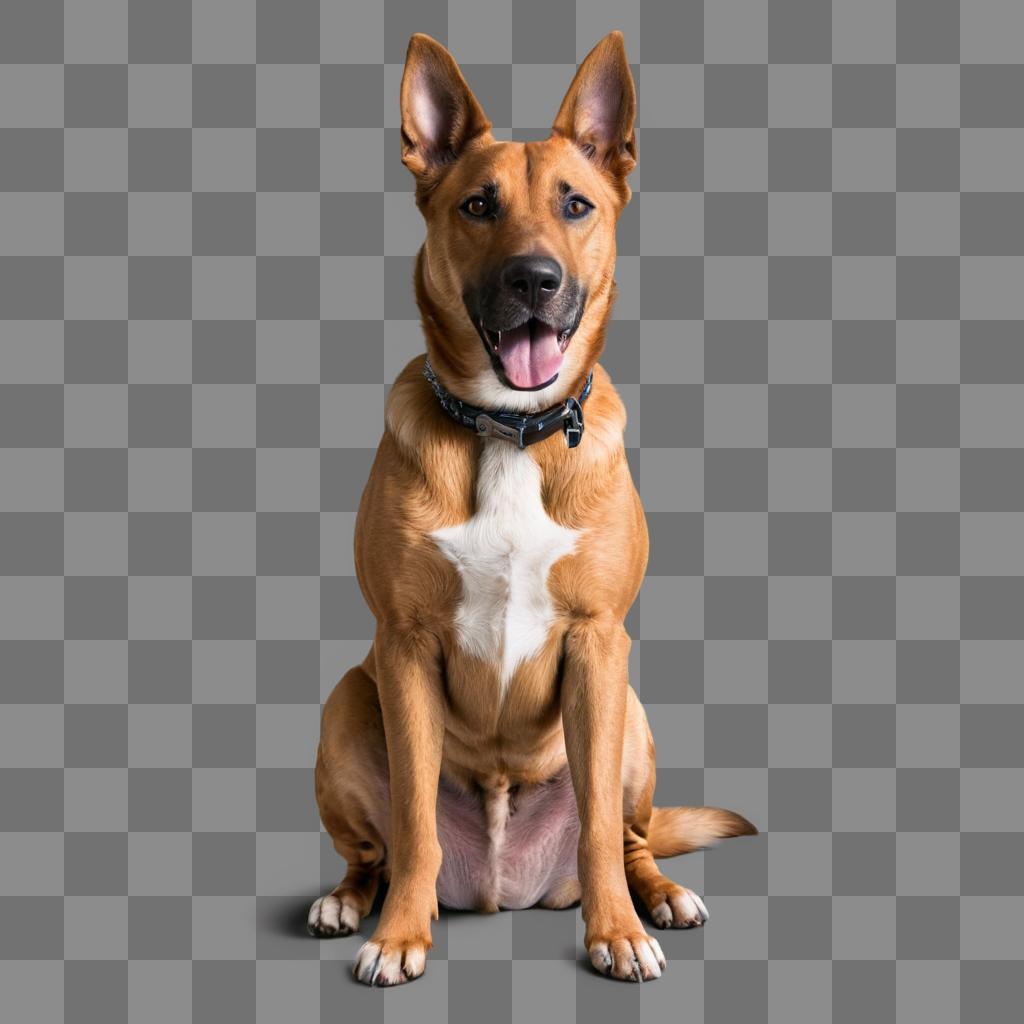}

\vspace{1pt}
\includegraphics[width=0.245\linewidth]{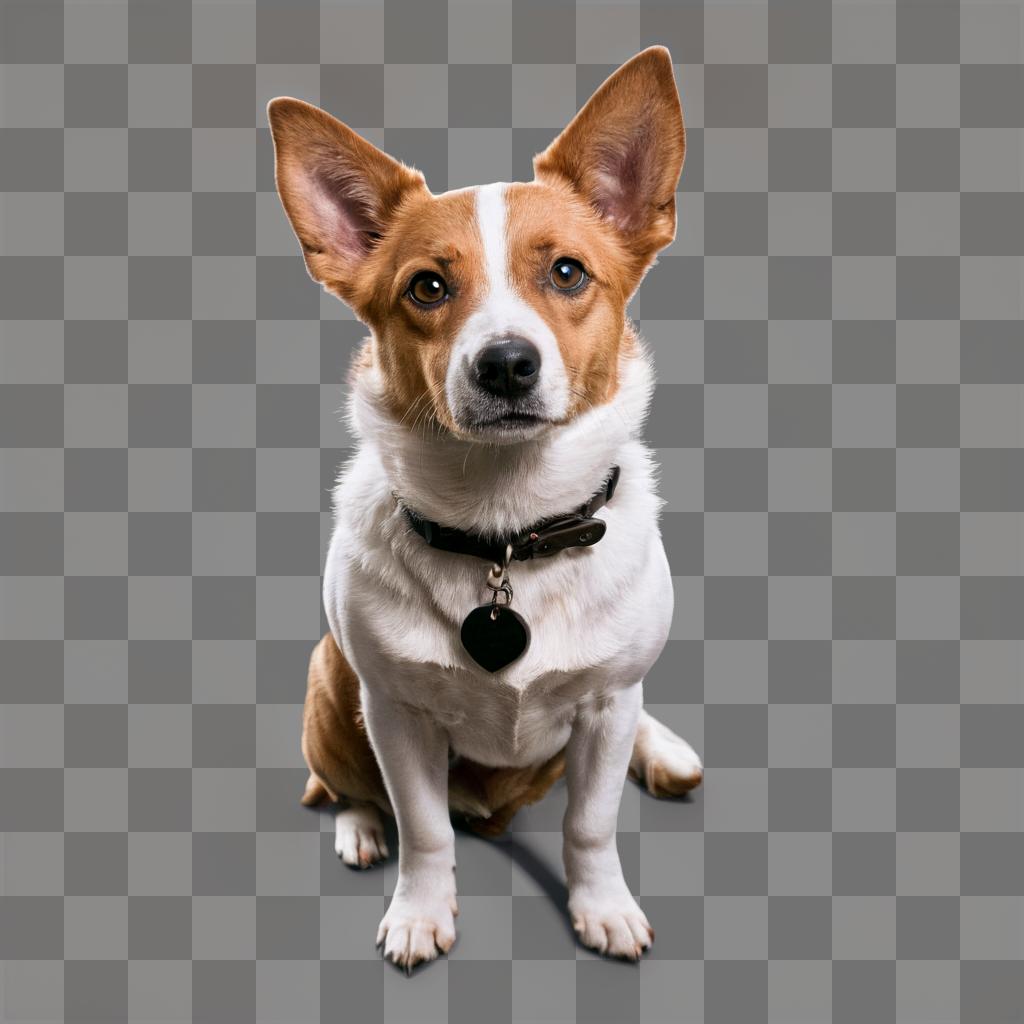}\hfill
\includegraphics[width=0.245\linewidth]{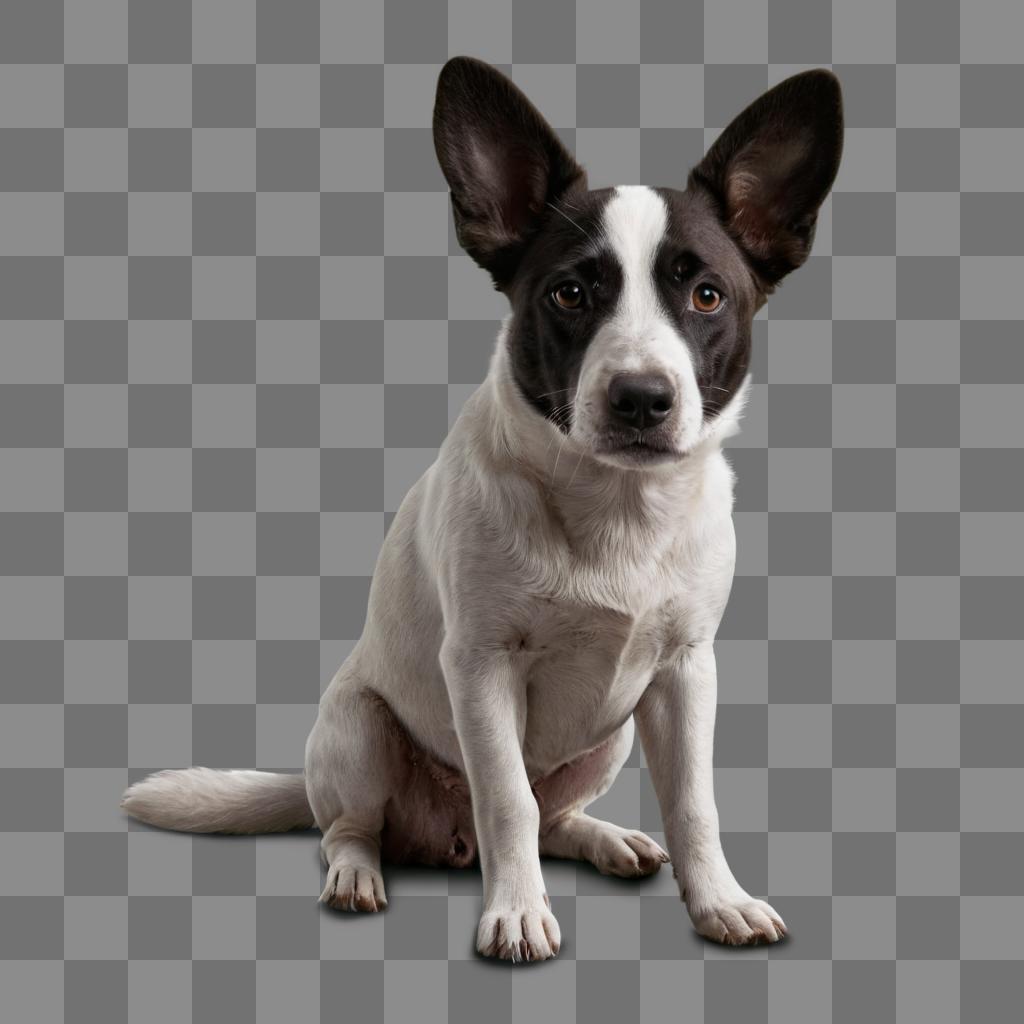}\hfill
\includegraphics[width=0.245\linewidth]{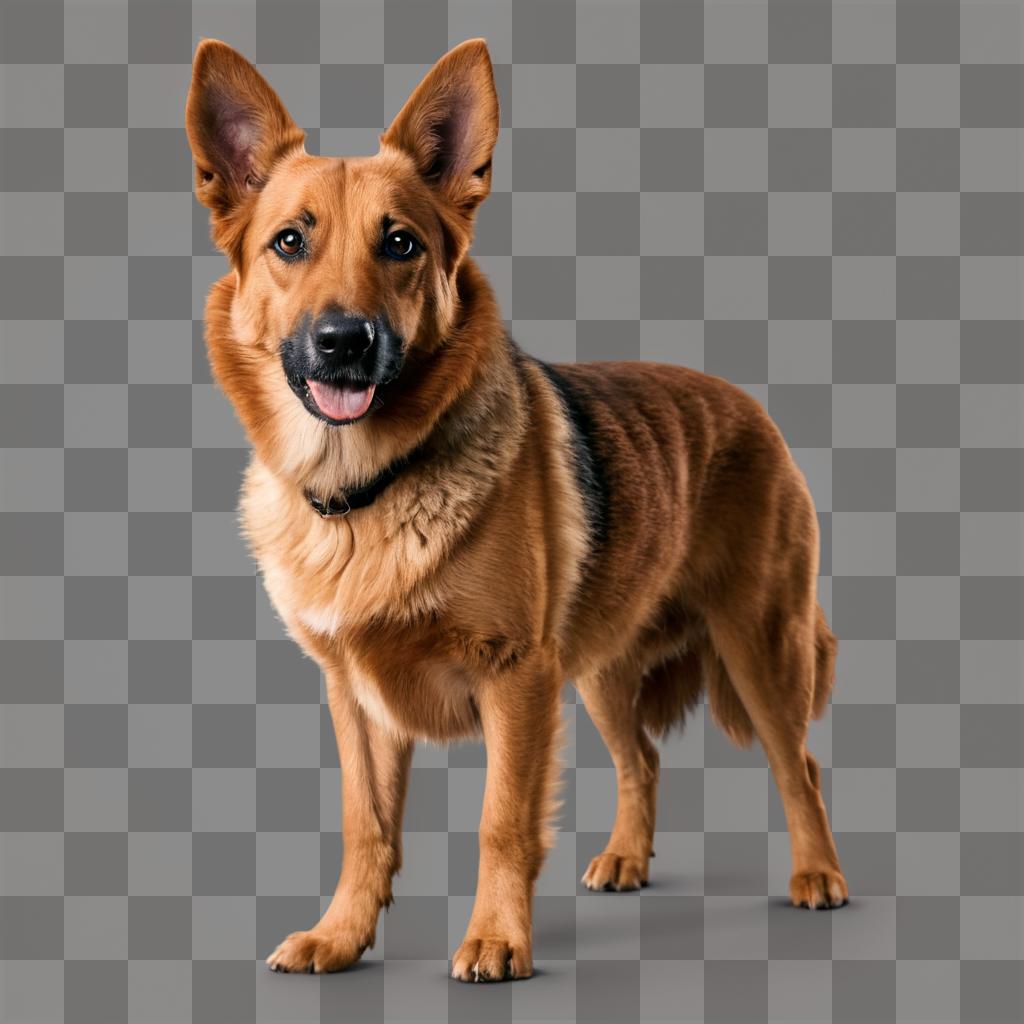}\hfill
\includegraphics[width=0.245\linewidth]{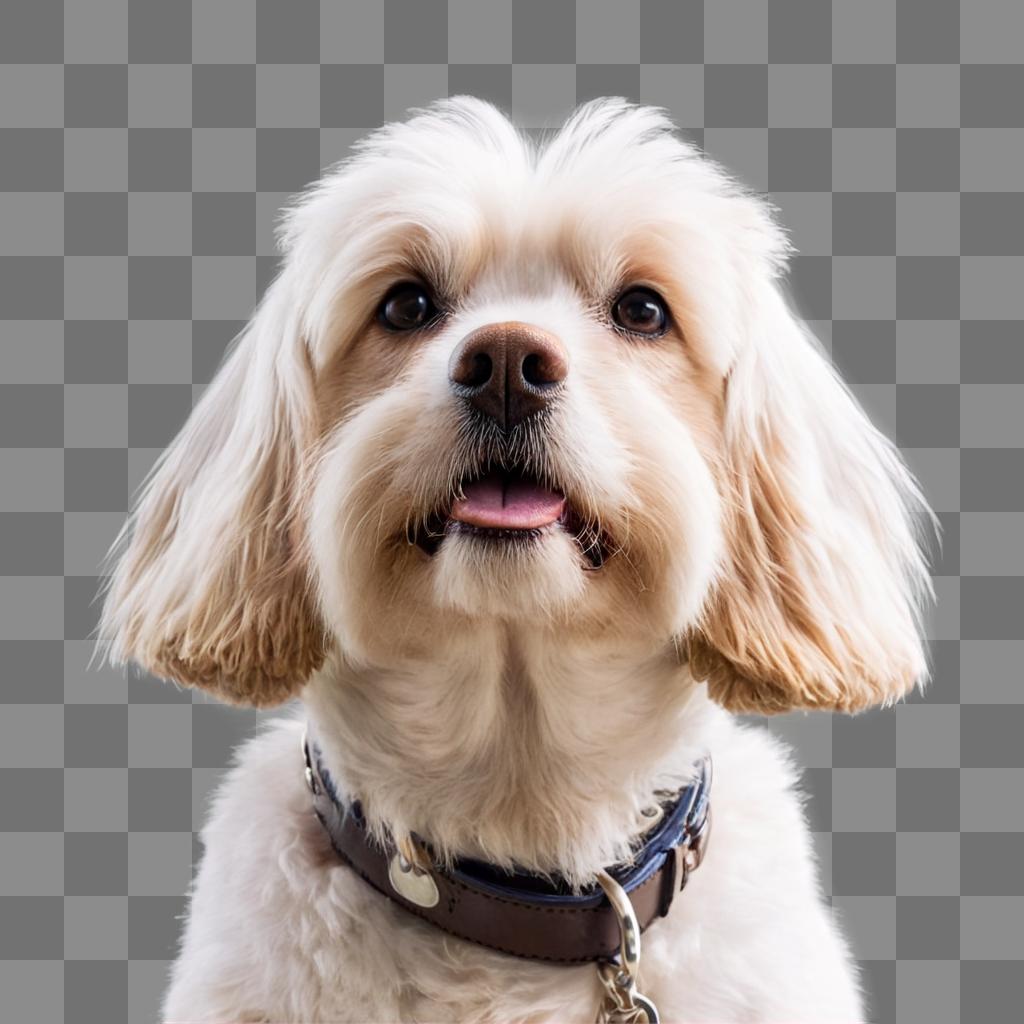}

\vspace{1pt}
\includegraphics[width=0.245\linewidth]{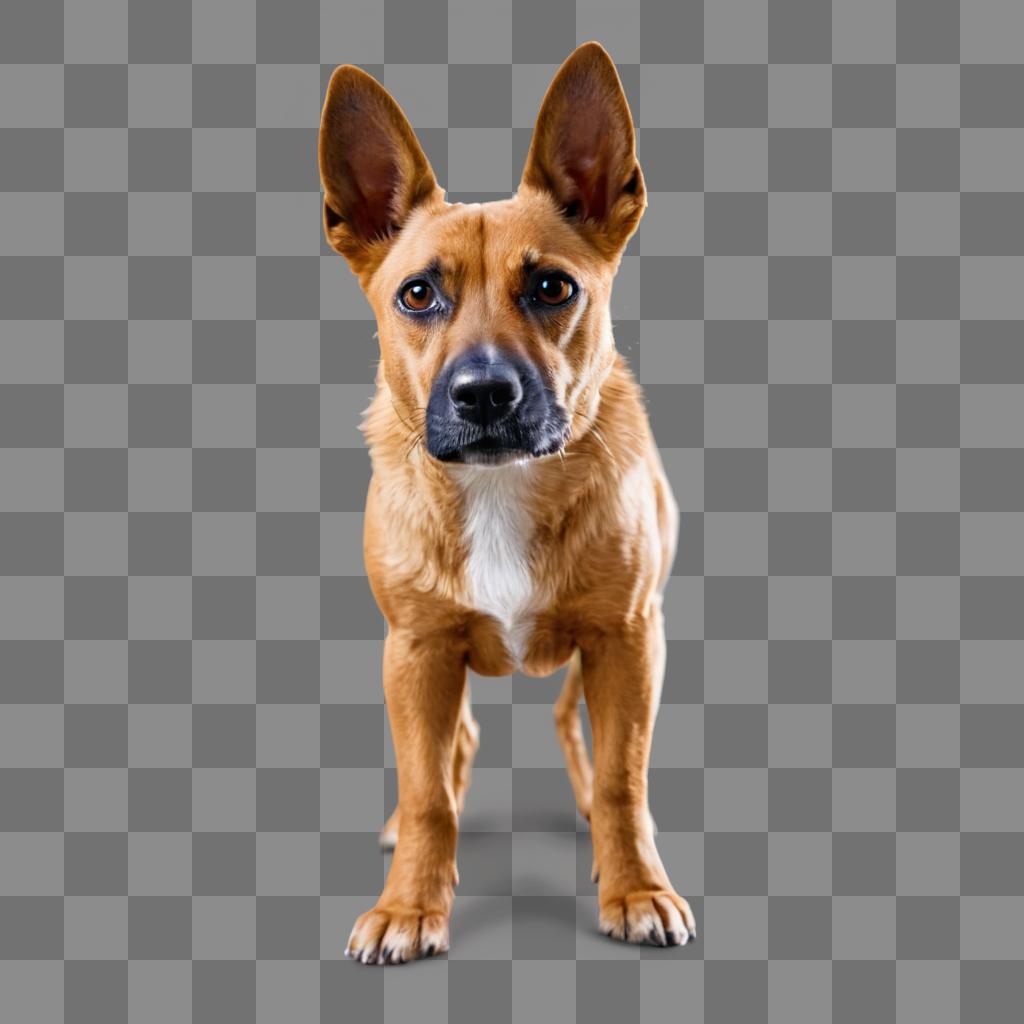}\hfill
\includegraphics[width=0.245\linewidth]{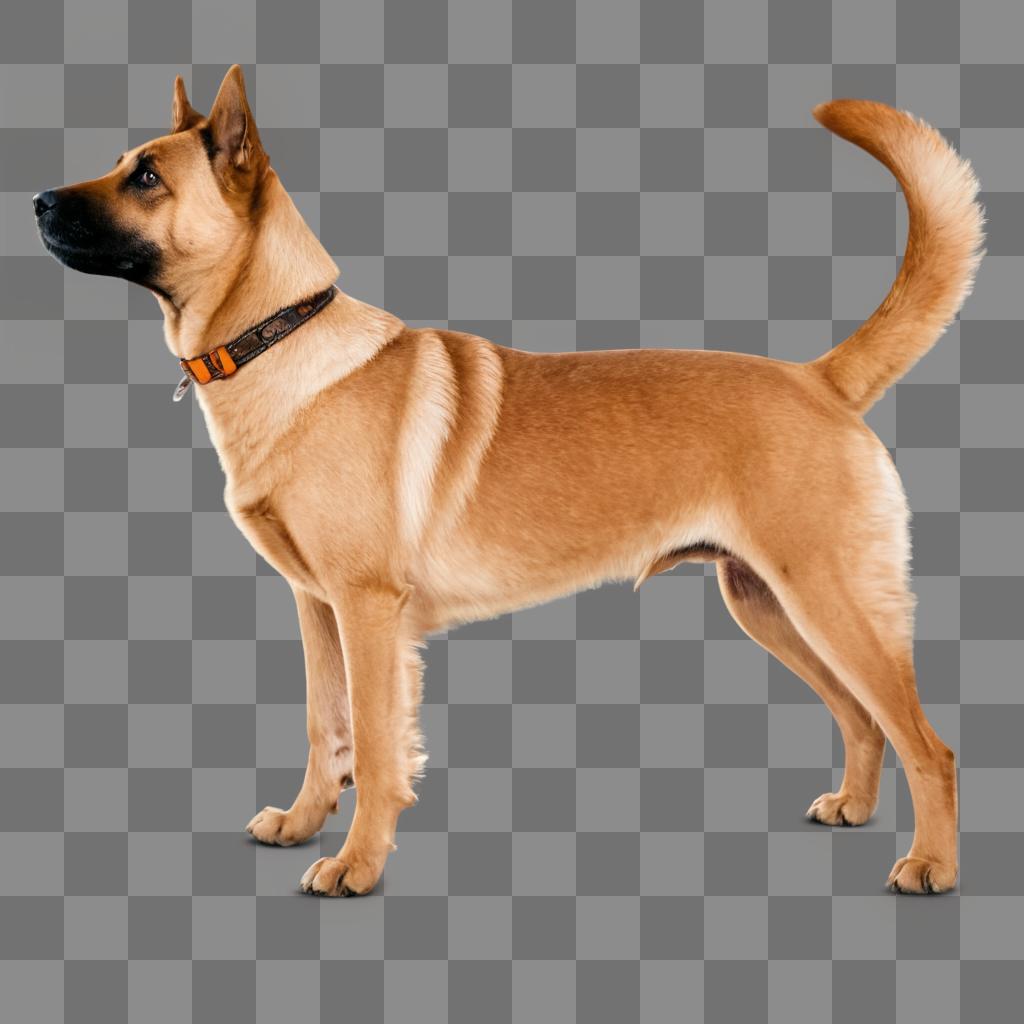}\hfill
\includegraphics[width=0.245\linewidth]{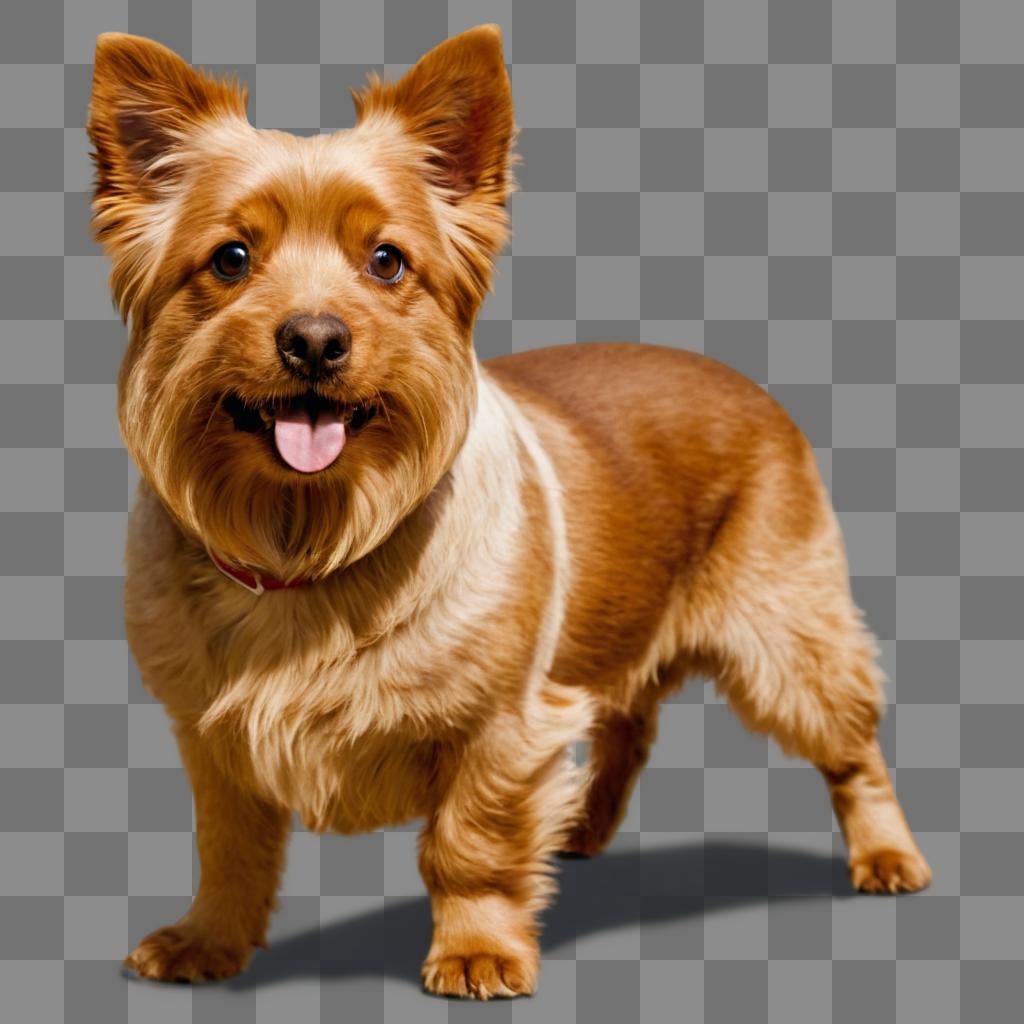}\hfill
\includegraphics[width=0.245\linewidth]{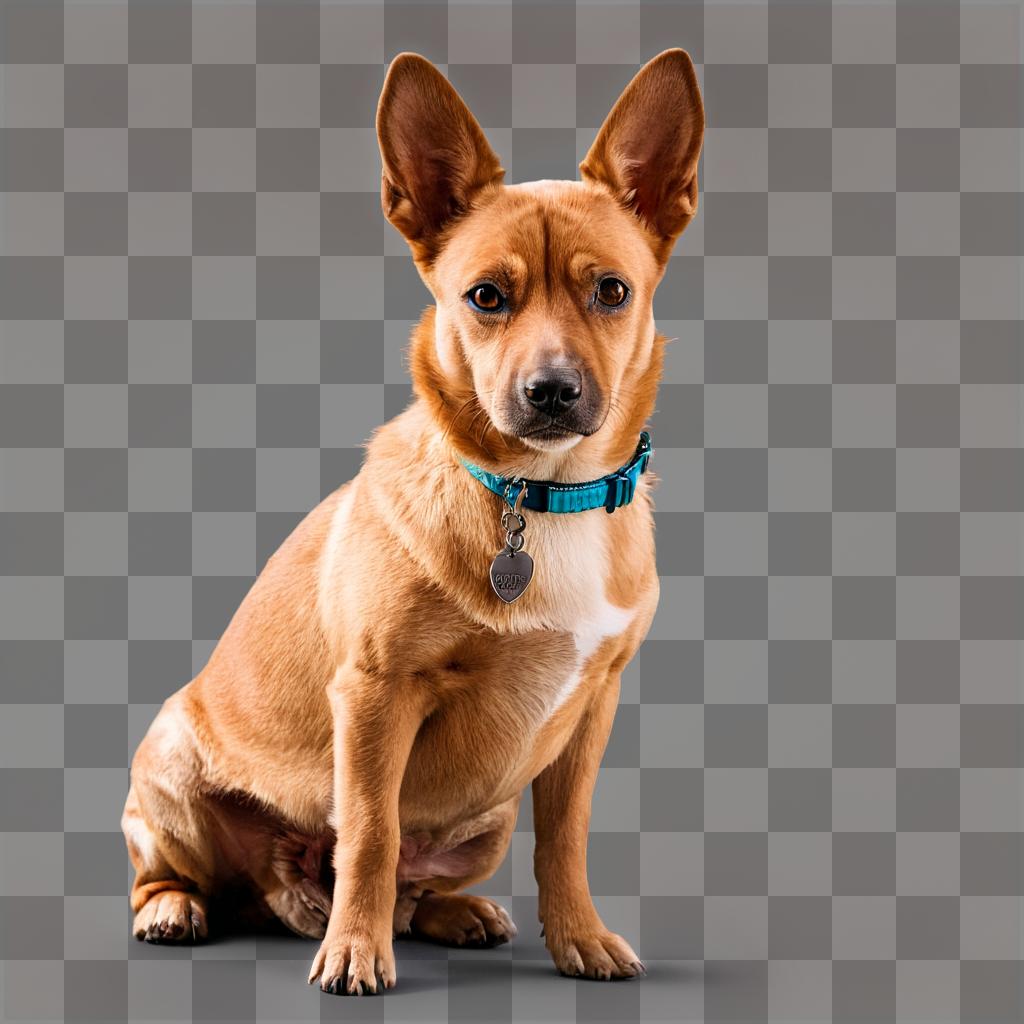}
\caption{Single Transparent Image Results \#7. The prompt is ``dog''. Resolution is $1024\times1024$.}
\label{fig:a7}
\end{minipage}
\end{figure*}

\begin{figure*}

\begin{minipage}{\linewidth}
\includegraphics[width=0.245\linewidth]{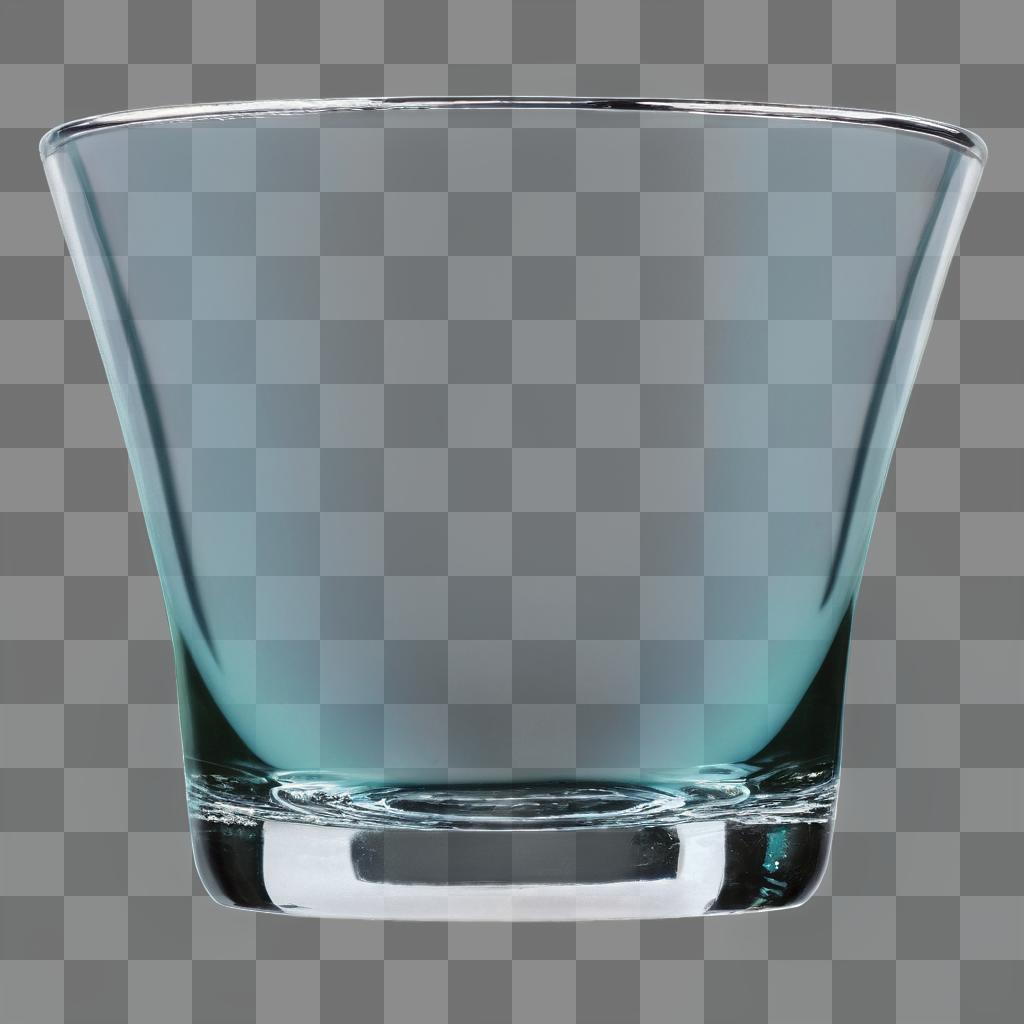}\hfill
\includegraphics[width=0.245\linewidth]{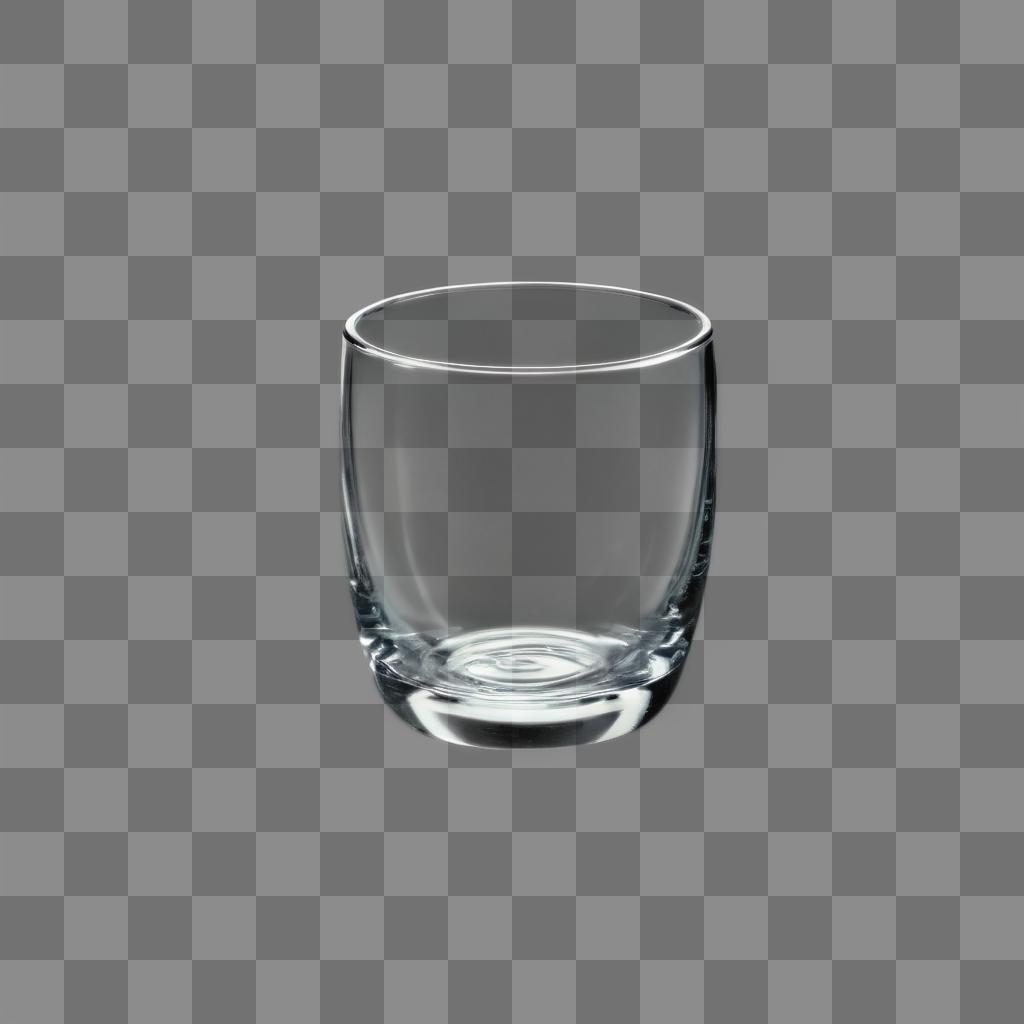}\hfill
\includegraphics[width=0.245\linewidth]{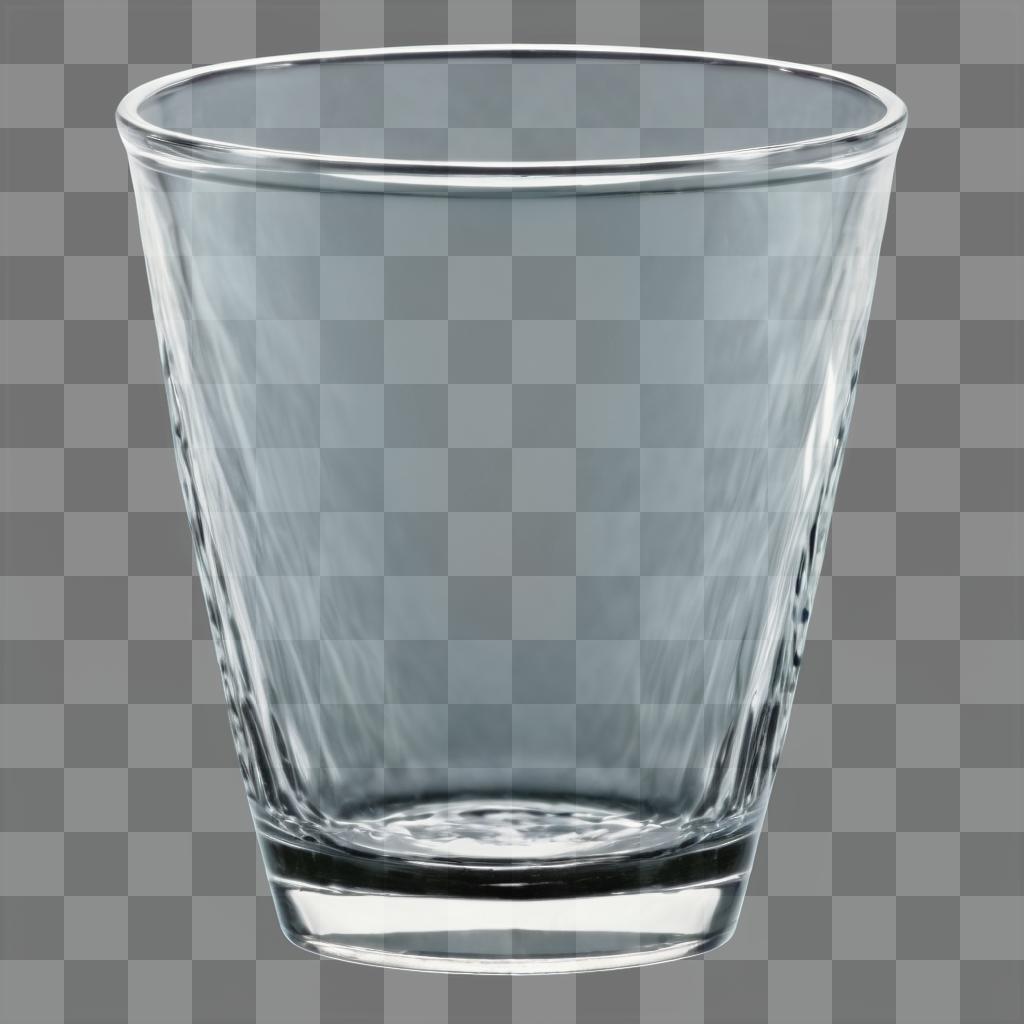}\hfill
\includegraphics[width=0.245\linewidth]{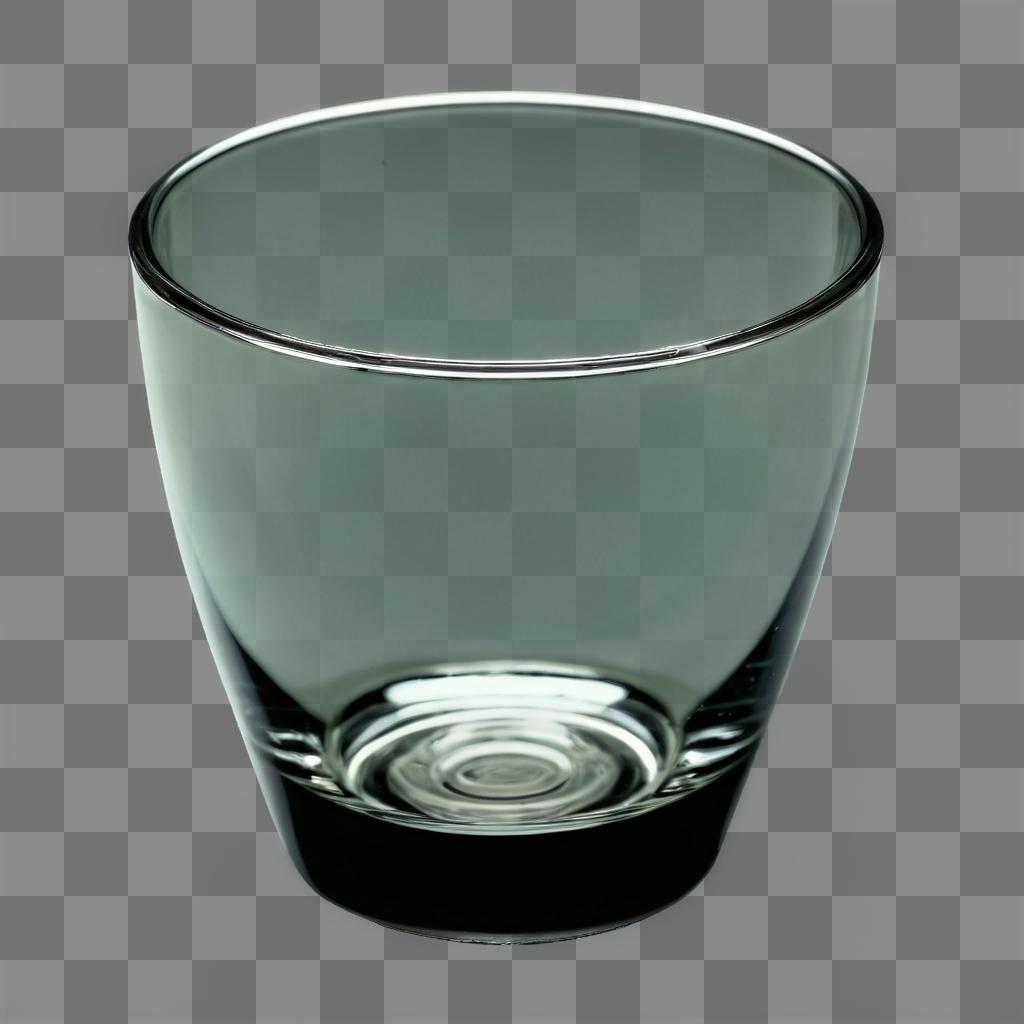}

\vspace{1pt}
\includegraphics[width=0.245\linewidth]{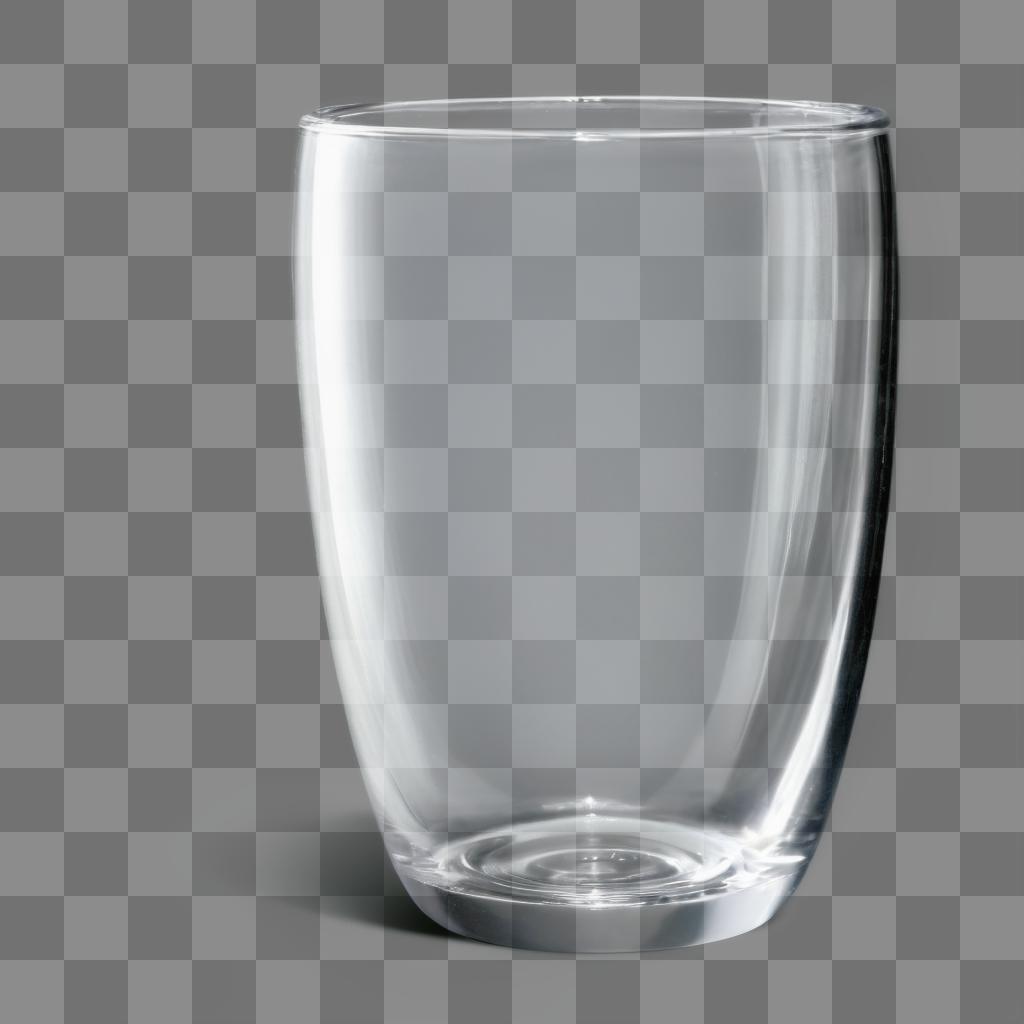}\hfill
\includegraphics[width=0.245\linewidth]{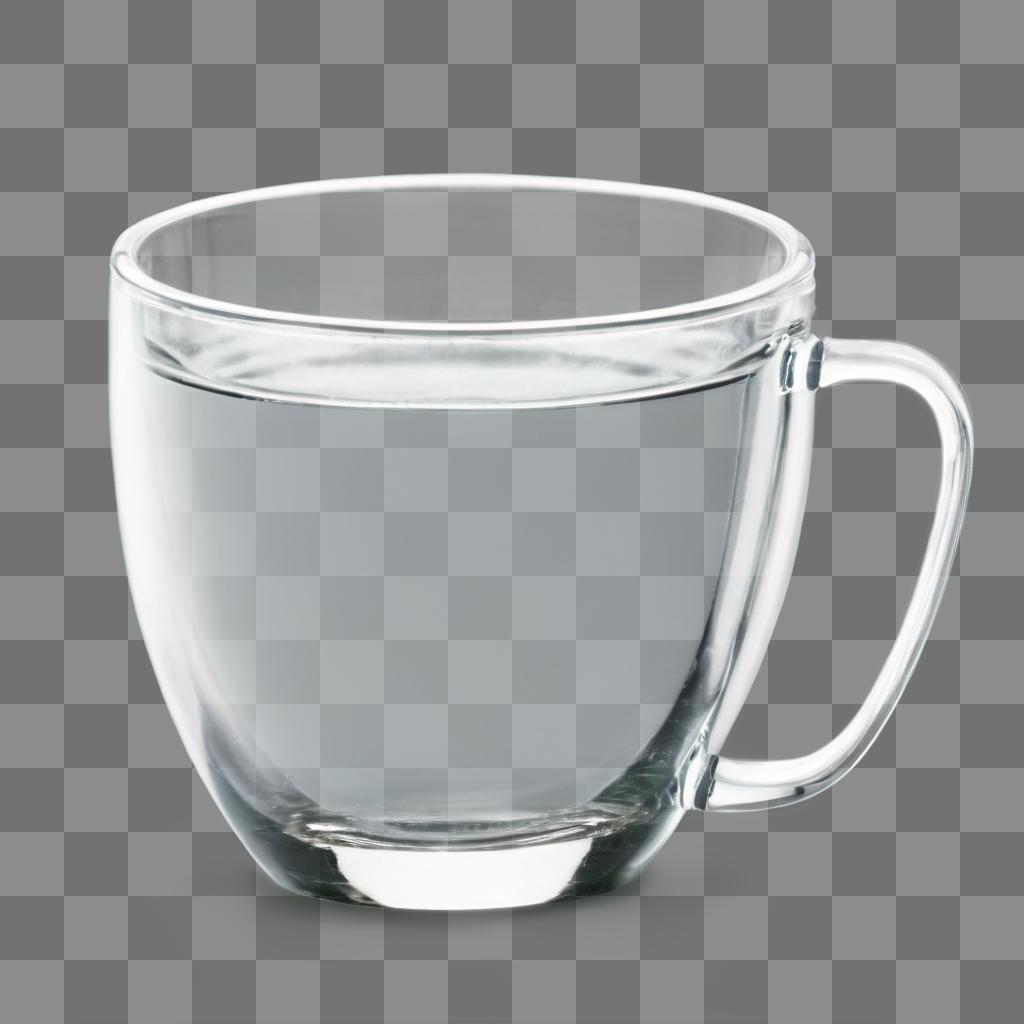}\hfill
\includegraphics[width=0.245\linewidth]{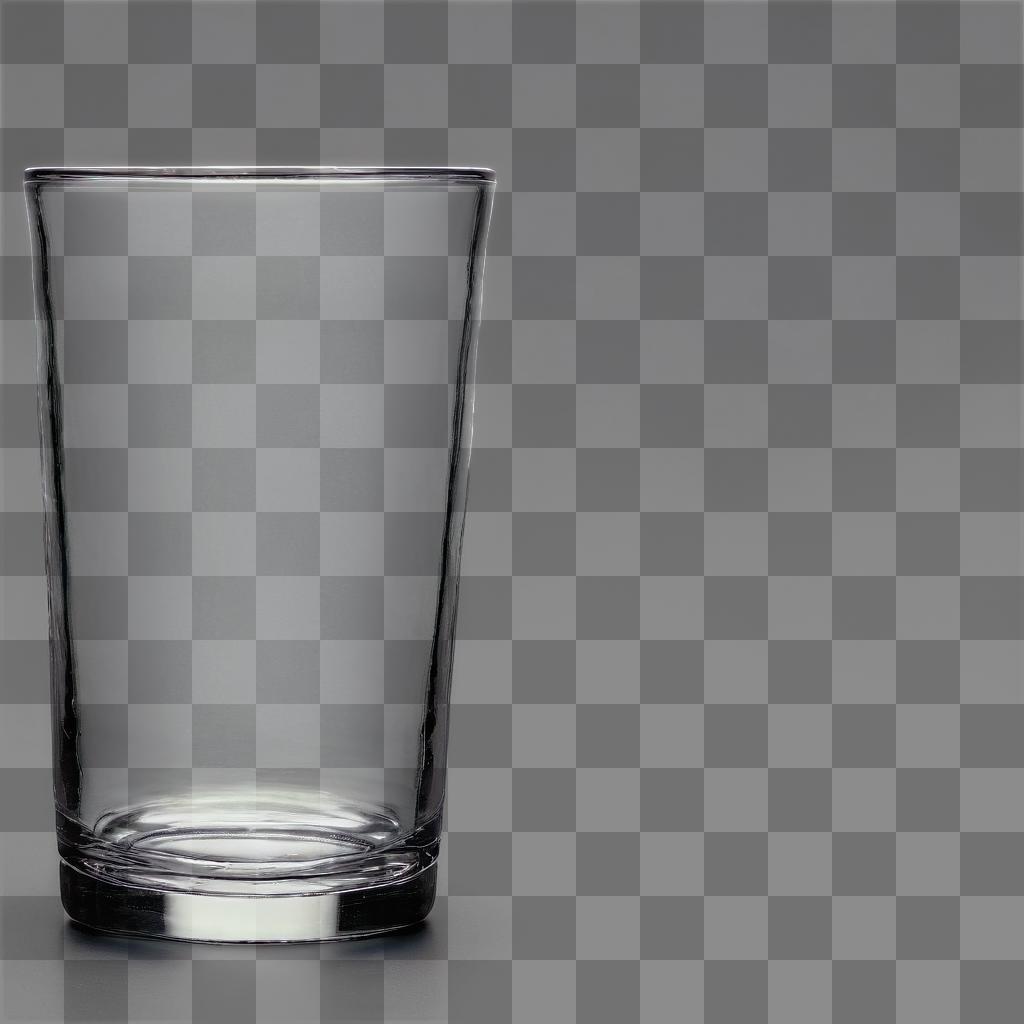}\hfill
\includegraphics[width=0.245\linewidth]{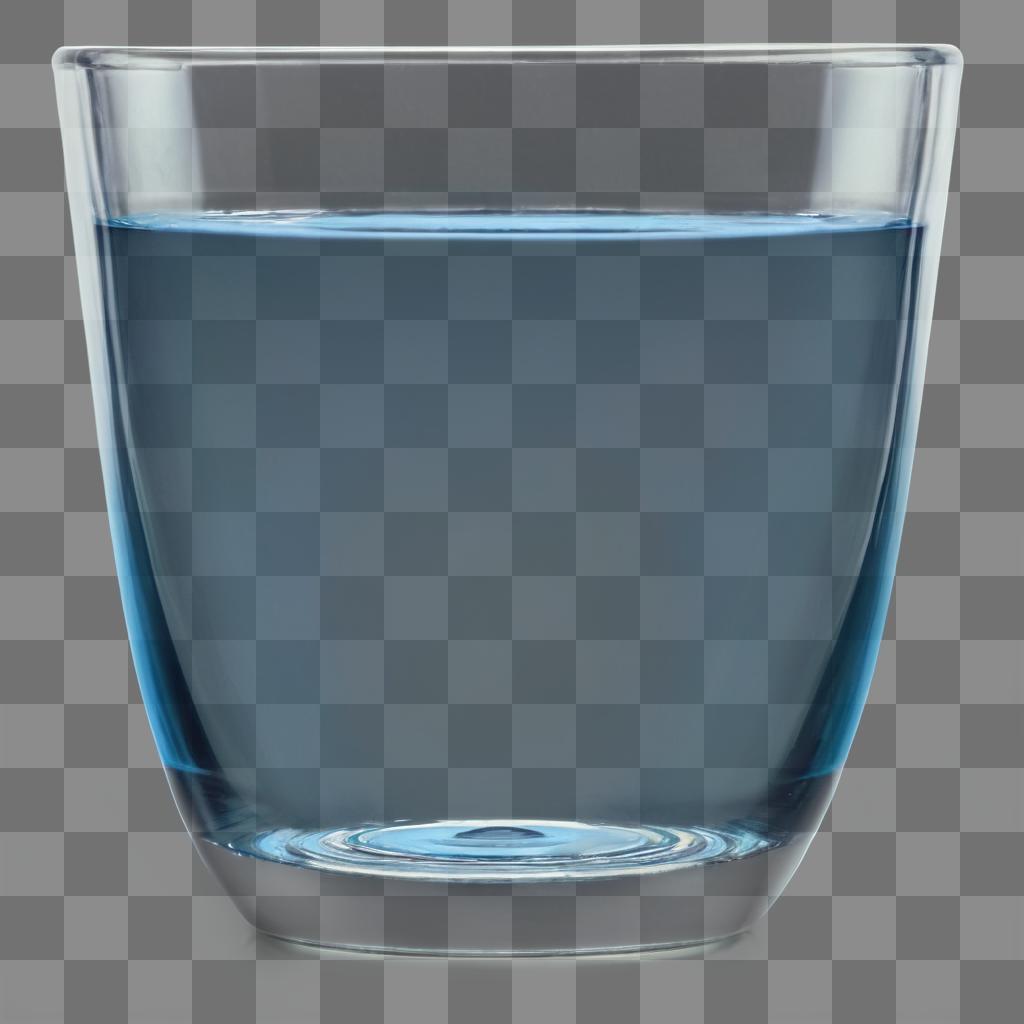}

\vspace{1pt}
\includegraphics[width=0.245\linewidth]{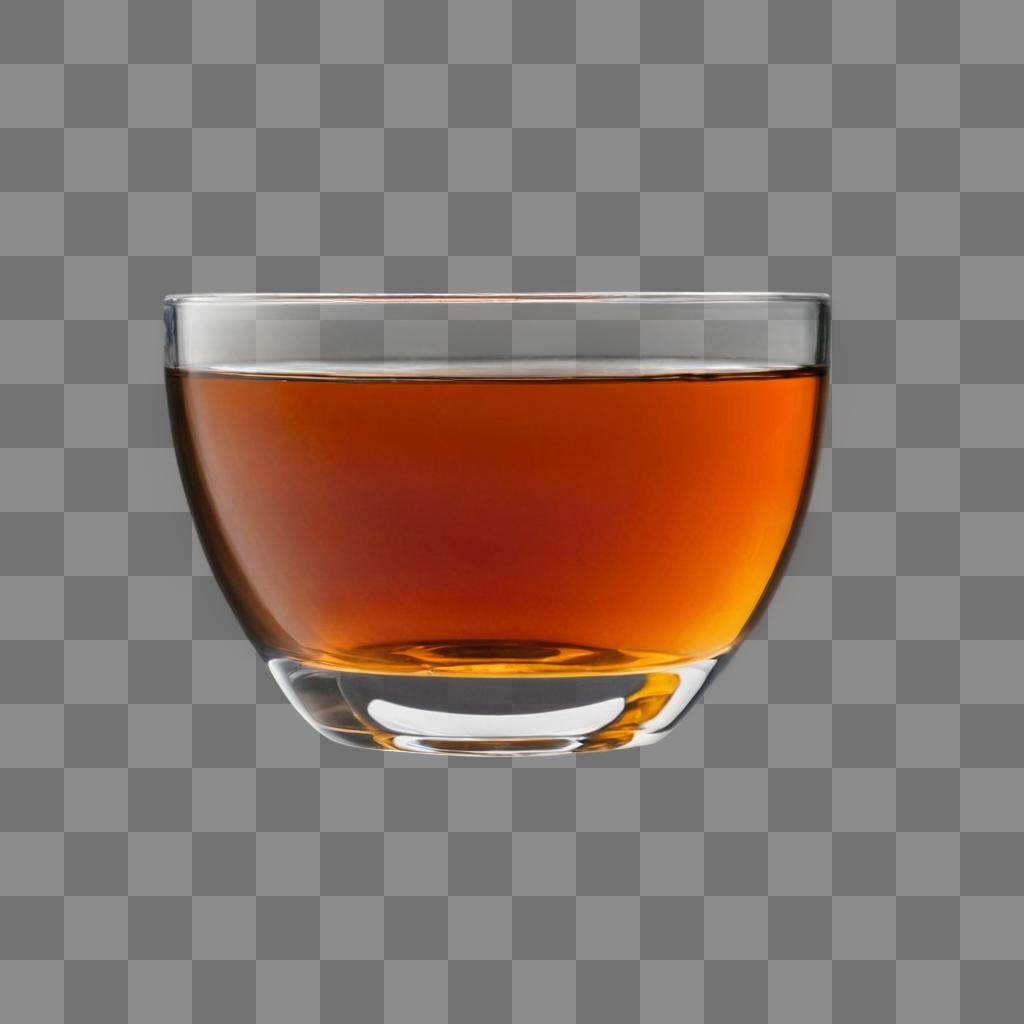}\hfill
\includegraphics[width=0.245\linewidth]{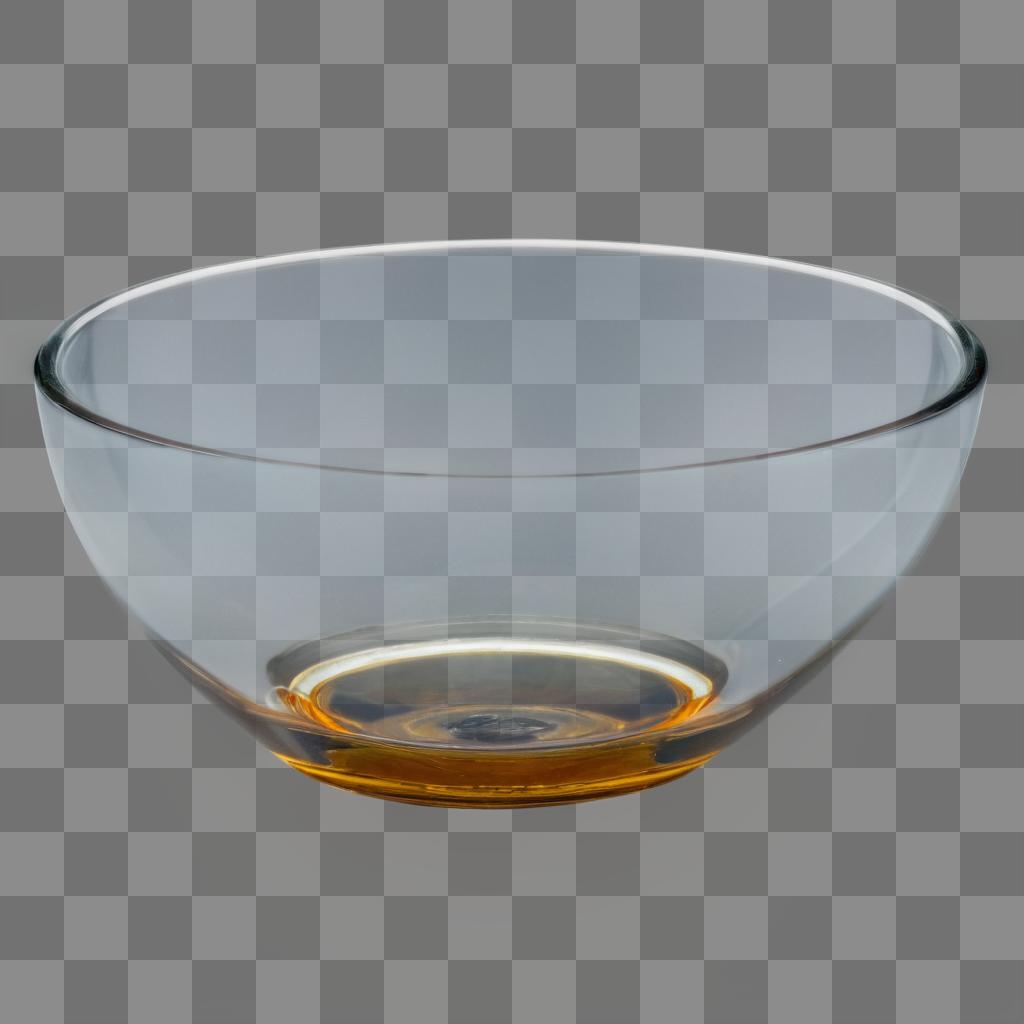}\hfill
\includegraphics[width=0.245\linewidth]{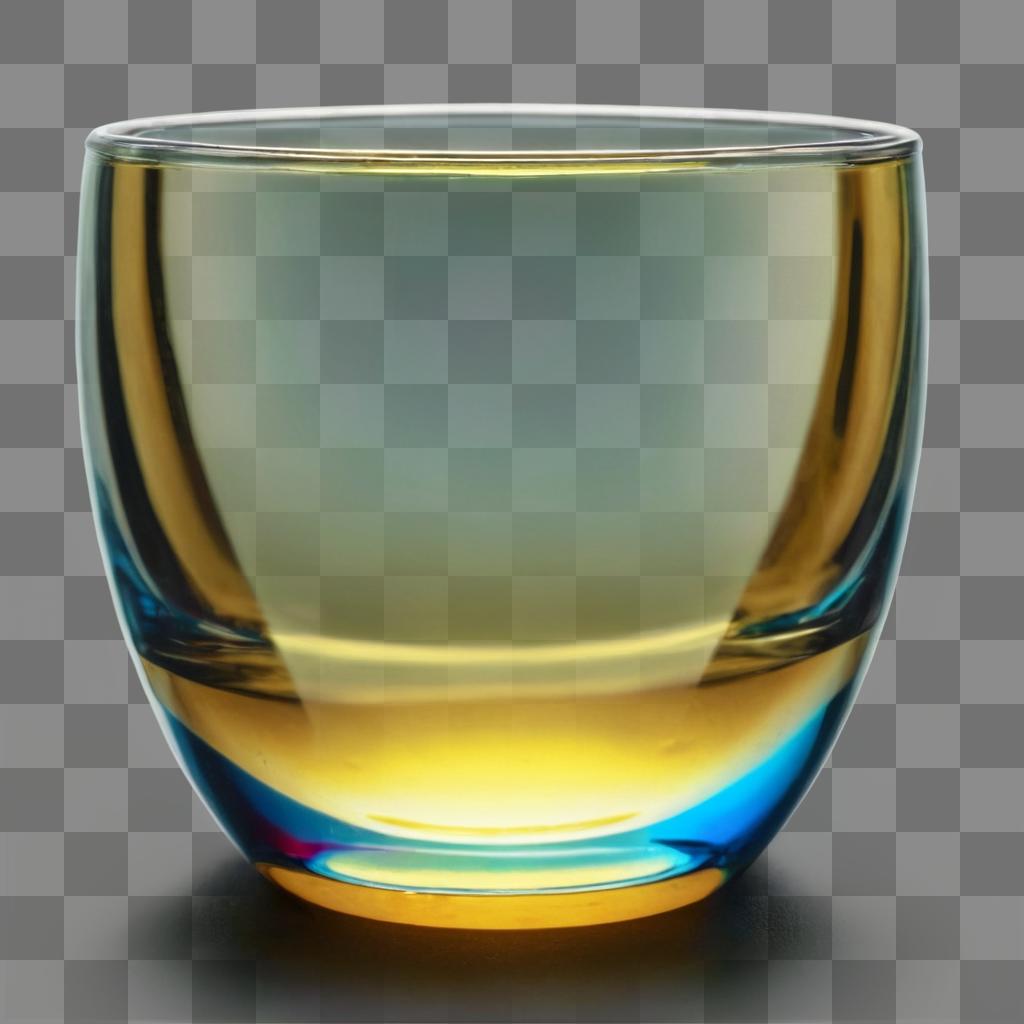}\hfill
\includegraphics[width=0.245\linewidth]{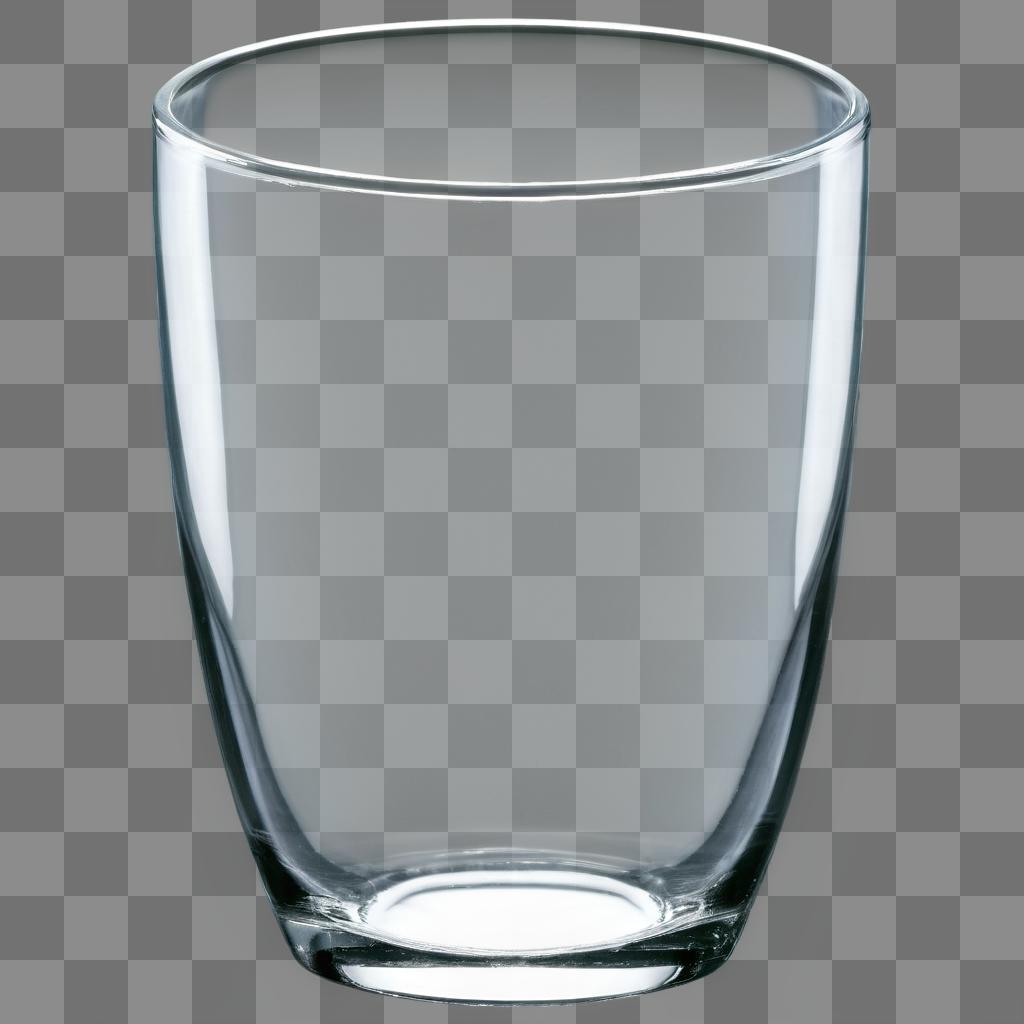}
\caption{Single Transparent Image Results \#8. The prompt is ``glass cup''. Resolution is $1024\times1024$.}
\label{fig:a8}
\end{minipage}
\end{figure*}

\begin{figure*}

\begin{minipage}{\linewidth}
\includegraphics[width=0.33\linewidth]{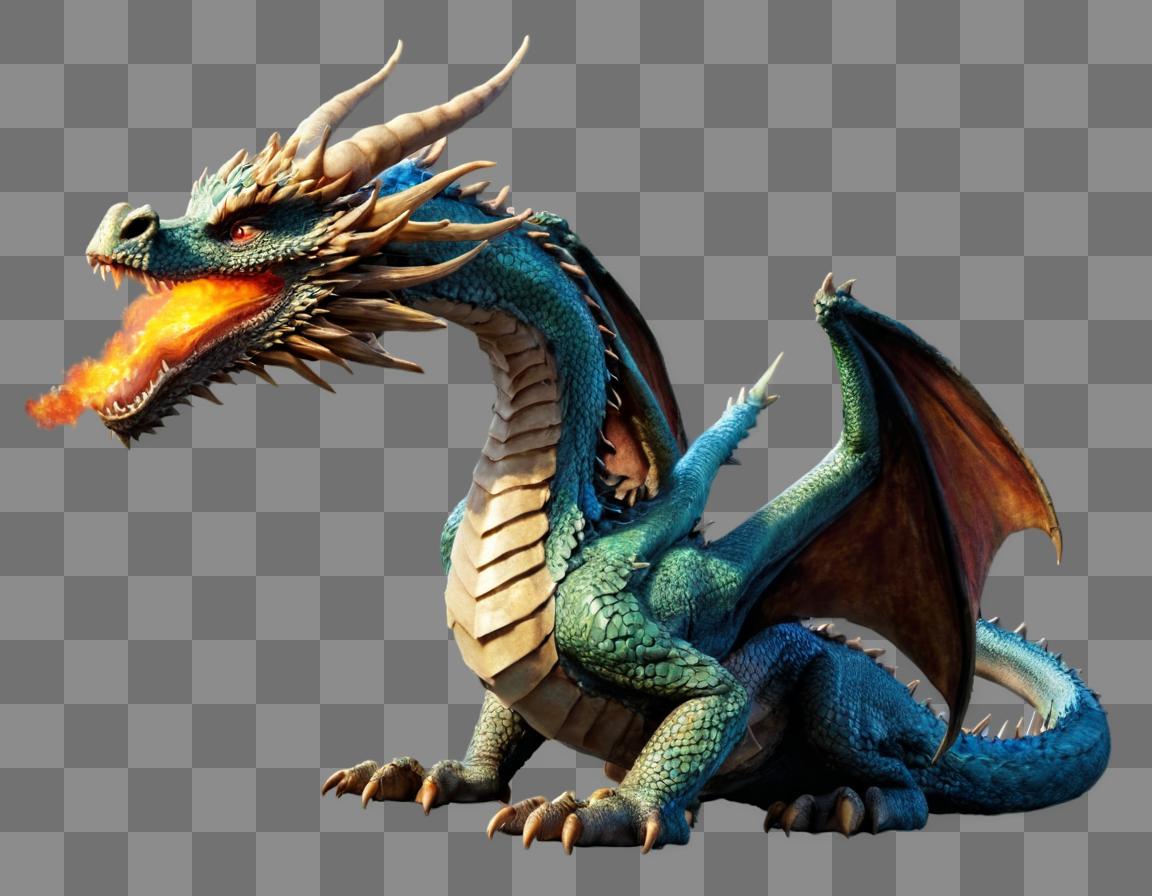}\hfill
\includegraphics[width=0.33\linewidth]{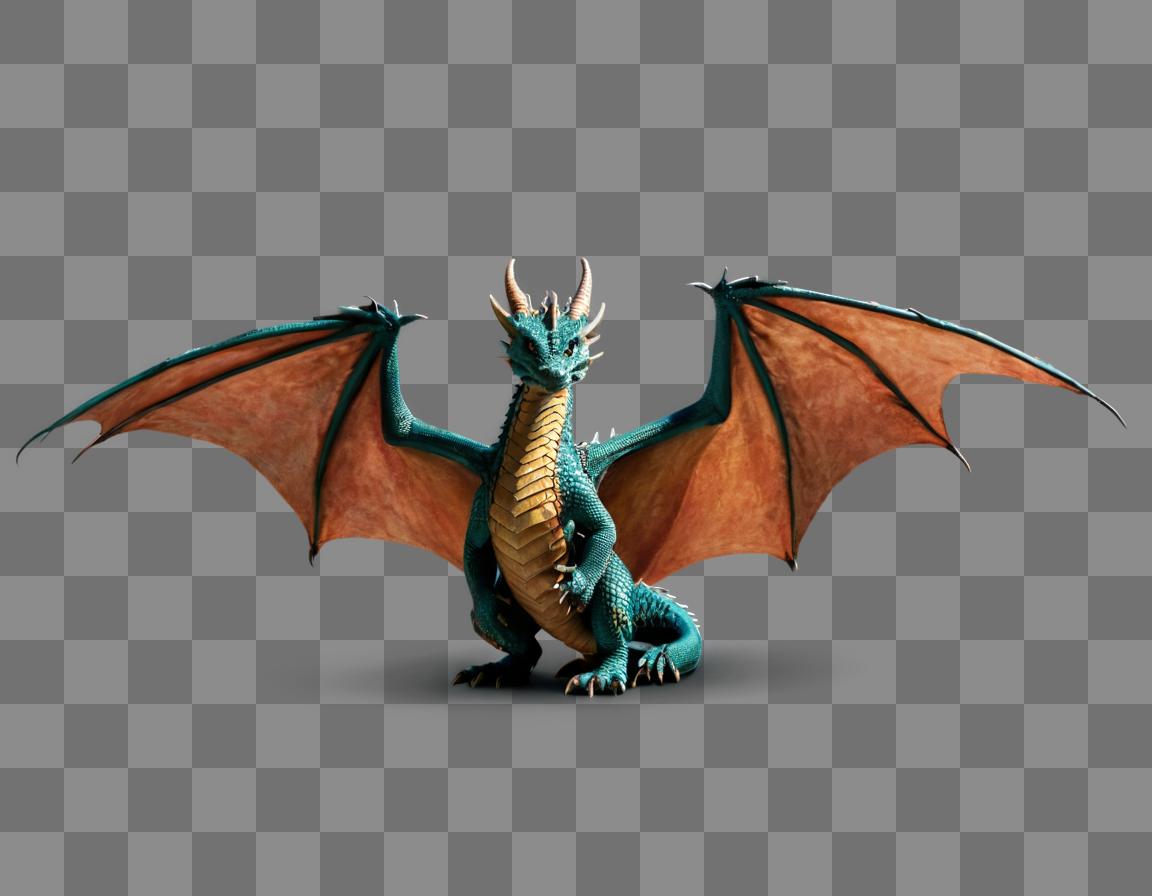}\hfill
\includegraphics[width=0.33\linewidth]{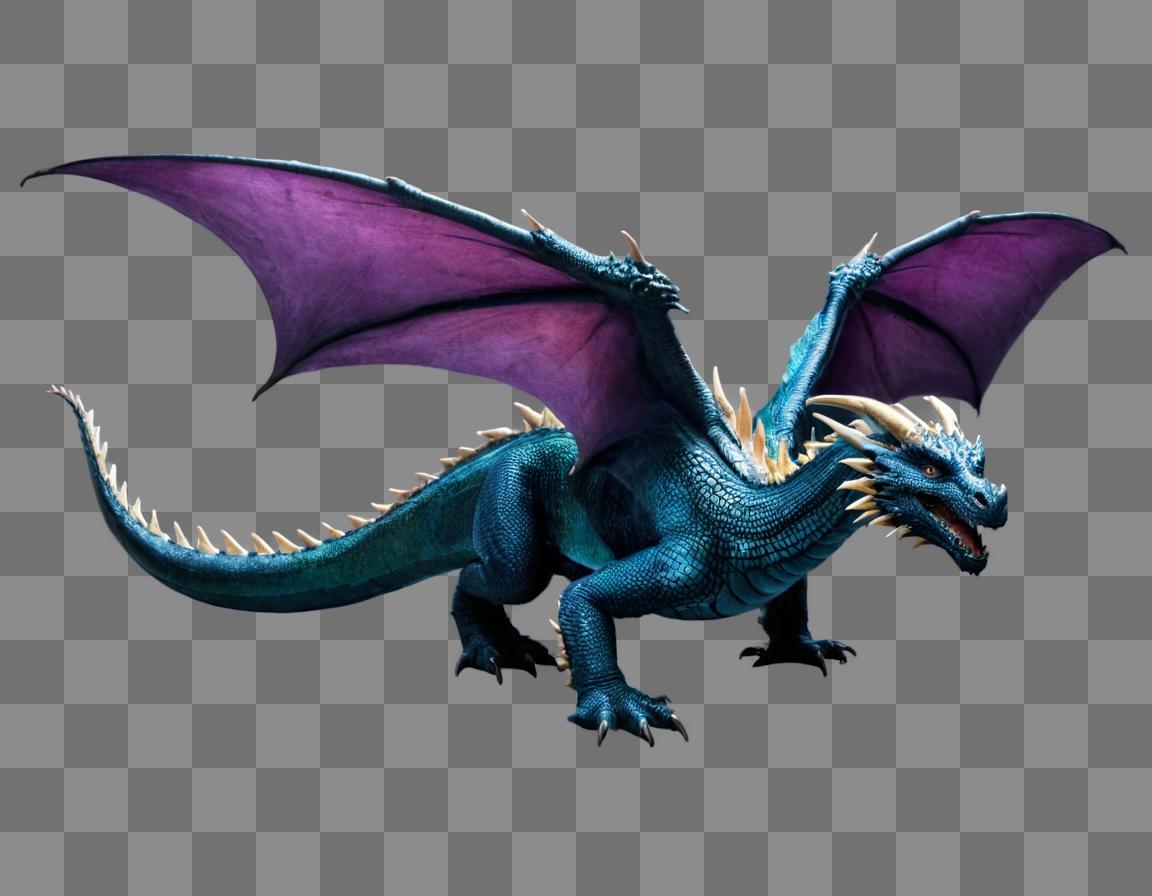}

\vspace{1pt}
\includegraphics[width=0.33\linewidth]{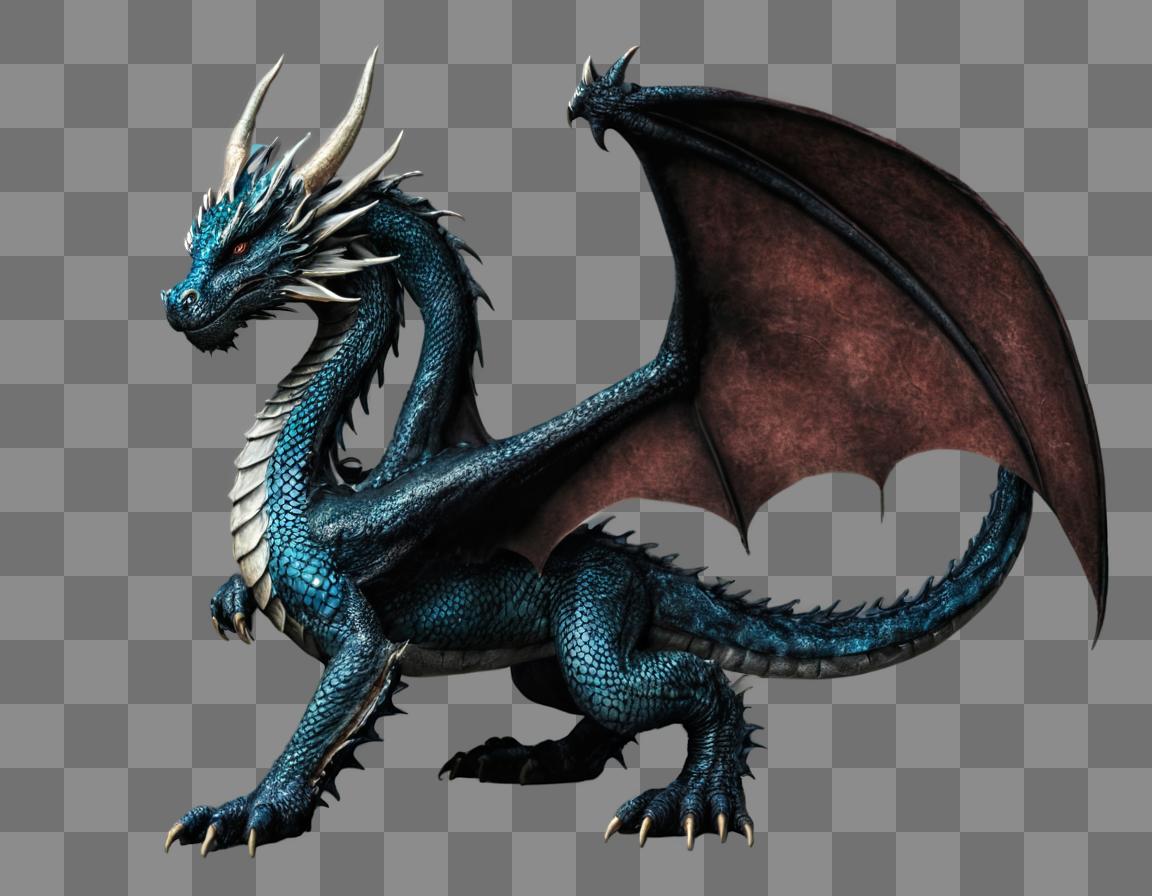}\hfill
\includegraphics[width=0.33\linewidth]{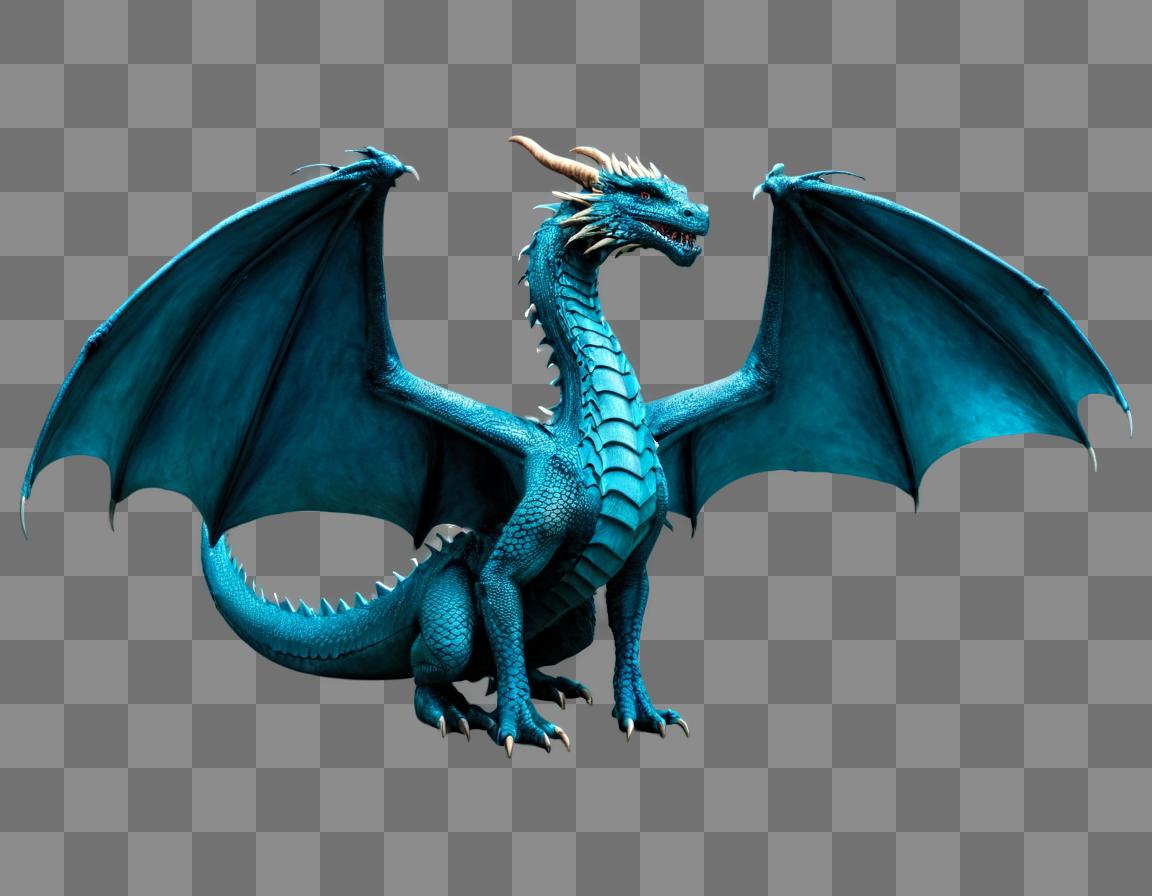}\hfill
\includegraphics[width=0.33\linewidth]{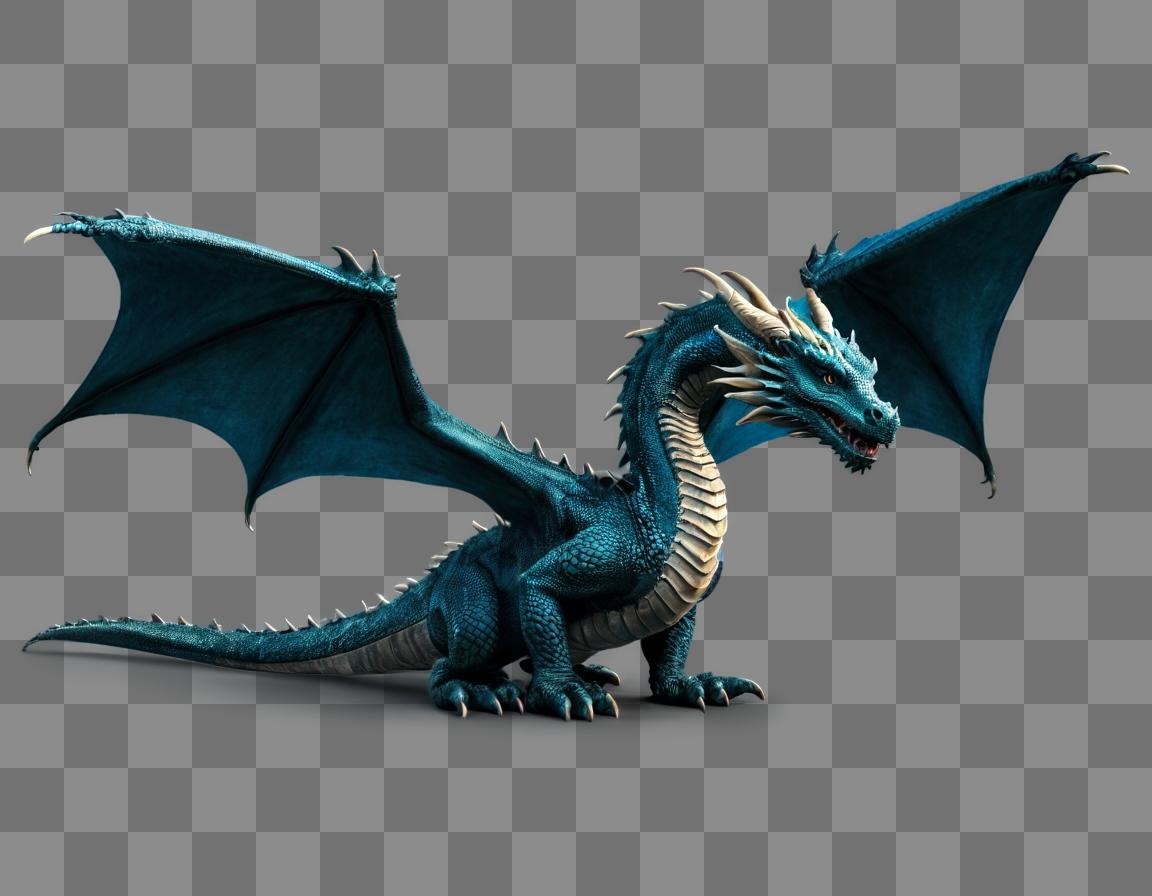}

\vspace{1pt}
\includegraphics[width=0.33\linewidth]{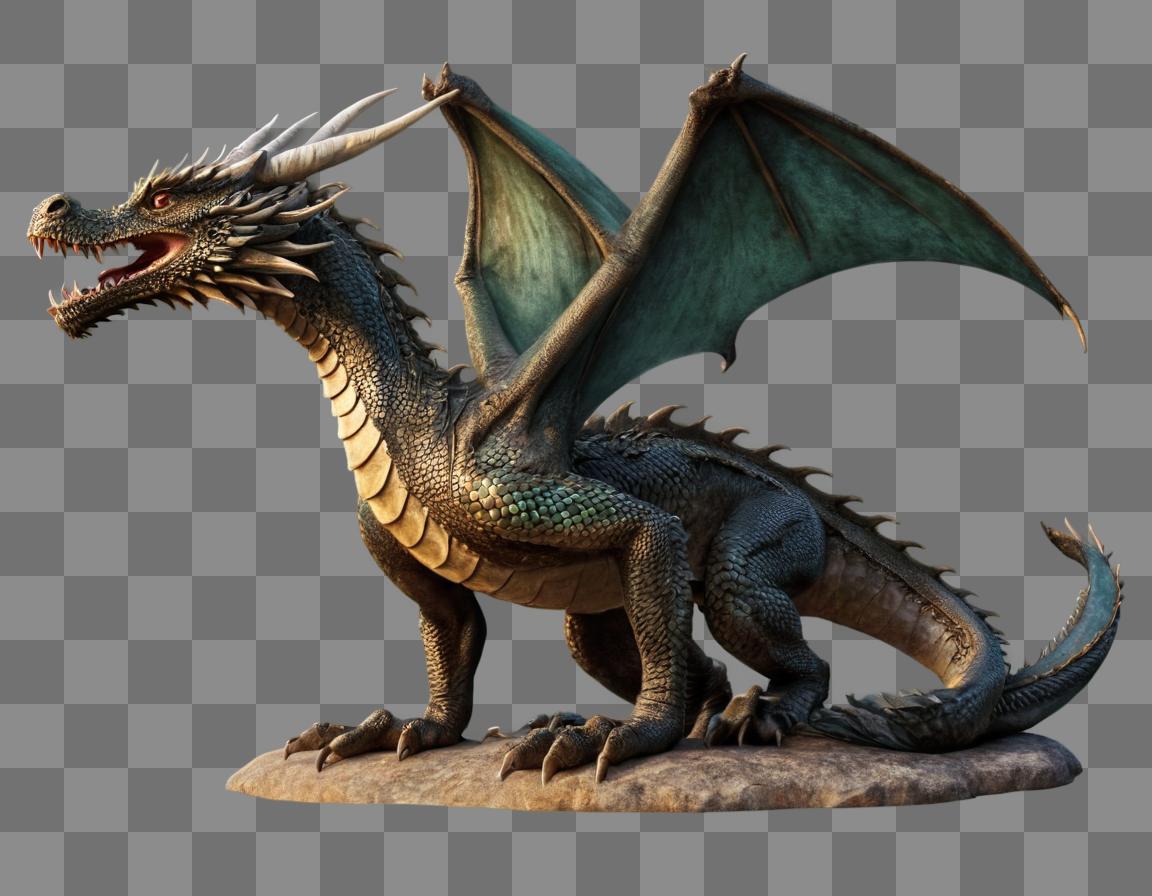}\hfill
\includegraphics[width=0.33\linewidth]{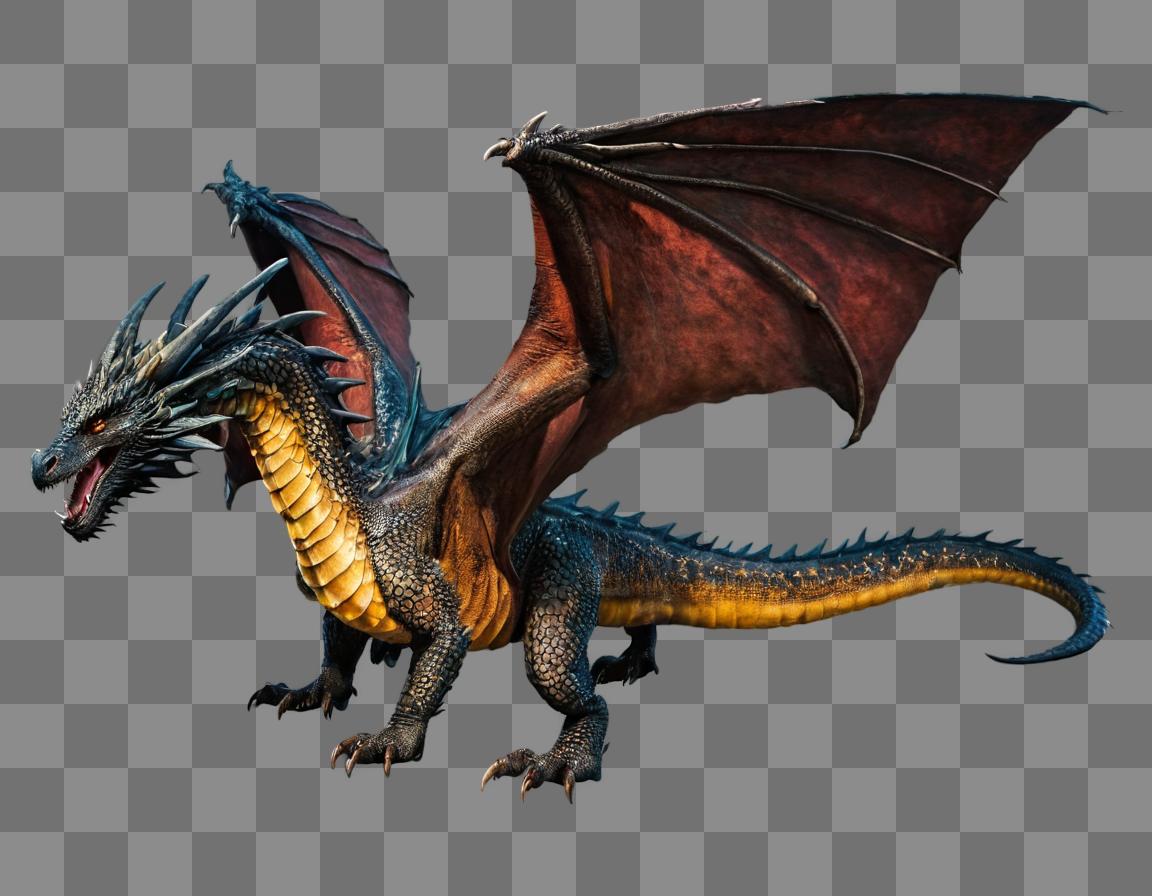}\hfill
\includegraphics[width=0.33\linewidth]{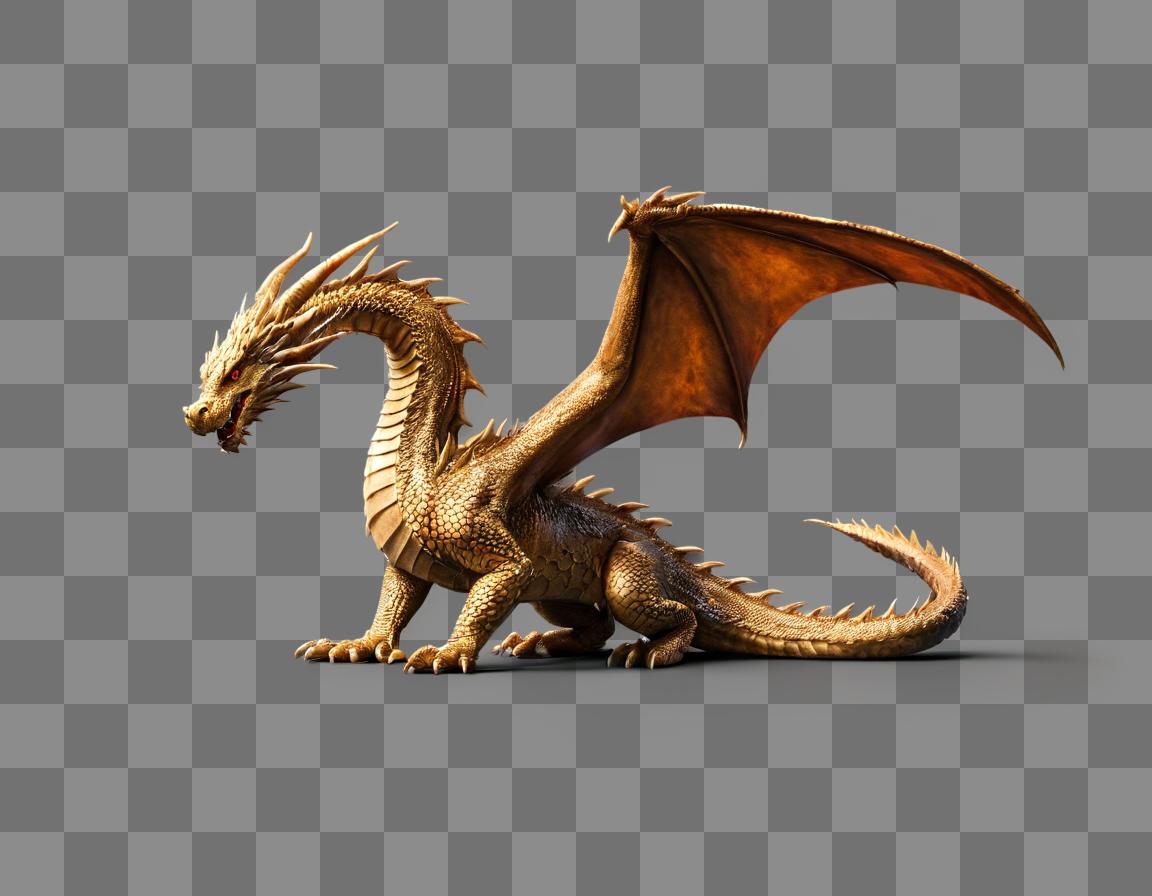}

\vspace{1pt}
\includegraphics[width=0.33\linewidth]{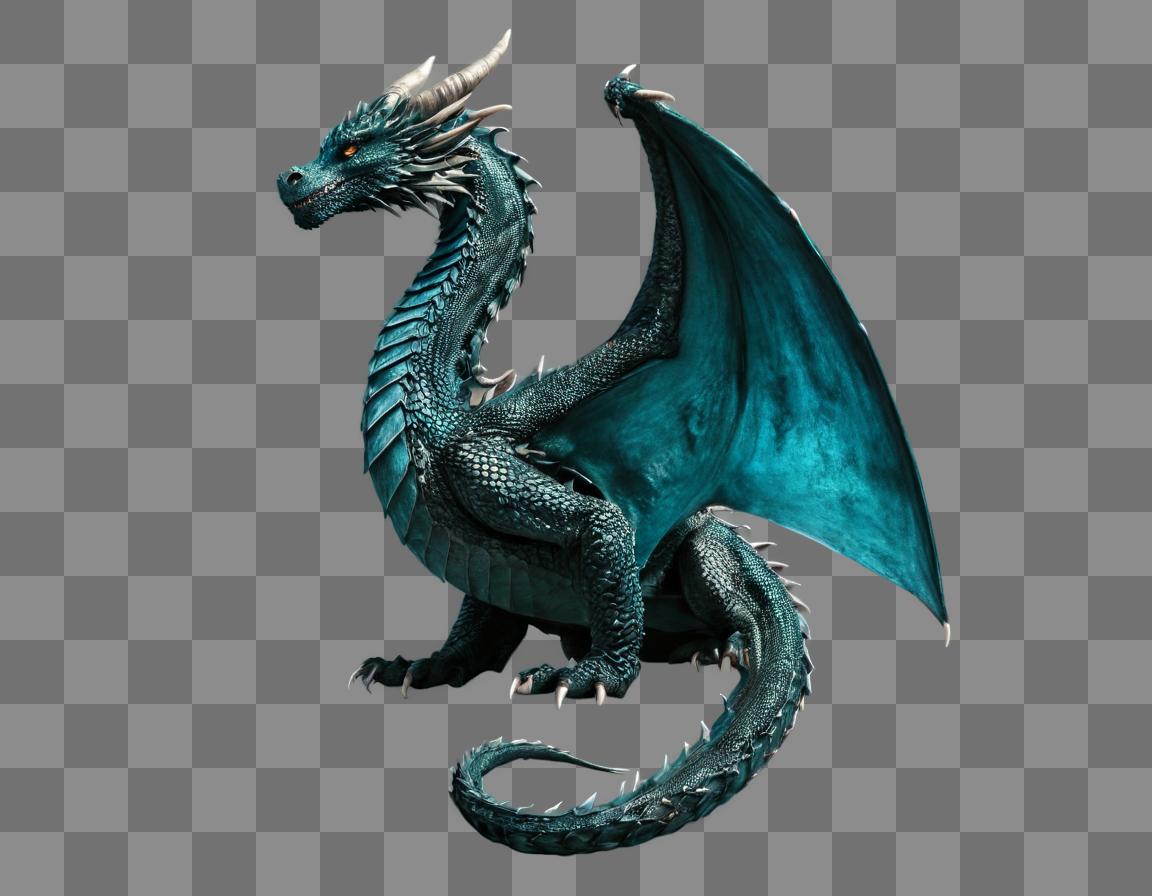}\hfill
\includegraphics[width=0.33\linewidth]{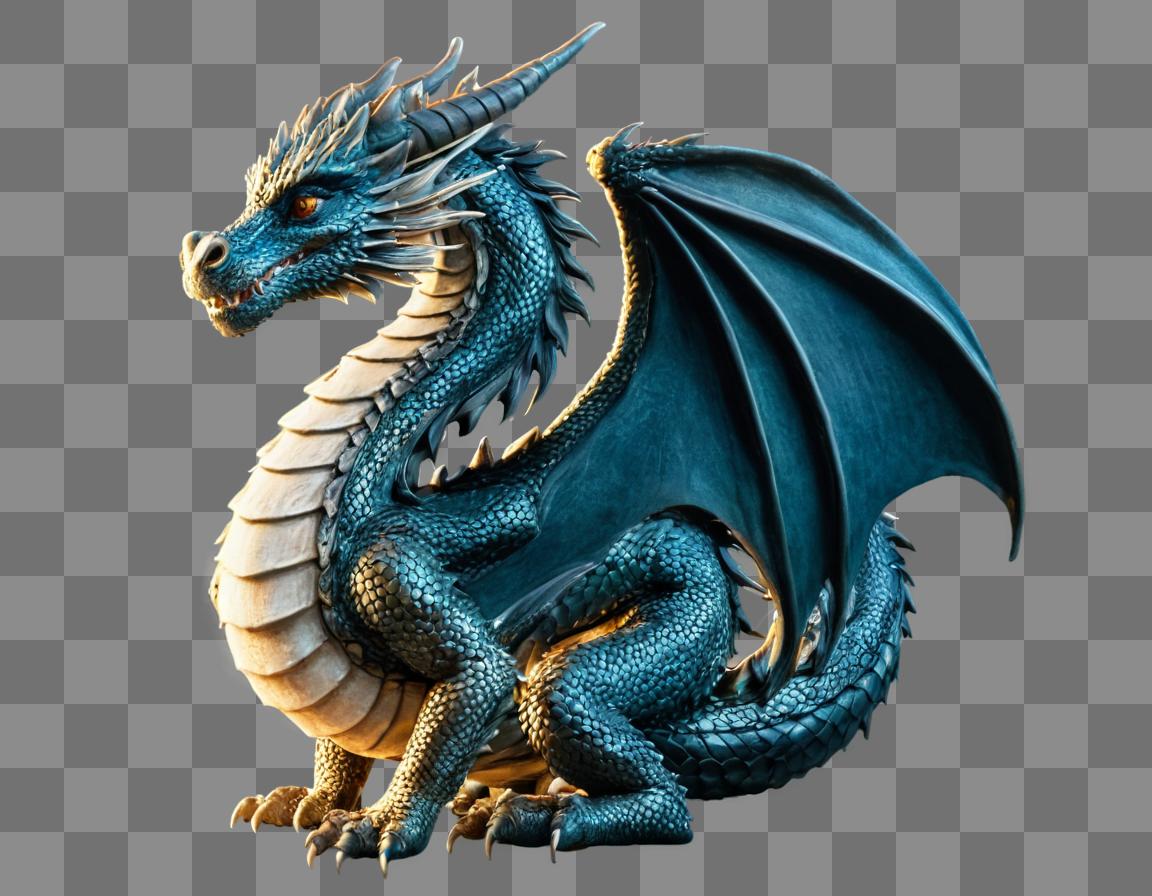}\hfill
\includegraphics[width=0.33\linewidth]{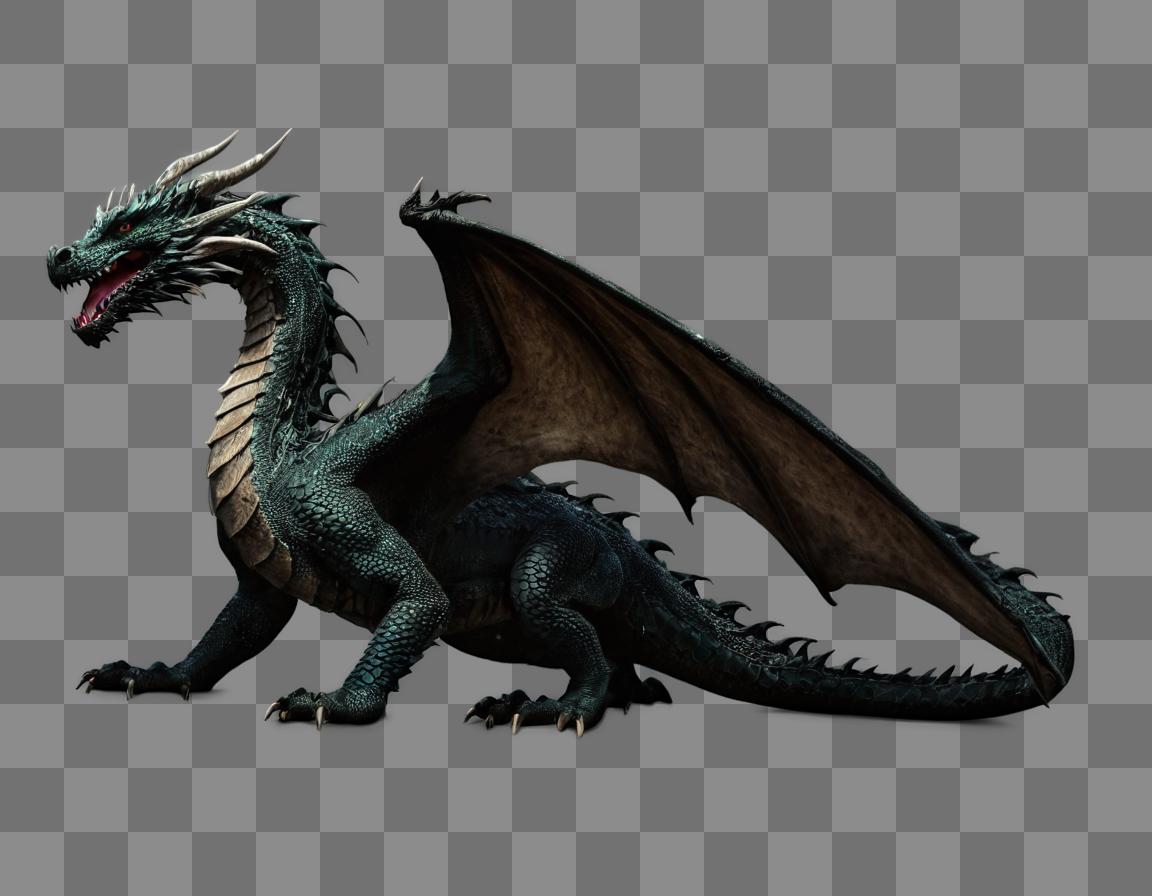}
\caption{Single Transparent Image Results \#9. The prompt is ``dragon''. Resolution is $1152\times896$.}
\label{fig:a9}
\end{minipage}
\end{figure*}

\begin{figure*}

\begin{minipage}{\linewidth}
\includegraphics[width=0.33\linewidth]{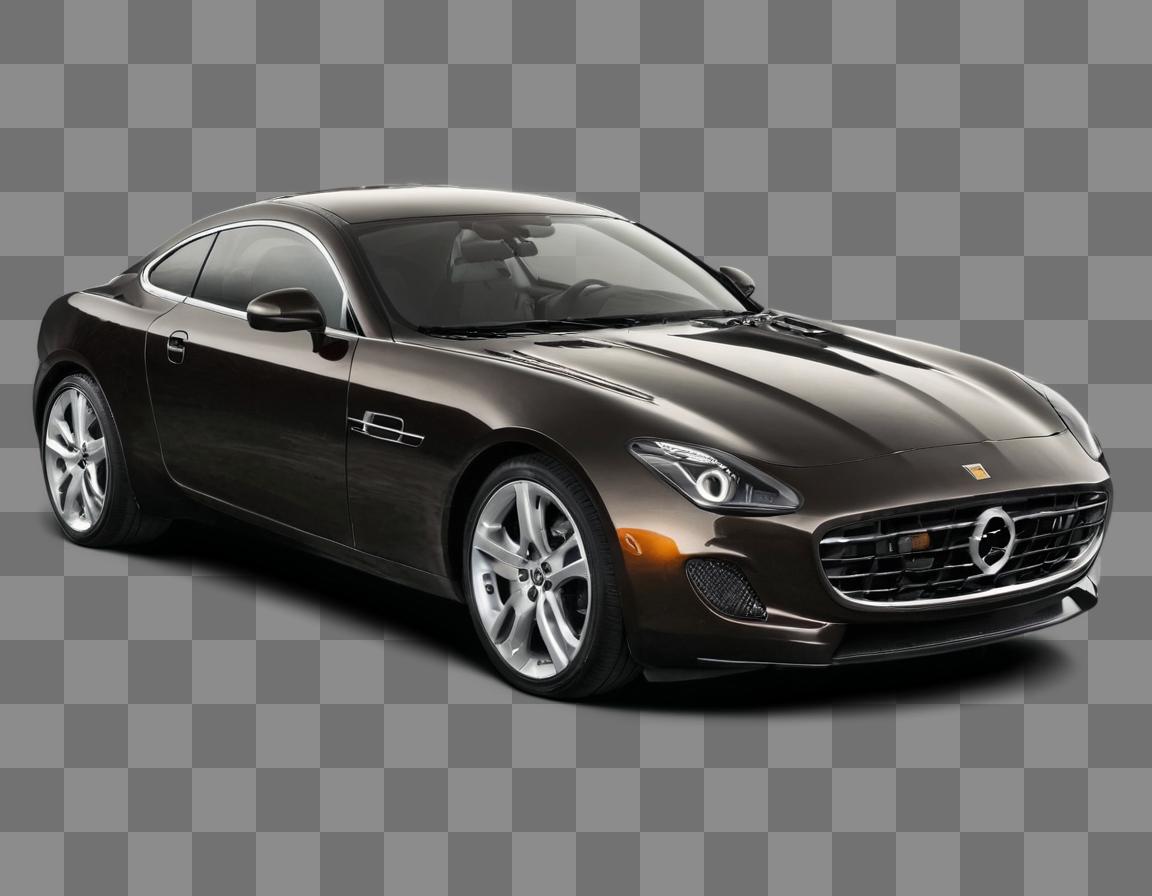}\hfill
\includegraphics[width=0.33\linewidth]{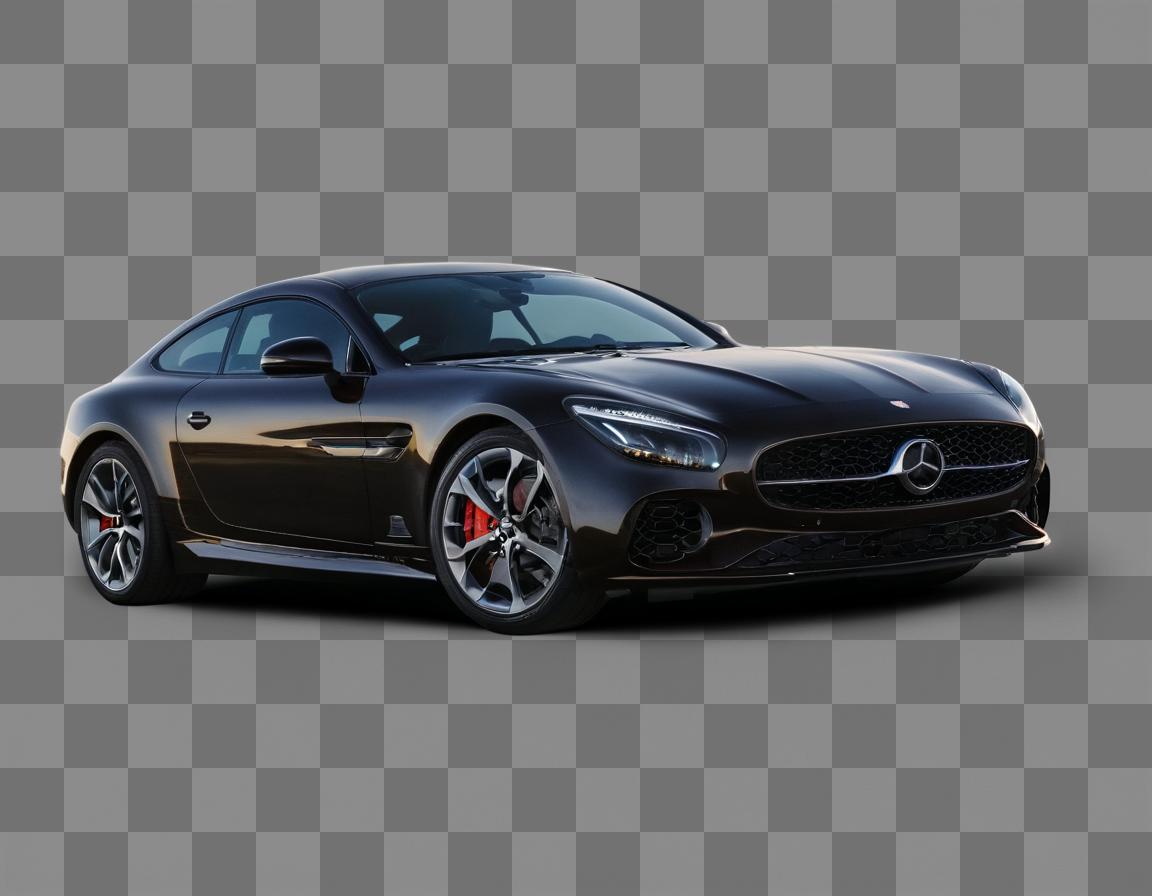}\hfill
\includegraphics[width=0.33\linewidth]{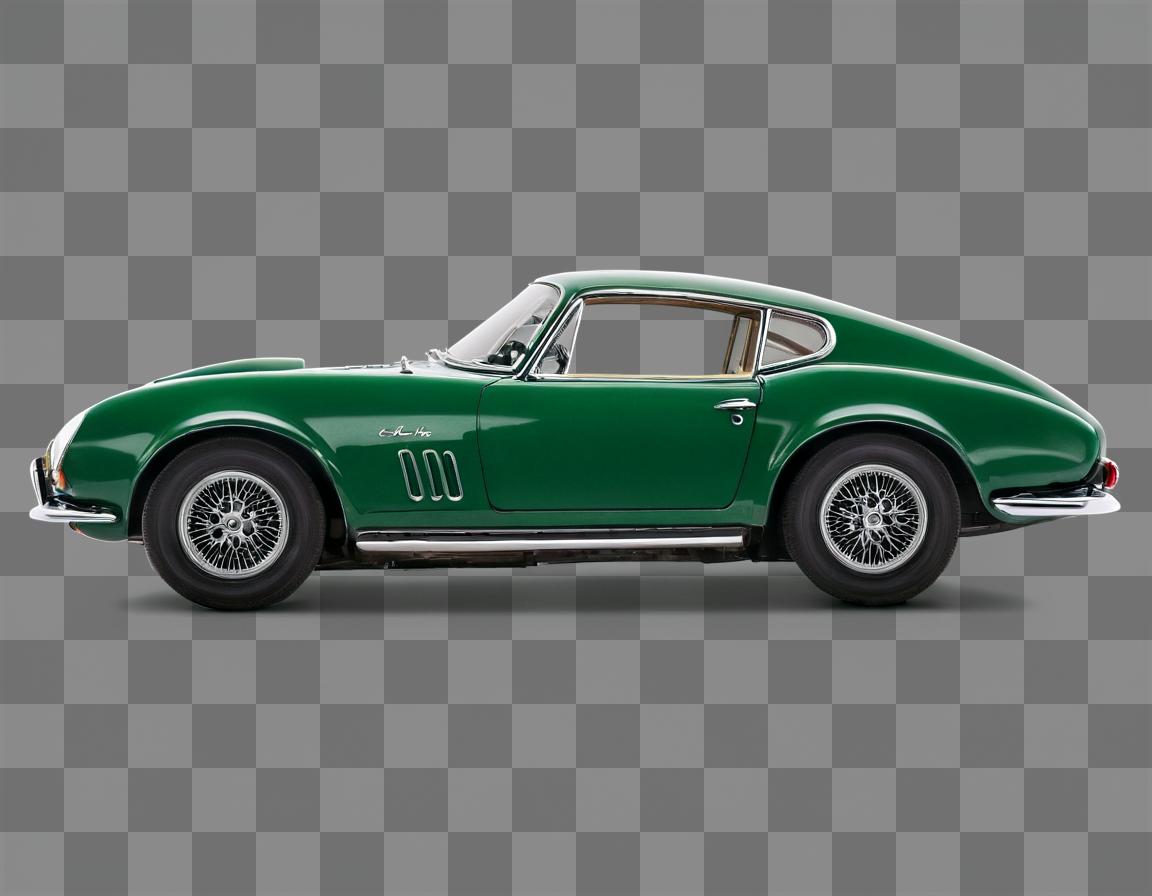}

\vspace{1pt}
\includegraphics[width=0.33\linewidth]{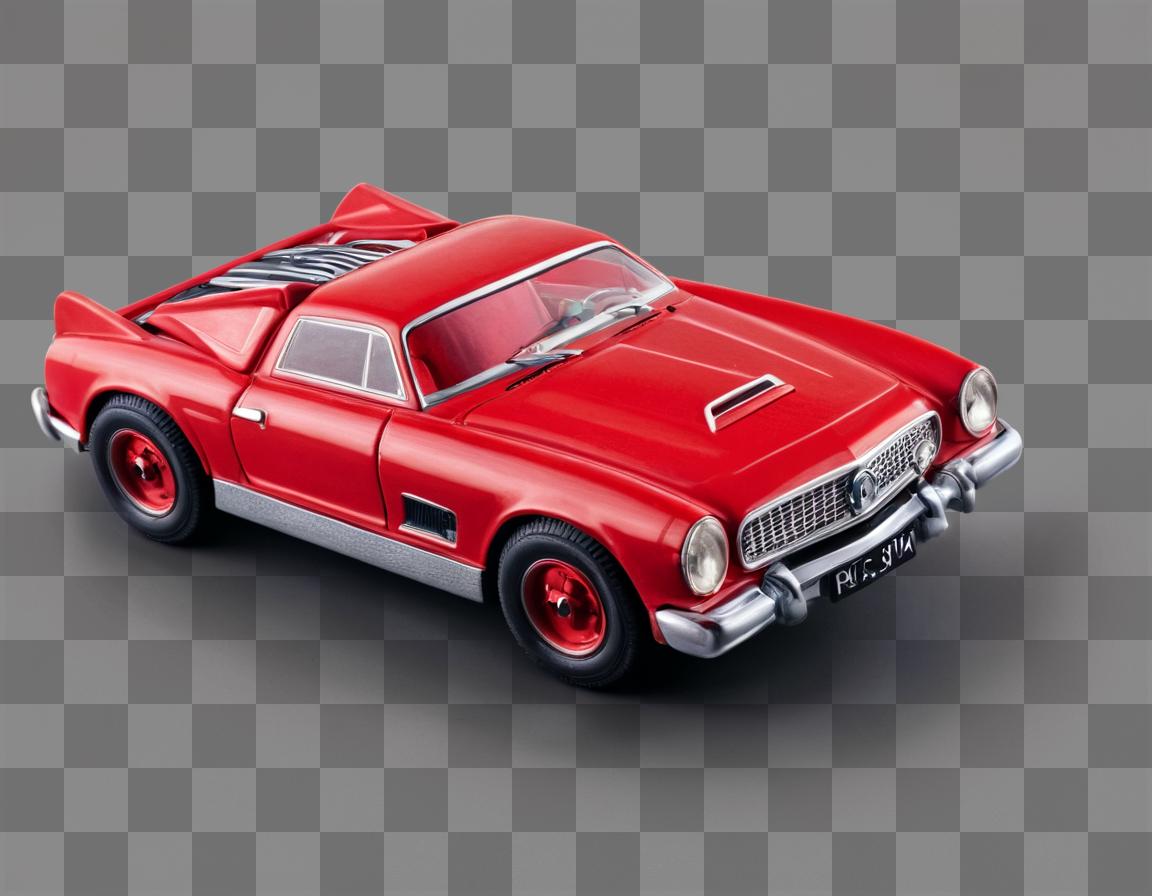}\hfill
\includegraphics[width=0.33\linewidth]{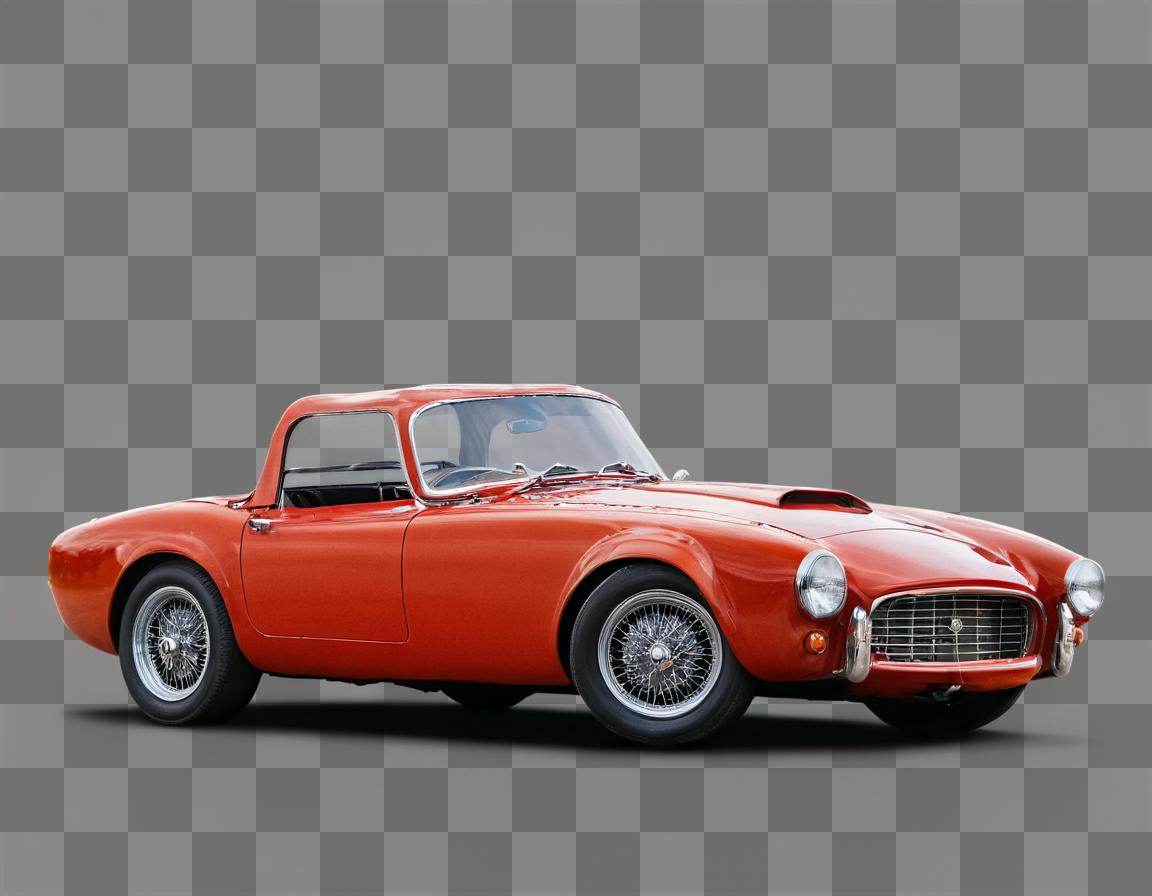}\hfill
\includegraphics[width=0.33\linewidth]{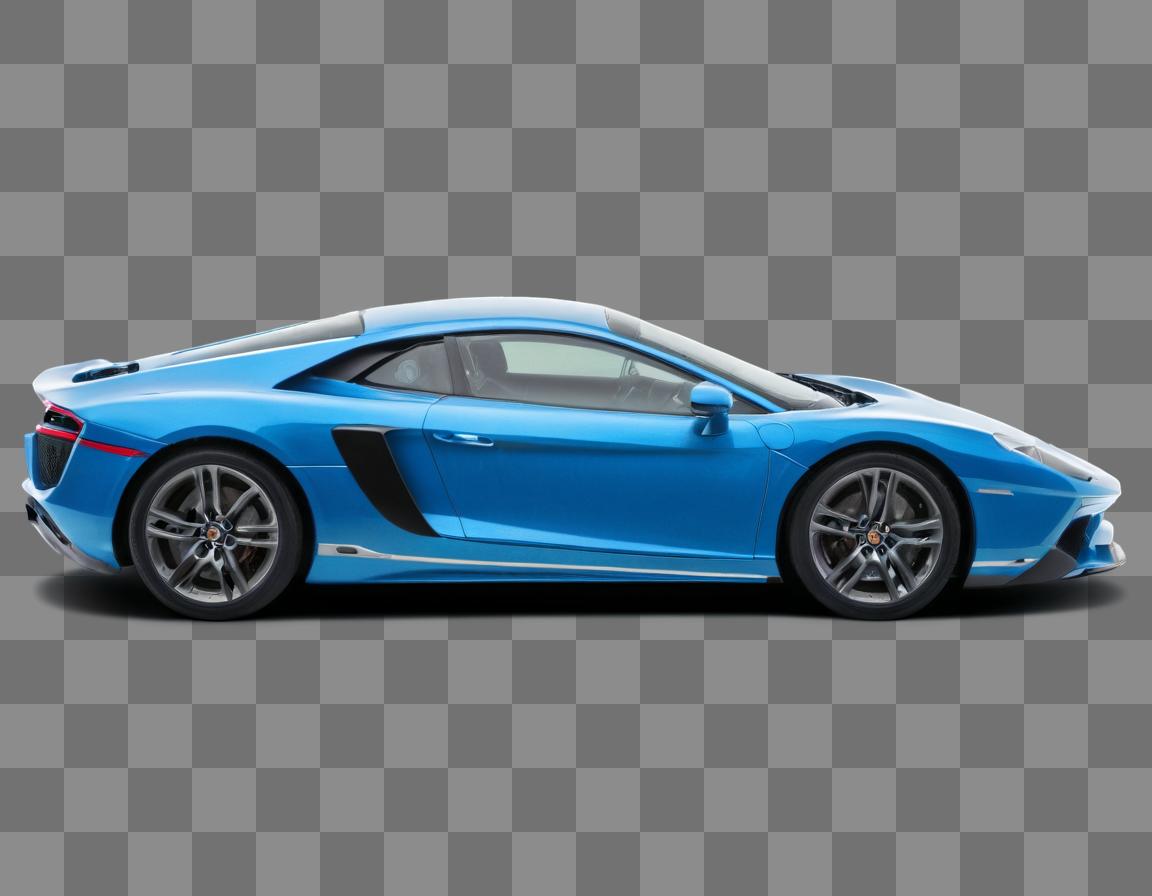}

\vspace{1pt}
\includegraphics[width=0.33\linewidth]{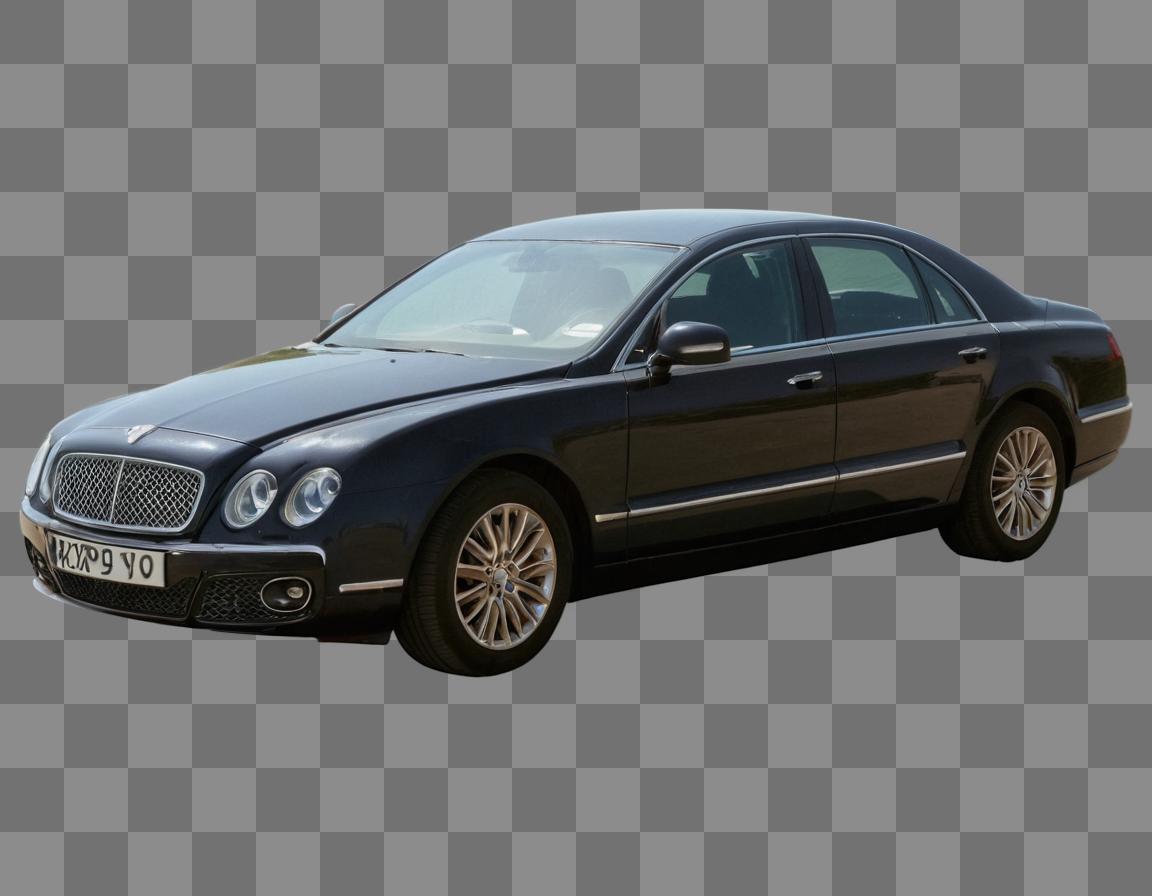}\hfill
\includegraphics[width=0.33\linewidth]{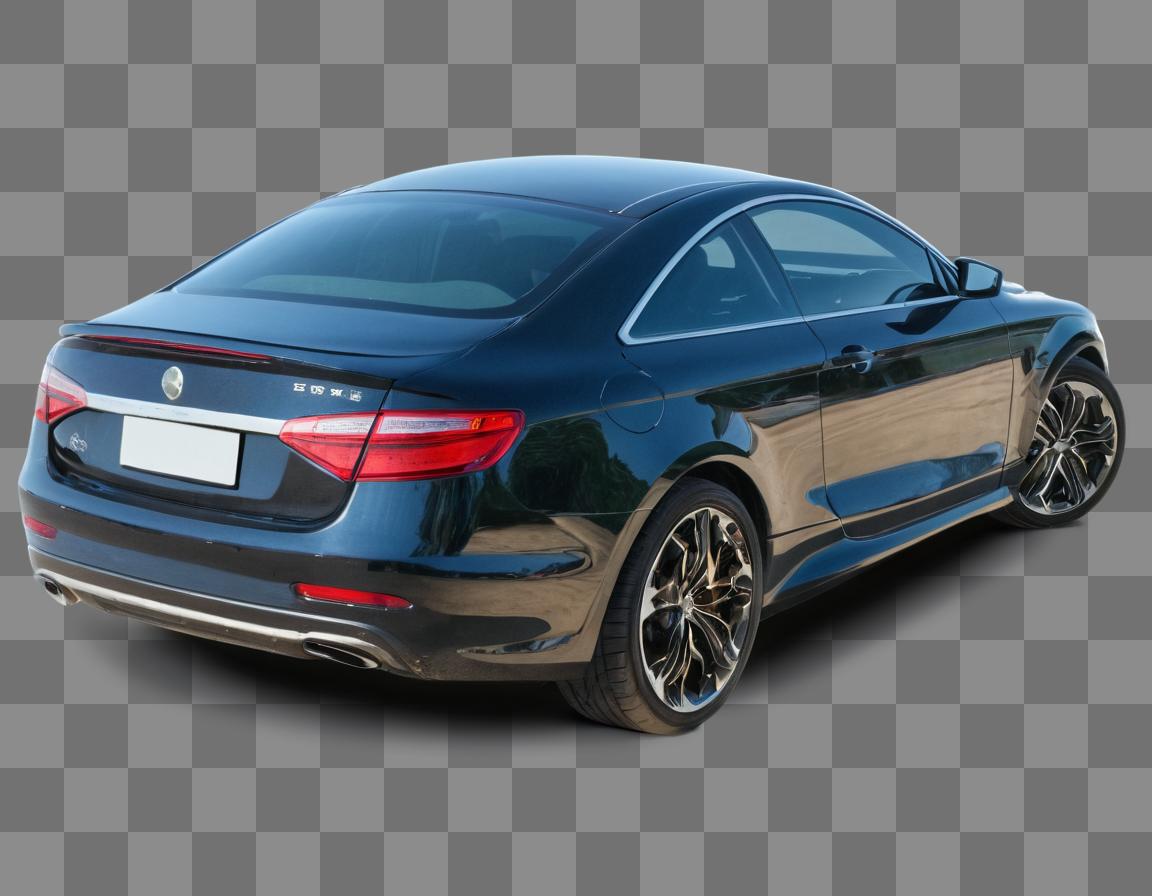}\hfill
\includegraphics[width=0.33\linewidth]{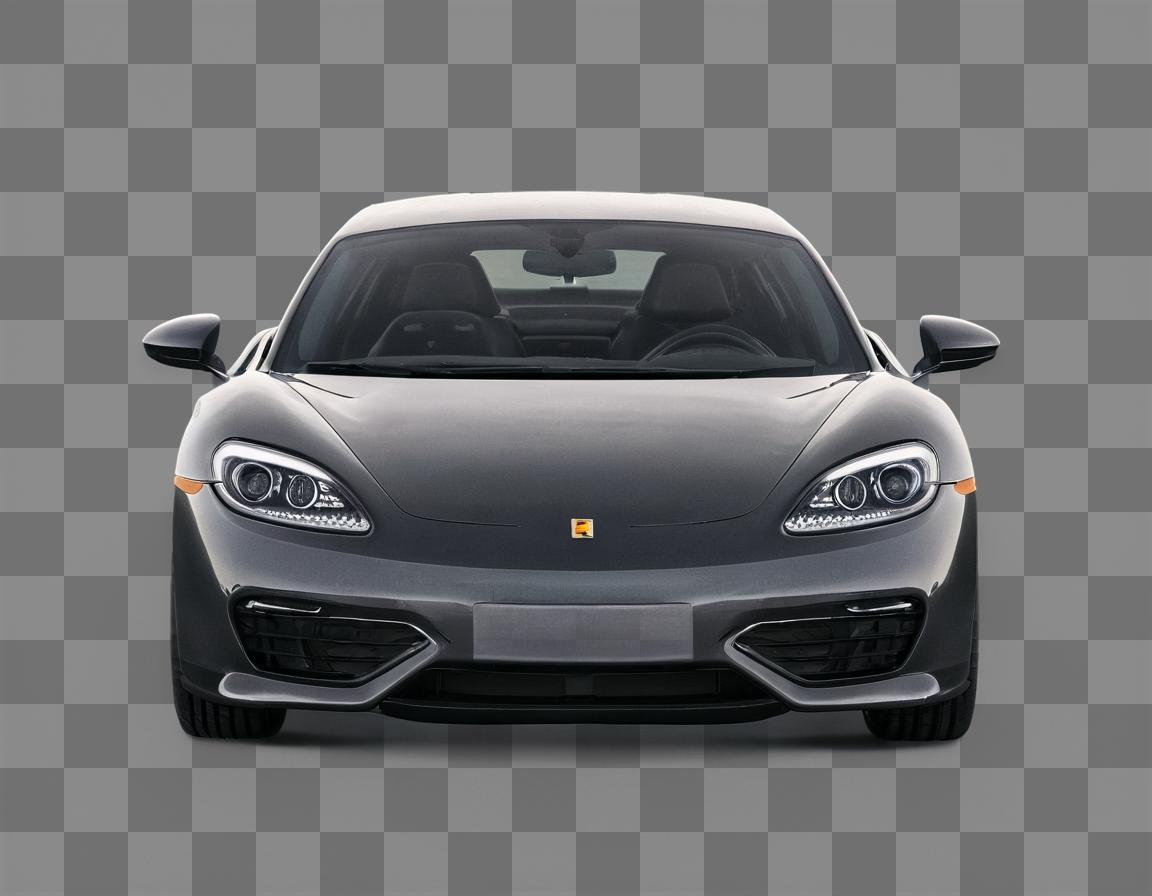}

\vspace{1pt}
\includegraphics[width=0.33\linewidth]{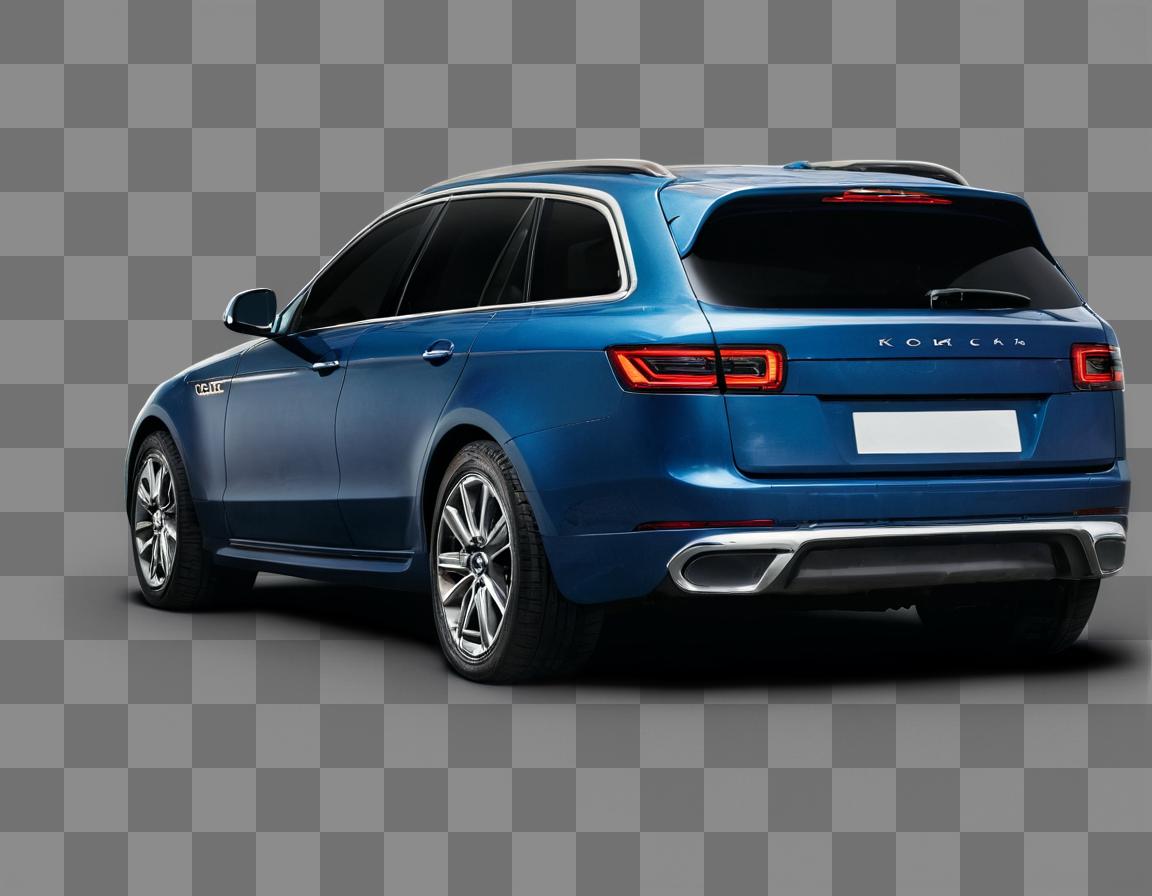}\hfill
\includegraphics[width=0.33\linewidth]{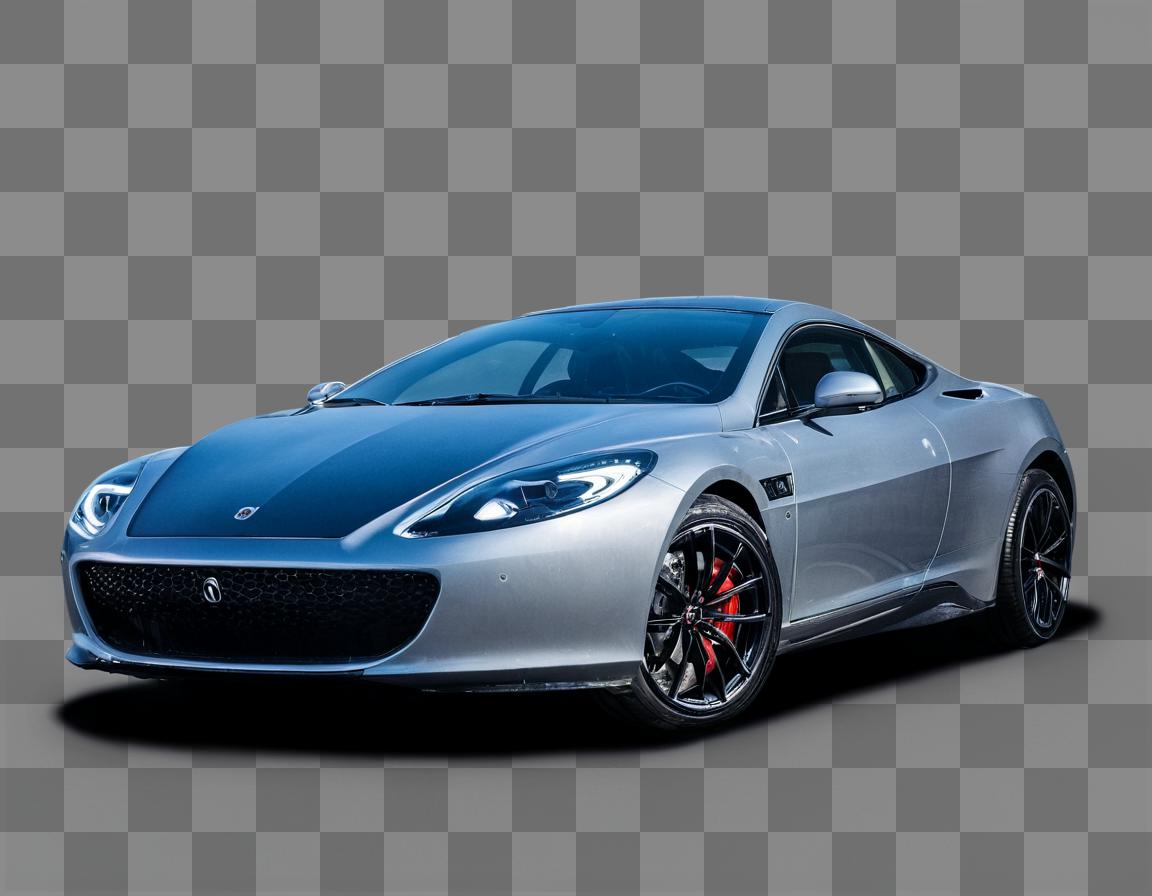}\hfill
\includegraphics[width=0.33\linewidth]{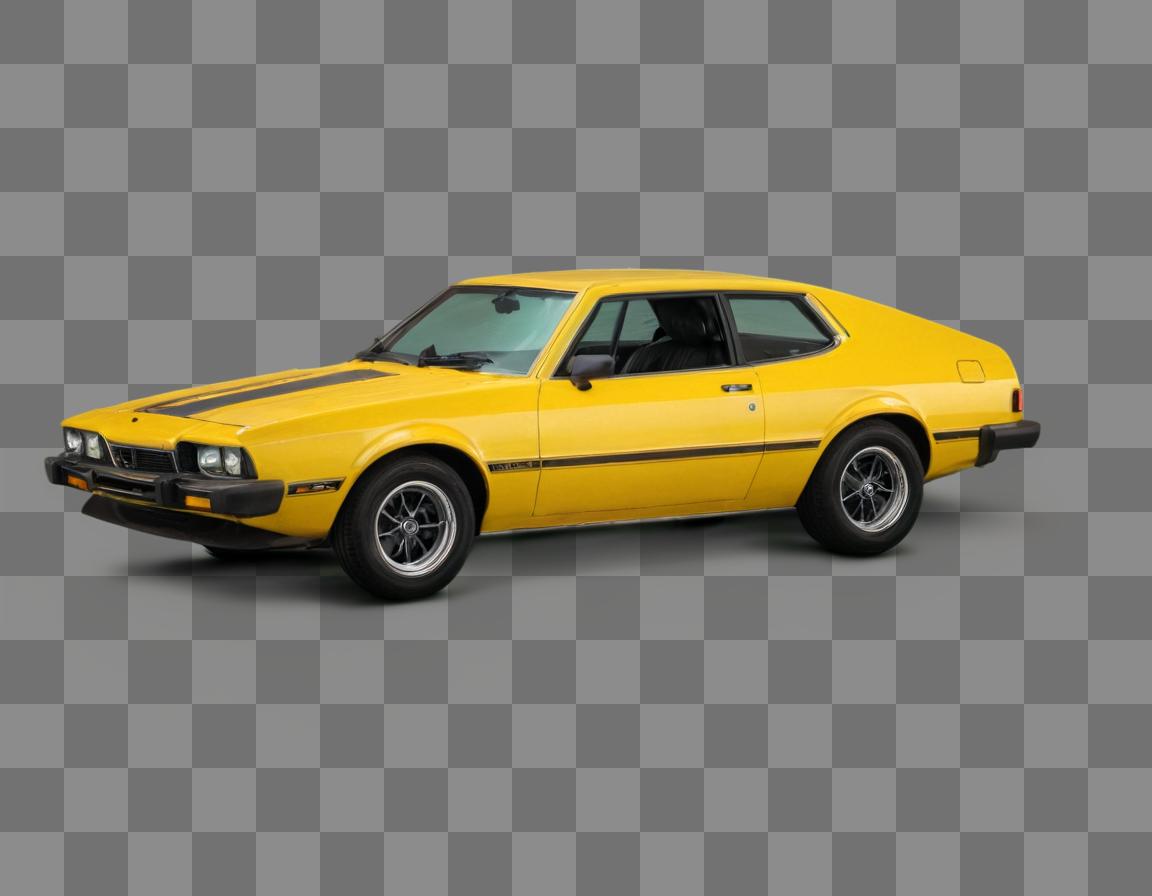}
\caption{Single Transparent Image Results \#10. The prompt is ``car''. Resolution is $1152\times896$.}
\label{fig:a10}
\end{minipage}
\end{figure*}

\begin{figure*}

\begin{minipage}{\linewidth}
\includegraphics[width=0.33\linewidth]{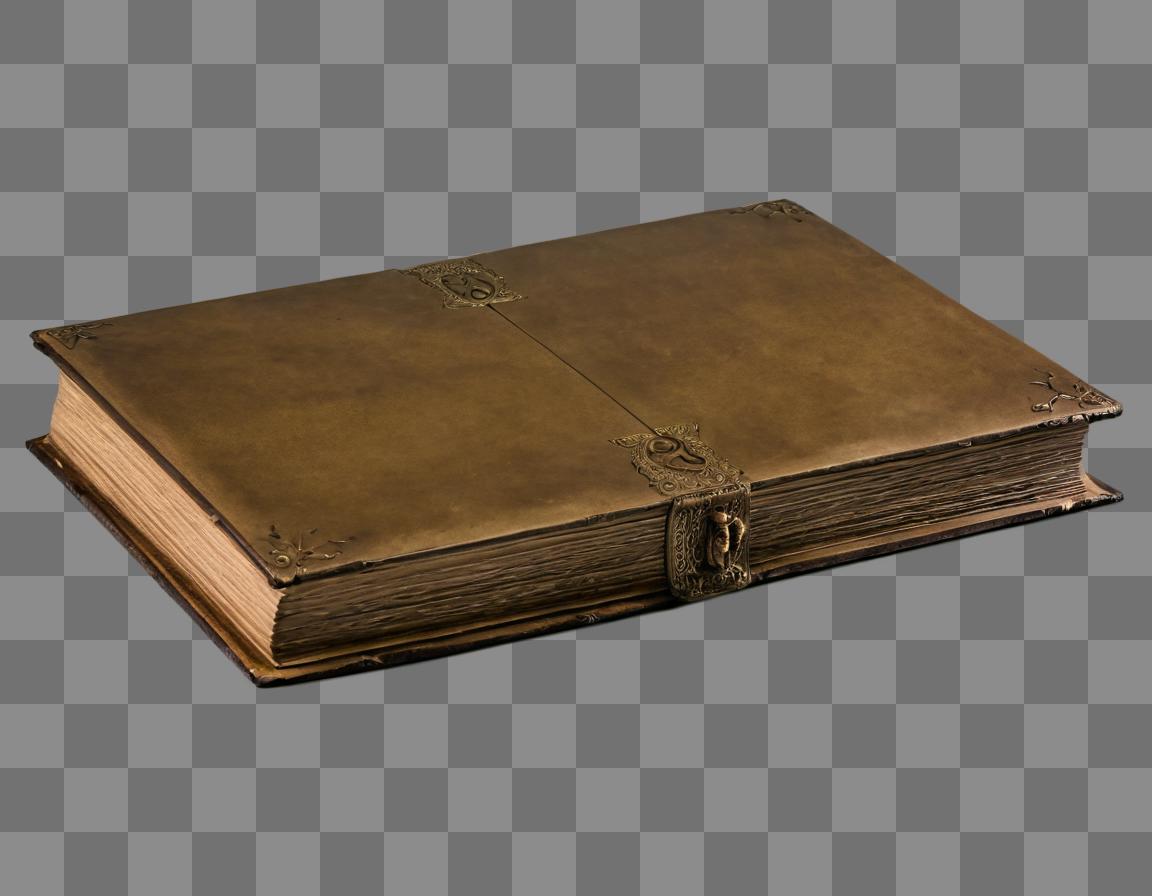}\hfill
\includegraphics[width=0.33\linewidth]{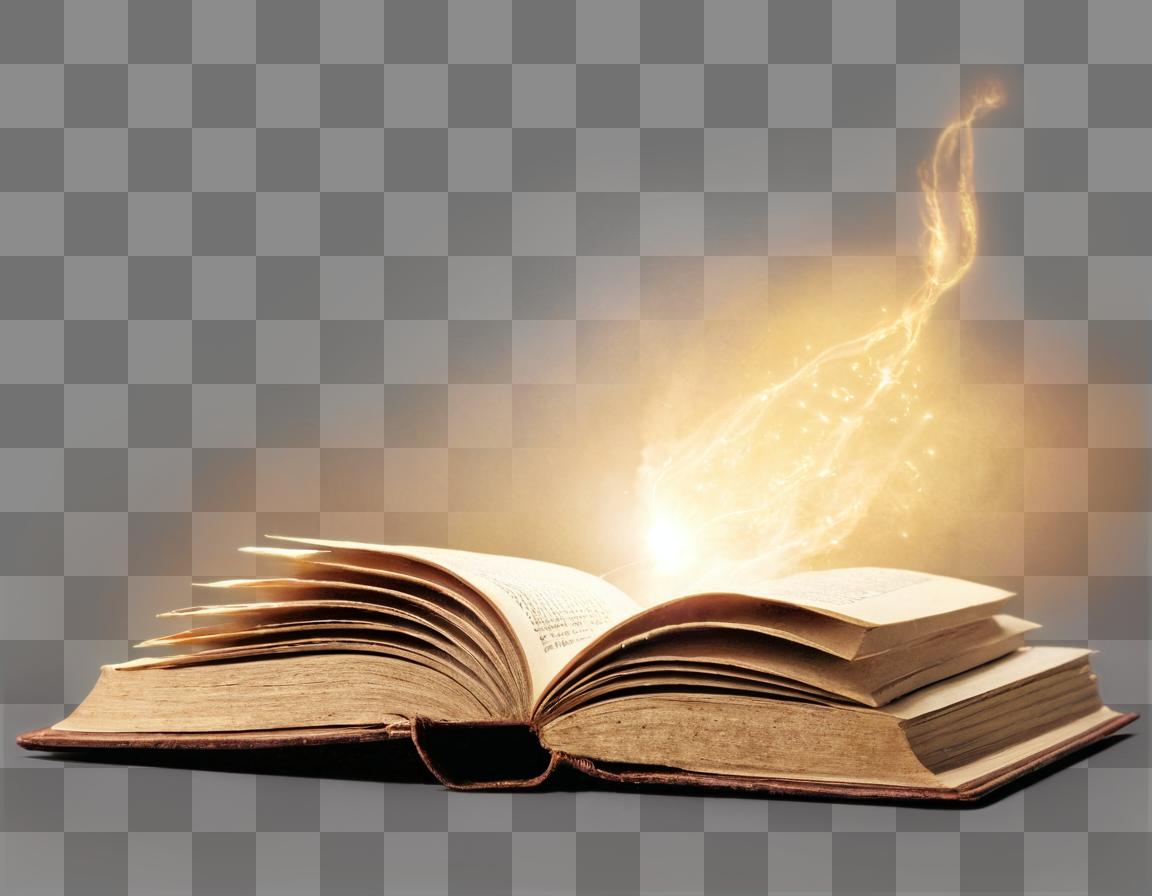}\hfill
\includegraphics[width=0.33\linewidth]{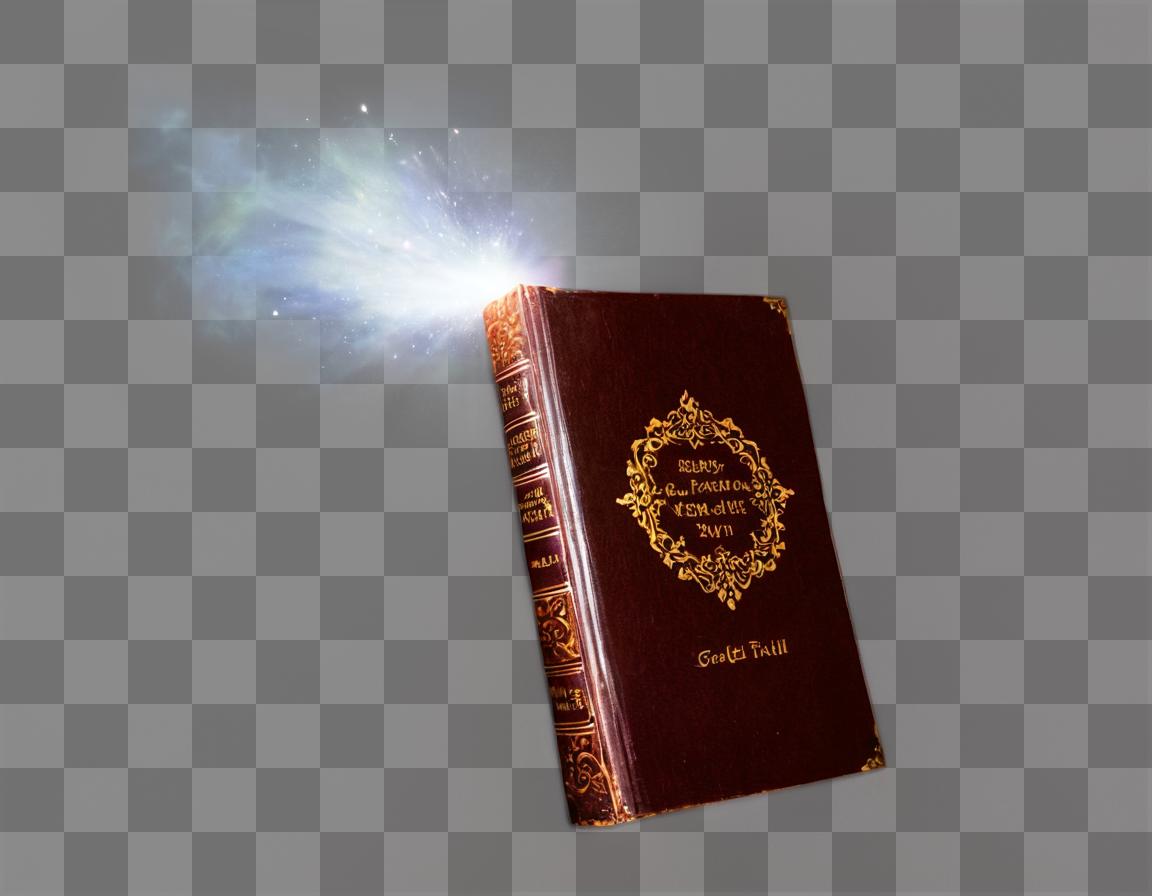}

\vspace{1pt}
\includegraphics[width=0.33\linewidth]{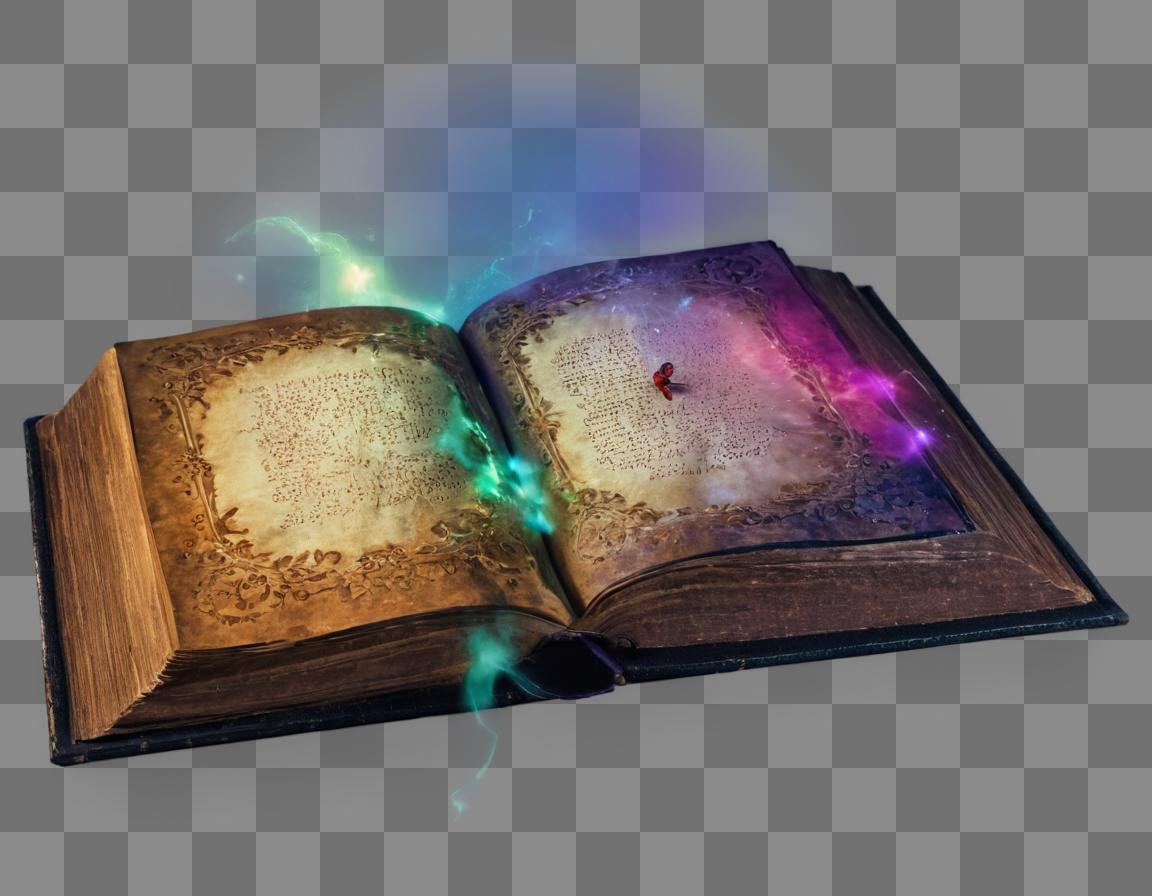}\hfill
\includegraphics[width=0.33\linewidth]{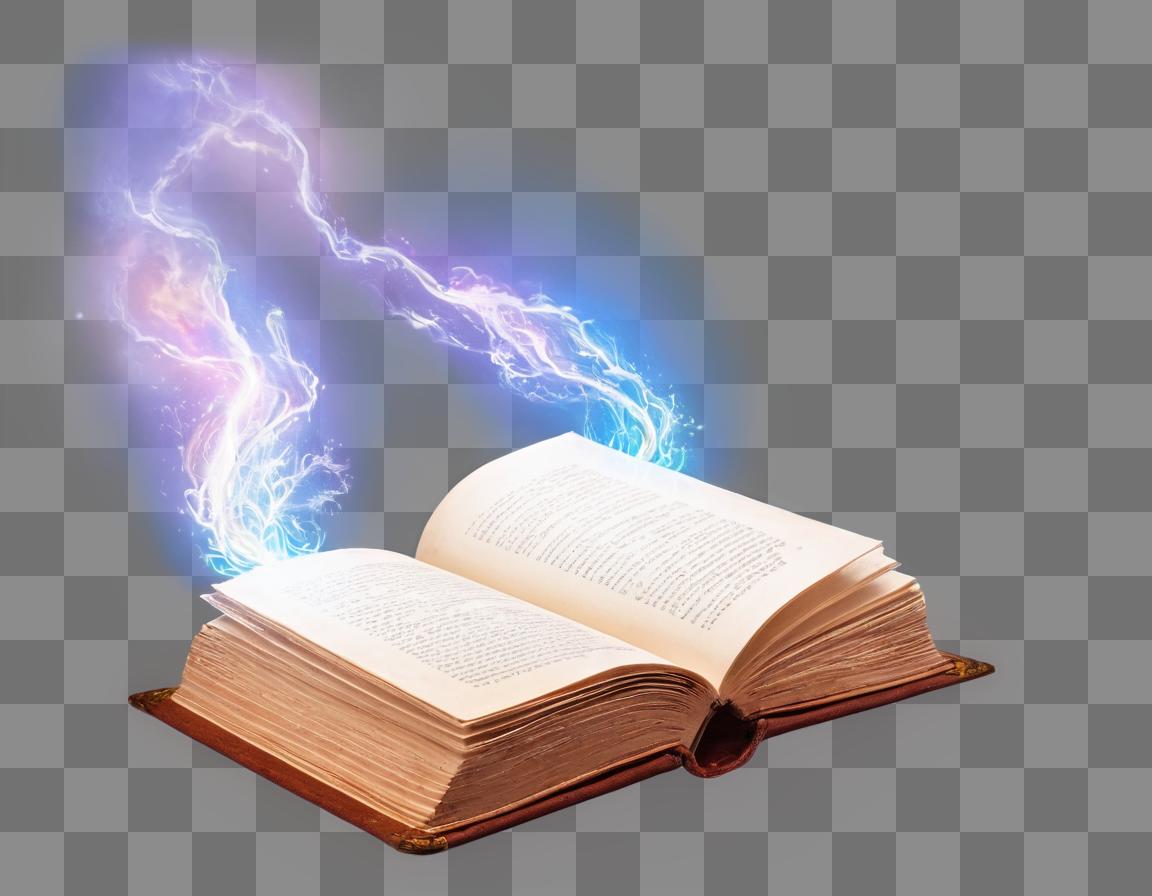}\hfill
\includegraphics[width=0.33\linewidth]{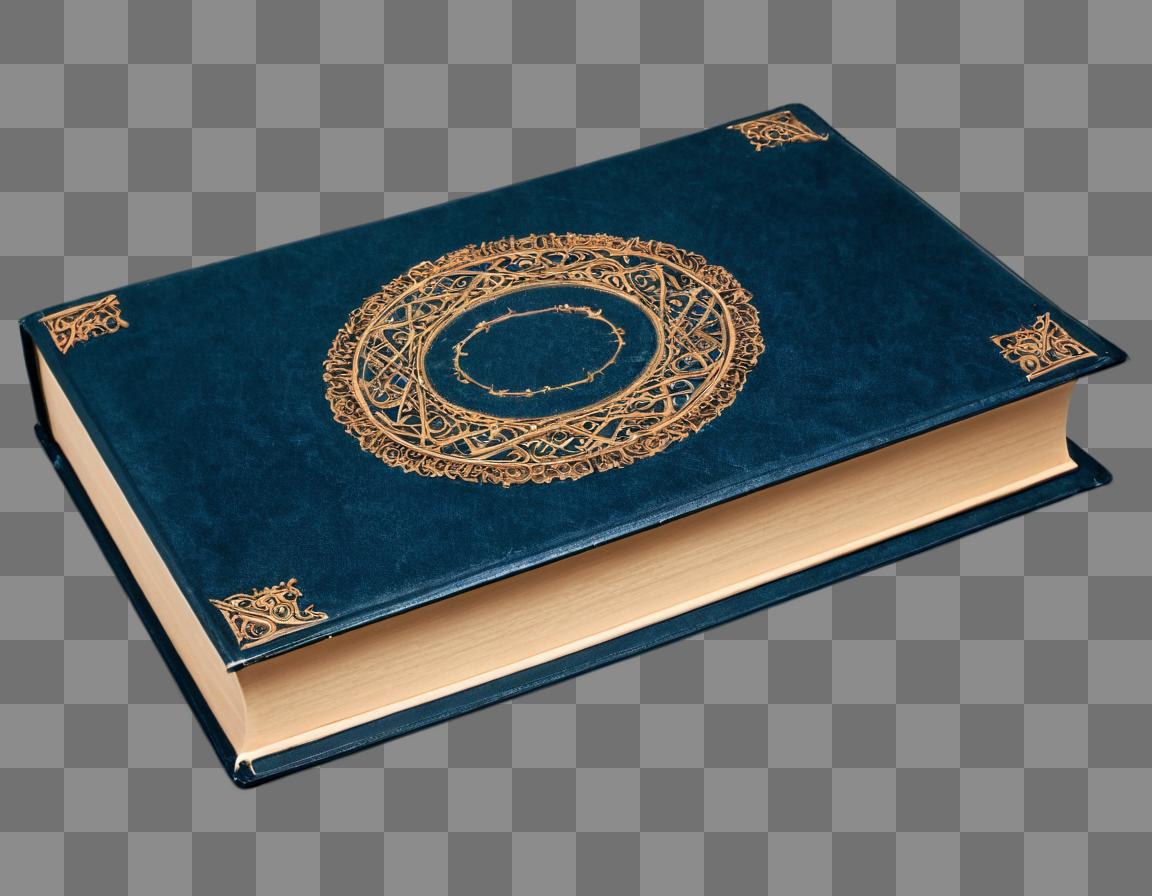}

\vspace{1pt}
\includegraphics[width=0.33\linewidth]{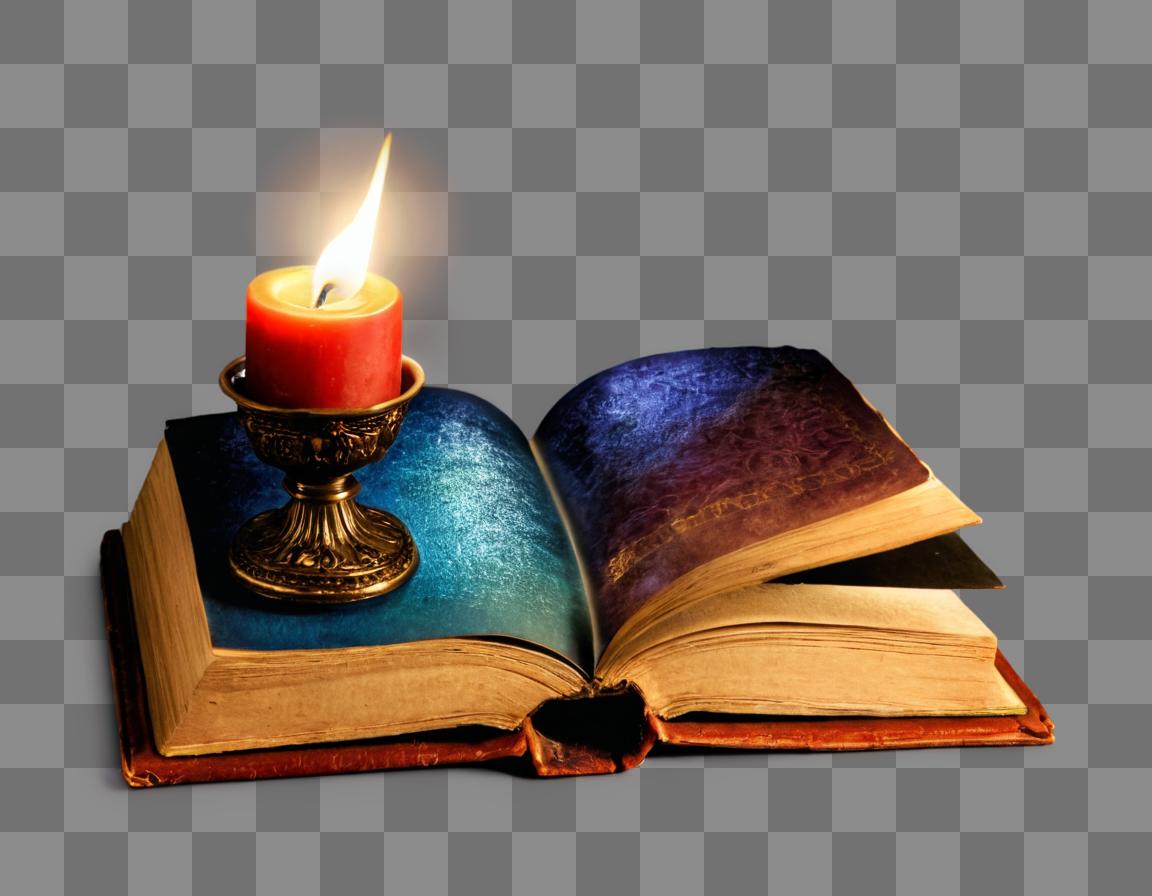}\hfill
\includegraphics[width=0.33\linewidth]{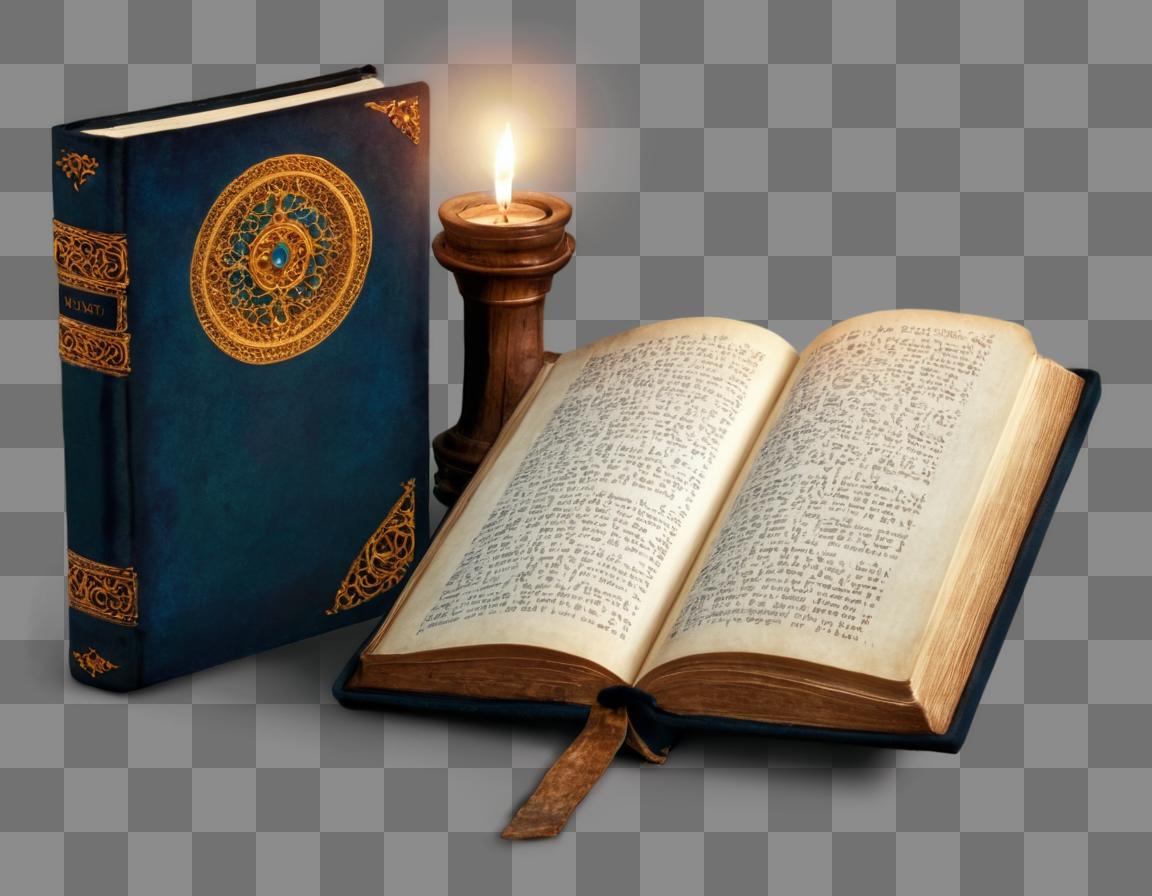}\hfill
\includegraphics[width=0.33\linewidth]{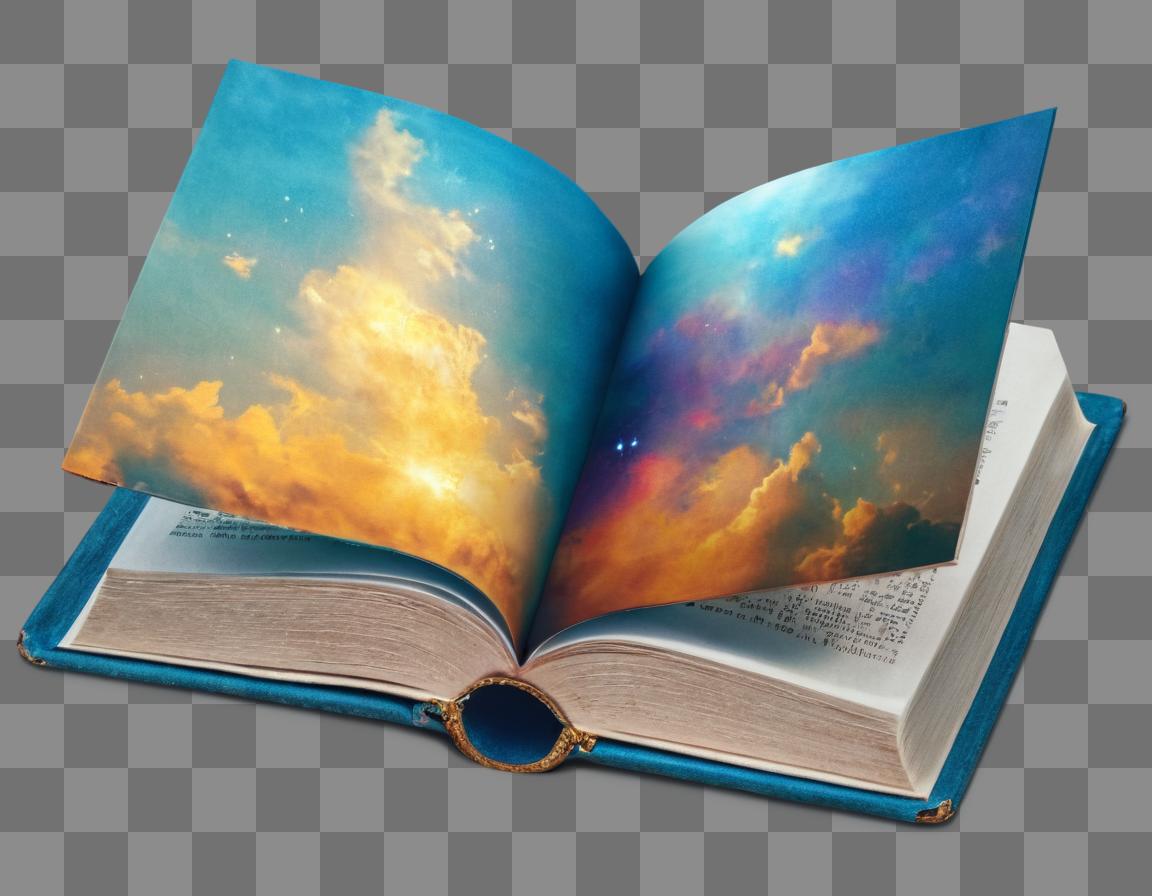}

\vspace{1pt}
\includegraphics[width=0.33\linewidth]{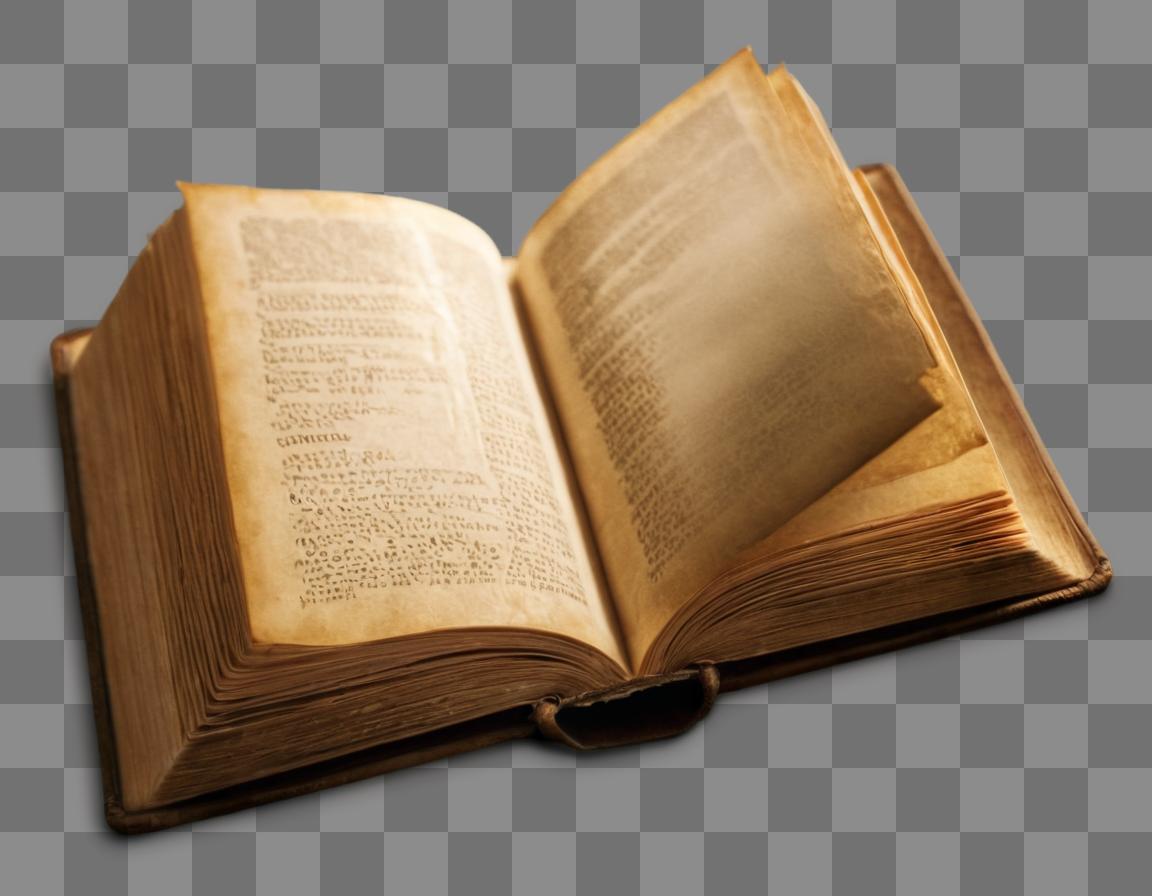}\hfill
\includegraphics[width=0.33\linewidth]{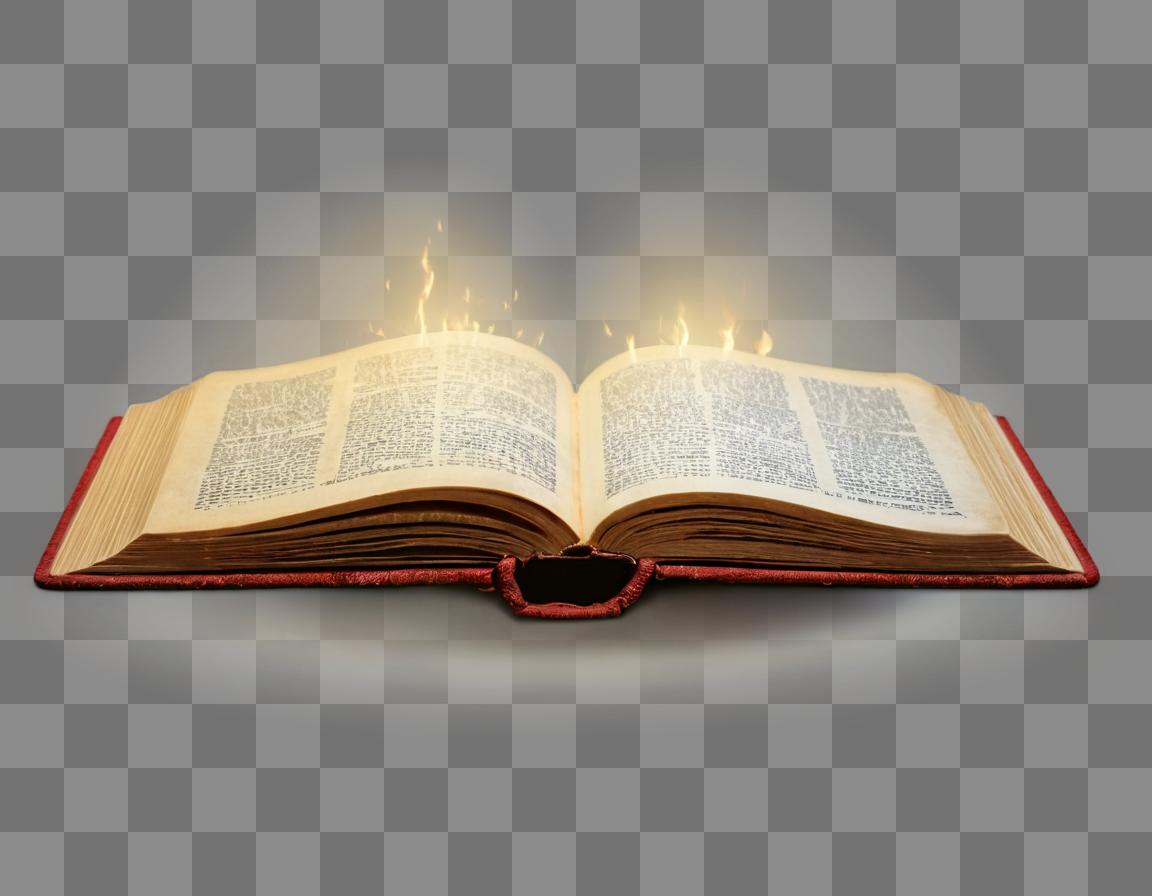}\hfill
\includegraphics[width=0.33\linewidth]{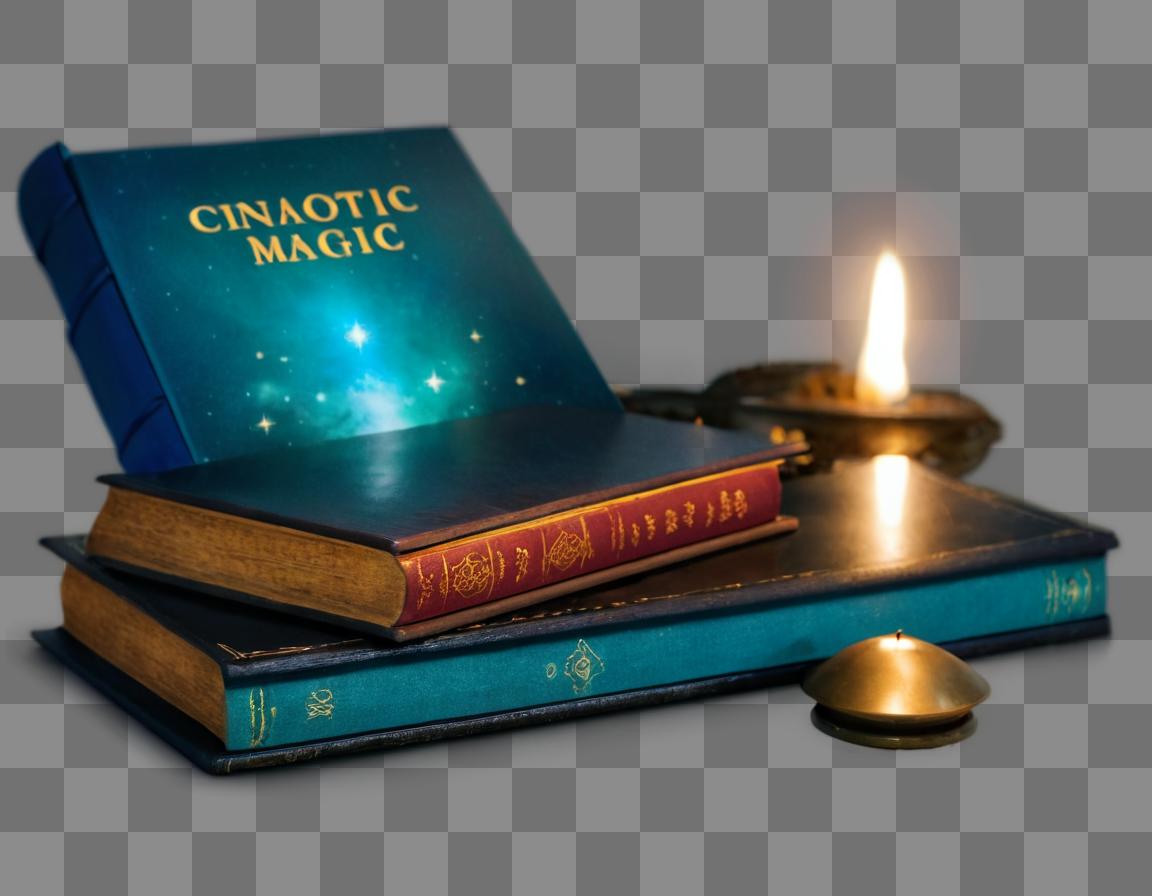}
\caption{Single Transparent Image Results \#11. The prompt is ``magic book''. Resolution is $1152\times896$.}
\label{fig:a11}
\end{minipage}
\end{figure*}

\begin{figure*}

\begin{minipage}{\linewidth}
\includegraphics[width=0.33\linewidth]{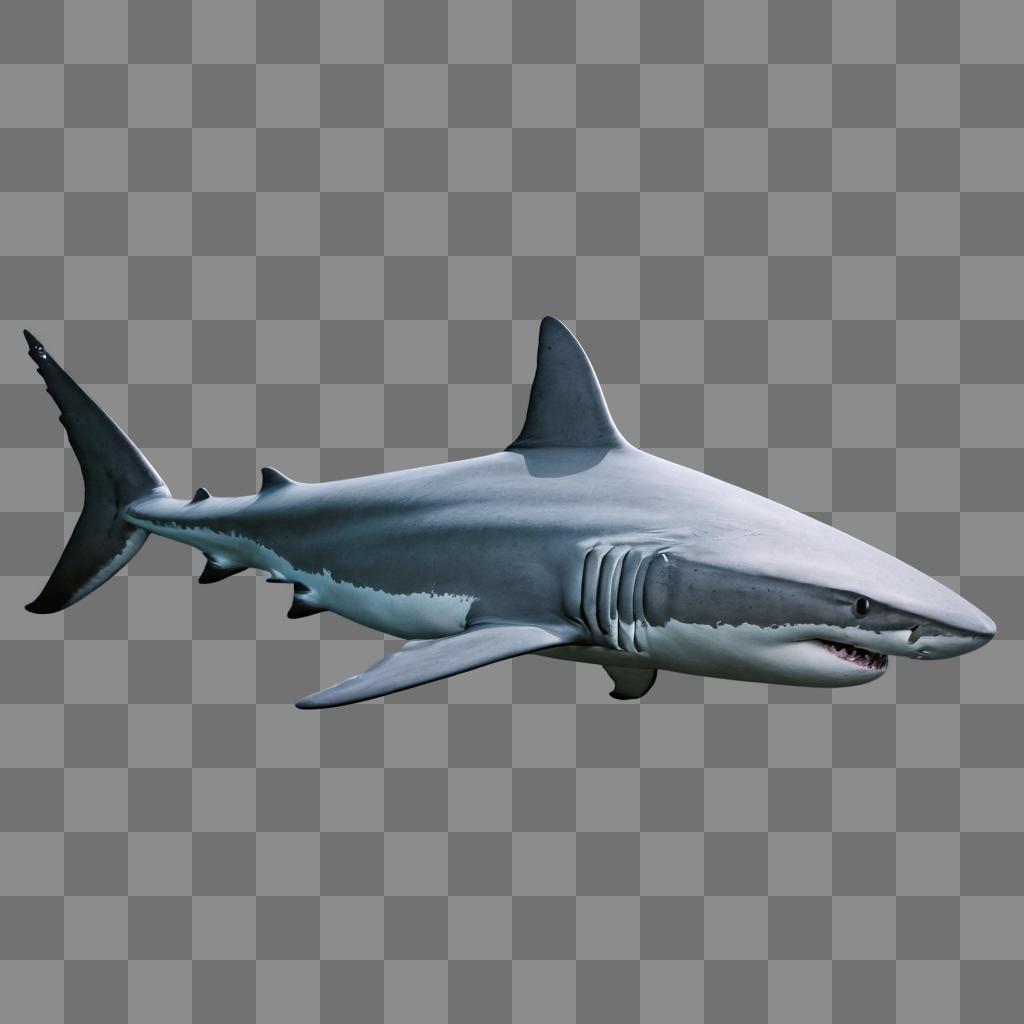}\hfill
\includegraphics[width=0.33\linewidth]{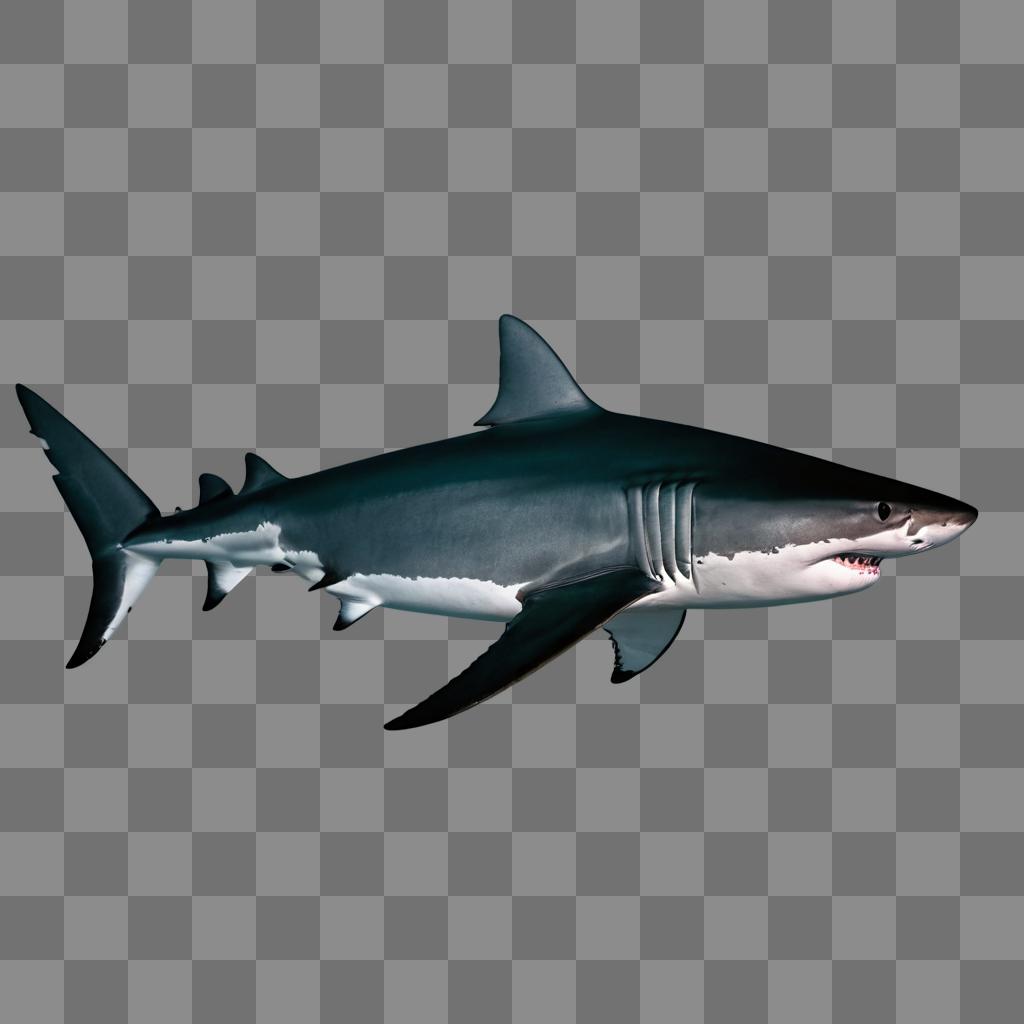}\hfill
\includegraphics[width=0.33\linewidth]{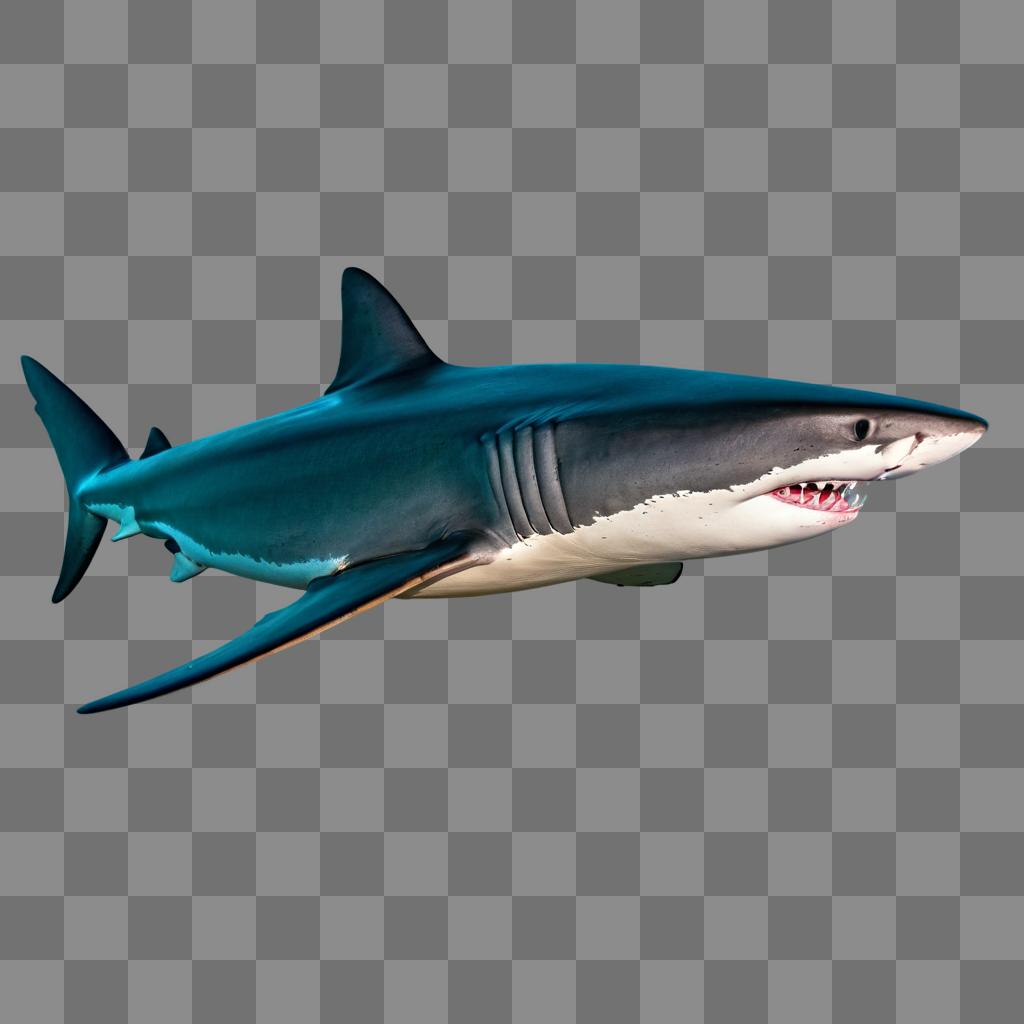}

\vspace{1pt}
\includegraphics[width=0.33\linewidth]{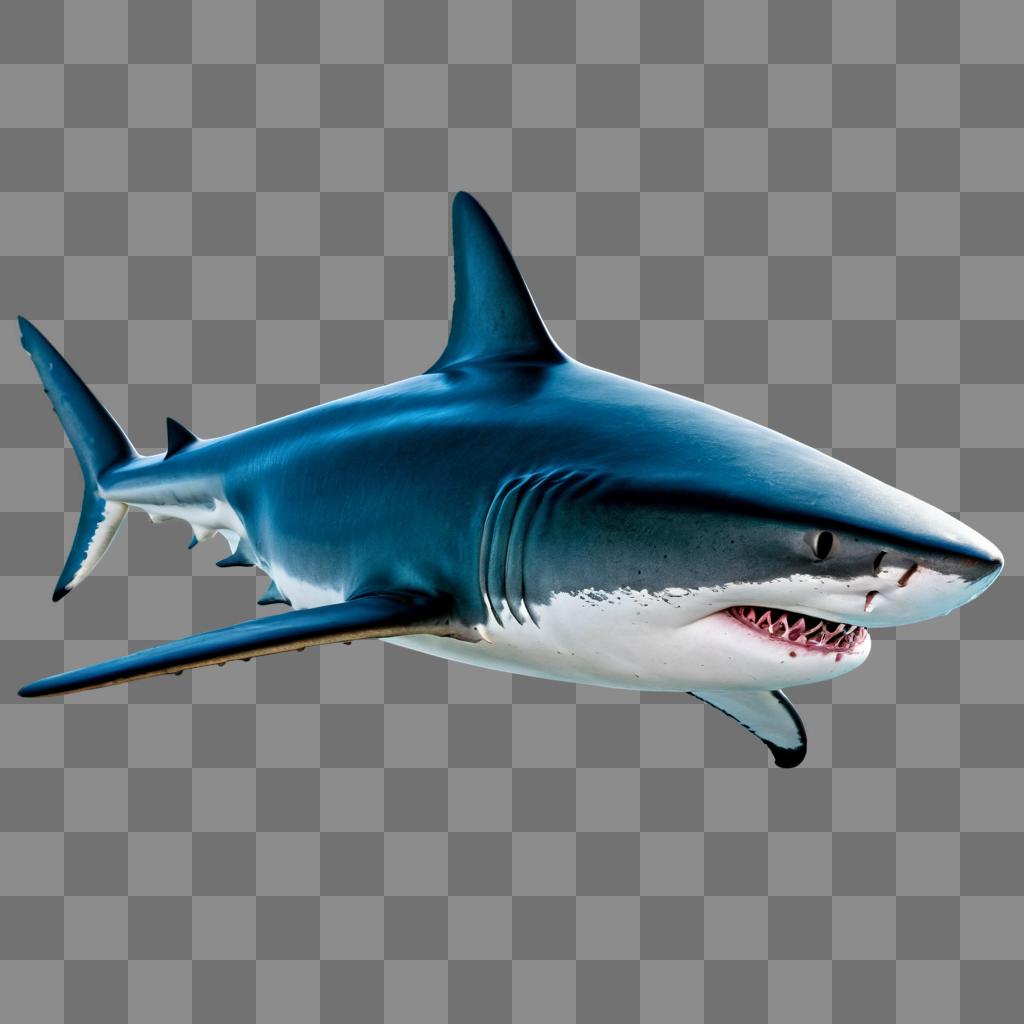}\hfill
\includegraphics[width=0.33\linewidth]{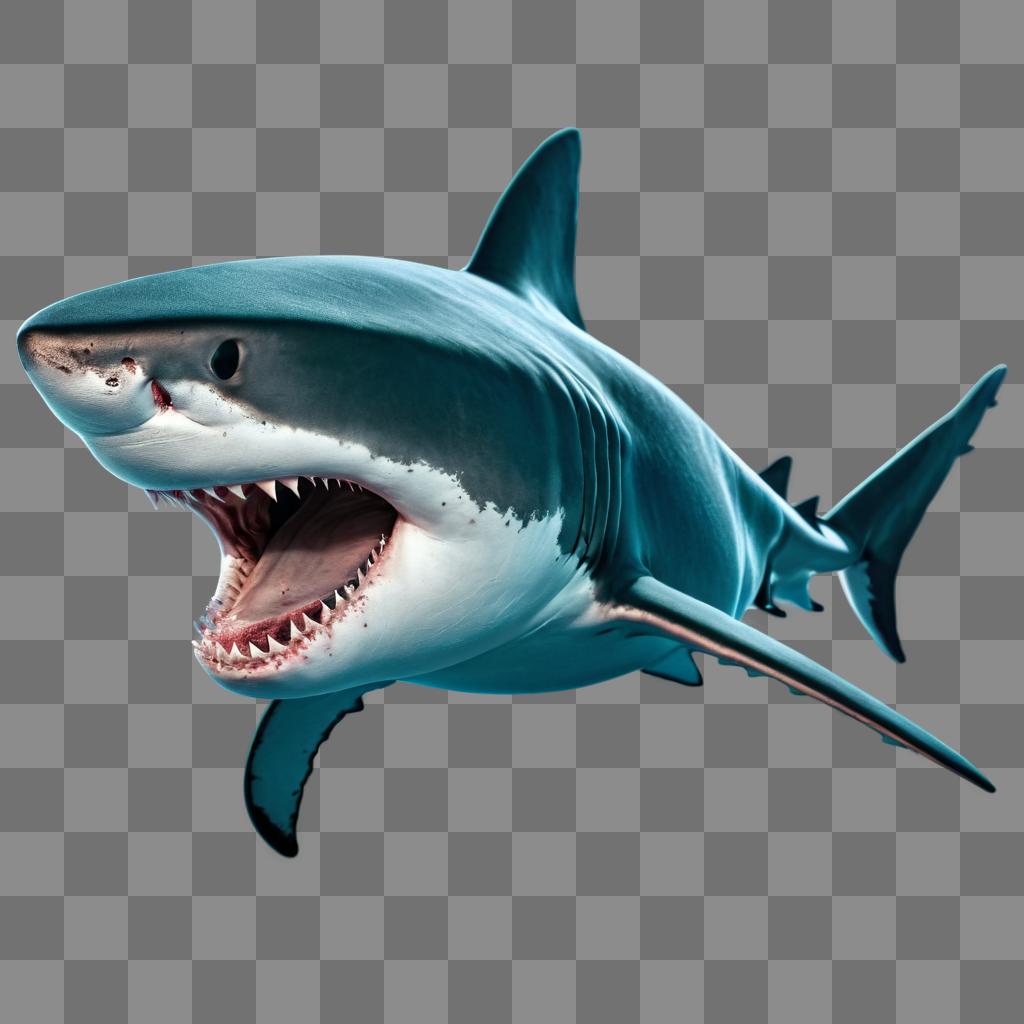}\hfill
\includegraphics[width=0.33\linewidth]{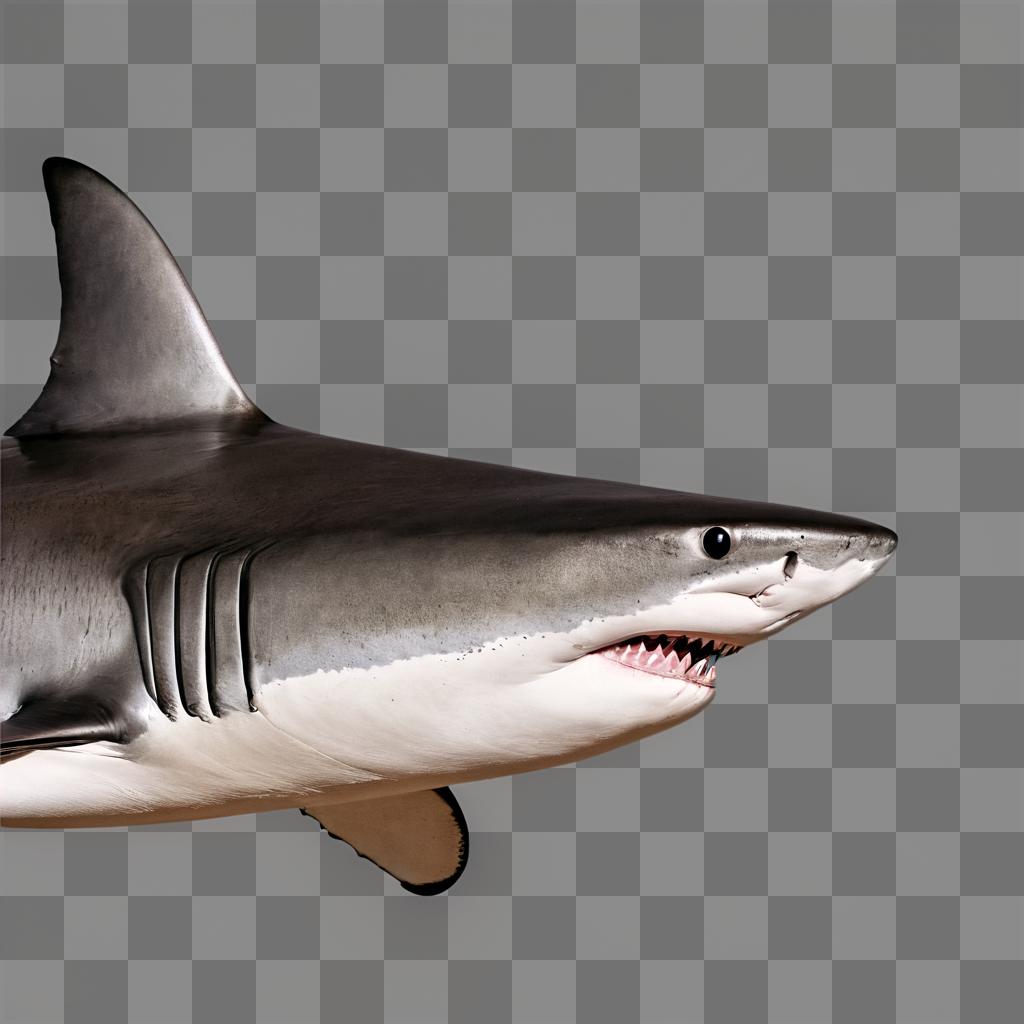}

\vspace{1pt}
\includegraphics[width=0.33\linewidth]{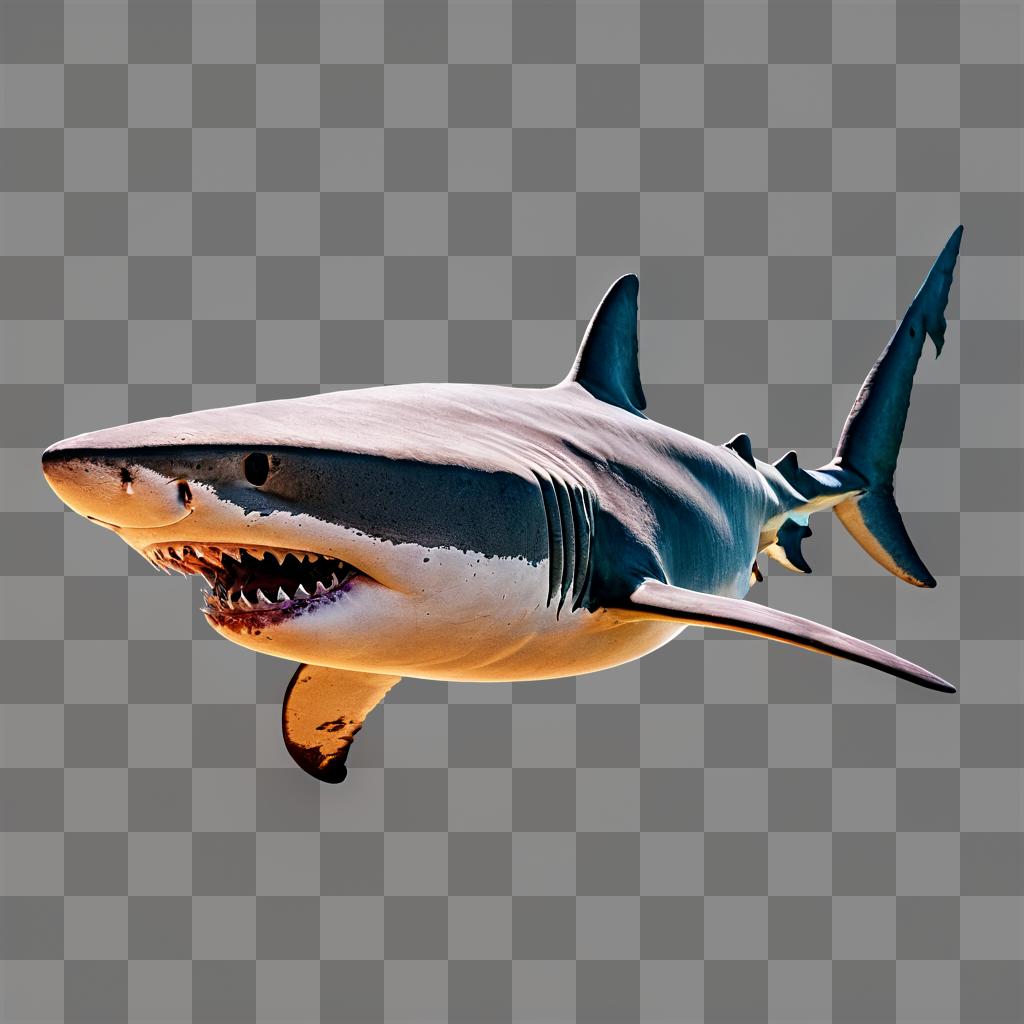}\hfill
\includegraphics[width=0.33\linewidth]{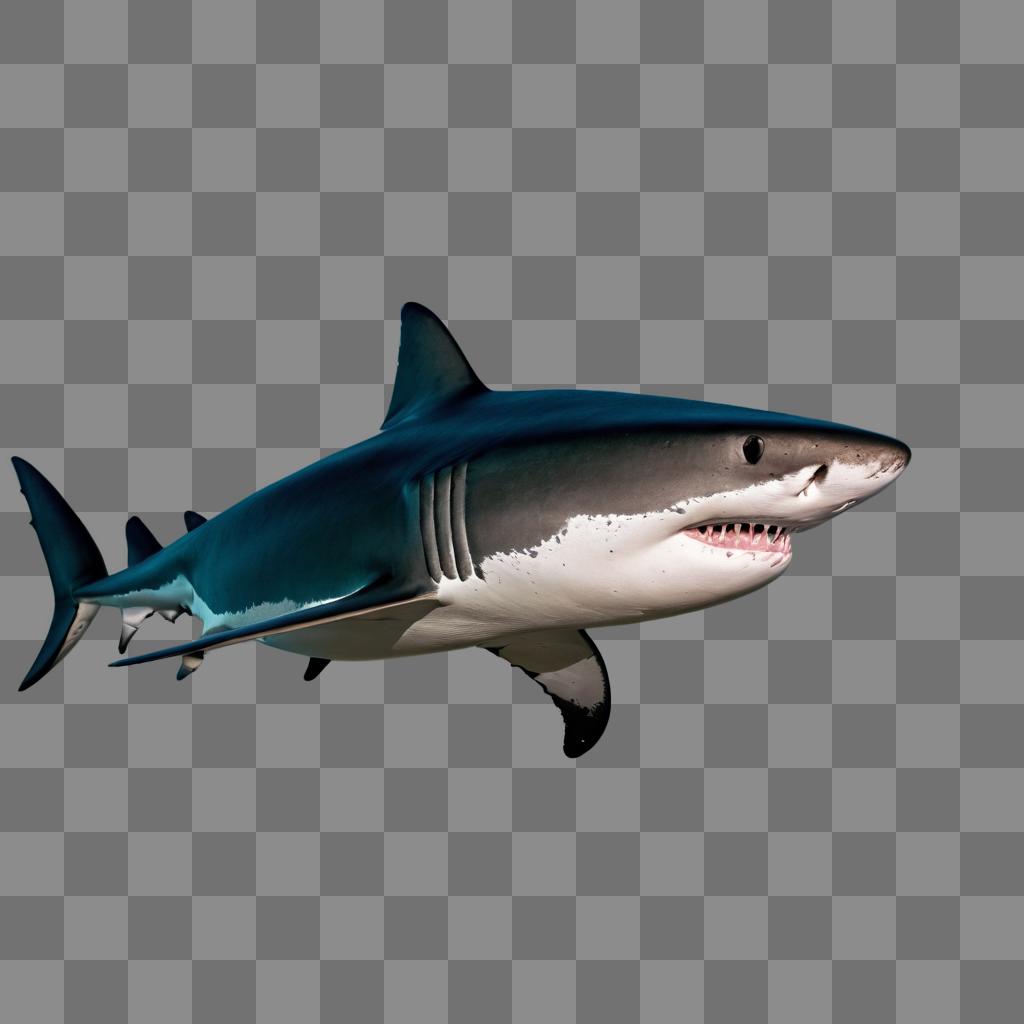}\hfill
\includegraphics[width=0.33\linewidth]{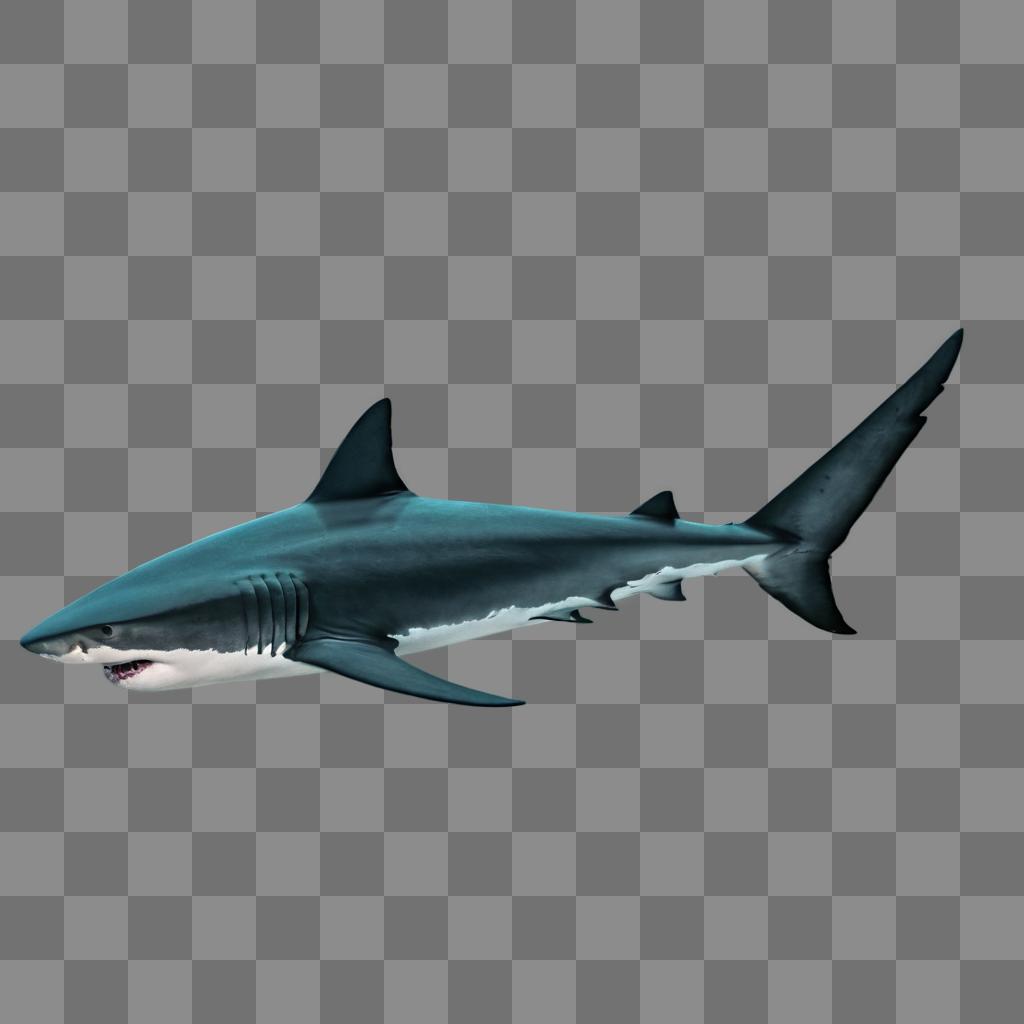}

\vspace{1pt}
\includegraphics[width=0.33\linewidth]{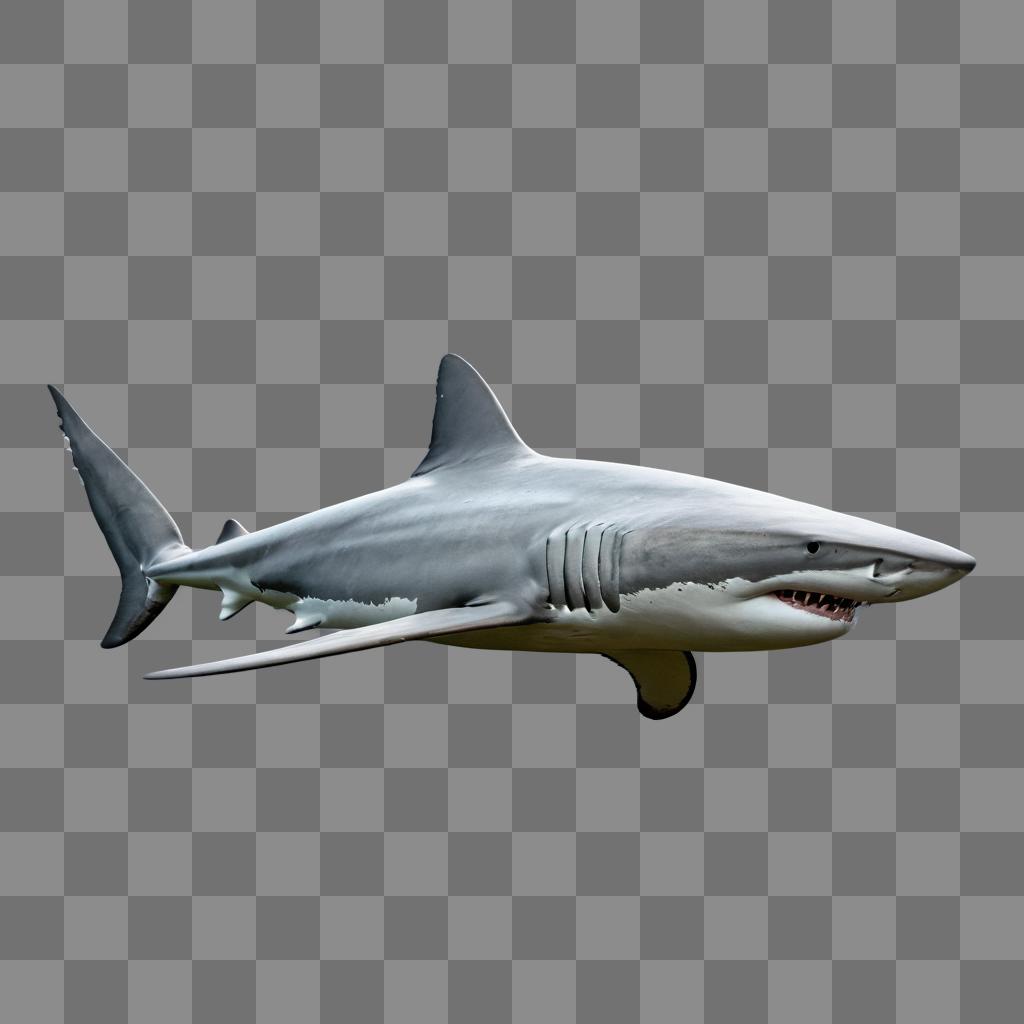}\hfill
\includegraphics[width=0.33\linewidth]{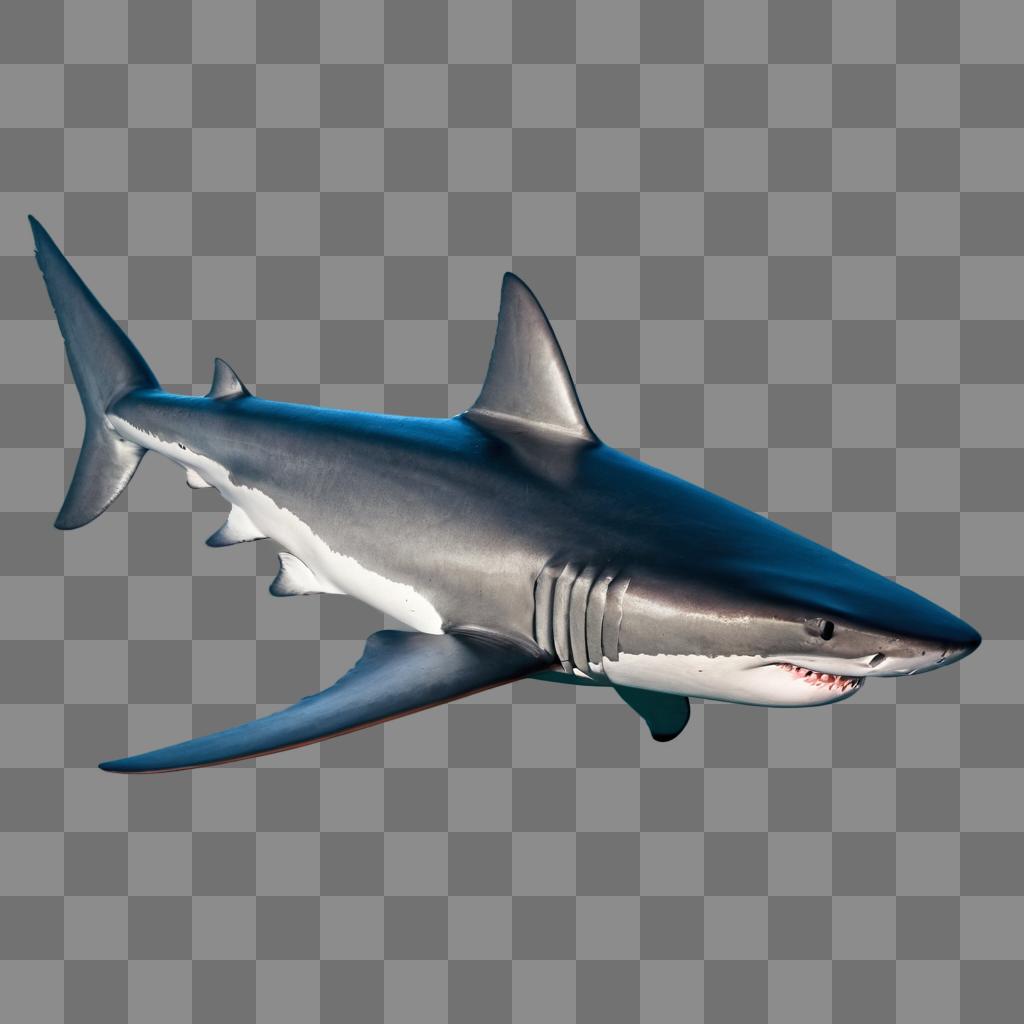}\hfill
\includegraphics[width=0.33\linewidth]{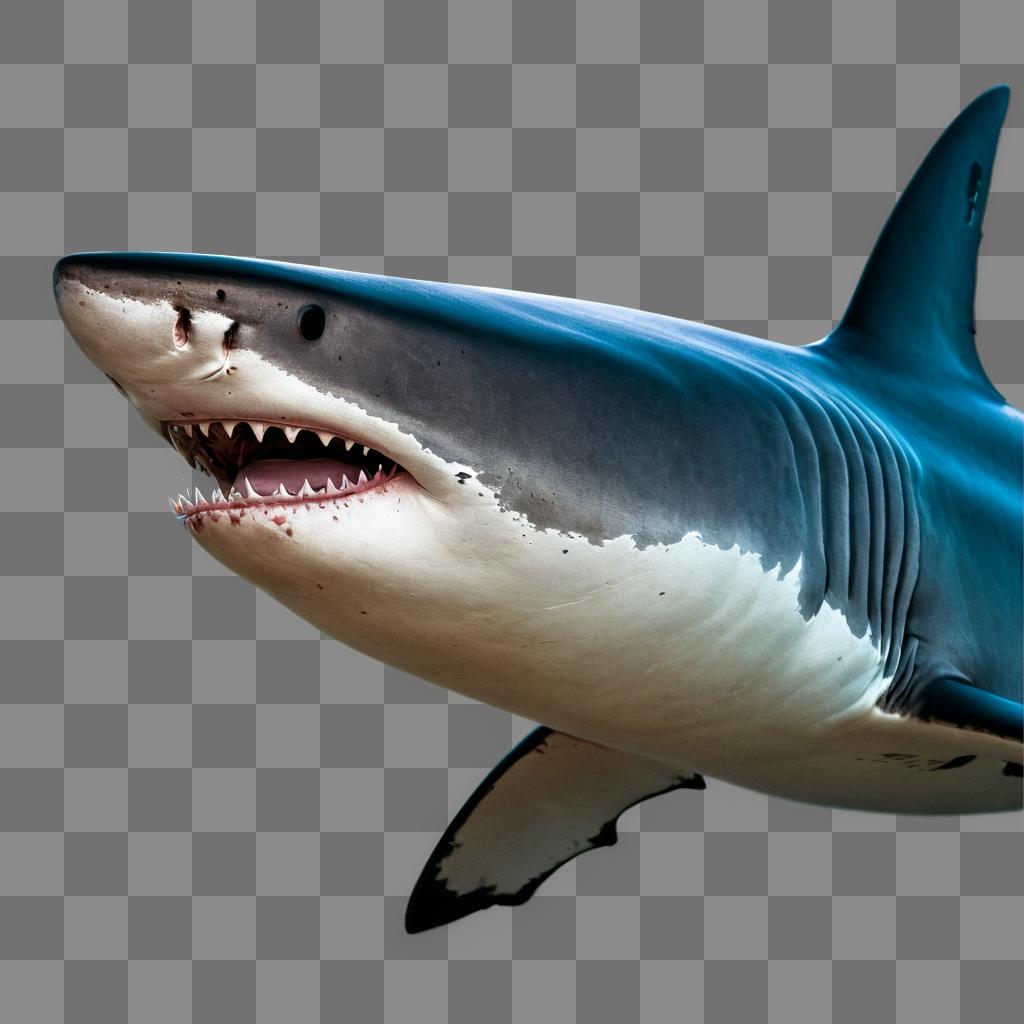}
\caption{Single Transparent Image Results \#12. The prompt is ``shark''. Resolution is $1024\times1024$.}
\label{fig:a12}
\end{minipage}
\end{figure*}

\begin{figure*}

\begin{minipage}{\linewidth}
\includegraphics[width=0.245\linewidth]{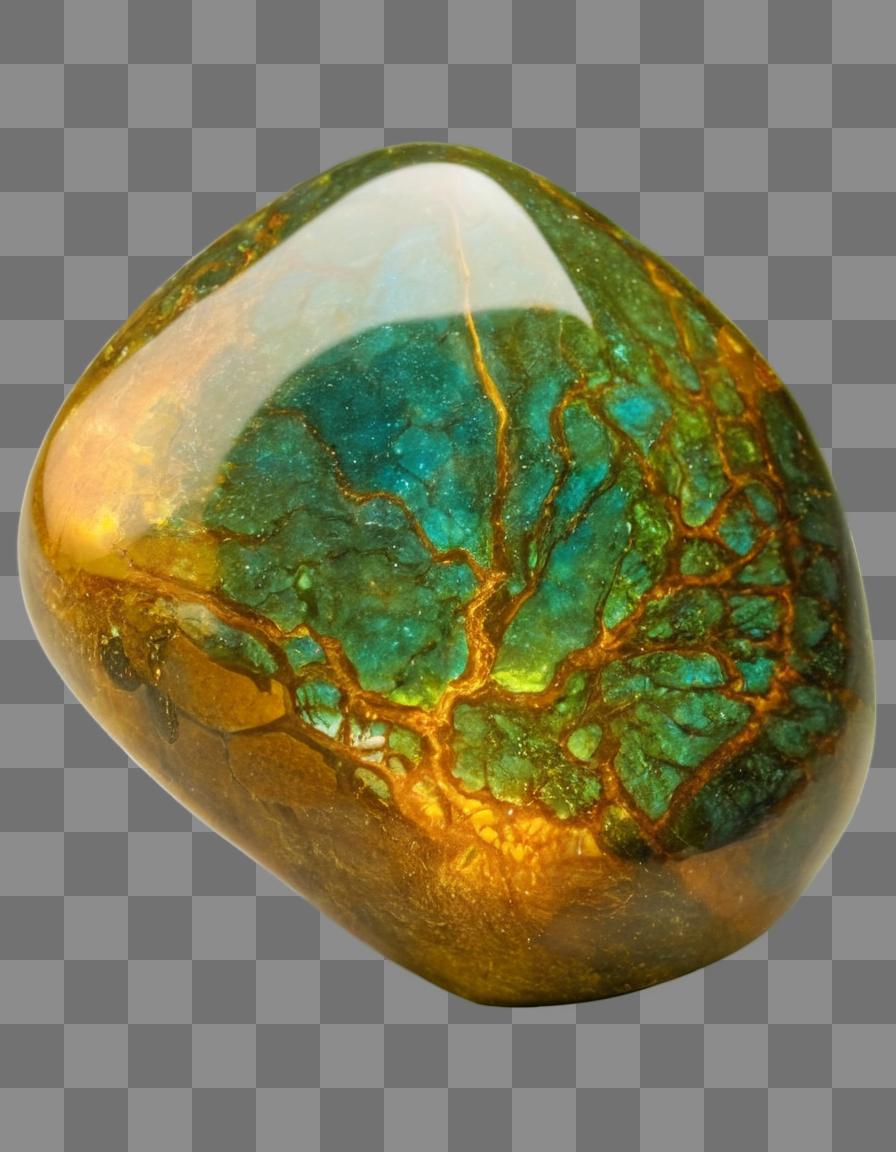}\hfill
\includegraphics[width=0.245\linewidth]{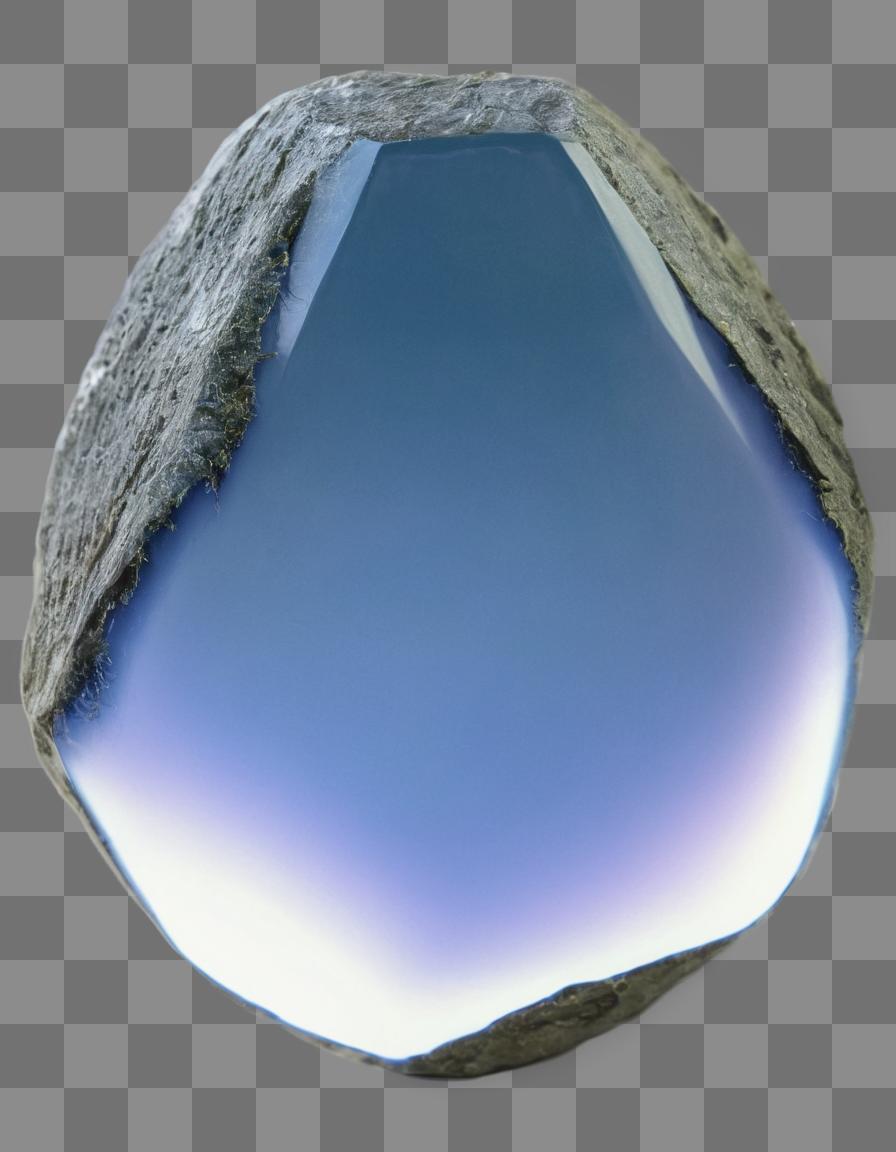}\hfill
\includegraphics[width=0.245\linewidth]{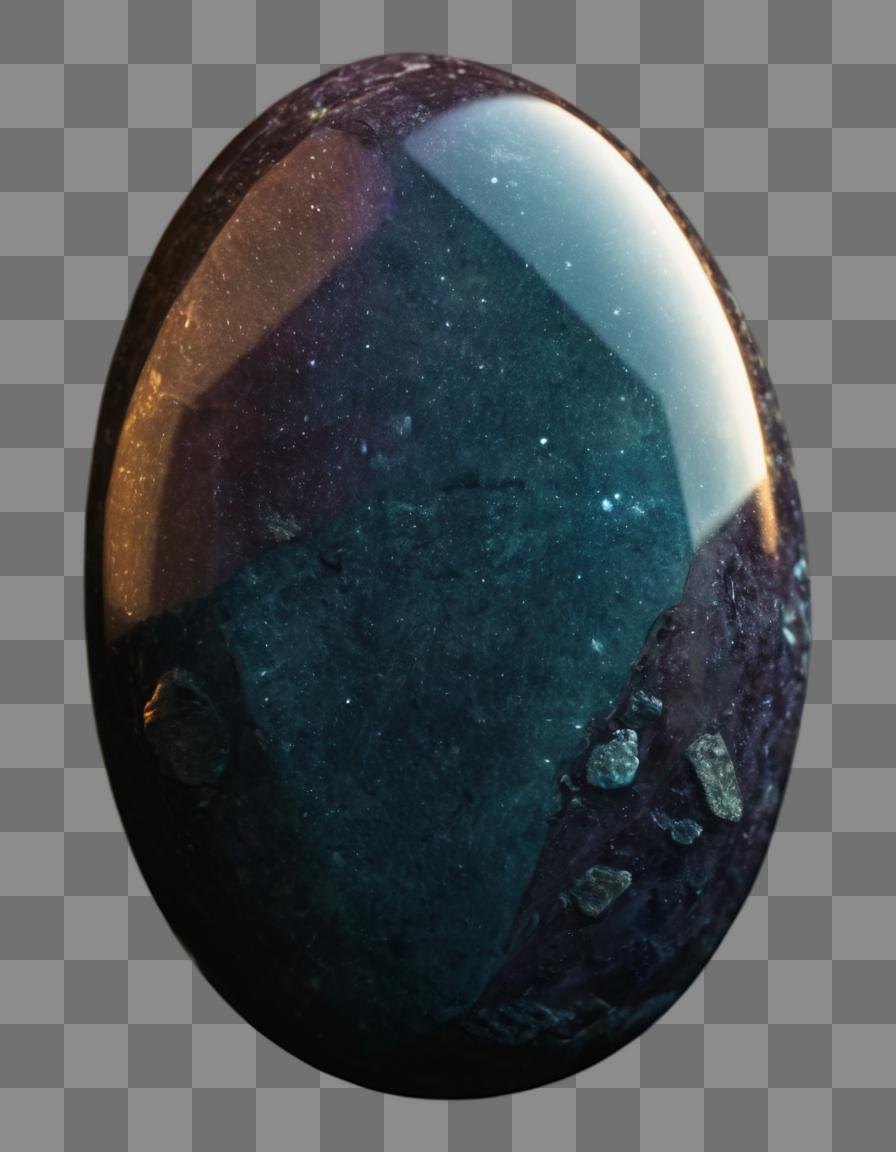}\hfill
\includegraphics[width=0.245\linewidth]{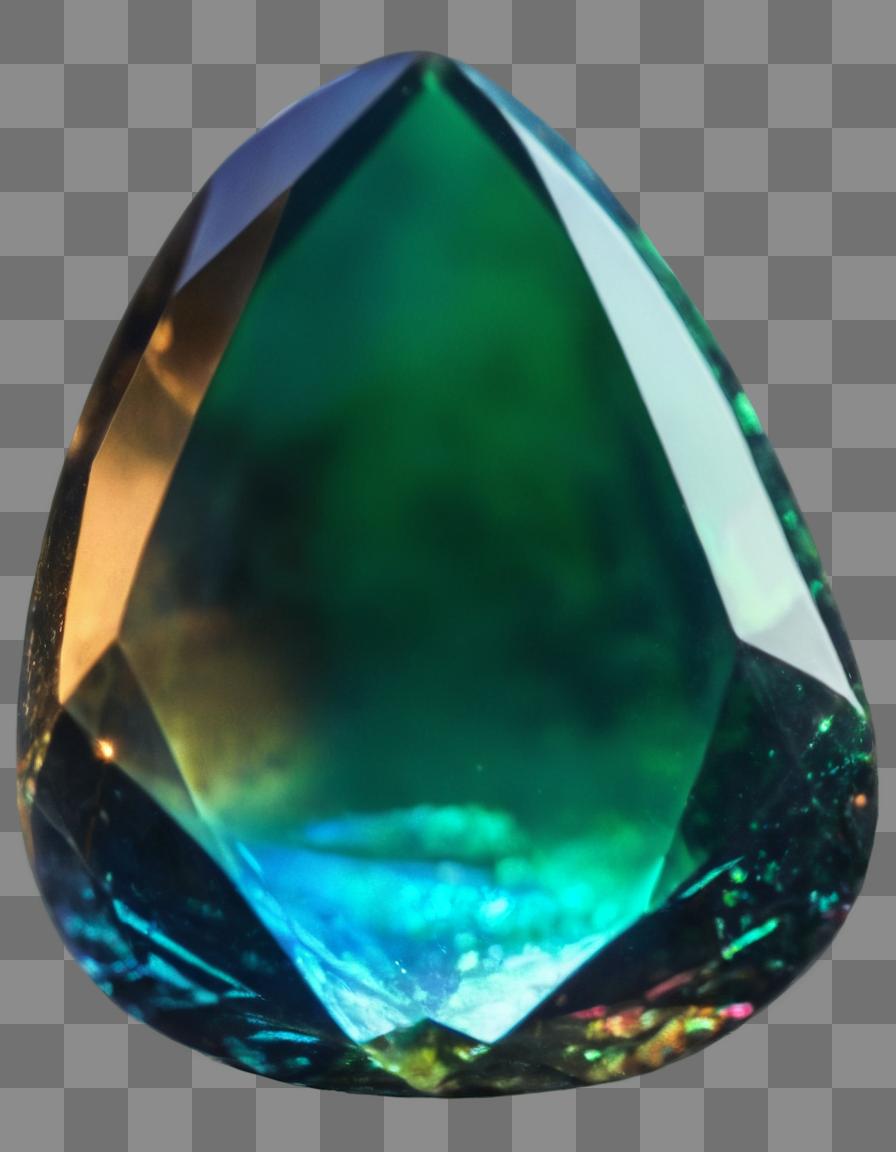}

\vspace{1pt}
\includegraphics[width=0.245\linewidth]{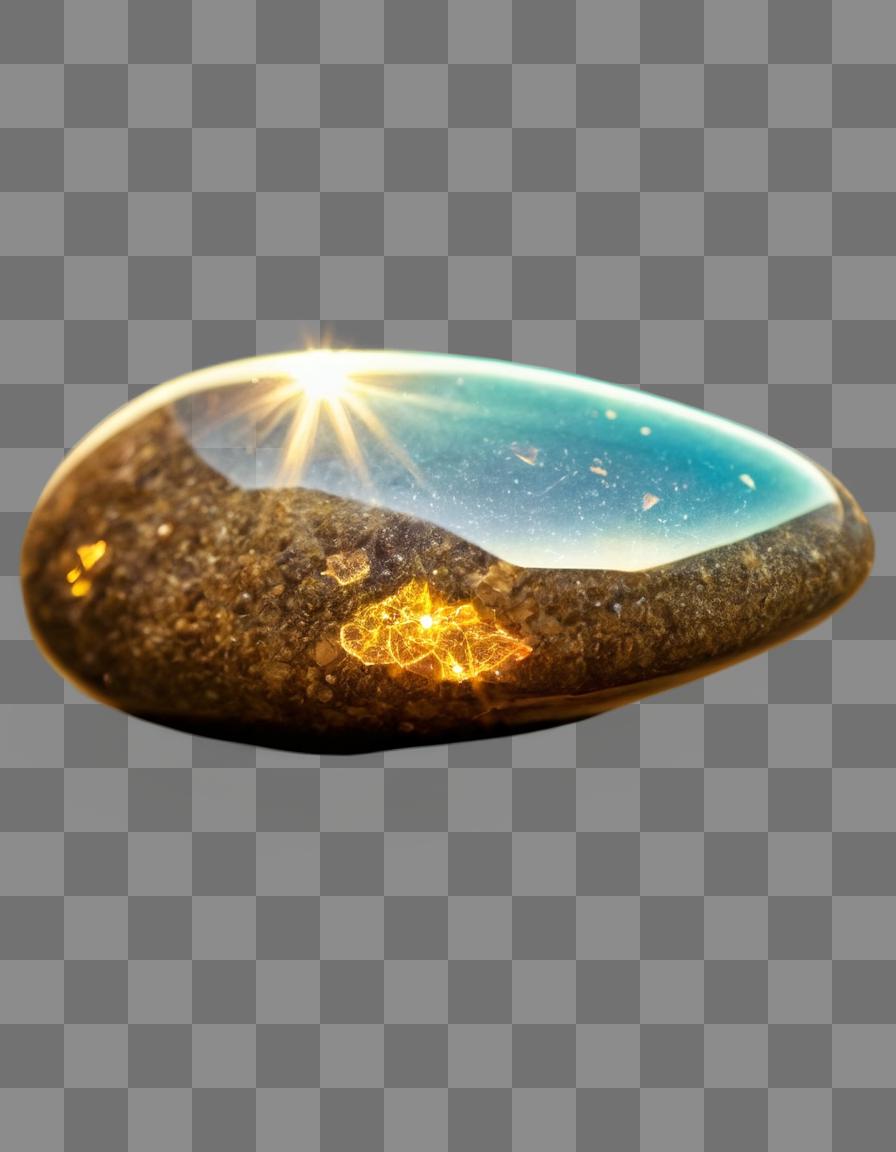}\hfill
\includegraphics[width=0.245\linewidth]{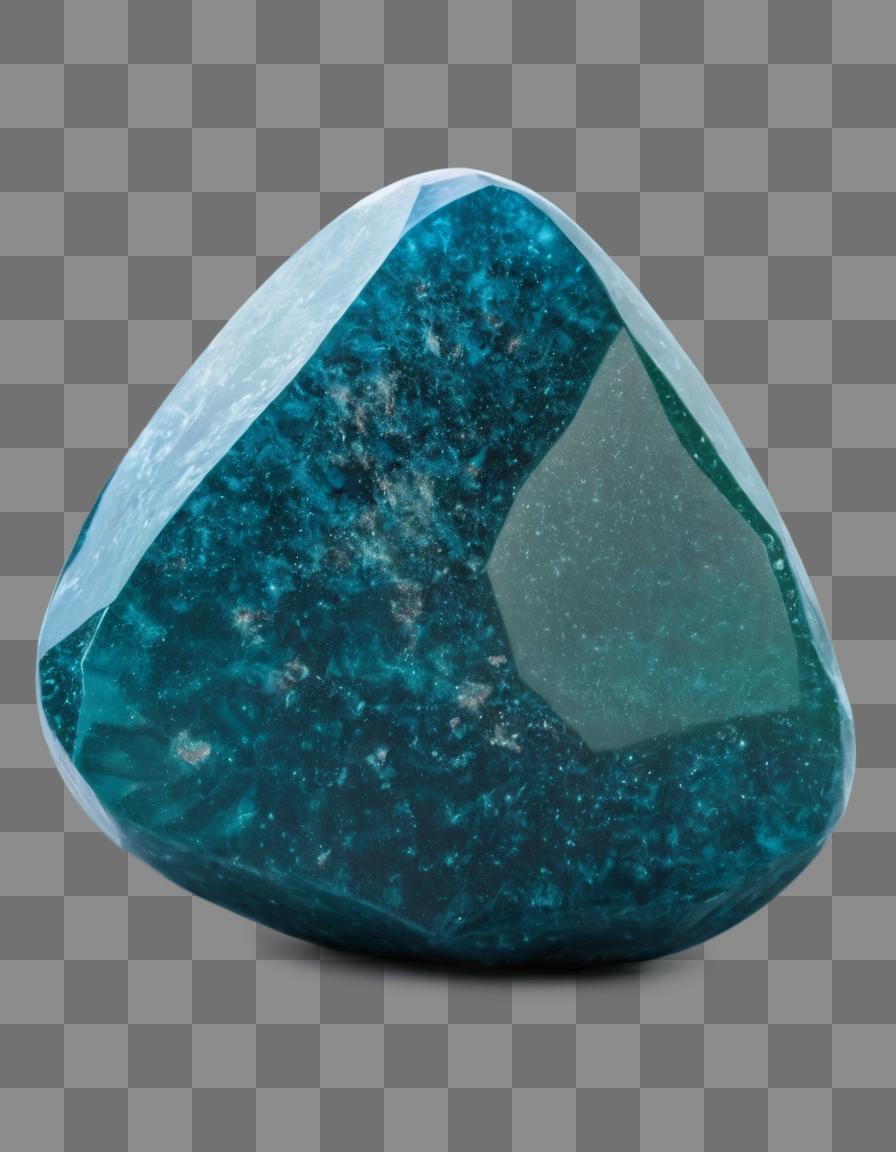}\hfill
\includegraphics[width=0.245\linewidth]{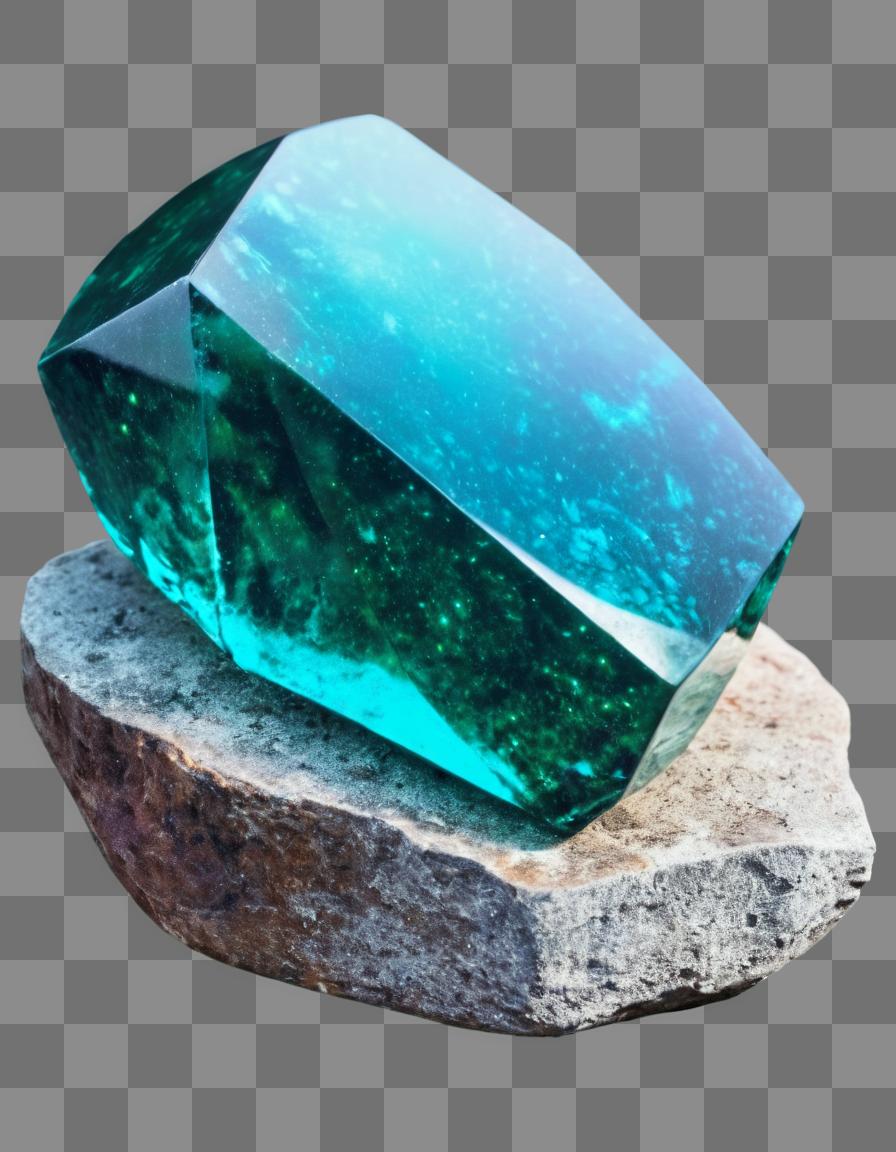}\hfill
\includegraphics[width=0.245\linewidth]{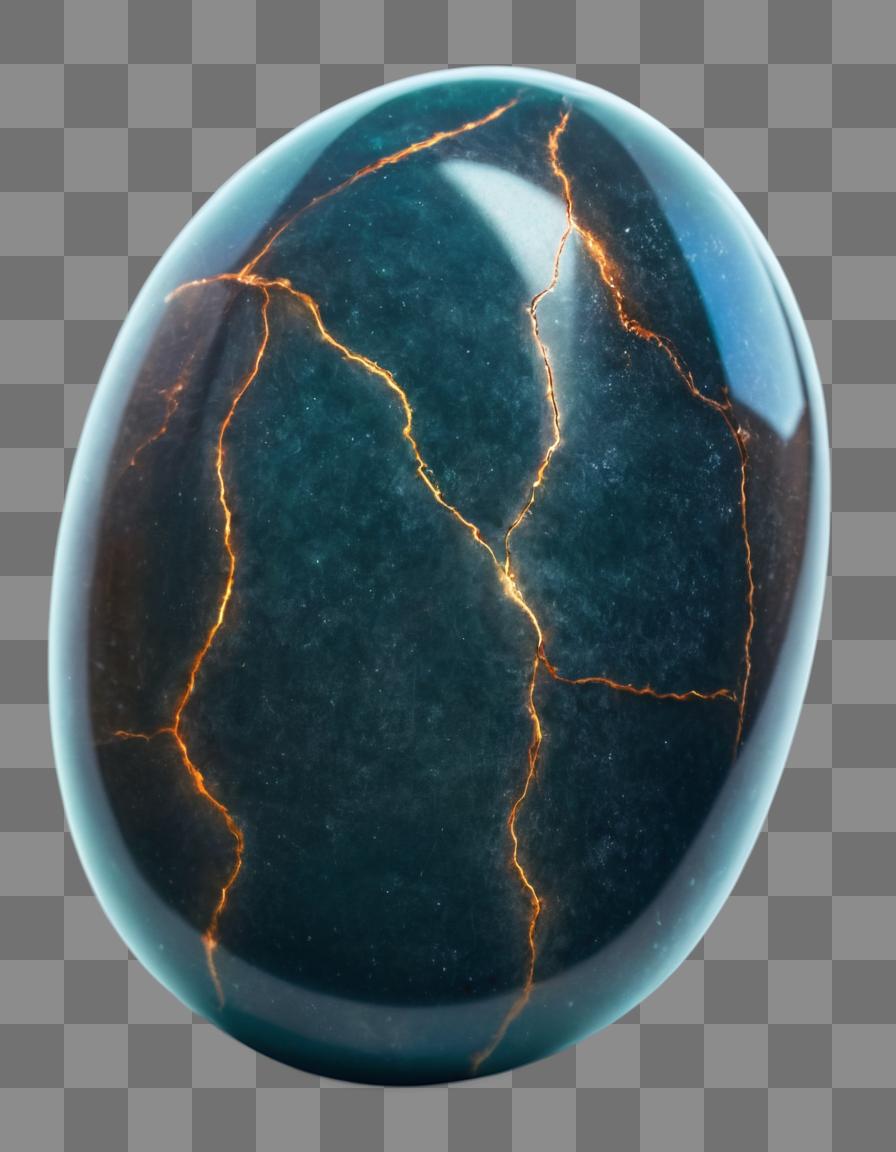}

\vspace{1pt}
\includegraphics[width=0.245\linewidth]{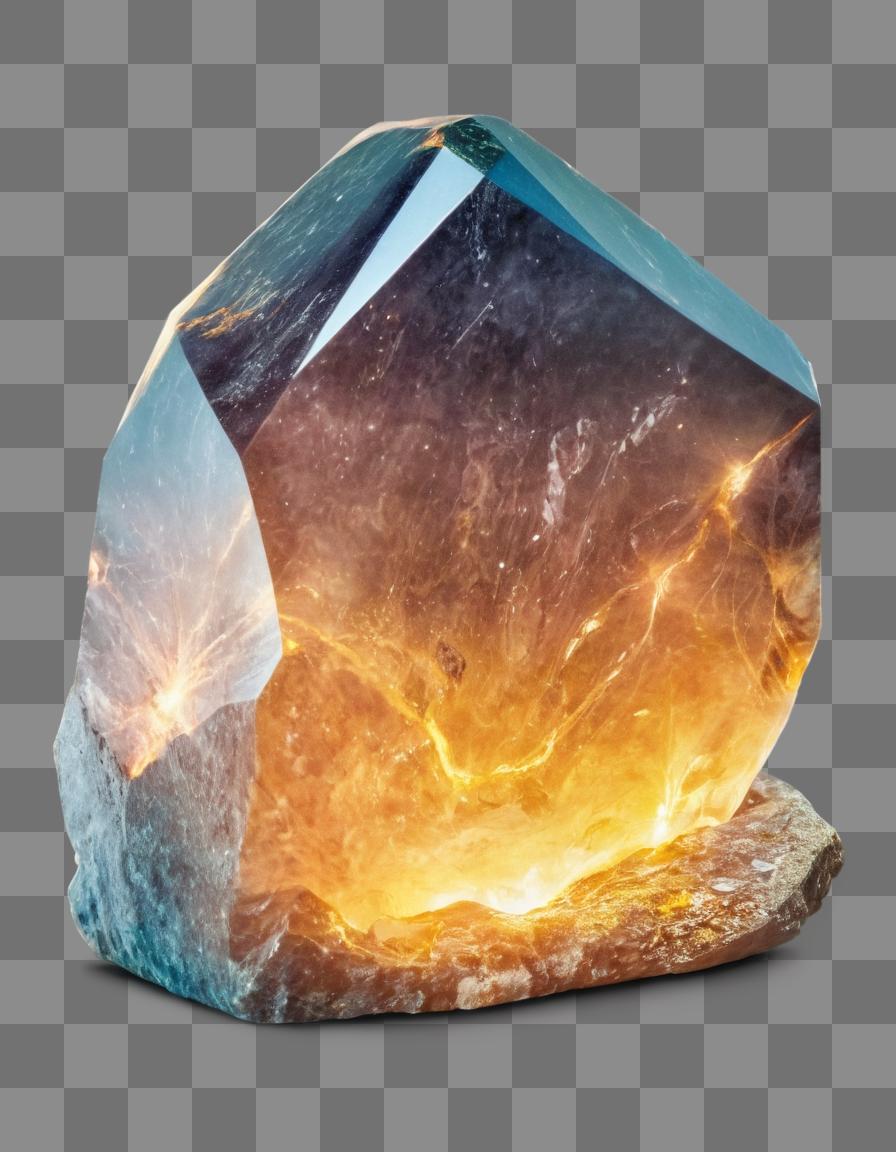}\hfill
\includegraphics[width=0.245\linewidth]{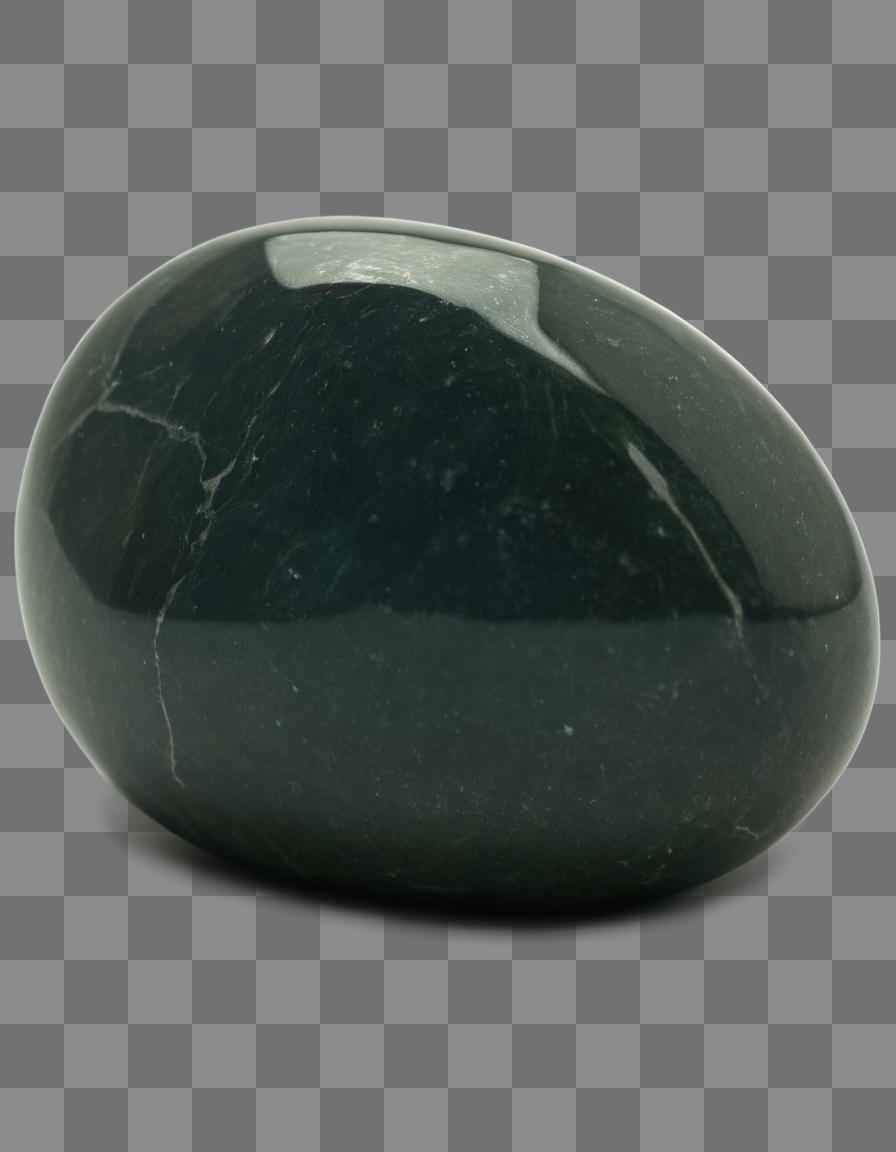}\hfill
\includegraphics[width=0.245\linewidth]{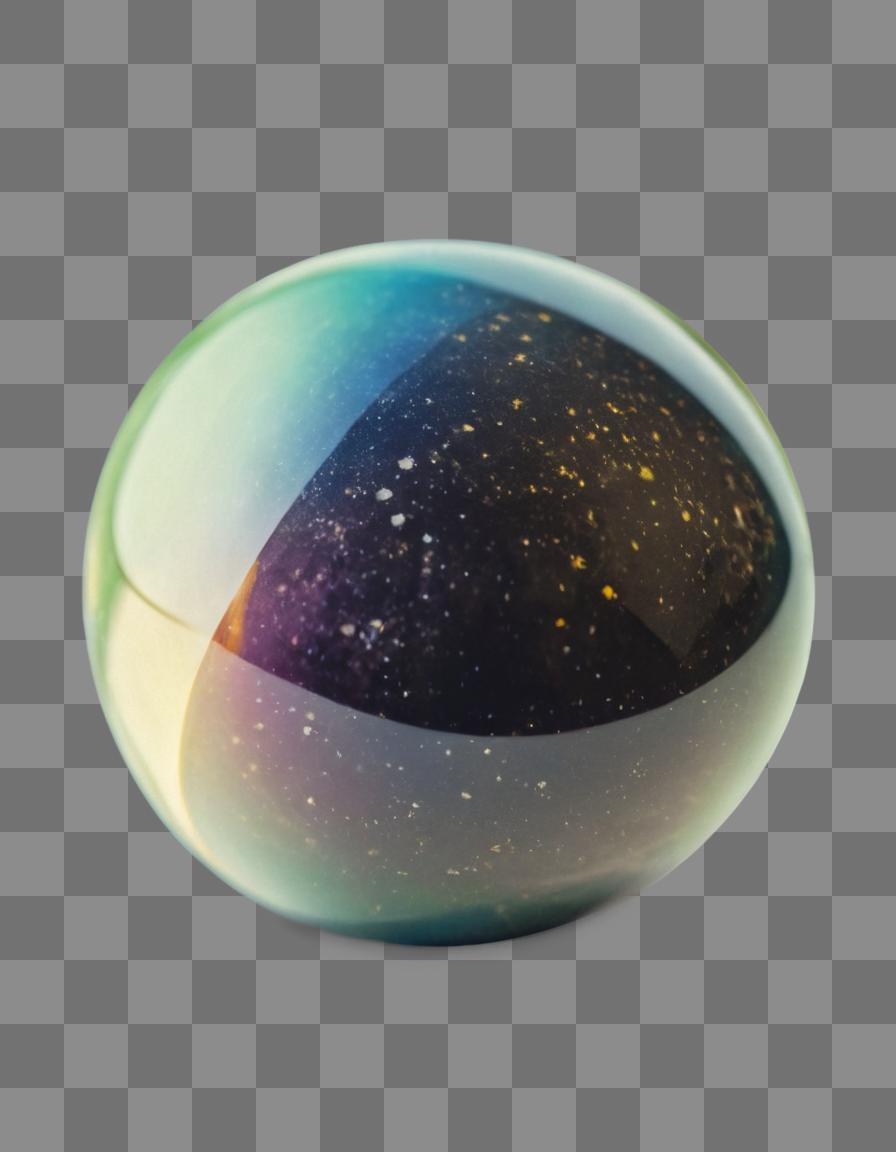}\hfill
\includegraphics[width=0.245\linewidth]{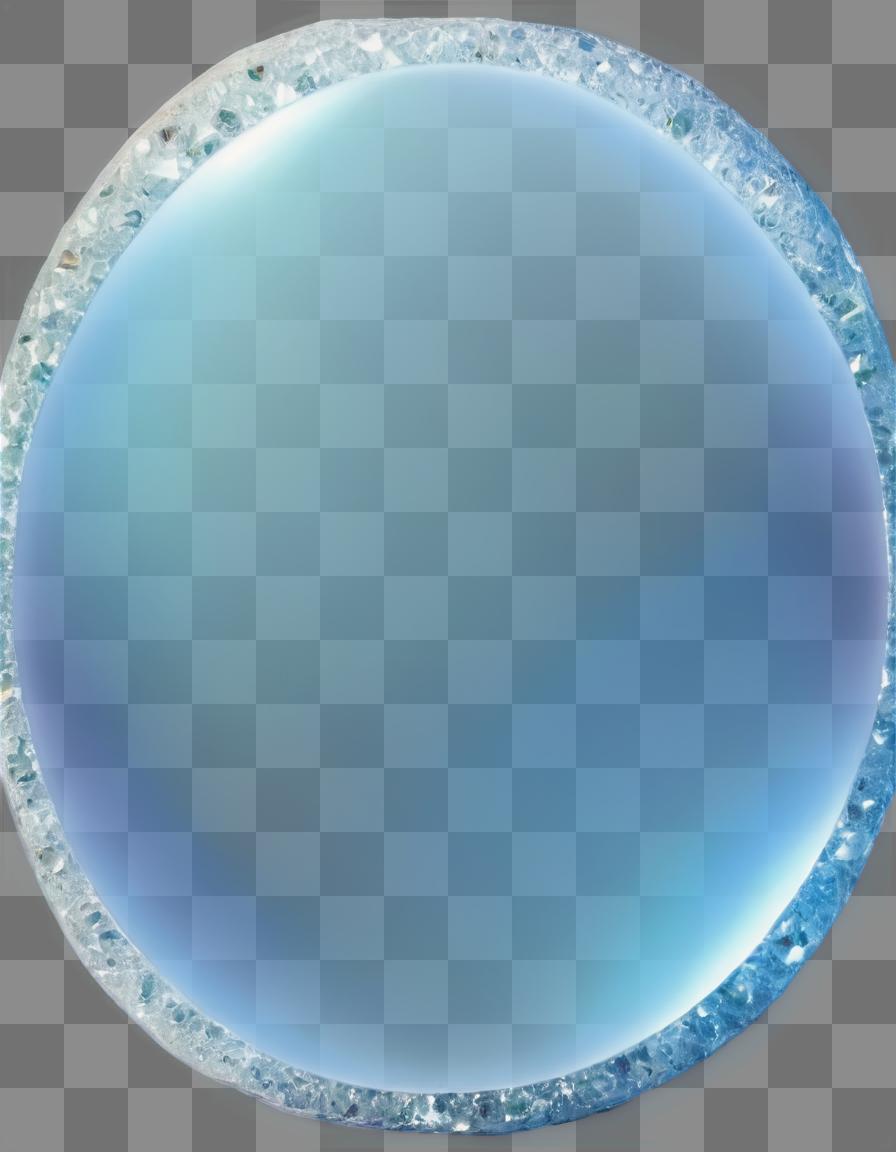}
\caption{Single Transparent Image Results \#13. The prompt is ``magic stone''. Resolution is $896\times1152$.}
\label{fig:a13}
\end{minipage}
\end{figure*}

\begin{figure*}

\begin{minipage}{\linewidth}
\includegraphics[width=0.245\linewidth]{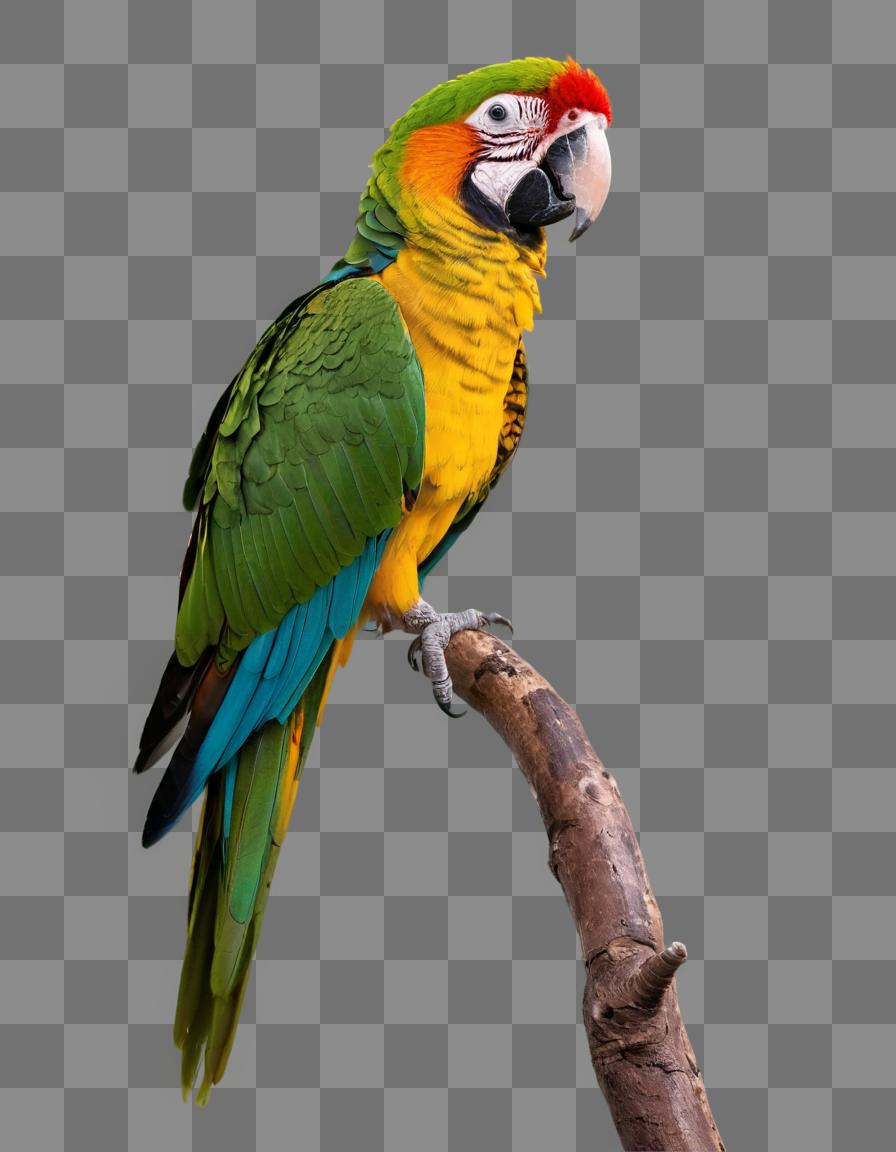}\hfill
\includegraphics[width=0.245\linewidth]{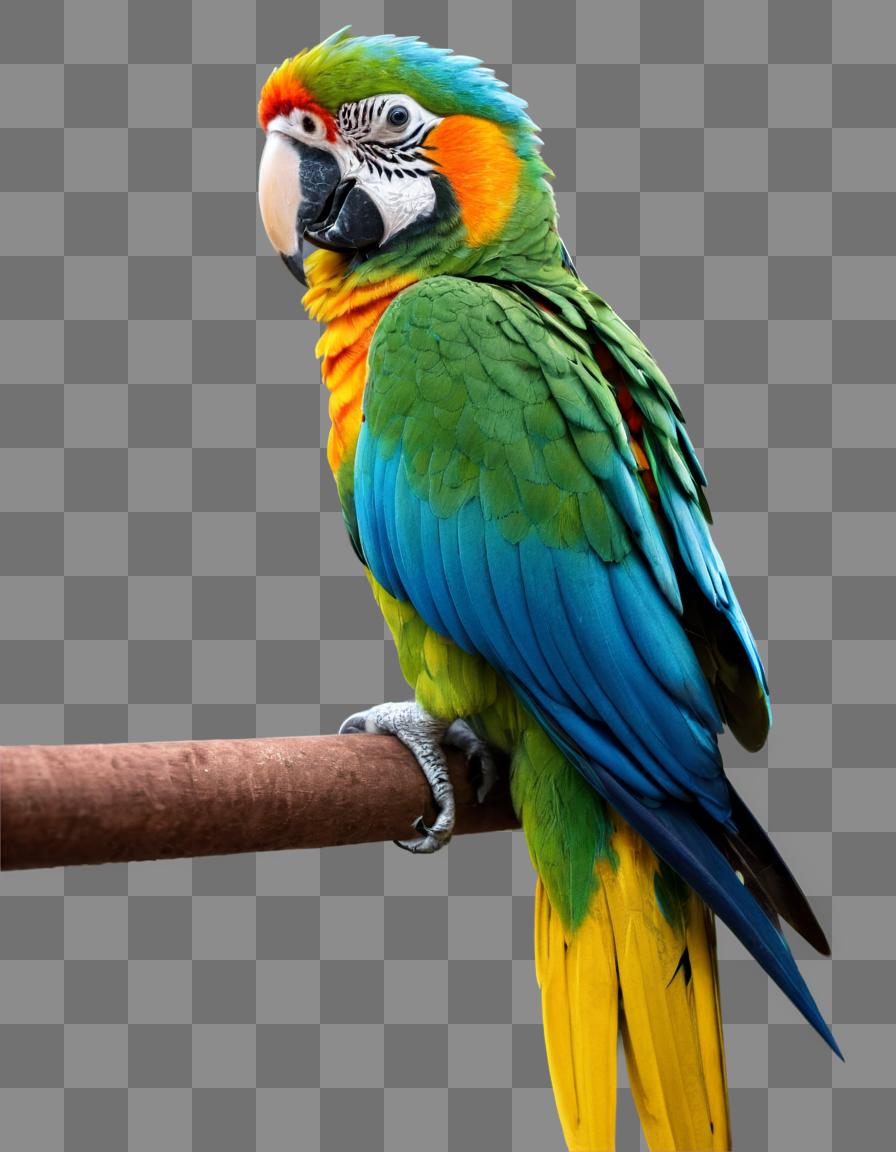}\hfill
\includegraphics[width=0.245\linewidth]{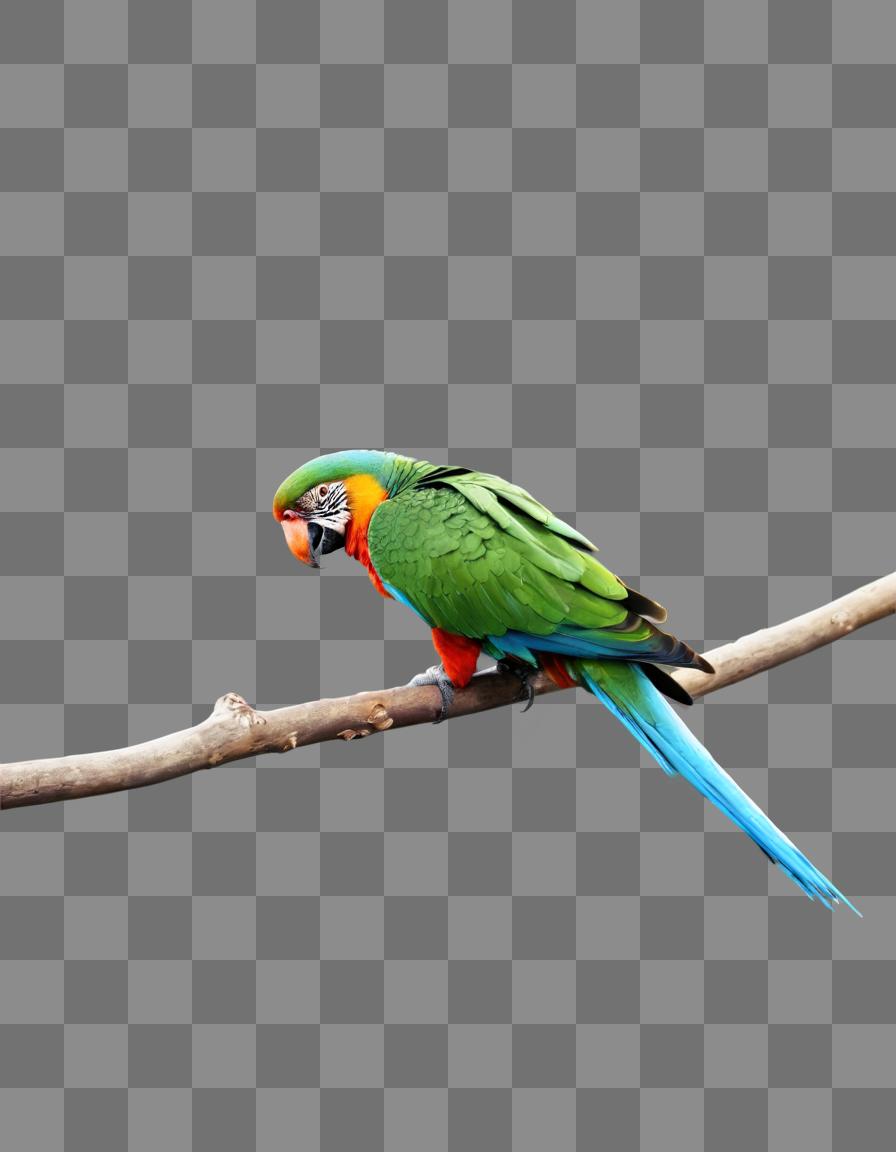}\hfill
\includegraphics[width=0.245\linewidth]{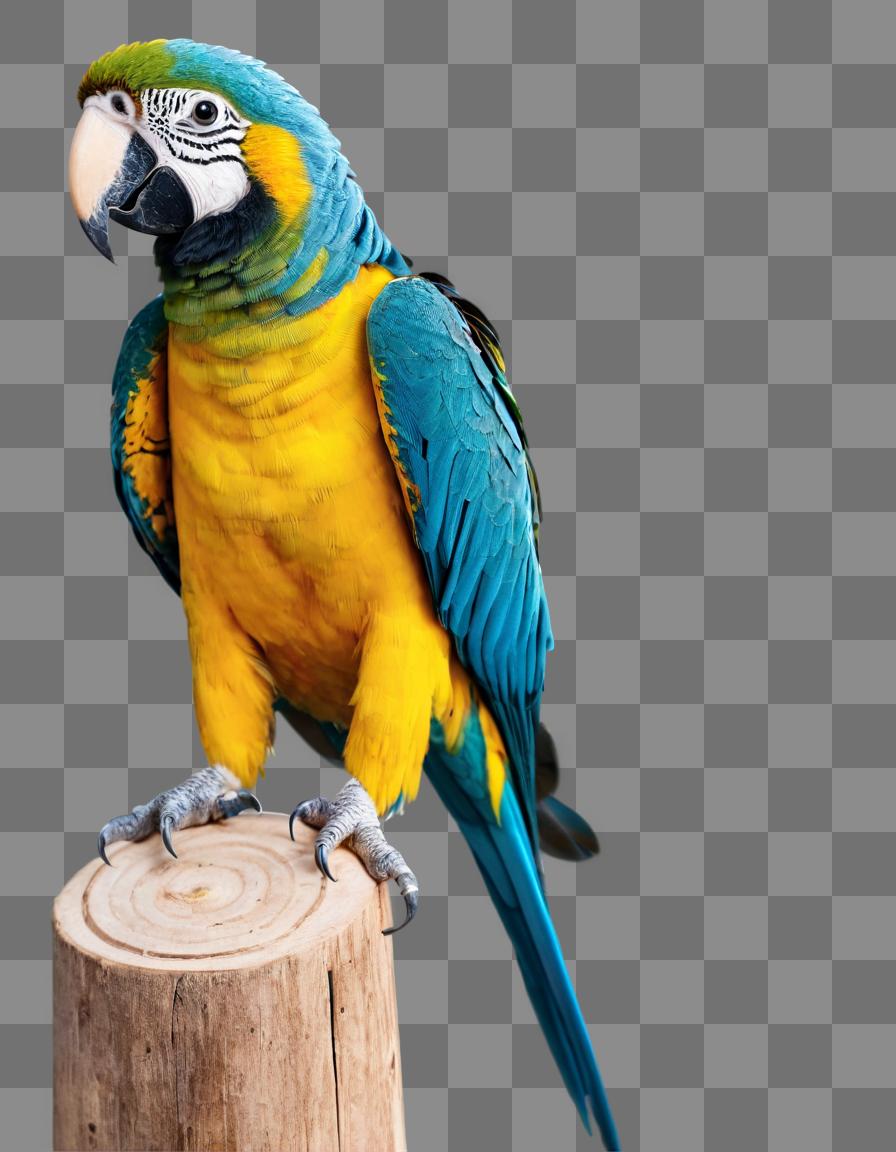}

\vspace{1pt}
\includegraphics[width=0.245\linewidth]{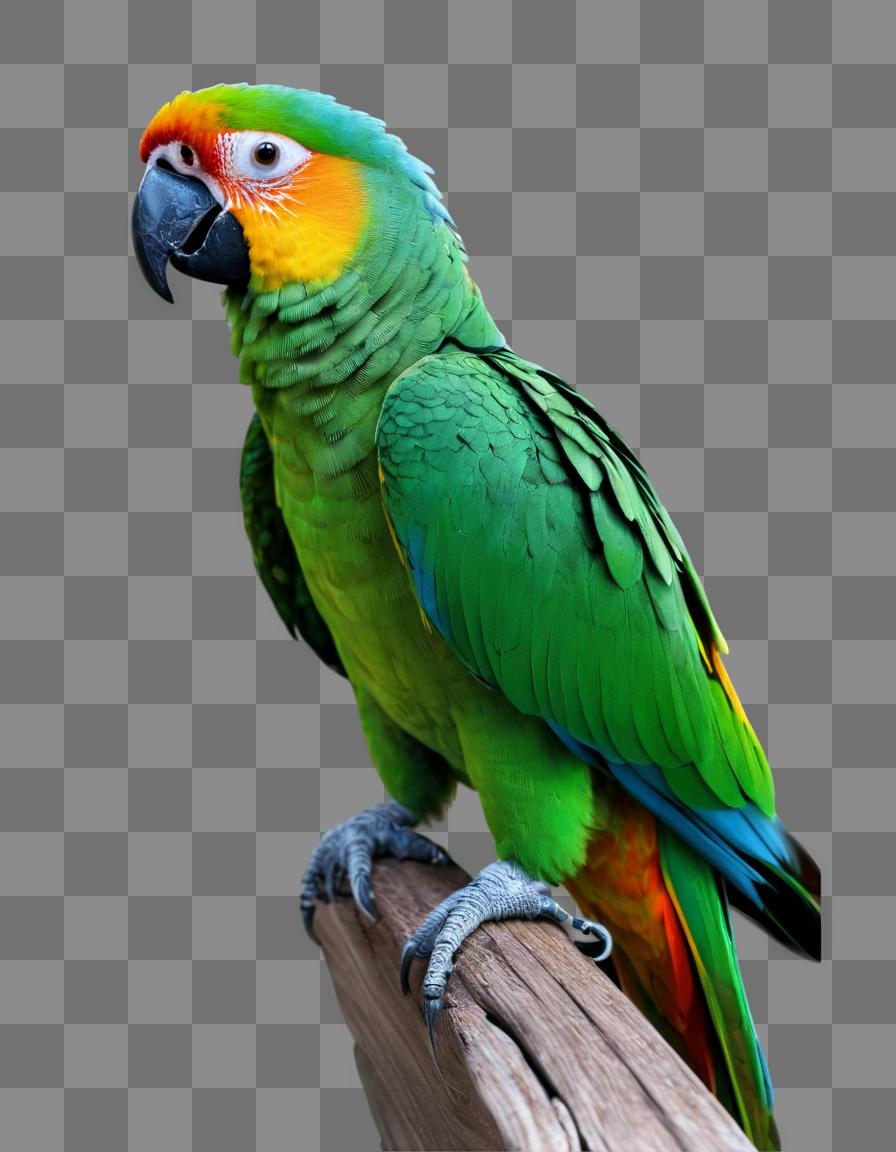}\hfill
\includegraphics[width=0.245\linewidth]{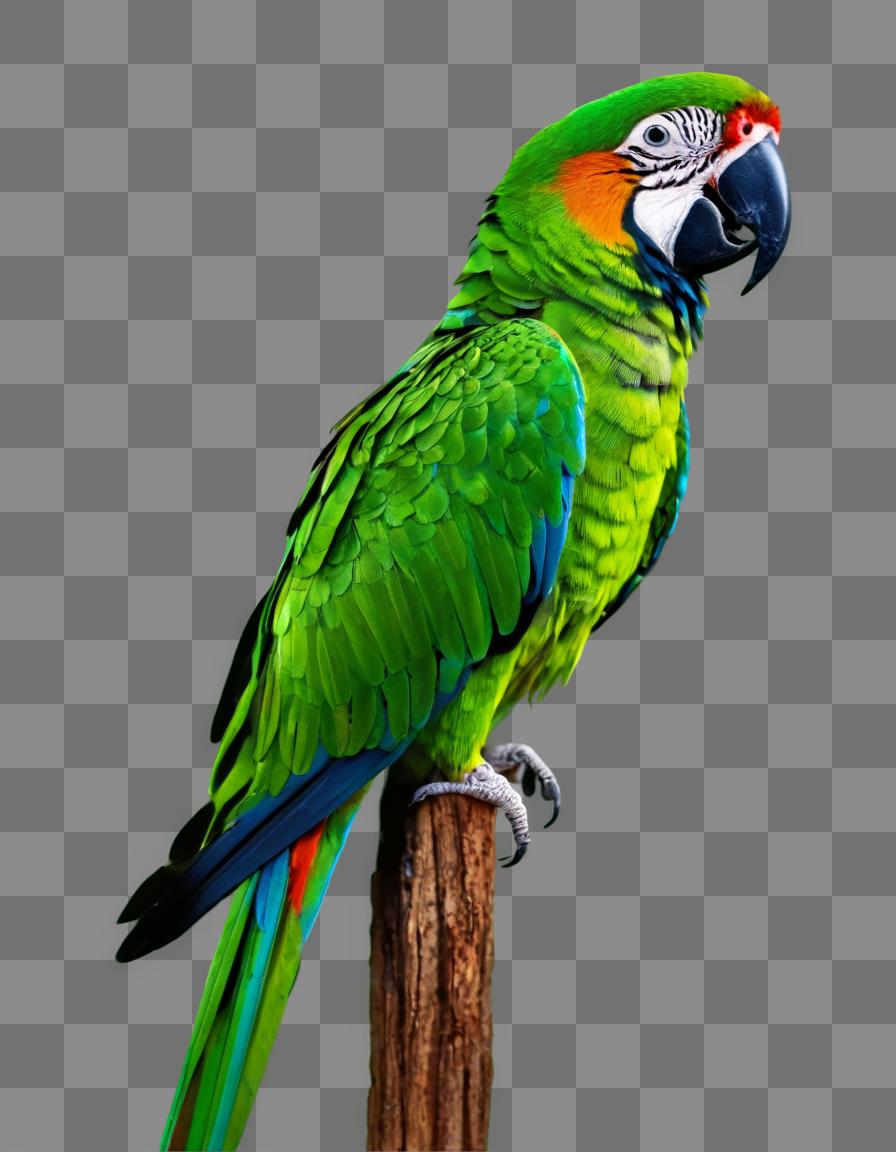}\hfill
\includegraphics[width=0.245\linewidth]{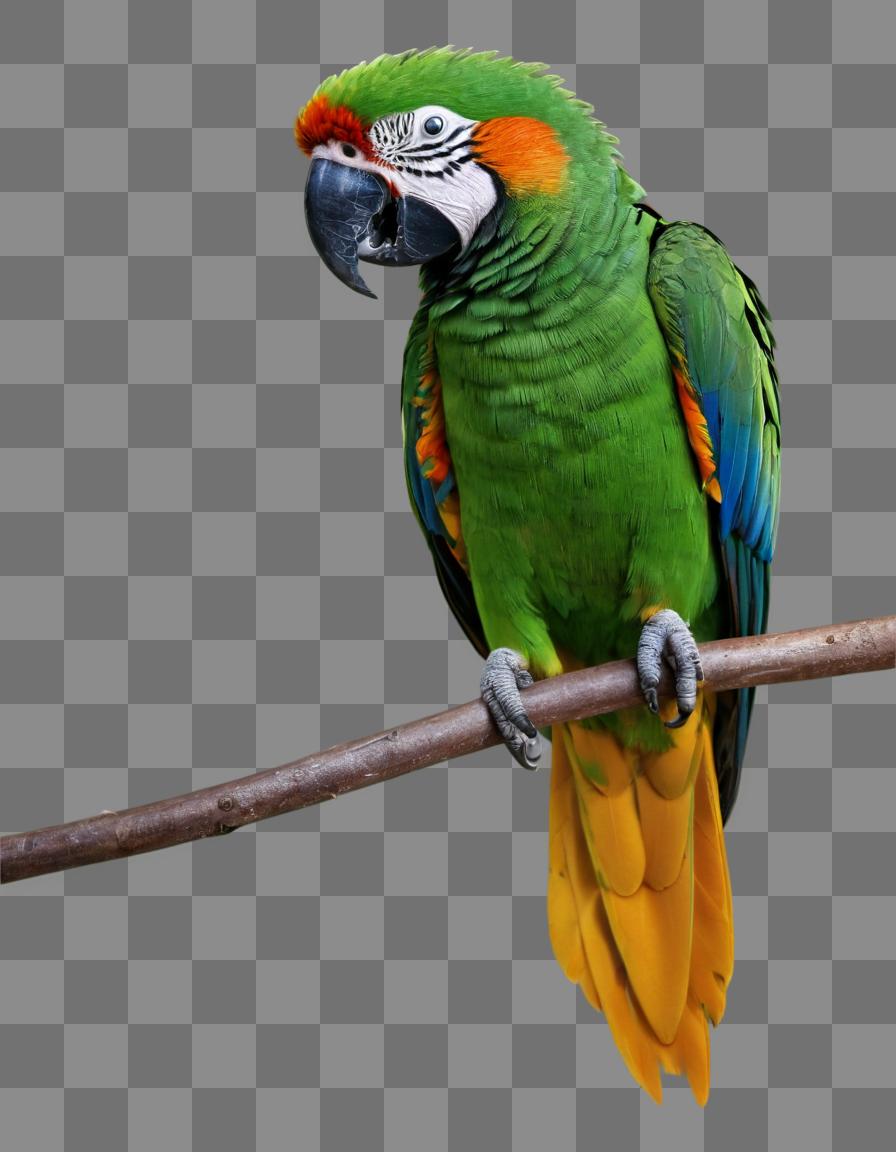}\hfill
\includegraphics[width=0.245\linewidth]{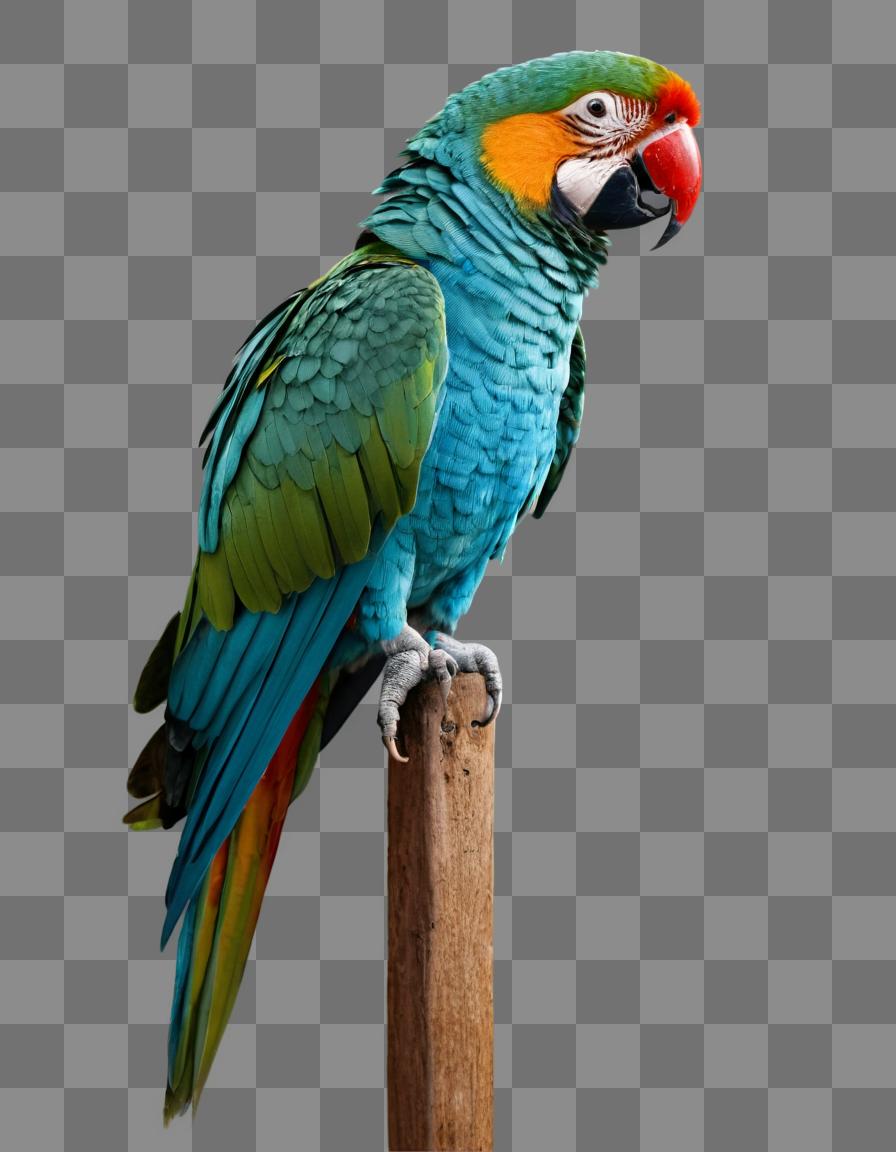}

\vspace{1pt}
\includegraphics[width=0.245\linewidth]{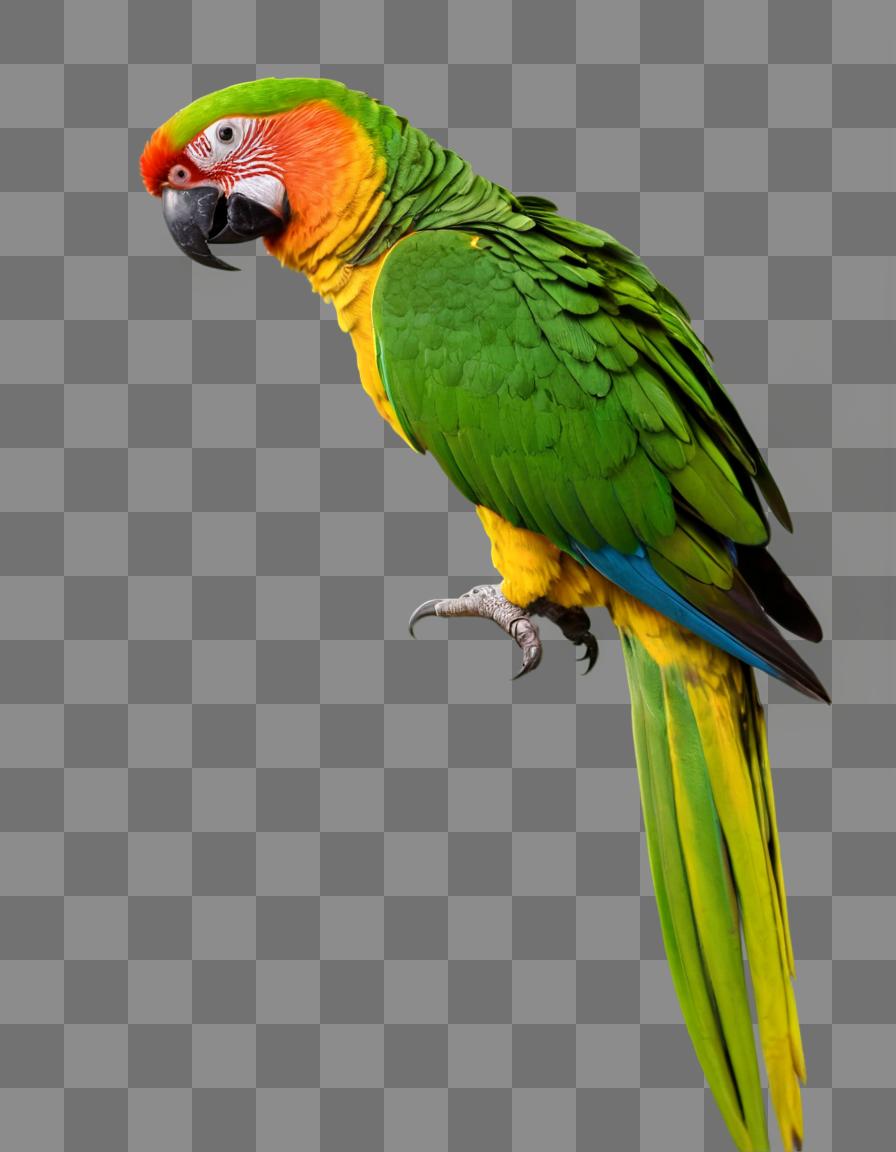}\hfill
\includegraphics[width=0.245\linewidth]{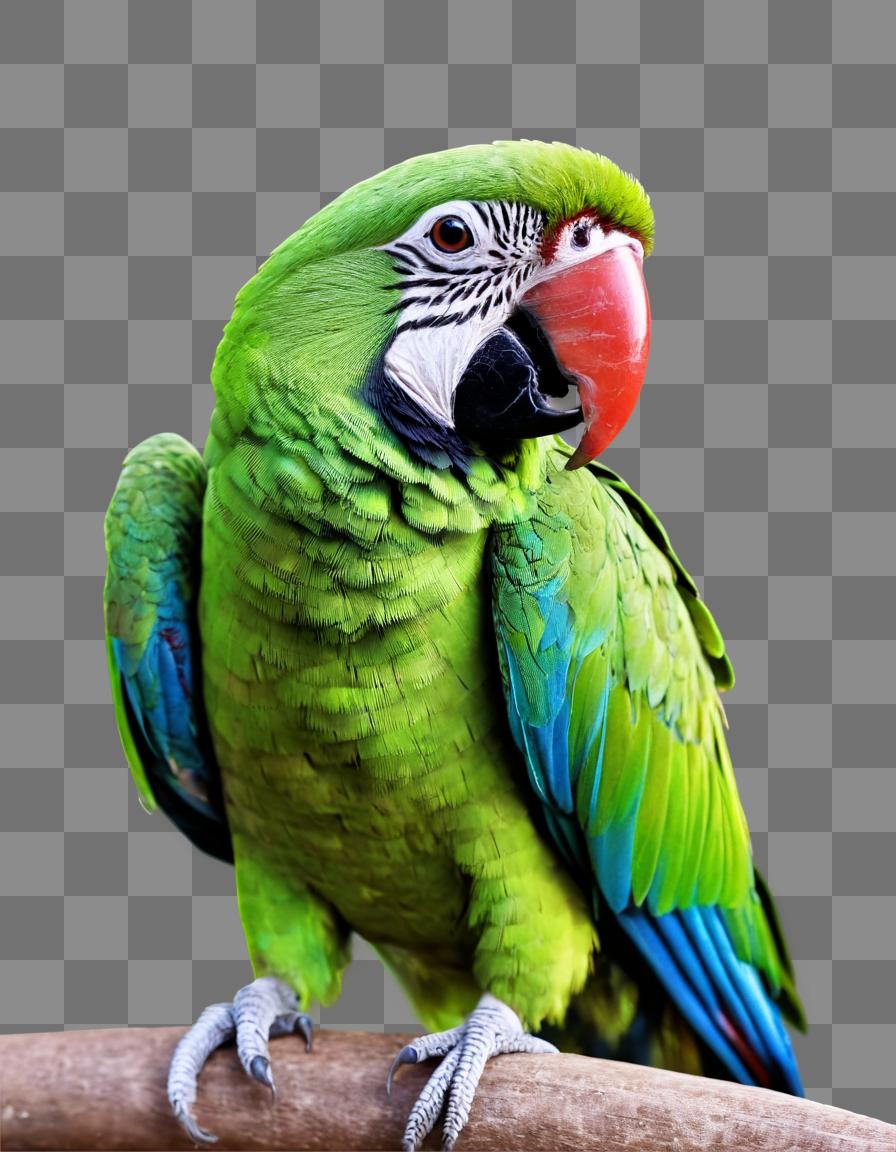}\hfill
\includegraphics[width=0.245\linewidth]{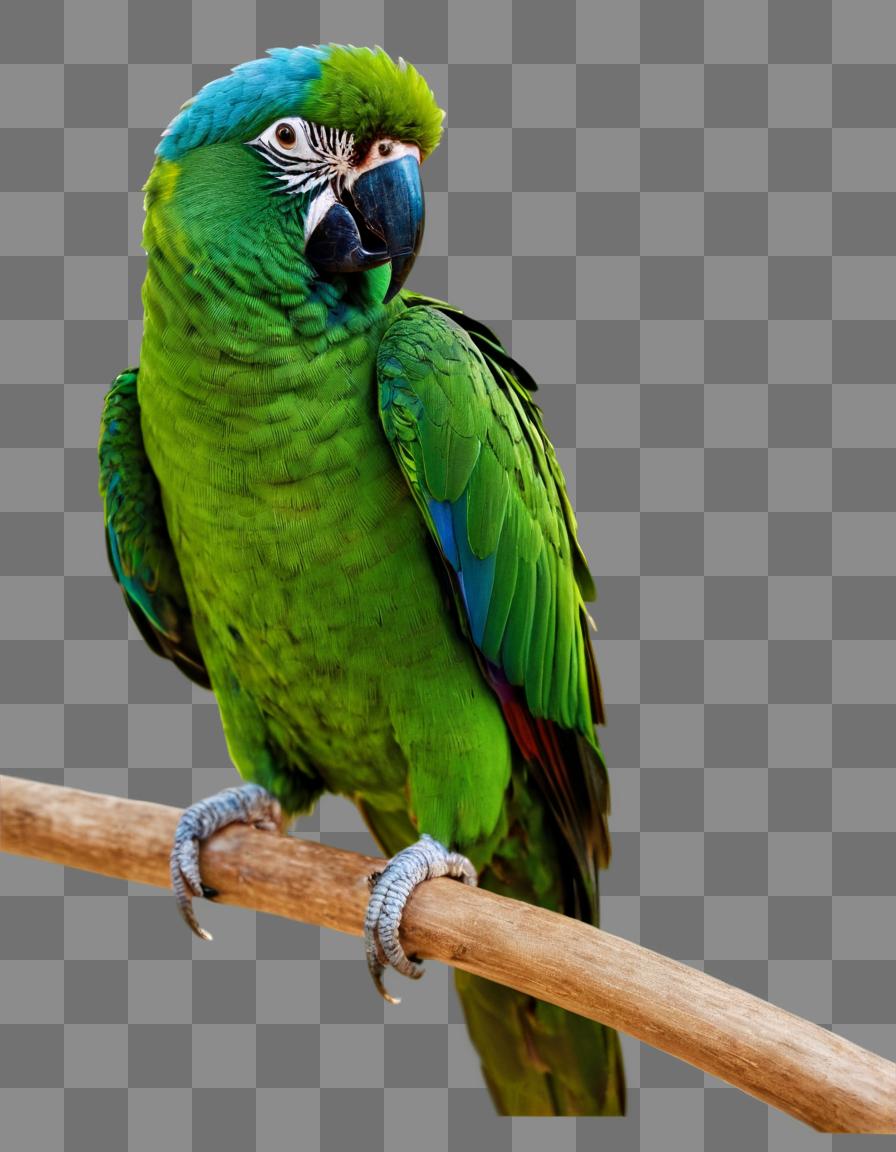}\hfill
\includegraphics[width=0.245\linewidth]{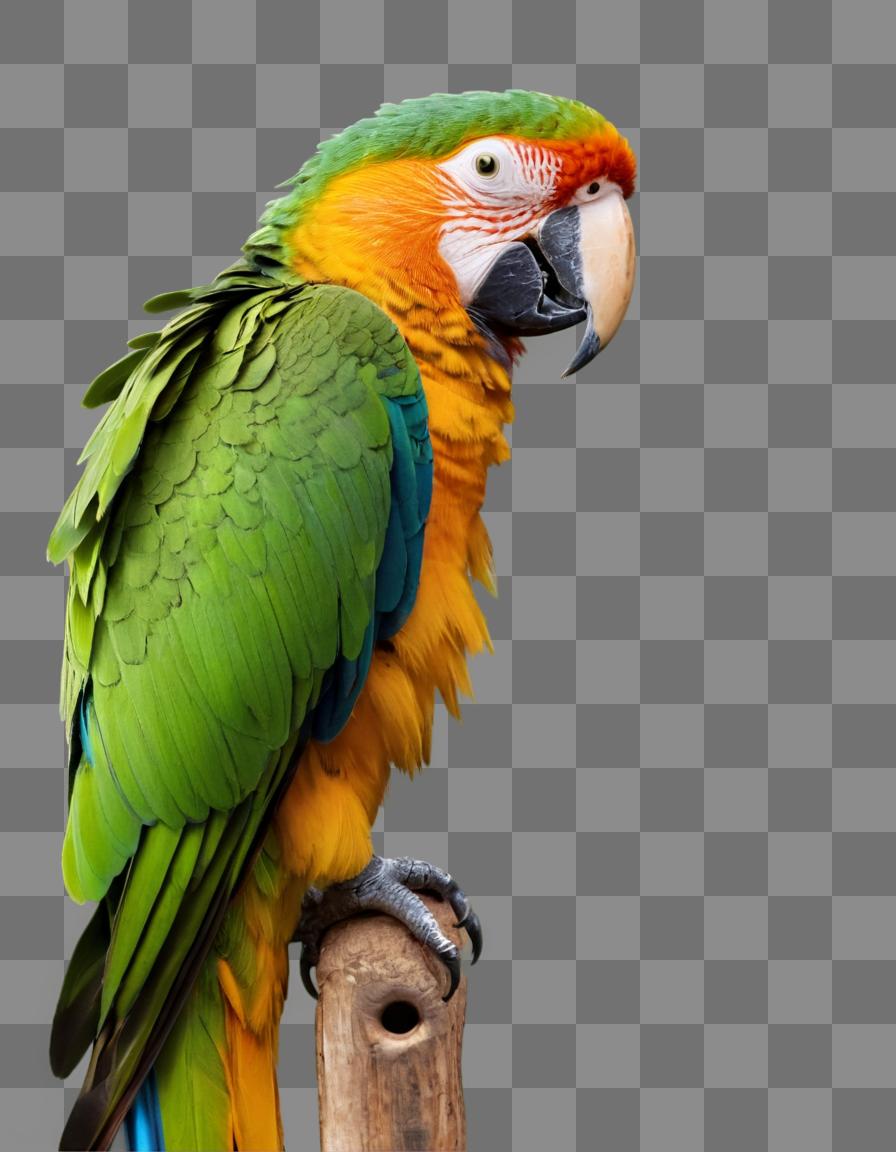}
\caption{Single Transparent Image Results \#14. The prompt is ``parrot, green fur''. Resolution is $896\times1152$.}
\label{fig:a14}
\end{minipage}
\end{figure*}

\begin{figure*}

\begin{minipage}{\linewidth}
\includegraphics[width=0.245\linewidth]{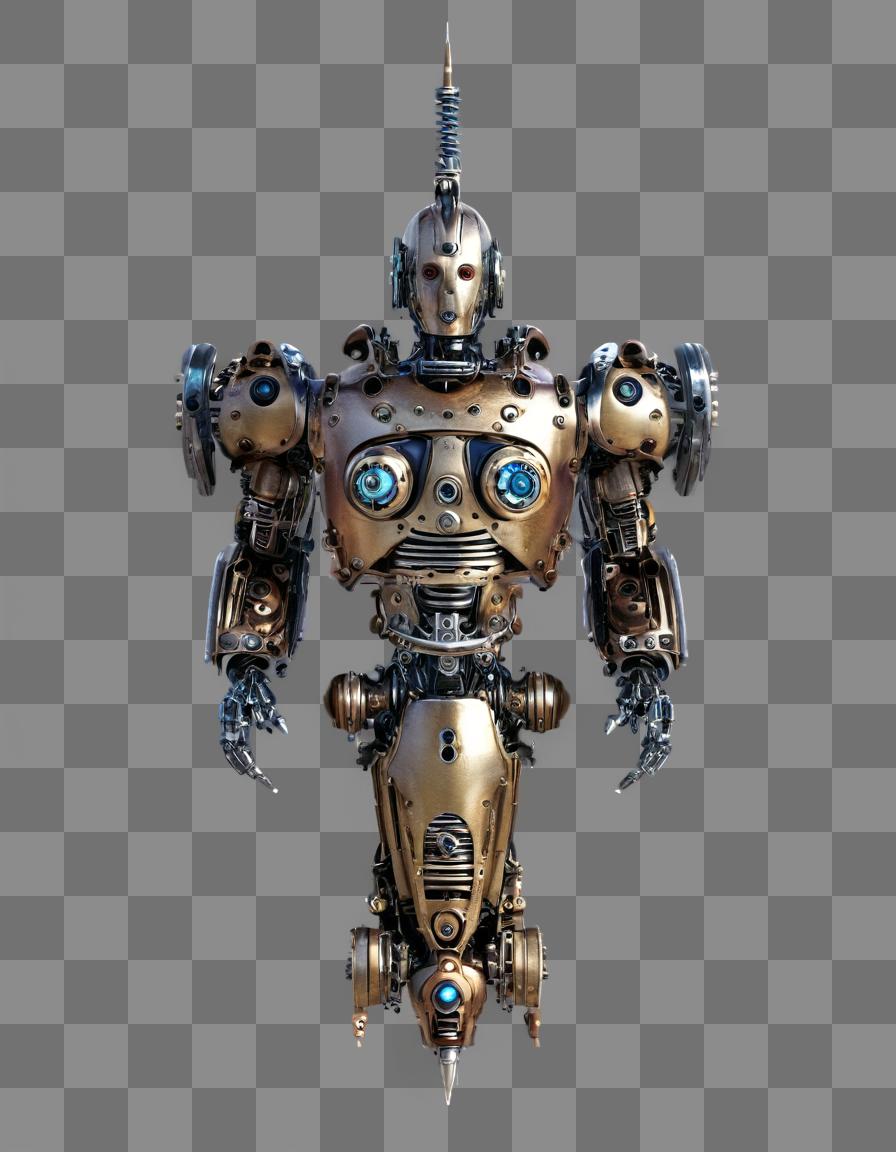}\hfill
\includegraphics[width=0.245\linewidth]{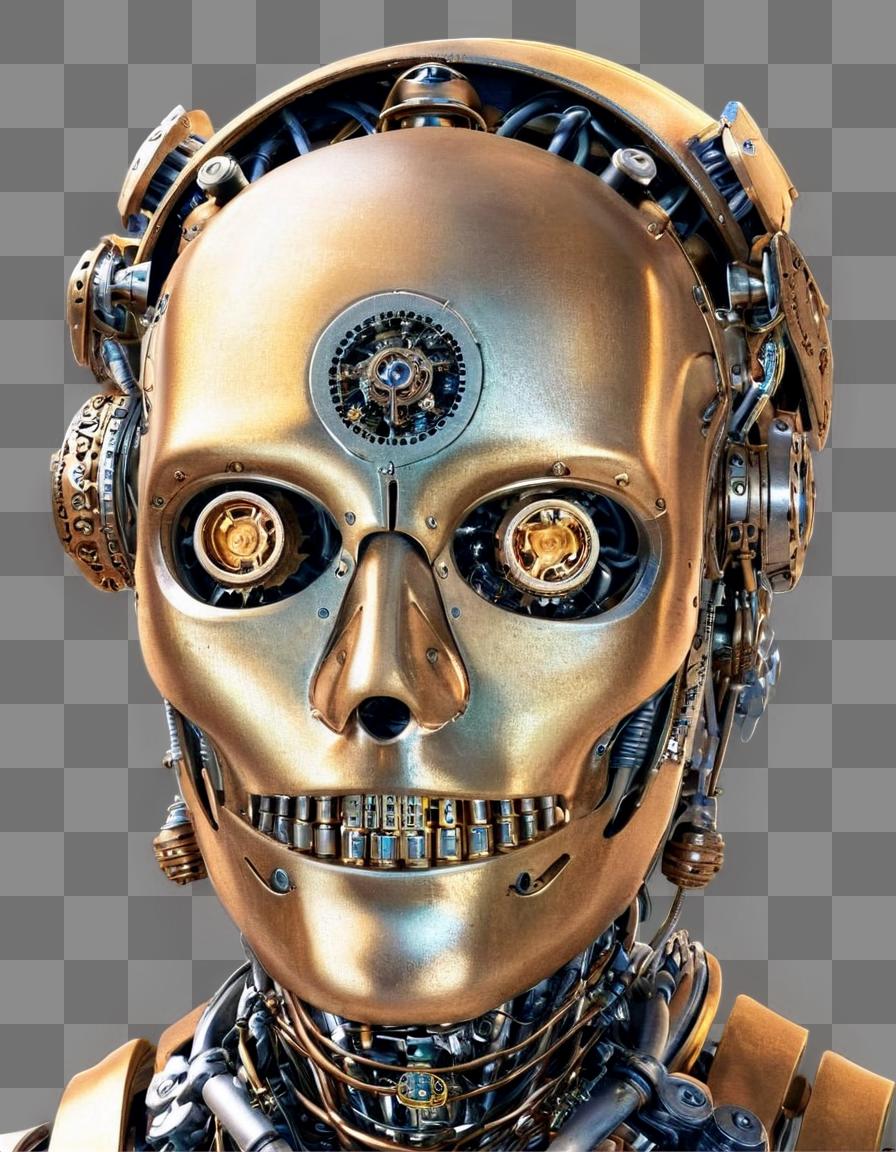}\hfill
\includegraphics[width=0.245\linewidth]{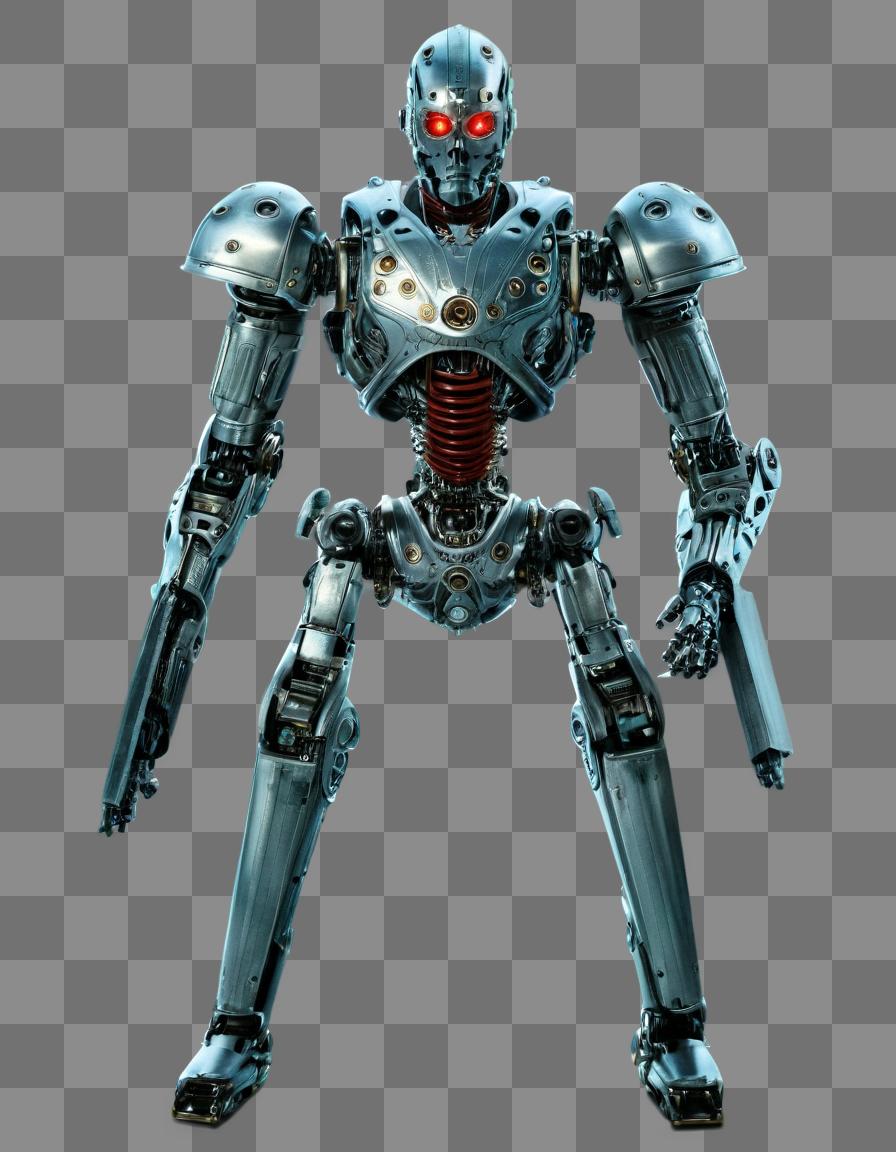}\hfill
\includegraphics[width=0.245\linewidth]{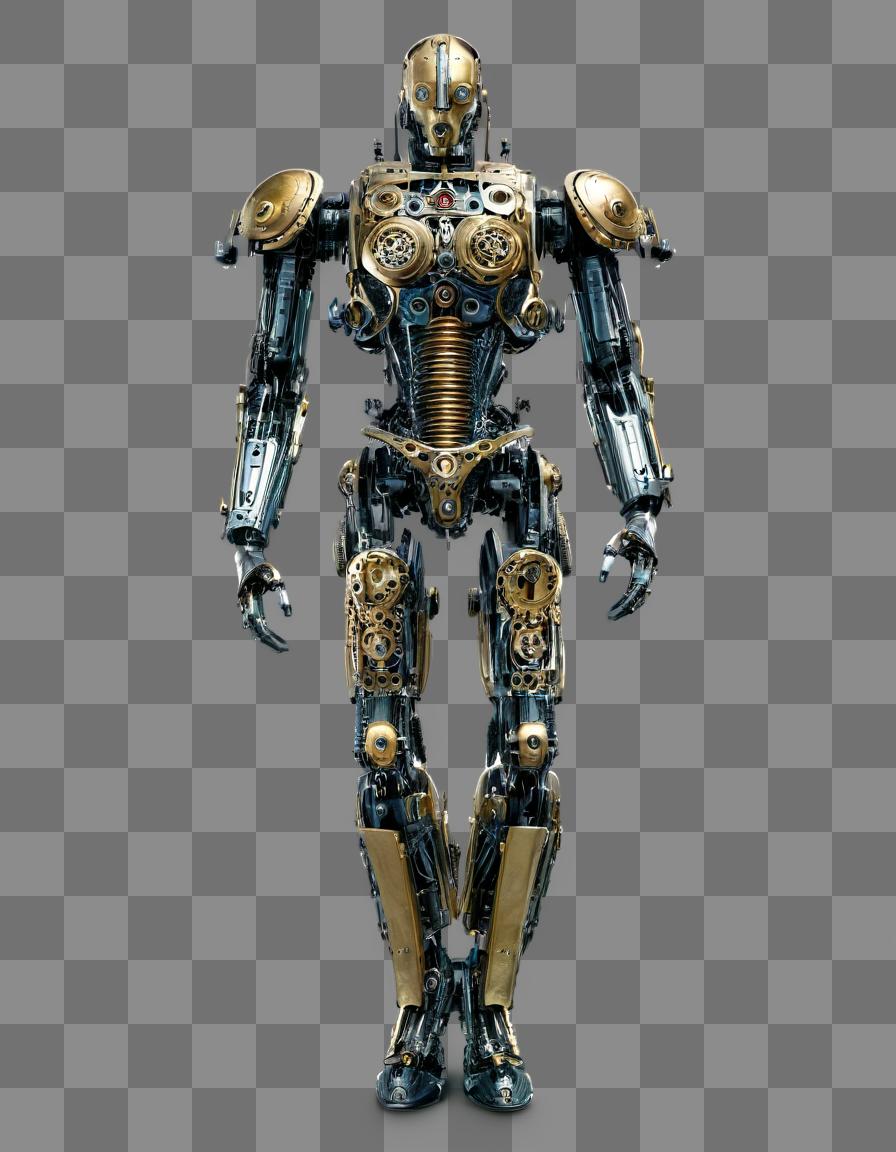}

\vspace{1pt}
\includegraphics[width=0.245\linewidth]{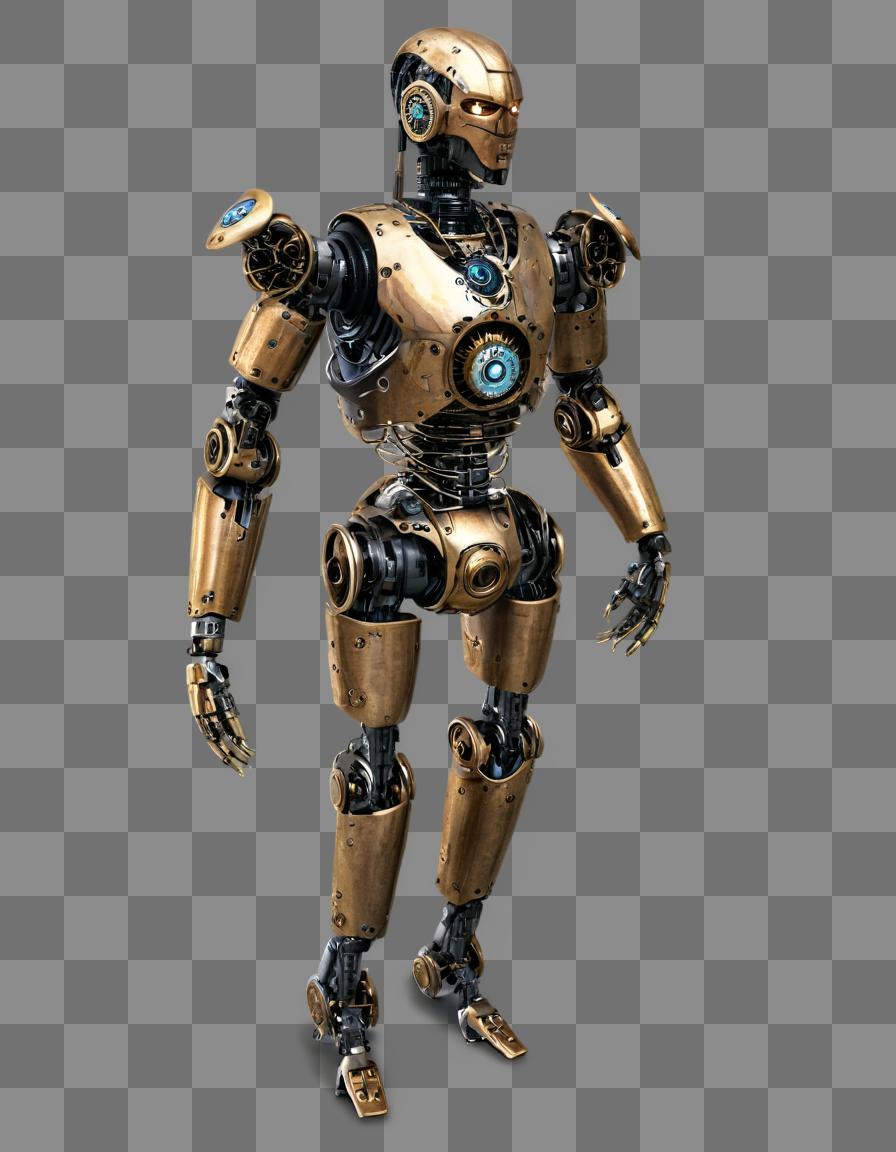}\hfill
\includegraphics[width=0.245\linewidth]{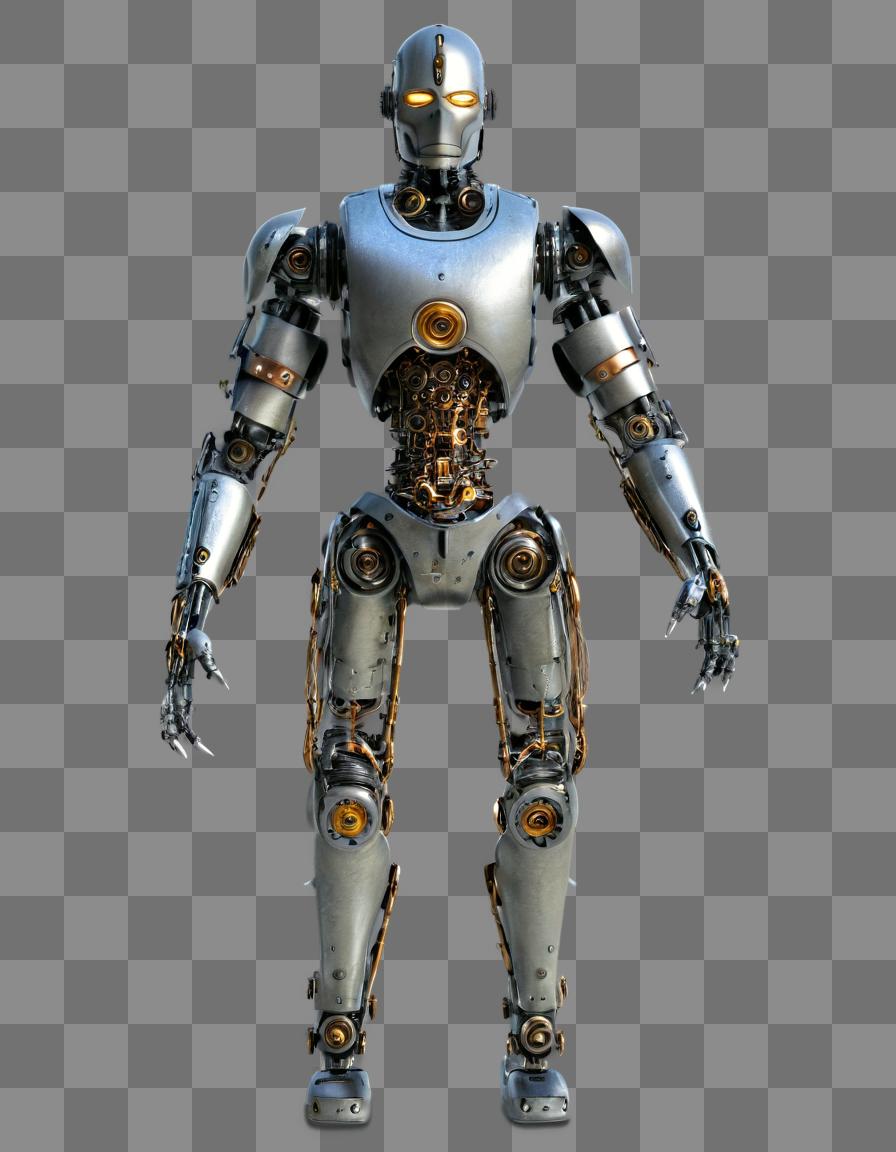}\hfill
\includegraphics[width=0.245\linewidth]{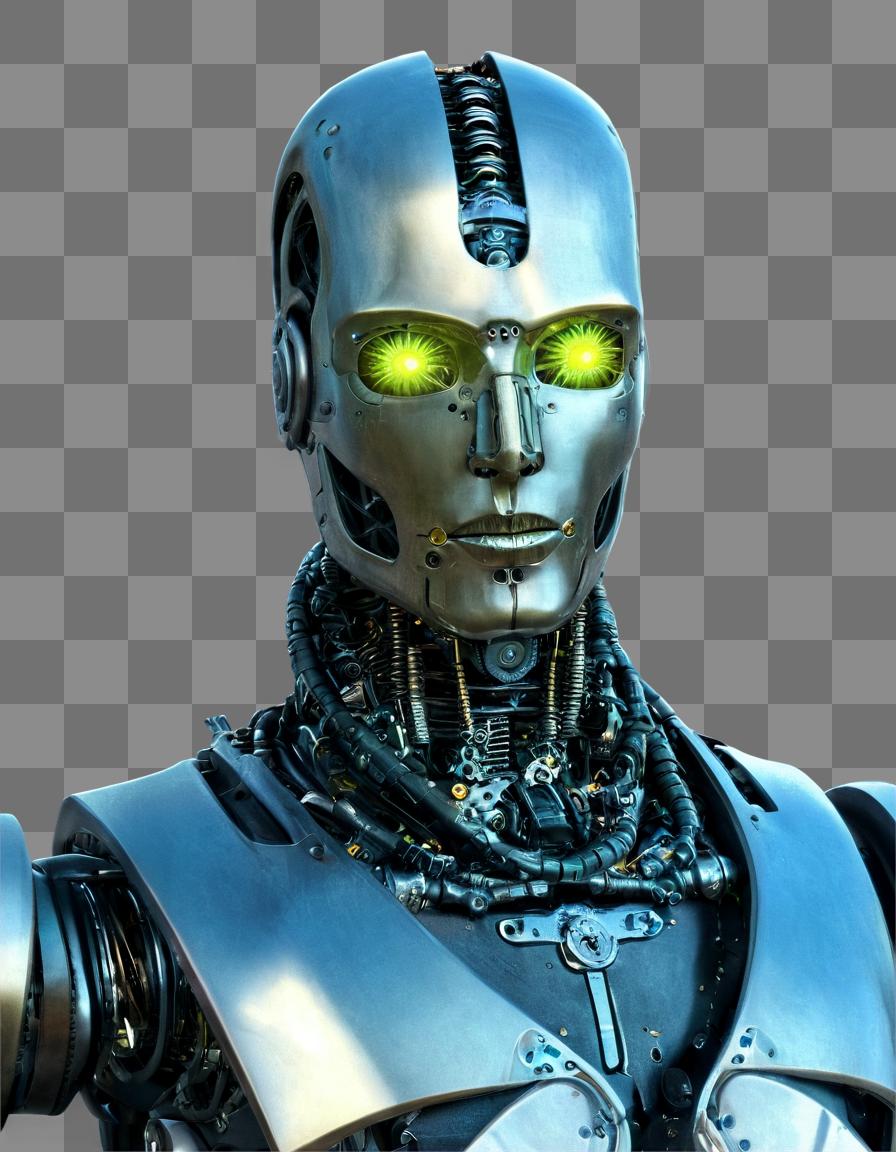}\hfill
\includegraphics[width=0.245\linewidth]{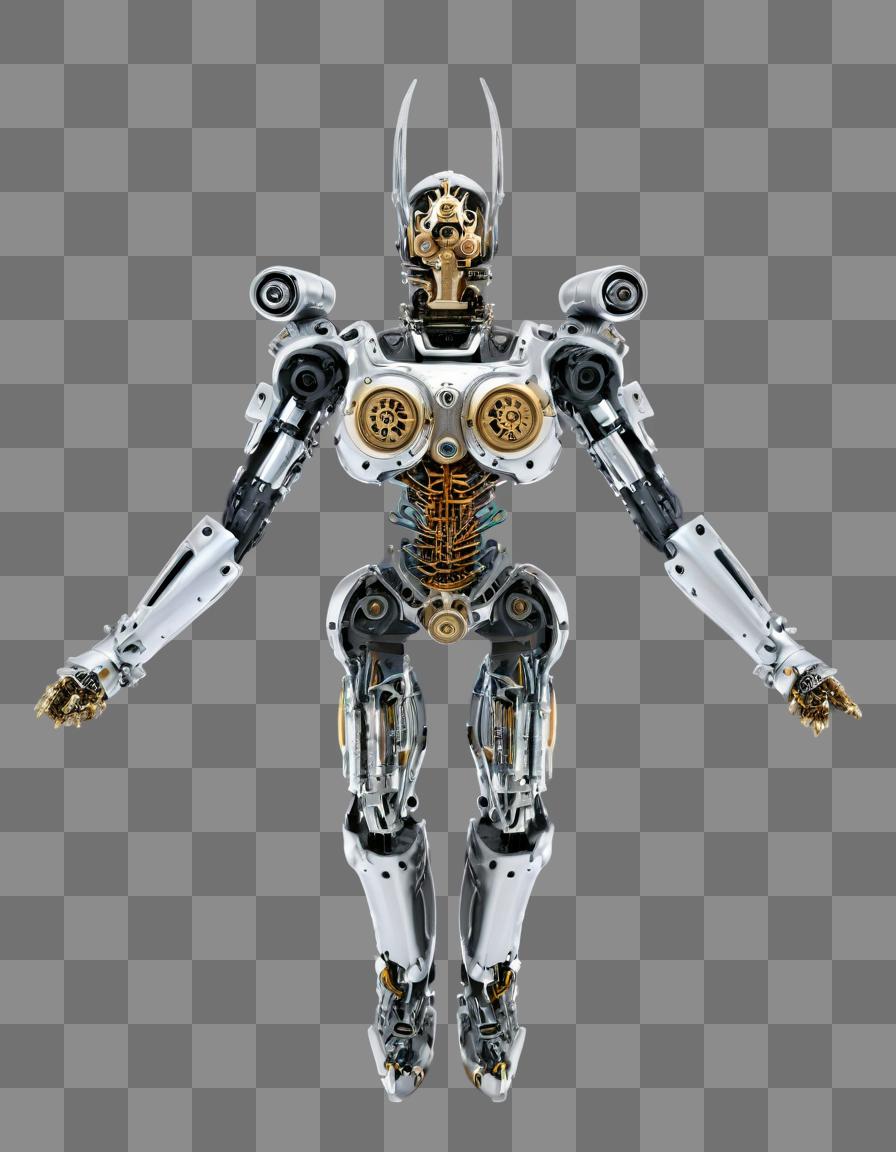}

\vspace{1pt}
\includegraphics[width=0.245\linewidth]{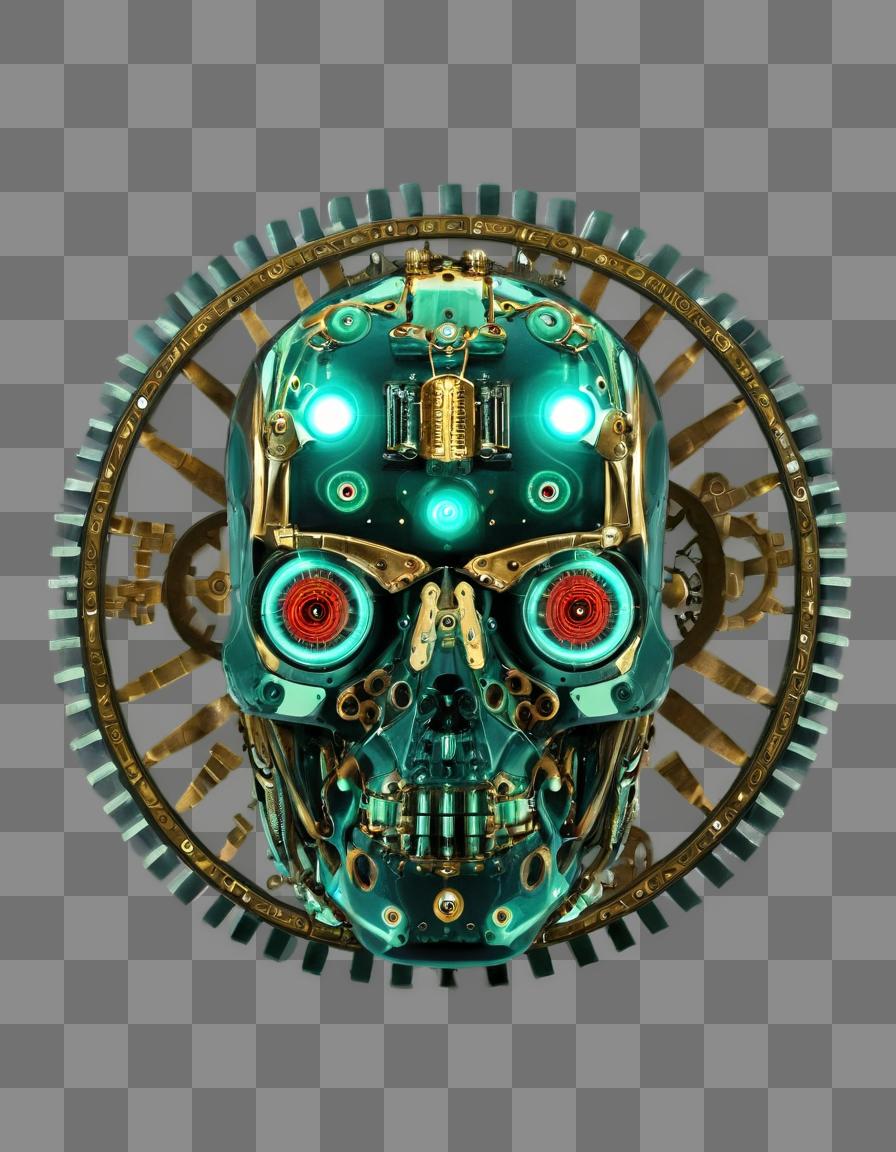}\hfill
\includegraphics[width=0.245\linewidth]{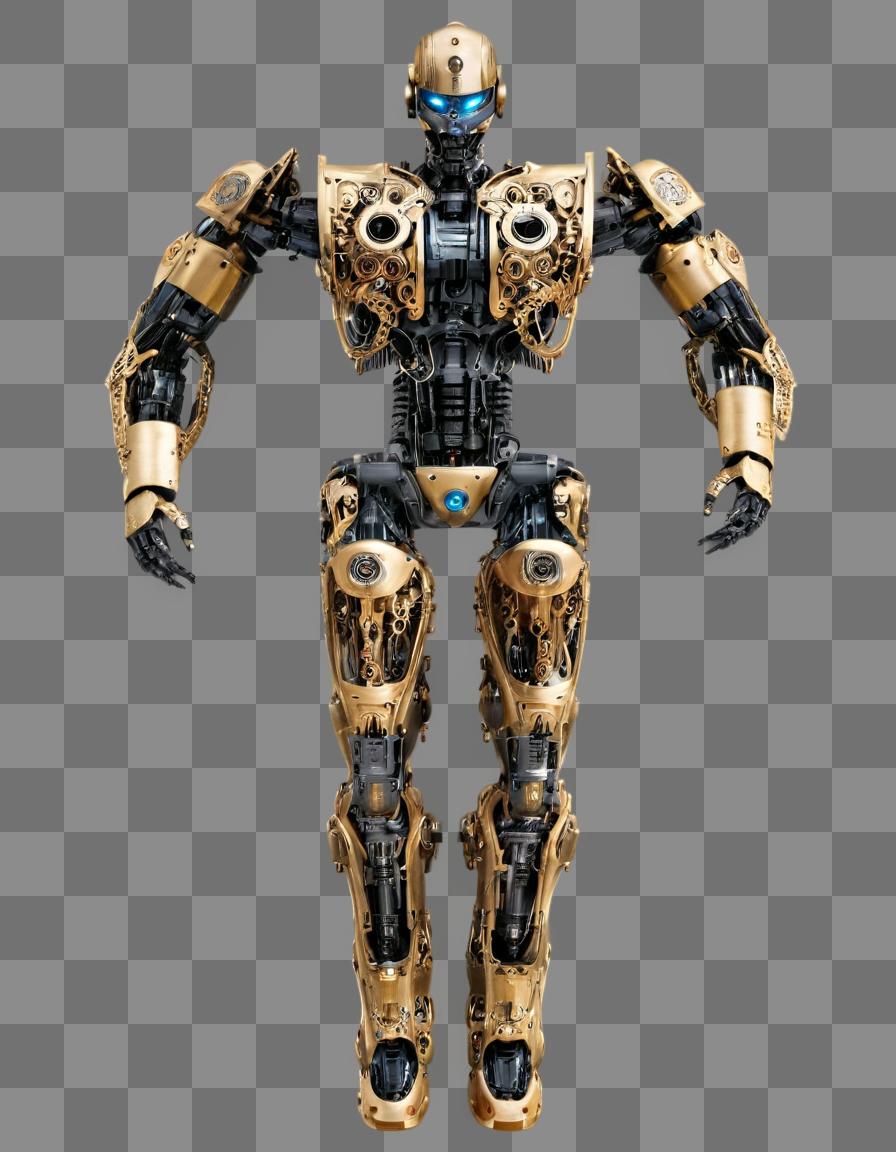}\hfill
\includegraphics[width=0.245\linewidth]{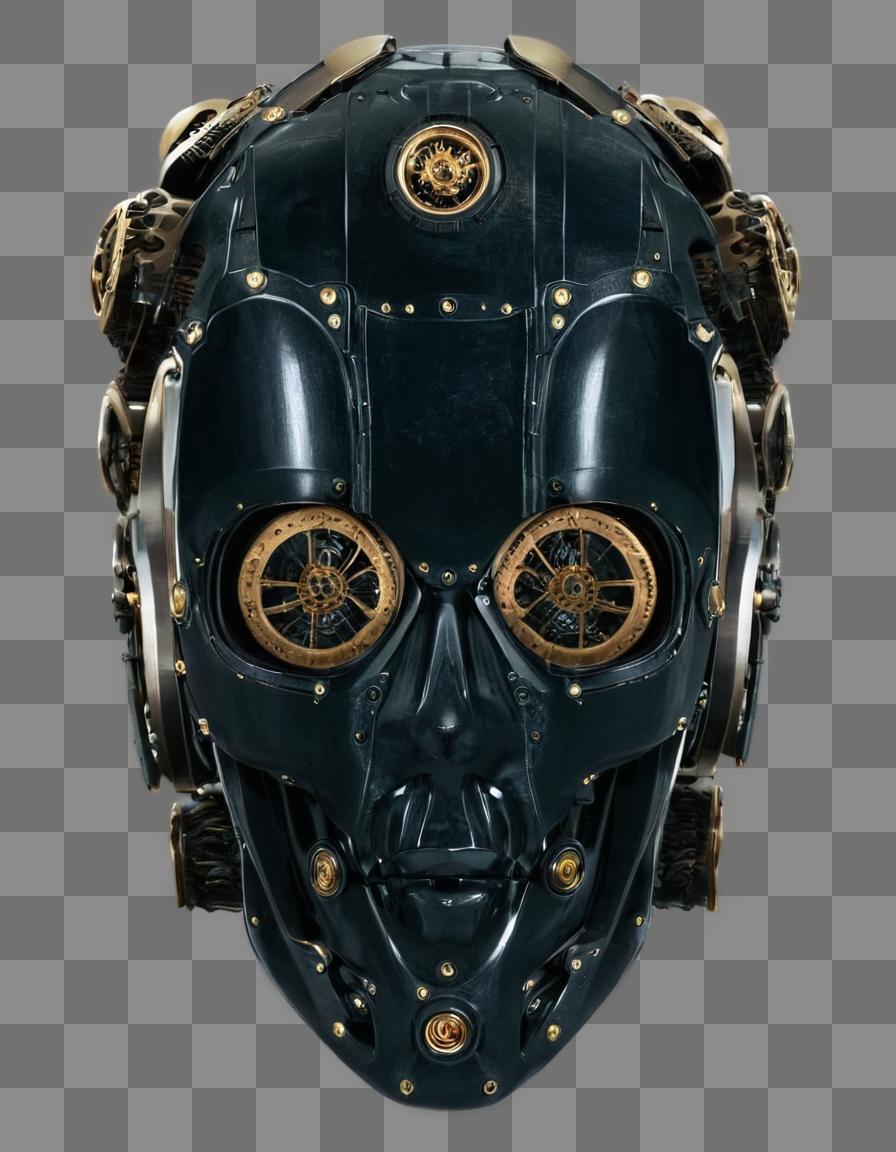}\hfill
\includegraphics[width=0.245\linewidth]{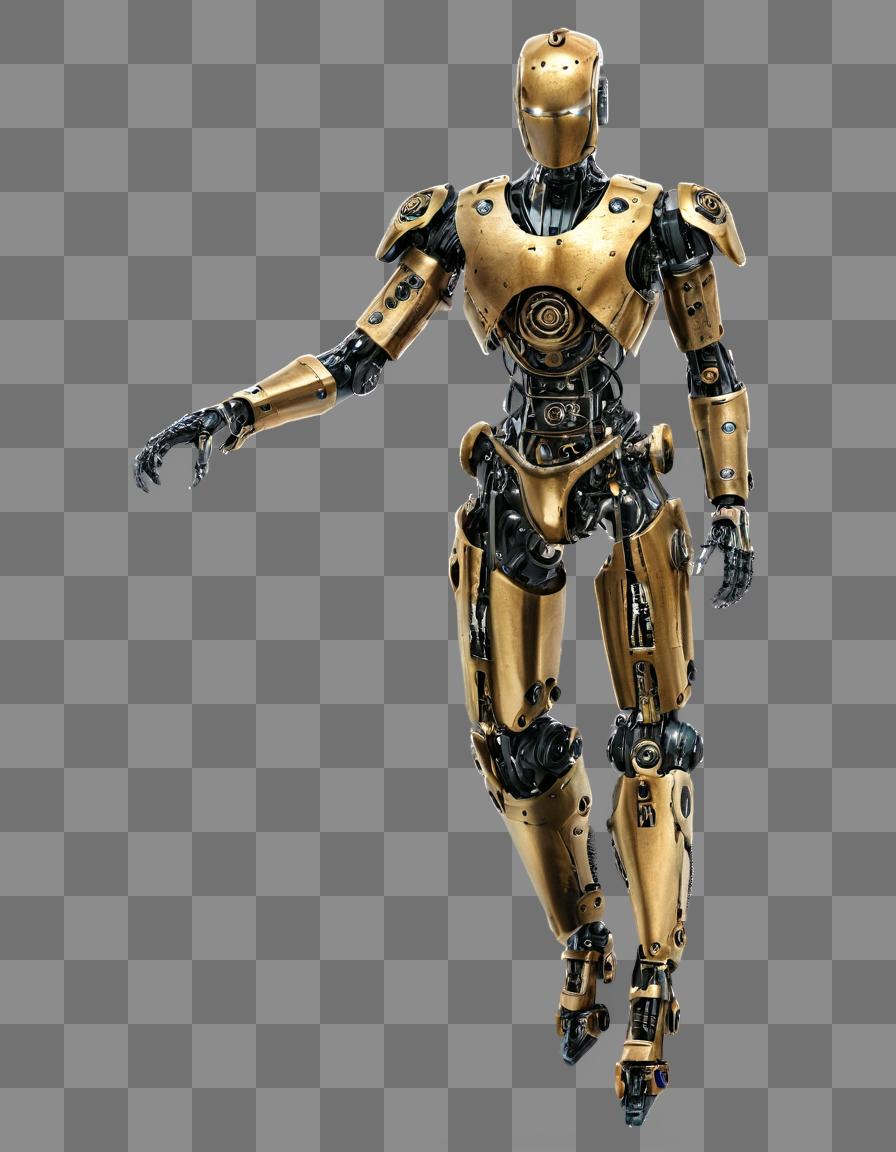}
\caption{Single Transparent Image Results \#15. The prompt is ``cyber steampunk robot''. Resolution is $896\times1152$.}
\label{fig:a15}
\end{minipage}
\end{figure*}

\begin{figure*}

\begin{minipage}{\linewidth}
\includegraphics[width=0.245\linewidth]{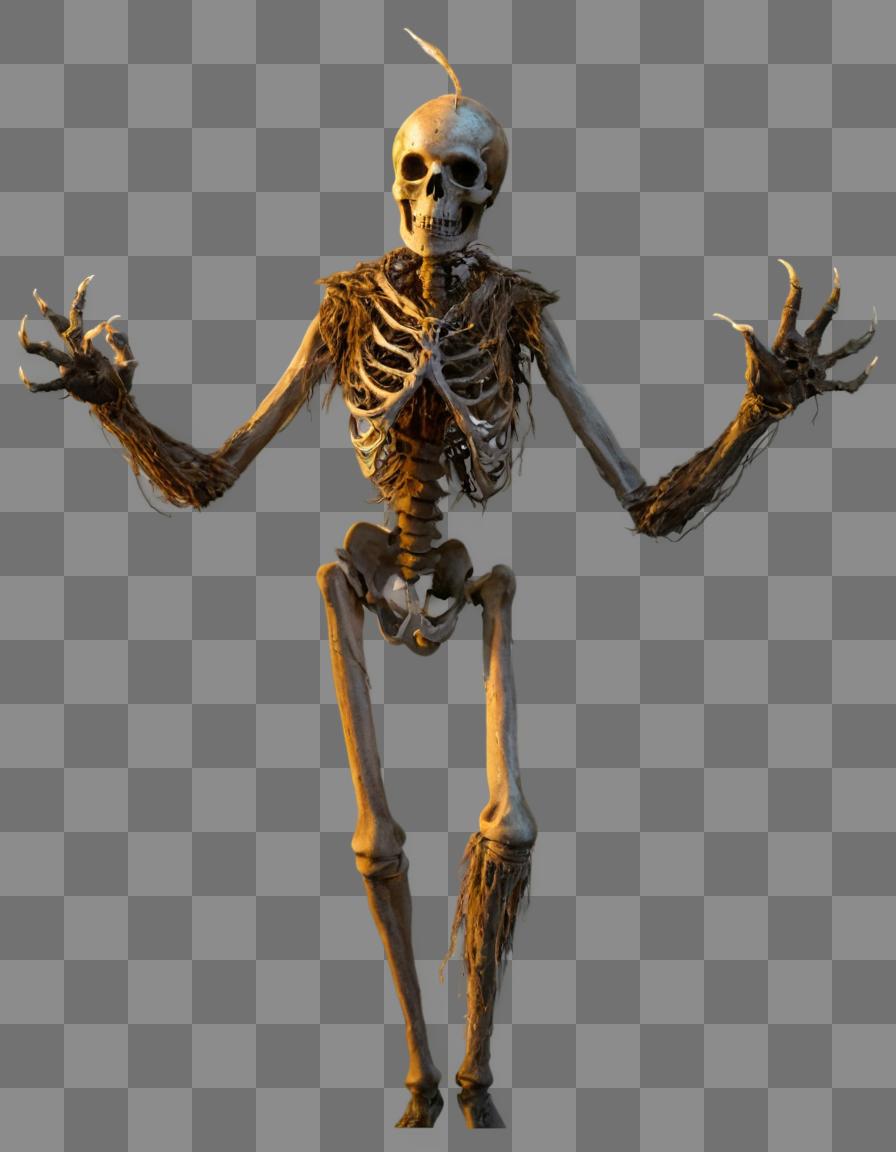}\hfill
\includegraphics[width=0.245\linewidth]{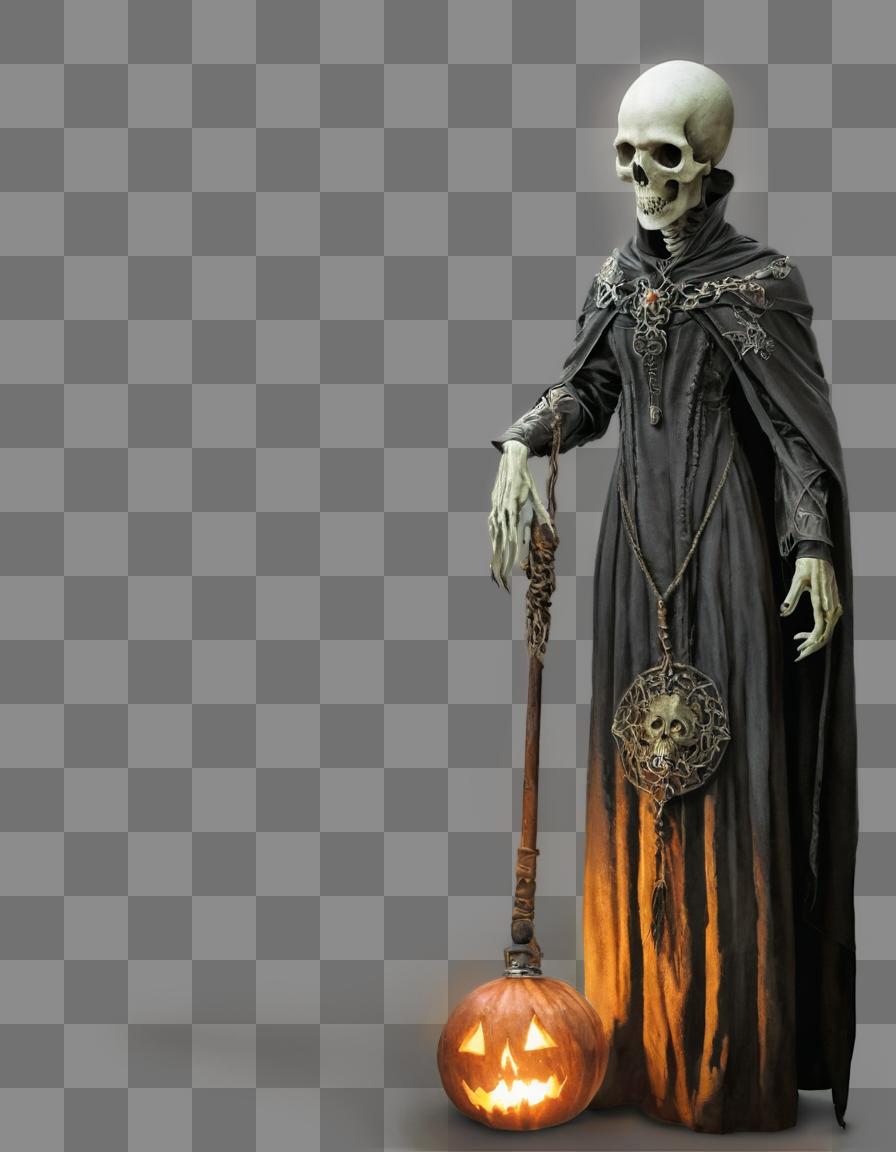}\hfill
\includegraphics[width=0.245\linewidth]{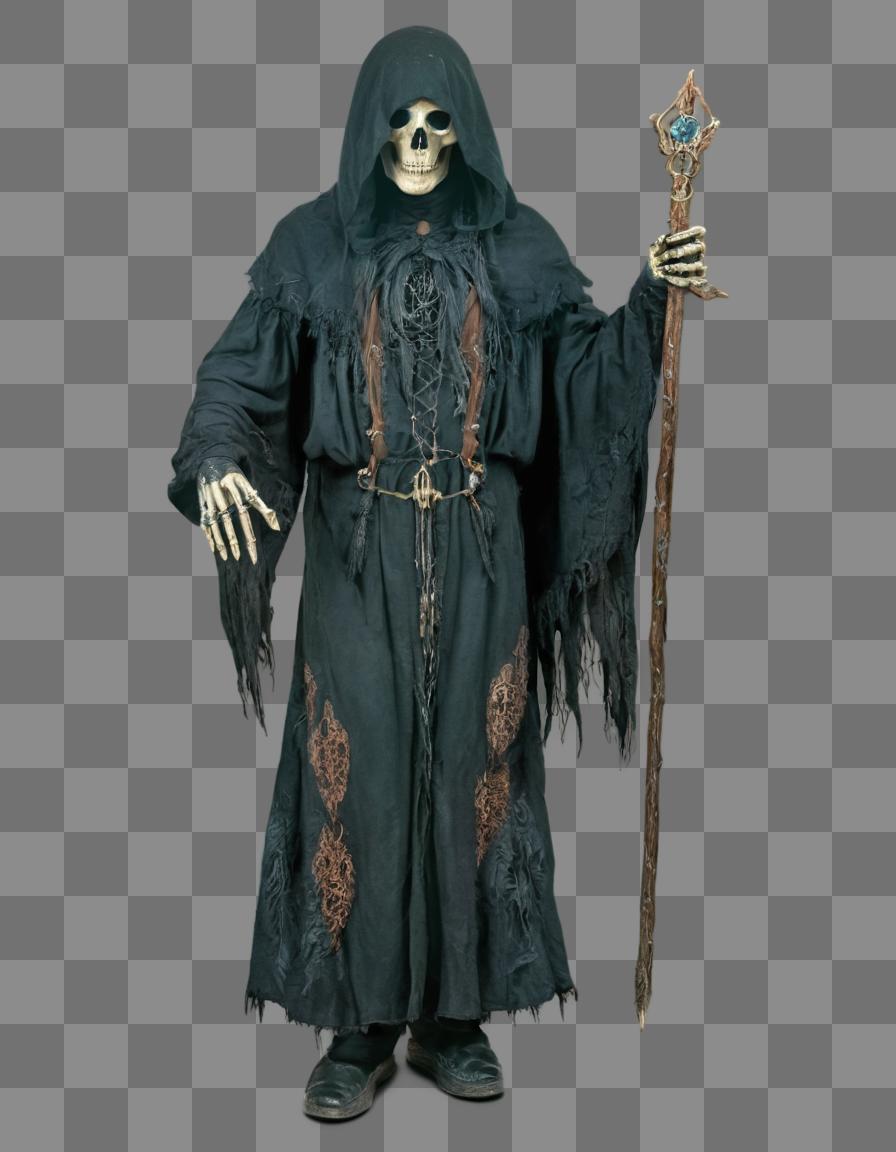}\hfill
\includegraphics[width=0.245\linewidth]{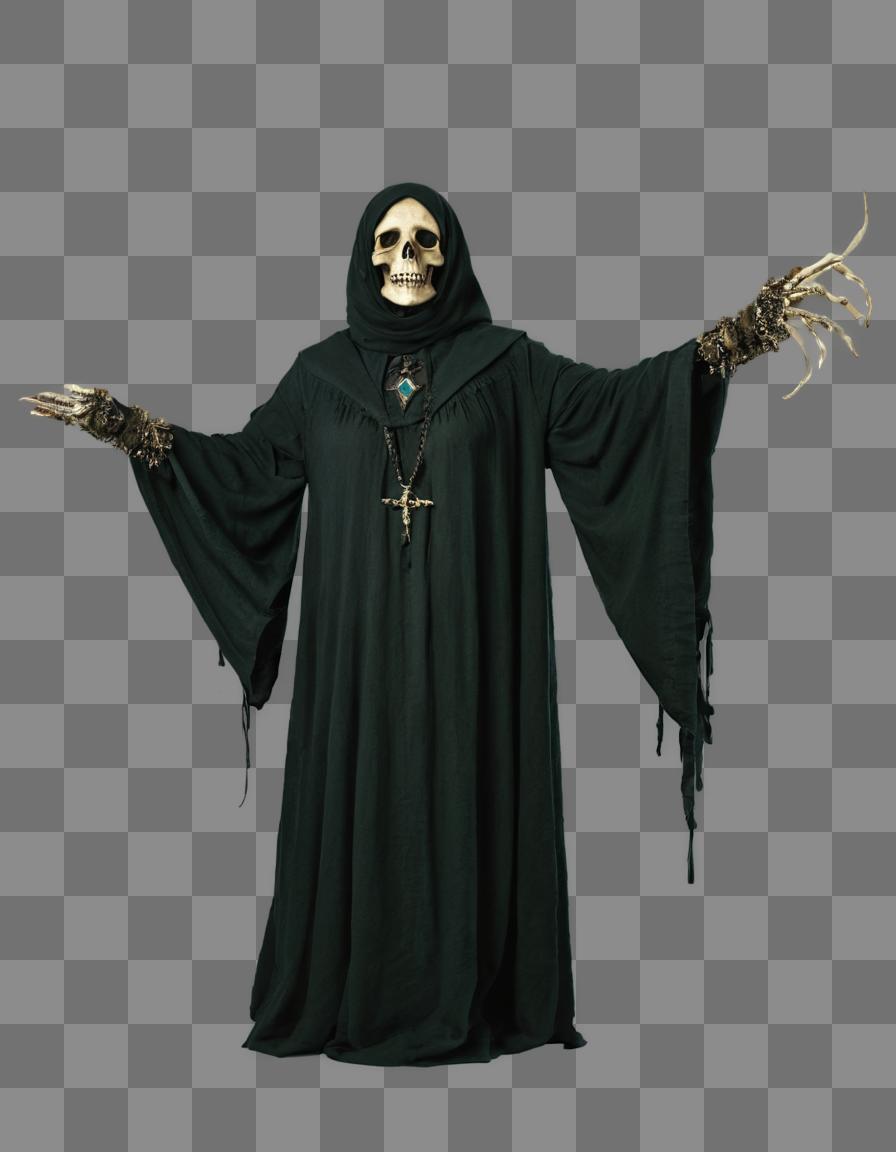}

\vspace{1pt}
\includegraphics[width=0.245\linewidth]{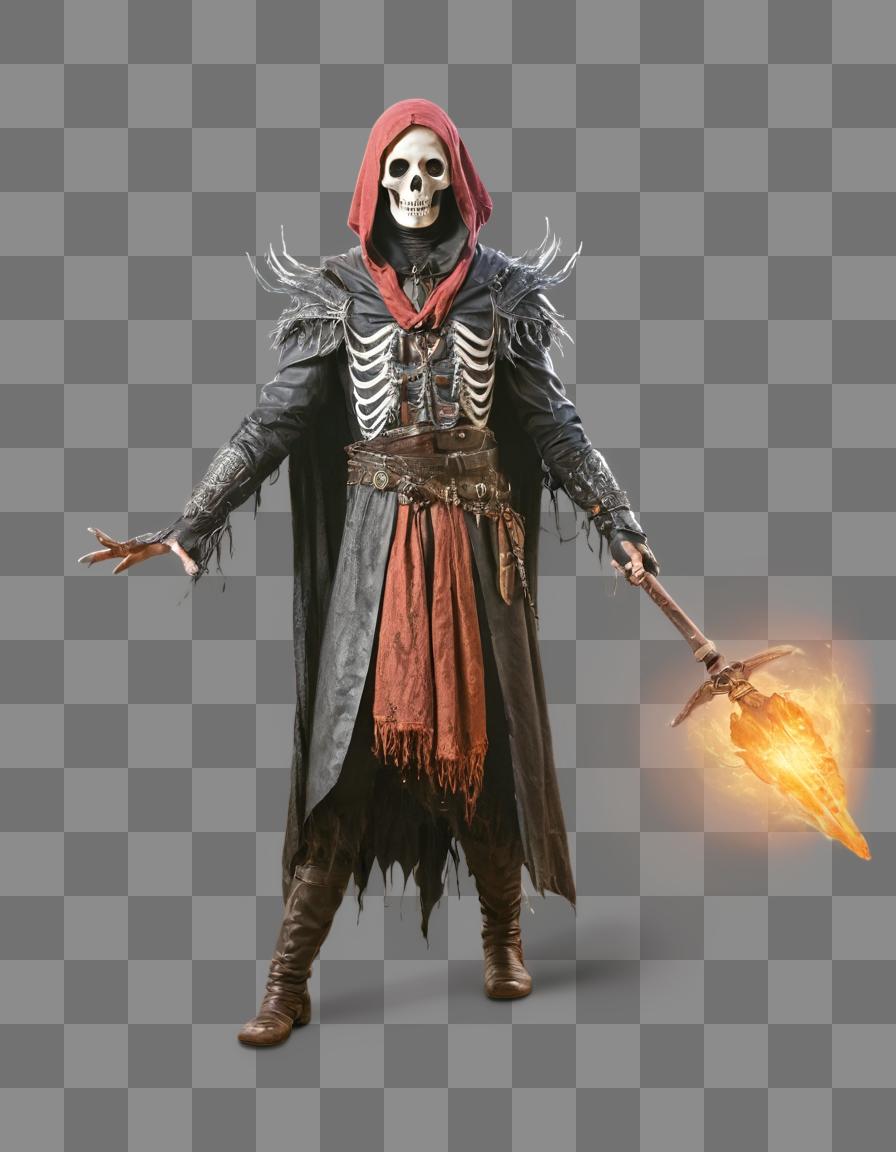}\hfill
\includegraphics[width=0.245\linewidth]{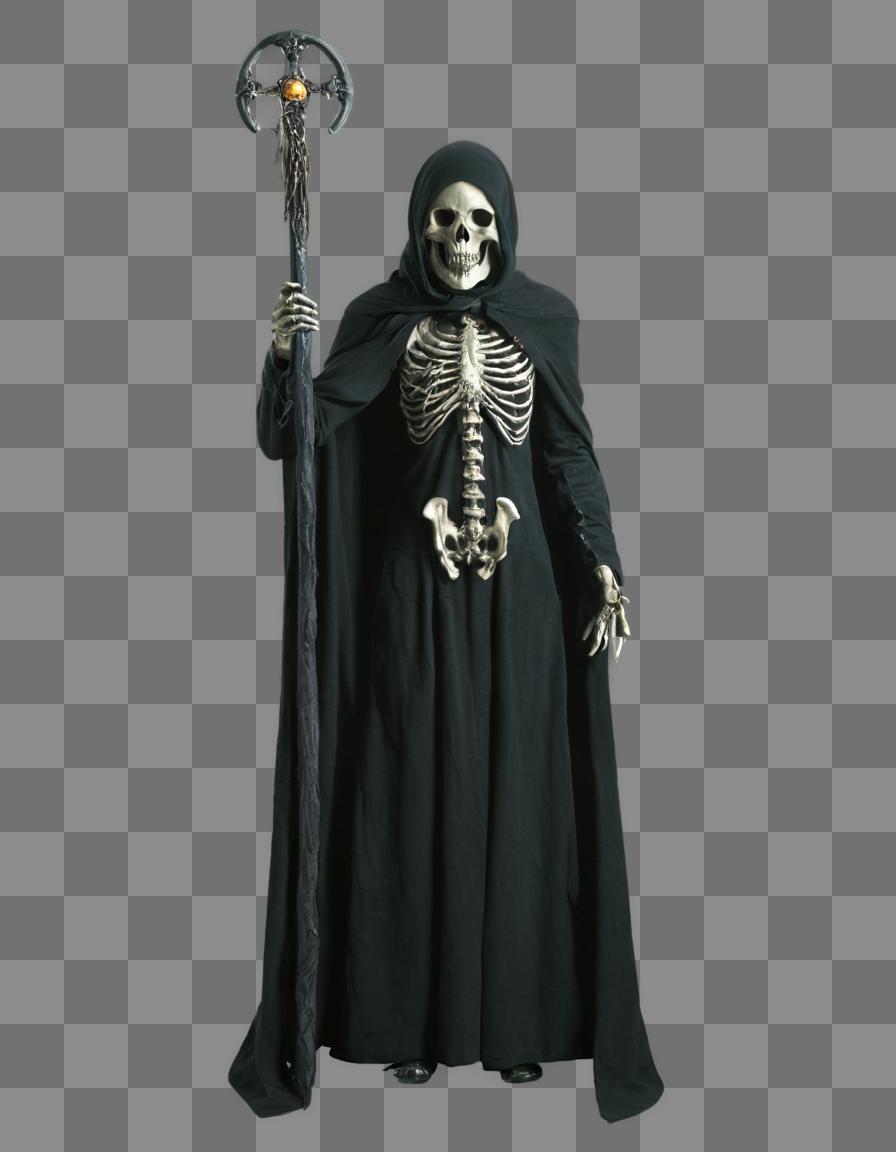}\hfill
\includegraphics[width=0.245\linewidth]{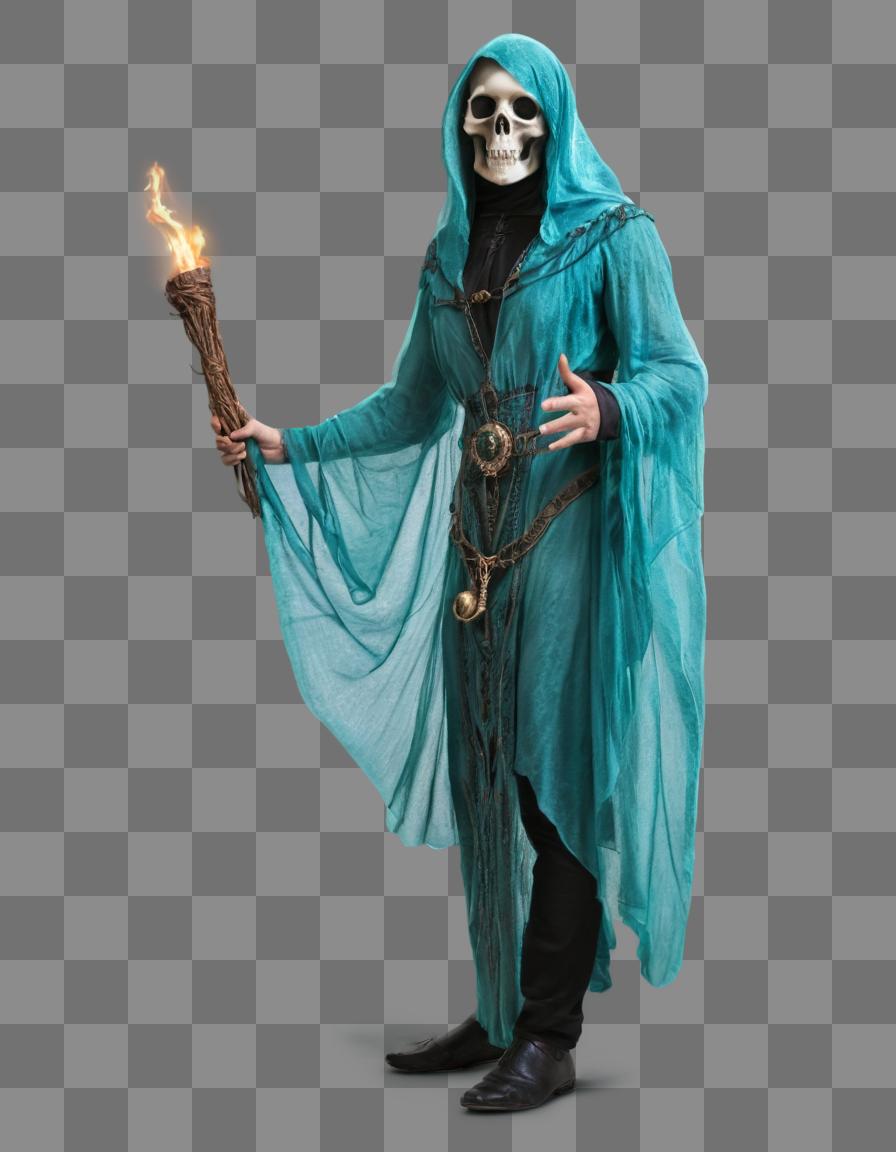}\hfill
\includegraphics[width=0.245\linewidth]{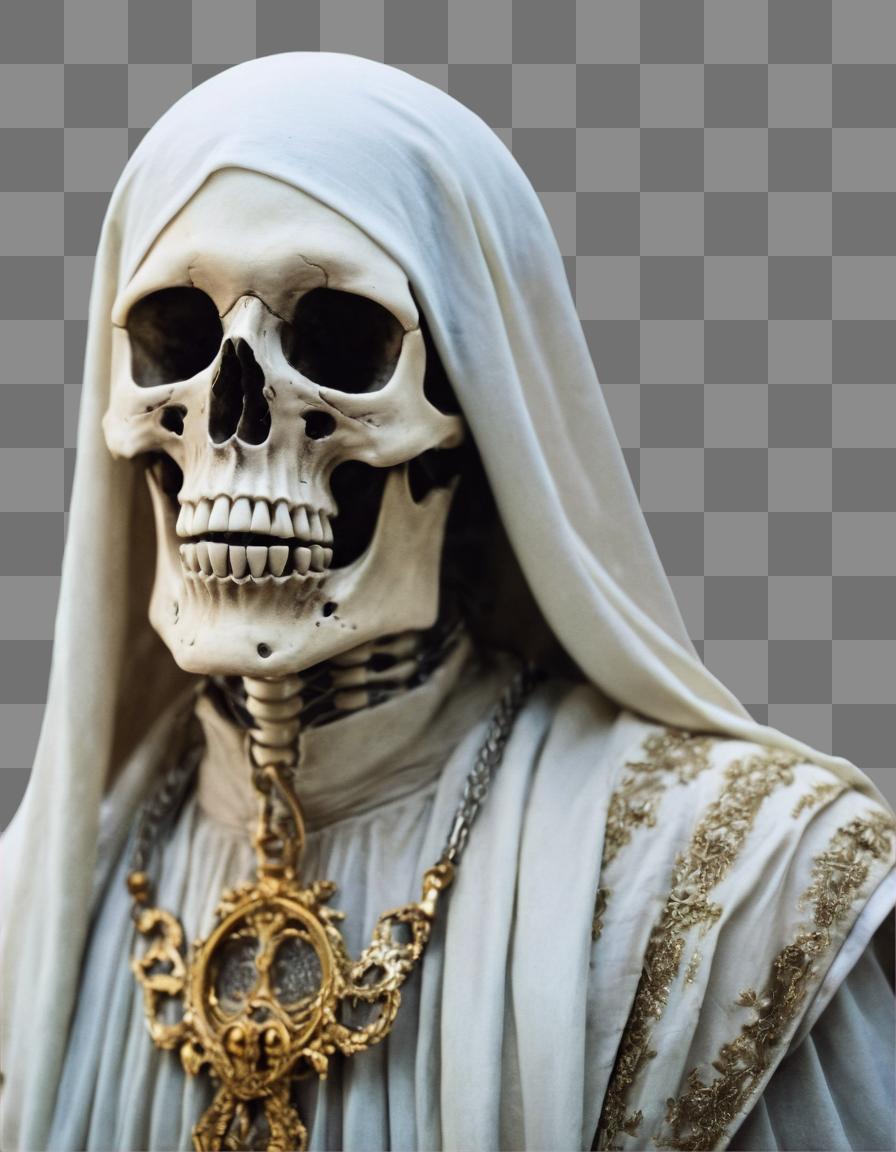}

\vspace{1pt}
\includegraphics[width=0.245\linewidth]{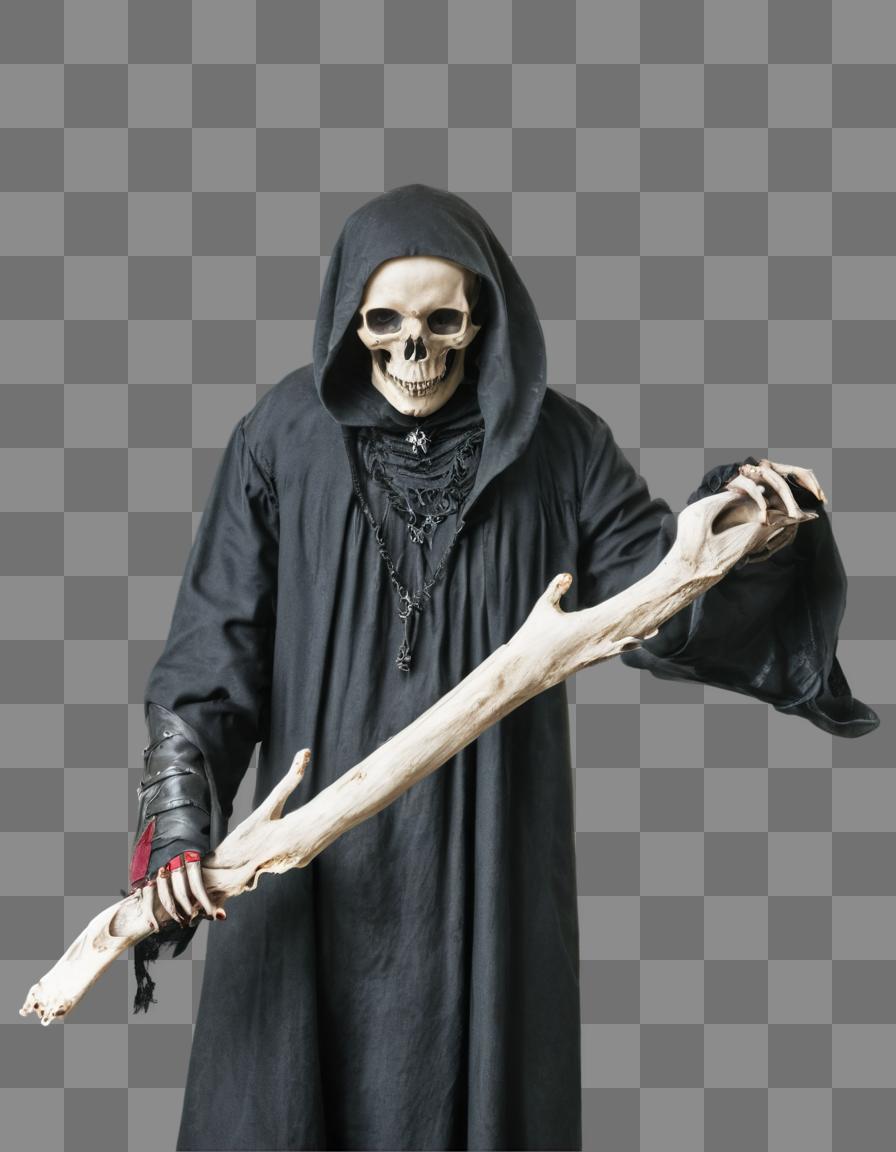}\hfill
\includegraphics[width=0.245\linewidth]{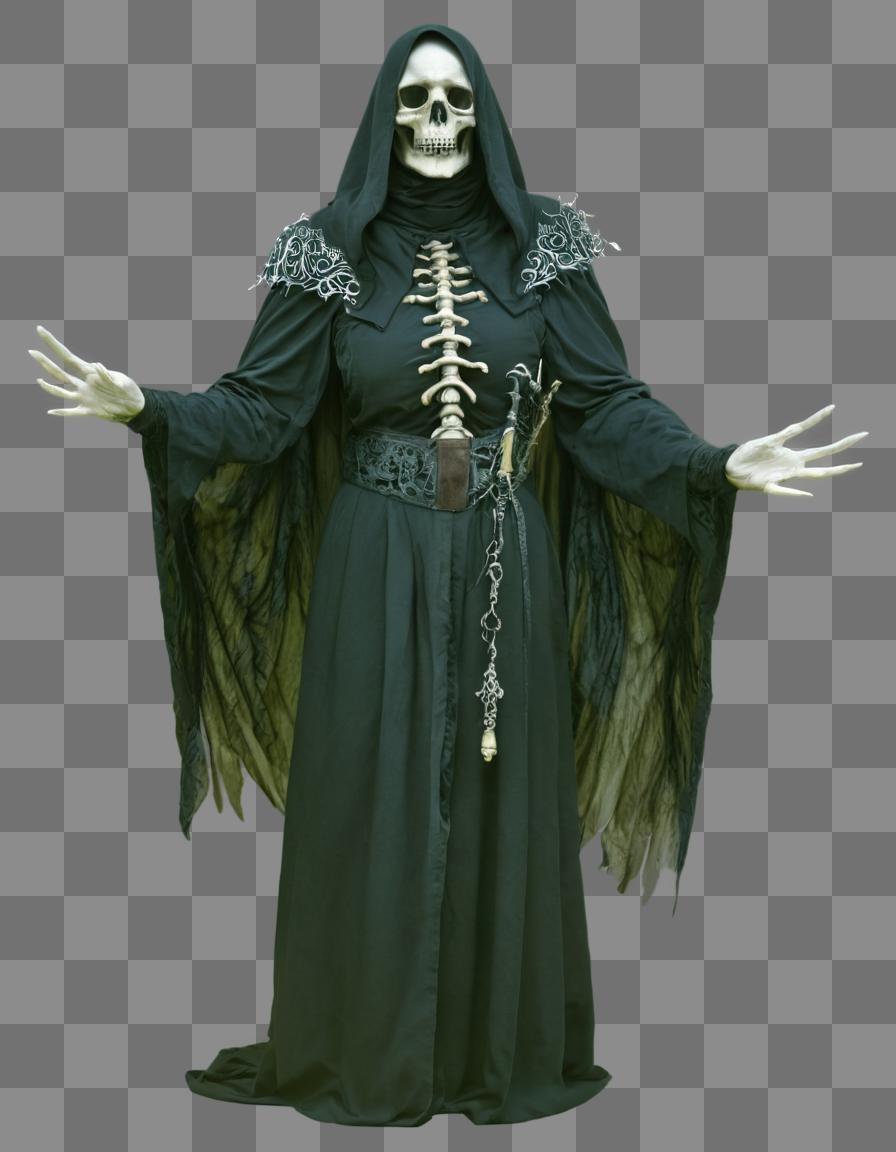}\hfill
\includegraphics[width=0.245\linewidth]{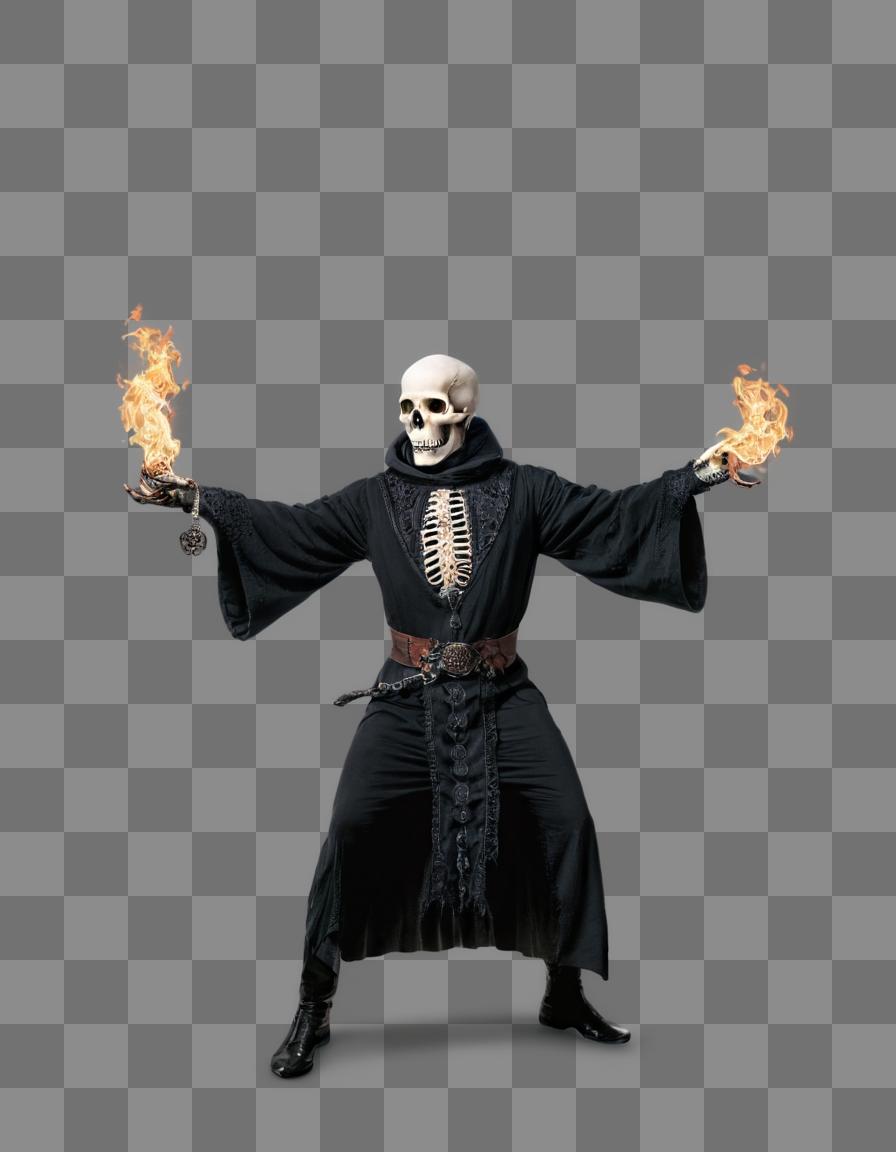}\hfill
\includegraphics[width=0.245\linewidth]{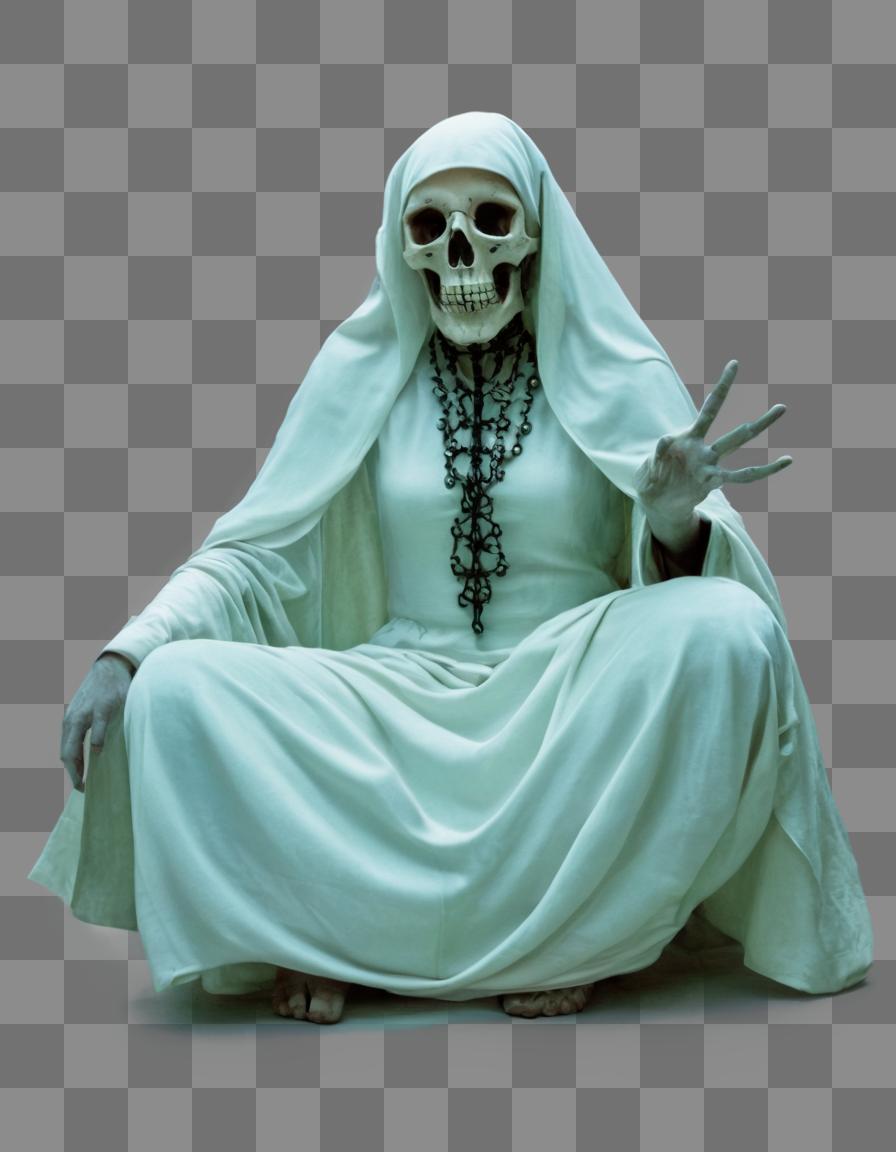}
\caption{Single Transparent Image Results \#16. The prompt is ``necromancer''. Resolution is $896\times1152$.}
\label{fig:a16}
\end{minipage}
\end{figure*}

We present additional results for single transparent images, from Figure~\ref{fig:a1} to Figure~\ref{fig:a16}.

\section{Multiple Transparent Layers}
	
\begin{figure*}

\begin{minipage}{\linewidth}
\includegraphics[width=0.245\linewidth]{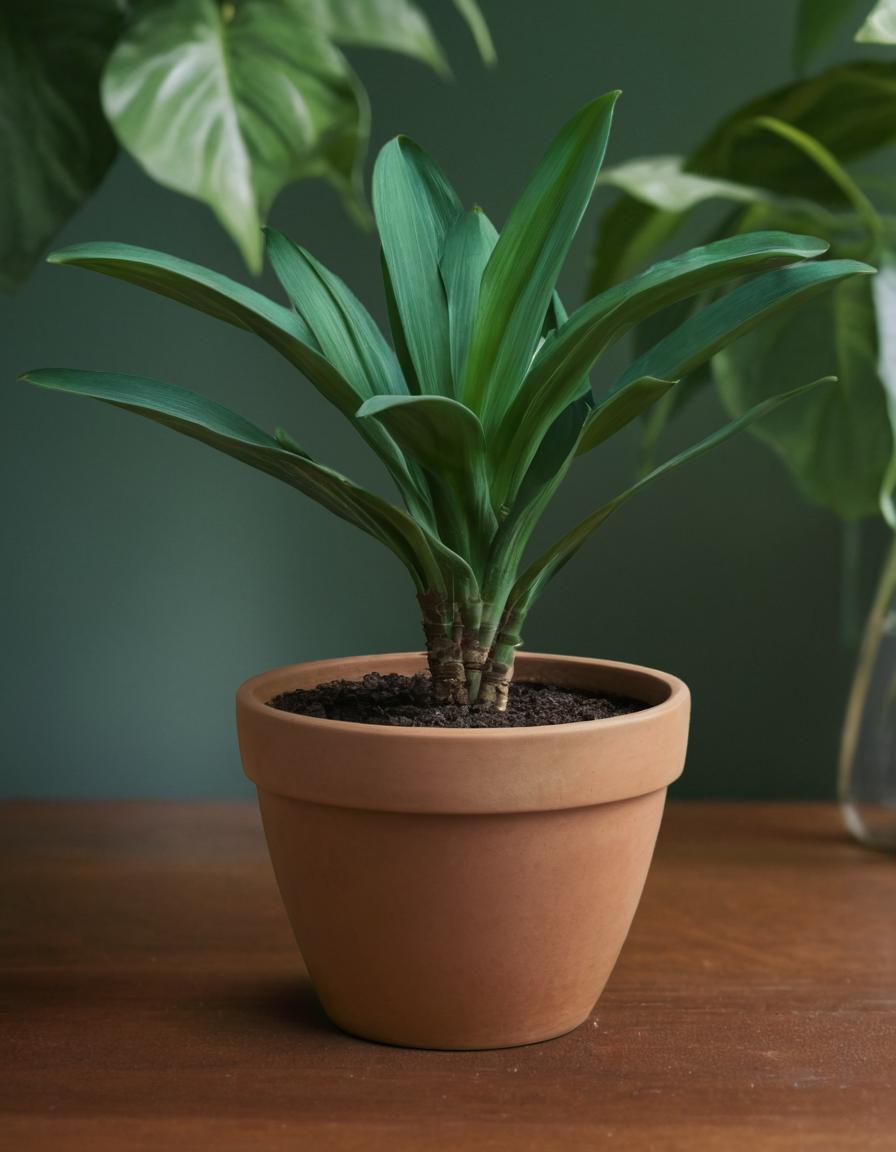}\hfill
\includegraphics[width=0.245\linewidth]{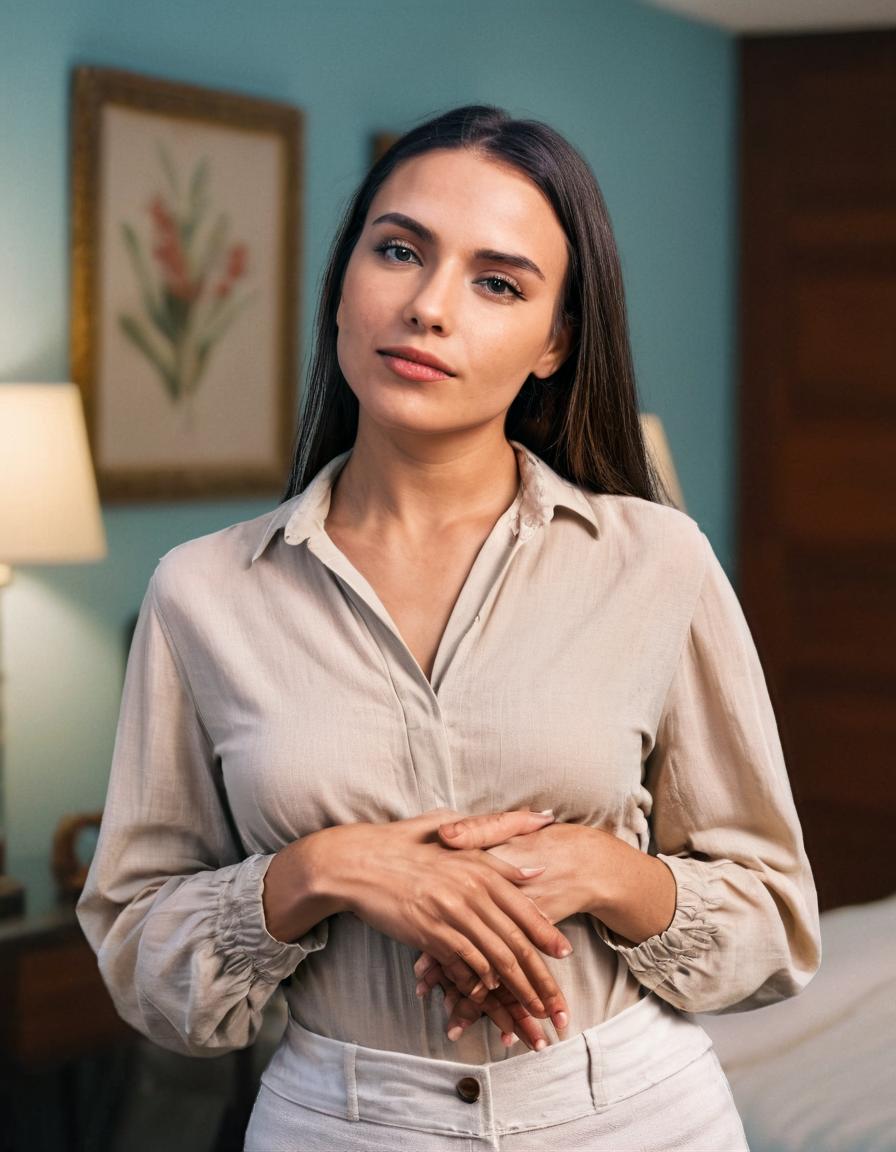}\hfill
\includegraphics[width=0.245\linewidth]{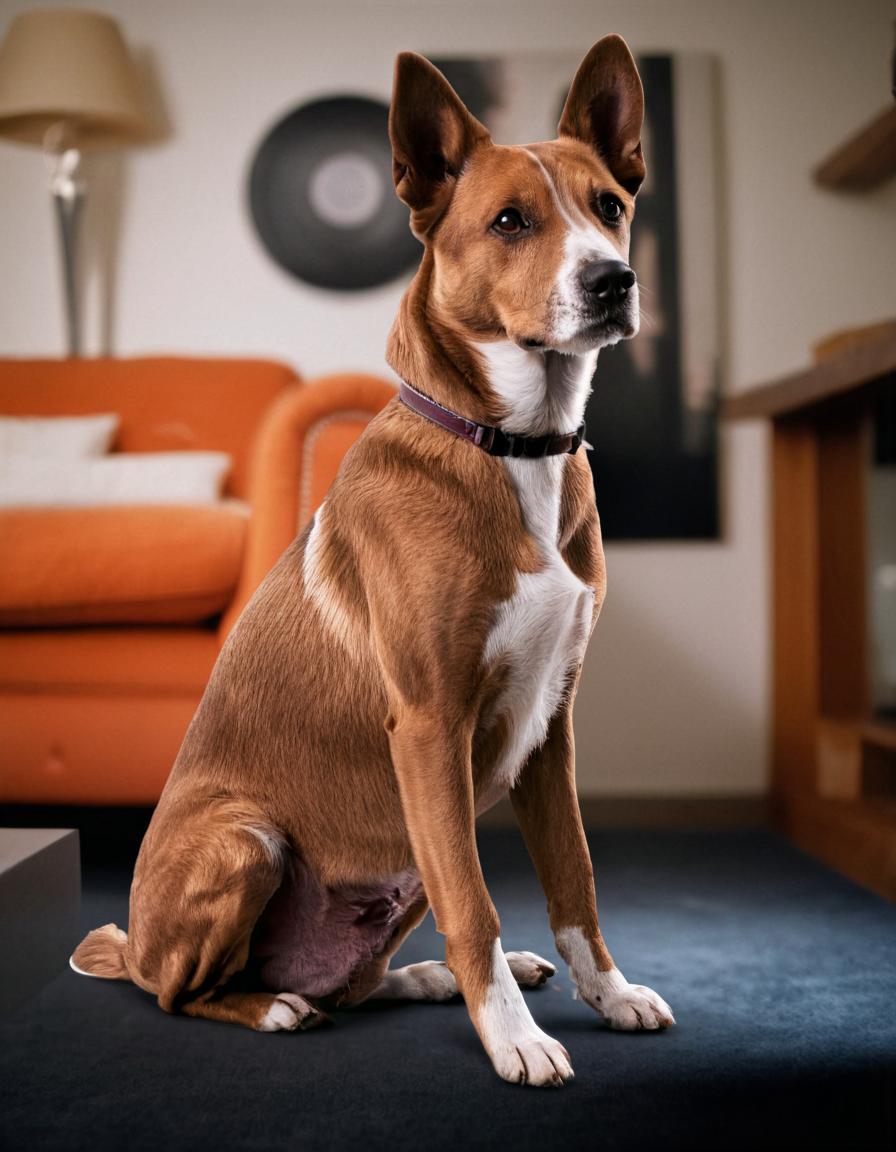}\hfill
\includegraphics[width=0.245\linewidth]{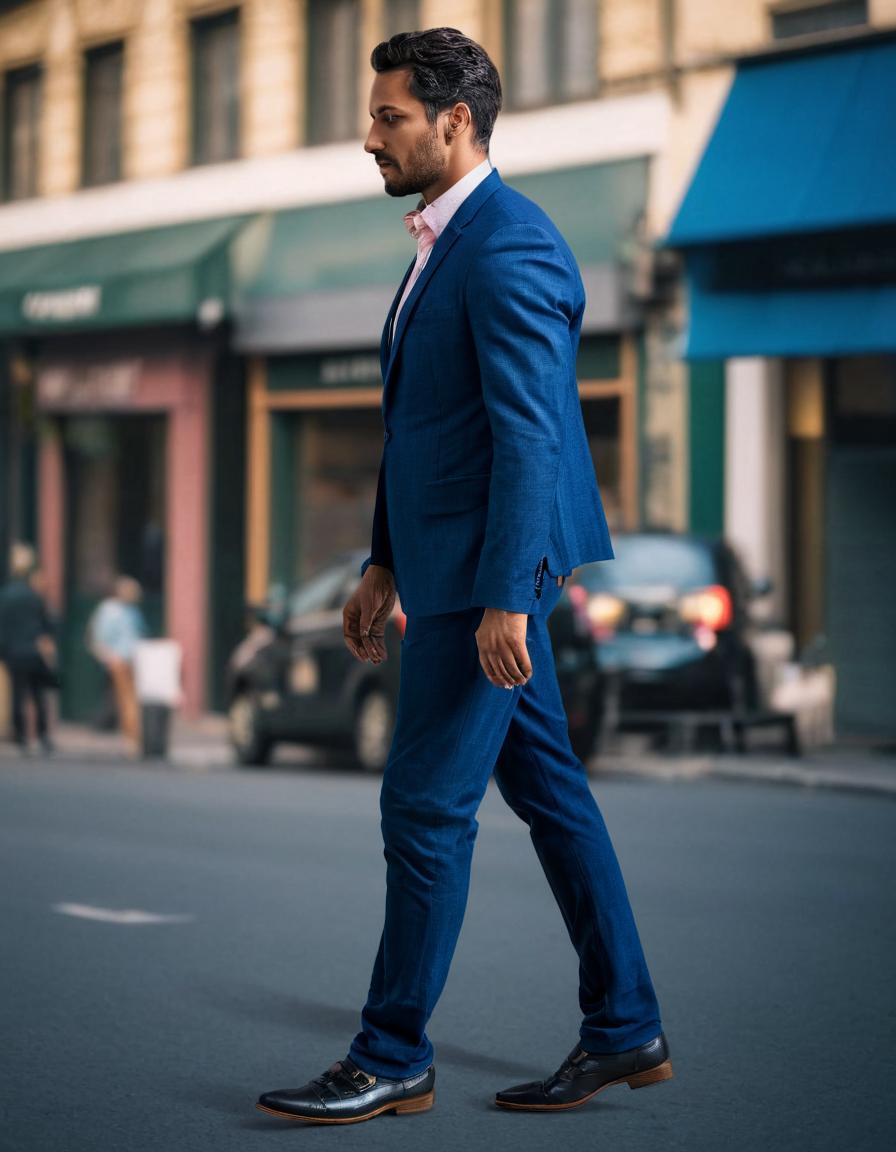}

\vspace{1pt}
\includegraphics[width=0.245\linewidth]{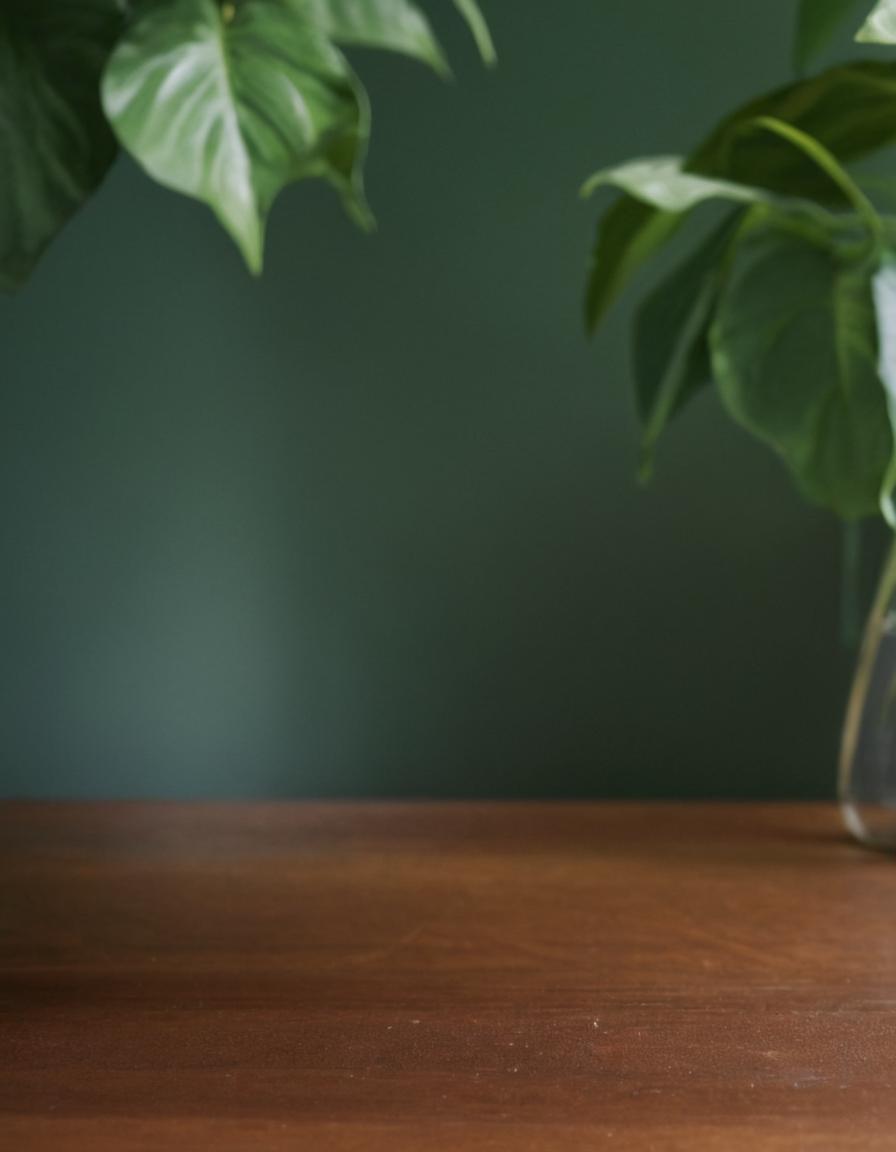}\hfill
\includegraphics[width=0.245\linewidth]{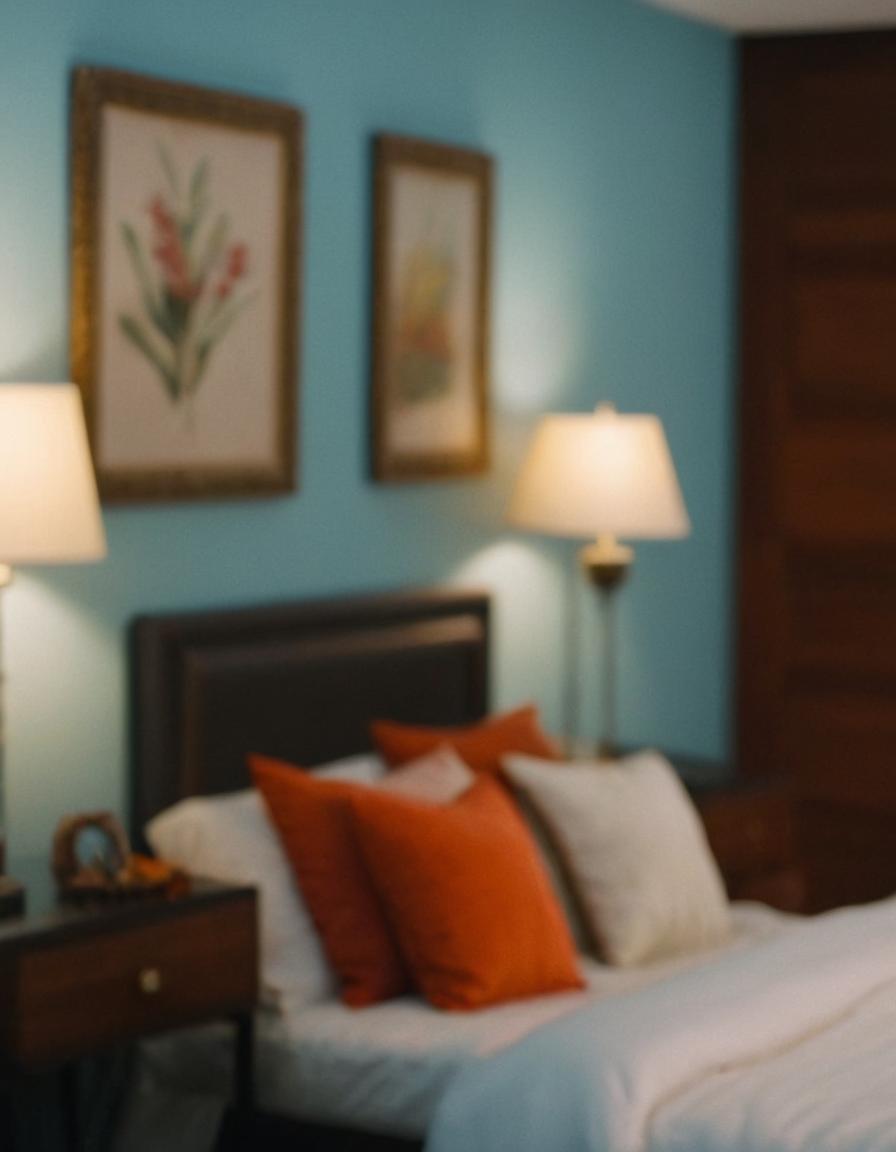}\hfill
\includegraphics[width=0.245\linewidth]{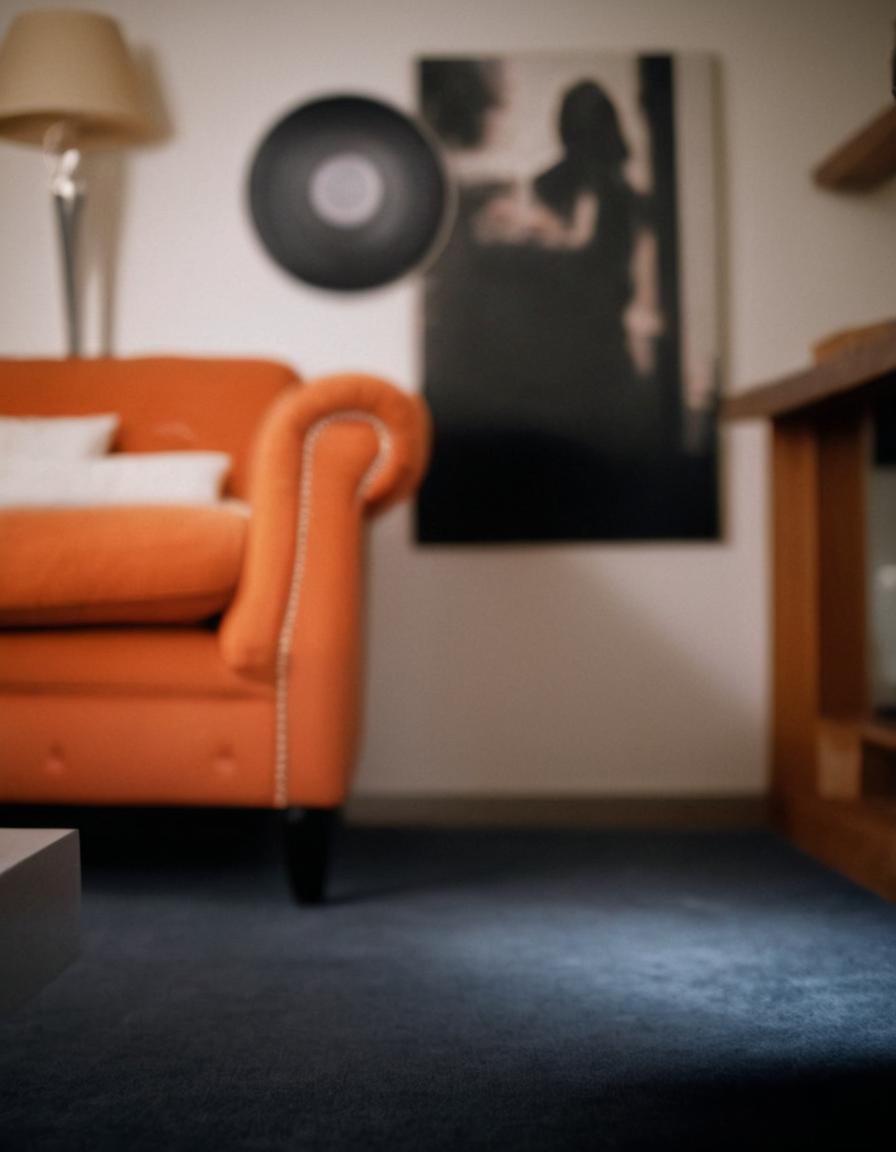}\hfill
\includegraphics[width=0.245\linewidth]{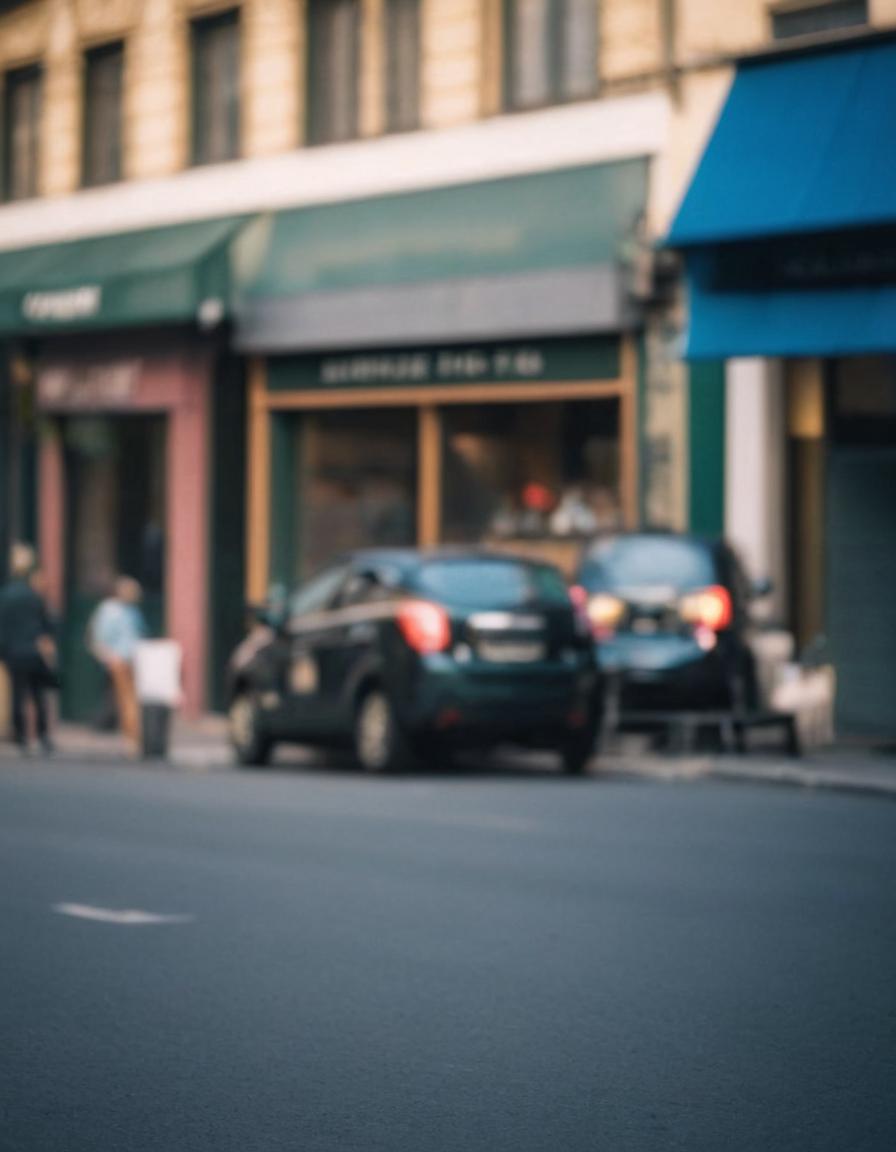}

\vspace{1pt}
\includegraphics[width=0.245\linewidth]{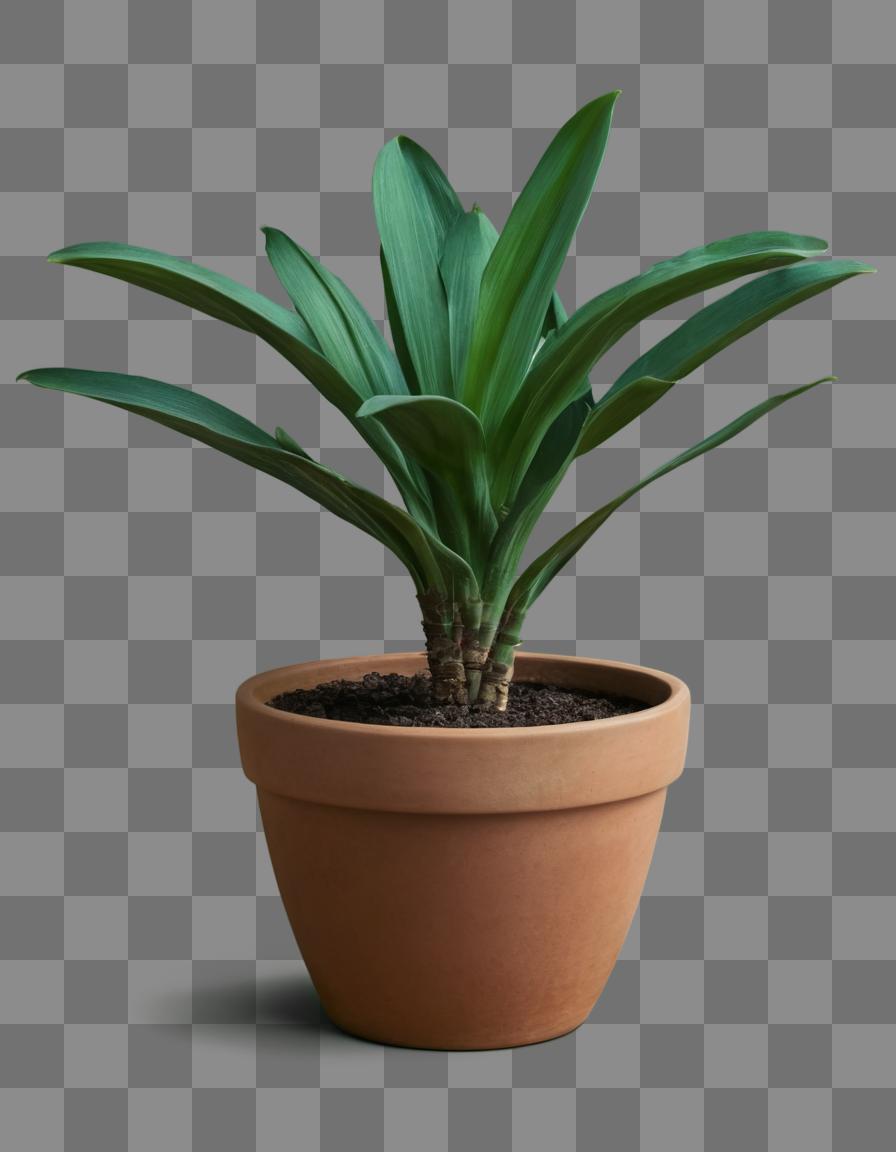}\hfill
\includegraphics[width=0.245\linewidth]{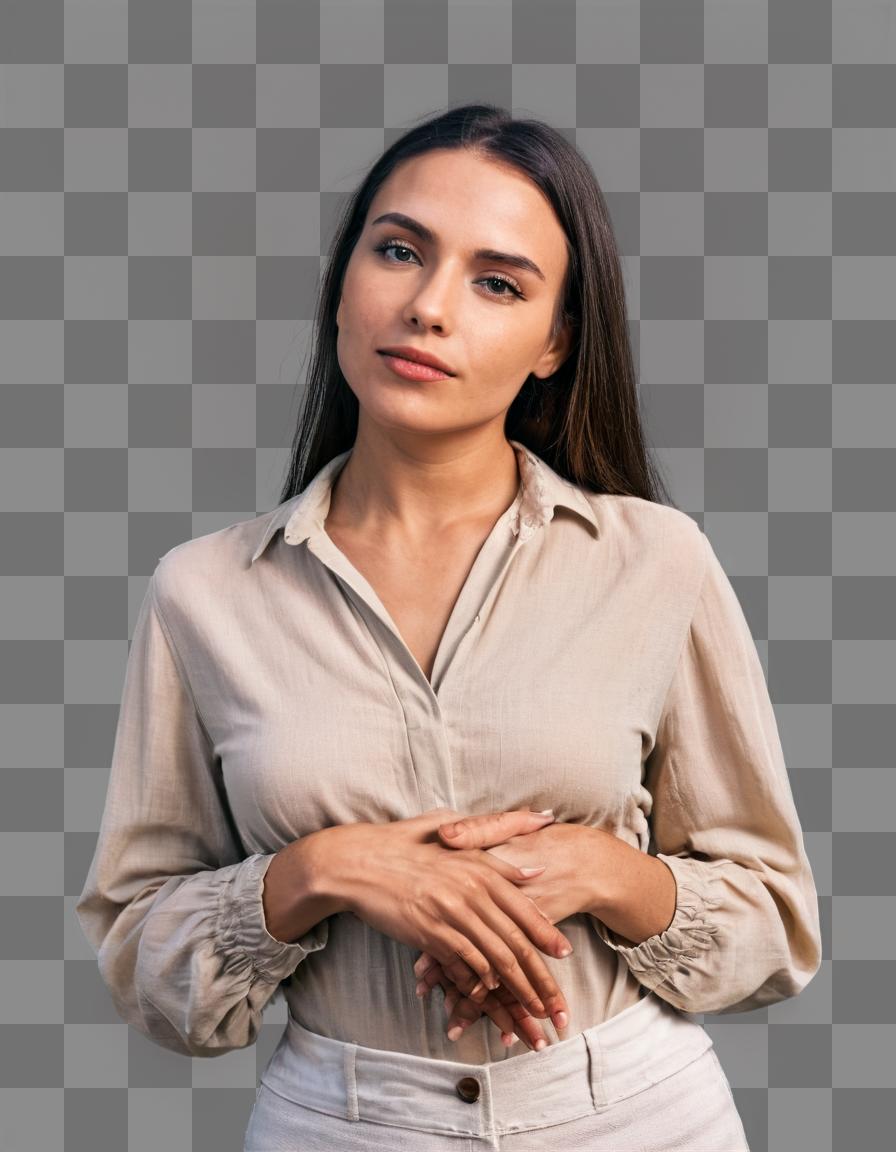}\hfill
\includegraphics[width=0.245\linewidth]{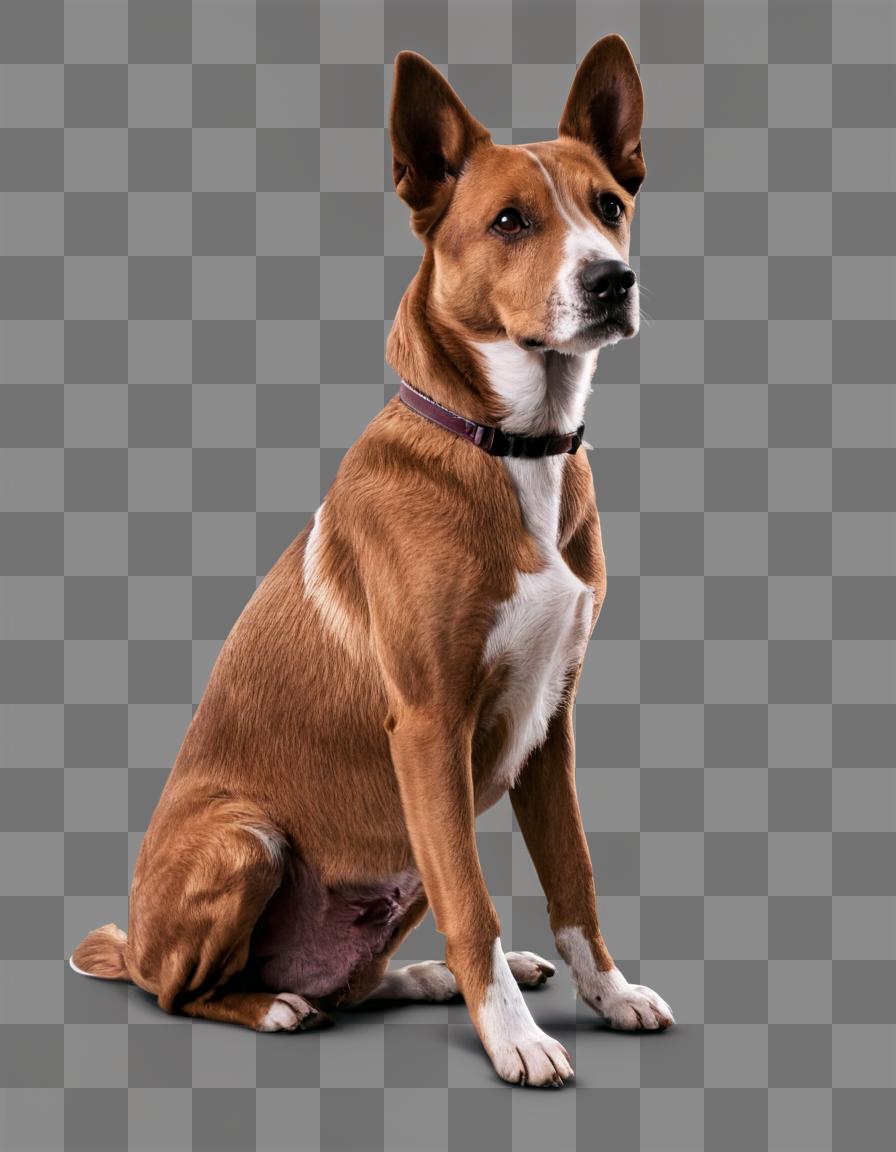}\hfill
\includegraphics[width=0.245\linewidth]{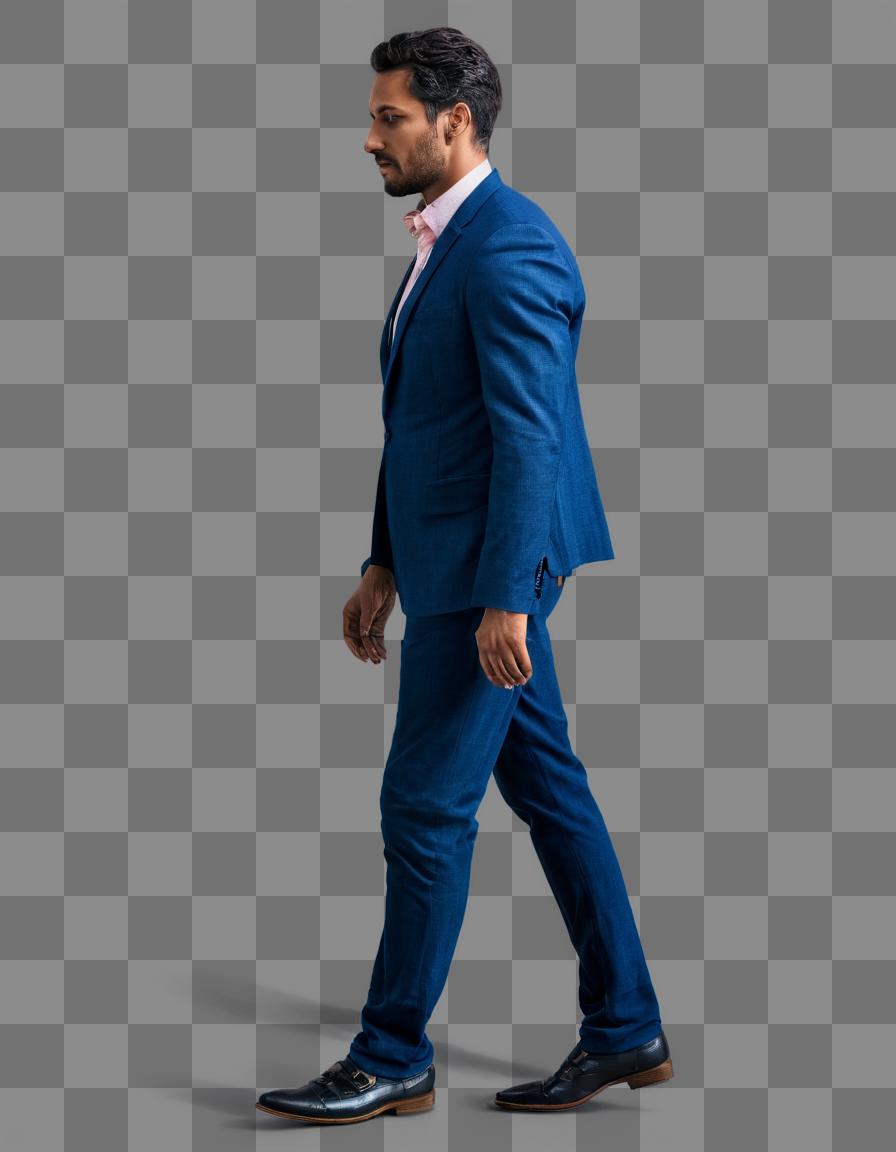}
\caption{Multi-layer Results \#1. The prompts are ``plant on table'', ``woman in room'', ``dog on floor'', ``man walking on street''. Resolution is $896\times1152$.}
\label{fig:c1}
\end{minipage}
\end{figure*}

\begin{figure*}

\begin{minipage}{\linewidth}
\includegraphics[width=0.245\linewidth]{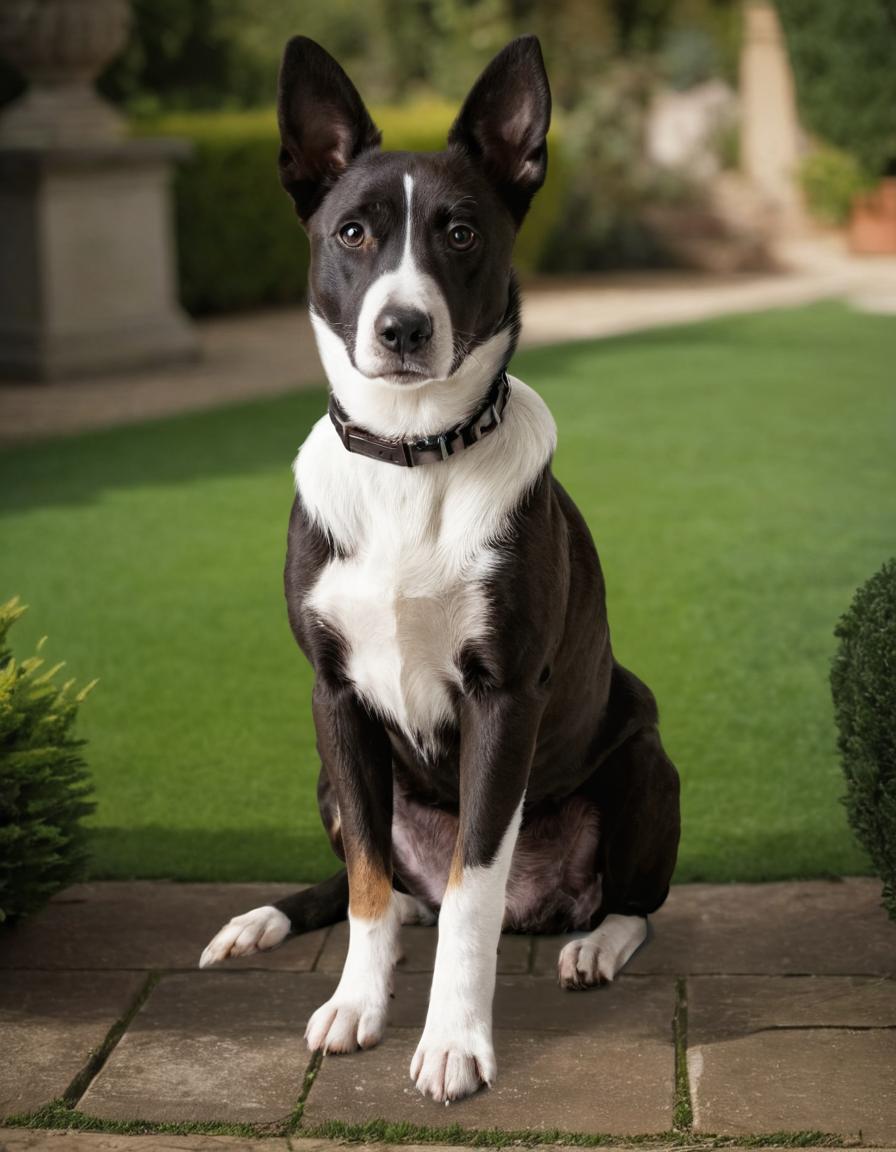}\hfill
\includegraphics[width=0.245\linewidth]{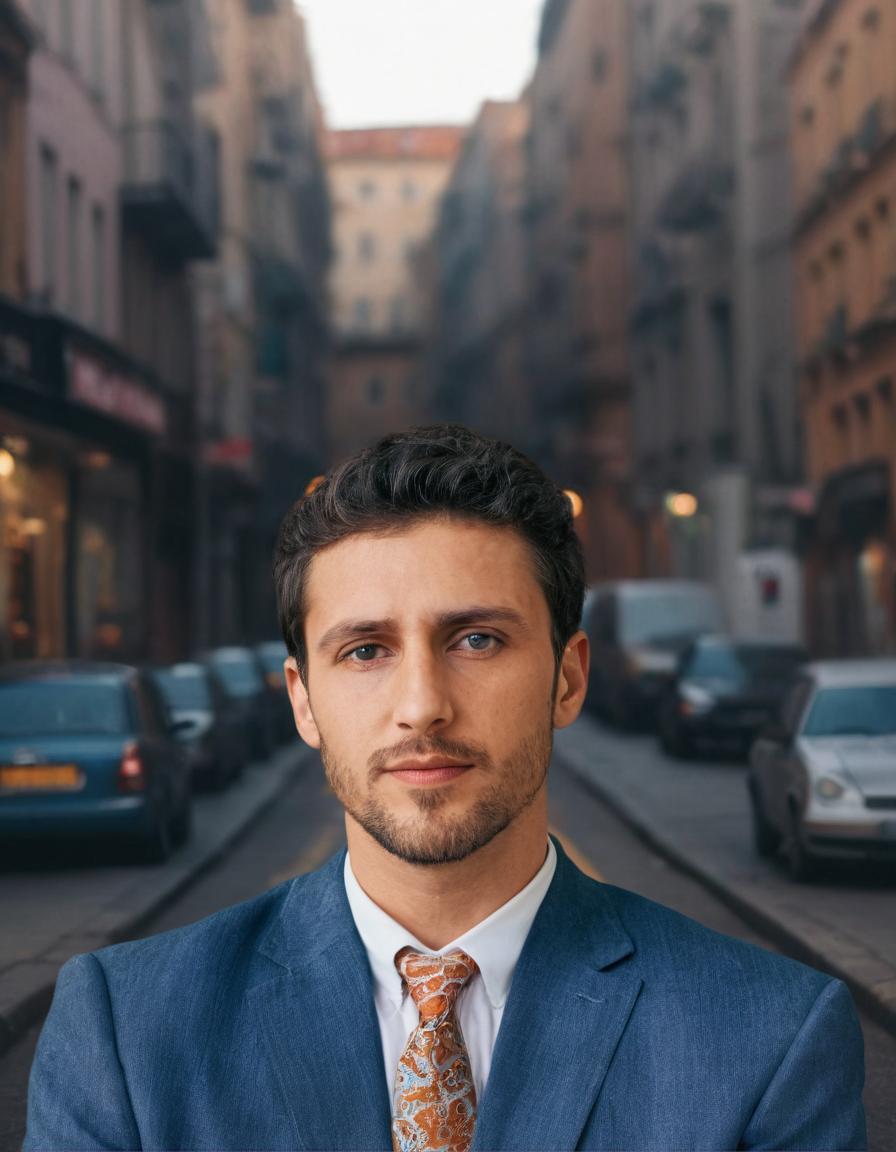}\hfill
\includegraphics[width=0.245\linewidth]{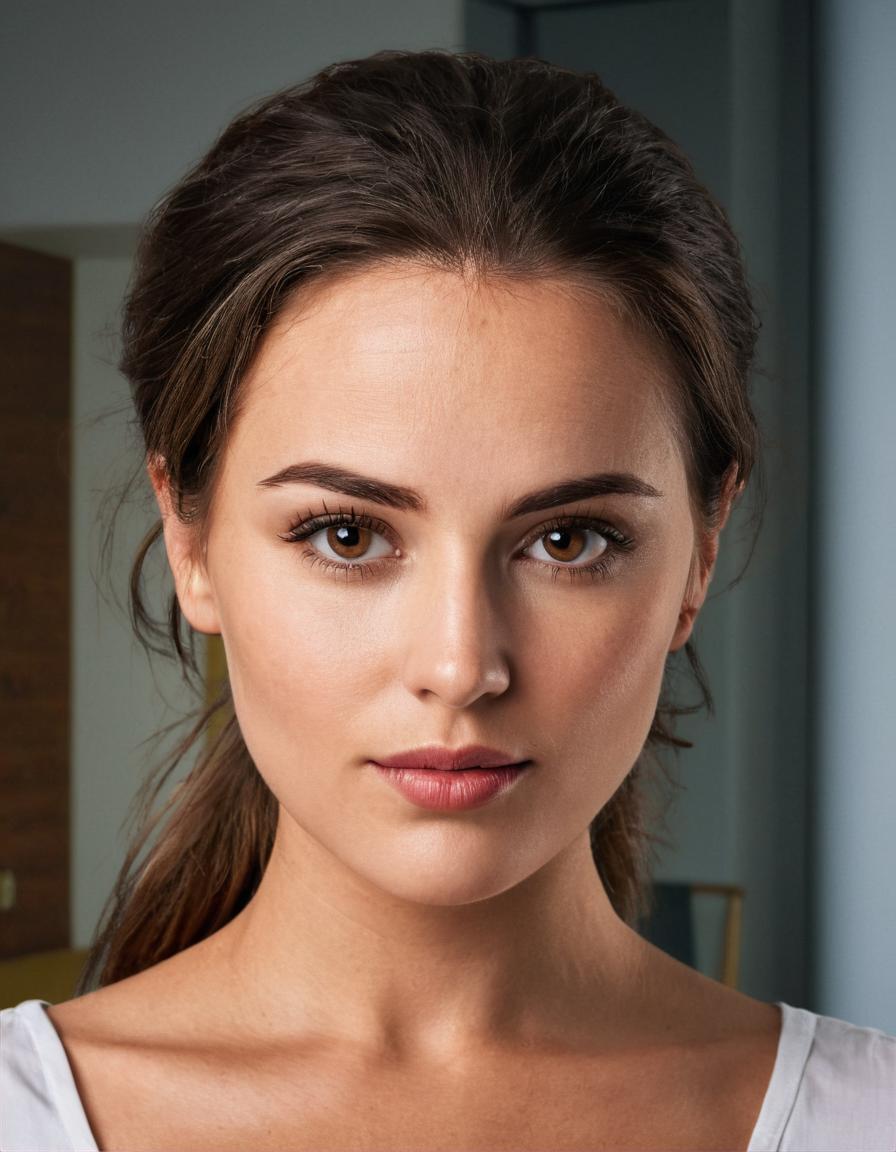}\hfill
\includegraphics[width=0.245\linewidth]{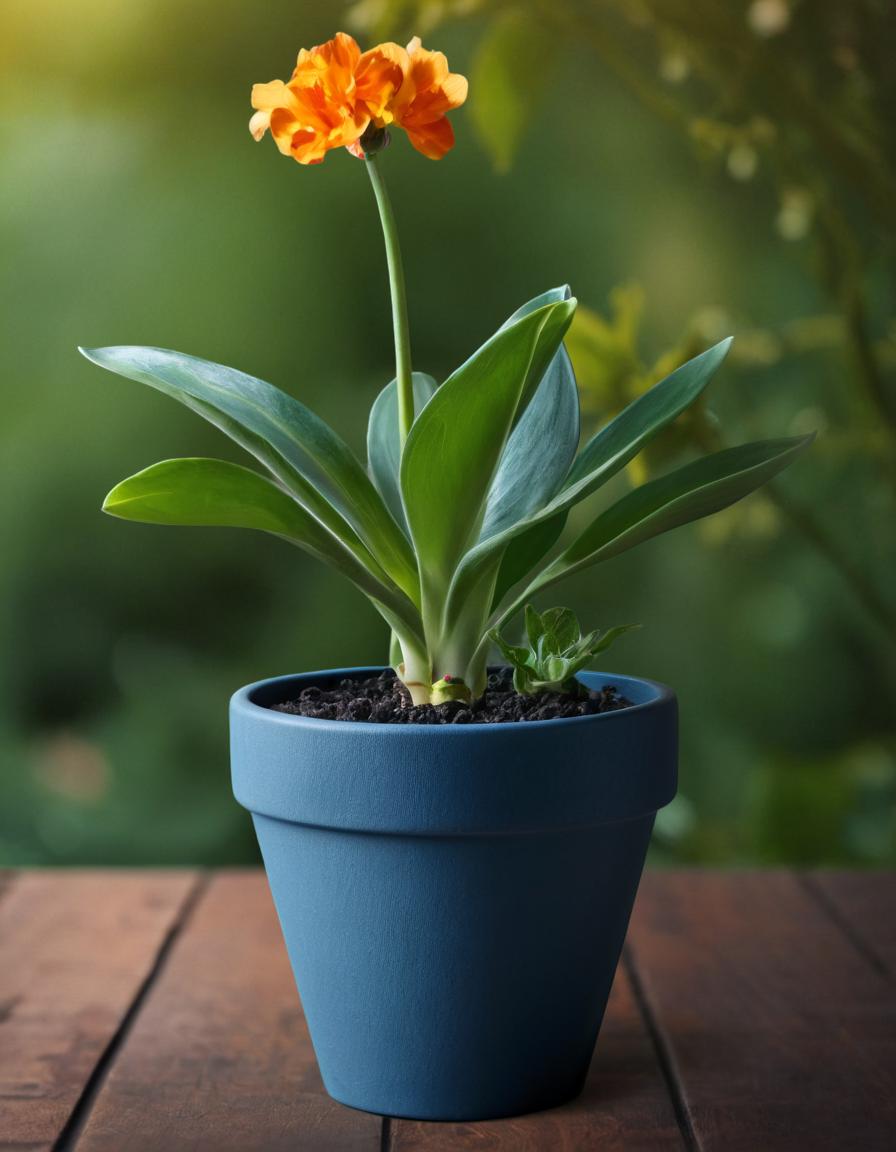}

\vspace{1pt}
\includegraphics[width=0.245\linewidth]{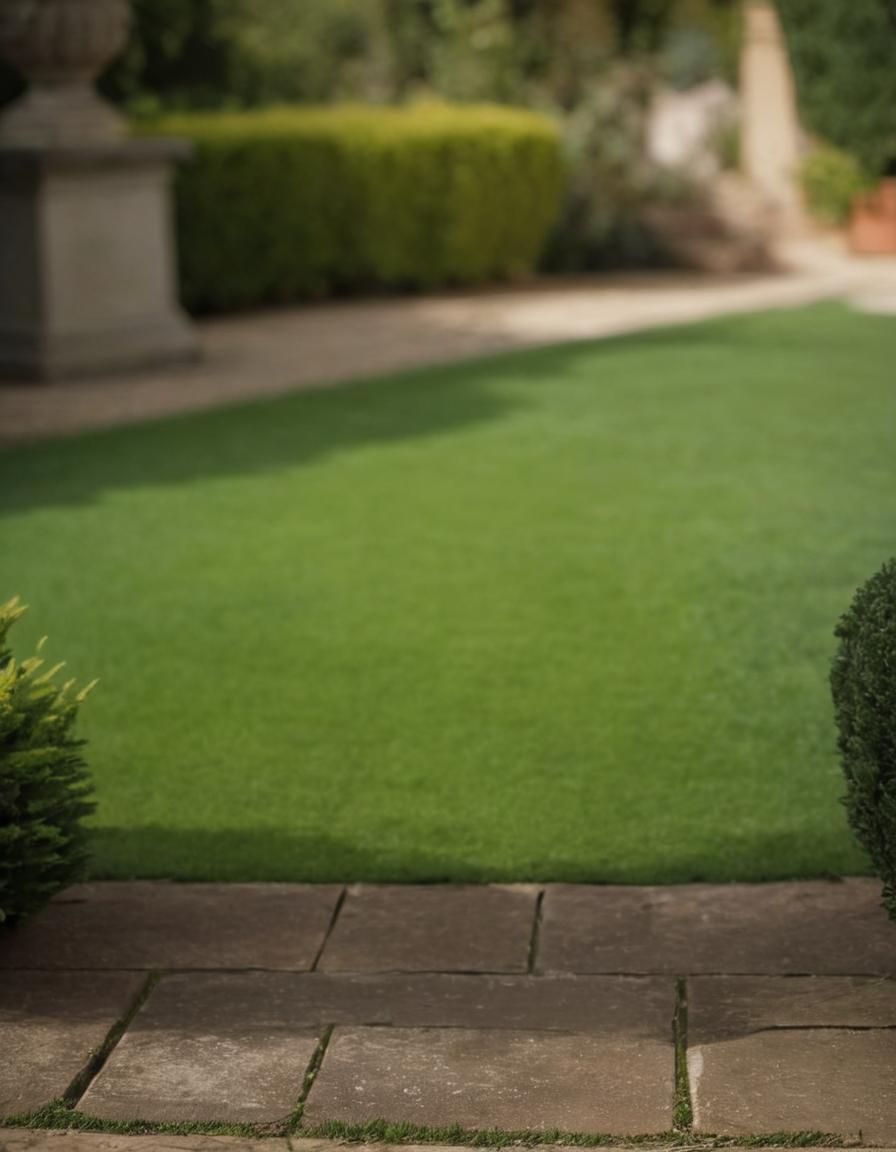}\hfill
\includegraphics[width=0.245\linewidth]{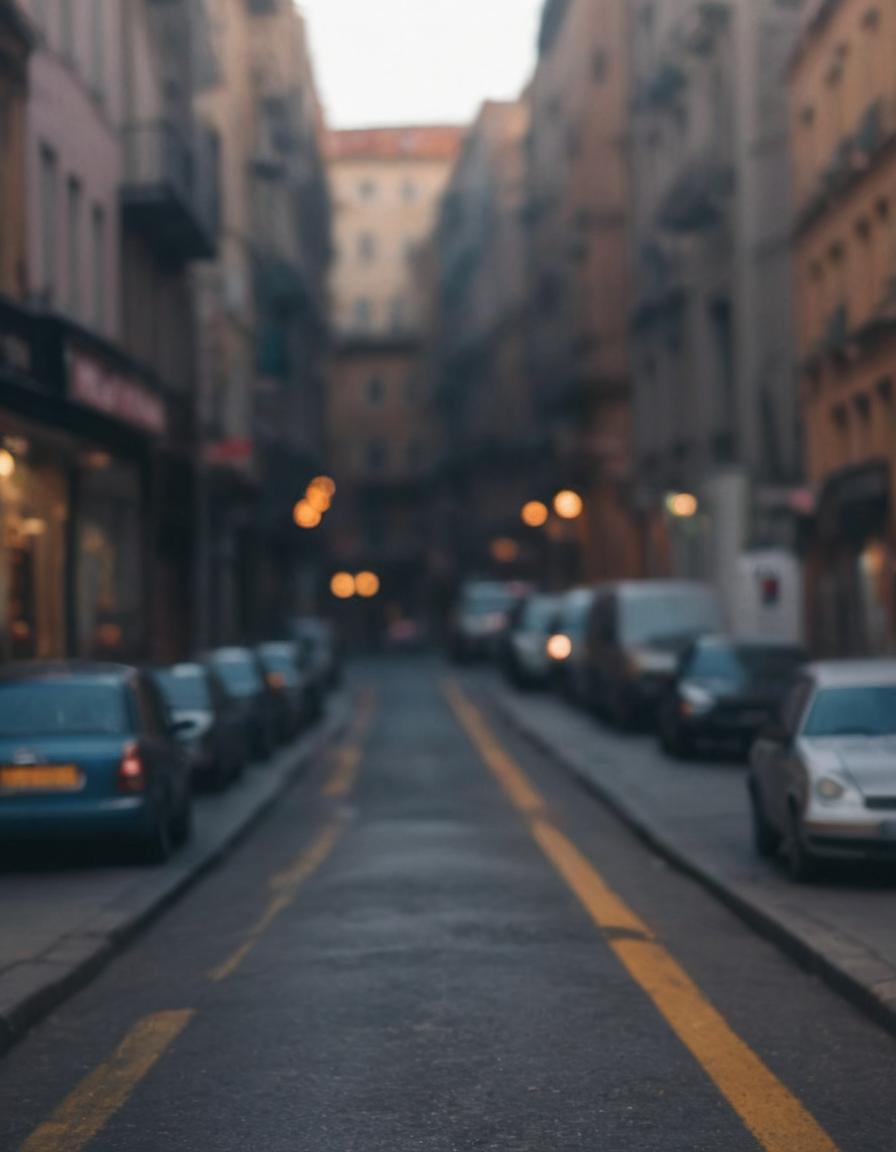}\hfill
\includegraphics[width=0.245\linewidth]{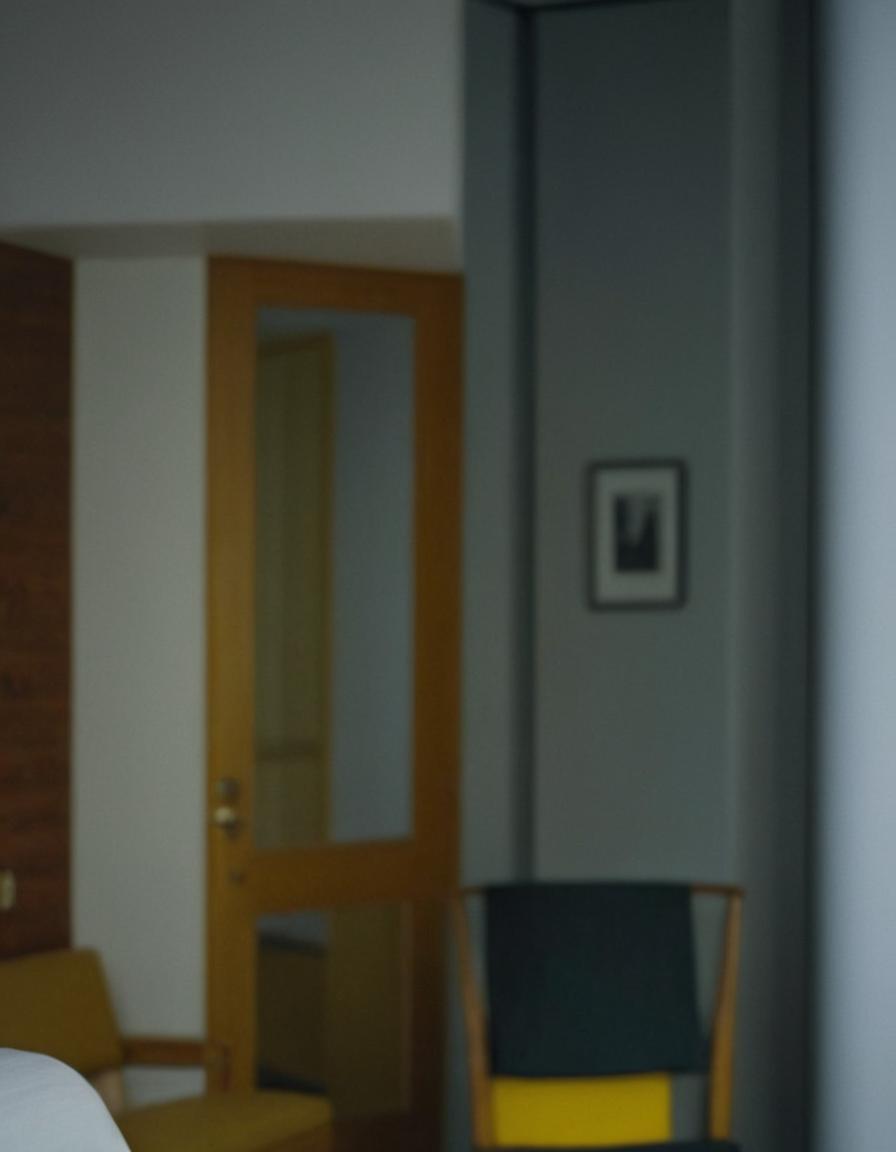}\hfill
\includegraphics[width=0.245\linewidth]{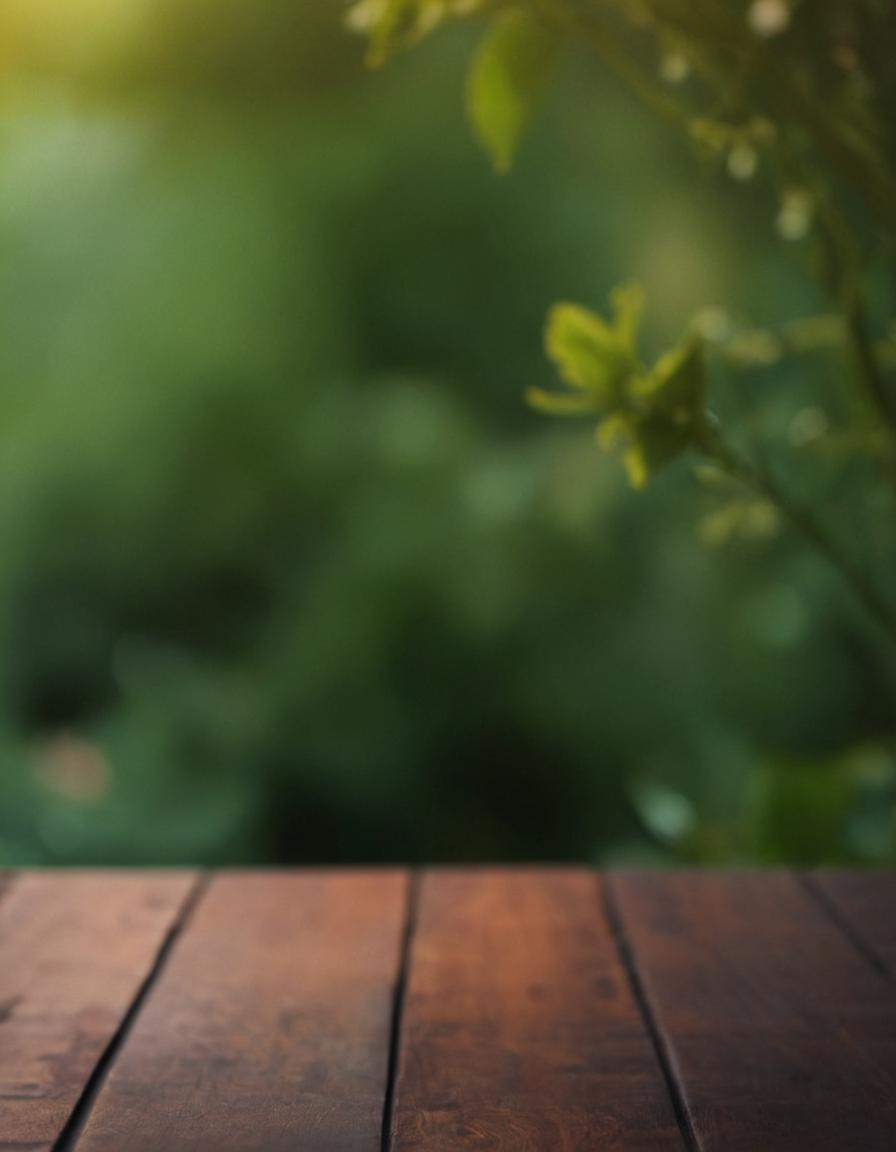}

\vspace{1pt}
\includegraphics[width=0.245\linewidth]{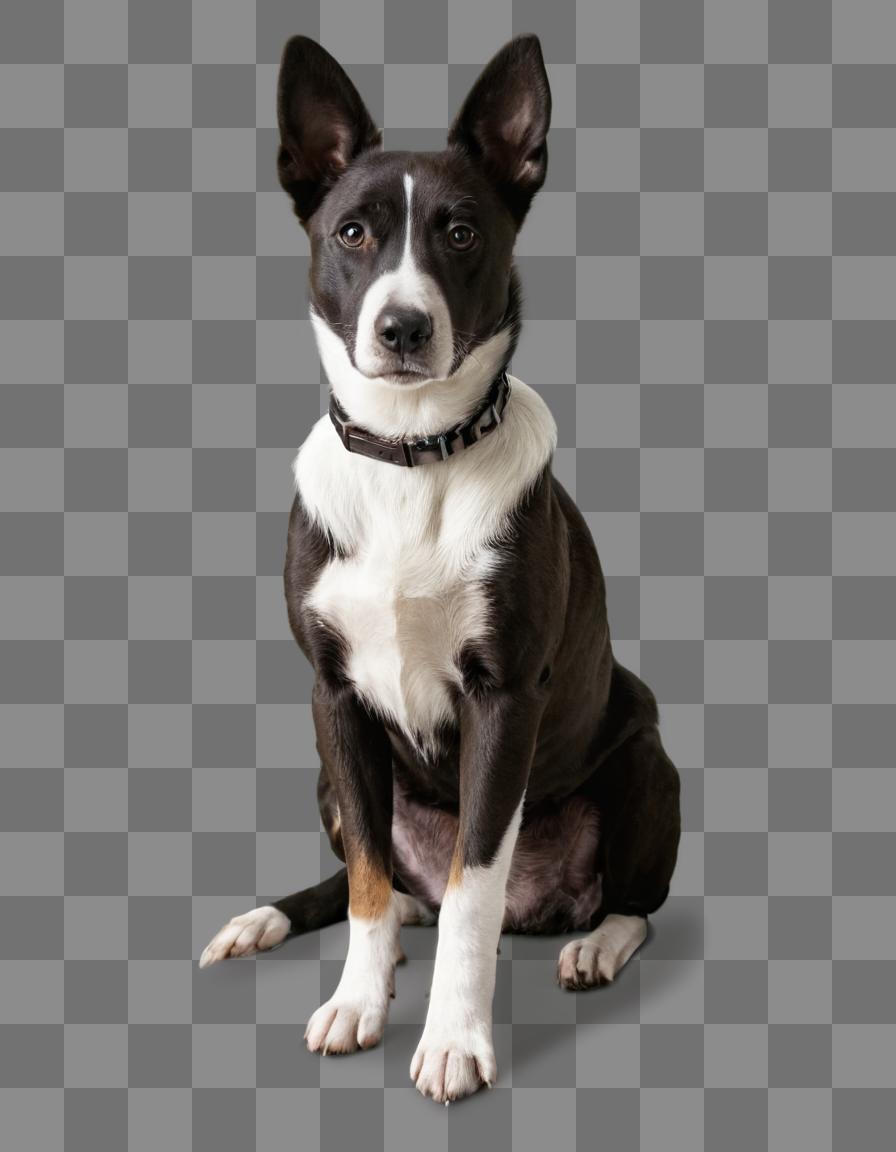}\hfill
\includegraphics[width=0.245\linewidth]{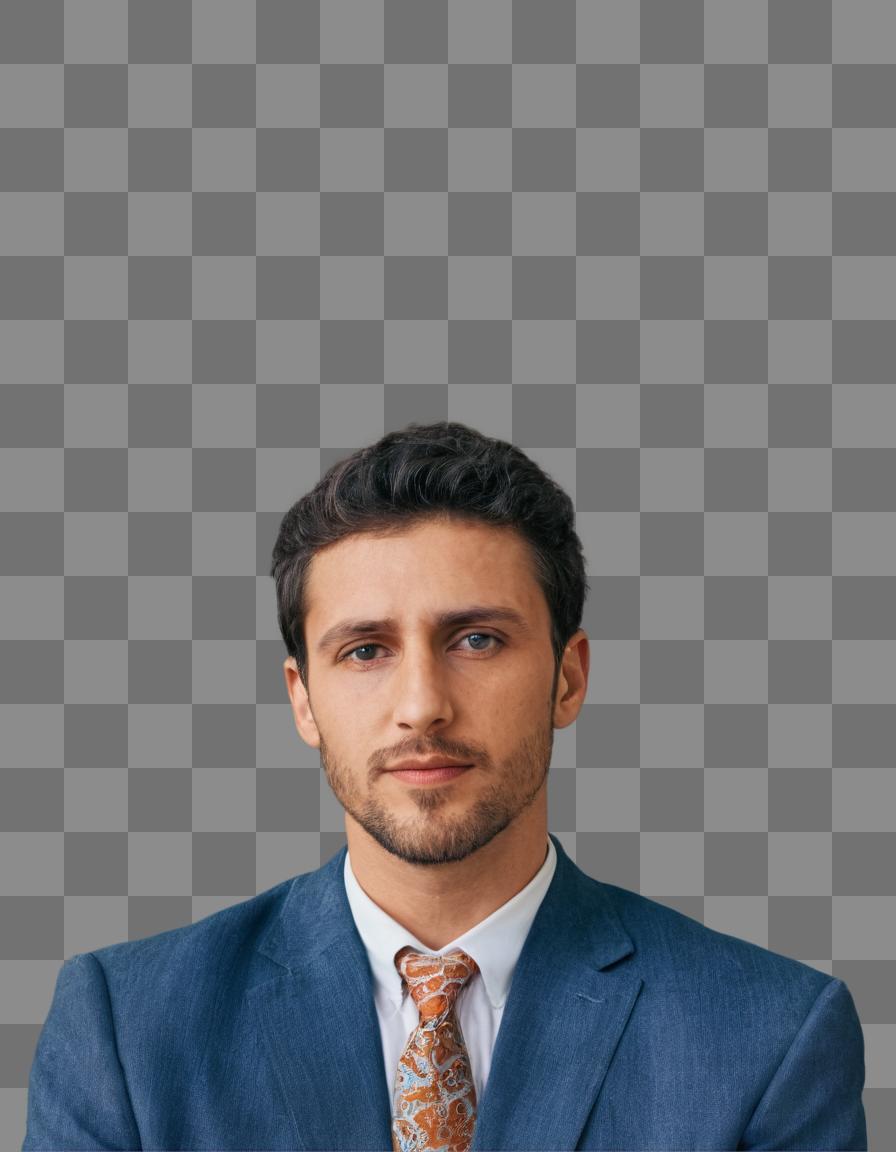}\hfill
\includegraphics[width=0.245\linewidth]{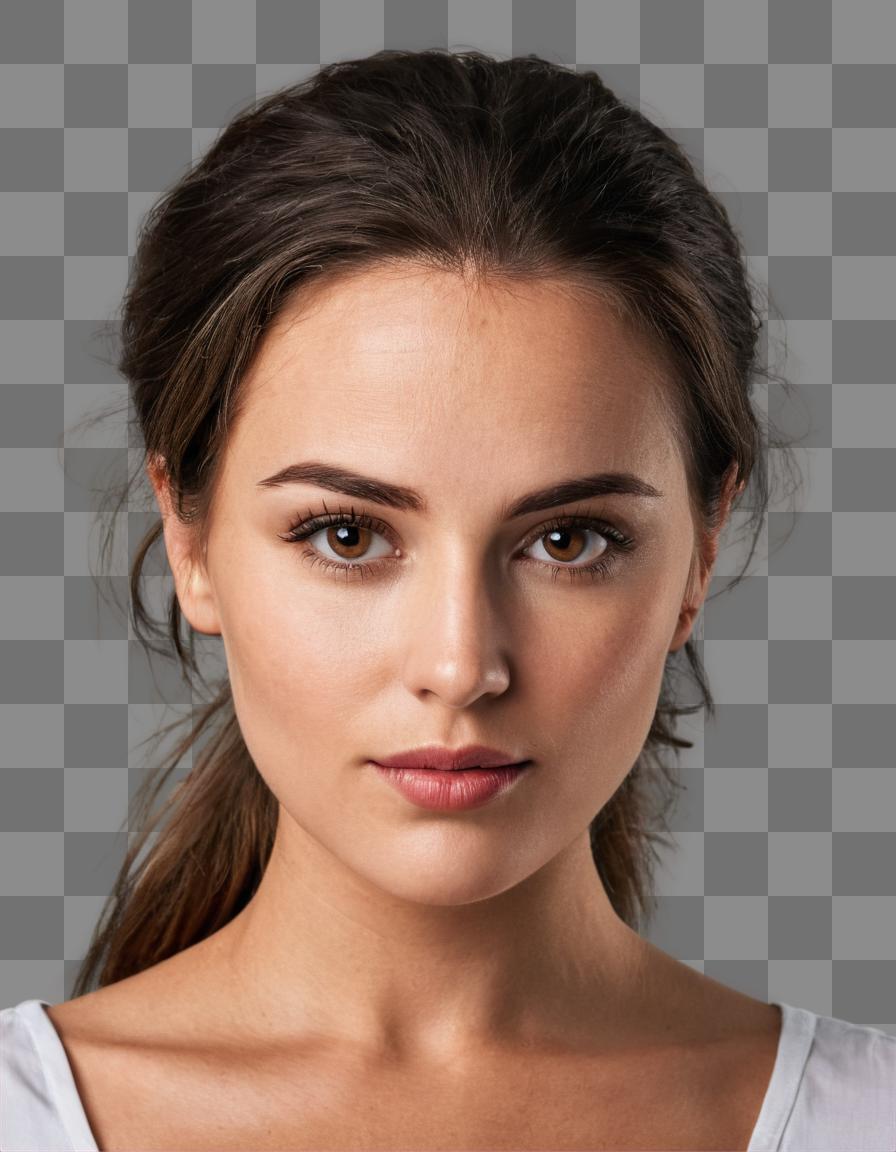}\hfill
\includegraphics[width=0.245\linewidth]{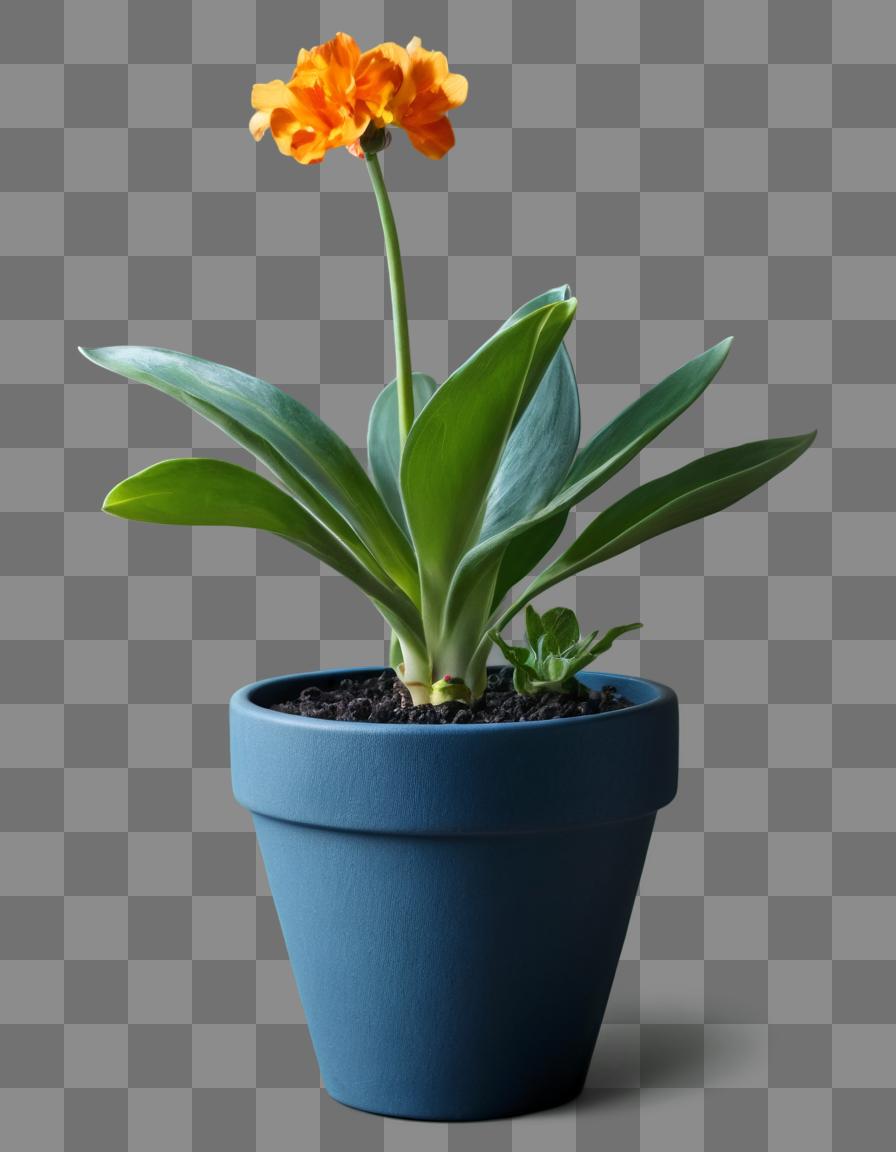}
\caption{Multi-layer Results \#2. The prompts are ``dog in garden'', ``man in street'', ``woman, closeup'', ``plants on table''. Resolution is $896\times1152$.}
\label{fig:c2}
\end{minipage}
\end{figure*}

\begin{figure*}

\begin{minipage}{\linewidth}
\includegraphics[width=0.245\linewidth]{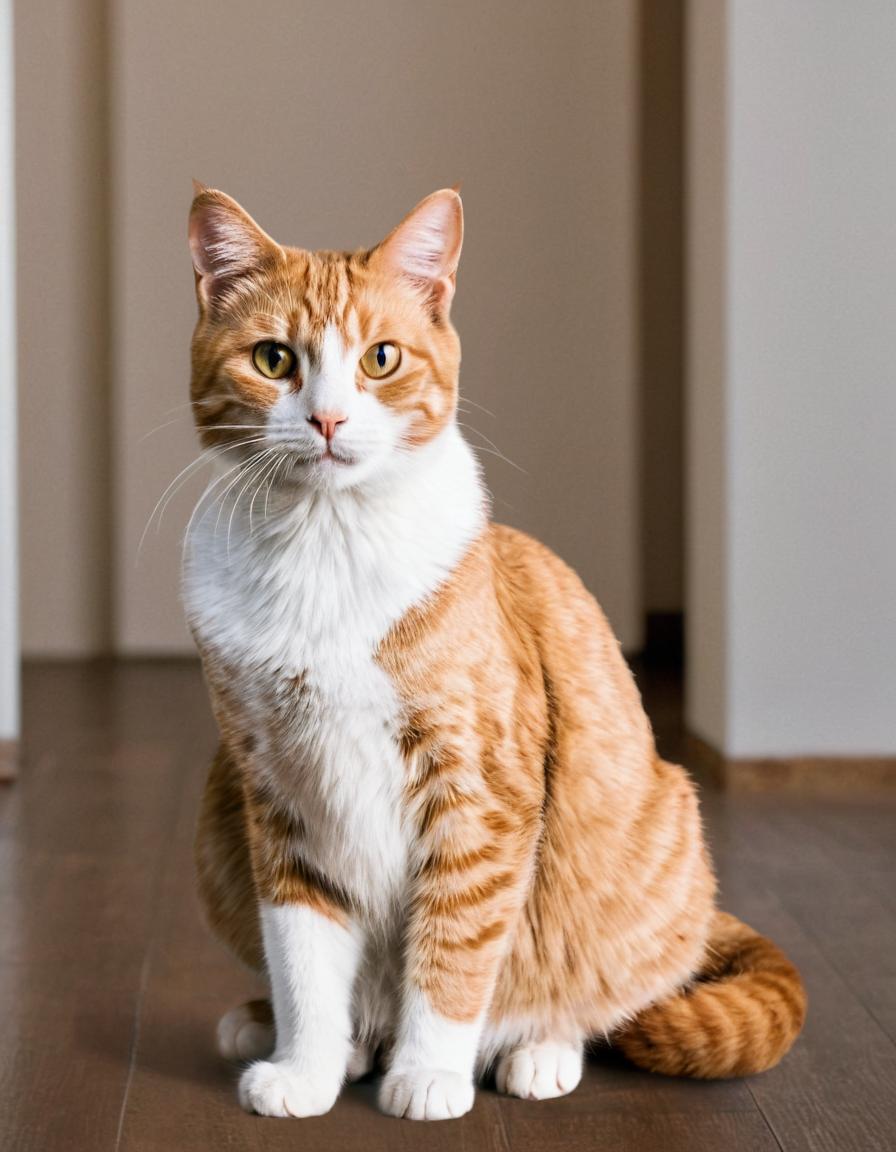}\hfill
\includegraphics[width=0.245\linewidth]{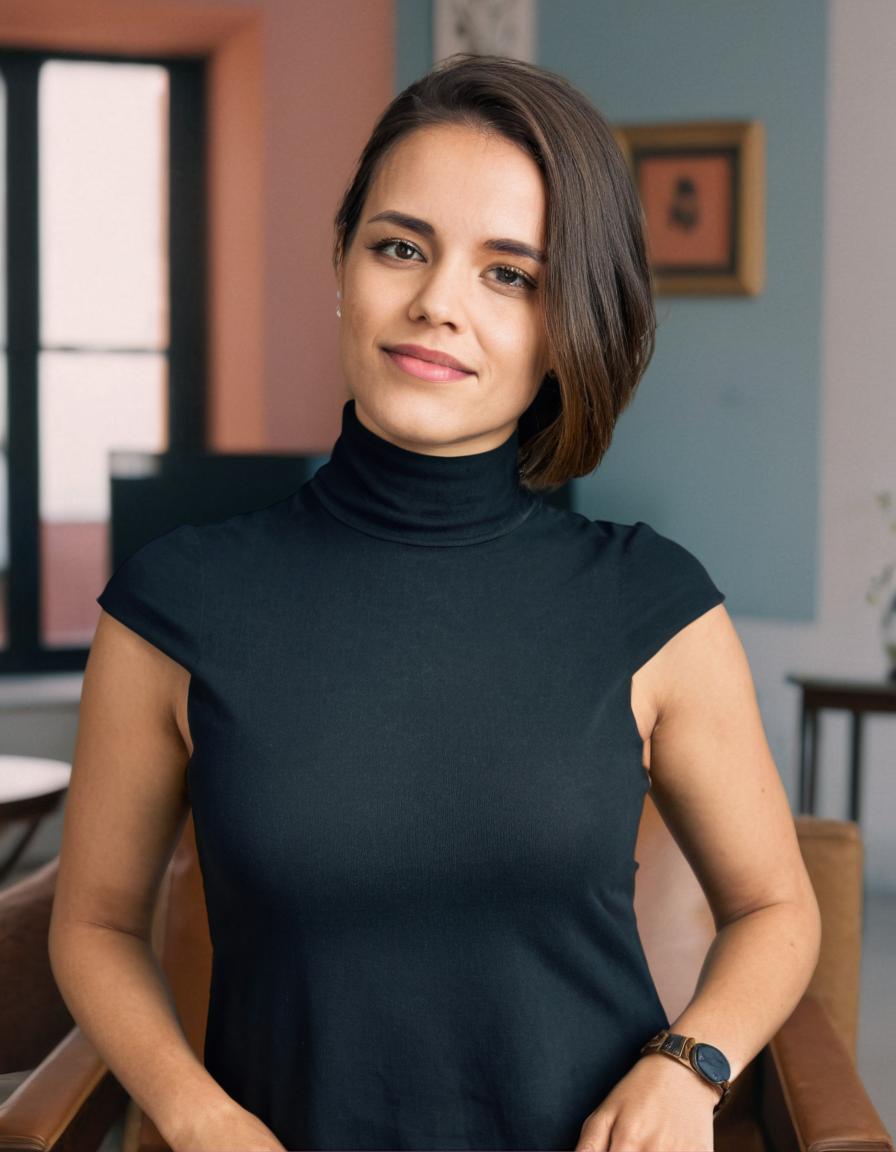}\hfill
\includegraphics[width=0.245\linewidth]{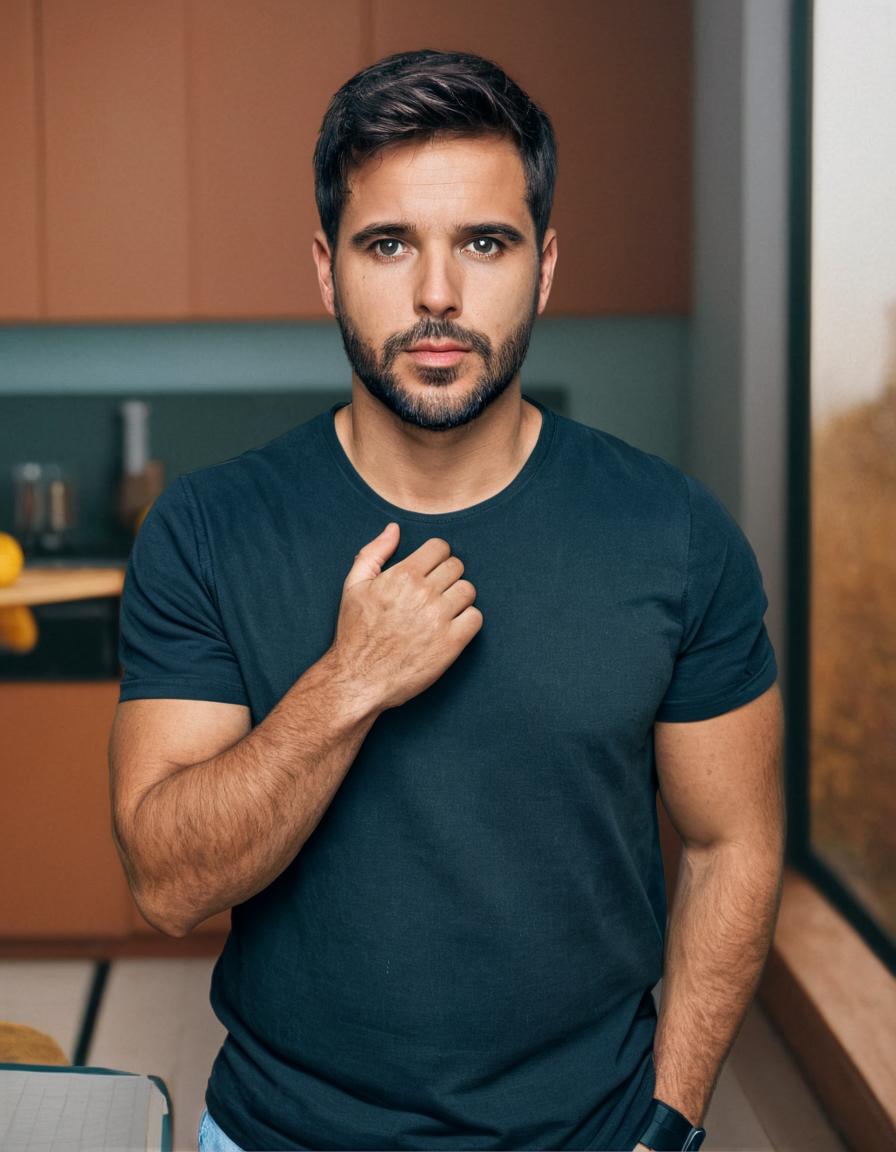}\hfill
\includegraphics[width=0.245\linewidth]{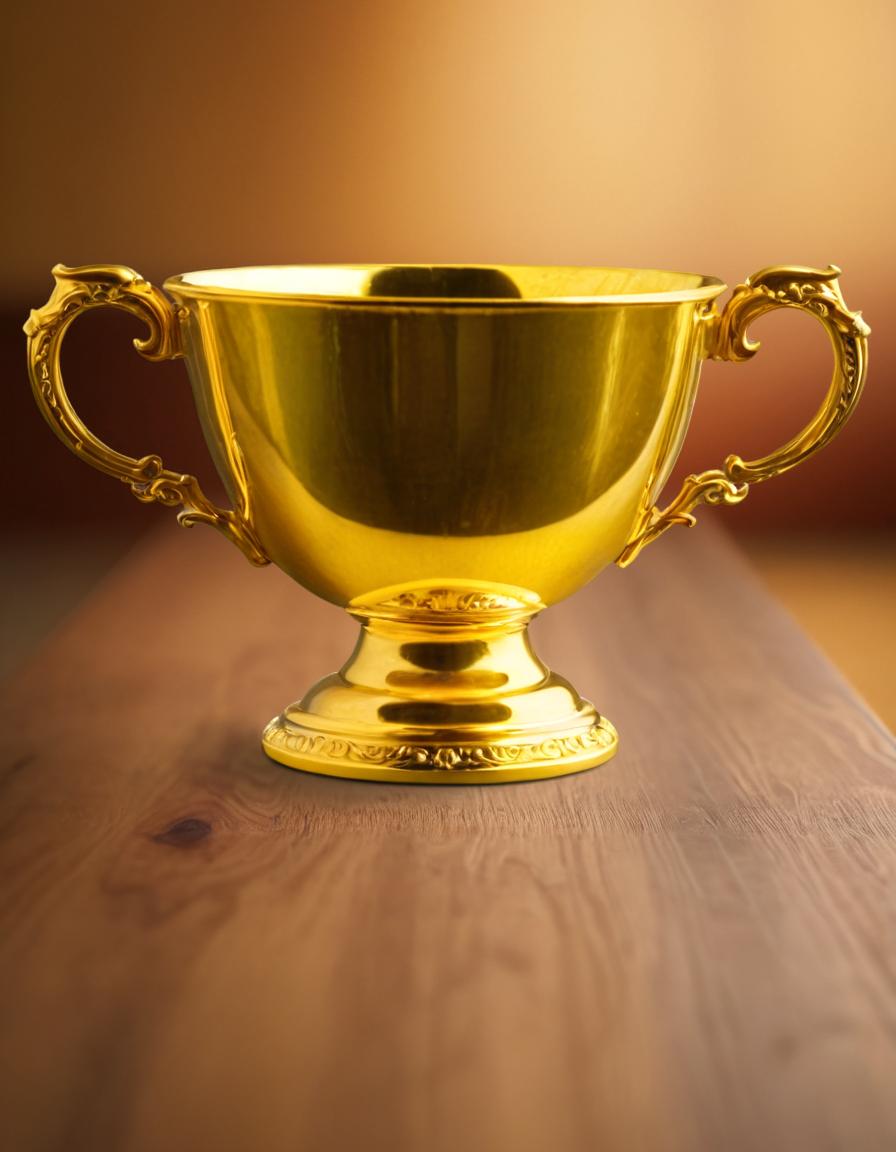}

\vspace{1pt}
\includegraphics[width=0.245\linewidth]{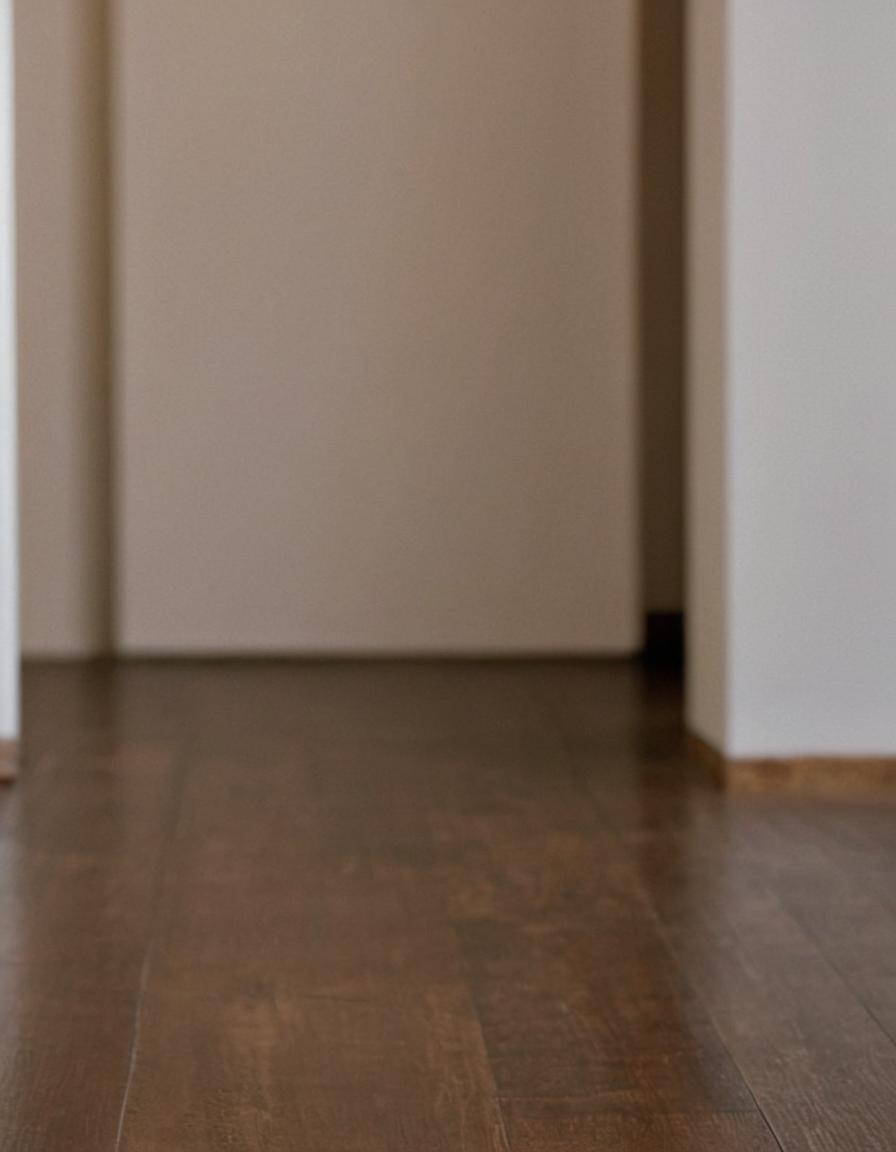}\hfill
\includegraphics[width=0.245\linewidth]{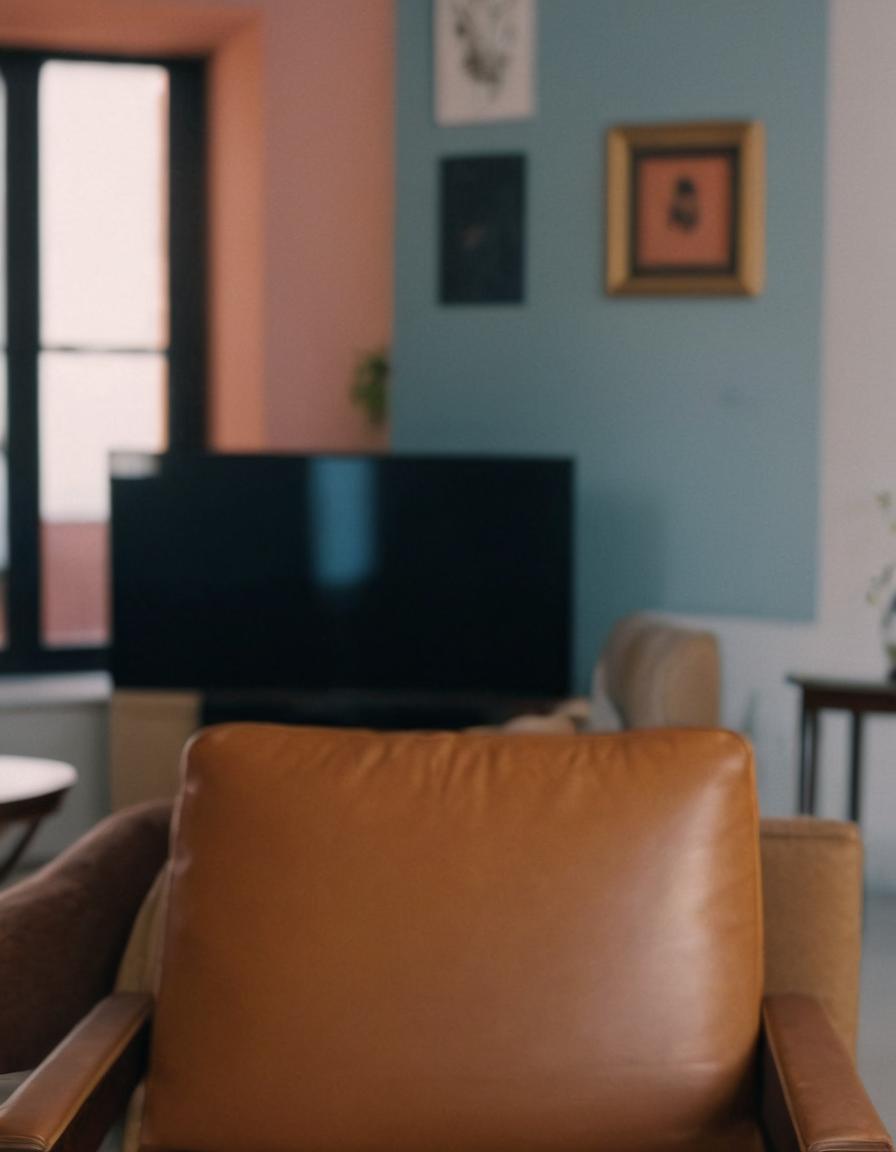}\hfill
\includegraphics[width=0.245\linewidth]{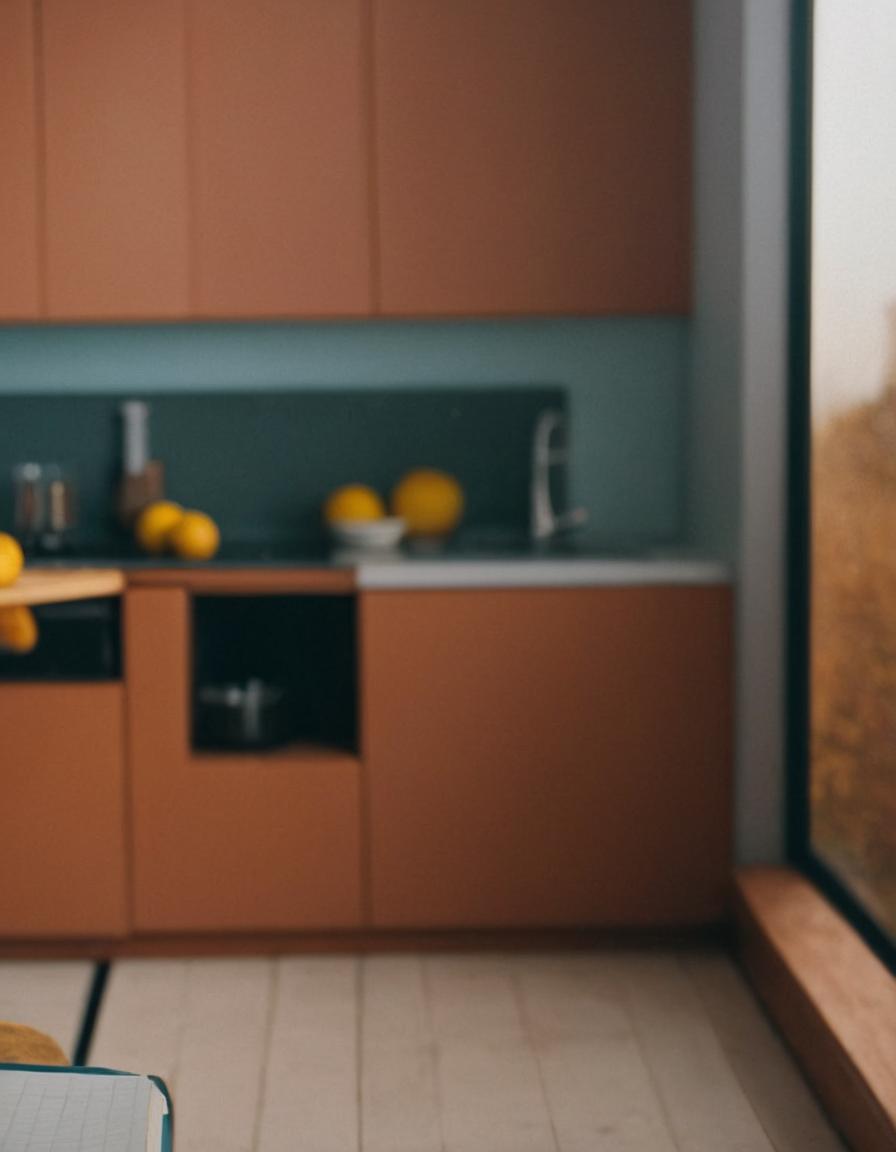}\hfill
\includegraphics[width=0.245\linewidth]{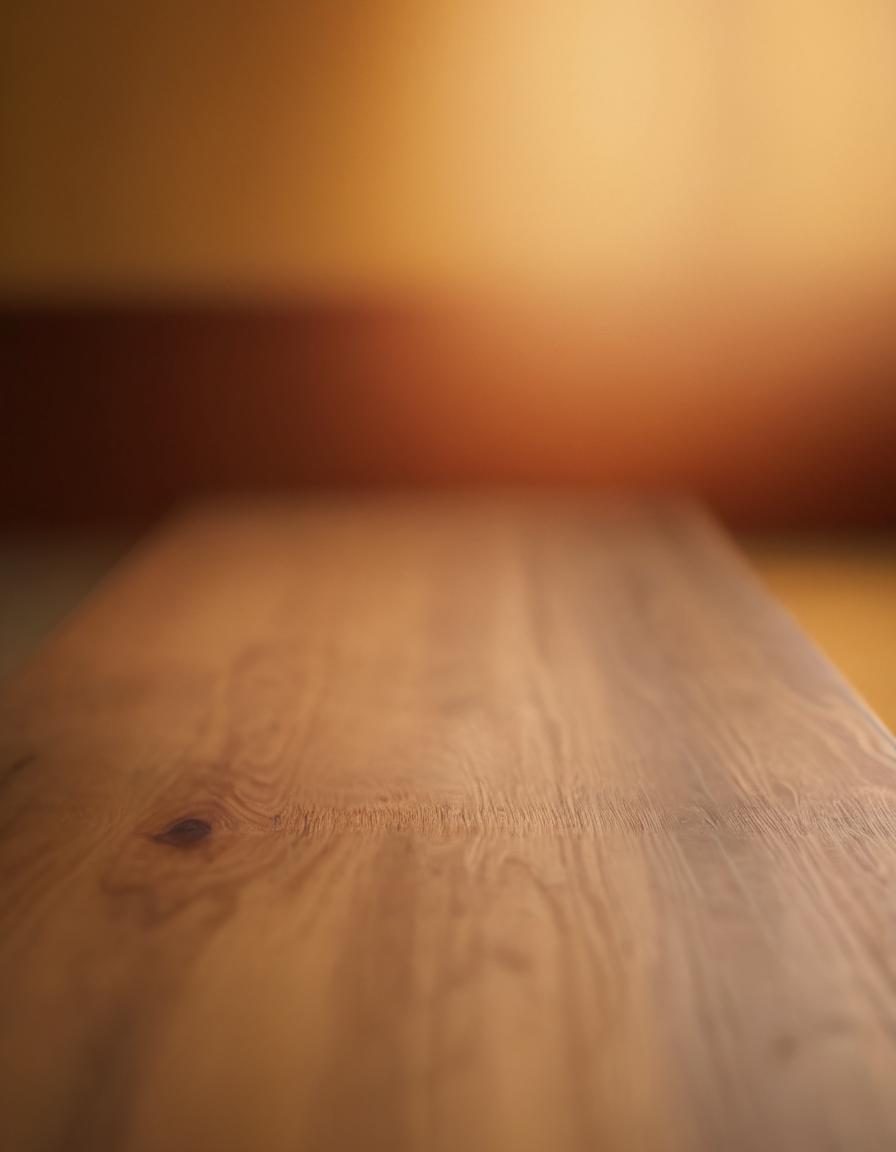}

\vspace{1pt}
\includegraphics[width=0.245\linewidth]{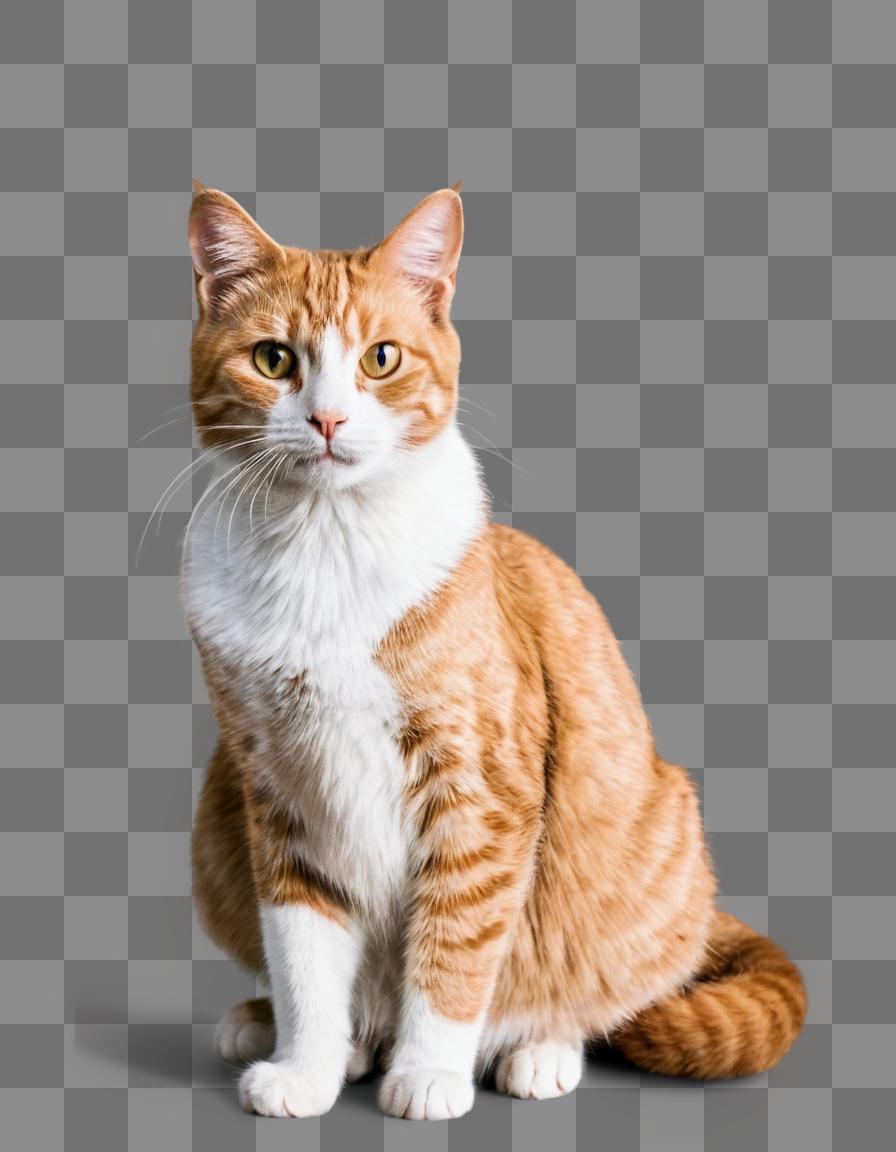}\hfill
\includegraphics[width=0.245\linewidth]{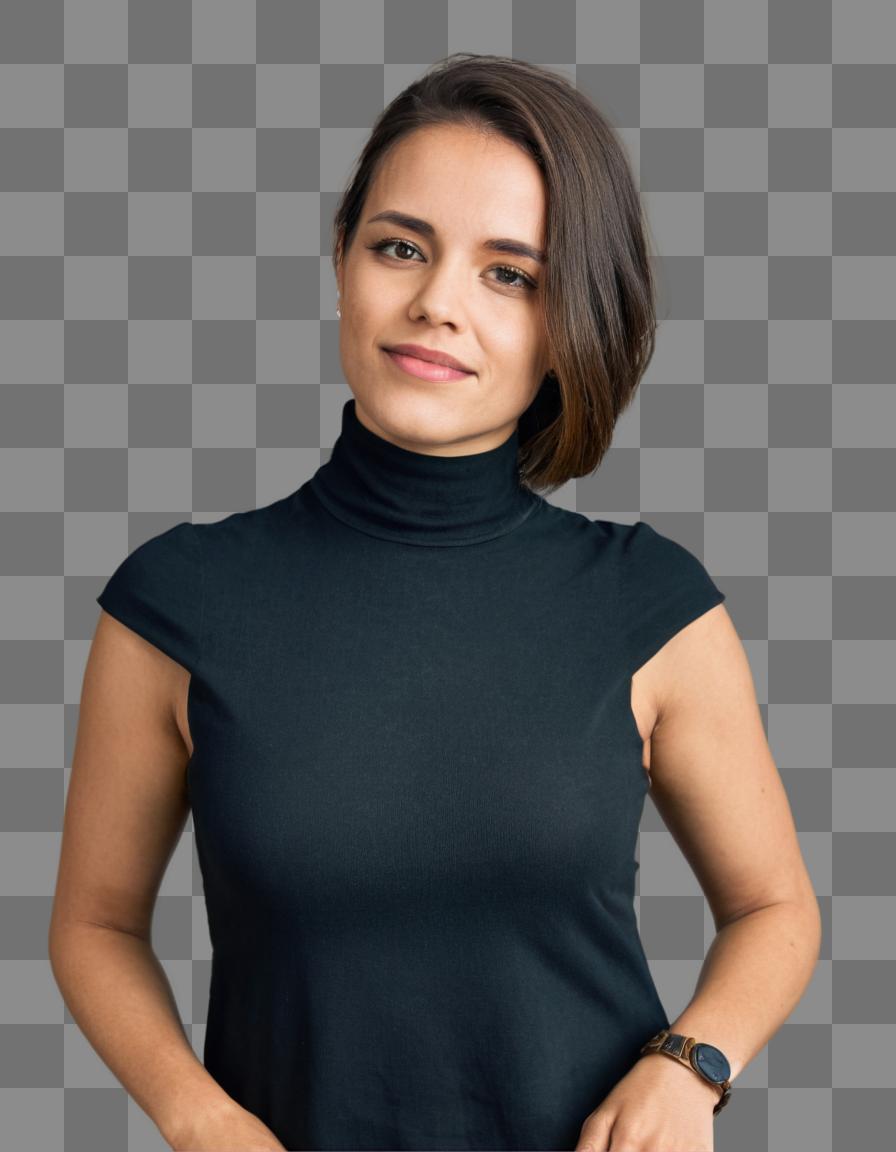}\hfill
\includegraphics[width=0.245\linewidth]{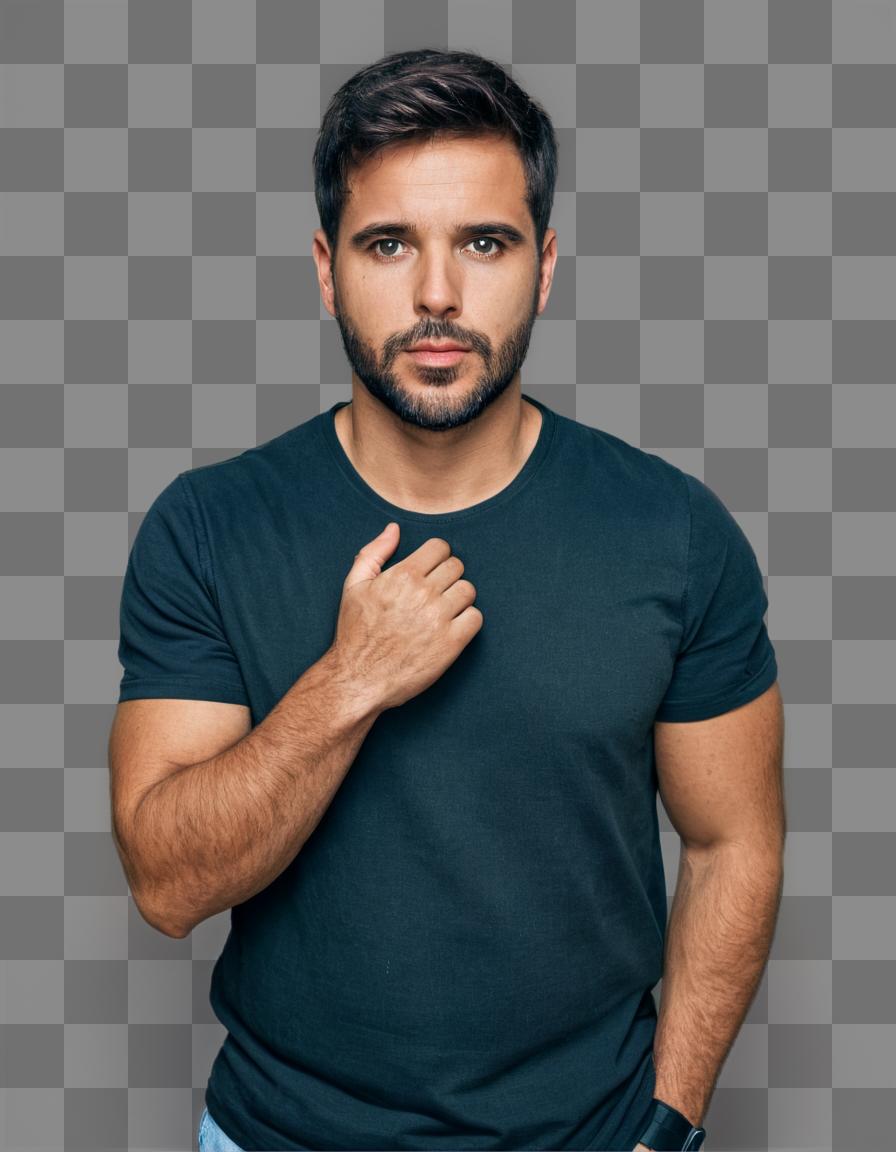}\hfill
\includegraphics[width=0.245\linewidth]{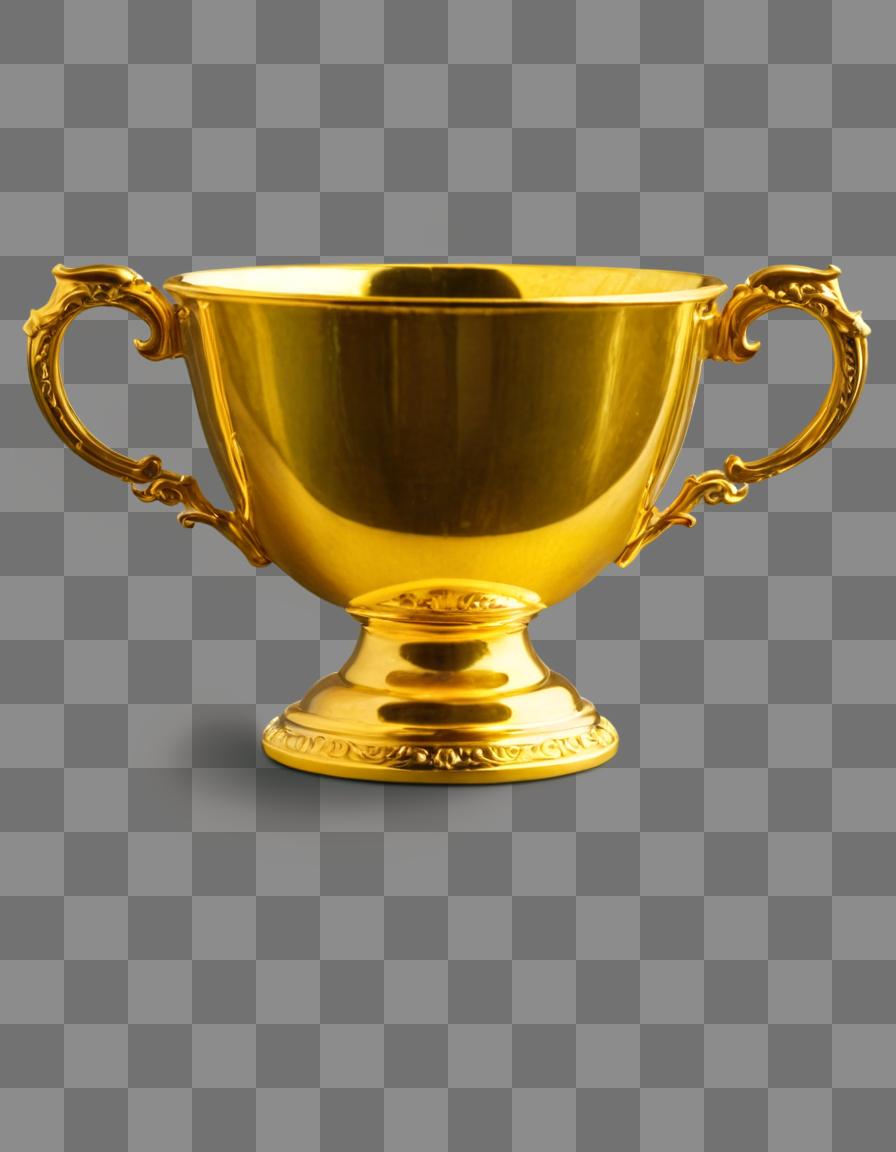}
\caption{Multi-layer Results \#3. The prompts are ``cat on floor'', ``woman in room'', ``man in room'', ``golden cup''. Resolution is $896\times1152$.}
\label{fig:c3}
\end{minipage}
\end{figure*}

We present additional results for multiple transparent layers, from Figure~\ref{fig:c1} to Figure~\ref{fig:c3}.

\section{Foreground-Conditioned Backgrounds}
	
\begin{figure*}

\begin{minipage}{\linewidth}
\includegraphics[width=0.2\linewidth]{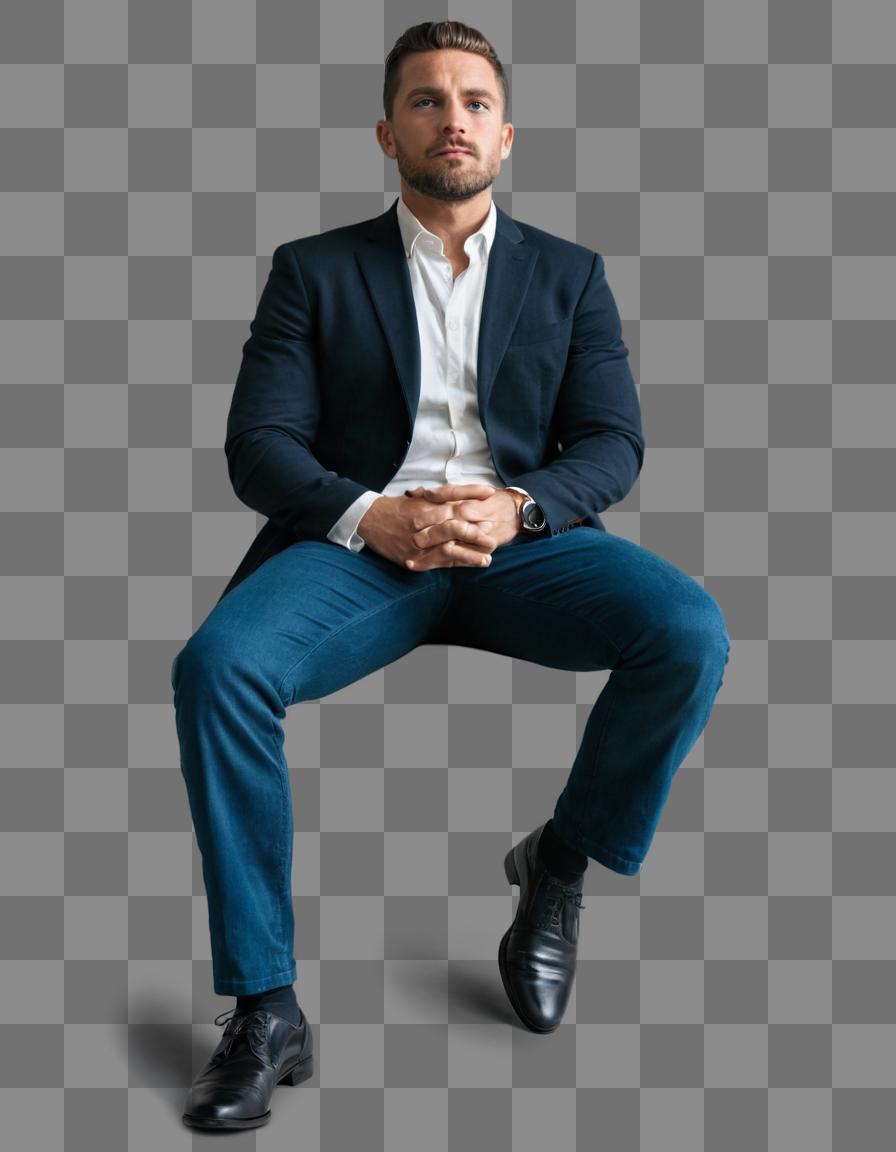}\hfill
\includegraphics[width=0.2\linewidth]{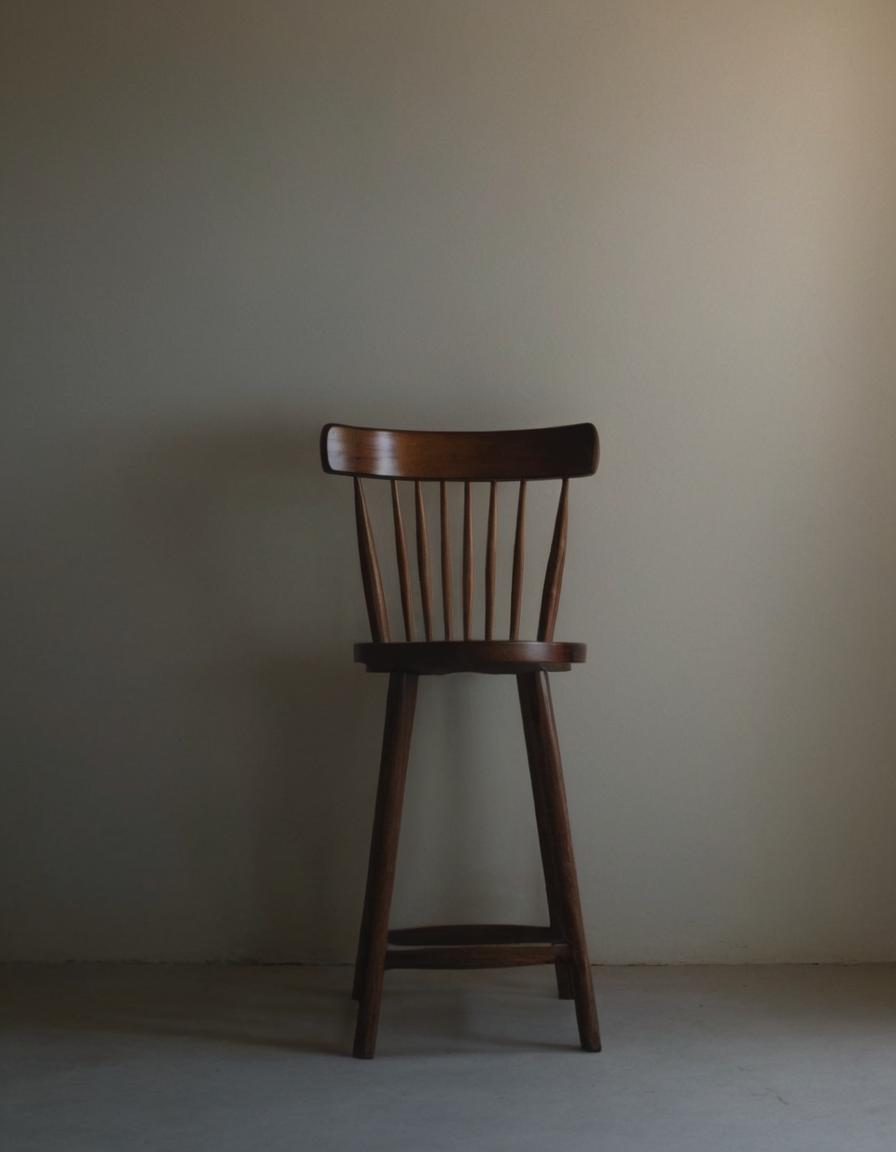}\hfill
\includegraphics[width=0.2\linewidth]{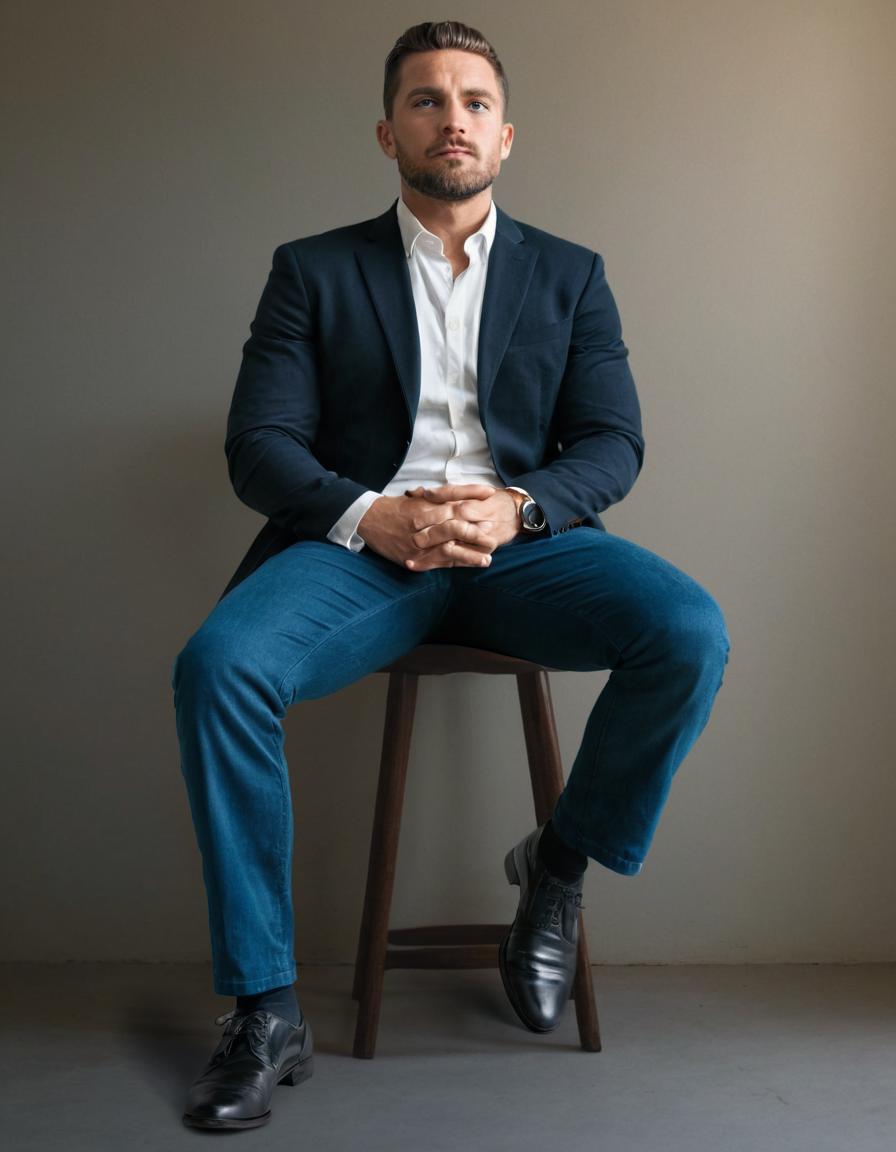}\hfill
\includegraphics[width=0.2\linewidth]{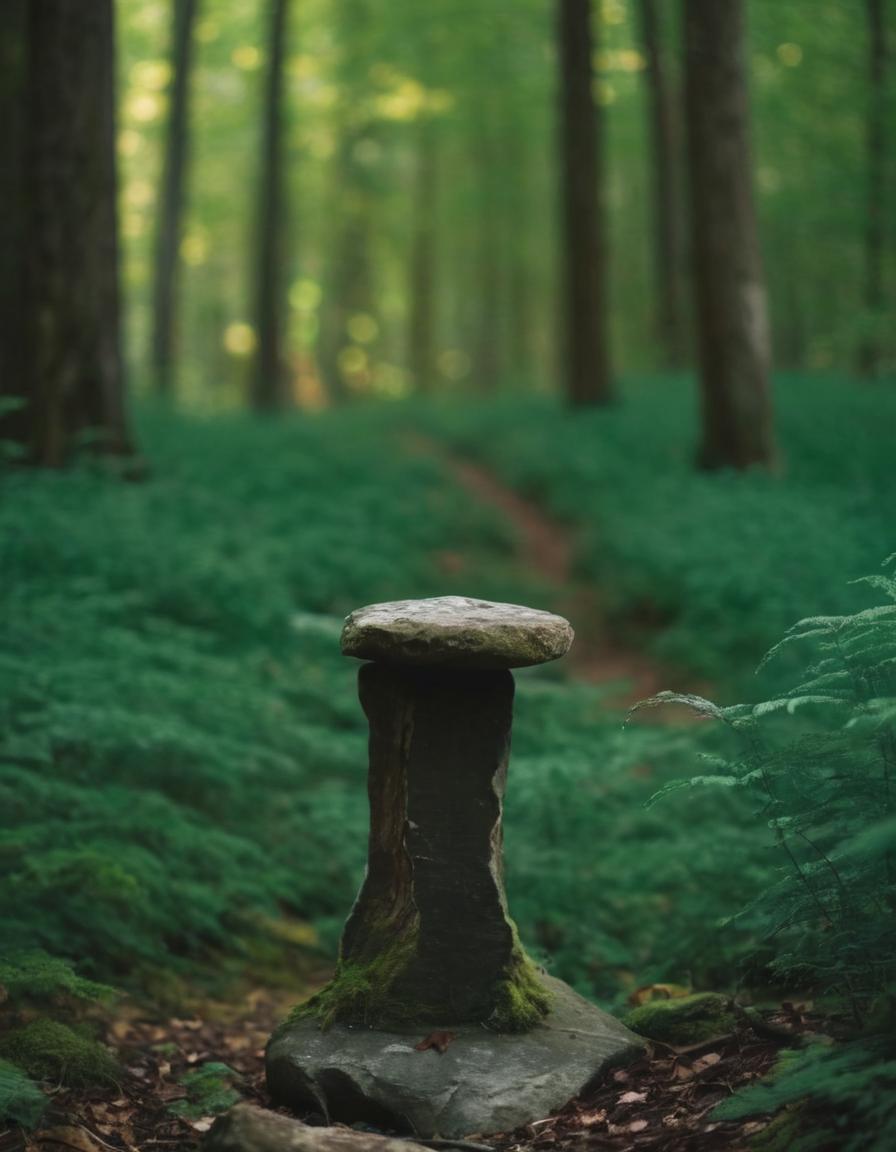}\hfill
\includegraphics[width=0.2\linewidth]{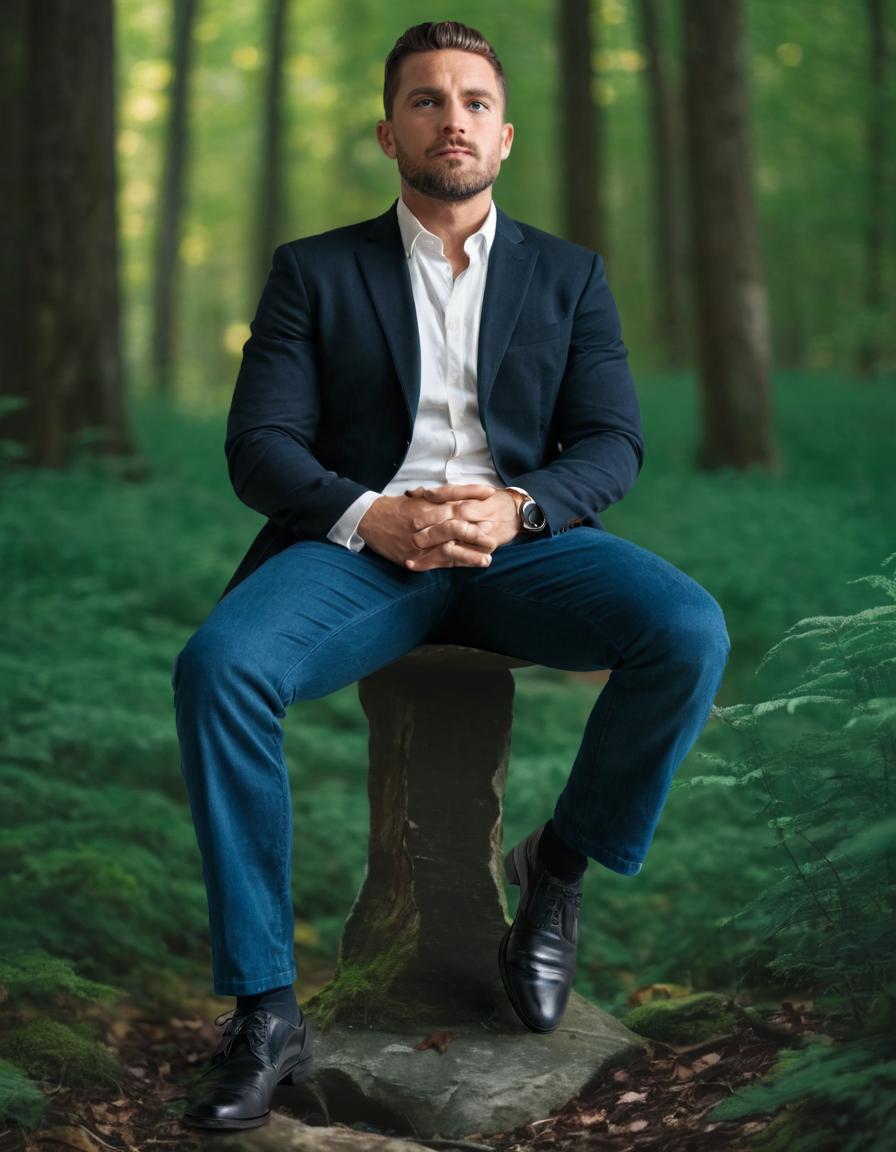}

\vspace{1pt}
\includegraphics[width=0.2\linewidth]{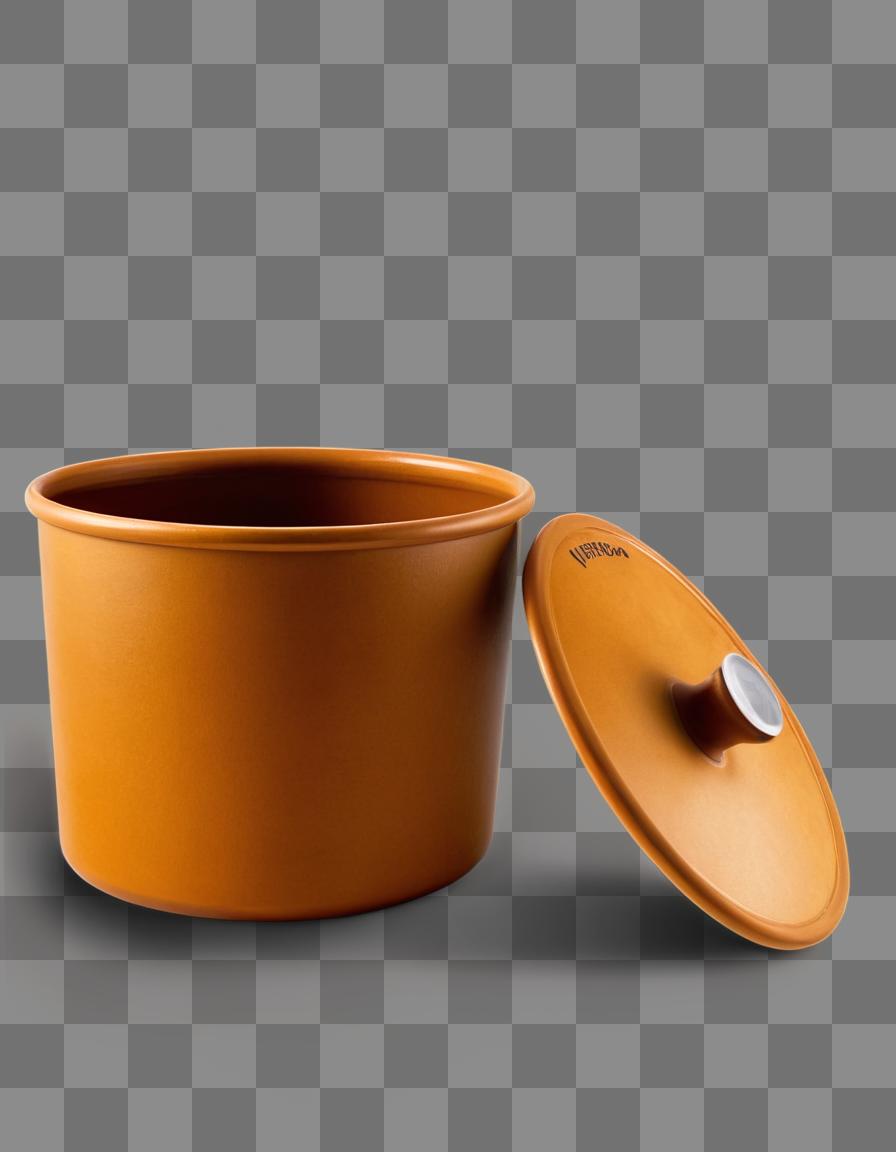}\hfill
\includegraphics[width=0.2\linewidth]{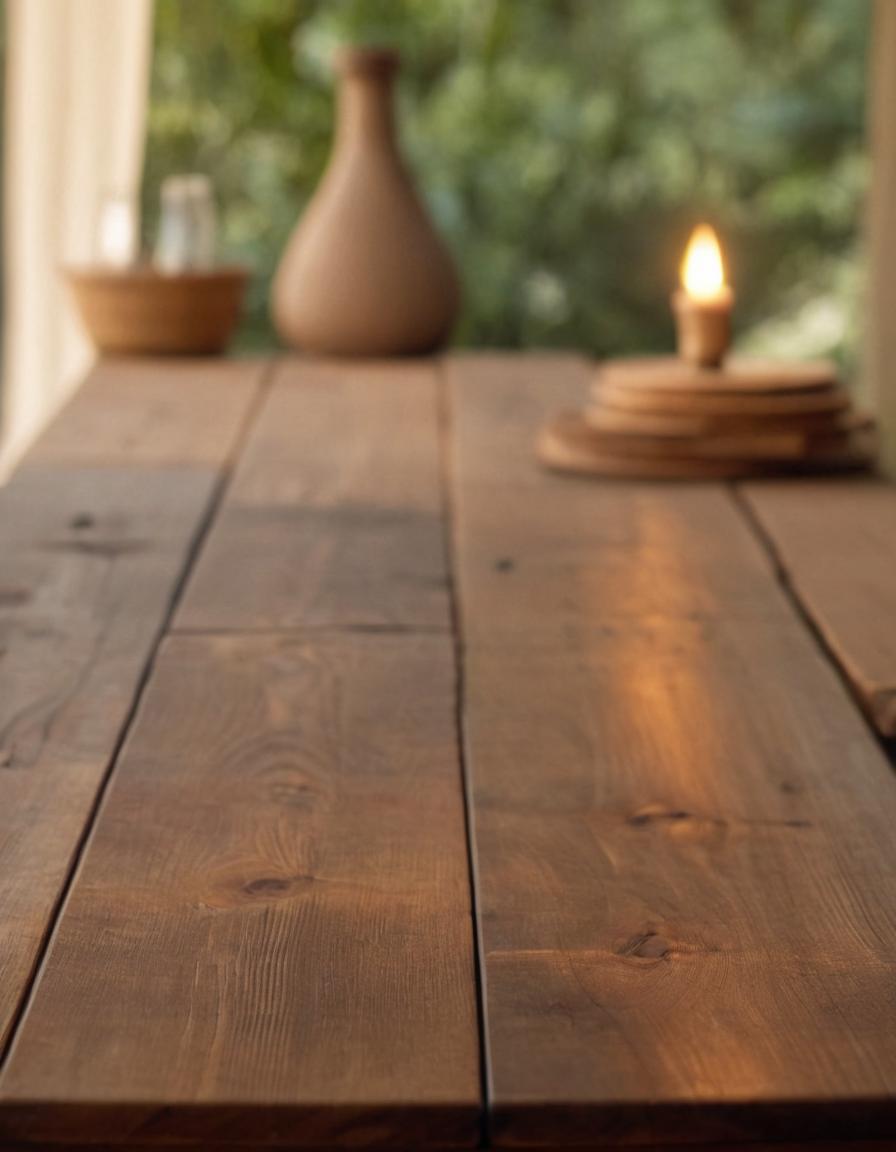}\hfill
\includegraphics[width=0.2\linewidth]{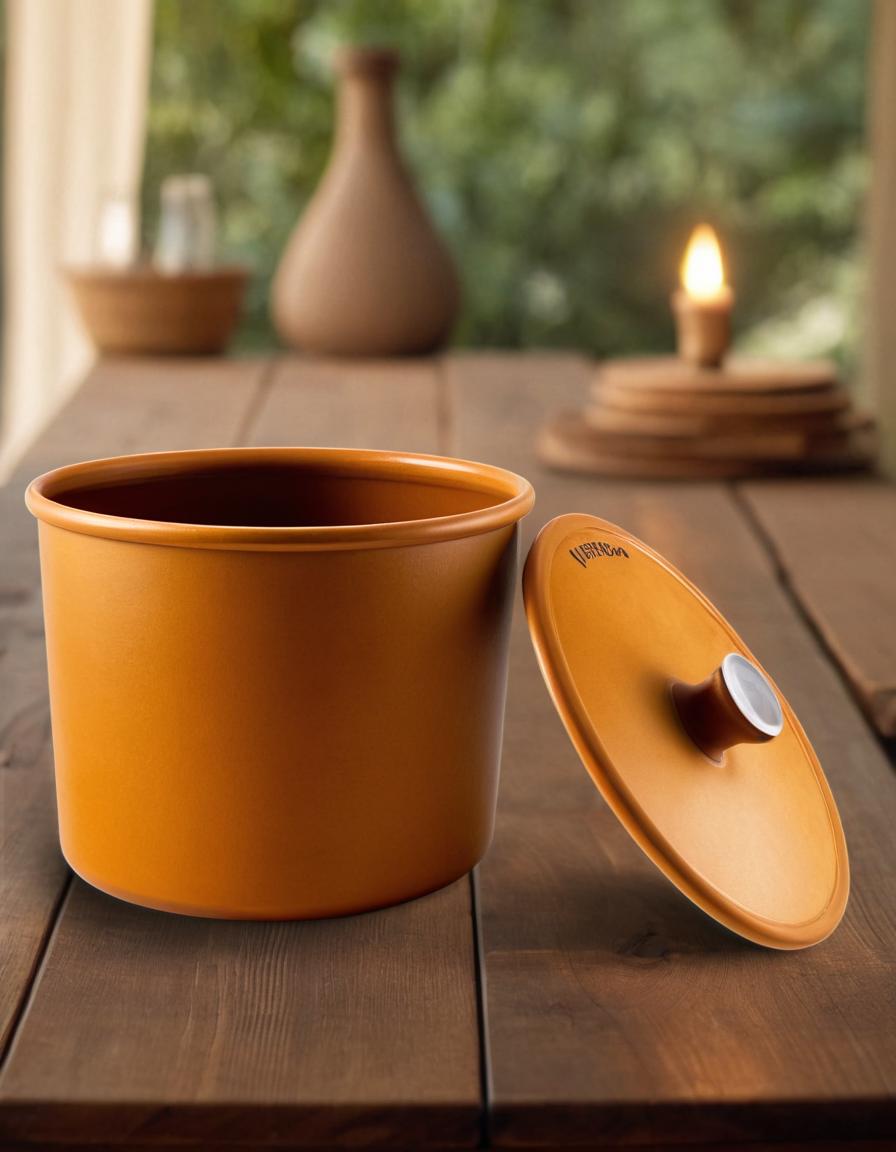}\hfill
\includegraphics[width=0.2\linewidth]{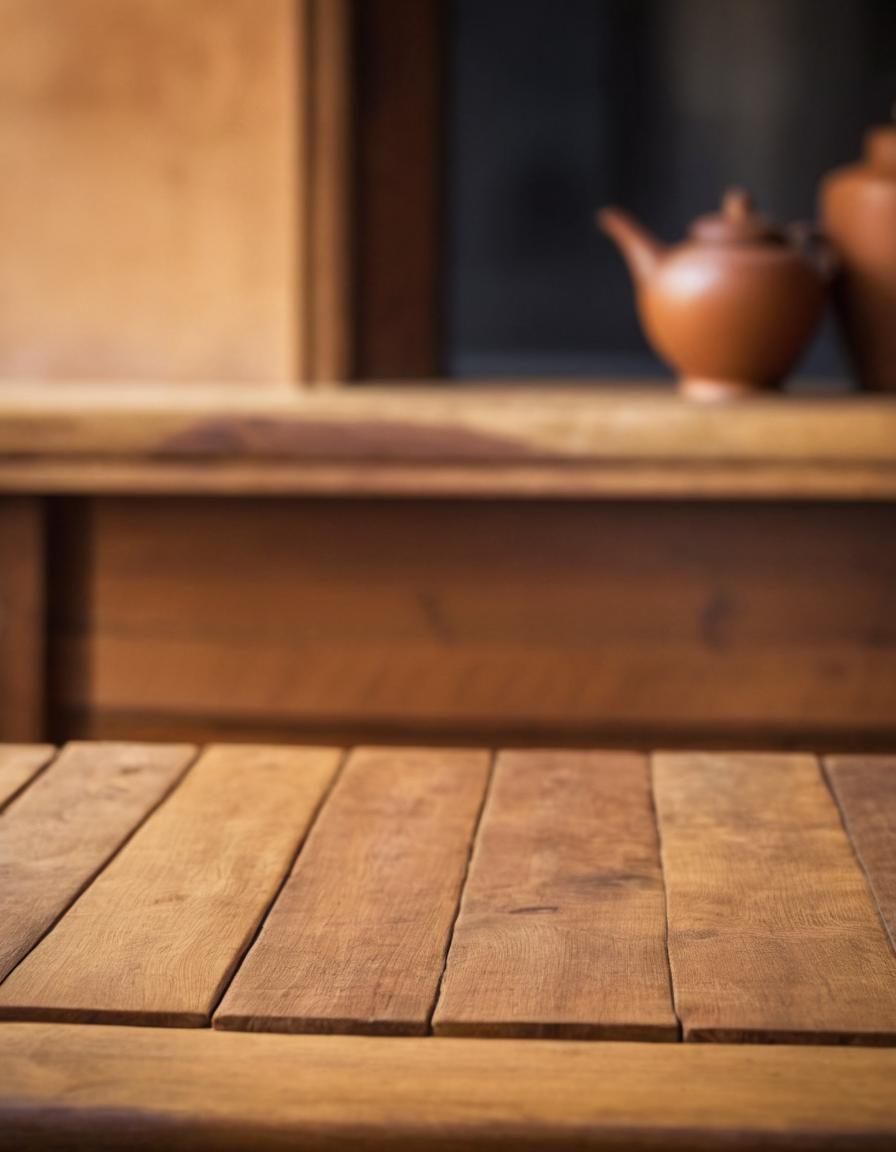}\hfill
\includegraphics[width=0.2\linewidth]{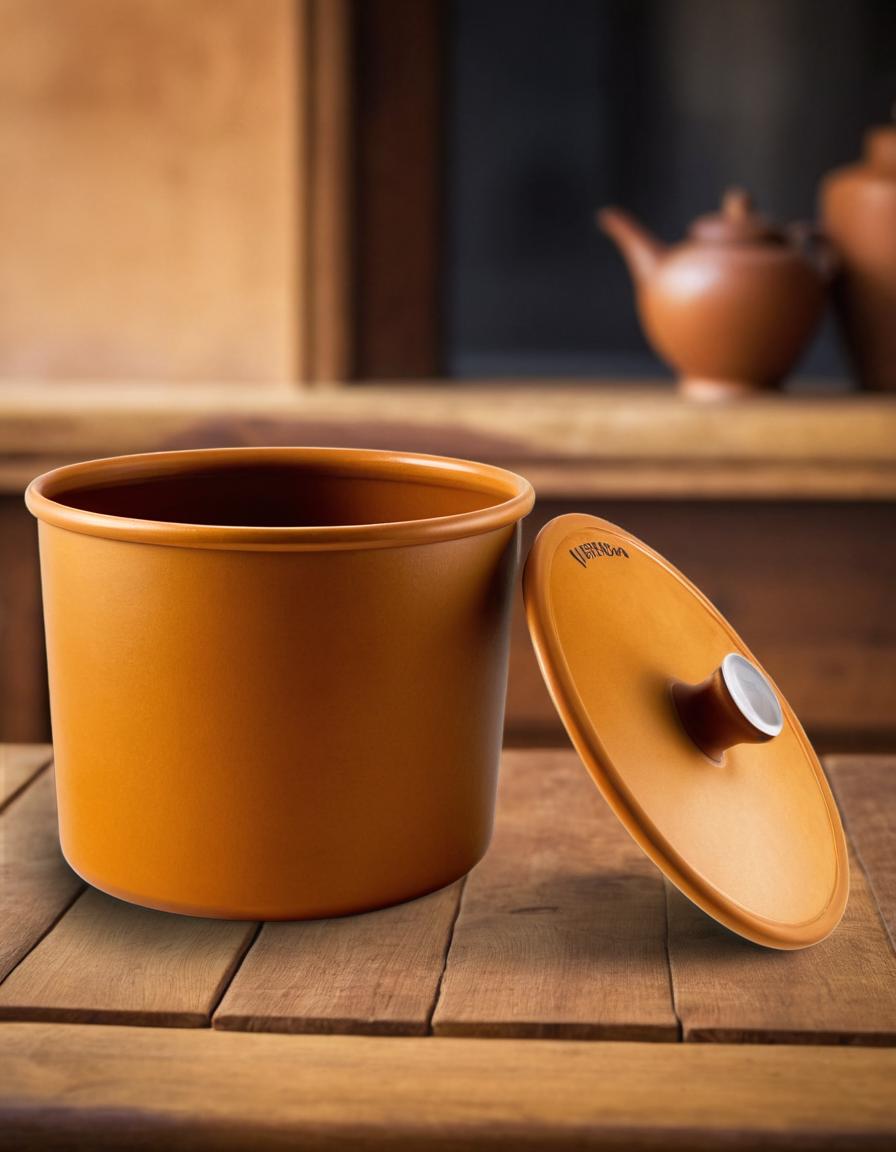}

\vspace{1pt}
\includegraphics[width=0.2\linewidth]{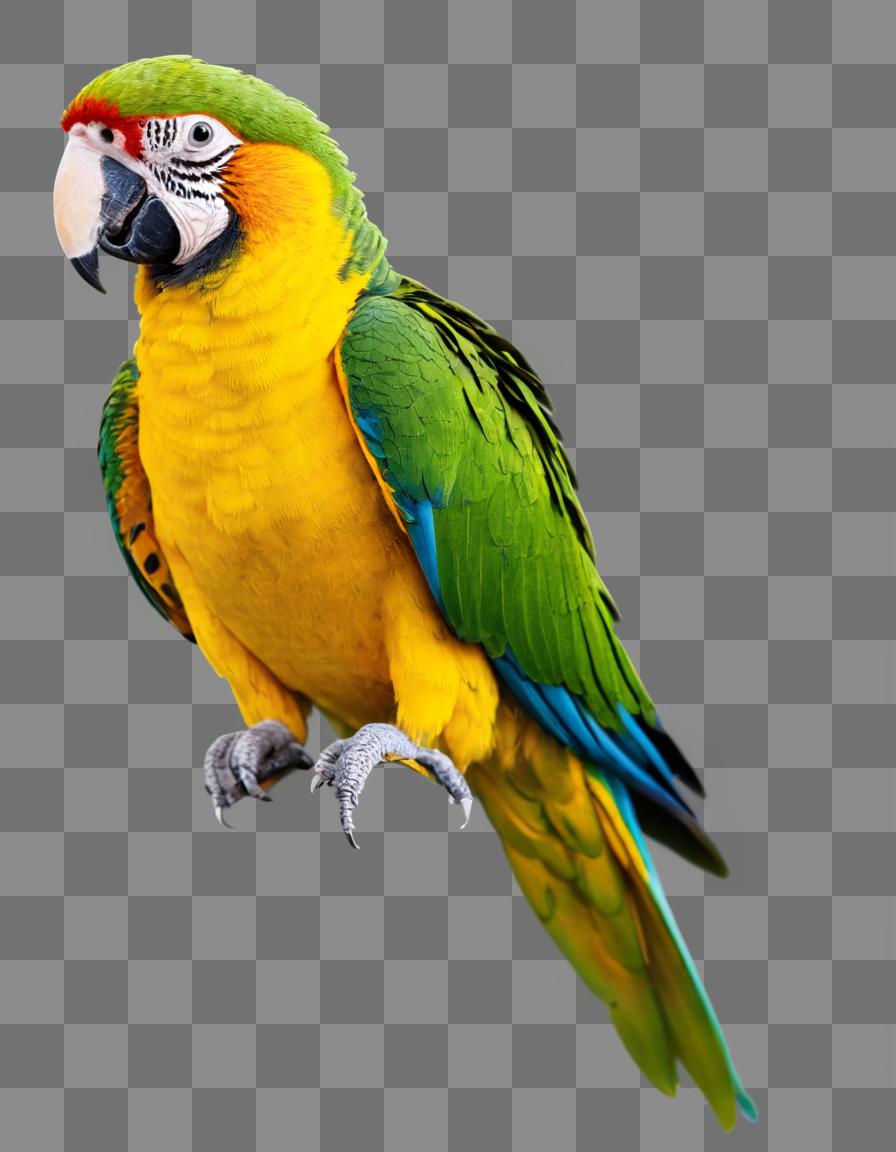}\hfill
\includegraphics[width=0.2\linewidth]{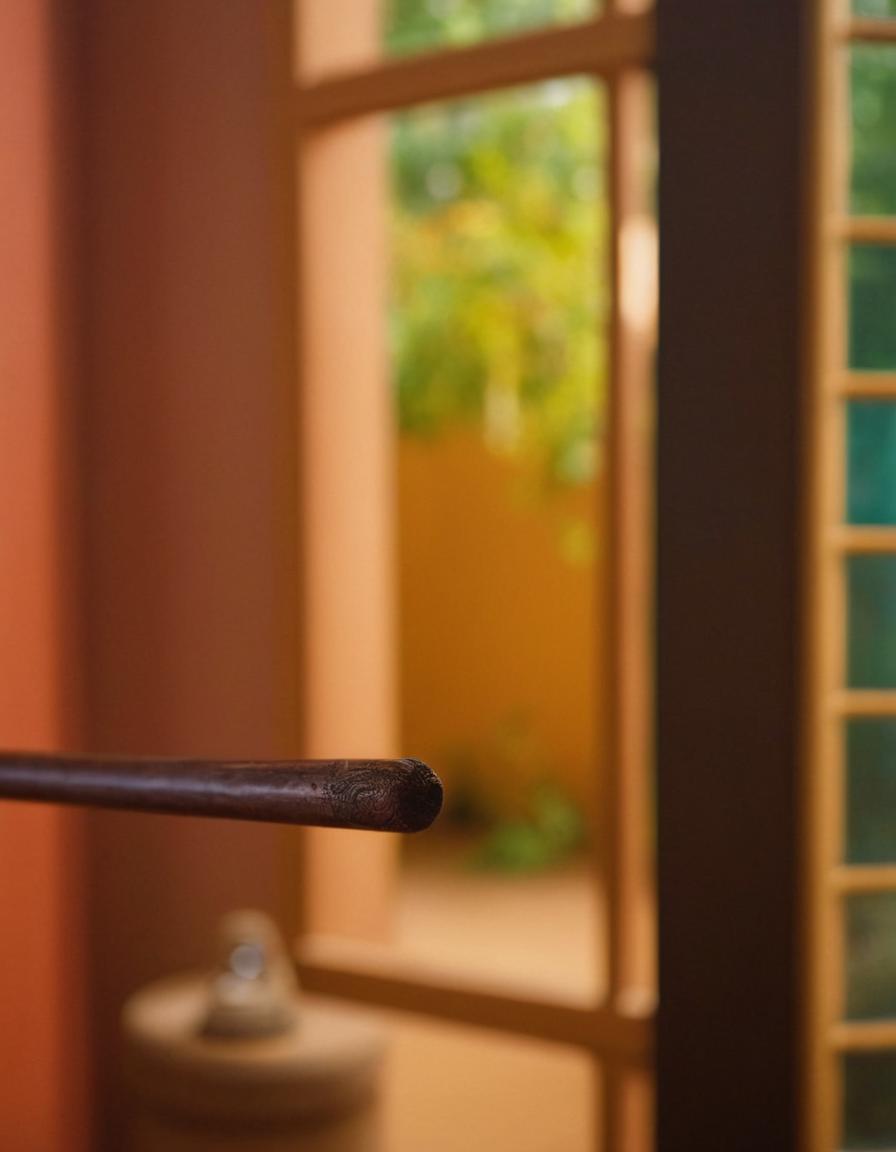}\hfill
\includegraphics[width=0.2\linewidth]{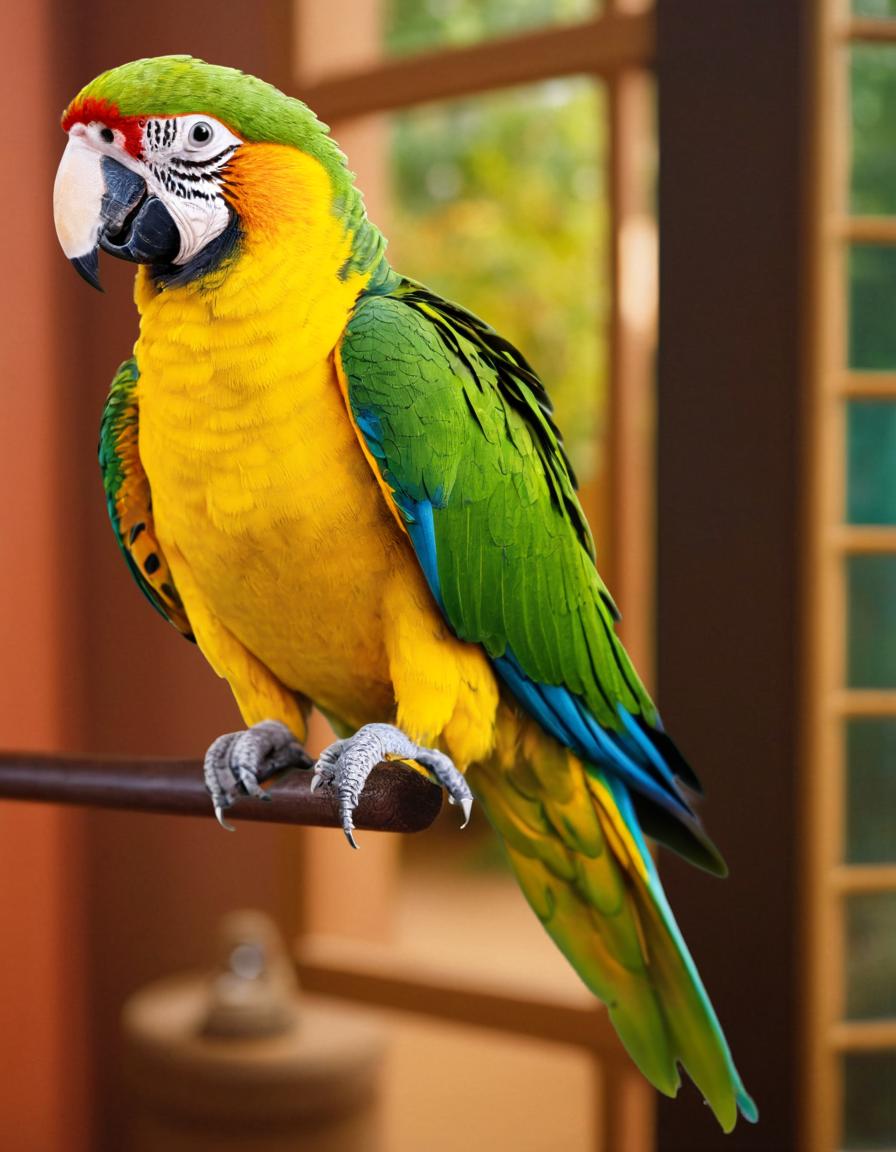}\hfill
\includegraphics[width=0.2\linewidth]{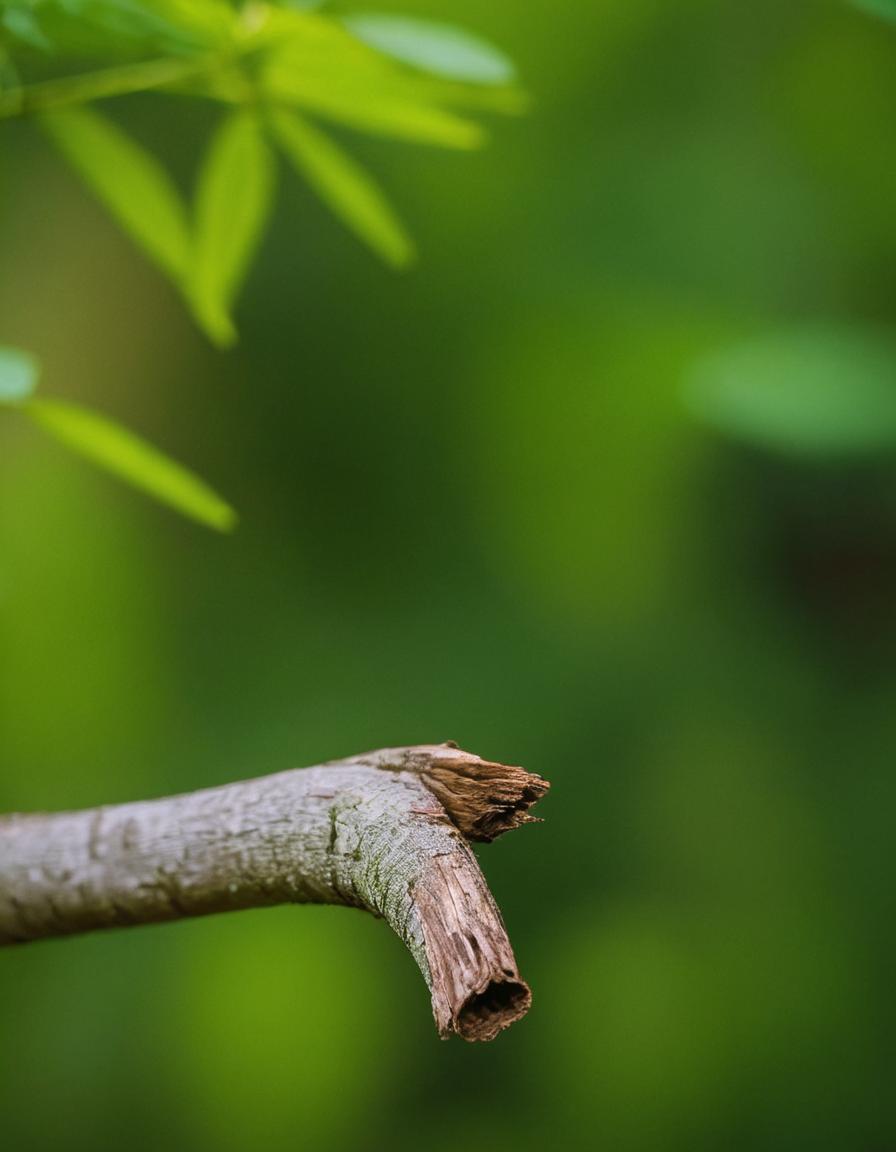}\hfill
\includegraphics[width=0.2\linewidth]{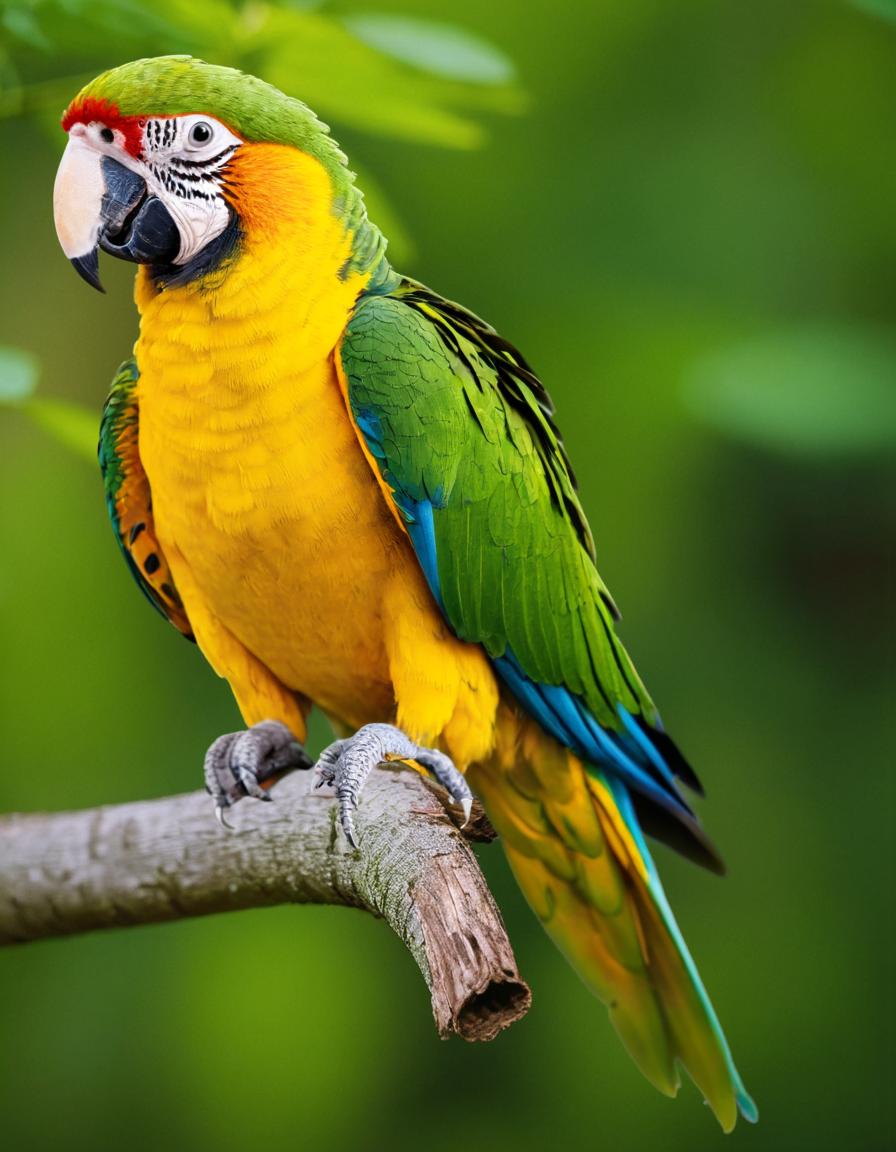}
\caption{Foreground-conditioned Background Results \#1. The left-most images are inputs. The prompts are ``man sitting on chair'', ``man sitting in forest'', ``pots on wood table'', ``parrot in room'', ``parrot in forest''. Resolution is $896\times1152$.}
\label{fig:f1}
\end{minipage}
\end{figure*}

\begin{figure*}

\begin{minipage}{\linewidth}
\includegraphics[width=0.2\linewidth]{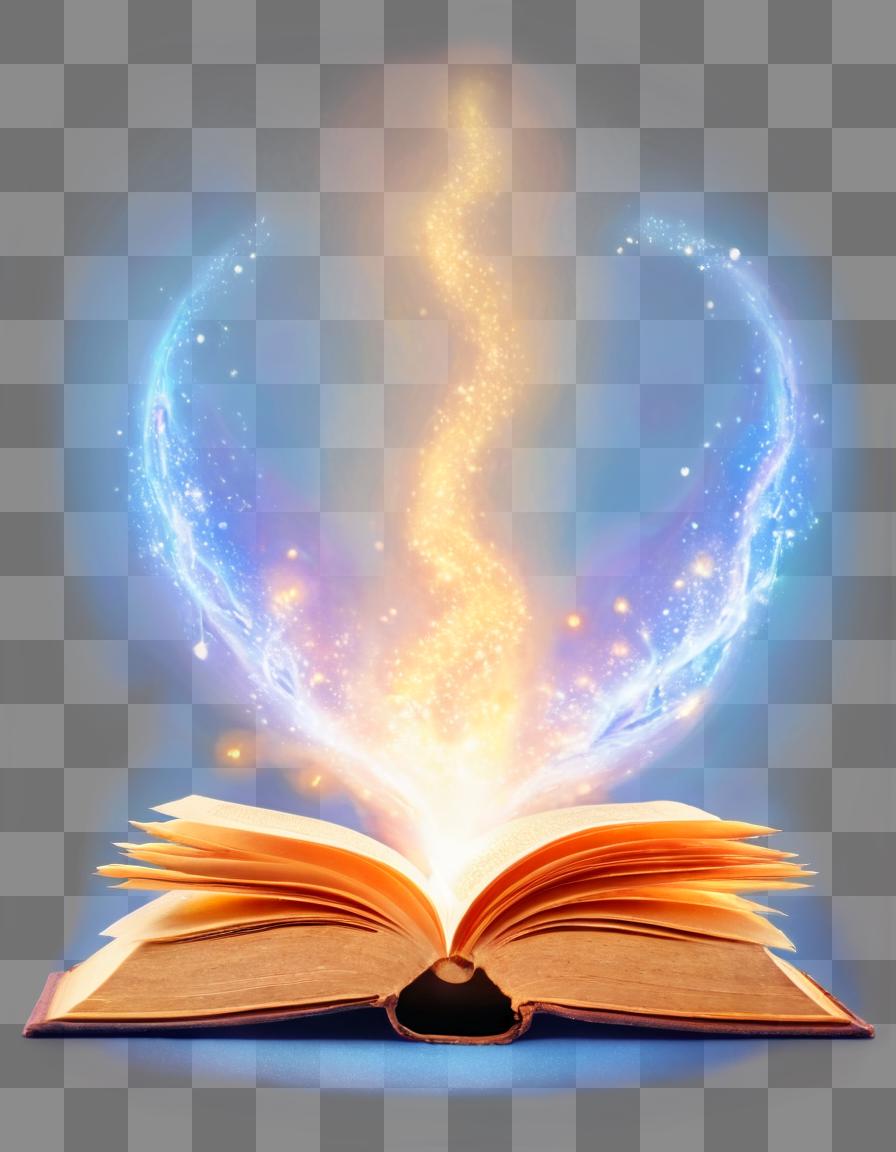}\hfill
\includegraphics[width=0.2\linewidth]{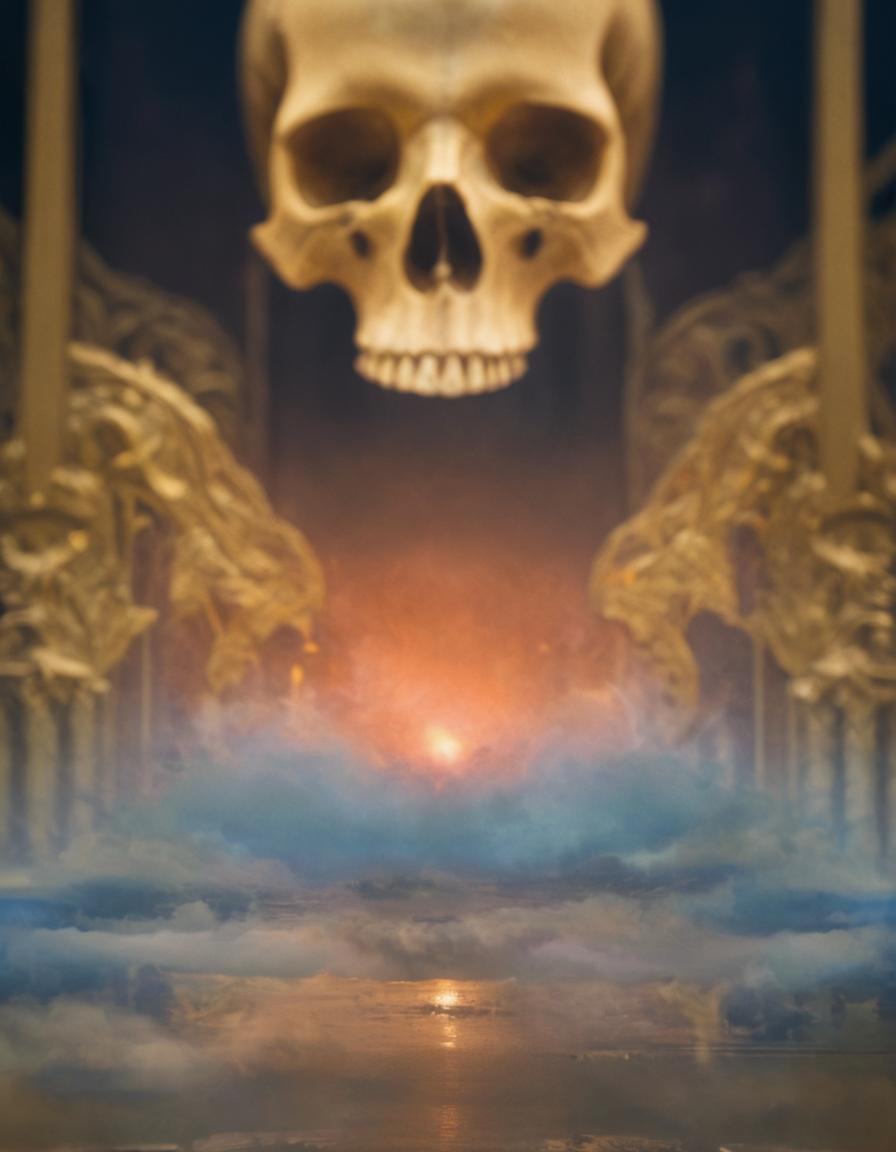}\hfill
\includegraphics[width=0.2\linewidth]{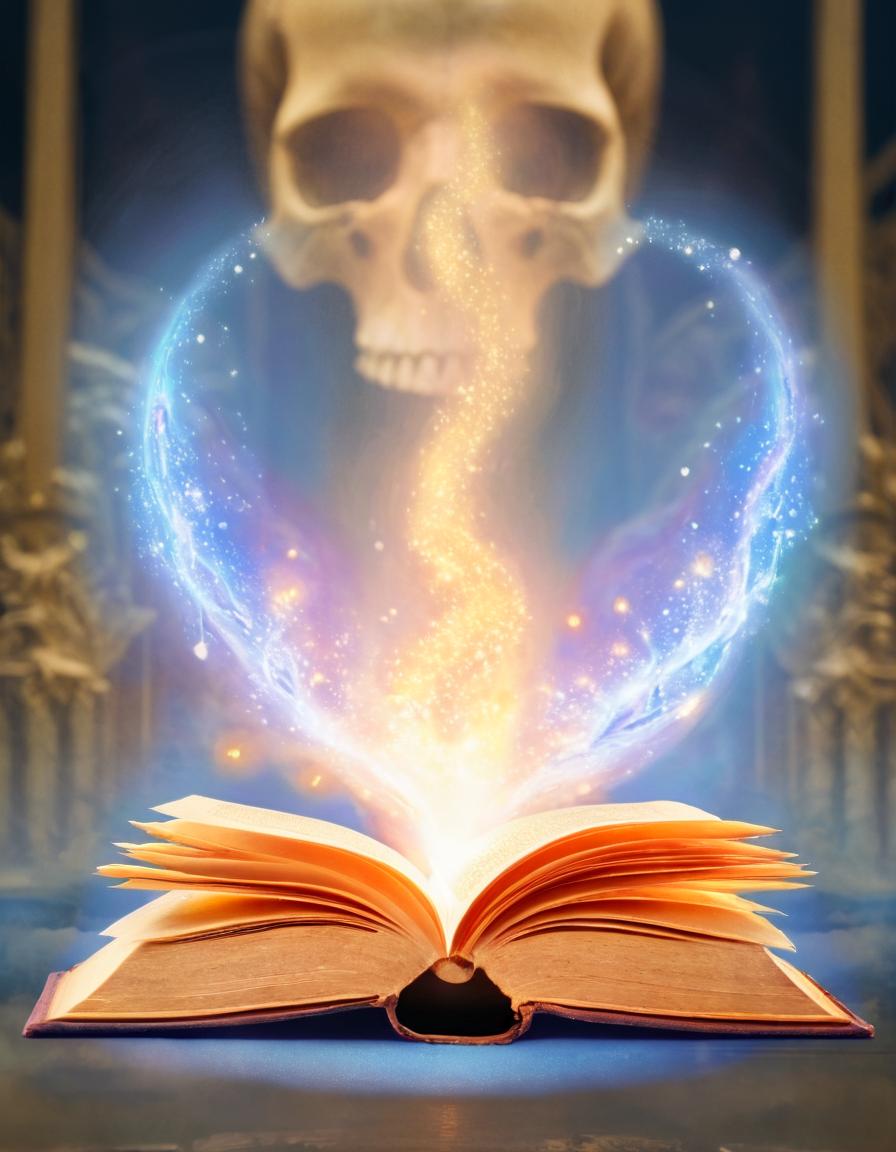}\hfill
\includegraphics[width=0.2\linewidth]{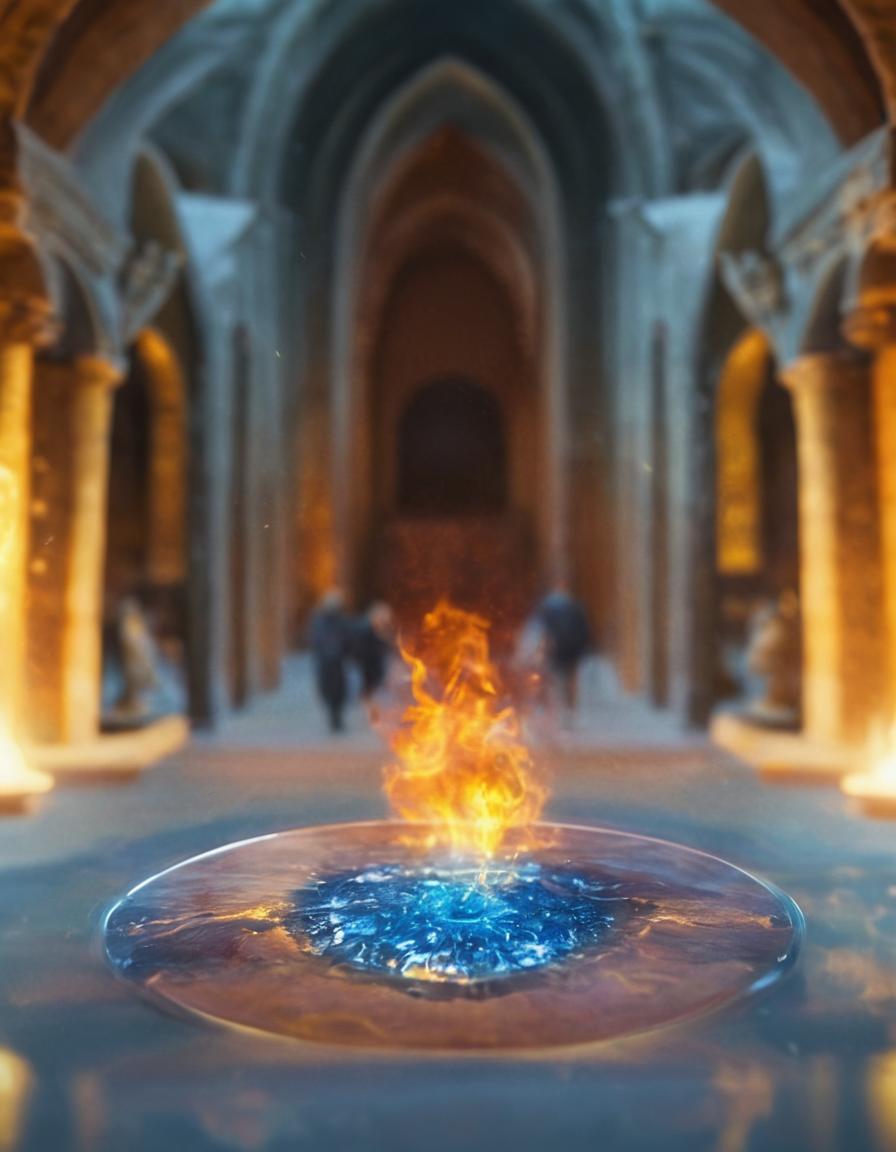}\hfill
\includegraphics[width=0.2\linewidth]{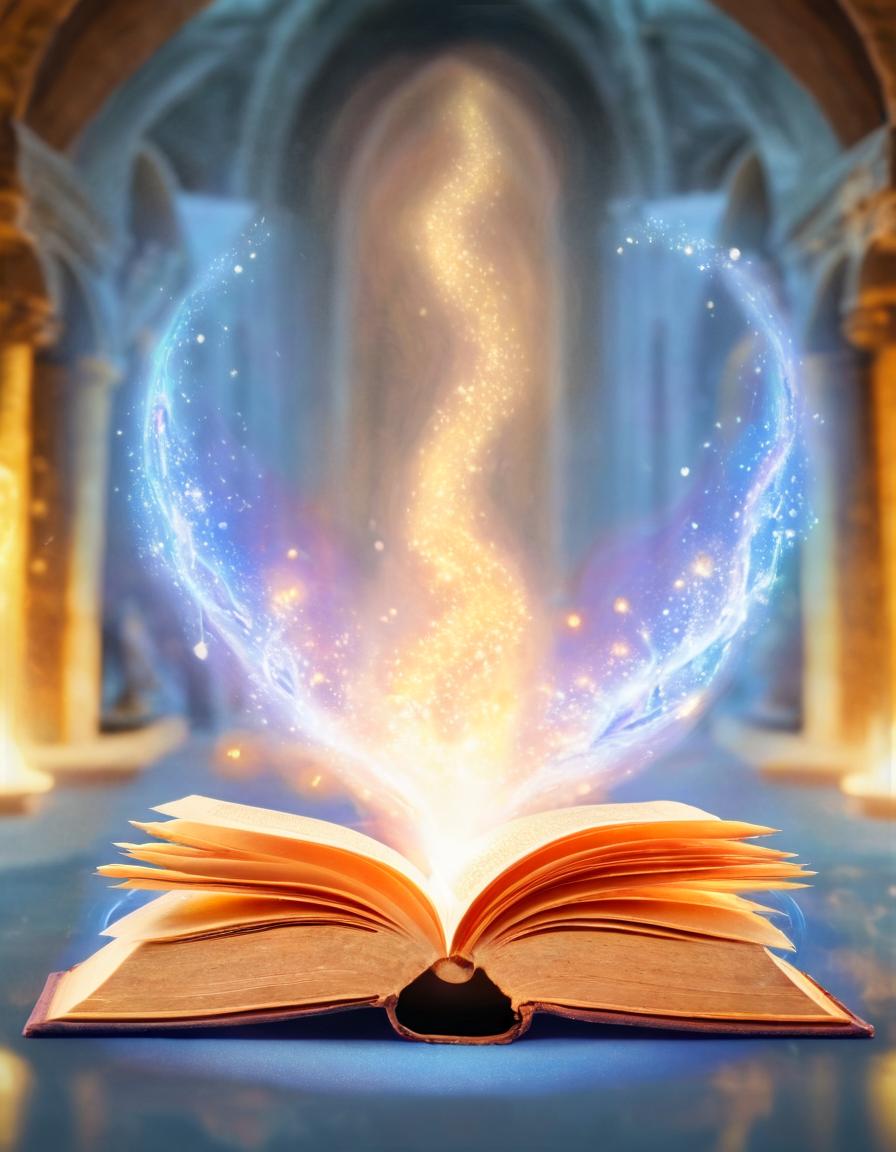}

\vspace{1pt}
\includegraphics[width=0.2\linewidth]{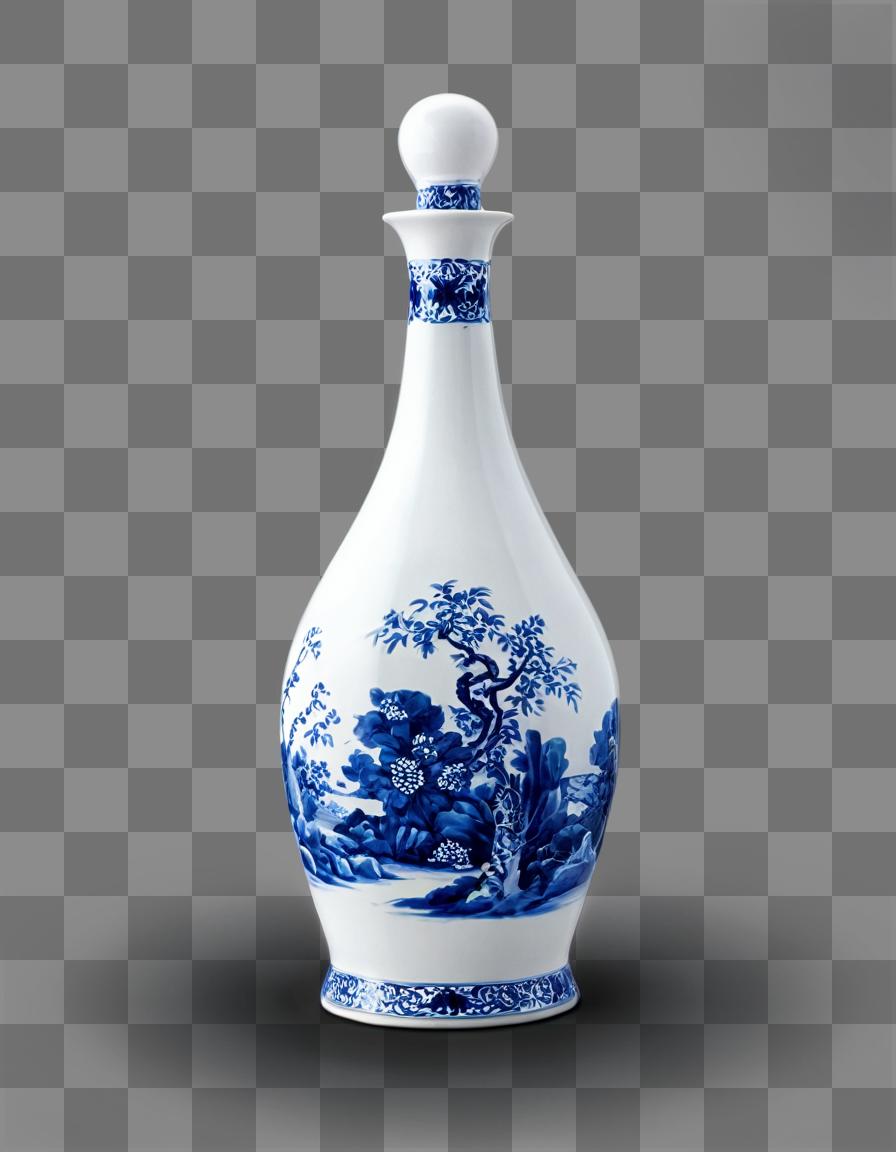}\hfill
\includegraphics[width=0.2\linewidth]{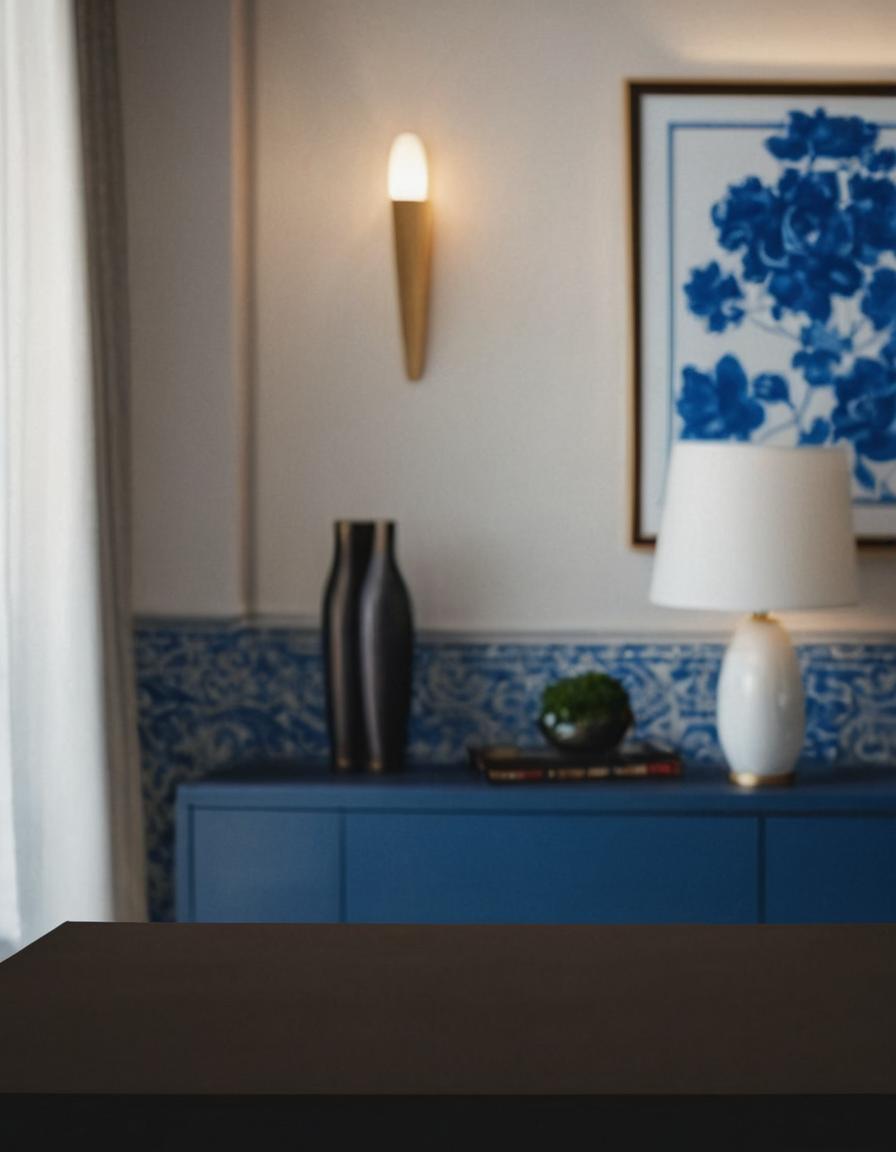}\hfill
\includegraphics[width=0.2\linewidth]{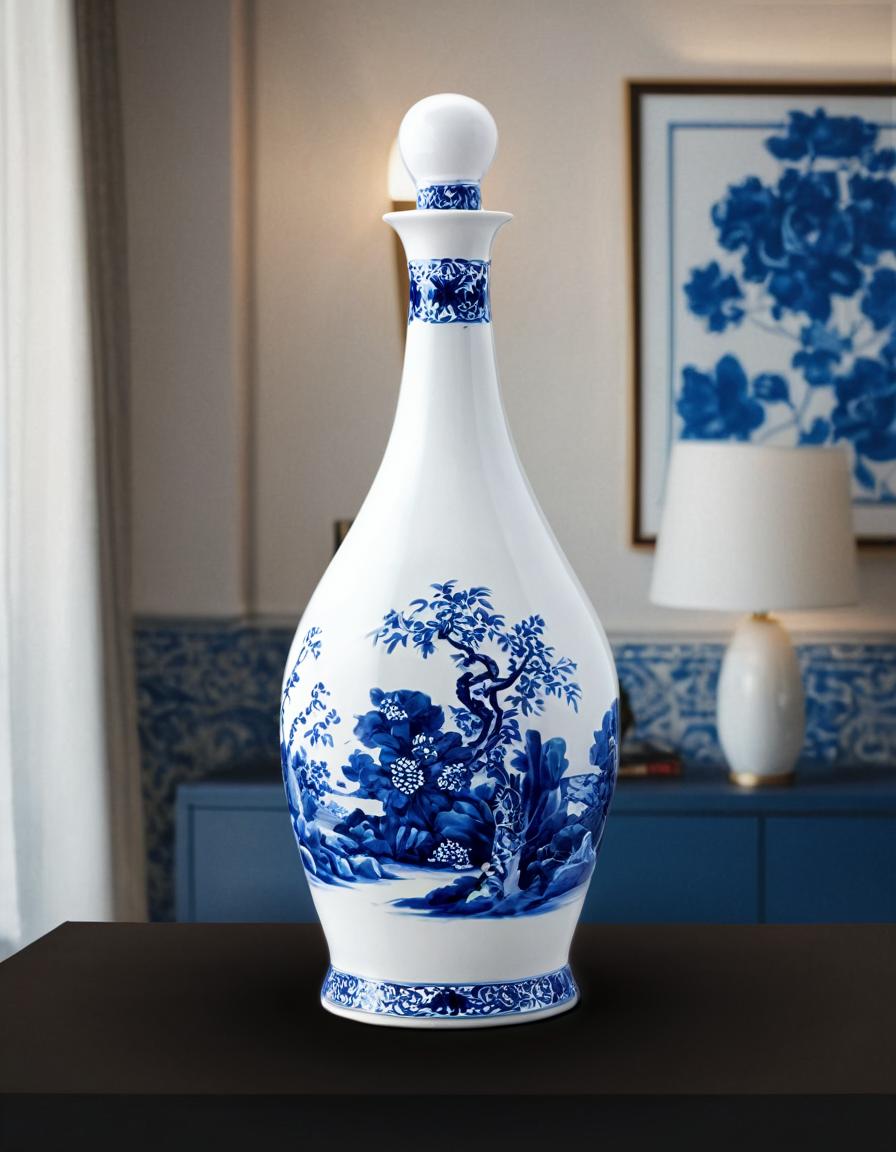}\hfill
\includegraphics[width=0.2\linewidth]{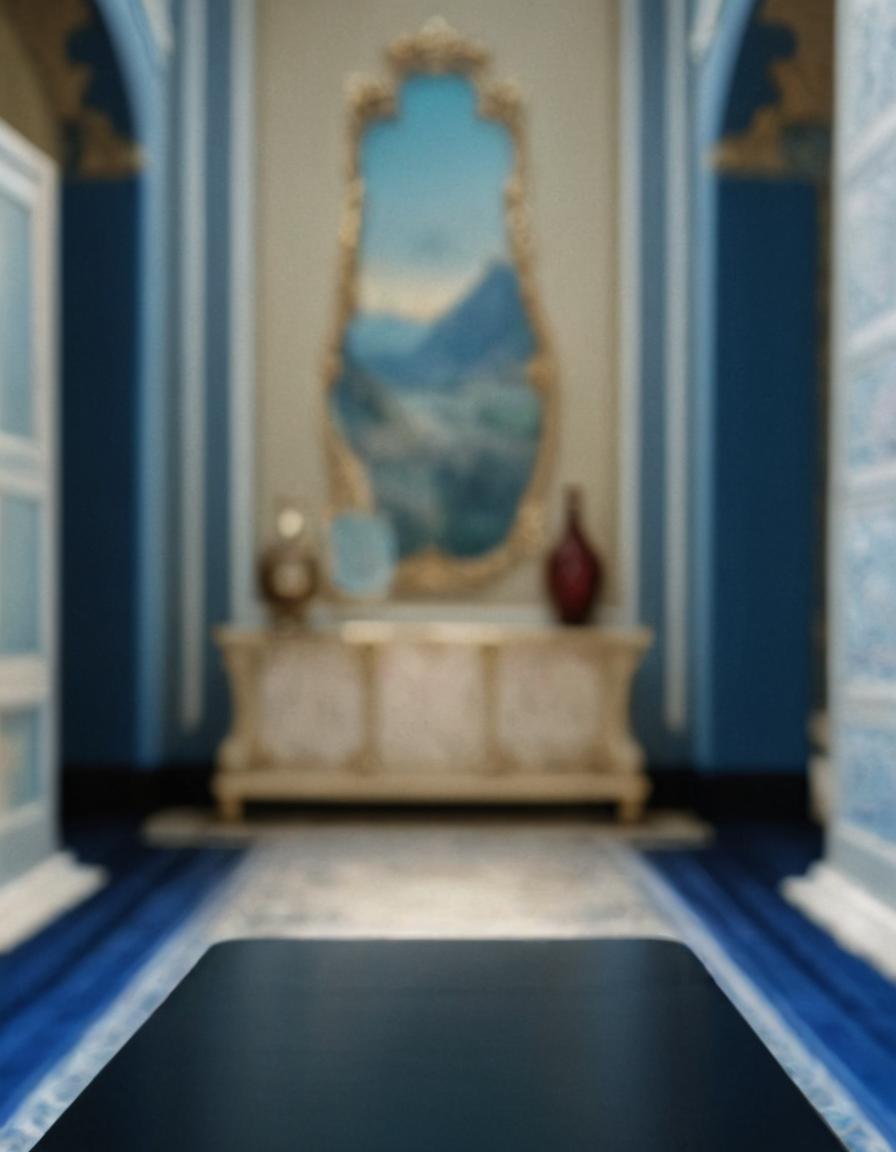}\hfill
\includegraphics[width=0.2\linewidth]{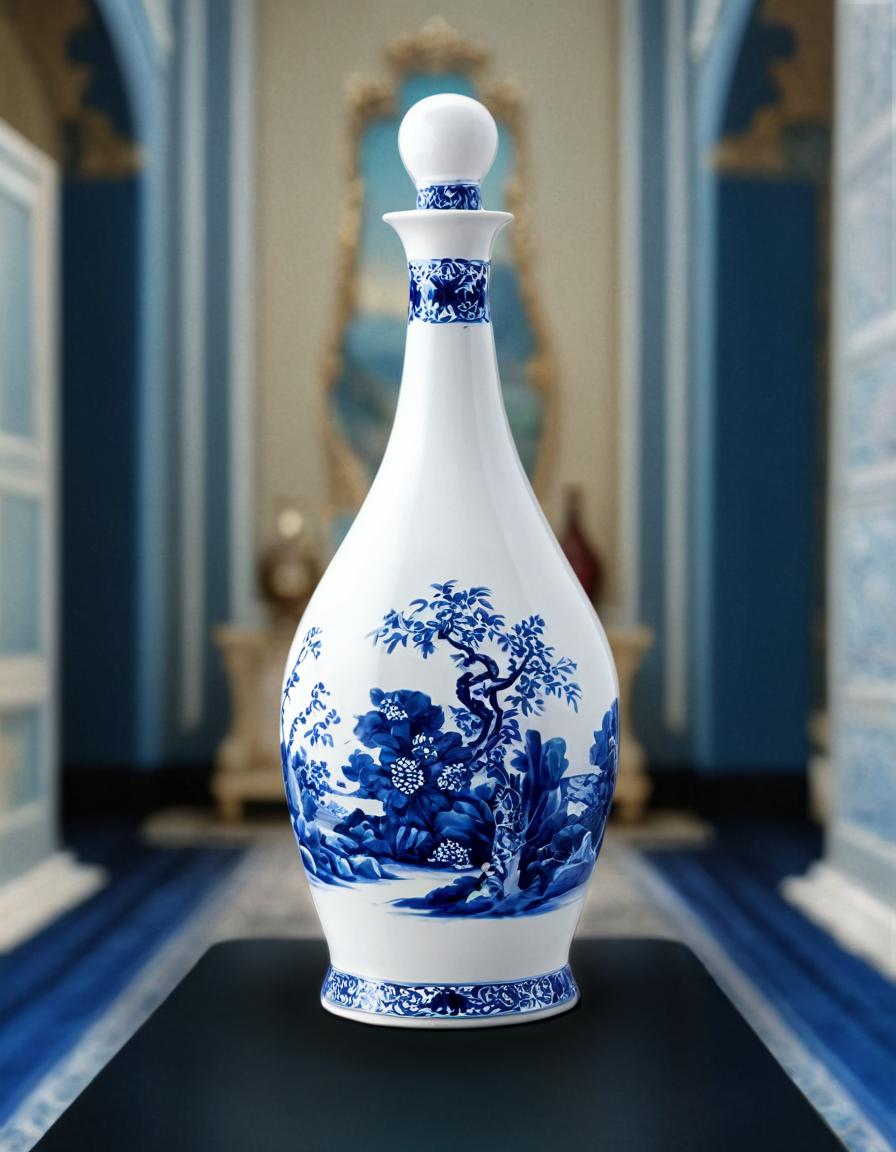}

\vspace{1pt}
\includegraphics[width=0.2\linewidth]{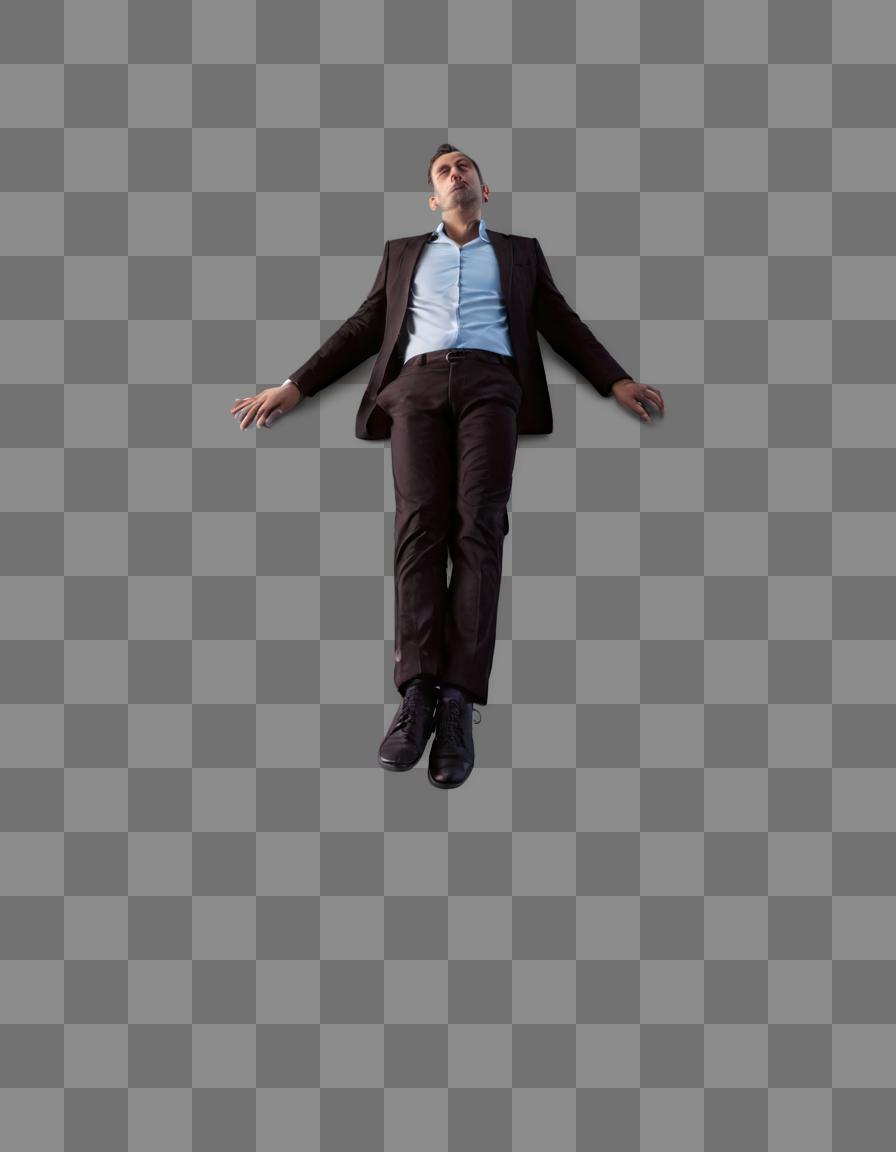}\hfill
\includegraphics[width=0.2\linewidth]{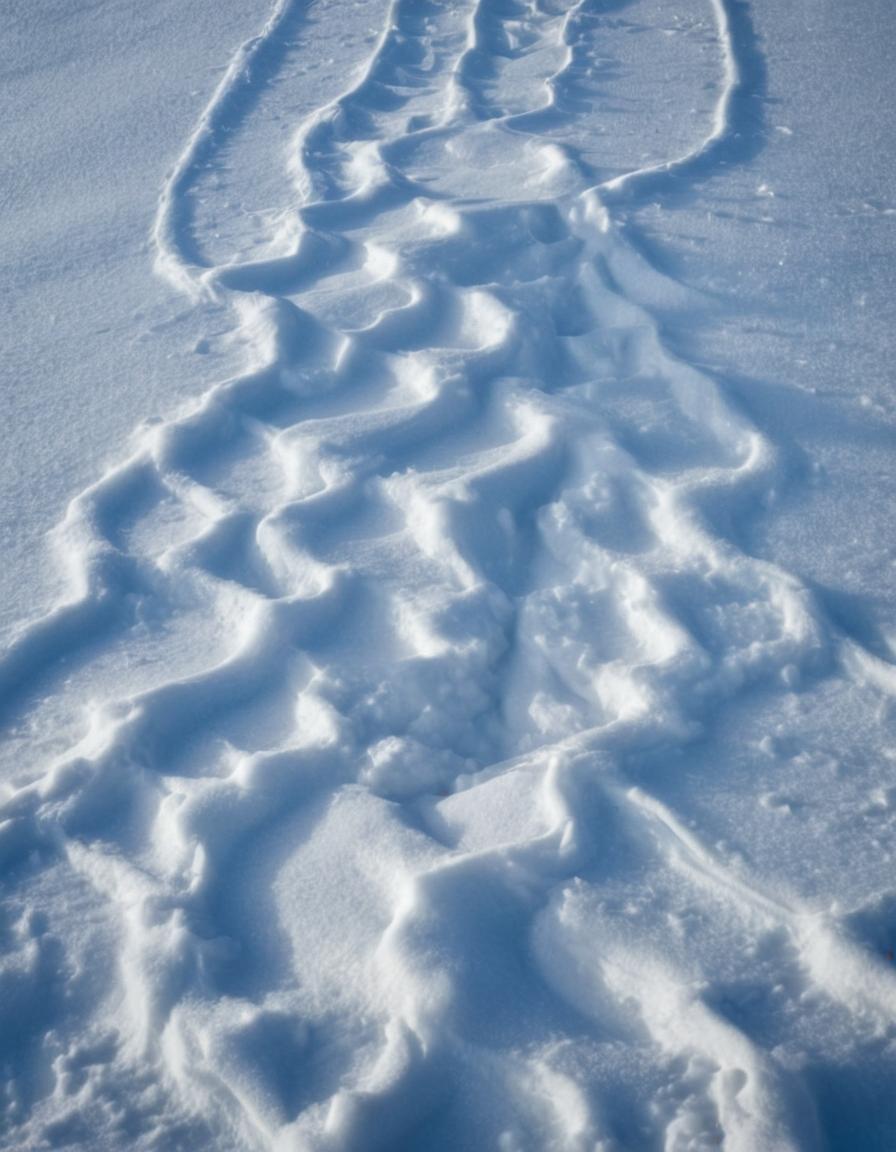}\hfill
\includegraphics[width=0.2\linewidth]{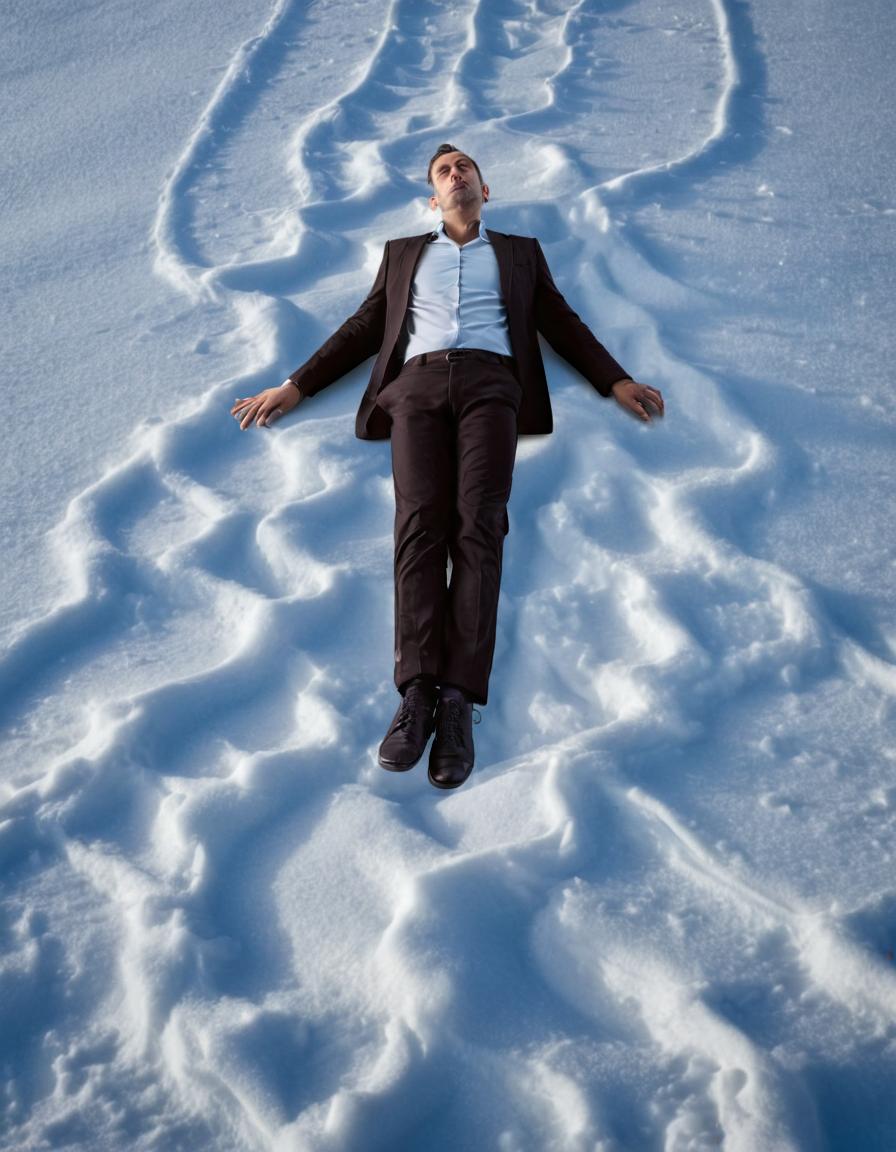}\hfill
\includegraphics[width=0.2\linewidth]{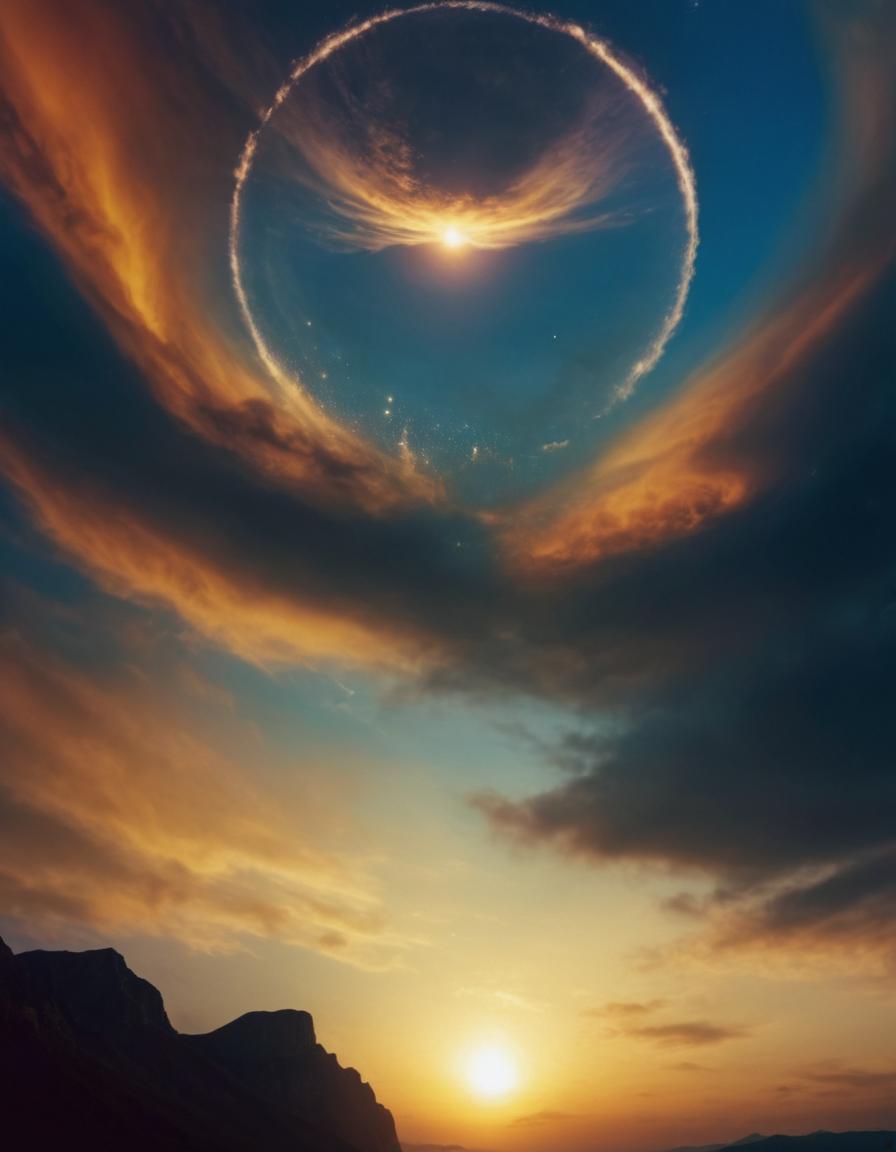}\hfill
\includegraphics[width=0.2\linewidth]{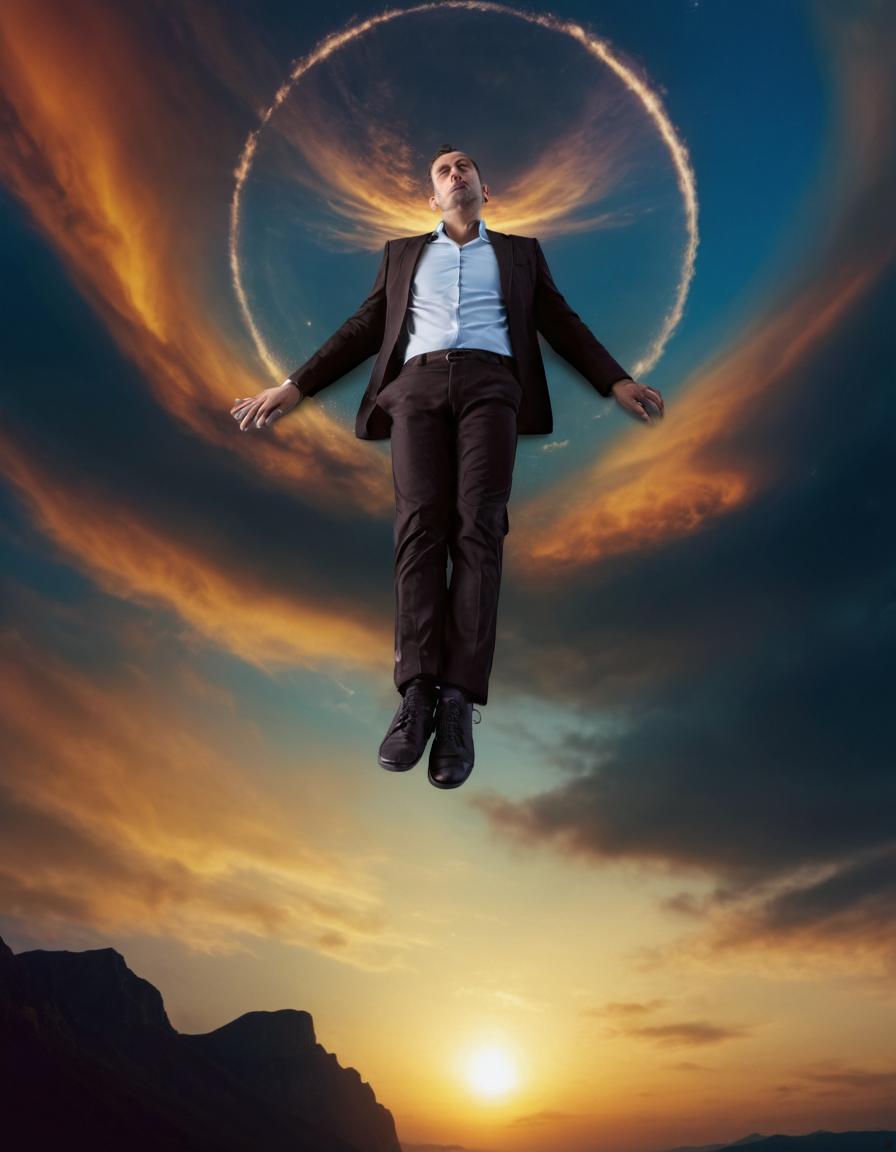}
\caption{Foreground-conditioned Background Results \#2. The left-most images are inputs. The prompts are ``magic book of death'', ``magic book of life'', ``blue and white porcelain vase in my home'', ``blue and white porcelain vase in the museum'', ``the man in the snow'', ``god of infinity''. Resolution is $896\times1152$.}
\label{fig:f2}
\end{minipage}
\end{figure*}

We present additional results for foreground-conditioned backgrounds, from Figure~\ref{fig:f1} to Figure~\ref{fig:f2}.

\section{Background-Conditioned Foregrounds}
	
\begin{figure*}

\begin{minipage}{\linewidth}
\includegraphics[width=0.2\linewidth]{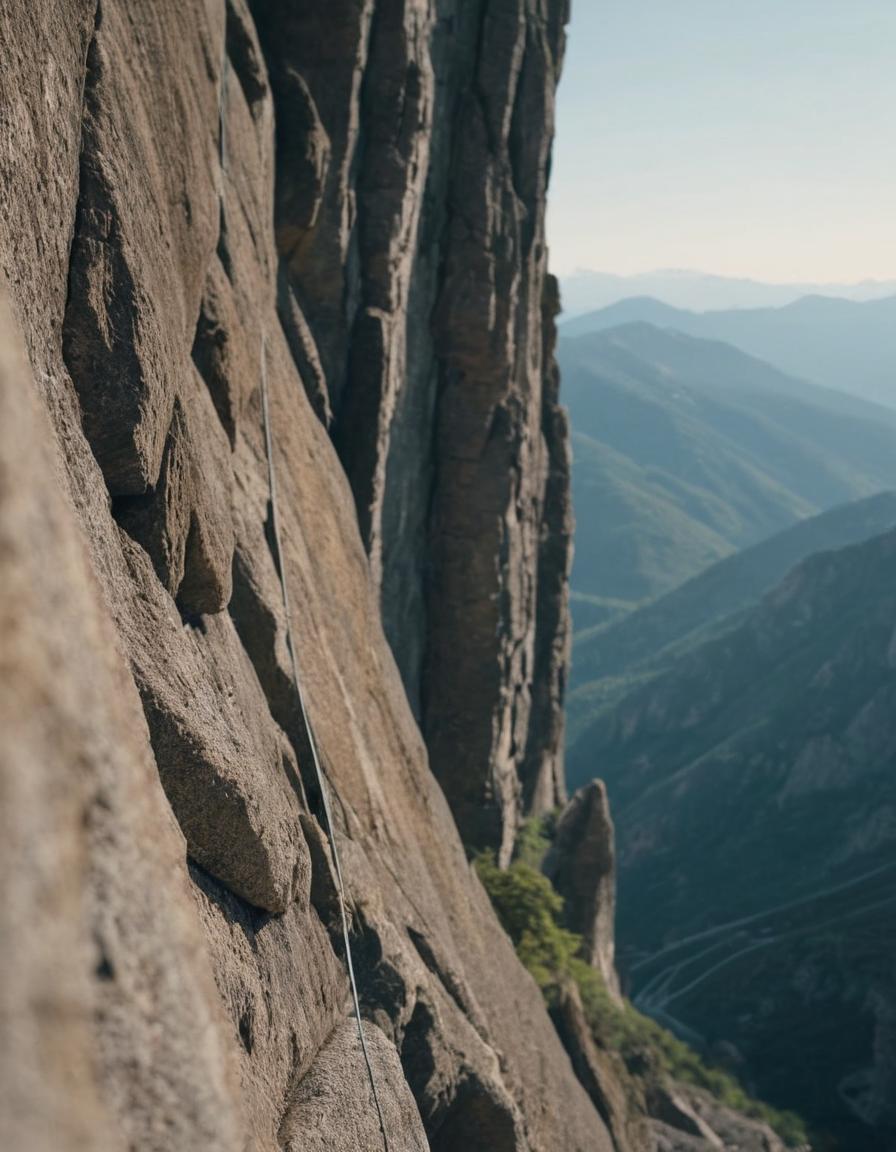}\hfill
\includegraphics[width=0.2\linewidth]{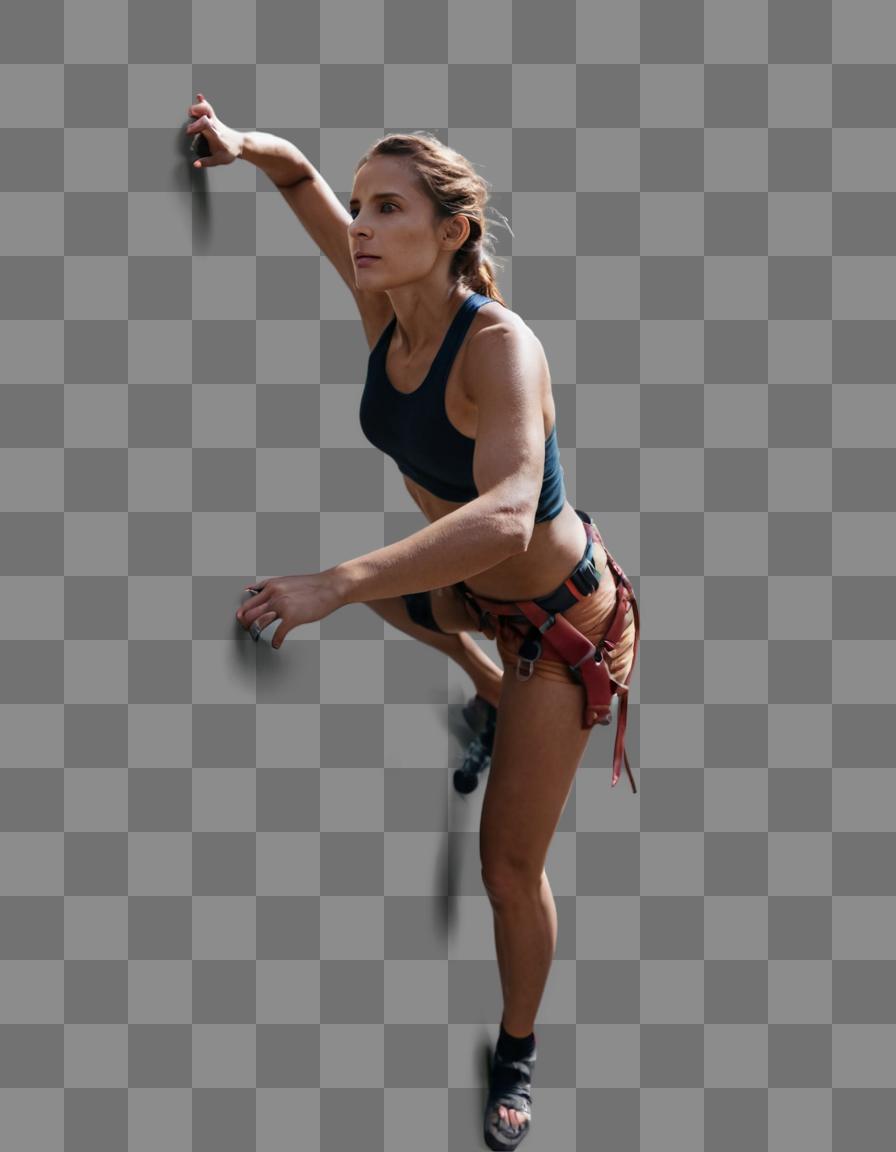}\hfill
\includegraphics[width=0.2\linewidth]{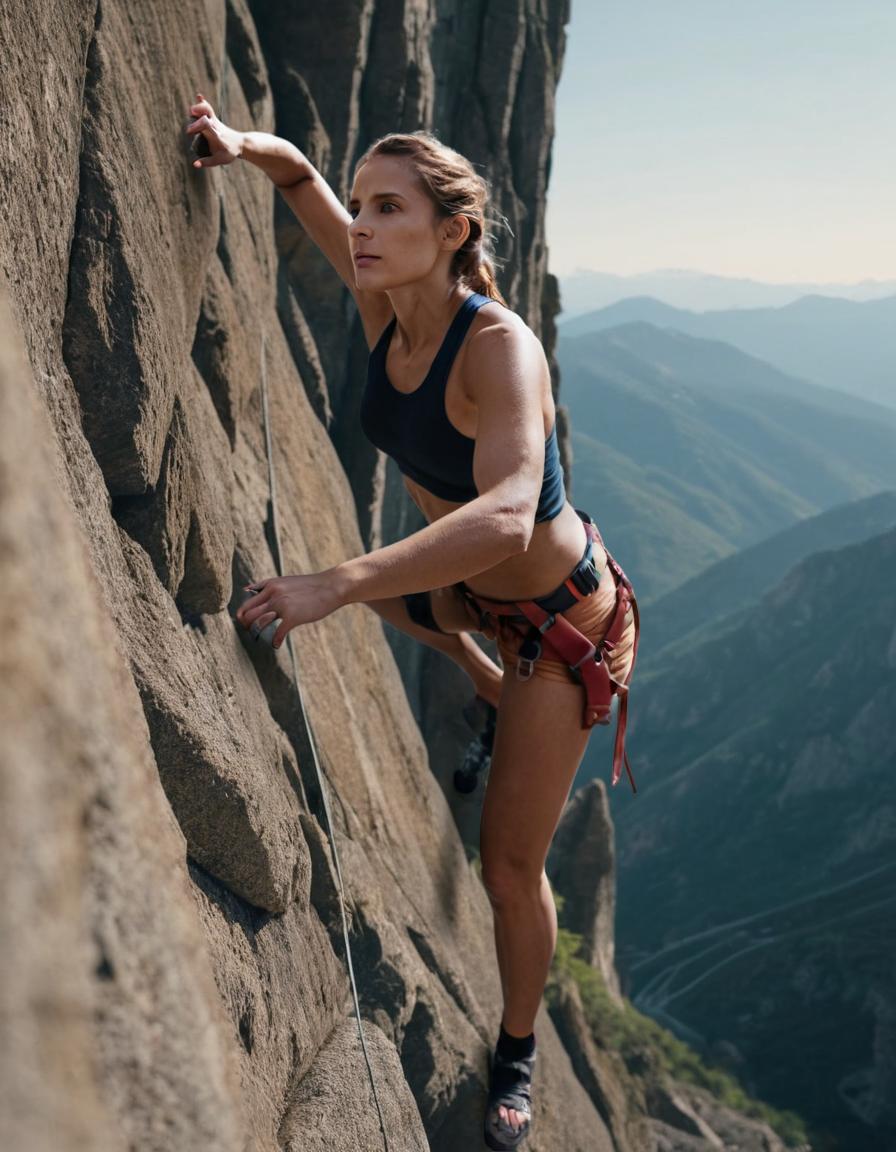}\hfill
\includegraphics[width=0.2\linewidth]{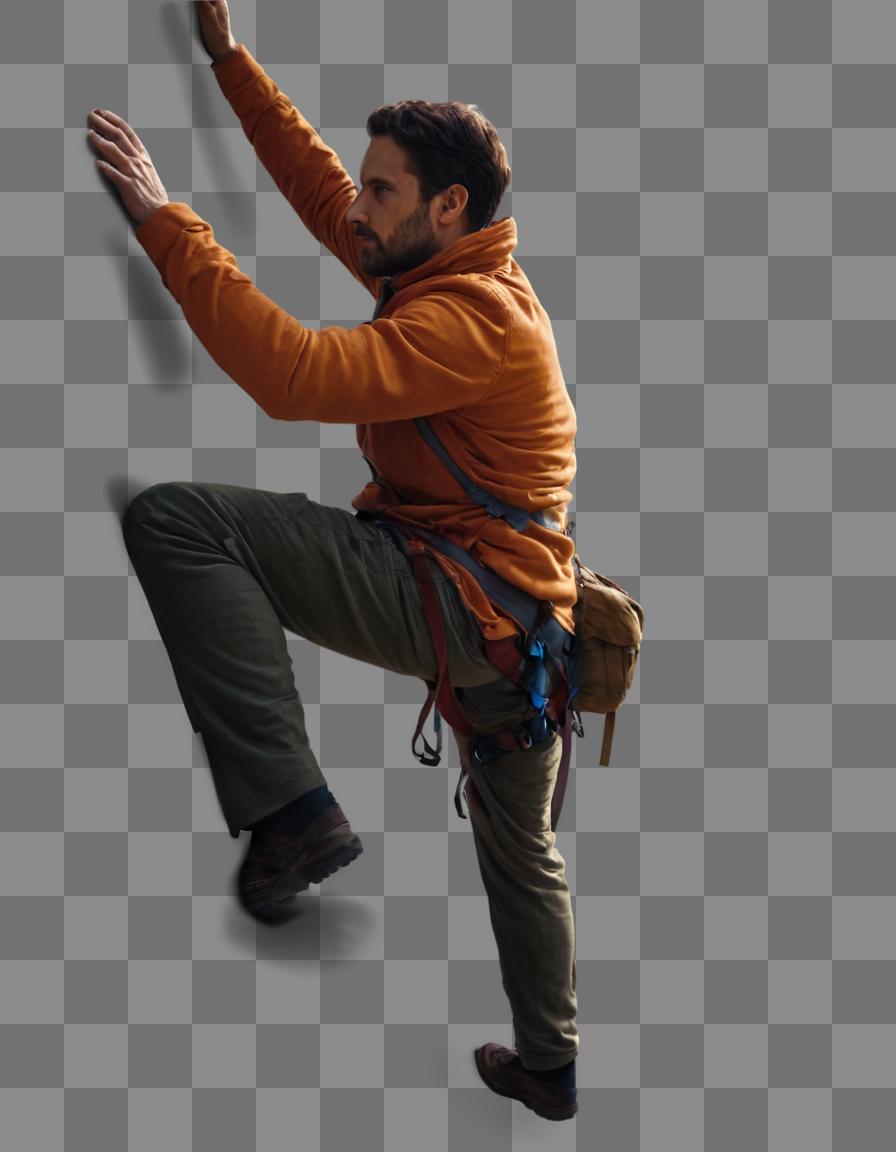}\hfill
\includegraphics[width=0.2\linewidth]{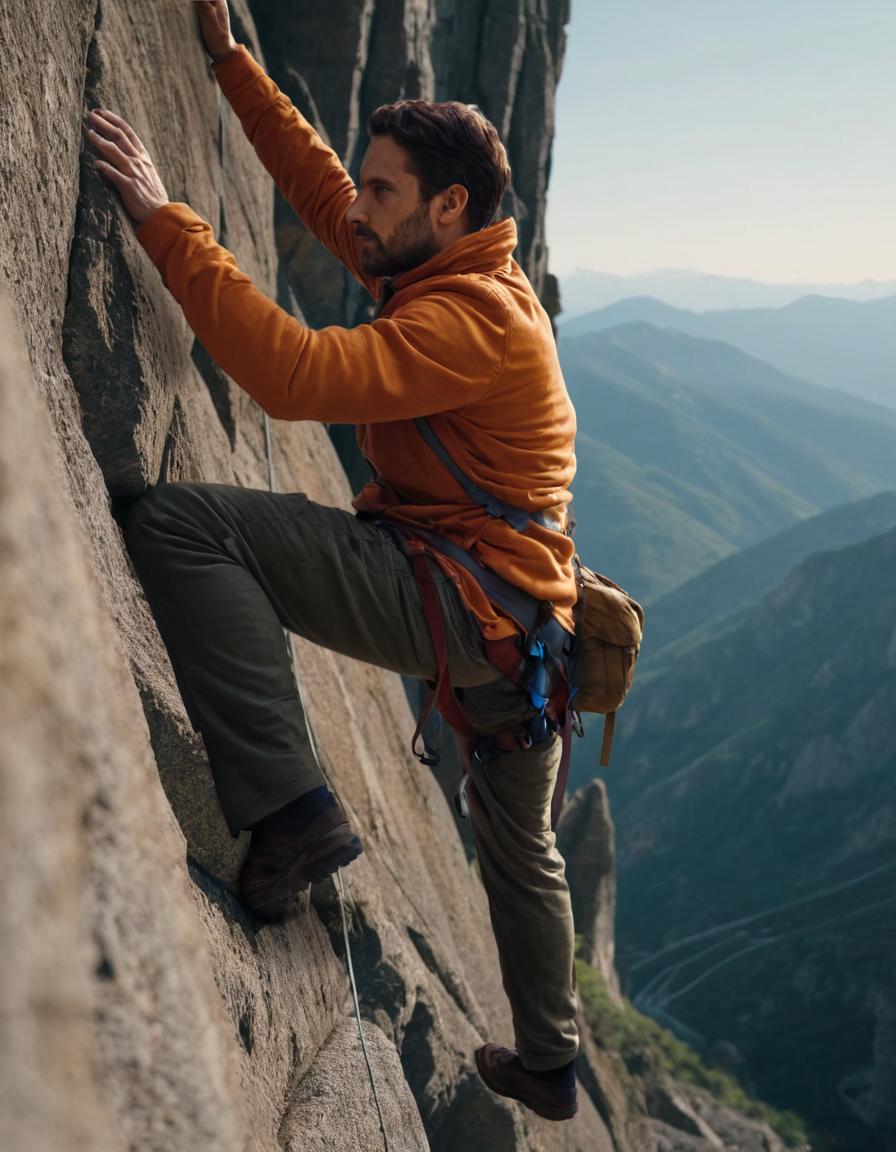}

\vspace{1pt}
\includegraphics[width=0.2\linewidth]{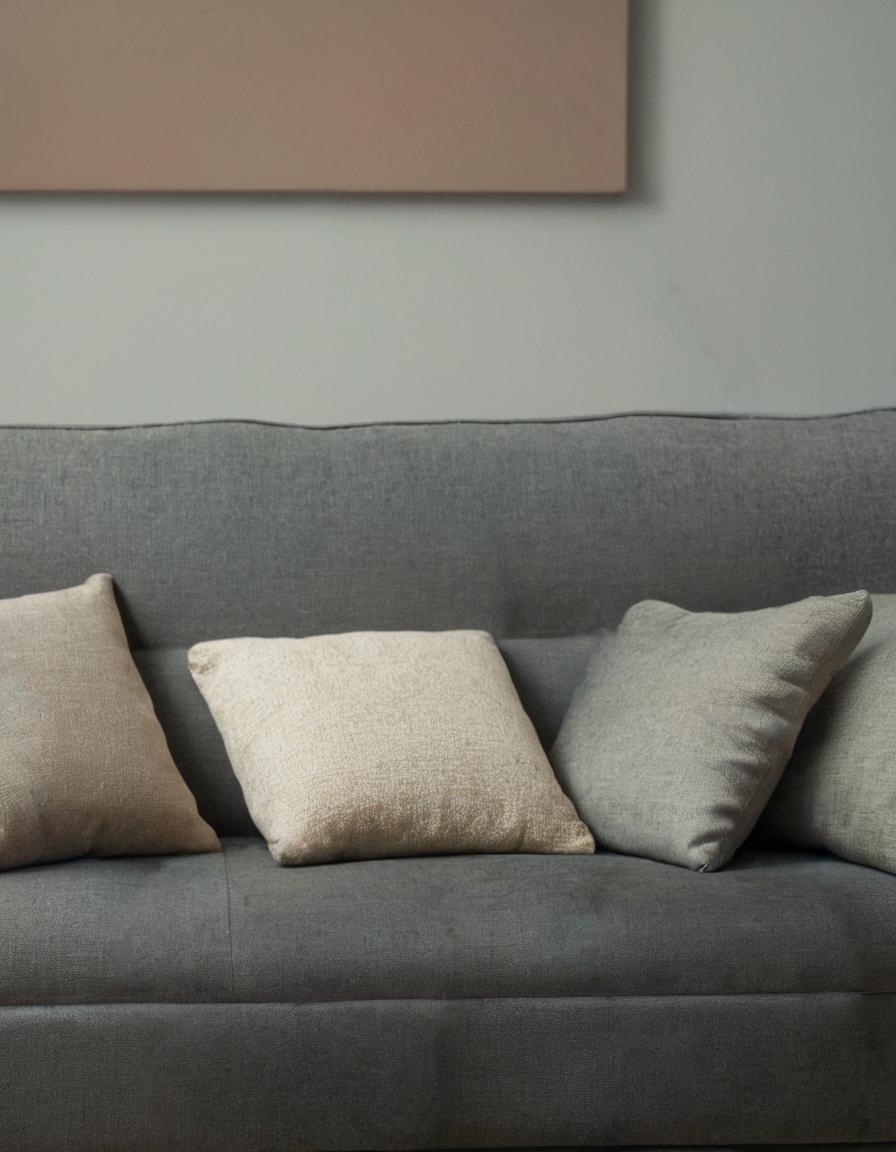}\hfill
\includegraphics[width=0.2\linewidth]{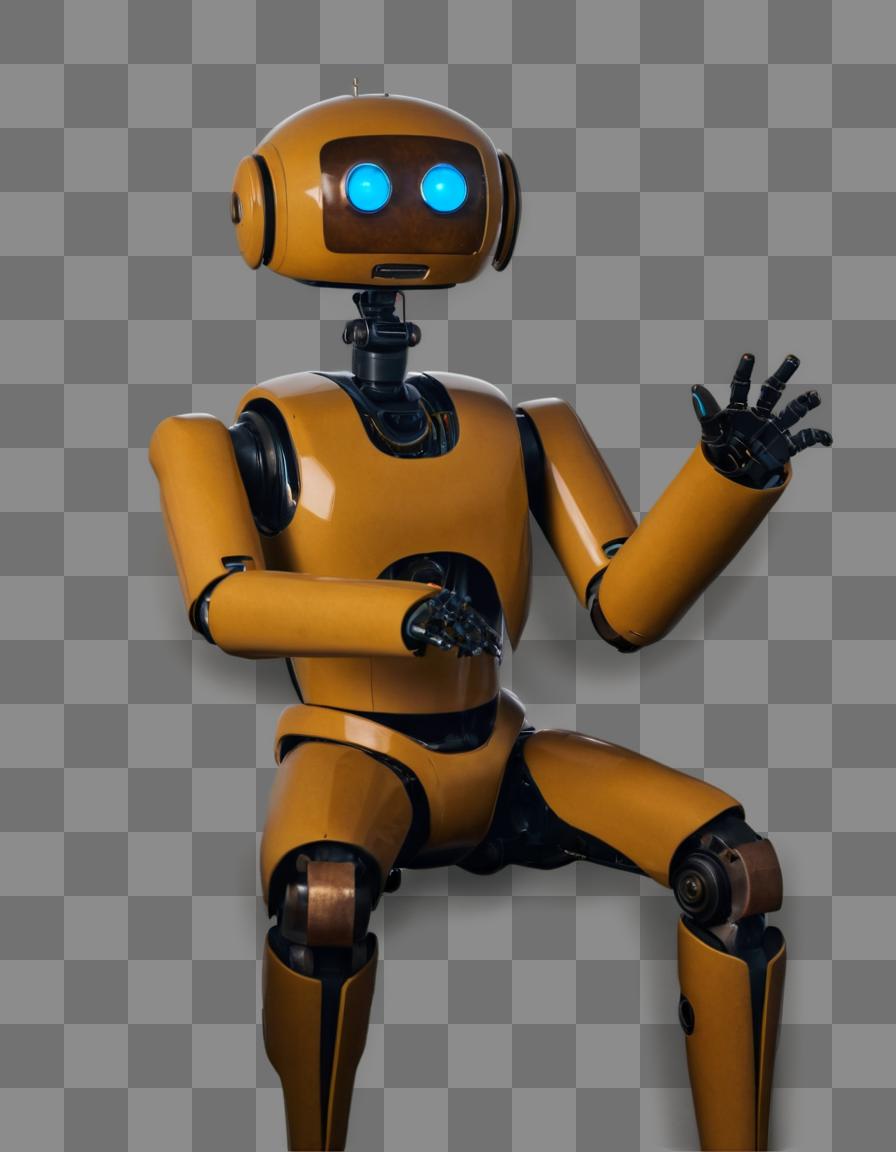}\hfill
\includegraphics[width=0.2\linewidth]{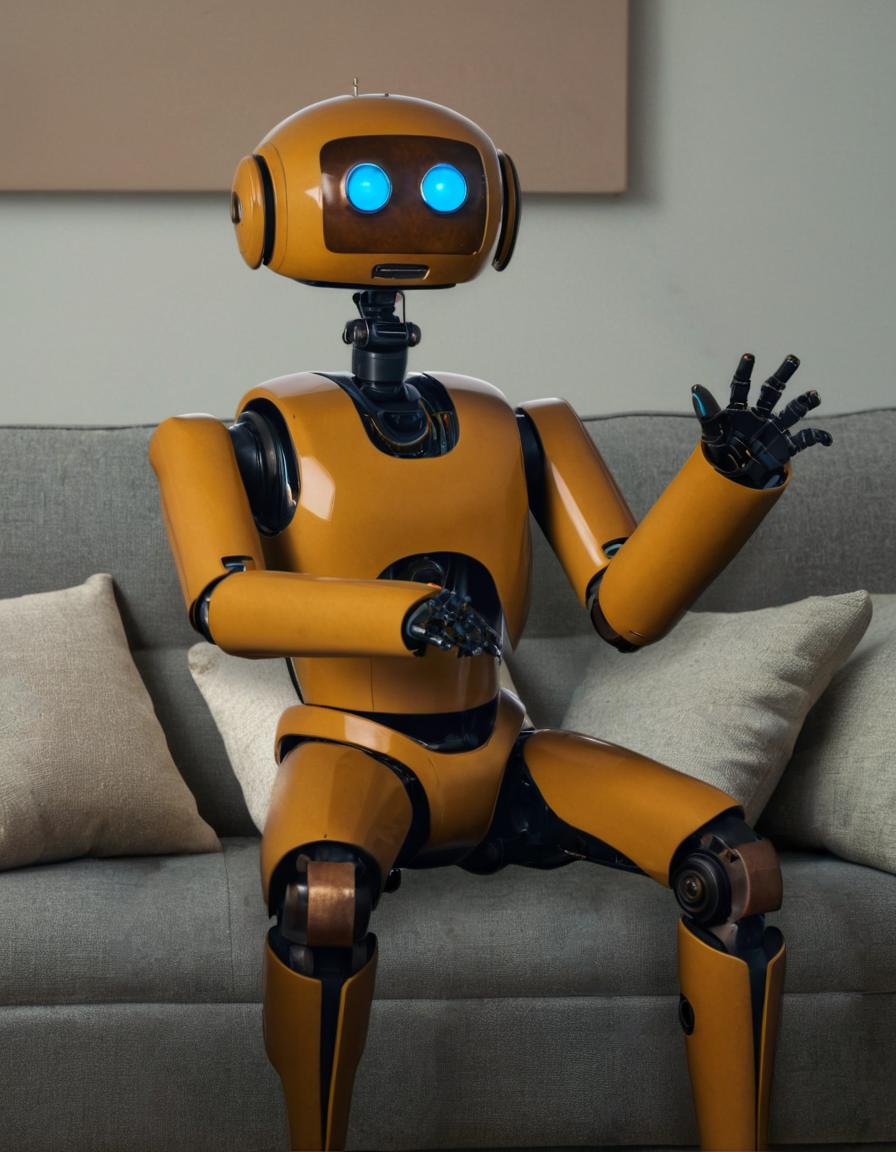}\hfill
\includegraphics[width=0.2\linewidth]{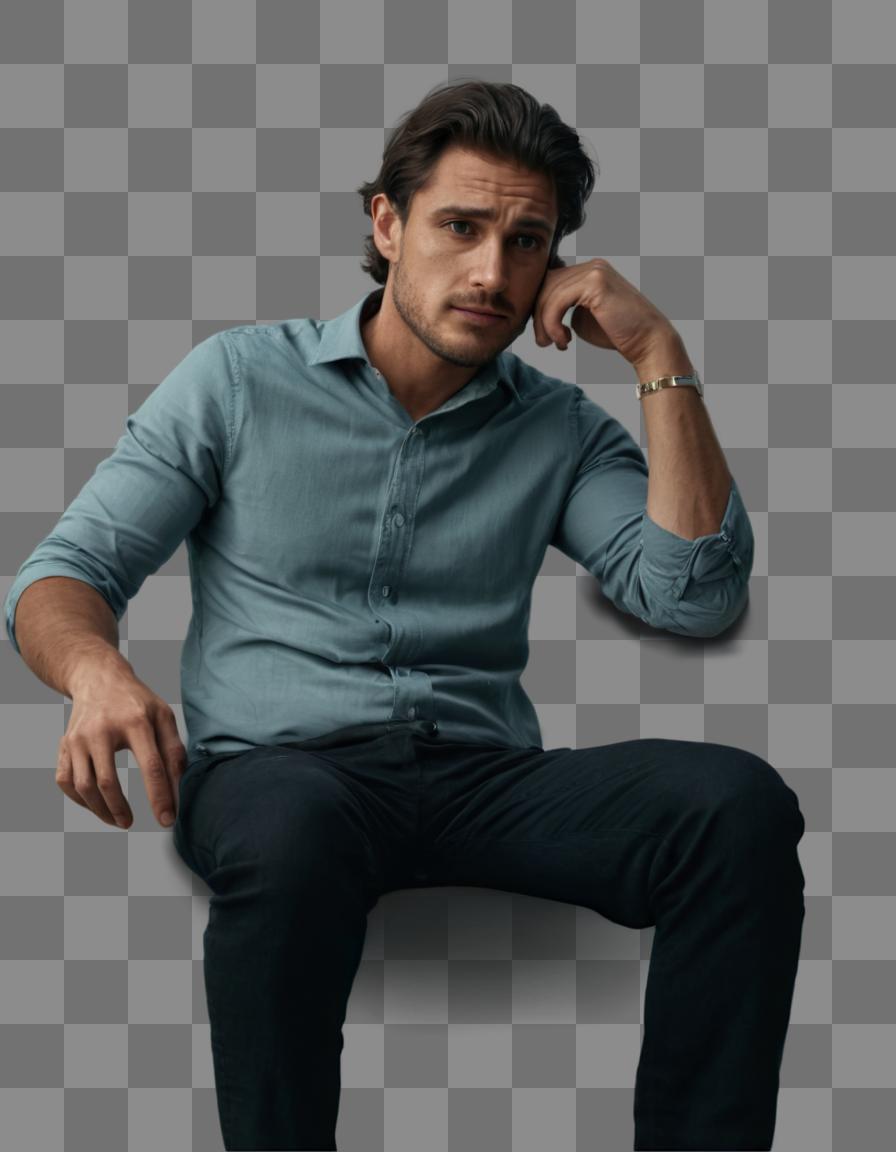}\hfill
\includegraphics[width=0.2\linewidth]{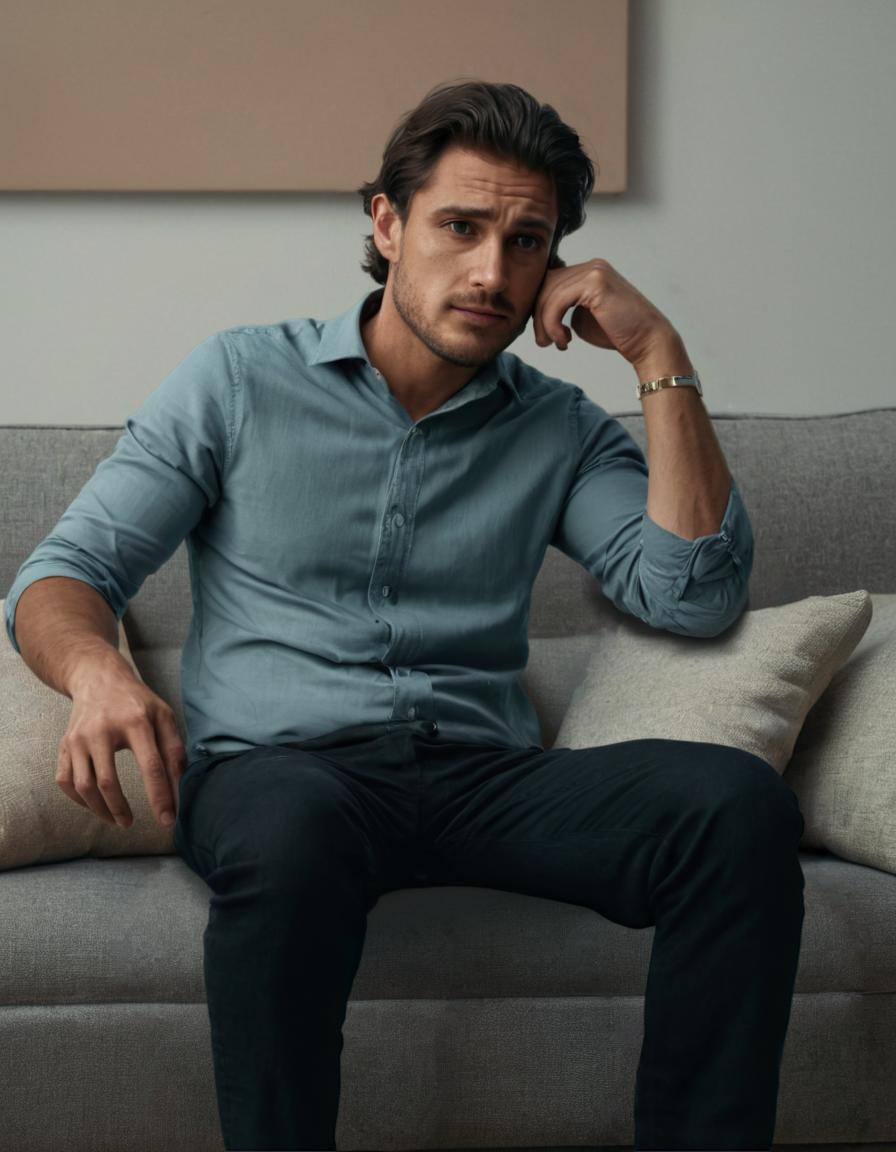}

\vspace{1pt}
\includegraphics[width=0.2\linewidth]{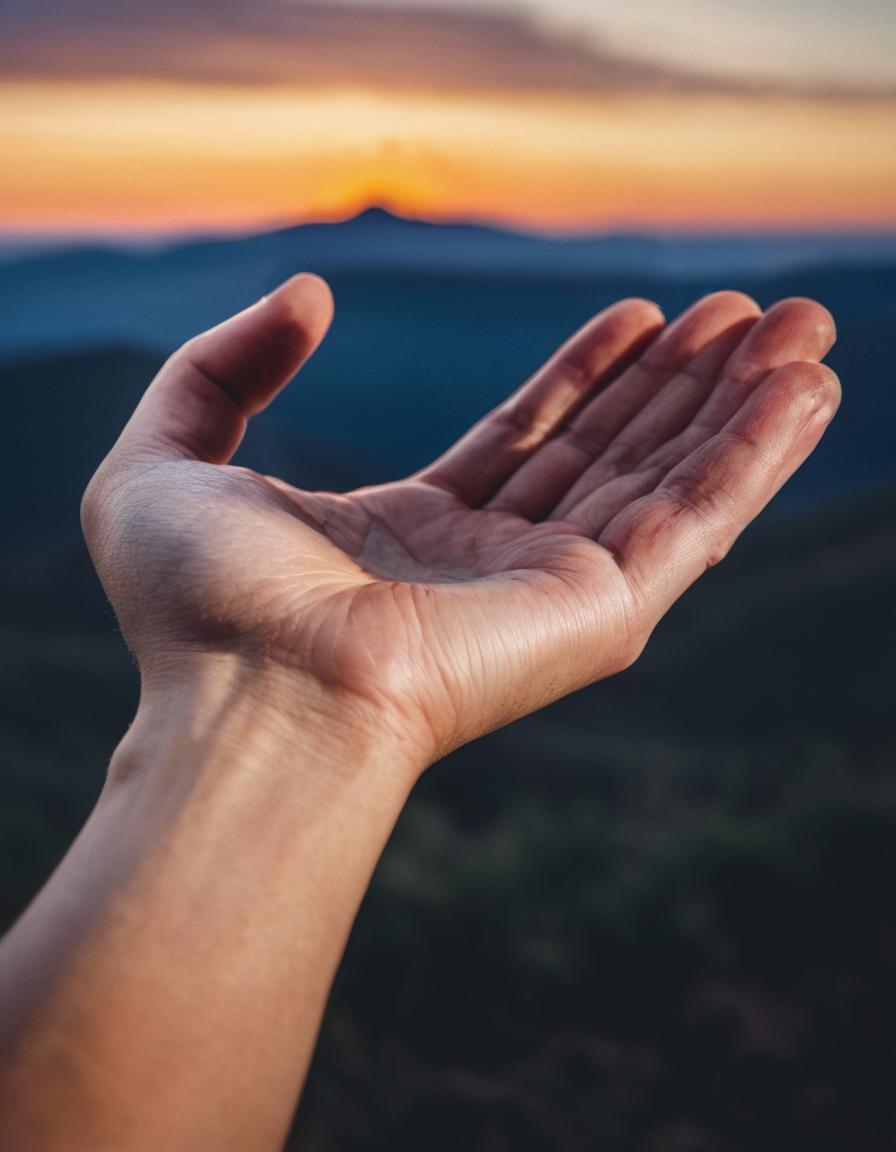}\hfill
\includegraphics[width=0.2\linewidth]{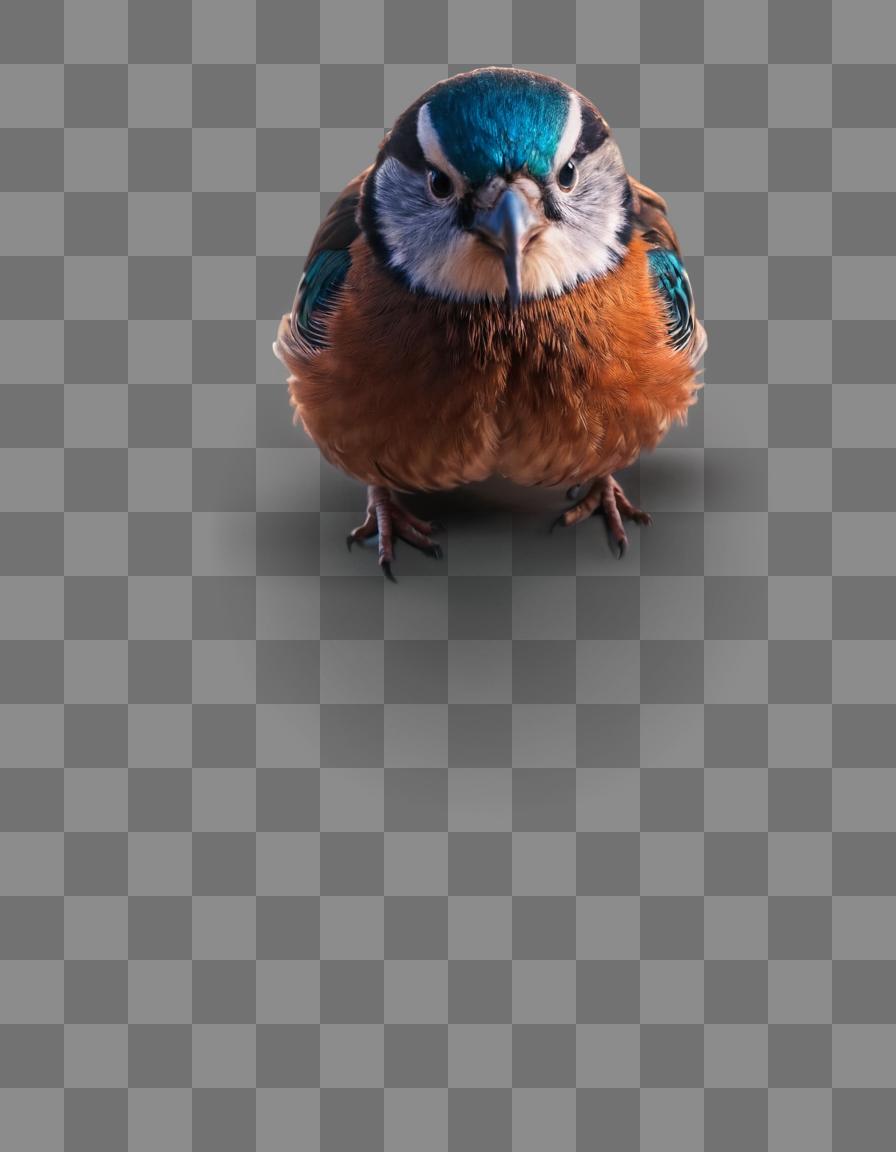}\hfill
\includegraphics[width=0.2\linewidth]{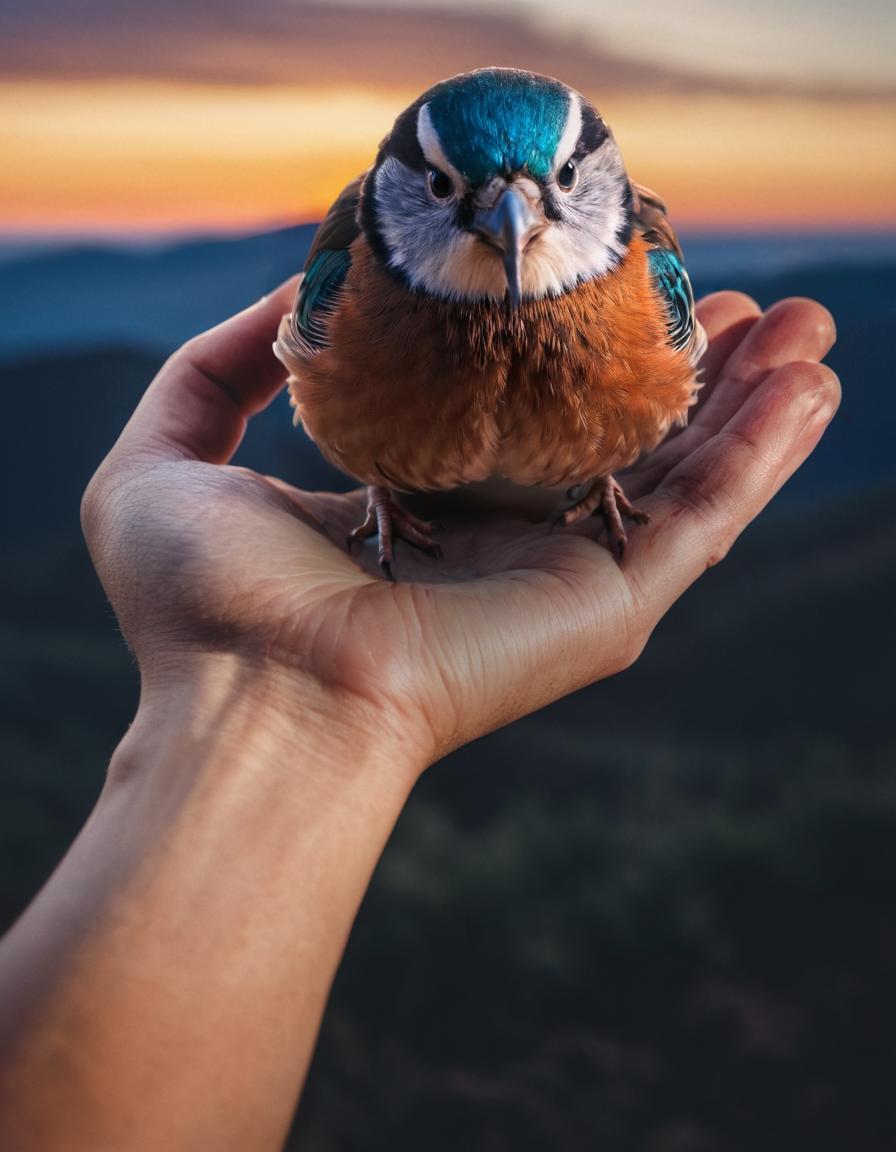}\hfill
\includegraphics[width=0.2\linewidth]{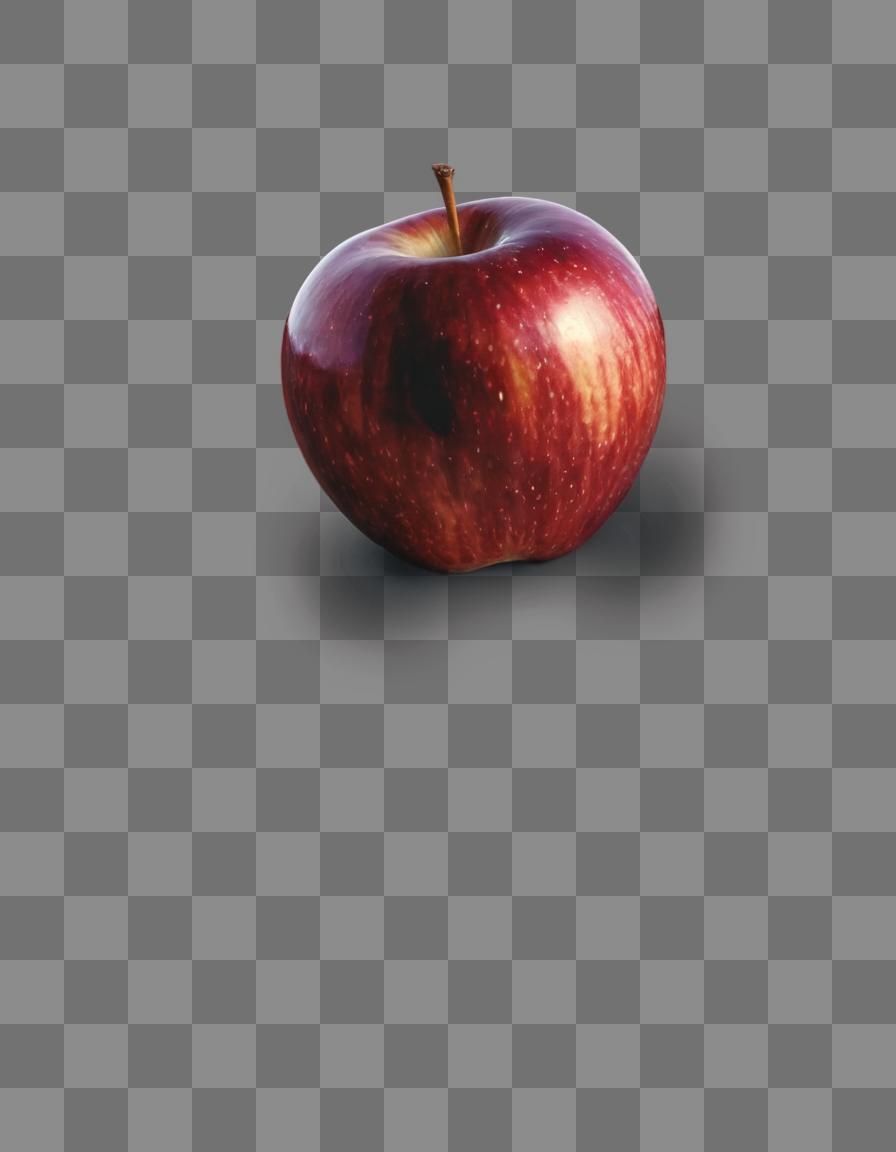}\hfill
\includegraphics[width=0.2\linewidth]{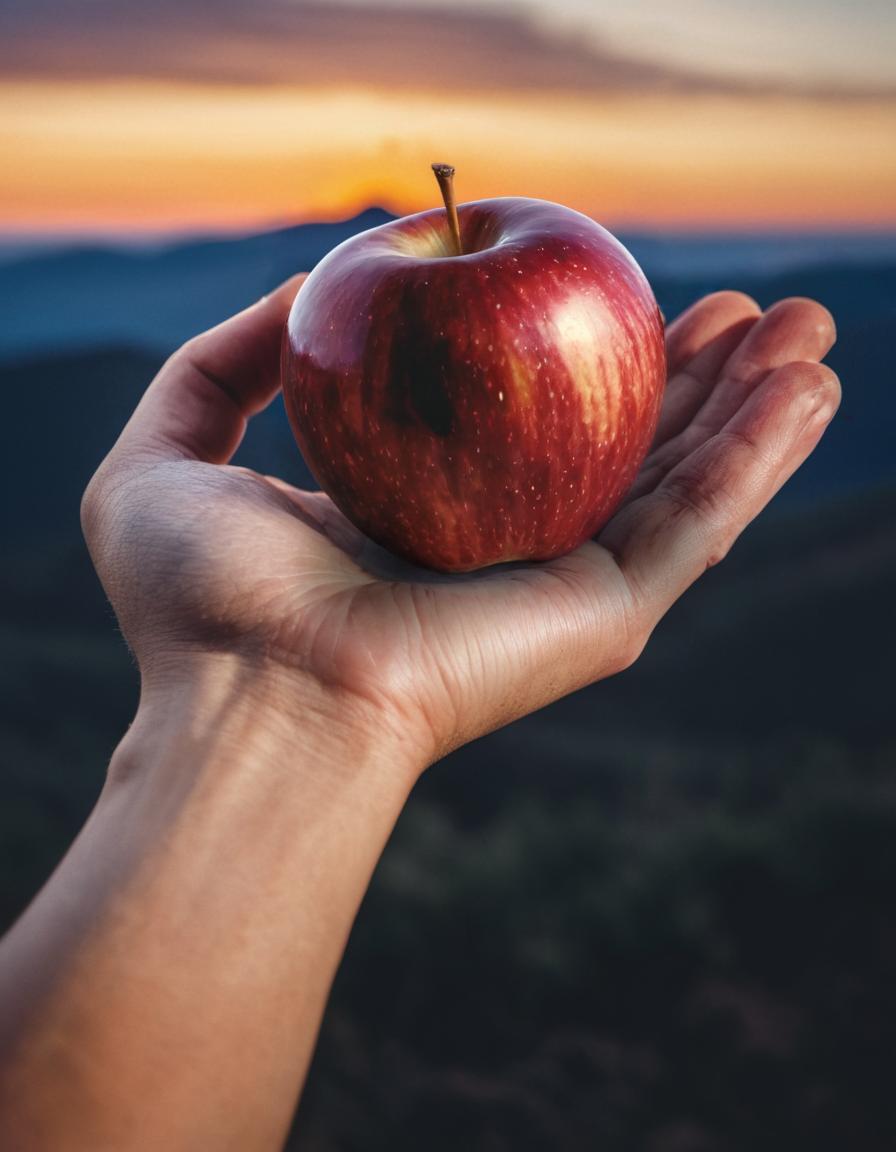}
\caption{Background-conditioned Foreground Results \#1. The left-most images are inputs. The prompts are ``woman climbing mountain'', ``man climbing mountain'', ``robot in sofa waving hand'', ``man in sofa'', ``bird on hand'', ``apple on hand''. Resolution is $896\times1152$.}
\label{fig:b1}
\end{minipage}
\end{figure*}

We present additional results for background-conditioned foregrounds in Figure~\ref{fig:b1}.

\end{document}